\documentclass[letterpaper,11pt]{report}

\usepackage{sunyitms}       
\usepackage{ulem}       
\usepackage{url}	
\usepackage{pstricks}
\usepackage{epsfig}
\usepackage{amsmath}
\usepackage{graphicx}
\usepackage{here}
\usepackage{longtable}
\usepackage{multirow}
\usepackage{caption}
\usepackage{subfig}

\title{Adapting Stochastic Search for Real-time Dynamic Weighted Constraint Satisfaction}
\author{Gregory E. Hasseler}
\date{May 8, 2010}
\airforceclearance{Distribution A: Approved for public release; distribution is unlimited. \\ (Approval AFRL PA \#88ABW-2010-4175)}

\previousdegree{B.S., State University of New York Institute of Technology, 2008}

\advisor{William J. Confer}

\professor{Christopher W. Urban\\
Lecturer\\
Department of Computer Science}

\professor{William J. Confer, Chair\\
Assistant Professor \\
Department of Computer Science}

\professor{Patricia E. Murphy\\
Assistant Professor \\
School of Arts and Sciences}

\style{Journal of Approximation Theory (together with the style known as
``sunyitms'').  Bibliography follows van Leunen's {\it A} {\it Handbook}
{\it for} {\it Scholars}.}

\software{The document preparation package \TeX{} (specifically \LaTeX2e)
together with the style-file {\tt sunyitms.sty}.}

\begin{document}

%

\begin{romanpages}      
\TitlePage 
\CopyrightPage  

%
%
\begin{vita}
Gregory Hasseler began his career in the field of computer science in 2001. He started by teaching himself C++ and Visual Basic .NET.

When he started college in the fall of 2005, he decided to continue pursuing his interests in the field of computer science. The following semester, he began working for the Computer Science Department at the State University of New York Institute of Technology as a systems administrator underneath the careful guidance of Mr. Nick Merante, Mr. Nicholas Gasparovich, and Mr. Stephen Naimoli.

In the summer of 2008, Greg began an internship at the USAF Rome Research Site underneath the careful mentorship of Mr. Anthony Ford. It is at this internship where Greg developed a strong computer science research interest. He continued to work for the USAF until the spring of 2010, when he accepted a position as a computer scientist with ATC--NY.
\end{vita}

%
%
\begin{abstract}
This work presents two new algorithms for performing constraint satisfaction. The first algorithm presented, \emph{DMaxWalkSat}, is a constraint solver specialized for solving dynamic, weighted constraint satisfaction problems. The second algorithm, \emph{RDMaxWalkSat}, is a derivative of DMaxWalkSat that has been modified into an anytime algorithm, and hence support real--time constraint satisfaction. DMaxWalkSat is shown to offer performance advantages in terms of solution quality and run--time over its parent constraint solver, MaxWalkSat. RDMaxWalkSat is shown to support anytime operation. The introduction of these algorithms brings another tool to the areas of computer science that naturally represent problems as constraint satisfaction problems, an example of which is the robust coherence algorithm.
\end{abstract}

%
%
\begin{acknowledgments}
This work is dedicated to my grandparents: Edward and Evelyn Peller, and Donald and Ruth Hasseler.

Without the help of my adviser, Dr. William Confer, and the rest of my committee, this work would not have happened.

Mr. Anthony Ford, Mr. Jerry Dussault, and Mr. Alex Sisti of USAF AFRL/RIS have also been instrumental in this work. Anthony, Jerry and Al are responsible for getting me interested in the field of constraint satisfaction and have provided me immeasurable help not just on this thesis, but throughout my professional career thus far.

In addition, Mrs. Dawn Nelson and Ms. Jennifer Lampe have contributed much support to me and have spent far too many hours proof reading.

Finally, without my parents and their support, I don't know if any of this work would have been possible. I am greatly indebted to them.
\end{acknowledgments}

\StylePage
\tableofcontents

%
\listoffigures
\listoftables
\listofalgorithms

\end{romanpages}        

\normalem       

%

%

%

%

%

%

%

%

\chapter{Introduction}

Constraint satisfaction problems have applicability for describing a wide variety of problems. The full value of problem description as a constraint satisfaction problem cannot be achieved without the availability of a CSP solver. Many CSP solvers exist, however, very few support dynamic constraint satisfaction problems, and even fewer are capable of operating effectively in a real--time environment.

This work presents two CSP solvers. The first, \emph{DMaxWalkSat}, adapts an existing stochastic search based CSP solver for dynamic constraint satisfaction problems. The second, \emph{RDMaxWalkSat}, adapts DMaxWalkSat for effective operation in real--time environments. Both algorithms are likely to find applicability in fields requiring dynamic constraint satisfaction or real--time constraint satisfaction. Artificial Intelligence and Assisted Planning are two examples of fields likely to benefit.

This work performs two studies, each focused on a particular solver. The first study in this work examines DMaxWalkSat, while the second study examines RDMaxWalkSat. The DMaxWalkSat study generated a number of dynamic constraint satisfaction problems and compared DMaxWalkSat's effectiveness solving such problems against the non--dynamic constraint satisfaction solver, MaxWalkSat. The comparison found that for 96.7\% of the problems tested, DMaxWalkSat solved constraint additions more effectively than MaxWalkSat, and for 95.6\% of the problems, it performed constraint removals more effectively. The RDMaxWalkSat study utilized RDMaxWalkSat's anytime algorithm capability to test the quality of the generated solution at arbitrary points in time. The results indicate a non--linear relationship between solution score and time the solver was allowed to run. The results also show that for 66\% of the problems tested, substantial improvements to the solution can be made 25\% of the time it takes a comparable stochastic search based CSP solver to solve the same problem.
\chapter{Related Work}

\section{Background} 
\begin{figure}
	\centering
	\includegraphics[width=3in]{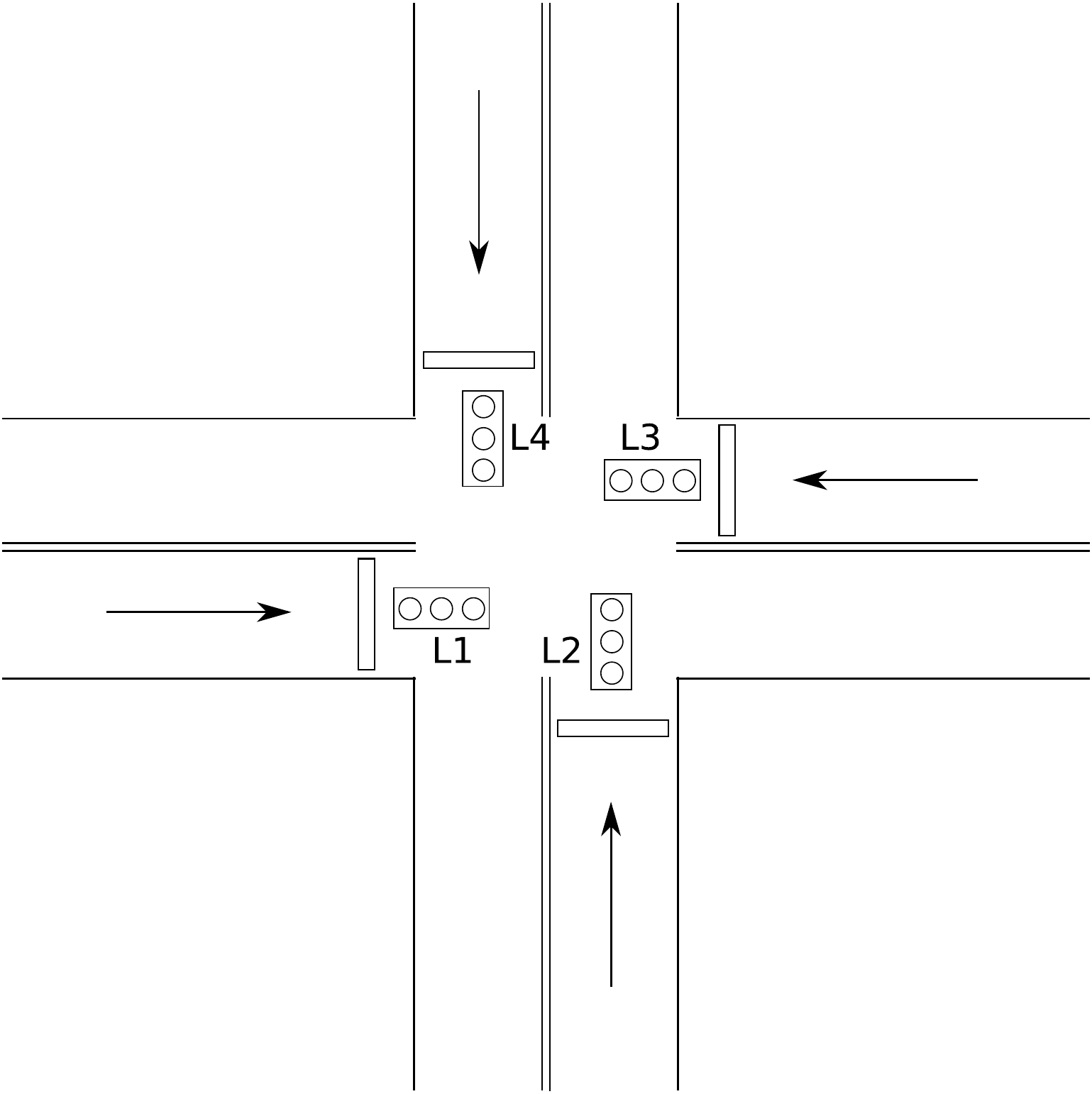}
	\caption{An example four-way intersection with traffic lights.} 
	\label{fig_trafficlights}
\end{figure}

A constraint satisfaction problem (CSP) deals with assigning values to variables in accordance with a set of restrictions, or constraints. The simple act of driving to work leads most people to encounter an instance of an extremely simple CSP: a traffic light. In the case of traffic lights for a four-way intersection, only one road going through the intersection may have a green or yellow light at one time. Consider Figure \ref{fig_trafficlights}. If light $L1$ or $L3$ and lights $L4$ or $L2$ simultaneously present green or yellow lights, an accident is likely to occur. If light $L2$ or $L4$ and lights $L1$ or $L3$ simultaneously present green or yellow lights, an accident is also likely to occur. Depending on the intended operation of the traffic lights, for example, with one or more lanes of traffic having an advanced turn, other constraints not shown here may also be applicable.

In Figure \ref{fig_trafficlights}, there are four traffic lights (lights L1-L4) to assign colors to. Each light may be represented by a \emph{variable}. The collection of variables representing the lights is set $V$. Each traffic light can show one color at any time. The only possible colors to show are red, yellow, or green. The collection of these colors is referred to as the \emph{domain} for each variable in set $V$ because the set contains the only possible values that may be assigned to any of the variables. Each variable has its own domain, even if the domain for one variable is the same as the domain for another variable. The set of domains in the traffic light example is set $D$. The colors may not be assigned to the traffic lights arbitrarily. Instead, the assignment made to each traffic light may not violate any of the \emph{constraints}. The set of constraints is called set $C$. The traffic light CSP example may be formally stated as $CSP = (V,D,C)$, where
	\begin{align*}
		V =& \{L1,L2,L3,L4\} \\
		D =&
			\{\{red,yellow,green\},
			\{red,yellow,green\},
			\{red,yellow,green\},
			\{red,yellow,green\}\} \\
		C =& 
			\{
			L1_{green} \mbox{ NAND } L2_{green},
			L1_{green} \mbox{ NAND } L2_{yellow},
			L1_{green} \mbox{ NAND } L4_{green}, \\
		  & 
		  	L1_{green} \mbox{ NAND } L4_{yellow},
		  	L1_{yellow} \mbox{ NAND } L2_{green},
		  	L1_{yellow} \mbox{ NAND } L2_{yellow}, \\
		  &
		  	L1_{yellow} \mbox{ NAND } L4_{green},
		  	L1_{yellow} \mbox{ NAND } L4_{yellow},
		  	L2_{green} \mbox{ NAND } L3_{green}, \\
		  &
		  	L2_{green} \mbox{ NAND } L3_{yellow},
		  	L2_{yellow} \mbox{ NAND } L3_{green},
		  	L2_{yellow} \mbox{ NAND } L3_{yellow}, \\
		  &
		  	L3_{green} \mbox{ NAND } L4_{green},
		  	L3_{green} \mbox{ NAND } L4_{yellow},
		  	L3_{yellow} \mbox{ NAND } L4_{green}, \\
		  &
		  	L3_{yellow} \mbox{ NAND } L4_{yellow}
		  	\}
	\end{align*}

CSPs may vary in the number of constrained variables, the number of values each respective variable may take, and the number of constraints. This work focuses only on binary boolean CSPs. A binary CSP is a CSP where no constraint is on more than two variables. A boolean CSP is a CSP where each variable's domain is $\{T,F\}$.

\section{SAT vs. MAX--SAT vs. Weighted MAX--SAT}
Not all CSPs have a possible solution assignment that satisfies all the constraints. This class of problems is known as unsatisfiable (UNSAT). The class of CSPs with a solution assignment that satisfies all the constraints is known as satisfiable (SAT) \cite{Dechter_2003_Constraint}. A traditional or classical CSP solver seeks to determine whether or not a CSP is SAT or UNSAT. If it is SAT, the solver seeks to find a solution assignment. However, if the CSP is UNSAT, then no matter the amount of computation performed by the solver, no solution assignment satisfying all the constraints will ever be found! In such cases, a different type of solver, known as a Maximum Satisfiability (MAX--SAT) solver, may be used. Instead of immediately terminating and declaring the problem as UNSAT, a MAX--SAT solver seeks to find the solution assignment that satisfies the greatest number of constraints possible.

In the case of a MAX--SAT CSP, some constraints may be more important to satisfy than others. This importance is encoded by associating a weight with each constraint. Commonly, each associated weight is an integer value greater than 0 and is considered to be the cost of not satisfying the corresponding constraint. A CSP composed of weighted constraints is known as a Weighted MAX--SAT CSP (WCSP). A WCSP solver tries to satisfy the combination of constraints that will minimize the cost of the solution. Another description of WCSP is that the associated weights are positive scores and the objective of WCSP solvers is to satisfy the combination of constraints that will result in the highest possible cumulative score.

\section{Dynamic CSP}
Traditional CSPs are sometimes not sufficient for accurately modeling and solving problems. A typical class of examples is from the field of Artificial Intelligence (AI). Despite many AI problems being modeled as static CSPs, some of these problems are more accurately modeled as dynamic constraint satisfaction problems (DCSPs). CSPs that model a problem that may be changed by its environment may change with maturation of user requirements, or may change as the result of actions of other agents in the same network \cite{Verfaillie_1994_Solution, Dechter_1988_Belief} are usually best represented as a DCSP. DCSPs are commonly represented as a sequence of CSPs \cite{Dechter_1988_Belief}, as is shown in Figure \ref{fig_dcsp_as_csp}.

\section{Constraint Expression}
The constraints for the problem shown in Figure \ref{fig_trafficlights} were expressed as \\
	$L1_{green}$ $\mbox{NAND}$ $L2_{green}$,
	$L1_{green}$ $\mbox{NAND}$ $L2_{yellow}$,
	$L1_{green}$ $\mbox{NAND}$ $L4_{green}$,
	$L1_{green}$ $\mbox{NAND}$ $L4_{yellow}$,
  	$L1_{yellow}$ $\mbox{NAND}$ $L2_{green}$,
  	$L1_{yellow}$ $\mbox{NAND}$ $L2_{yellow}$,
  	$L1_{yellow}$ $\mbox{NAND}$ $L4_{green}$,
  	$L1_{yellow}$ $\mbox{NAND}$ $L4_{yellow}$,
  	$L2_{green}$ $\mbox{NAND}$ $L3_{green}$,
  	$L2_{green}$ $\mbox{NAND}$ $L3_{yellow}$,
  	$L2_{yellow}$ $\mbox{NAND}$ $L3_{green}$,
  	$L2_{yellow}$ $\mbox{NAND}$ $L3_{yellow}$,
  	$L3_{green}$ $\mbox{NAND}$ $L4_{green}$,
  	$L3_{green}$ $\mbox{NAND}$ $L4_{yellow}$,
  	$L3_{yellow}$ $\mbox{NAND}$ $L4_{green}$,
  	$L3_{yellow}$ $\mbox{NAND}$ $L4_{yellow}$.
  	
This expression format is not the format used by a majority of solvers. Most solvers require that their constraints be expressed in conjunctive normal form (CNF). Through the use of De Morgan's law, the NAND operation used in the expression of the constraints for Figure \ref{fig_trafficlights} may be expressed in an equivalent CNF. The equivalent CNF is, $F$, where
\begin{align*}
F	=& 	\neg L1_{green} \vee \neg L2_{green},
		\neg L1_{green} \vee \neg L2_{yellow},
		\neg L1_{green} \vee \neg L4_{green},
		\neg L1_{green} \vee \neg L4_{yellow}, \\
	&	\neg L1_{yellow} \vee \neg L2_{green},
		\neg L1_{yellow} \vee \neg L2_{yellow},
		\neg L1_{yellow} \vee \neg L4_{green},
		\neg L1_{yellow} \vee \neg L4_{yellow}, \\
	&	\neg L2_{green} \vee \neg L3_{green},
		\neg L2_{green} \vee \neg L3_{yellow},
		\neg L2_{yellow} \vee \neg L3_{green},
		\neg L2_{yellow} \vee \neg L3_{yellow}, \\
	&	\neg L3_{green} \vee \neg L4_{green},
		\neg L3_{green} \vee \neg L4_{yellow},
		\neg L3_{yellow} \vee \neg L4_{green},
		\neg L3_{yellow} \vee \neg L4_{yellow}
\end{align*} 
The  $\wedge$ symbol indicating conjunction is omitted from between each conjunct because it is implicit in CNF. In the case of $F$, each clause is a conjunct.

\begin{figure}
	\centering
	\includegraphics{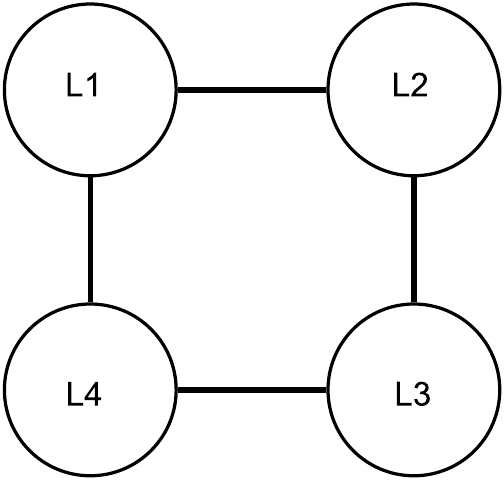}
	\caption{The constraint graph representing the constraints present between the traffic lights from the example given in Figure \ref{fig_trafficlights}.} 
	\label{fig_trafficlights_cg}
\end{figure}

Another popular format for expressing constraints, although not commonly used by solvers, are constraint graphs. Constraint graphs are typically used as visual aids for understanding the structure of a CSP. The constraint graphs shown in this work are primal due to this work's focus on binary CSPs. However, if there is a need to develop a constraint graph for non-binary constraints (also known as $k$-ary constraints), hypergraphs \cite{Rossi_2006_Handbook} could be used. In the case of primal constraint graphs, each node represents a different variable and an edge between nodes represents a constraint among the connected nodes. In the case of hypergraph constraint graphs, each node represents a different variable and a hyperedge connecting a set of nodes represents a constraint among all the connected nodes. Figure \ref{fig_trafficlights_cg} shows a constraint graph for the traffic light example.

\section{Solver Proliferation}
The performance and accuracy of solvers have not always scaled well with CSPs as the number of variables and constraints change. For example, brute force solvers may be appropriate for some small CSPs. However, the amount of computation required for brute force solvers grows exponentially as the CSP becomes more complex, severely limiting the practicality of them. For brute force solvers, the number of candidate solutions that must be tested is the cardinality of the Cartesian product of all the variable domains \cite{Kumar_1992_Algorithms}. The desire to solve as many CSP instances as possible in the shortest amount of time has led to the development of solvers that use more advanced kinds of search and heuristics than simple brute force solvers.

\section{Complete vs. Incomplete Solvers}
An extremely important characteristic of a CSP solver is whether it is complete or incomplete. According to Tsang, a complete solver tries to find all solutions for a particular CSP, whereas an incomplete solver tries to find a solution for a particular CSP \cite{Tsang_1993_Foundations}. Some applications of a CSP solver may require that all solutions be found. In those applications, a complete solver should be used. In other applications, an incomplete solver may be sufficient.

\section{Solution Techniques for SAT}
A variety of solution techniques for SAT solving exist. This work will survey the most common basic techniques.

\subsection{Constraint propagation}
Constraint propagation, sometimes referred to as arc consistency, is a fairly straight forward technique for solving CSPs. Constraint propagation builds off of constraint graphs. Basic constraint propagation starts by examining the domain for a given variable $v$. It then proceeds by examining the domain of another variable, $w$, that $v$ is constrained with. The next step that occurs is the elimination of any values from $w$'s domain that would violate the constraint(s) between $v$ and $w$. The process then continues for each variable until only consistent values in the domain of each variable remain. It is important to note that if a domain for a variable is reduced, it is necessary to reprocess all variables in order to ensure that a previously satisfied constraint has not become violated. Additionally, it is important to note that constraint propagation is directional \cite{Kumar_1992_Algorithms}.

The result of performing constraint propagation on a CSP yields a set of possible assignments. However, it does not necessarily yield all possible assignments. Constraint propagation is an incomplete constraint solving technique. Although constraint propagation may find a CSP to be arc inconsistent if it eliminates all values from a variables domain, it does not necessarily mean that the CSP is UNSAT. Due to its incomplete nature, a CSP that is found to be arc inconsistent may later be found to be arc consistent when traversed in a different order.

\subsection{Stochastic search}
Another popular approach to searching that is used in many different CSP solvers is that of stochastic search \cite{Smyth_2003_Iterated, Jiang_1995_Solving}. Stochastic search algorithms explore the search space of possible solutions through the application of various heuristics. Because not all heuristics will cause the search space of possible solutions to be exhaustively examined, stochastic search based solvers are generally incomplete \cite{Tsang_1993_Foundations}. Stochastic search based algorithms generally yield shorter solution times than complete search techniques \cite{Tsang_1993_Foundations, Frank_1997_Learning, Selman_1992_New}. A benefit of the generally shorter solution times is that stochastic search based solvers are sometimes able to solve problems that complete search based solvers are otherwise unable to due to time limitations.

Although there are many different classes of approach to stochastic search solvers, a very popular approach is that of hill--climbing \cite{Tsang_1993_Foundations}.

\subsubsection{Hill--climbing solvers}
Naive hill--climbing solvers are generally very simple. The basis of hill--climbing solvers is that the solver can non--deterministically move throughout the search space from possible solution assignment to possible solution assignment. The next possible solution assignment the solver moves to in the search space is the best possible solution assignment it can see from its current position. The situation eventually occurs where the solver sees no better possible solution assignment than its current location. When this occurs, it means that the solver has moved into a local--maxima or onto a plateau. A local--maxima occurs when the solver believes it has found the best solution assignment. A plateau occurs if all other possible solution assignments the solver can move to are equal in quality. Unfortunately, because of the non--deterministic traversal of the search space, a local--maxima or plateau may not also be a global--maxima. When a solver has found a local--maxima or a plateau that is not also a global--maxima, it is said to be ``trapped'' \cite{Tsang_1993_Foundations}.

The following questions remain unanswered:
	\begin{itemize}
		\item Where does the solver start in the search space?
		\item How does the solver determine what the next possible moves are?
		\item How does the solver measure the quality/fitness of the possible next moves?
		\item How does the solver determine if it has become trapped in local--maxima or plateau that is not also a global--maxima?
		\item If the solver can detect that it has become trapped, how does it free itself?
	\end{itemize}
The answers to these questions differentiate various hill--climbing solvers from one another.

\subsection{Backtracking search}
Unlike constraint propagation and stochastic search, backtracking search is a complete search technique. Unfortunately, due to its complete nature, backtracking search may take an extremely long time to find a solution assignment. In a worst case scenario, a backtracking search CSP solver will try every possible value for every variable before concluding the CSP it is operating on is UNSAT.

A simple form of backtracking search is chronological backtracking. A chronological backtracking search solver tries to extend a partial assignment to another variable without violating any constraints. In the case that a solver cannot extend a partial assignment to another variable, the solver will ``backtrack'' and undo the previous assignment \cite{Tsang_1993_Foundations}. By undoing a previous assignment, the solver hopes to be able to extend the assignment in a different way that will not also lead to a dead end. If the solver has undone all partial assignments and is still unable to find a new partial assignment without violating any constraints, then the CSP it is trying to solve is considered UNSAT.

Like stochastic search CSP solvers, backtracking search CSP solvers have to make a variety of decisions that can drastically effect their performance. These decisions include:
\begin{itemize}
	\item How does the solver decide which variable to start the partial assignment with?
	\item How does the solver decide which variable to try to extend the partial assignment to next?\footnote{The order in which the variables are explored can have profound effects on the performance of the algorithm because it directly effects the amount of search space that must be examined \cite{Dechter_2003_Constraint}.}
	\item How can the solver reduce the number of times it must backtrack?
	\item How can the solver reduce the number of times it has to search through previously explored areas of the search space?
\end{itemize}
Again, the way different solvers make these decisions differentiate them.

\subsection{Hybrid search}
Hybrid search CSP solvers combine elements of both stochastic search and backtracking search solvers. Prestwich states that this is sometimes done because ``neither backtracking nor [stochastic] search is seen as adequate for all problems'' \cite{Prestwich_2001_Local}. A common hybrid approach is using a stochastic solver to generate an initial solution that a backtracking solver will then try to refine into a better solution.

\section{Solution Techniques for WCSP}
Many WCSP solvers are adaptations of CSP solvers. WCSP solvers may utilize all the same techniques as CSP solvers such as constraint propagation, stochastic search, backtracking search, and hybrid search. This work describes an adaptation of stochastic search for WCSP and an adaptation of backtracking search for WCSP.

\subsection{Adaptation of Stochastic Search for WCSP}
The adaptation of stochastic search for WCSP is fairly straightforward. For the stochastic search based CSP solver WalkSat, the required modification is changing the fitness function that is used to decide the next move \cite{Jiang_1995_Solving}. WalkSat's stock fitness function measures how many constraints each next possible move will violate. The function then compares the results for each next possible move to the current number of violated constraints for the current position. It then returns the next possible move that minimizes the number of violated constraints. The WCSP version of WalkSat, called MaxWalkSat, changes the fitness function to one that computes a score for each next possible move by summing the associated weights of the constraints that would be violated by the move. Again, the function compares these scores to the score of the current position and returns the next possible position that minimizes the score.

\subsection{Adaptation of Backtracking Search for WCSP}
According to Borchers and Furman \cite{Borchers_1999_Two, Givry_2003_Solving}, the adaptation of backtracking search for WCSP solving requires slightly more adaptation than stochastic search. A backtracking based WCSP solver begins by executing a stochastic search based WCSP solver on the same WCSP instance. The cost of the solution found by the stochastic search based WCSP solver is used to establish an initial best cost, or upper bound. The solver then begins the same as a backtracking search based CSP solver by attempting to extend partial assignments into full assignment. The difference, however, is that after extending a partial assignment, the solver sums the cost of all constraints violated by the assignment and assigns this sum as the lower bound. The advantage of establishing a lower bound is that when a partial assignment has an equal or higher cost than a previous partial assignment, the solver knows that the current partial assignment cannot be extended in any way that would result in a lower cost. The detection of this condition results in the solver pruning the remaining branches in the search space that originate from the current partial assignment. The resulting behavior of the solver is identical to that of a backtracking based CSP solver after encountering a dead--end assignment. However, if the lower bound is less than the upper bound, the solver will continue its search along the current branch in the solution space. When the solver has extended the partial assignment to a full solution assignment, it compares the cost of the resulting assignment to the upper bound. If the cost of the solution is less than the upper bound, than the solver has been able to find a better solution than the initial one found by the stochastic search based solver. Otherwise, the costs of the two solutions are equal.

\section{Solution Techniques for DCSP}
\begin{figure}
	\centering
	\includegraphics{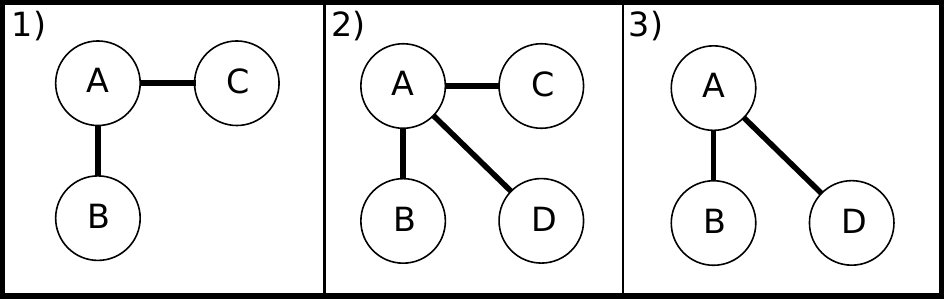}
	\caption{DCSP shown as a sequence of CSPs, with each CSP having to be solved from scratch. Box 1 shows the initial CSP. Box 2 shows the next CSP in the sequence after adding a constraint constraining variables A and B. Box 3 shows the following CSP in the sequence after removing the constraint constraining variables A and C.} 
	\label{fig_dcsp_as_csp}
\end{figure}

\begin{figure}
	\centering
	\includegraphics{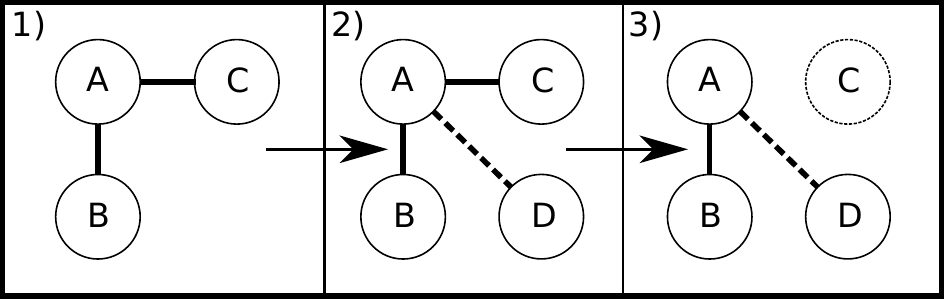}
	\caption{DCSP shown as a single CSP instance. Box 1 shows the initial CSP. Box 2 shows the CSP instance after adding a constraint constraining variables A and D. Box 3 shows the CSP instance after removing the constraint constraining variables A and C.} 
	\label{fig_dcsp_as_dcsp}
\end{figure}

DCSPs may be solved either as a sequence of CSPs, with each CSP in the sequence being the result of adding or removing constraints to the previous CSP, or using a specialized DCSP solver. Figure \ref{fig_dcsp_as_csp} shows an example sequence of CSPs for a DCSP. In contrast, Figure \ref{fig_dcsp_as_dcsp} shows the changes to a single CSP instance over time that occur when using a DCSP solver.

Solving a DCSP as a sequence of static CSPs may require more time than necessary to reach a solution and yield unstable solutions over time \cite{Verfaillie_1994_Solution}. Solution efficiency and stability are especially important for DCSP applications such as mission scheduling, because a new solution resulting from a constraint modification may disregard work that has already begun based upon a previous solution \cite{Bresina_2005_Activity}.

\subsection{Stochastic Search for DCSP}
A number of techniques exist for adapting stochastic search in general for DCSP \cite{Miguel_2000_Dynamic}.

\subsubsection{Local and Heuristic Repair}
Local repair is a technique for adapting stochastic search for DCSP elaborated on by Minton et al.\ in \cite{Minton_1992_Minimizing}. Local repair starts with an old, suboptimal solution assignment. In this way, local repair is a form of heuristic repair \cite{Verfaillie_1994_Solution}. Heuristic repair is a repair process that is guided by a particular heuristic. Stochastic search is then resumed from the suboptimal solution. As the stochastic search proceeds, the changes it makes to the assignment are considered to be local repairs.

\subsubsection{Local Change}
As described by Miguel and Shen in \cite{Miguel_2000_Dynamic}, local change is a DCSP technique very similar to local repair. The difference between local change and local repair is that local change tries to be more efficient by focusing only on the variables that are involved in a violated constraint.

\subsubsection{Constraint Recording}
Constraint recording is described by Miguel and Shen in \cite{Miguel_2000_Dynamic}. Constraint recording is similar to clause learning in backtracking search. However, where as clause learning in backtracking search focuses on preventing the same dead end partial assignments from being repeated in search, constraint recording infers new constraints from the current problem that were not necessarily expressed in the previous problem.

\subsection{Backtracking Search for DCSP}
The adoption of backtracking search for DCSP is more complex than the adoption of stochastic search for DCSPs. Verfaillie and Schiex propose a technique utilizing dynamic backtracking in \cite{Verfaillie_1994_Dynamic} for DCSP solving. Verfaillie and Schiex's algorithm works by making use of the eliminating explanations produced by dynamic backtracking \cite{Ginsberg_1993_Dynamic}. Their algorithm, Dynamic Dynamic Backtracking (ddbt), is based on the belief that most of the eliminating explanations produced from one CSP in the DCSP sequence will hold for the next CSP in the sequence. By preserving the eliminating explanations from the previous CSP, a significant amount of search space can be pruned for the current CSP in the sequence. ddbt operates on Verfaillie's assumption ``that the non--destructive backtracking mechanism of Dynamic Backtracking'' \cite{Verfaillie_1994_Dynamic} will help preserve solution stability.

\section{Significant Developments in CSP Solving}
Significant developments have been made in the field of constraint processing. Due to the vast number of developments, only a subset of the developments we feel to be the most influential are briefly highlighted in this work.

\subsection{Constraint Propagation Developments}
Significant developments that have been made in constraint propagation based CSP solving are now briefly highlighted.

\subsubsection{AC--3}
A.K. Mackworth created a new algorithm for constraint propagation called AC--3. AC--3 is an improvement over AC--1 because it only reexamines the variables that may have been affected by a domain reduction on another variable. AC--1 reexamines all variables whether or not they may have been effected by a domain reduction on another variable \cite{Kumar_1992_Algorithms}. AC--3 is a very popular choice for use in CSP solvers because of its simplicity and versatility \cite{Surynek_2004_New}.

\subsection{Stochastic Search Developments}
Significant developments in stochastic search based CSP solving are now briefly highlighted.

\subsubsection{Greedy Local Search}
In ~\cite{Selman_1992_New}, Selman et al.\ introduce a new stochastic search based solver known as GSAT. GSAT is unique in that it spends time considering alternative solutions to the current solution when deciding what move to make next. In addition, GSAT introduces the use of restarts to stochastic search solvers controlled through the use of two adjustable parameters, commonly named \emph{MAX--FLIPS} and \emph{MAX--TRIES}. GSAT uses restarts and sideways moves to avoid search getting stuck in plateaus and local maxima. Due to its stochastic nature, it is possible for GSAT to make errors. However, through their experimentation, Selman et al.\ determined that GSAT seems to make statistically minimal errors. However, they are careful to note that CSPs with a certain structure can ``mislead'' GSAT, resulting in increased solution times and lower score solutions.

\subsubsection{Mixed Random Walk}
A variation of GSAT called WalkSat is introduced by Selman, et al.\ in \cite{Selman_1995_Local}. A major drawback of GSAT is that it can still quite easily become stuck on a local maxima and plateaus in the search space. By introducing noise, WalkSat handles many cases where GSAT has become stuck on a local maxima or search plateau. WalkSat introduces noise into the search by flipping a random variable in a randomly selected unsatisfied clause. By choosing the variable to flip from an unsatisfied clause, WalkSat maximizes the chances that the variable flip will lead to an improvement in the solution assignment.

\subsubsection{Tabu Search}
Glover and Laguna introduce the metaheuristic tabu search in \cite{Glover_1993_Modern} and \cite{Glover_1998_Tabu}. Tabu search is unique, because unlike most stochastic search based solvers, it follows the assumption that intelligent search is most likely more systematic than it is random. In order to systematically guide its search, tabu search emphasizes adaptive memory and responsive exploration. The adaptive memory tabu search uses focuses on recency, frequency, quality and influence. The flexibility gained from using adaptive memory gives tabu search an advantage over branch and bound solvers because it allows tabu search to more accurately discover influence throughout the search space. The responsive exploration of tabu search ``derives from the supposition that a bad strategic choice can yield more information than a good random choice'' \cite{Glover_1998_Tabu}. Responsive exploration also allows tabu search to explore different areas of the search space that seem as though they will yield promising results while not discarding any assignments found thus far.

\subsection{Backtracking Search Developments}
Significant developments in backtracking search based CSP solving are now briefly highlighted.

\subsubsection{Davis Putnam Logemann Loveland (DPLL) Procedure}
DPLL is a foundational piece of work put forth by Davis, Putnam, Logemann, and Loveland in \cite{Davis_1960_Computing} and \cite{Davis_1962_Machine}. DPLL was motivated by the authors' work in quantification theory. DPLL helped popularize expressing CSPs in CNF, and established a set of rules for solving CSPs that are still in use by many backtracking search based solvers today. These works established the rule for the elimination of one--literal clauses, the affirmative--negative rule, and the rule for eliminating atomic formulas.

\subsubsection{Clause Learning}
In \cite{Dechter_2003_Constraint}, Dechter reviews traditional clause learning. Backtracking search algorithms may encounter the same partial assignment that leads to a dead end numerous times. Making this discovery more than once is costly and results in repetition in the search space. Once a partial assignment has resulted in a dead end, the partial assignment may be analyzed and a minimal conflict set (no--good) determined. By adding this no--good as a new clause, the solver avoids repeatedly searching the same areas of the search space that lead to dead ends.

\subsubsection{Dependenc--Directed Backtracking and Dependency--Directed Backjumping}
A significant improvement to traditional clause learning is made by Stallman and Sussman in \cite{Stallman_1977_Forward}. In \cite{Stallman_1977_Forward}, Stallman and Sussman introduce dependency--directed backtracking, also known as dependency--directed backjumping and intelligent backtracking. Dependency--directed backtracking is triggered where a normal backtrack would occur. At this point, dependency--directed backtracking analyzes the failure and determines which variable assignment actually caused the failure. It then \emph{backjumps} directly to the variable assignment that led to the failure. Chronological backtracking potentially requires more computation to reach the variable assignment that caused the failure.

\subsubsection{Schiex and Verfaillie's No--good Recording}
In \cite{Schiex_1994_Nogood}, Schiex and Verfaillie propose a form of no--good recording that instead focuses on the justification for any no--good it wants to learn. No--good recording ``builds an approximate (polynomially bounded in space) description of the frontier of the space explored along with justifications relating the frontier to the CSP constraints.'' By recording an approximate description of the solution space thus far explored, Schiex and Verfaillie's algorithm records higher quality no--goods than most traditional approaches, such as that proposed by Dechter in \cite{Dechter_1990_Enhancement}. Through the recording of higher quality no--goods, solver's implementing this form of no--good recording are able to further prune the search space than solvers implementing traditional forms of no--good recording.

\subsubsection{Weak--commitment Search}
In \cite{Yokoo_1994_Weak}, Yokoo introduces weak--commitment search. Weak--commitment search recognizes that in backtracking search, if a bad decision was made regarding the partial assignment, there is no way for the search to revise its decision without performing an exhaustive search of the branch. In the case of large CSPs, this means that making a bad decision can have detrimental effects due to the large sizes of the branches in the search space. Weak--commitment search adds a mechanism for backtracking search to revise a bad decision. The ability for backtracking search to make such a revision substantially mitigates the damage that can be caused by a bad decision. According to Yokoo, weak-commitment search is very similar to iterative broadening, except that weak--commitment search preserves the search width when restarting.

\subsubsection{Branching Heuristics}
Marques--Silva concludes in \cite{Marques-Silva_1999_Impact} that search pruning techniques often have a larger impact than the selected branching heuristic, the heuristics used to decide which variable and assignment will be selected next. The branching heuristics Marques--Silva examines are BOHM, DLCS, DLIS, JW--OS, JW--TS, MOM, RAND, RDLIS. He evaluates these heuristics in the context of the GRASP CSP solver. Marques-Silva's evaluation revealed that although branching heuristics are important, they usually do not offer as much improvement as search pruning techniques.

\subsection{Hybrid Search Developments}
Significant developments in hybrid search are now briefly highlighted.

\subsubsection{Incomplete Dynamic Backtracking (IDB)}
In \cite{Prestwich_2001_Local}, Prestwich proposes a hybrid search based CSP solving technique called Incomplete Dynamic Backtracking (IDB). Prestwich recognized that although backtracking search often yields higher quality assignments than stochastic search, it frequently cannot scale as well, and therefore occasionally results in CSPs that are unsolvable due to time limitations. Prestwich determined that the property of stochastic search that allows it to scale better than backtracking search is that unlike backtracking search, stochastic search does not suffer from a strong commitment to early assignments. To counteract this commitment bias, when IDB encounters a dead end, it randomly selects a variable to backtrack to, leaving later assignments unchanged. The introduction of this stochastic element by IDB mitigates backtracking search's strong commitment to early assignments.

\chapter{The Problem of Real--Time Dynamic Weighted Constraint Satisfaction}

\section{Distributed Exploratory Episodic Planning (DEEP)}
The focus of this work stems from a project at the United States Air Force  (USAF) Research Laboratory, Information Directorate, called Distributed Exploratory Episodic Planning (DEEP). The objective of the DEEP project is to develop an assisted planning system that is mixed--initiative and uses a set of distributed episodic case bases maintained by autonomous agents~\cite{Ford_2008_Synthesizing}.

\subsection{Motivation behind the DEEP project}
In today's battle environments, optimal solution of the problems facing military commanders require more knowledge than that which is typically provided by their military training. How can commanders be provided the additional knowledge required to make good decisions and optimally solve the problems they face in battle? The DEEP project is designed to provide an answer to this question.

\subsection{The theory behind the DEEP project}
The DEEP project recognizes that knowledge is the result of experience. It further recognized that experience results from ``direct participation in or observation of an event or activity'' ~\cite{Ford_2007_Creating}. The individual conventionally draws their knowledge from their own experiences. It is not possible for an individual to have had as many experiences and therefore draw as much knowledge from those experiences as the collective experiences from a group of individuals. The provision of a collective of experiences from various individuals and a mechanism for semantically reasoning over that collective will make available to commanders the knowledge required to optimally solve battlefield problems.

\subsection{A primary challenge facing the DEEP project}
Although the DEEP project faces some significant challenges in various areas of active research, the primary challenge this work is intended to help the project overcome derives from the project's analogical reasoning algorithm~\cite{Ford_2007_Creating}.

The project's analogical reasoning algorithm utilizes a technique known as robust coherence~\cite{Ford_2008_Synthesizing, Ford_2009_Robust} to reason across stored experiences. Robust coherence represents a powerful enhancement over traditional deliberative coherence by addressing the idealism objection to deliberative coherence. Robust coherence is applied by first selecting a set of applicable experiences from the available collective of experiences. The second step to be performed is the analysis of the interactions between experiences. The resulting analysis is then encoded as a CSP problem, where the solution represents a dichotomy of experiences dichotomized into a set of maximally coherent and non--maximally coherent experiences.

The DEEP project's chosen CSP encoding associates a weight with each constraint representing the amount of coherence to be ``gained'' by satisfying the constraint.

\section{CSP Solver Qualities}
\label{csp_solver_qualities}
The CSP solver implemented by the robust coherence technique as part of the DEEP project's analogical reasoning algorithm must not be arbitrarily selected. The project's application of robust coherence necessitates a solver that meets a unique set of requirements.

\subsection{WCSP capable}
The chosen CSP encoding is that of a WCSP. Therefore, a WCSP solver is required to solve it. Because a solution from a WCSP solver is the least cost solution to the WCSP, when viewing the associated weights as the benefited score of satisfying the constraints, the solution is also the solution that maximizes the score of the satisfied constraints. Viewing the WCSP in the robust coherence implementation in this fashion makes obvious the solution represents the maximally coherent set of experiences.

\subsection{DWCSP capable}
The WCSP solver utilized by the project's analogical reasoning algorithm will initially operate on a fixed WCSP instance. As the sets of experiences available to the analogical reasoner change, the set of applicable experiences may also change. In the cases that a new applicable experience has become available or an experience currently in use becomes no longer applicable, and the robust coherence mechanism has updated its analysis, it is required that the WCSP solver component of the robust coherence mechanism need not be restarted from scratch. This requirement is satisfied by a WCSP solver that supports the dynamic addition and removal of constraints to the WCSP.

\subsection{Performance in real--time environments}
Although the idealized application of the DEEP project is to early Command and Control (C2) planning, potential applications of the project cannot be guaranteed to preclude real--time C2 planning. A shortcoming of the robust coherence mechanism used by the project that is immediately brought to light when considering it for real--time application is the mechanism's WCSP solver. Since the time required to solve a WCSP is unpredictable, having to do so is a real risk to achieving real--time performance. The WCSP solver implemented by the robust coherence mechanism must therefore support real--time performance.

\chapter{Real--time Dynamic Stochastic Search Based WCSP Solving}
This work focuses on the development of two algorithms. The first algorithm is called DMaxWalkSat, and the second is called RDMaxWalkSat. DMaxWalkSat extends MaxWalkSat for DWCSP solving, while RDMaxWalkSat extends DMaxWalkSat for real--time DWCSP solving. The following sections define DMaxWalkSat and RDMaxWalkSat.

\section{WCSP: MaxWalkSat}
MaxWalkSat is a generalization of WalkSat, which was proposed by Selman, et al.\ in \cite{Selman_1994_Noise} for solving WCSPs. MaxWalkSat itself is proposed in \cite{Jiang_1995_Solving} and is reproduced in Algorithm \ref{alg_MaxWalkSat}.

\begin{algorithm}
	\caption{Pseudocode algorithm for MaxWalkSat as proposed in \cite{Jiang_1995_Solving}.}
	\label{alg_MaxWalkSat}
	\begin{algorithmic}[1]
	\STATE \textbf{procedure} MaxWalkSat(weighted\_clauses, hard\_limit, max\_flips, target, max\_tries, noise)
	\STATE $m \leftarrow$ random truth assignment for variables appearing in weighted\_clauses
	\STATE $hard\_unsat \leftarrow$ clauses unsatisfied by $M$ with weight $\geq$ hard\_limit
	\STATE $soft\_unsat \leftarrow$ clauses unsatisfied by $M$ with weight $<$ hard\_limit
	\STATE $bad \leftarrow$ sum of the weights of the clauses in hard\_unsat and soft\_unsat
	\FOR{$i \leftarrow 0$ \TO max\_tries}
		\FOR{$j \leftarrow 0$ \TO max\_flips}
			\IF{$bad <$ target}
				\STATE $i \leftarrow$ max\_tries $+$ 1
				\STATE $j \leftarrow$ max\_flips $+$ 1
			\ENDIF
			
			\IF{$hard\_unsat$ \NOT empty}
				\STATE $c \leftarrow$ random clause from $hard\_unsat$
			\ELSE
				\STATE $c \leftarrow$ random clause from $soft\_unsat$
			\ENDIF
			\STATE $heads \leftarrow$ random coin toss with probability noise of heads
			\IF{$heads$}
				\STATE $p \leftarrow$ randomly chosen variable from $c$
			\ELSE
				\FORALL{variables $q$ in $c$}
					\STATE $breakcount[q] \leftarrow 0$
					\FORALL{clauses $d$ containing $q$}
						\IF{$d$ is satisified by $M$, but unsatisfied by \NOT $q$}
							\STATE $breakcount[q] \leftarrow breakcount[q]$ $+$ weight of $d$
						\ENDIF
					\ENDFOR
				\ENDFOR
				
				\STATE $p \leftarrow q$ where $q = \min(breakcount)$
			\ENDIF
			
			\STATE Flip value assigned to $p$ in $m$
			\STATE Update $hard\_unsat$, $soft\_unsat$, and $bad$
		\ENDFOR
	\ENDFOR
	\PRINT "Solution: ", $m$
	\PRINT "Cost: ", $bad$
	\end{algorithmic}
\end{algorithm}

\section{DWCSP: DMaxWalkSat}
A defining requirement of DCSP solvers is the ability to add and remove constraints to the CSP. This requirement holds for DWCSP. The addition of the capability to add and remove constraints from MaxWalKSat forms the DMaxWalkSat DWCSP solving algorithm.

\subsection{Adding a Constraint}
Two steps are performed by DMaxWalkSat to add a constraint. The first step performed is the addition of the constraint itself to the internal list of constraints. The second step depends on whether or not the variables constrained by the added constraint have been added to the solver before. No action is taken for constrained variables that have already been added to the solver. Constrained variables that have not already been added to the solver are added. The pseudocode for adding a constraint is shown in Algorithm \ref{alg_dmaxwalksat_addition}.

\begin{algorithm}
	\caption{Pseudocode for adding a constraint}
	\label{alg_dmaxwalksat_addition}
	\begin{algorithmic}
		\REQUIRE Previous states from MaxWalkSat
		\STATE \textbf{procedure} Add\_Constraint(clause)
		\STATE $m \leftarrow$ random truth assignment for any new variables in clause
		\IF{weight of clause $\geq$ hard\_limit}
			\STATE add clause to $hard\_unsat$
		\ELSE
			\STATE add clause to $soft\_unsat$
		\ENDIF
		\STATE Repeat MaxWalkSat starting on line 6
	\end{algorithmic}
\end{algorithm}

\subsection{Removing a Constraint}
DMaxWalkSat performs one simple step to remove a constraint, the removal of the constraint from the internal list of constraints. Although some of the variables present in the solver may become unconstrained, no degradation is caused by allowing the variables to remain in the internal data structures of the solver. The pseudocode for removing a constraint is shown in Algorithm \ref{alg_dmaxwalksat_removal}.

\begin{algorithm}
	\caption{Pseudocode for removing a constraint}
	\label{alg_dmaxwalksat_removal}
	\begin{algorithmic}
		\REQUIRE Previous states from MaxWalkSat
		\STATE \textbf{procedure} Remove\_Constraint(clause)
		\IF{weight of clause $\geq$ hard\_limit}
			\STATE remove clause from $hard\_unsat$
		\ELSE
			\STATE remove clause from $soft\_unsat$
		\ENDIF
		\STATE Repeat MaxWalkSat starting on line 5
	\end{algorithmic}
\end{algorithm}


\section{Real--time DWCSP: RDMaxWalkSat}
\label{lbl_real--time}
Real--time performance implies that an algorithm either terminates in a regular interval or that it can be interrupted at an arbitrary point in time and still yield valid results. RDMaxWalkSat was created by changing DMaxWalkSat into an anytime algorithm.

\subsubsection{Anytime Algorithm}
DMaxWalkSat is made into an anytime algorithm by maintaining a ``best'' solution and making it available at any point in time during its execution. If the solver is queried for the solution before it has successfully completed its search, the best solution thus far is returned. In the case that the solver is queried for the solution before it has completed any searching, the initial randomly generated solution is returned since it is the best solution the solver has seen thus far.

Algorithm \ref{alg_MaxWalkSat} already maintains an optimal solution in the form of $m$. Therefore, the only change to MaxWalkSat or DMaxWalkSat required to make either be an anytime algorithm is the addition of an interruption mechanism. RDMaxWalkSat is an implementation of DMaxWalkSat with the addition of such an interruption mechanism.

\chapter{Implementation}
This work is characterized primarily as the effort to introduce dynamic and real--time performance to the field of weighted constraint satisfaction for application in the robust coherence algorithm. The implementation and evaluations of DMaxWalkSat and RDMaxWalkSat resulting from this effort are focused on demonstrating that RDMaxWalkSat satisfactorily meets the requirements of the constraint satisfaction solver requirements described in Section \ref{csp_solver_qualities}.

\section{Implementation phases}
RDMaxWalkSat was implemented in three phases. The first phase performed was the implementation of MaxWalkSat (681 lines of code). The second phase performed was the implementation of the necessary extensions to MaxWalkSat that forms DMaxWalkSat (an additional 130 lines of code). The third and final phase performed was the implementation of the necessary extensions to DMaxWalkSat that form RDMaxWalkSat (an additional 298 lines of code).

\section{Implementation details}
All three phases were completed using the C++ programming language. The implementations also made use of the cross--platform Boost libraries, as well as the POSIX threading library. The use of C++ and freely available, cross--platform libraries allows the resulting implementations to be easily ported across different platforms. The development platforms used were OS X and Linux.

\section{Utilities}
Various utilities that are not part of the core implementation were written to support this work.

\subsection{DimacsReader}
One utility created was a class named DimacsReader. DimacsReader operates on a slightly modified version of DIMACS designed for WCSP and supporting only those features of it that are minimally required for problem representation. DIMACS is a popular format for storing WCSP and CSP instances \cite{1993_Satisfiability}. DimacsReader allows MaxWalkSat, DMaxWalkSat, and RDMaxWalkSat to operate on a problem instance that has been saved as a corresponding DIMACS file. DimacsReader is also written in C++ (133 lines of code).

\subsection{ConstraintGrapher}
Another utility created was a class named ConstraintGrapher. ConstraintGrapher generates a constraint graph for a CSP instance. ConstraintGrapher uses the Python Graphviz library available from \cite{pygraphviz}. ConstraintGrapher was written using the Python programming language (283 lines of code).

\subsection{Miscellaneous Scripts}
A number of Python scripts were written to perform small tasks such as generating graphs of the results from the different experiments, etc. It is estimated that approximately 1028 lines of code make up such scripts.

\chapter{Experiments}

\section{A Common Problem Corpus}
\begin{figure}
    \centering   \includegraphics[width=6in]{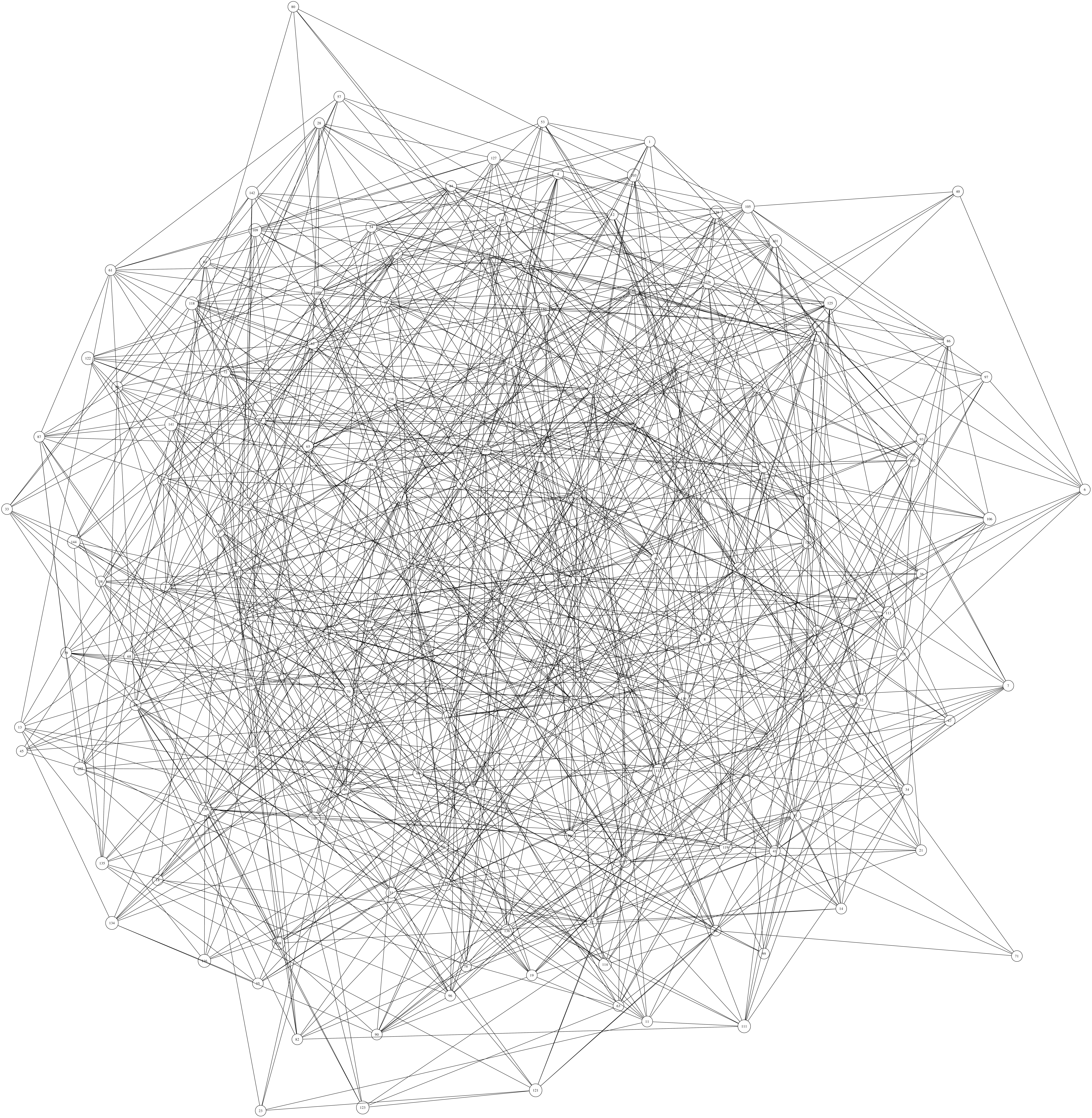}
    \caption{Constraint graph for lo/file\_rwpms\_wcnf\_L2\_V150\_C1000\_H150\_0.}
    \label{fig_lo/file_rwpms_wcnf_L2_V150_C1000_H150_0-cg}
\end{figure}

The experiments performed in this work operated on a common corpus of WCSPs. The corpus used was the set of 2-SAT random weighted partial MAX--SAT benchmark problems used in the 2009 Max--SAT evaluation \cite{solver_benchmarks}. Figure \ref{fig_lo/file_rwpms_wcnf_L2_V150_C1000_H150_0-cg} shows a constraint graph for one of the problems in the corpus. The constraint graphs for the rest of the problems in the corpus have been omitted due to the large number of constraints.

The selected corpus does not contain any DWCSPs, only WCSPs. This work generated DWCSPs from the WCSPs contained in the corpus. DWCSPs were generated from each WCSP in the corpus by using the first half the of constraints from the WCSP as the initial constraints. Constraints from the second half of the WCSP were then added 250 at a time to the initial set of constraints. Each addition marked a new sequence in the DWCSP. When no constraints in the second half of the WCSP remained to be added, the added constraints were then removed 250 at a time from the set of initial constraints. Each removal also marked a new sequence in the DWCSP.

\section{The Control Group}
When conducting an experiment, it is necessary to have a control group. The control group in this experiment is the MaxWalkSat solver.

MaxWalkSat was run on each problem in the corpus thirty times. Each trial was started with a randomly generated initial solution. The quality of each initial solution was calculated and recorded as well as the time the solver took. Graphs of the time versus the initial solution quality were generated and are shown in Appendix \ref{maxwalksat_graphs}. Figure \ref{fig_lo/file_rwpms_wcnf_L2_V150_C1000_H150_0-time_vs_initial_score} shows an example of one of the graphs for one of the problems (``lo/file\_rwpms\_wcnf\_L2\_V150\_C1000\_0'') in the corpus. The graphs show that, in general, as the quality of the initial solutions improved, the time the solver took decreased.

\begin{figure}
    \centering   \includegraphics[width=6in]{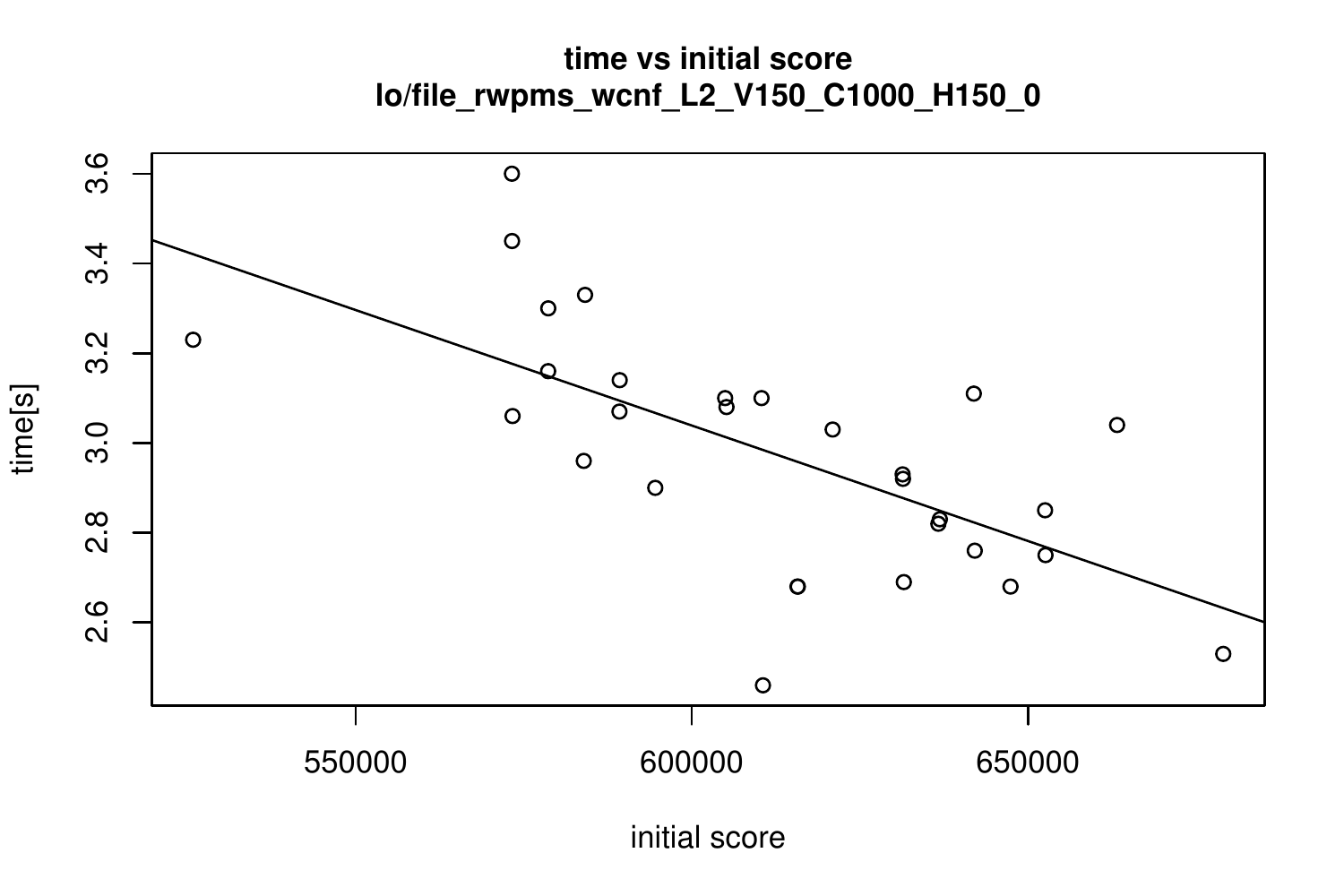}
    \caption{Graph for lo/file\_rwpms\_wcnf\_L2\_V150\_C1000\_H150\_0 showing the distribution of solution times against the quality of the randomly generated initial solution assignments.}
    \label{fig_lo/file_rwpms_wcnf_L2_V150_C1000_H150_0-time_vs_initial_score}
\end{figure}

The average times MaxWalkSat took to solve each problem in the corpus were calculated and are summarized in Table \ref{maxwalksat_runtimes}.

\section{DMaxWalkSat Experiment}
The first experiment performed investigated the DMaxWalkSat solver. We wanted to determine if this solver, a specialized DWCSP solver, offered any advantages when solving DWCSPs as compared to solving them with a WCSP solver. Advantages we expected we might see were substantially higher scores or significantly shorter run--times. The WCSP solver we used for this experiment was MaxWalkSat, which would be required to solve each DWCSP as a sequence of WCSP problems.

\subsection{Procedure}
This experiment started by first solving each initial DWCSP problem from the same randomly generated initial solution using DMaxWalkSat and MaxWalkSat. The constraint additions in each DWCSP were then processed. For each constraint addition, the constraint addition was added to the DMaxWalkSat solver using its native procedures for doing so. In the case of the MaxWalkSat solver, every constraint addition necessitated initializing a new instance of MaxWalkSat, but this time containing all of the previous constraints plus the new constraints from the constraint addition. The solvers were then each allowed to solve their constraints again. After all the constraint additions in the DWCSP were processed, the constraint removals were then processed in the same way. All DWCSPs were solved in this way 30 times. The run--time of each solver, the quality of the initial solution, and the quality of the generated solution were recorded for each trial.

\begin{figure}
	\centering
	\includegraphics[width=5.5in]{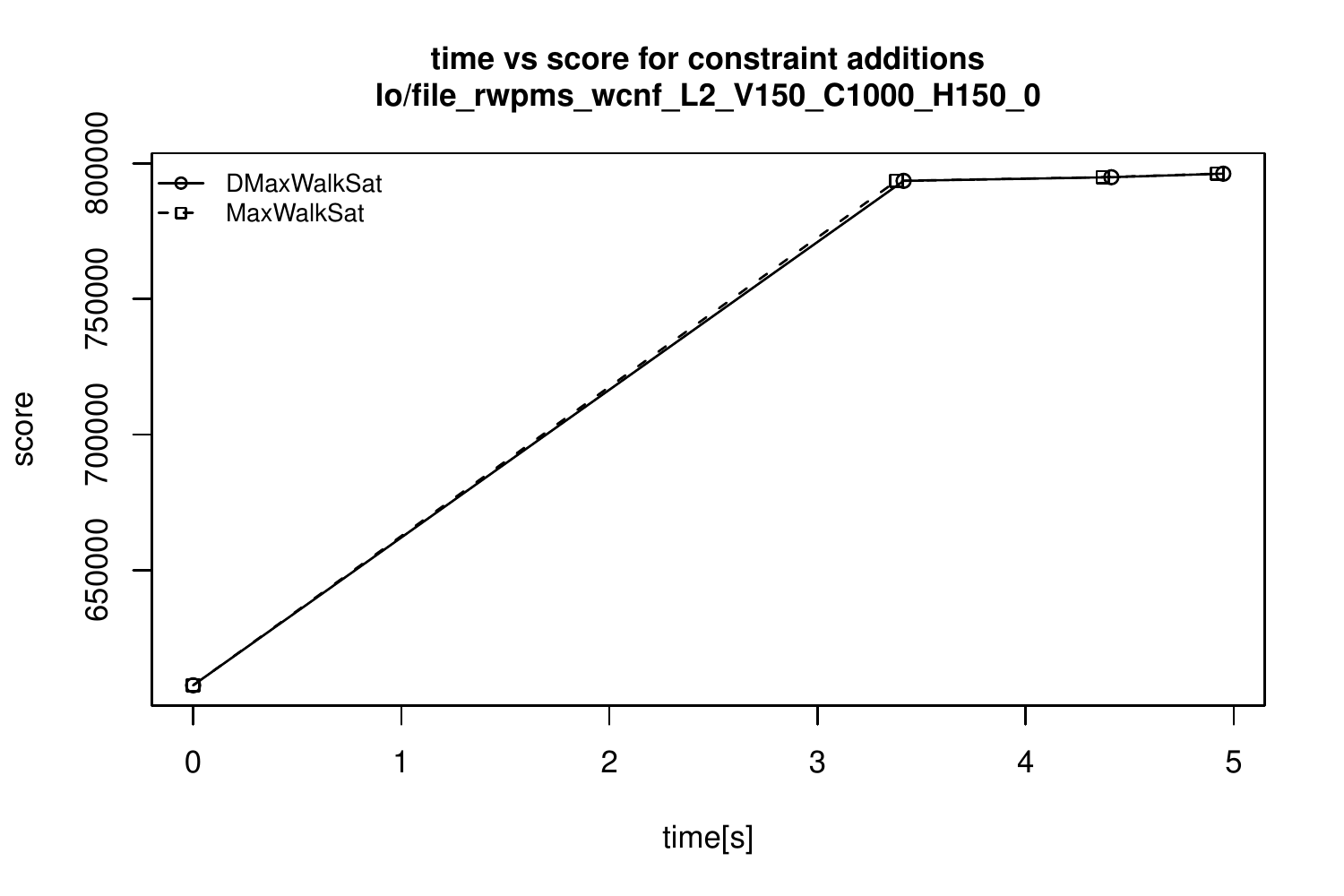}
	\caption{The solution scores for DMaxWalkSat and MaxWalkSat vs time for constraint additions.}
	\label{fig_dmaxwalksat_upward}
\end{figure}

\begin{figure}
	\centering
	\includegraphics[width=5.5in]{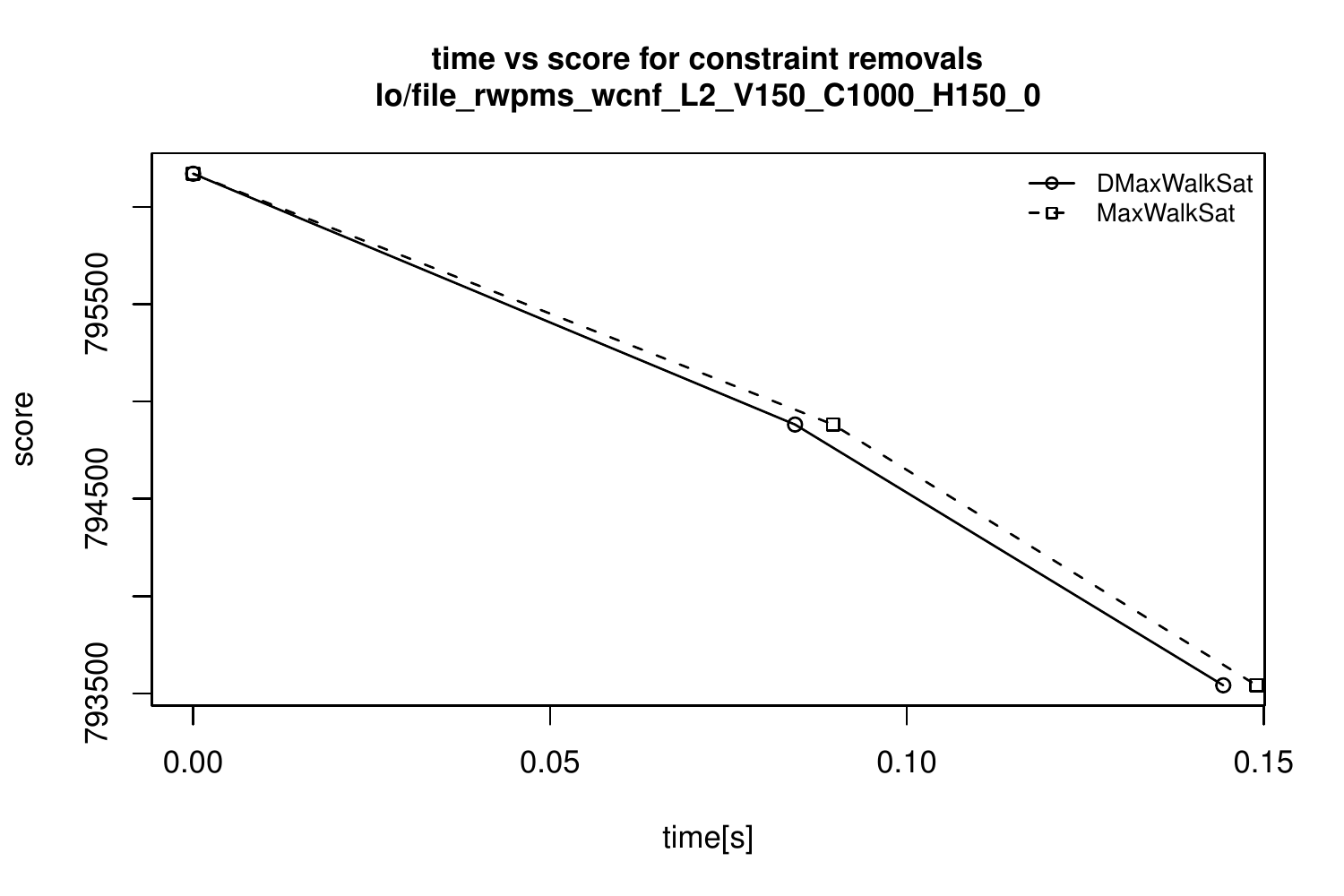}
	\caption{The solution scores for DMaxWalkSat and MaxWalkSat vs time for constraint removals.}
	\label{fig_dmaxwalksat_downward}
\end{figure}

\subsection{Results}
The mean upward and downward solution velocities of each solver were calculated for each problem in the corpus. In this work, upward solution velocity is the improvement in solution quality per unit time for constraint additions, while downward solution velocity is the improvement in solution quality per unit time for constraint removals. The upward and downward solution velocities are summarized in Table \ref{tbl_dmaxwalksat_velocities}. Figure \ref{fig_dmaxwalksat_upward} shows the run--time differences between DMaxWalkSat and MaxWalkSat for constraint additions for a problem in the corpus (``lo/file\_rwpms\_wcnf\_L2\_V150\_C1000\_0''), while Figure \ref{fig_dmaxwalksat_downward} shows the run--time differences for constraint removals. Graphs for the remaining problems are shown in Appendix \ref{dmaxwalksat_graphs}.

MaxWalkSat had a greater upward solution velocity than DMaxWalkSat for 3.3\% of the problems in the corpus. However, DMaxWalkSat had a greater solution velocity than MaxWalkSat for the remaining 96.7\% of the problems. MaxWalkSat had a greater downward solution velocity also for 3.3\% of the problems. DMaxWalkSat had a greater downward solution velocity for 95.6\% of the problems. For 1.1\% of the problems, MaxWalkSat and DMaxWalkSat had equal downward solution velocities.

\section{RDMaxWalkSat Experiment}
The final experiment performed analyzed the RDMaxWalkSat solver. We sought to determine whether solutions with significant improvements in quality over their initial solution could be gained from RDMaxWalkSat at arbitrary points in time.

\subsection{Procedure}
For this experiment, the time thresholds each solution score would be sampled at were first calculated. The time thresholds were calculated for each problem by taking 0\%, 25\%, 50\%, and 75\% of the time taken by MaxWalkSat (which itself represents 100\%) to solve the same problem. RDMaxWalkSat was then run on each problem with samples taken at each time threshold a total of 30 times. The solution scores from each sampling were recorded for each problem.

\subsection{Results}
For each problem, the mean solution scores were calculated for all sampled thresholds. Graphs were then generated showing the mean solution score in comparison to the time thresholds and can be seen in Appendix \ref{rdmaxwalksat_graphs}. Figure \ref{fig_rdmaxwalksat_lo/file_rwpms_wcnf_L2_V150_C1000_H150_0} shows an example graph for one of the problems (``lo/file\_rwpms\_wcnf\_L2\_V150\_C1000\_H150\_0''). The difference in the mean solution score between each set of adjacent time thresholds was then calculated and is summarized in Table \ref{tbl_rdmaxwalksat_deltas}. 

\begin{figure}
	\centering
	\includegraphics[width=5.5in]{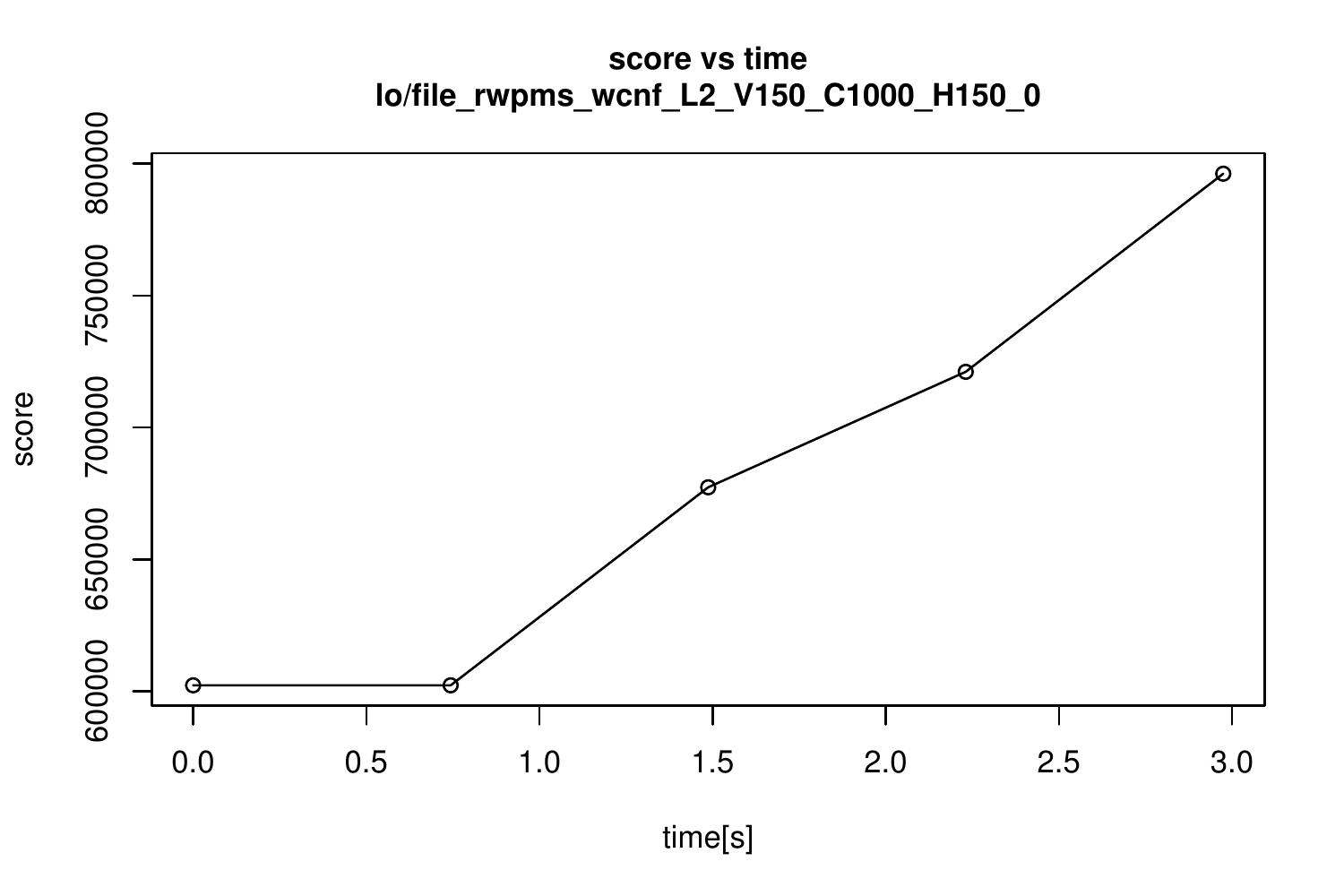}
	\caption{Score vs time for lo/file\_rwpms\_wcnf\_L2\_V150\_C1000\_H150\_0.}
	\label{fig_rdmaxwalksat_lo/file_rwpms_wcnf_L2_V150_C1000_H150_0}
\end{figure}

None of the problems showed equal improvement in solution score among all sample times. This indicates a non--linear relationship between solver time and solution score for the tested problems. In all cases, the interval between the 75\% and 100\% time thresholds had the greatest improvement in solution score. Unfortunately, the results indicate that the last time interval is when the solution score improved the most. Therefore, decreasing the duration of the last time interval is likely to have adverse effects on the solution score. Had the solution score improved more in earlier intervals, it would be less costly to decrease the duration of the last interval. However, the results do indicate that substantial improvements can be made to the solution score in less than 100\% of the time required by MaxWalkSat.

 As shown in some of the graphs in Appendix \ref{rdmaxwalksat_graphs}, Figure \ref{fig_rdmaxwalksat_lo/file_rwpms_wcnf_L2_V150_C1000_H150_0}, and Table \ref{tbl_rdmaxwalksat_deltas}, some of the solution score deltas for the first time sample were 0. This phenomenon occurred only on problems that have a relatively small run--time. We believe that in these instances, the time interval was smaller than the time required for the solver to make a first pass. However, we note that even in these cases, the solver returned a viable solution.

\section{Future Studies and Potential Applications}
Both DMaxWalkSat and RDMaxWalkSat inherit three parameters from MaxWalkSat. The parameters are the maximum number of flips, the maximum number of tries, and the amount of noise. Neither of the studies in this work changed these parameters from their default values. RDMaxWalkSat features a fourth tunable parameter, the time for which it should be allowed to run. Future researchers may elect to study the performance of these algorithms after incorporating some kind of adaptive learning mechanism that would be responsible for tuning these parameters based on previous performances on a class of problems. For example, a genetic algorithm may be used.

Another future study may wish to examine the effects DMaxWalkSat constraint addition and removal have on solution stability. This study was more interested in solution quality, and therefore did not make any attempts to examine solution stability.

\chapter{Conclusions}
The increasingly frequent application of constraint satisfaction techniques for modeling problems introduces an increased need for a variety of different CSP solvers. Further, the specialized nature of these applications increases the need for specialized CSP solvers, such as WCSP solvers, DCSP solvers, DWCSP solvers, and real--time capable variants.

This work introduced a new DWCSP solver, DMaxWalkSat, and a new real--time capable DWCSP solver, RDMaxWalkSat. DMaxWalkSat performed better on DWCSP problems on average than the stochastic search based solver it derives from, MaxWalkSat. Likewise, RDMaxWalkSat performed better on real--time DWCSP problems than did MaxWalkSat. These algorithms are each small and non--complex, allowing for implementation by even the most casual users.

The algorithms proposed in this work allow for the implementation of new and exciting algorithms, which themselves make use of DWCSP or real--time DWCSP problems. Previously, algorithms making use of certain specialized CSP solvers, such as robust coherence, could not be easily implemented without substantial efforts because an appropriate constraint solver did not exist.


\bibliographystyle{plain}
\bibliography{thesis}

\begin{appendix}
\chapter*{Appendices\addcontentsline{toc}{chapter}{Appendices}}
\begin{singlespace}

\chapter{MaxWalkSat Initial Score vs. Time Graphs\label{maxwalksat_graphs}}
\begin{figure}[H]
    \centering
    \includegraphics[height=3.5in]{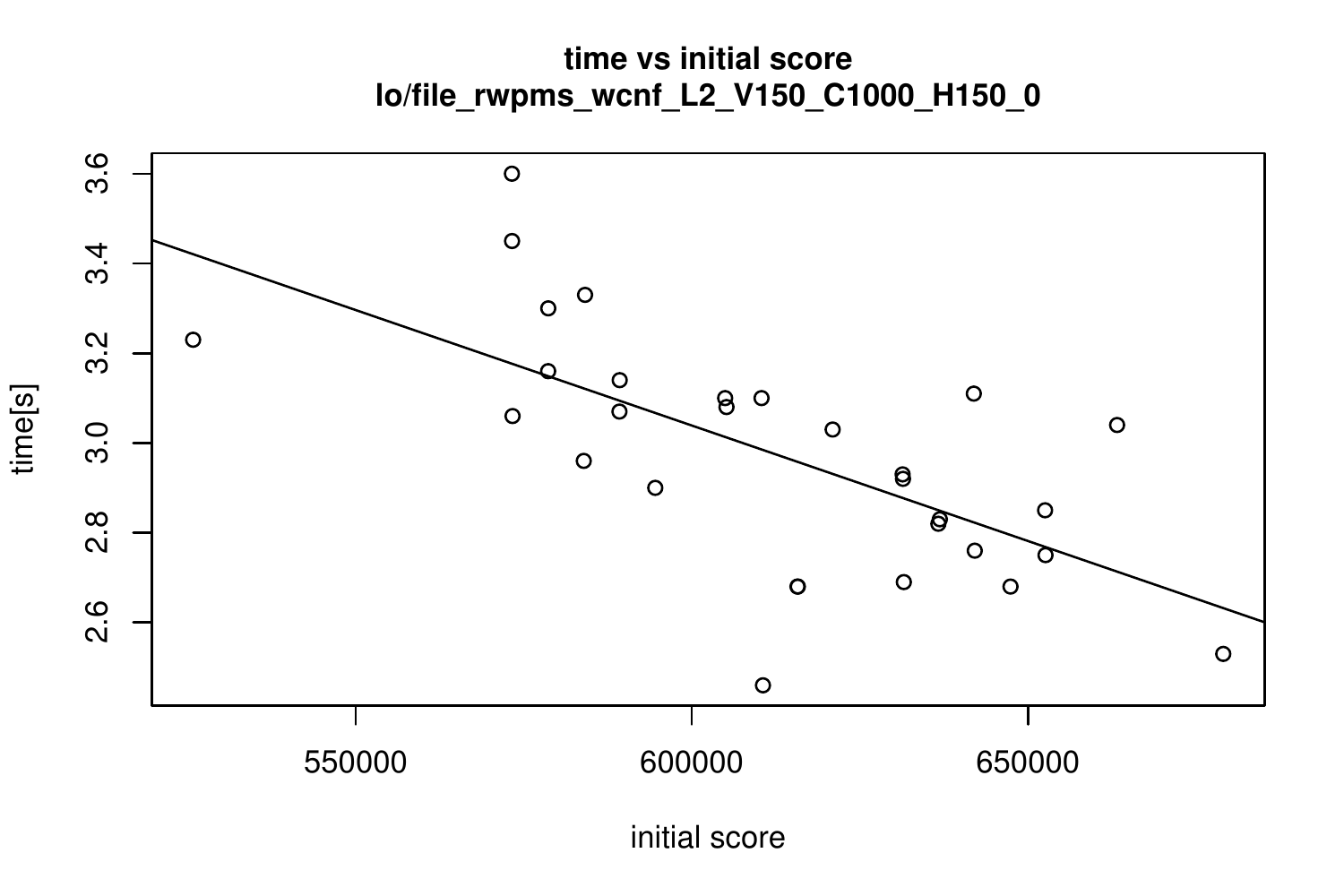}
    \label{fig_lo/file_rwpms_wcnf_L2_V150_C1000_H150_0/file_rwpms_wcnf_L2_V150_C1000_H150_0-time_vs_initial_score}
\end{figure}

\begin{figure}[H]
    \centering
    \includegraphics[height=3.5in]{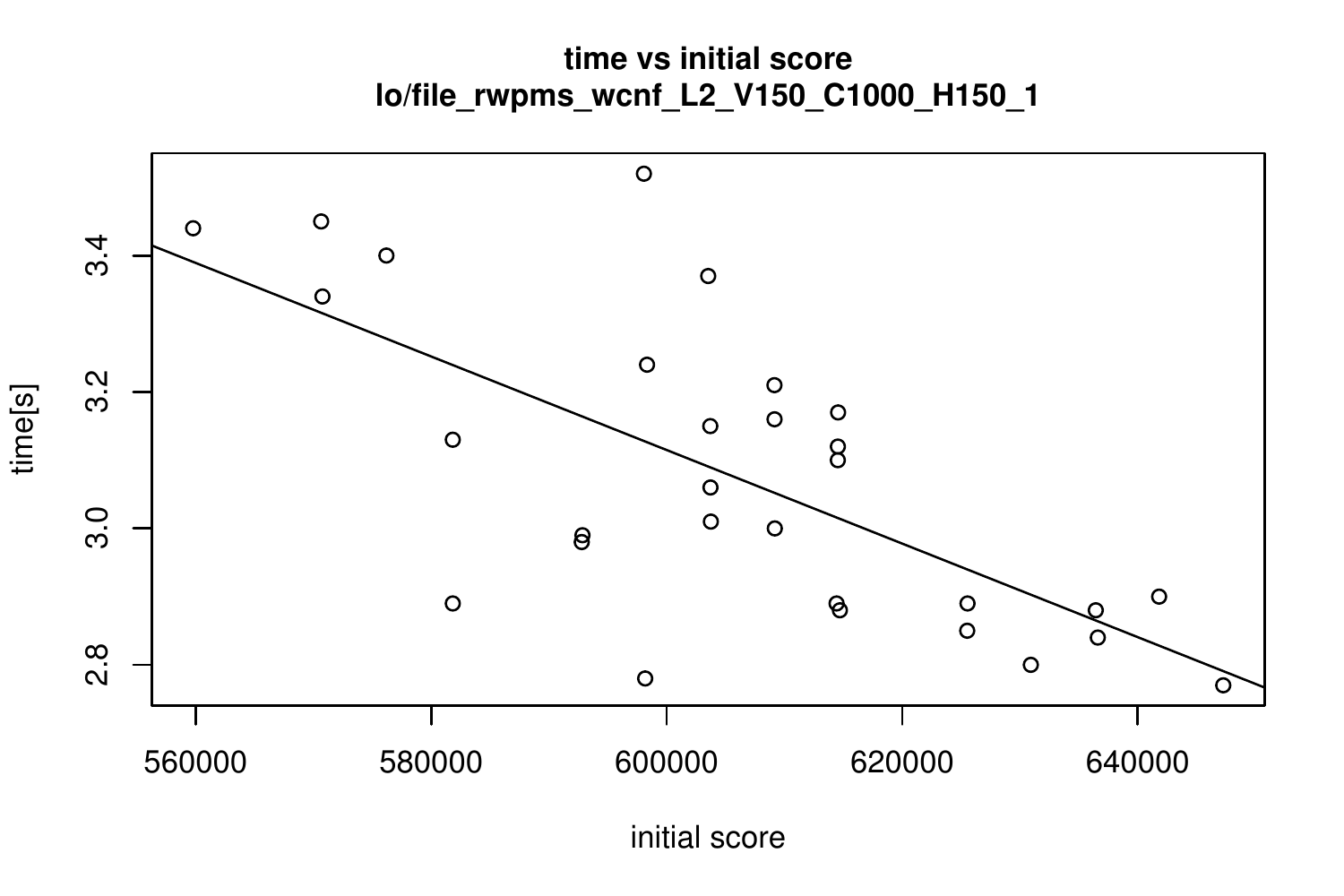}
    \label{fig_lo/file_rwpms_wcnf_L2_V150_C1000_H150_1/file_rwpms_wcnf_L2_V150_C1000_H150_1-time_vs_initial_score}
\end{figure}

\begin{figure}[H]
    \centering
    \includegraphics[height=3.5in]{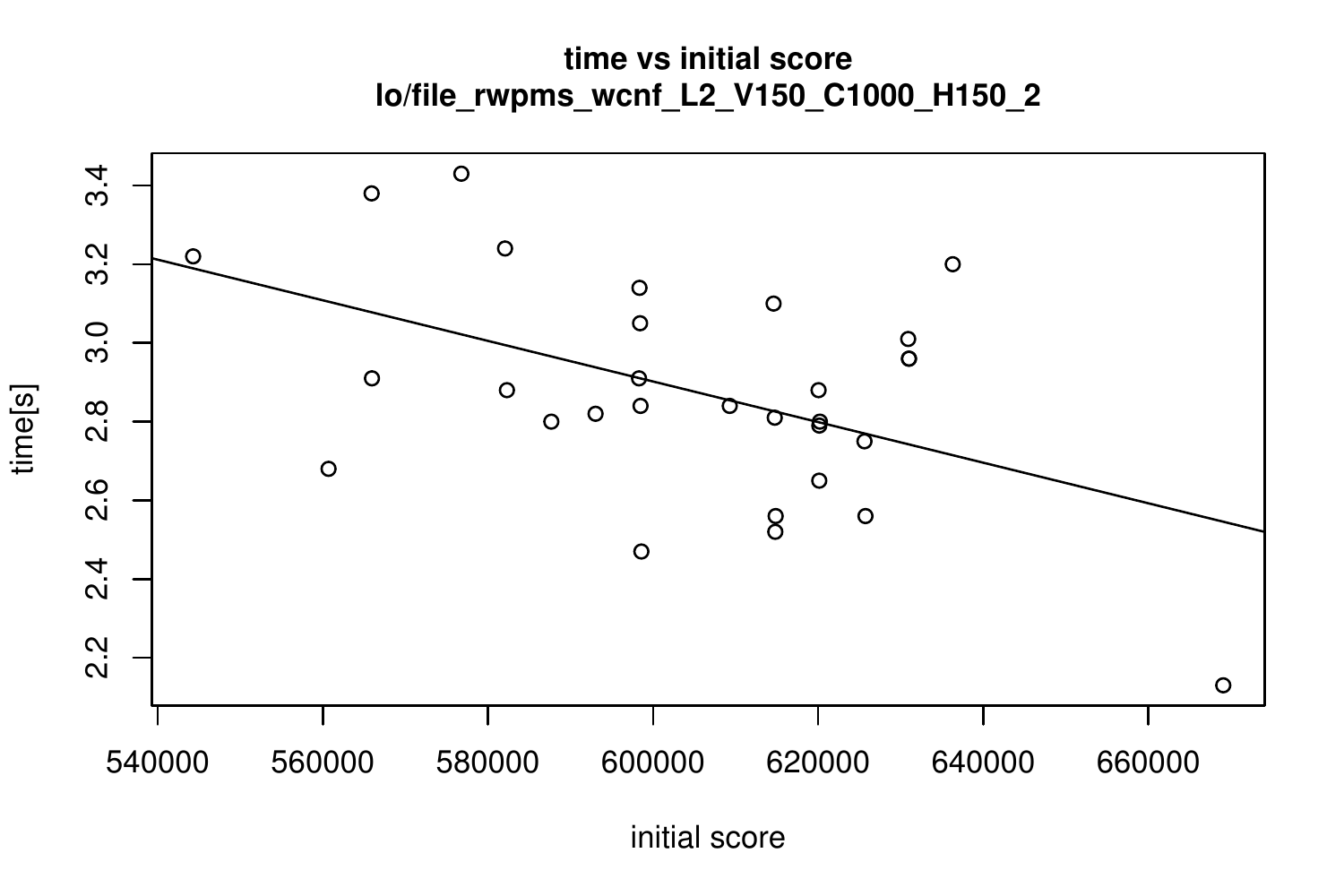}
    \label{fig_lo/file_rwpms_wcnf_L2_V150_C1000_H150_2/file_rwpms_wcnf_L2_V150_C1000_H150_2-time_vs_initial_score}
\end{figure}

\begin{figure}[H]
    \centering
    \includegraphics[height=3.5in]{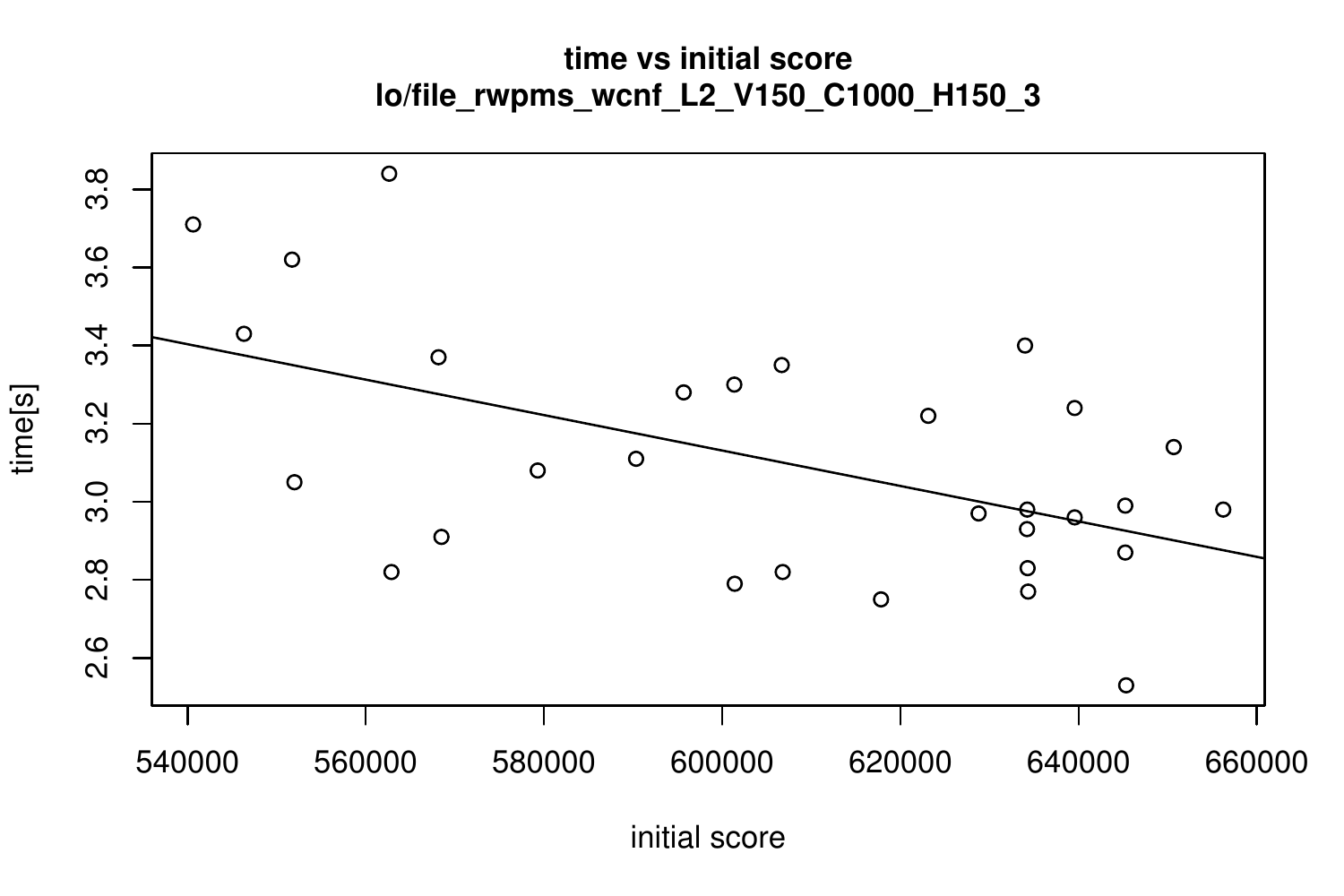}
    \label{fig_lo/file_rwpms_wcnf_L2_V150_C1000_H150_3/file_rwpms_wcnf_L2_V150_C1000_H150_3-time_vs_initial_score}
\end{figure}

\begin{figure}[H]
    \centering
    \includegraphics[height=3.5in]{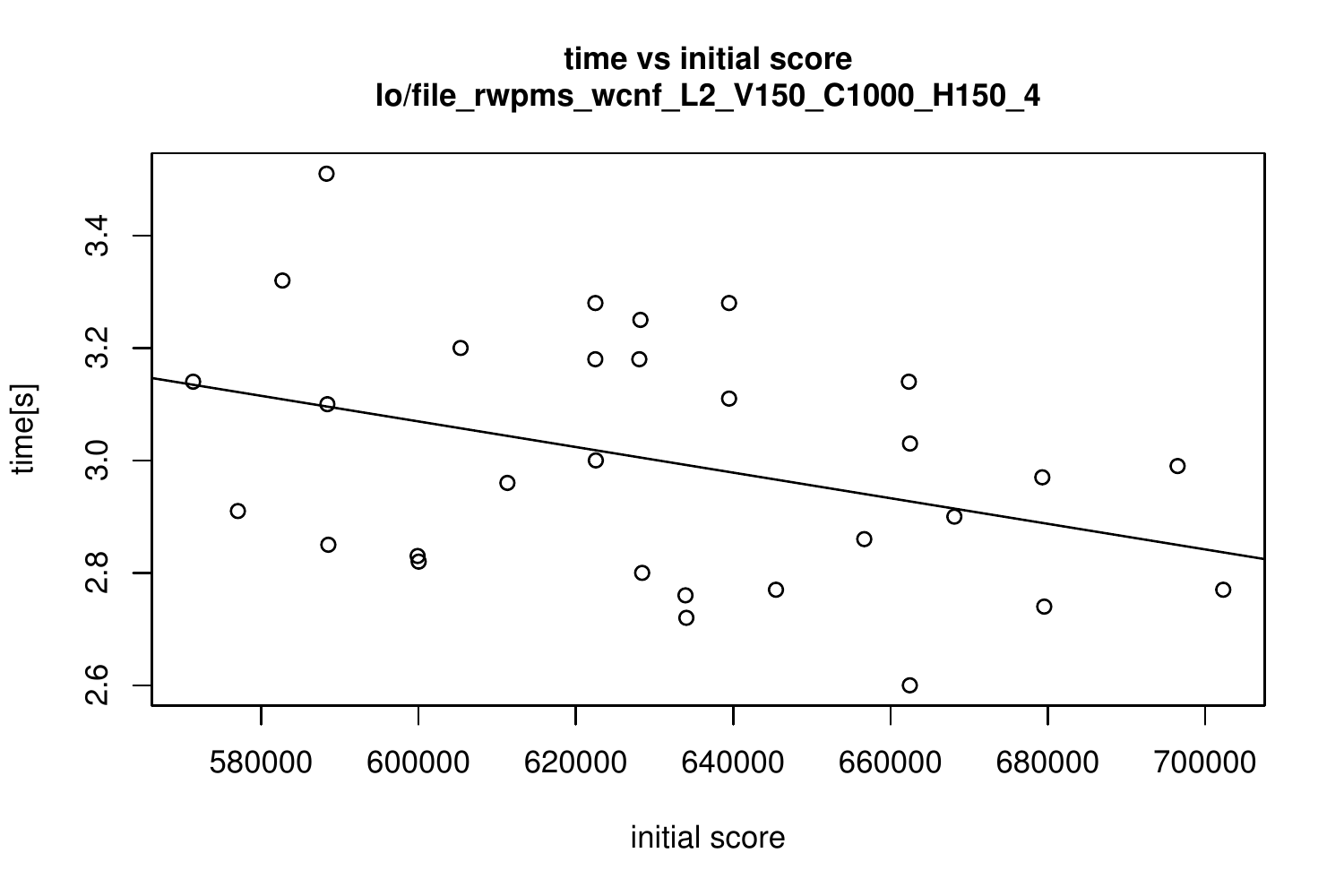}
    \label{fig_lo/file_rwpms_wcnf_L2_V150_C1000_H150_4/file_rwpms_wcnf_L2_V150_C1000_H150_4-time_vs_initial_score}
\end{figure}

\begin{figure}[H]
    \centering
    \includegraphics[height=3.5in]{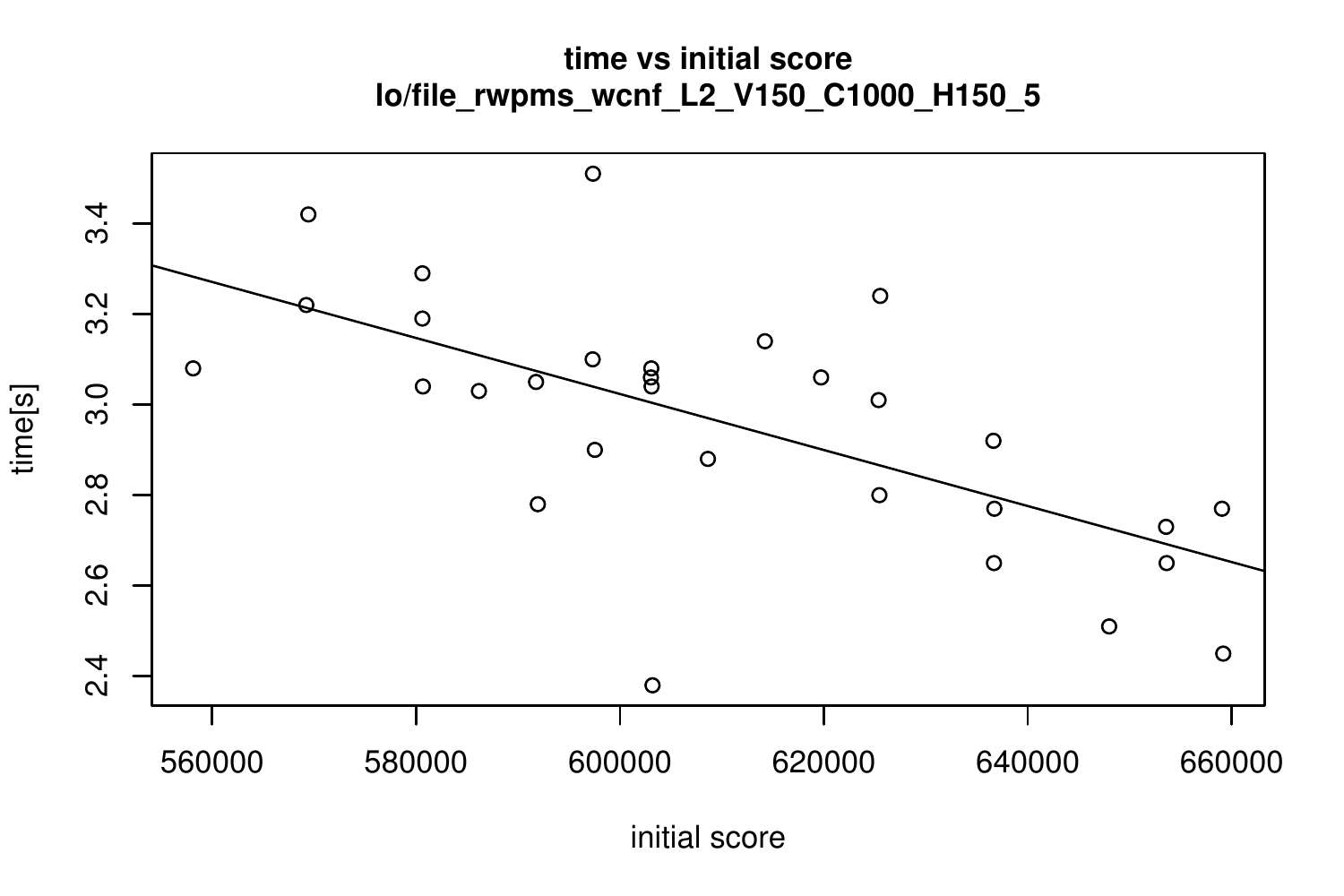}
    \label{fig_lo/file_rwpms_wcnf_L2_V150_C1000_H150_5/file_rwpms_wcnf_L2_V150_C1000_H150_5-time_vs_initial_score}
\end{figure}

\begin{figure}[H]
    \centering
    \includegraphics[height=3.5in]{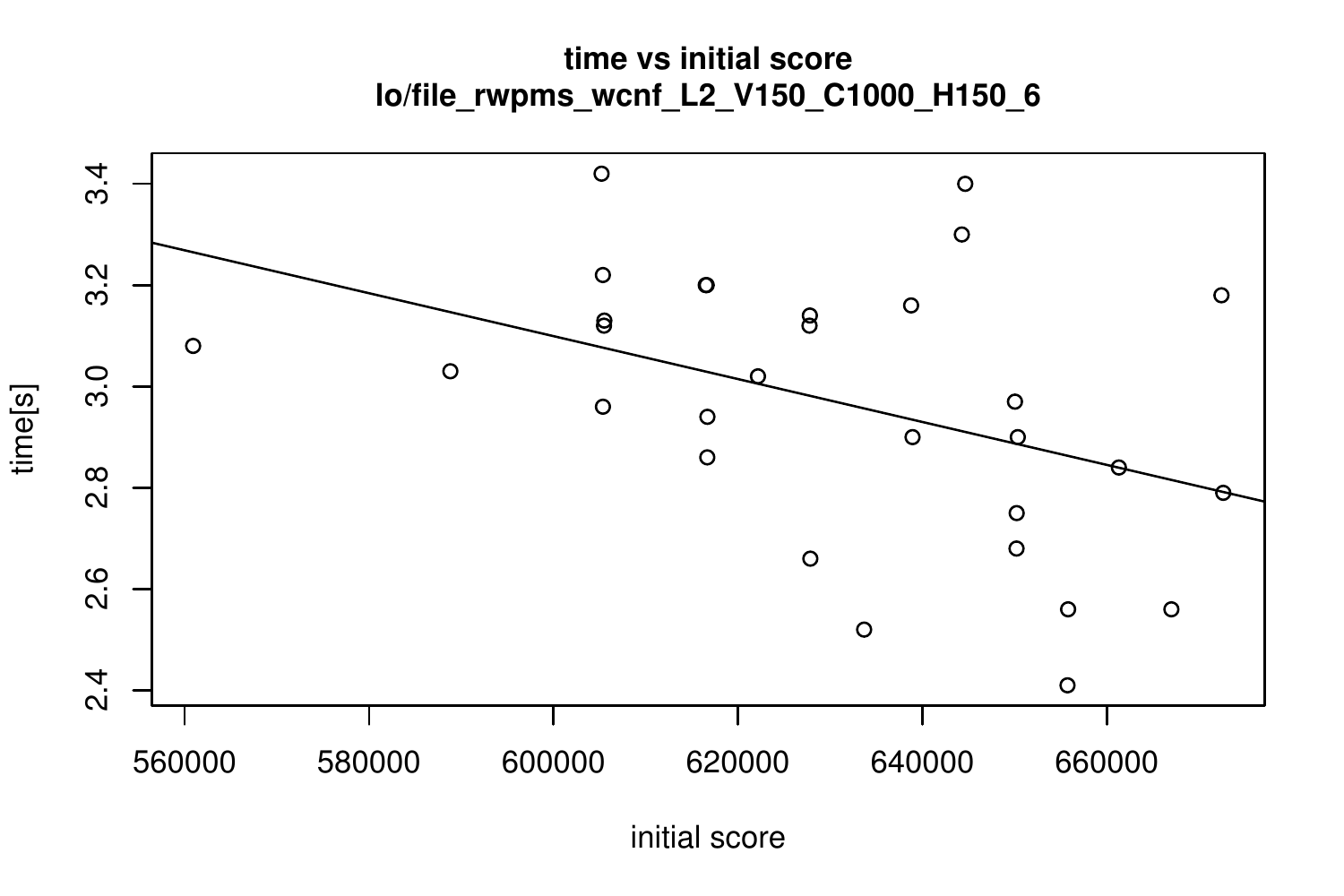}
    \label{fig_lo/file_rwpms_wcnf_L2_V150_C1000_H150_6/file_rwpms_wcnf_L2_V150_C1000_H150_6-time_vs_initial_score}
\end{figure}

\begin{figure}[H]
    \centering
    \includegraphics[height=3.5in]{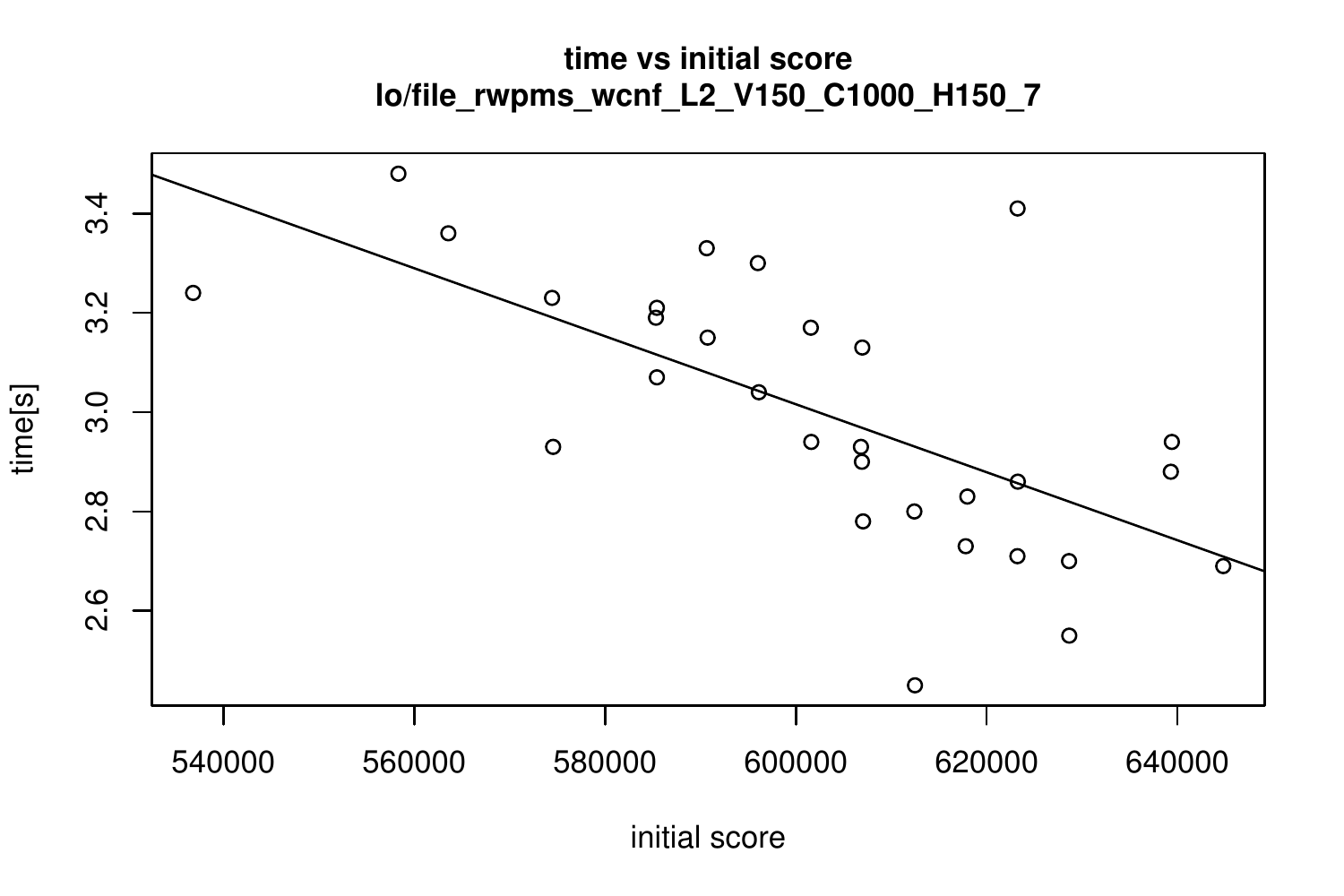}
    \label{fig_lo/file_rwpms_wcnf_L2_V150_C1000_H150_7/file_rwpms_wcnf_L2_V150_C1000_H150_7-time_vs_initial_score}
\end{figure}

\begin{figure}[H]
    \centering
    \includegraphics[height=3.5in]{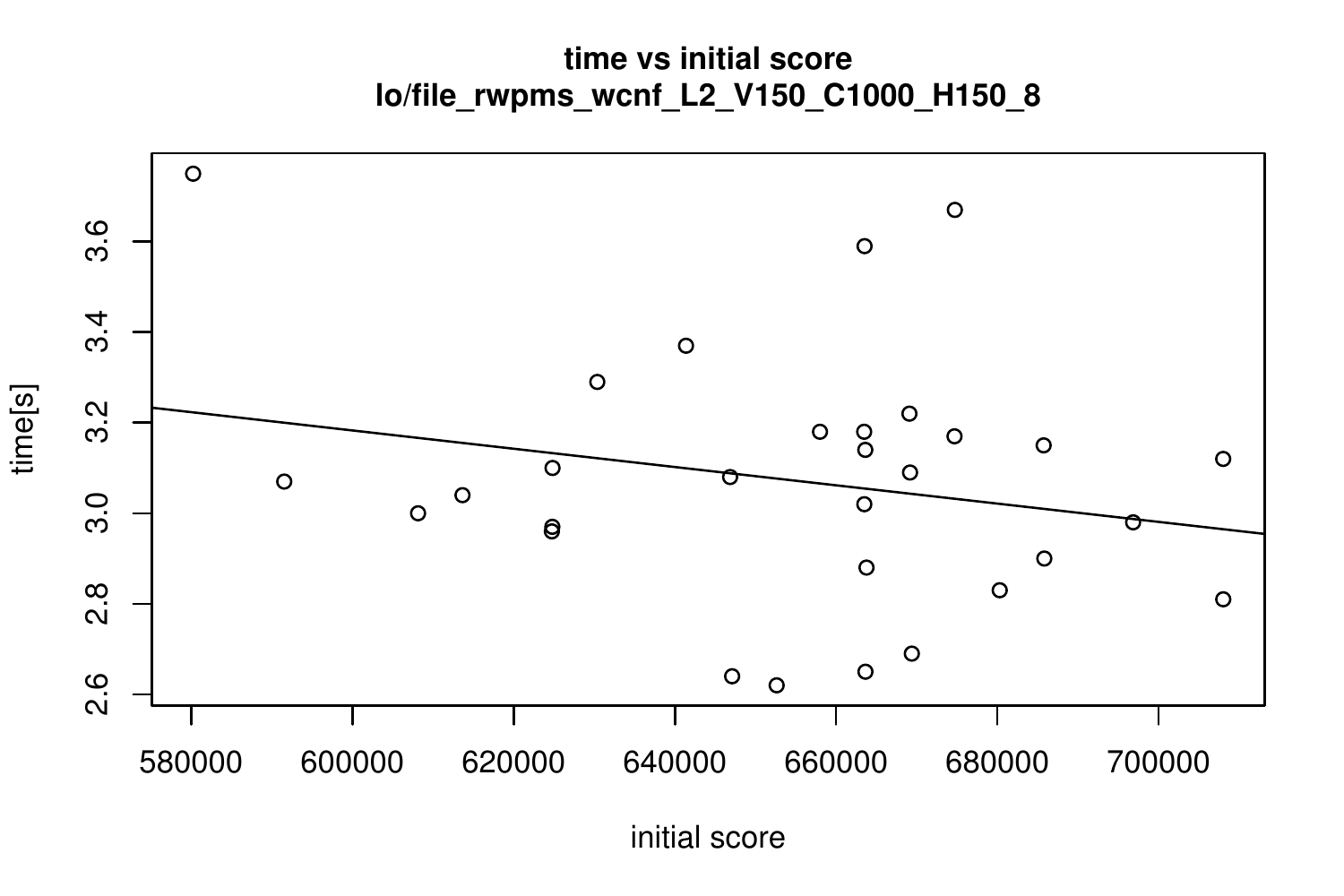}
    \label{fig_lo/file_rwpms_wcnf_L2_V150_C1000_H150_8/file_rwpms_wcnf_L2_V150_C1000_H150_8-time_vs_initial_score}
\end{figure}

\begin{figure}[H]
    \centering
    \includegraphics[height=3.5in]{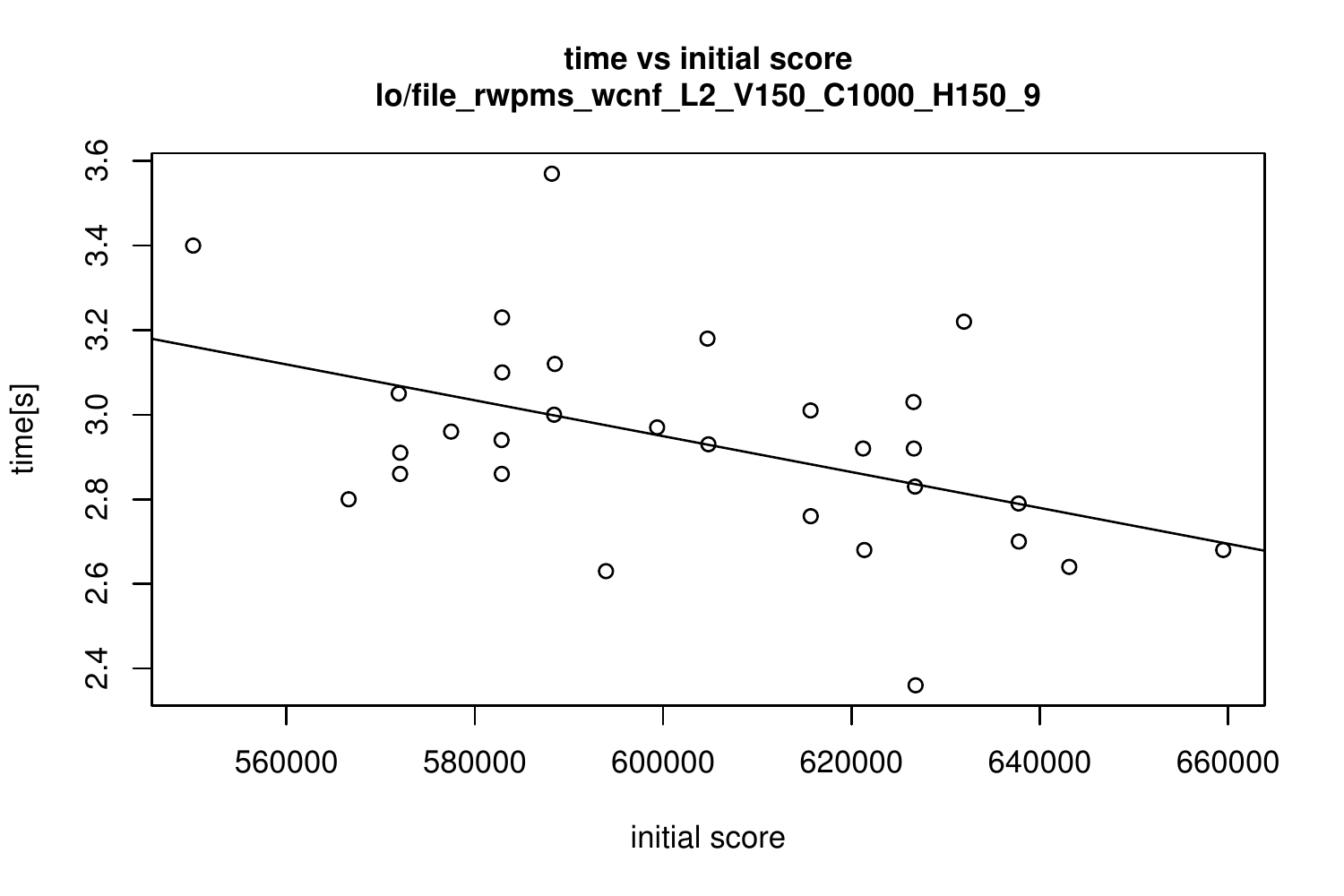}
    \label{fig_lo/file_rwpms_wcnf_L2_V150_C1000_H150_9/file_rwpms_wcnf_L2_V150_C1000_H150_9-time_vs_initial_score}
\end{figure}

\begin{figure}[H]
    \centering
    \includegraphics[height=3.5in]{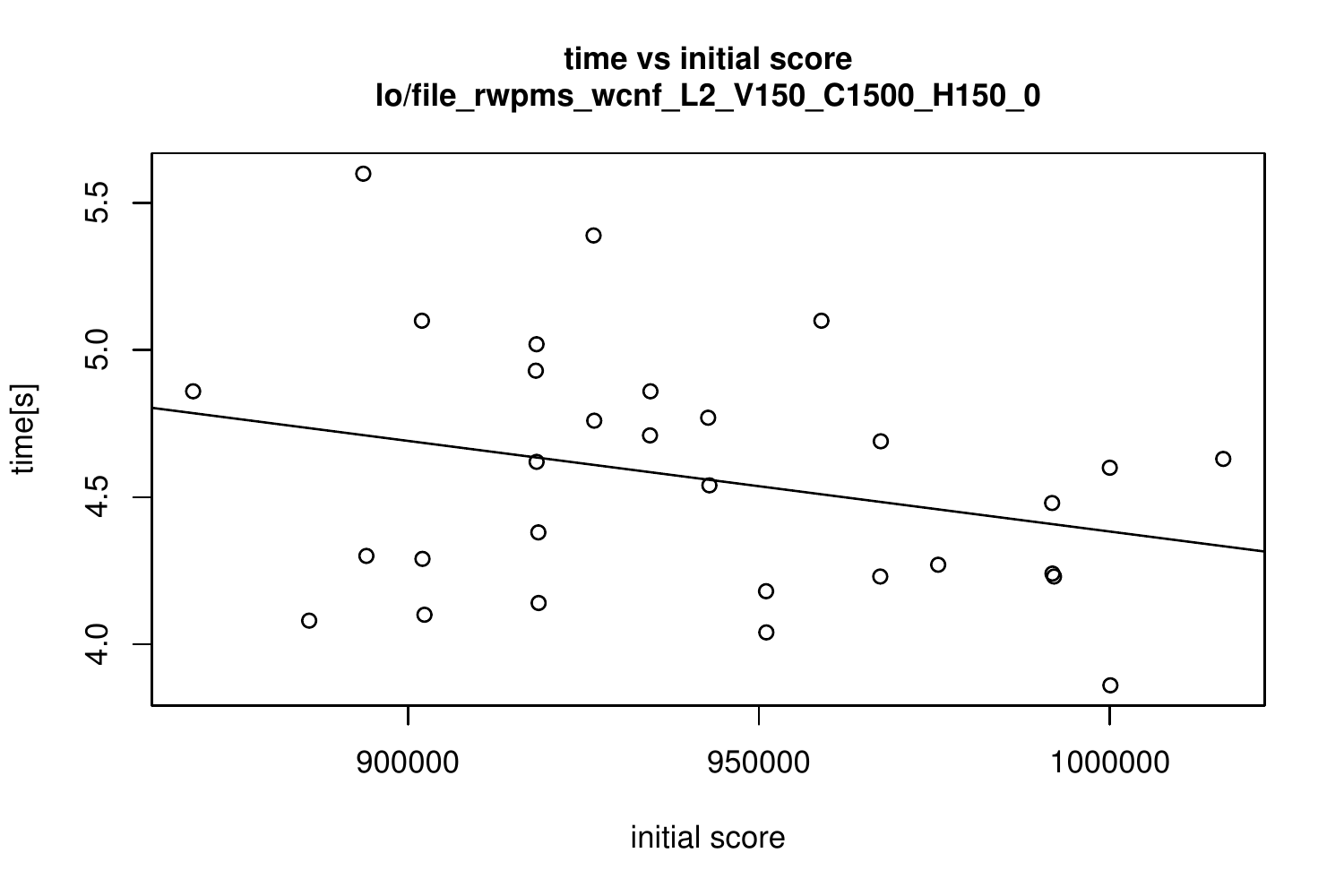}
    \label{fig_lo/file_rwpms_wcnf_L2_V150_C1500_H150_0/file_rwpms_wcnf_L2_V150_C1500_H150_0-time_vs_initial_score}
\end{figure}

\begin{figure}[H]
    \centering
    \includegraphics[height=3.5in]{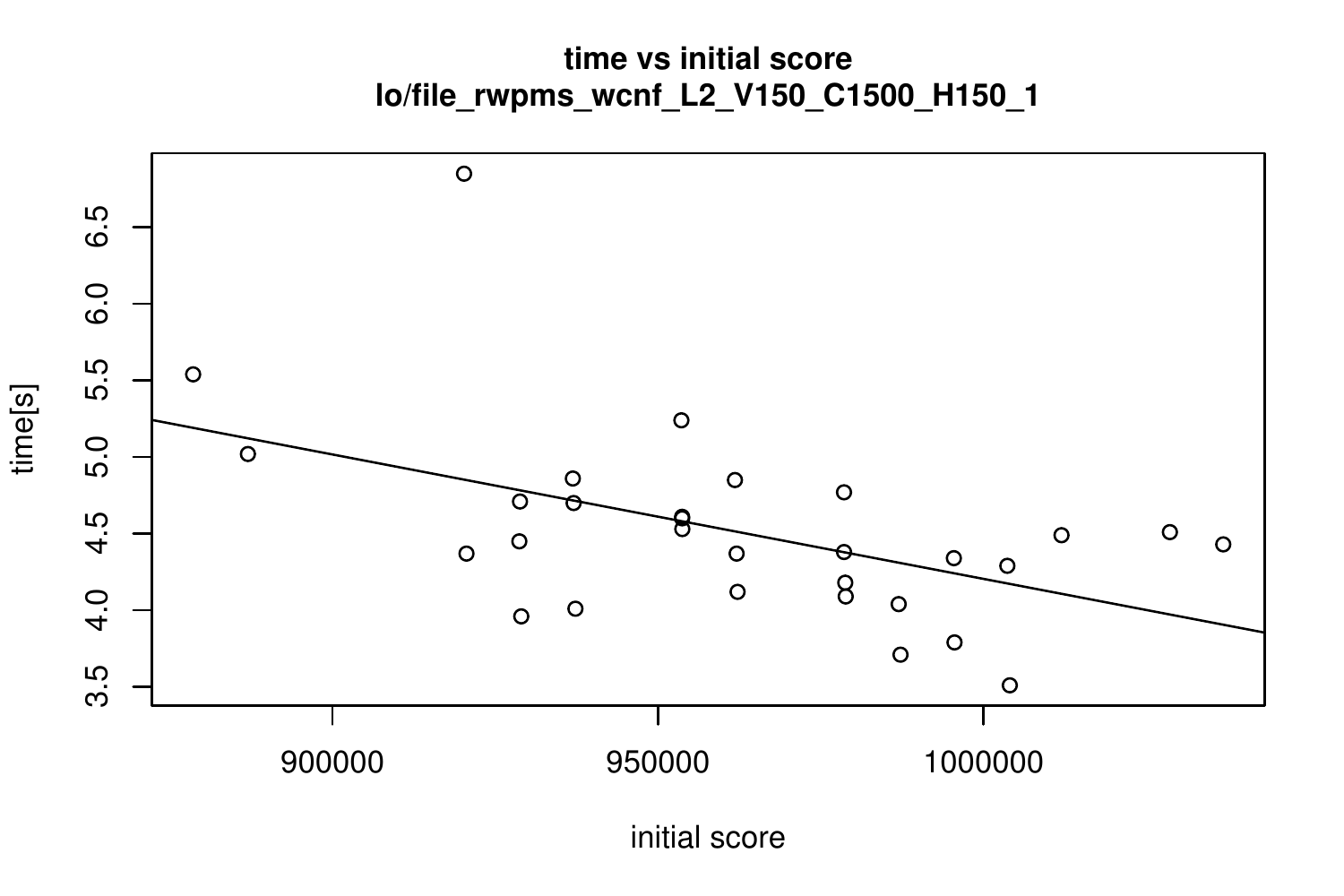}
    \label{fig_lo/file_rwpms_wcnf_L2_V150_C1500_H150_1/file_rwpms_wcnf_L2_V150_C1500_H150_1-time_vs_initial_score}
\end{figure}

\begin{figure}[H]
    \centering
    \includegraphics[height=3.5in]{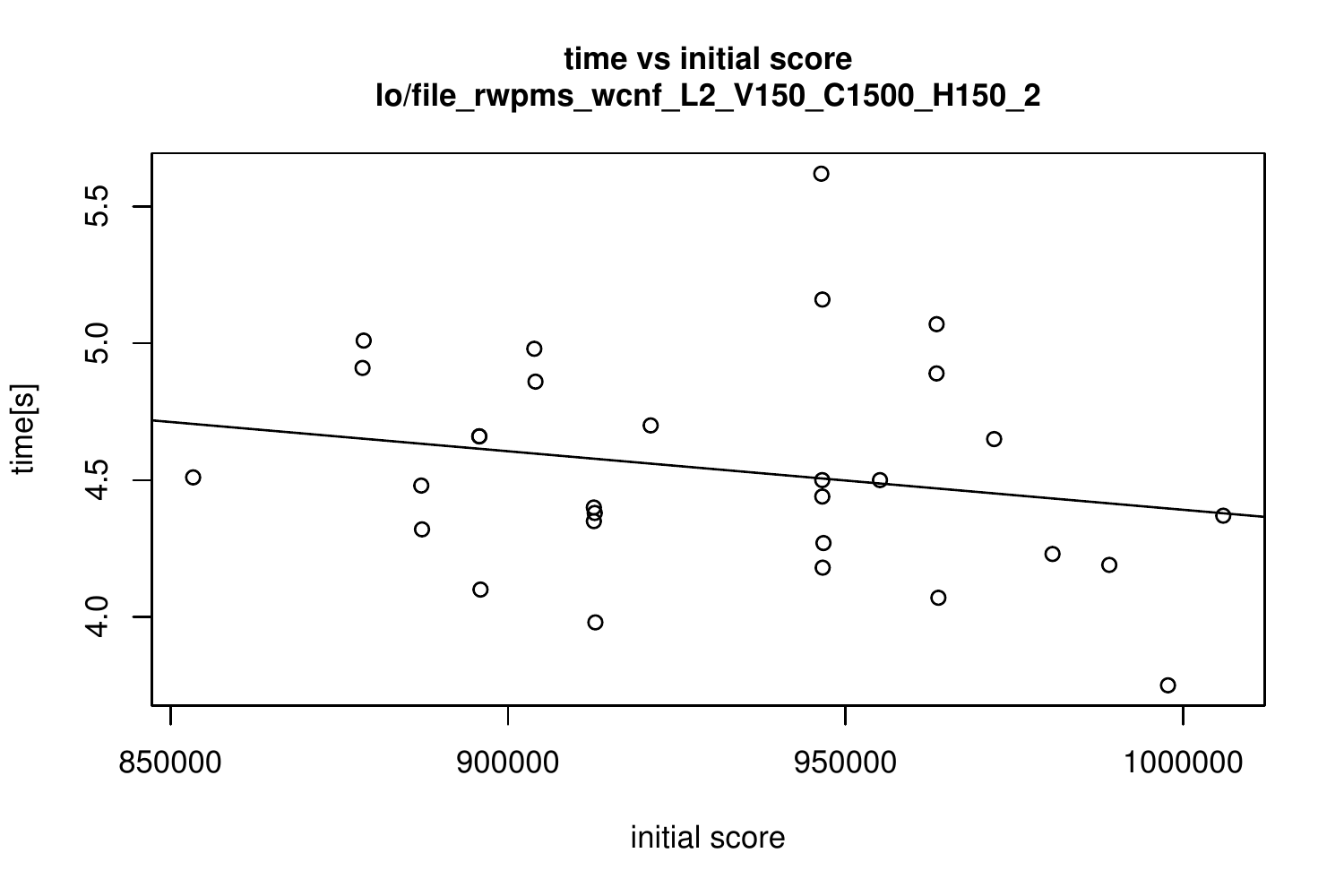}
    \label{fig_lo/file_rwpms_wcnf_L2_V150_C1500_H150_2/file_rwpms_wcnf_L2_V150_C1500_H150_2-time_vs_initial_score}
\end{figure}

\begin{figure}[H]
    \centering
    \includegraphics[height=3.5in]{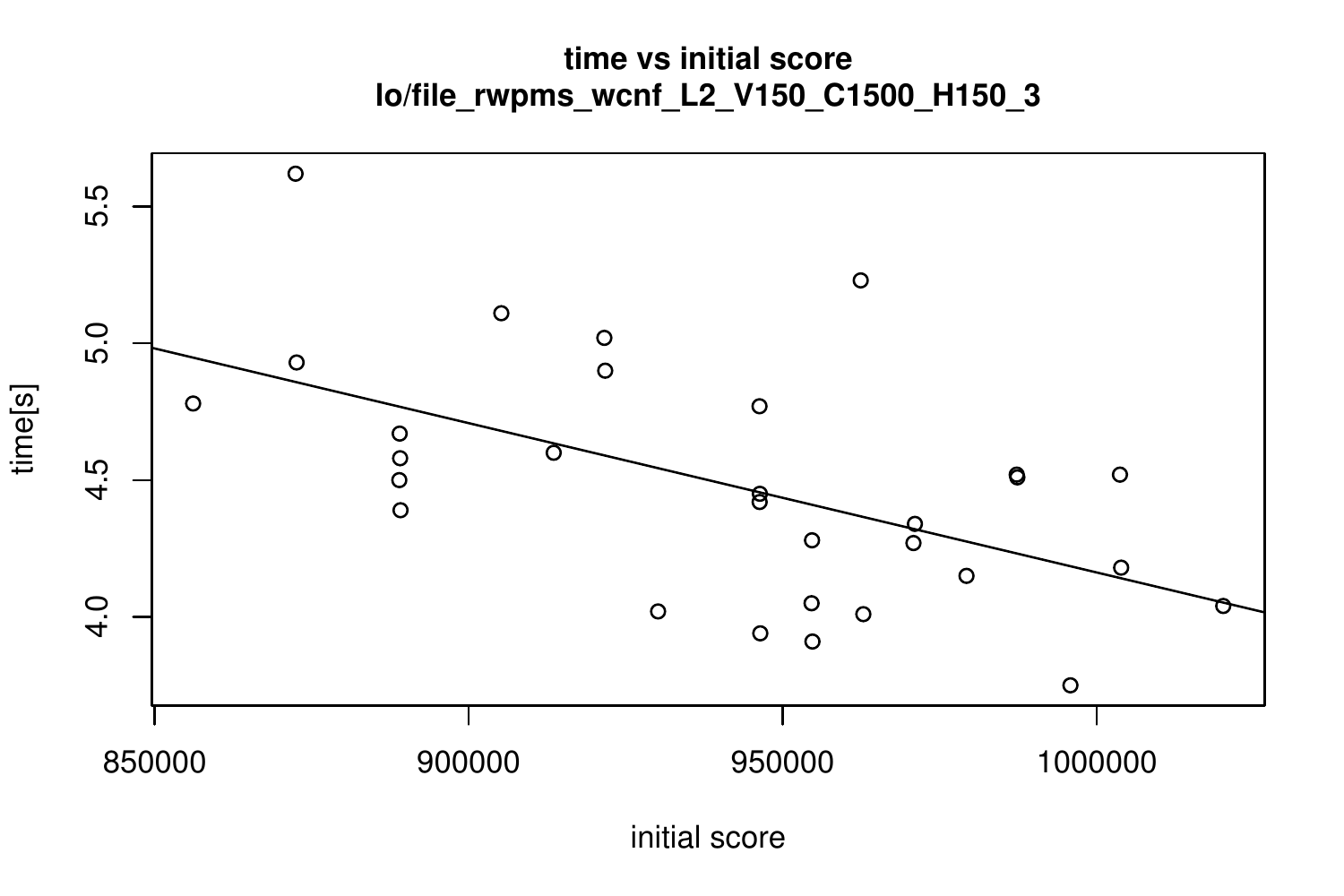}
    \label{fig_lo/file_rwpms_wcnf_L2_V150_C1500_H150_3/file_rwpms_wcnf_L2_V150_C1500_H150_3-time_vs_initial_score}
\end{figure}

\begin{figure}[H]
    \centering
    \includegraphics[height=3.5in]{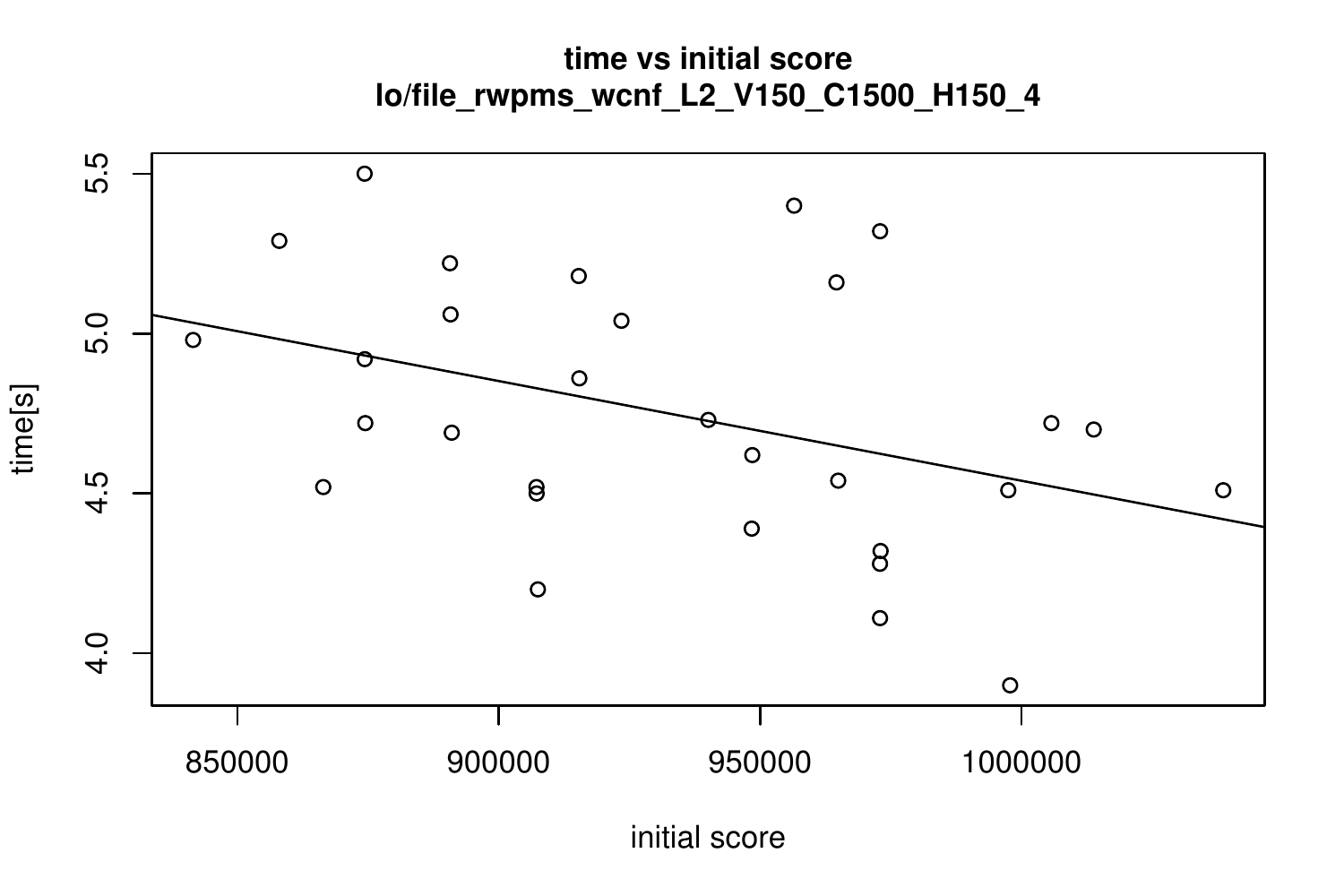}
    \label{fig_lo/file_rwpms_wcnf_L2_V150_C1500_H150_4/file_rwpms_wcnf_L2_V150_C1500_H150_4-time_vs_initial_score}
\end{figure}

\begin{figure}[H]
    \centering
    \includegraphics[height=3.5in]{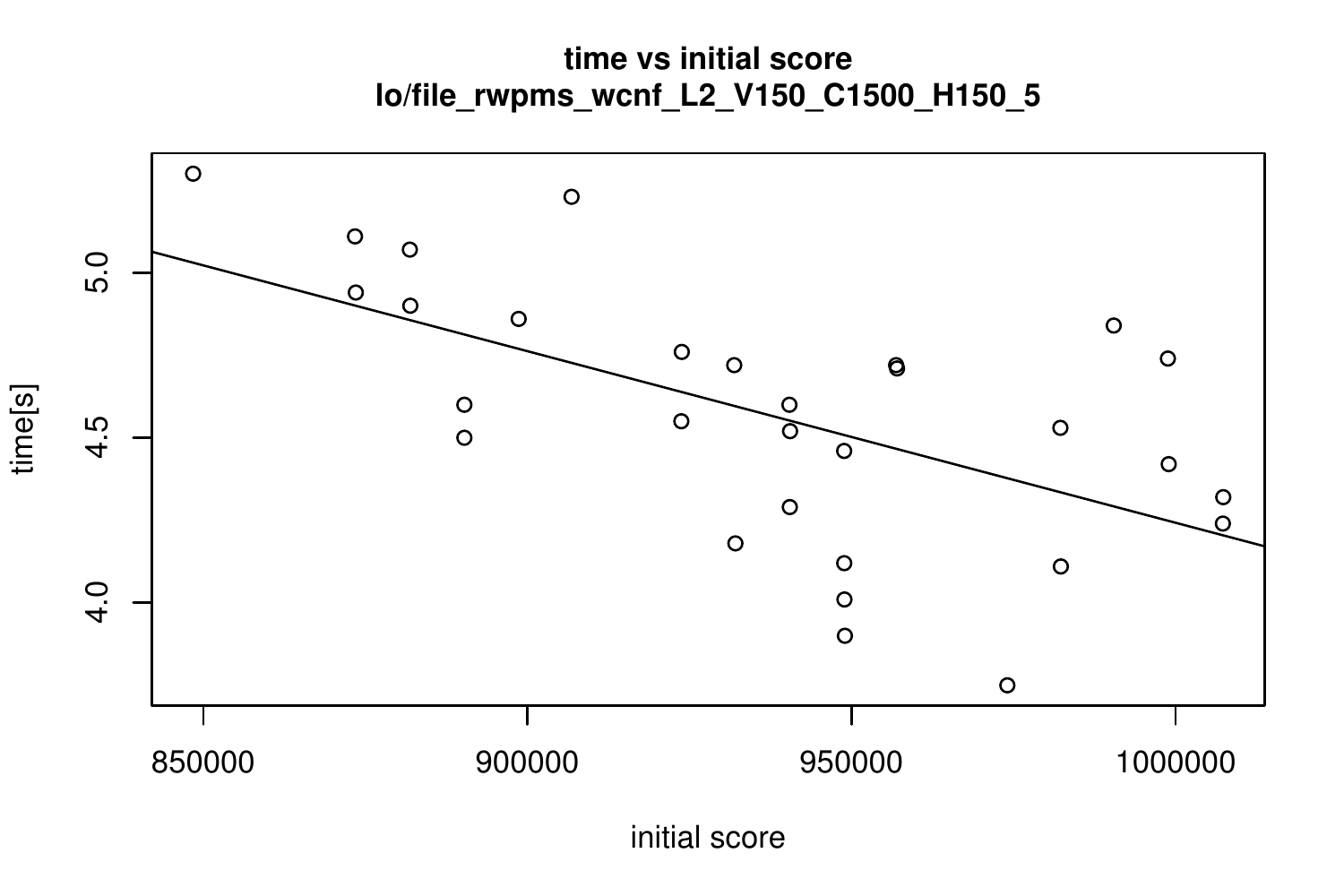}
    \label{fig_lo/file_rwpms_wcnf_L2_V150_C1500_H150_5/file_rwpms_wcnf_L2_V150_C1500_H150_5-time_vs_initial_score}
\end{figure}

\begin{figure}[H]
    \centering
    \includegraphics[height=3.5in]{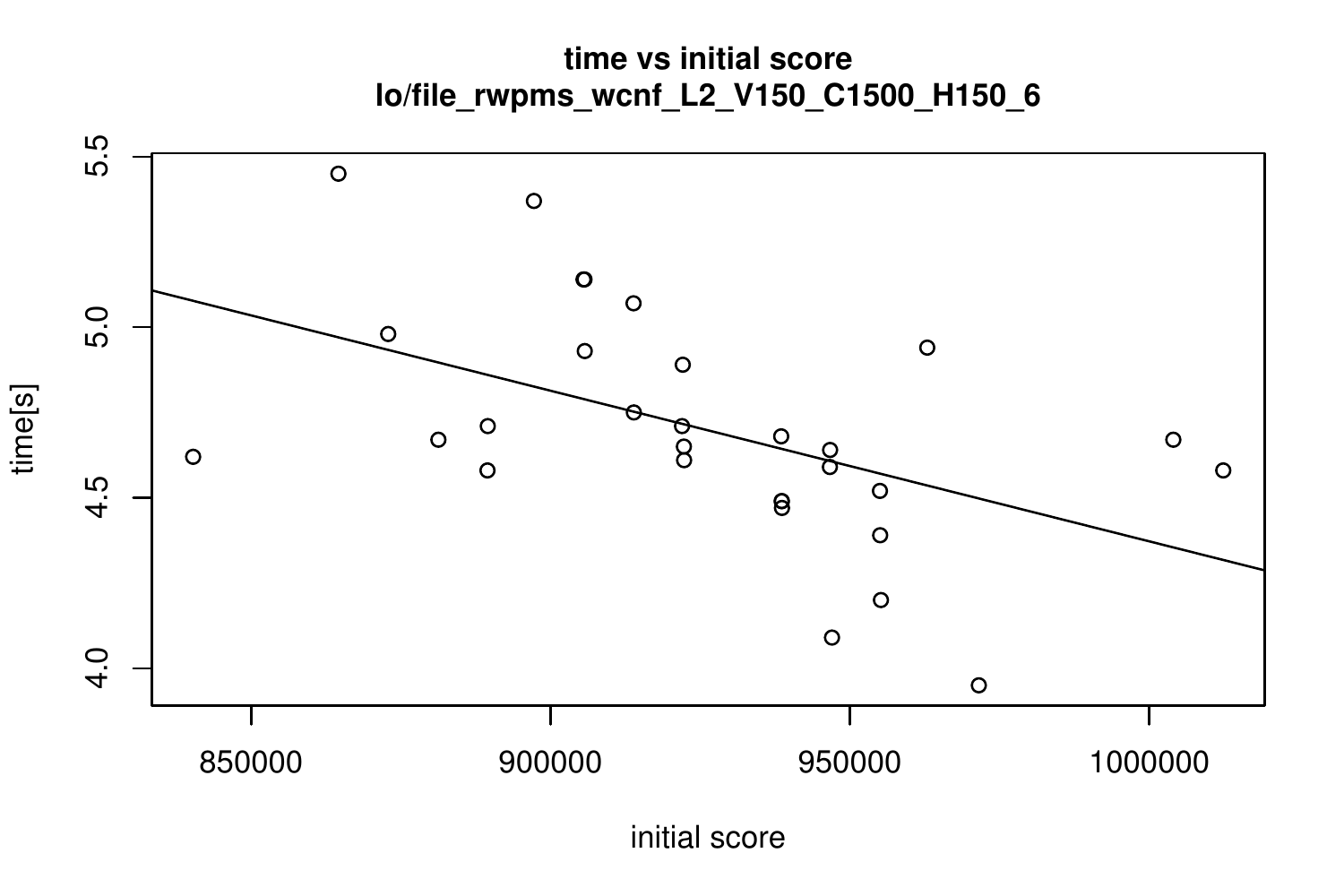}
    \label{fig_lo/file_rwpms_wcnf_L2_V150_C1500_H150_6/file_rwpms_wcnf_L2_V150_C1500_H150_6-time_vs_initial_score}
\end{figure}

\begin{figure}[H]
    \centering
    \includegraphics[height=3.5in]{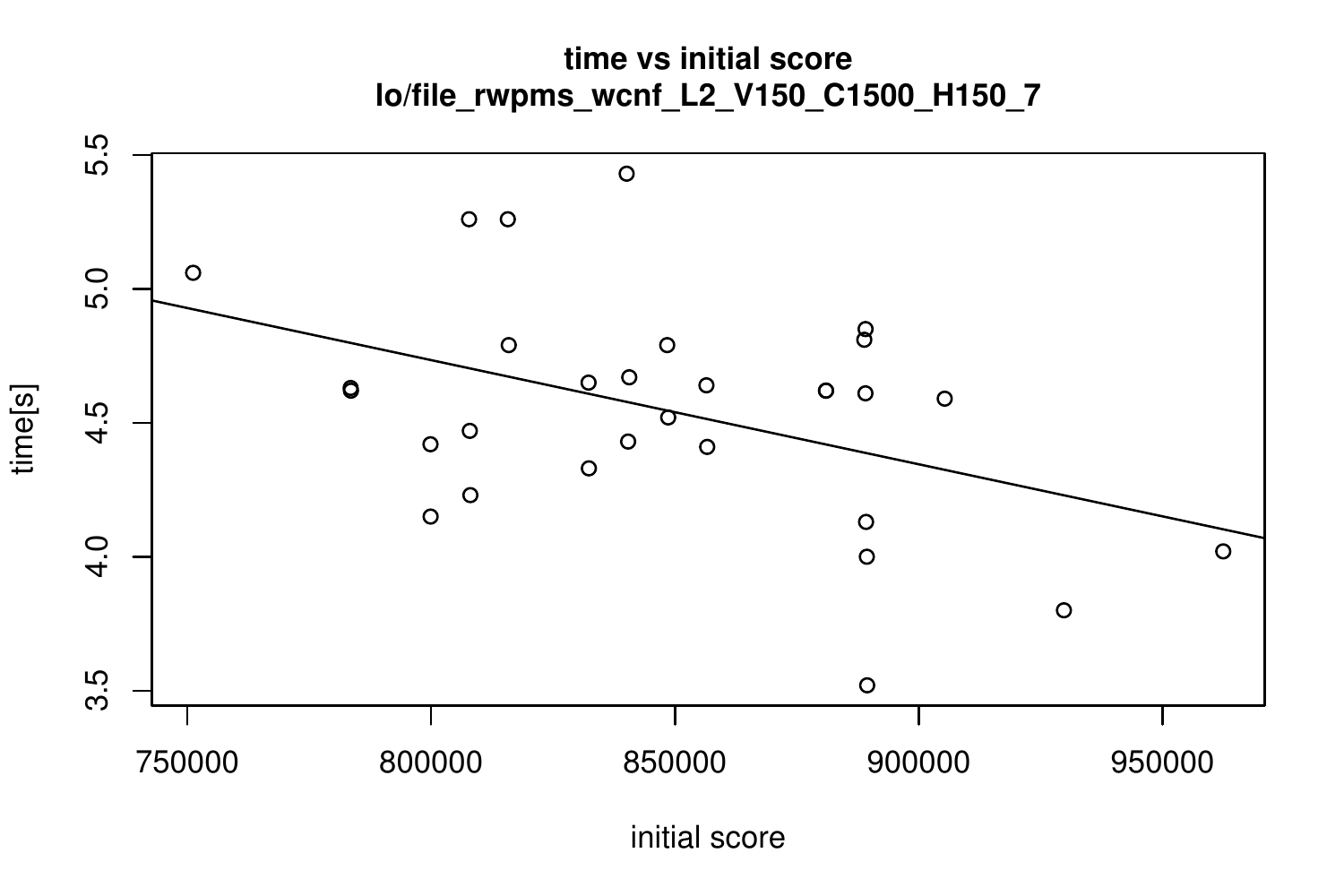}
    \label{fig_lo/file_rwpms_wcnf_L2_V150_C1500_H150_7/file_rwpms_wcnf_L2_V150_C1500_H150_7-time_vs_initial_score}
\end{figure}

\begin{figure}[H]
    \centering
    \includegraphics[height=3.5in]{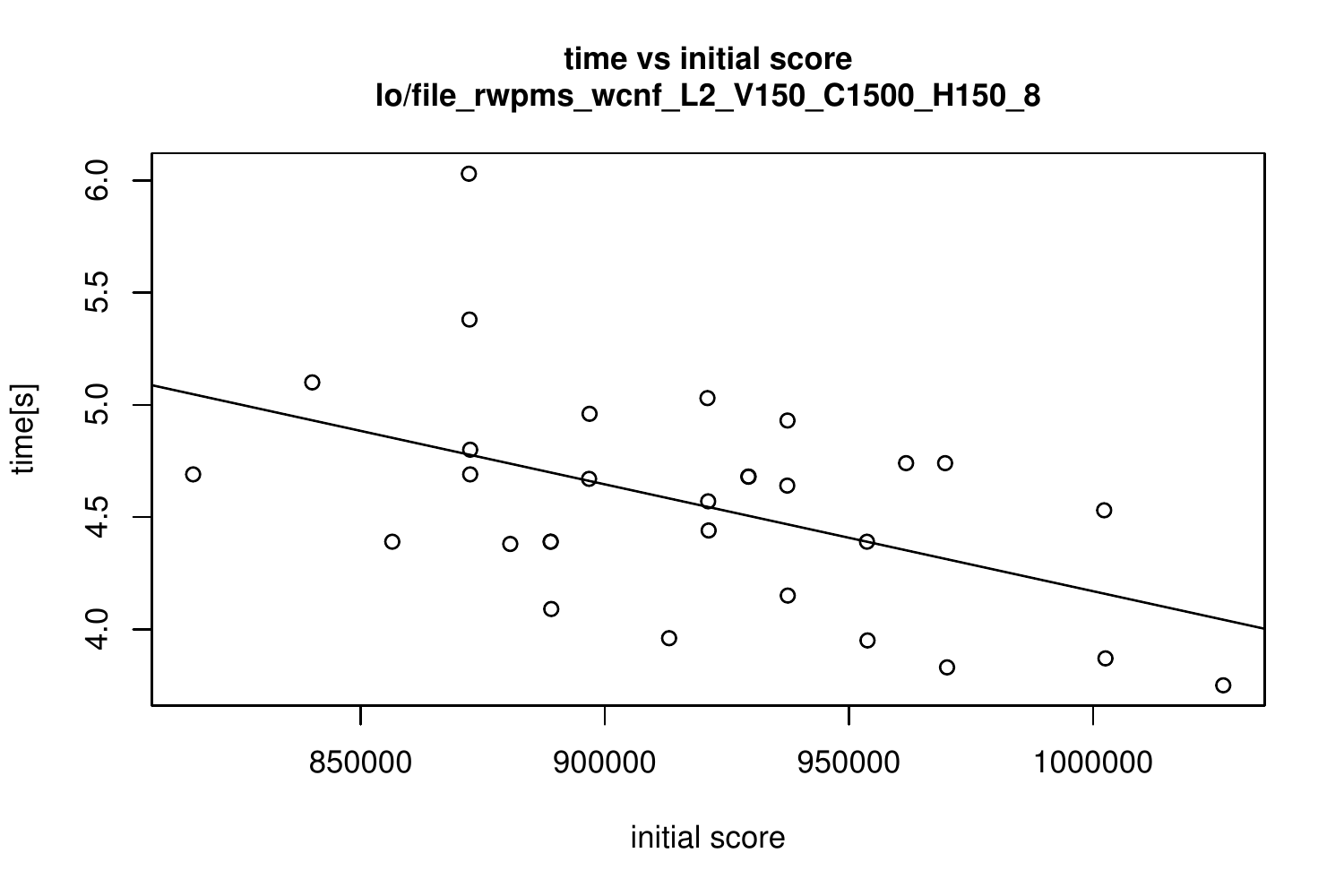}
    \label{fig_lo/file_rwpms_wcnf_L2_V150_C1500_H150_8/file_rwpms_wcnf_L2_V150_C1500_H150_8-time_vs_initial_score}
\end{figure}

\begin{figure}[H]
    \centering
    \includegraphics[height=3.5in]{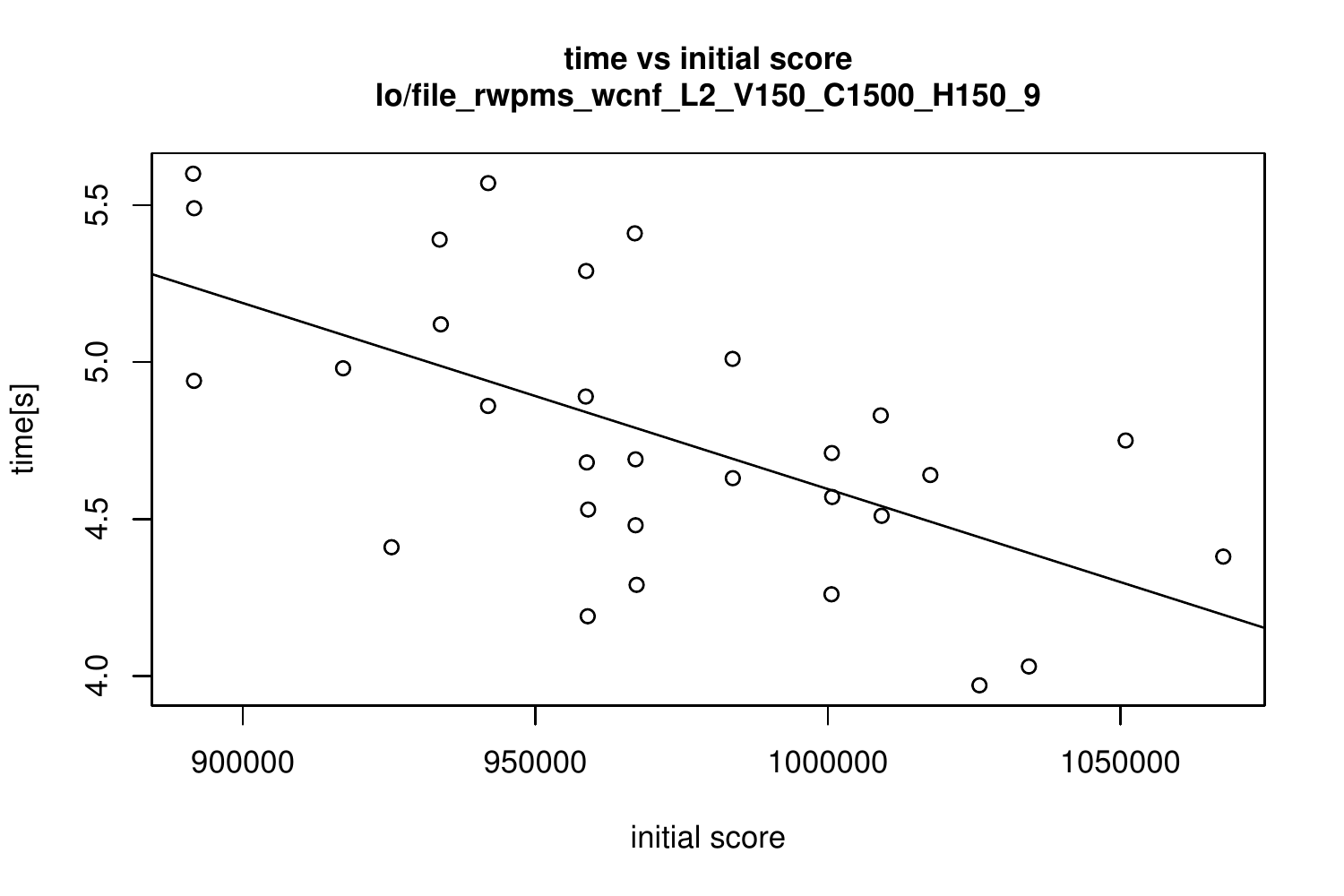}
    \label{fig_lo/file_rwpms_wcnf_L2_V150_C1500_H150_9/file_rwpms_wcnf_L2_V150_C1500_H150_9-time_vs_initial_score}
\end{figure}

\begin{figure}[H]
    \centering
    \includegraphics[height=3.5in]{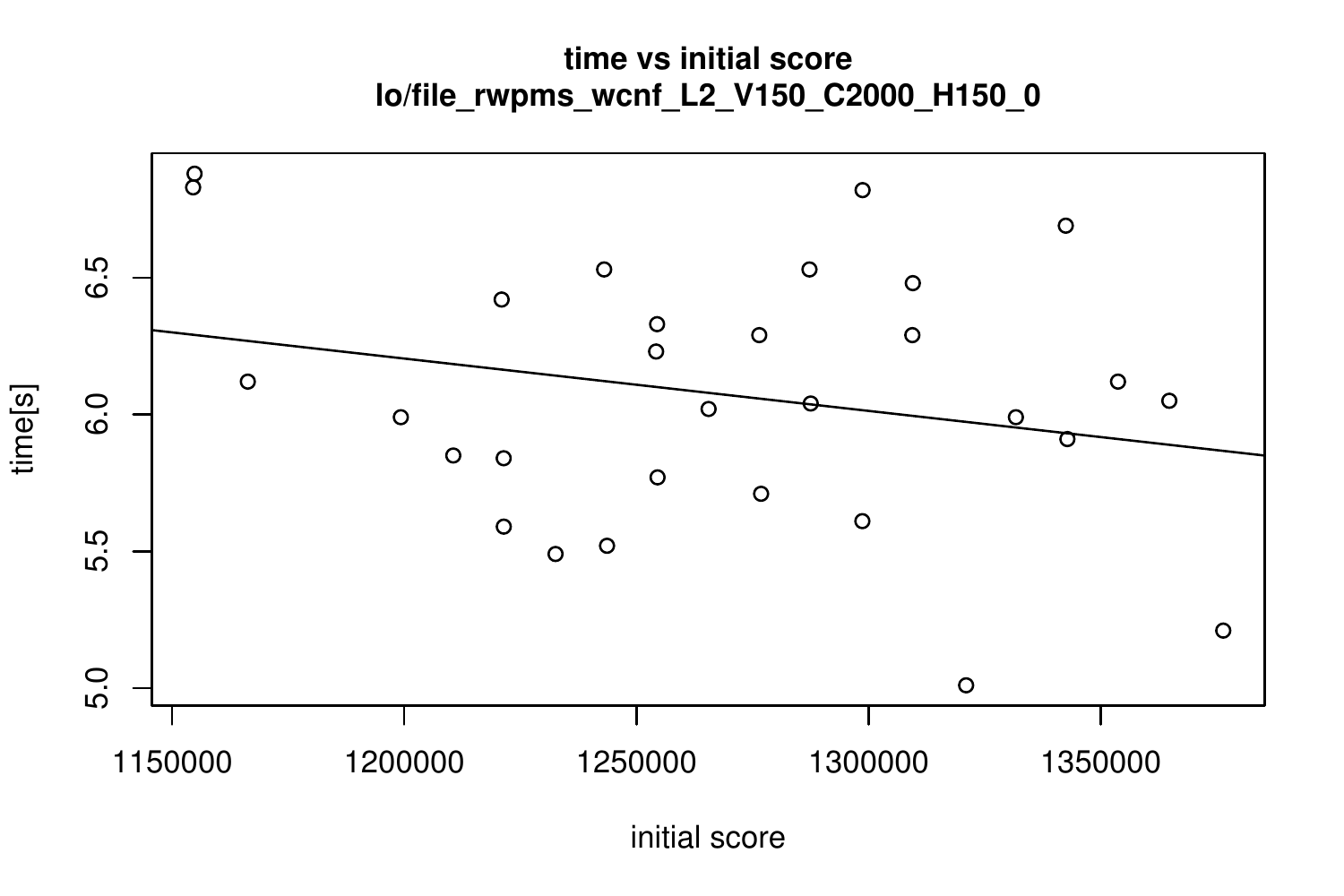}
    \label{fig_lo/file_rwpms_wcnf_L2_V150_C2000_H150_0/file_rwpms_wcnf_L2_V150_C2000_H150_0-time_vs_initial_score}
\end{figure}

\begin{figure}[H]
    \centering
    \includegraphics[height=3.5in]{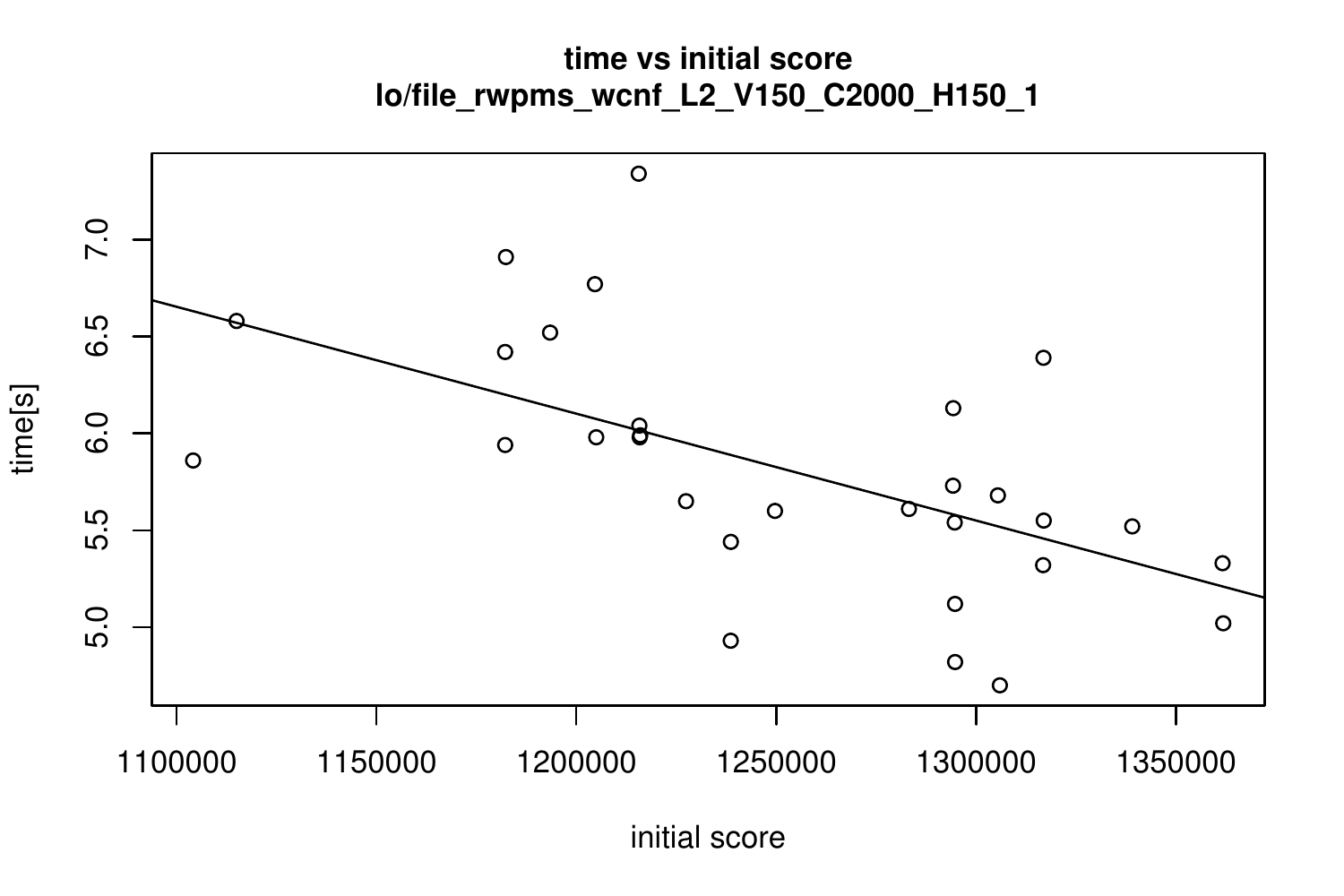}
    \label{fig_lo/file_rwpms_wcnf_L2_V150_C2000_H150_1/file_rwpms_wcnf_L2_V150_C2000_H150_1-time_vs_initial_score}
\end{figure}

\begin{figure}[H]
    \centering
    \includegraphics[height=3.5in]{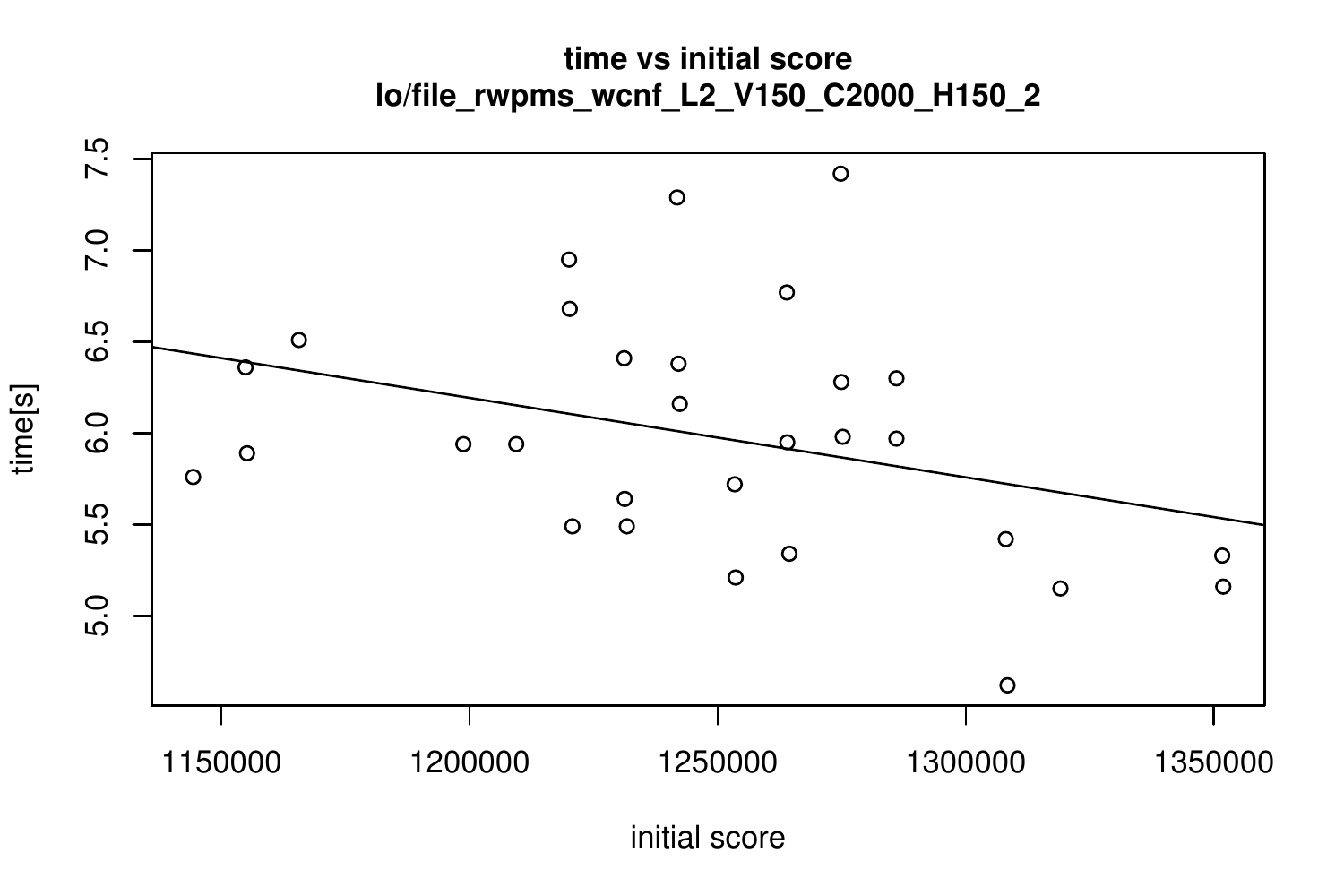}
    \label{fig_lo/file_rwpms_wcnf_L2_V150_C2000_H150_2/file_rwpms_wcnf_L2_V150_C2000_H150_2-time_vs_initial_score}
\end{figure}

\begin{figure}[H]
    \centering
    \includegraphics[height=3.5in]{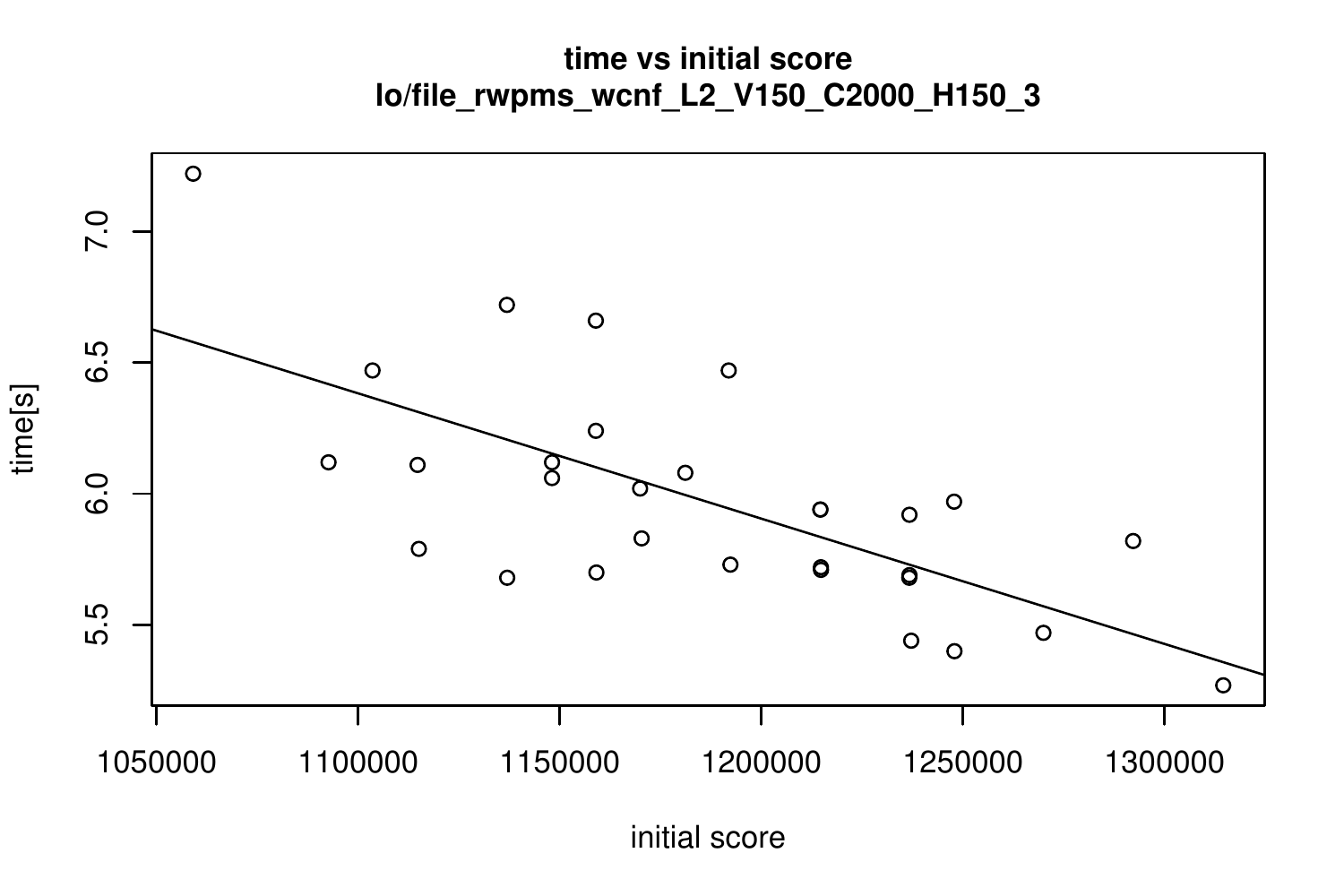}
    \label{fig_lo/file_rwpms_wcnf_L2_V150_C2000_H150_3/file_rwpms_wcnf_L2_V150_C2000_H150_3-time_vs_initial_score}
\end{figure}

\begin{figure}[H]
    \centering
    \includegraphics[height=3.5in]{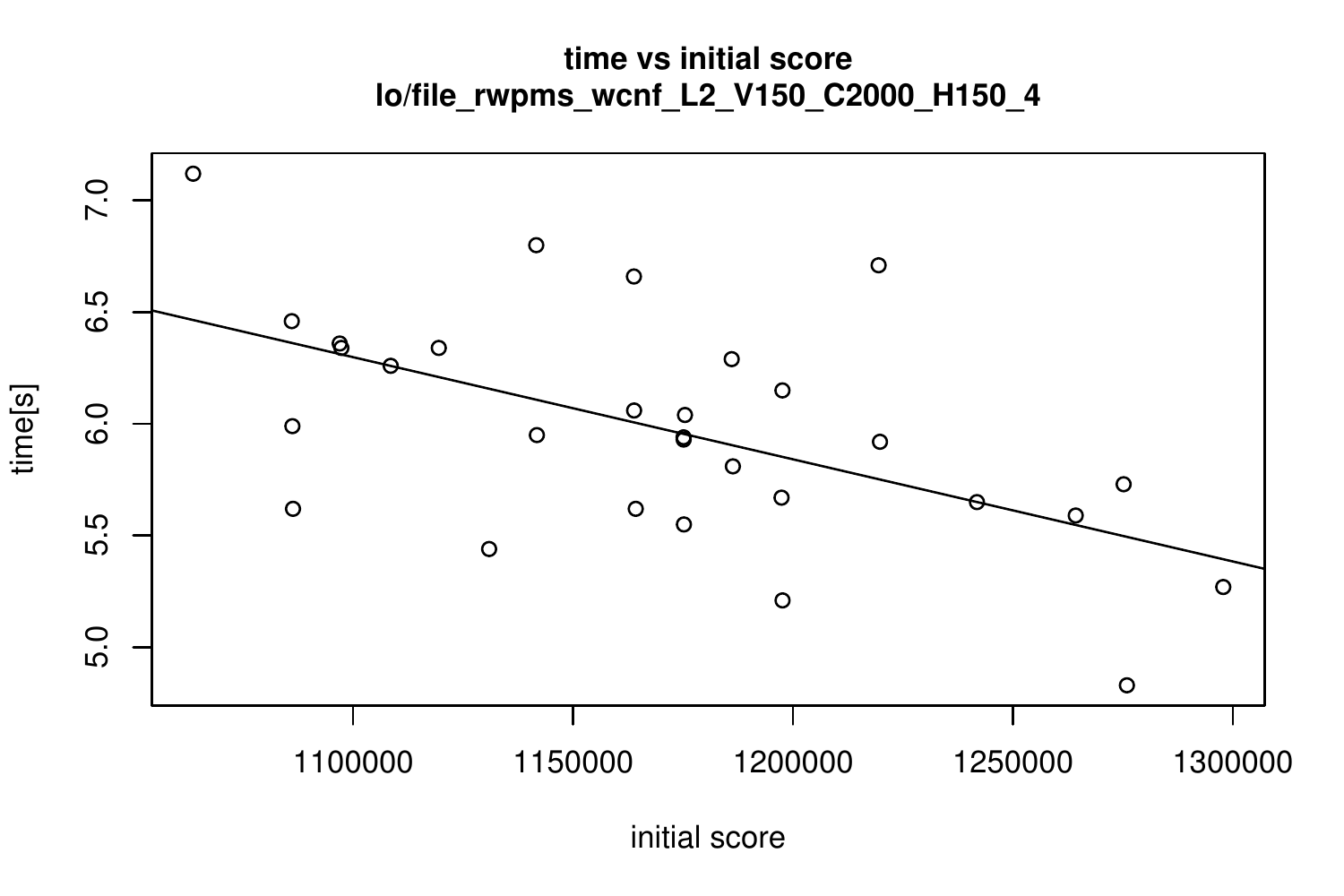}
    \label{fig_lo/file_rwpms_wcnf_L2_V150_C2000_H150_4/file_rwpms_wcnf_L2_V150_C2000_H150_4-time_vs_initial_score}
\end{figure}

\begin{figure}[H]
    \centering
    \includegraphics[height=3.5in]{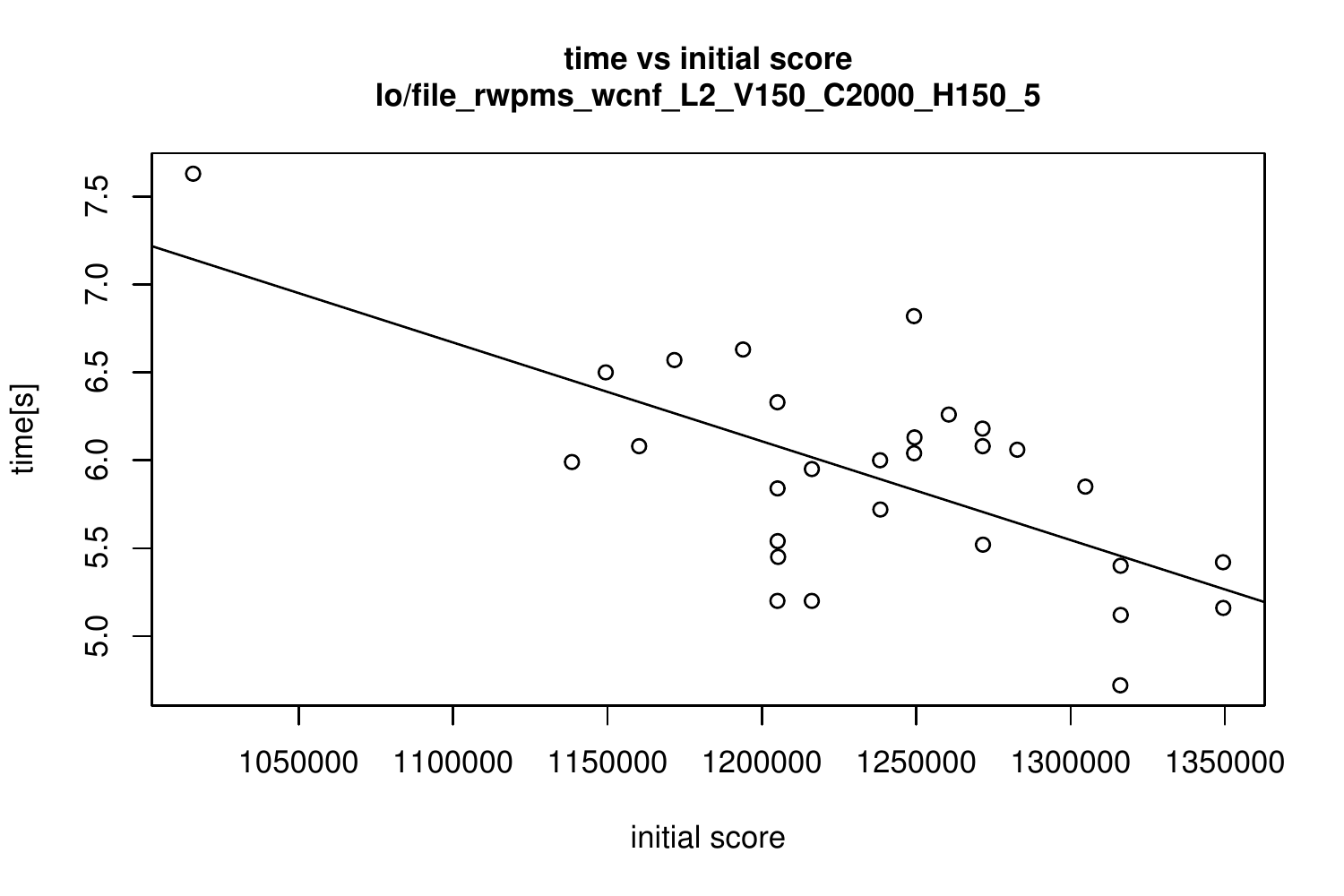}
    \label{fig_lo/file_rwpms_wcnf_L2_V150_C2000_H150_5/file_rwpms_wcnf_L2_V150_C2000_H150_5-time_vs_initial_score}
\end{figure}

\begin{figure}[H]
    \centering
    \includegraphics[height=3.5in]{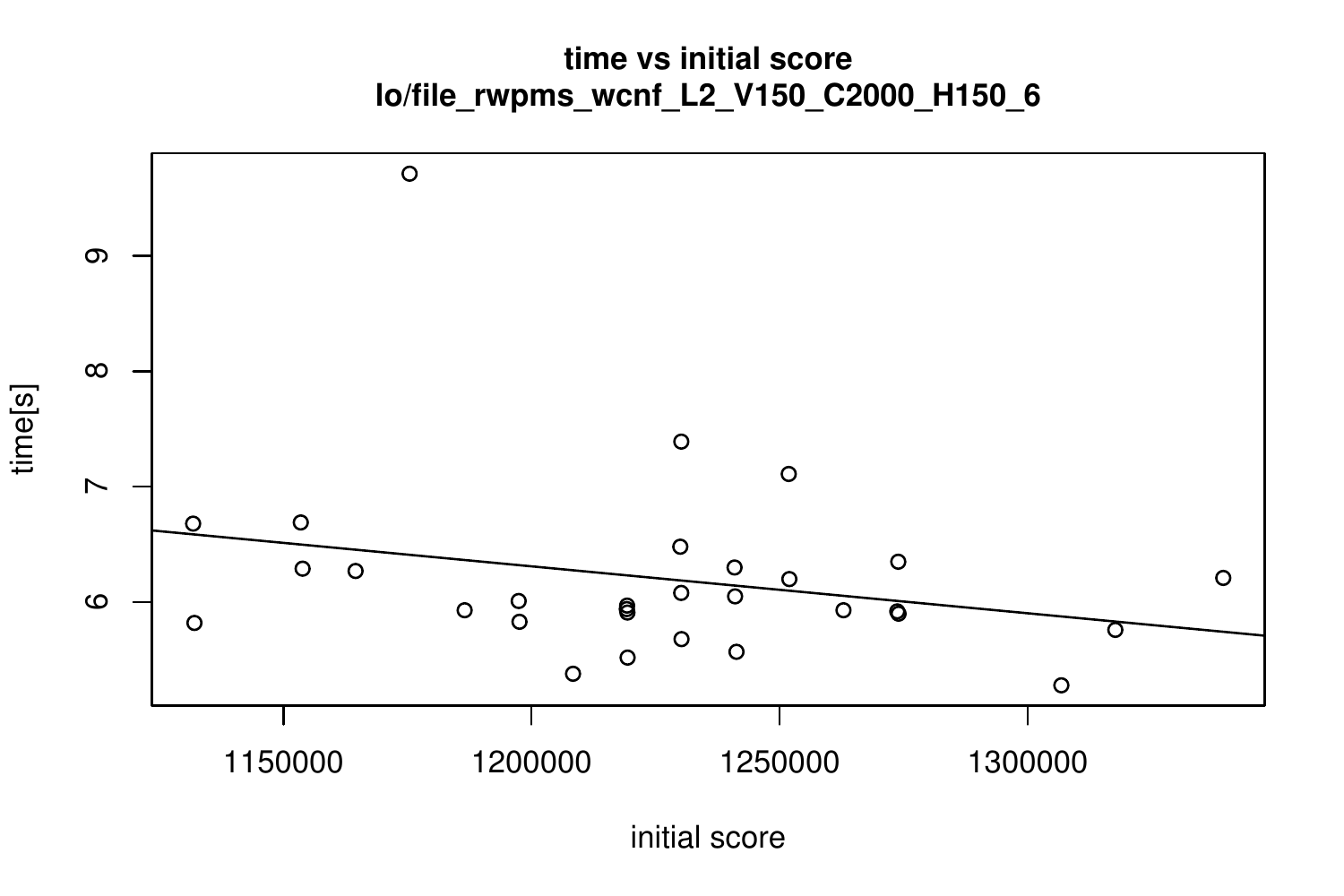}
    \label{fig_lo/file_rwpms_wcnf_L2_V150_C2000_H150_6/file_rwpms_wcnf_L2_V150_C2000_H150_6-time_vs_initial_score}
\end{figure}

\begin{figure}[H]
    \centering
    \includegraphics[height=3.5in]{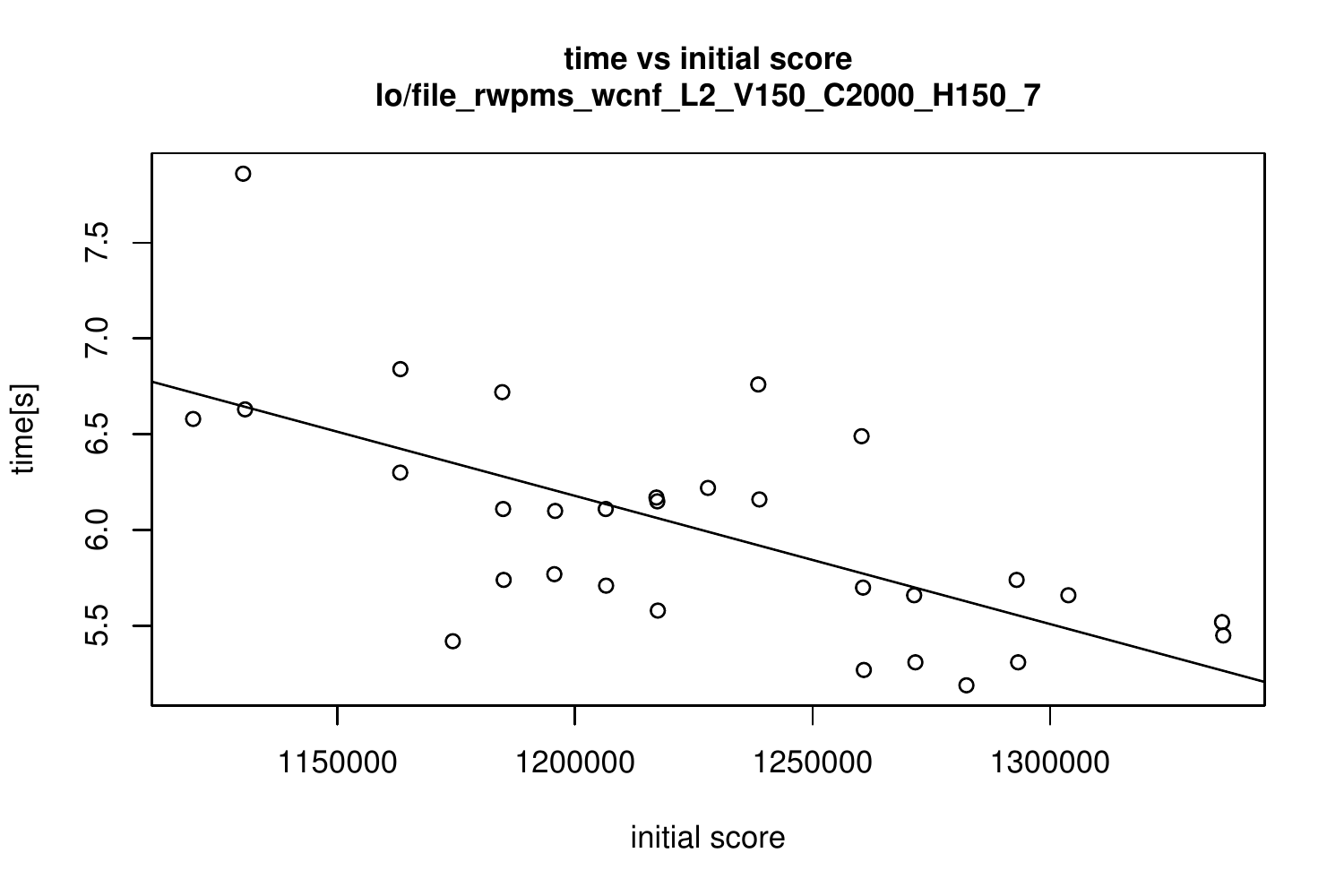}
    \label{fig_lo/file_rwpms_wcnf_L2_V150_C2000_H150_7/file_rwpms_wcnf_L2_V150_C2000_H150_7-time_vs_initial_score}
\end{figure}

\begin{figure}[H]
    \centering
    \includegraphics[height=3.5in]{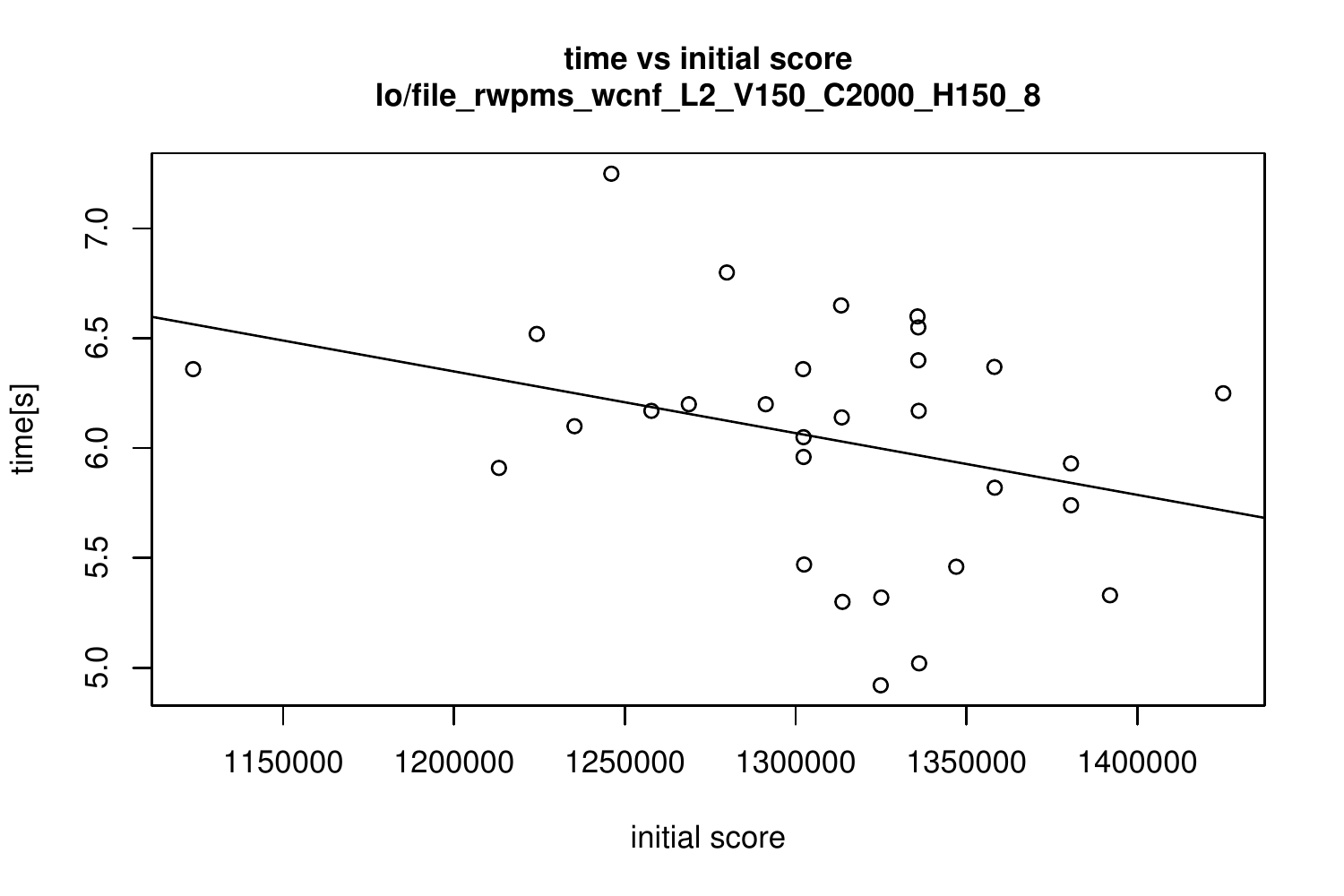}
    \label{fig_lo/file_rwpms_wcnf_L2_V150_C2000_H150_8/file_rwpms_wcnf_L2_V150_C2000_H150_8-time_vs_initial_score}
\end{figure}

\begin{figure}[H]
    \centering
    \includegraphics[height=3.5in]{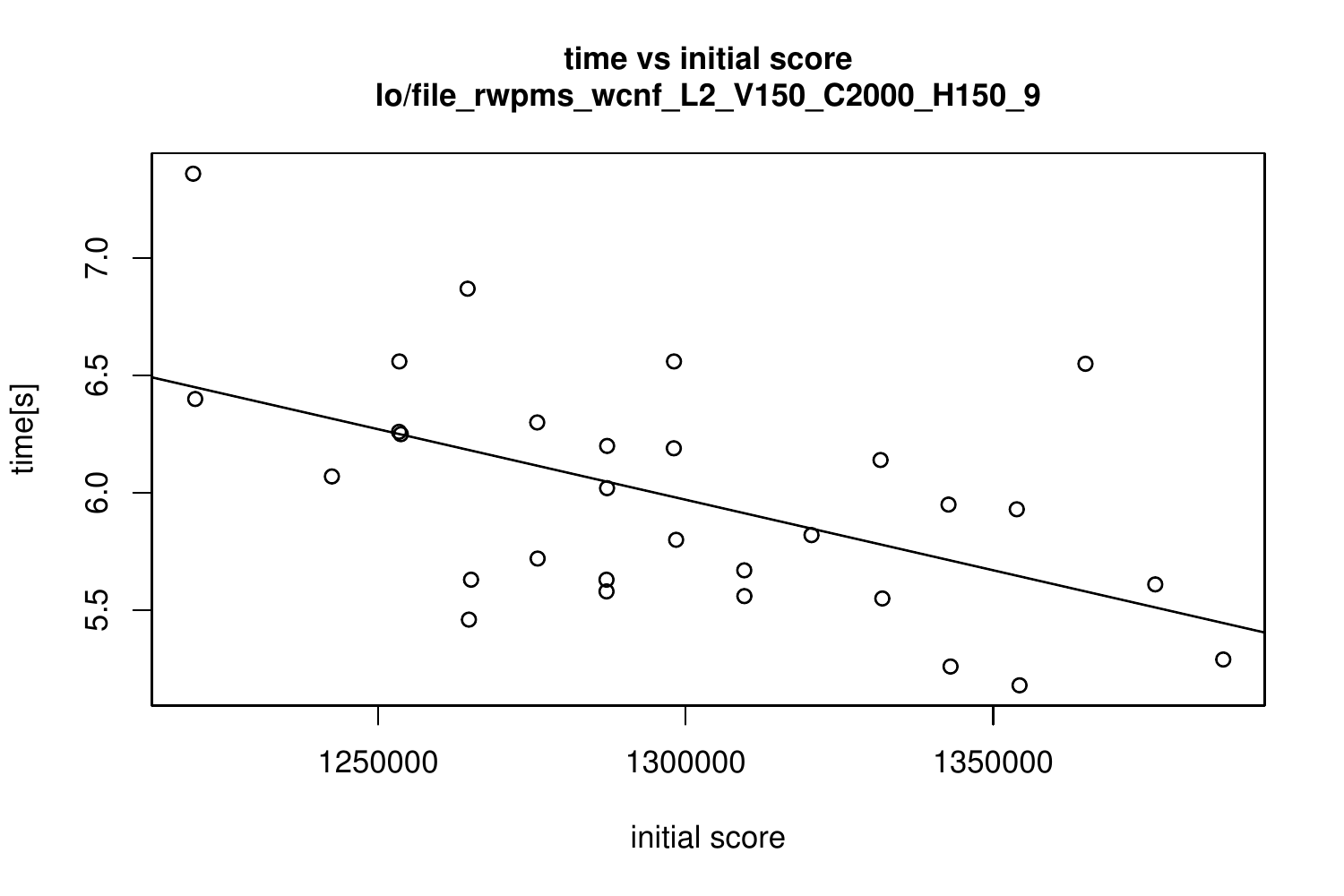}
    \label{fig_lo/file_rwpms_wcnf_L2_V150_C2000_H150_9/file_rwpms_wcnf_L2_V150_C2000_H150_9-time_vs_initial_score}
\end{figure}

\begin{figure}[H]
    \centering
    \includegraphics[height=3.5in]{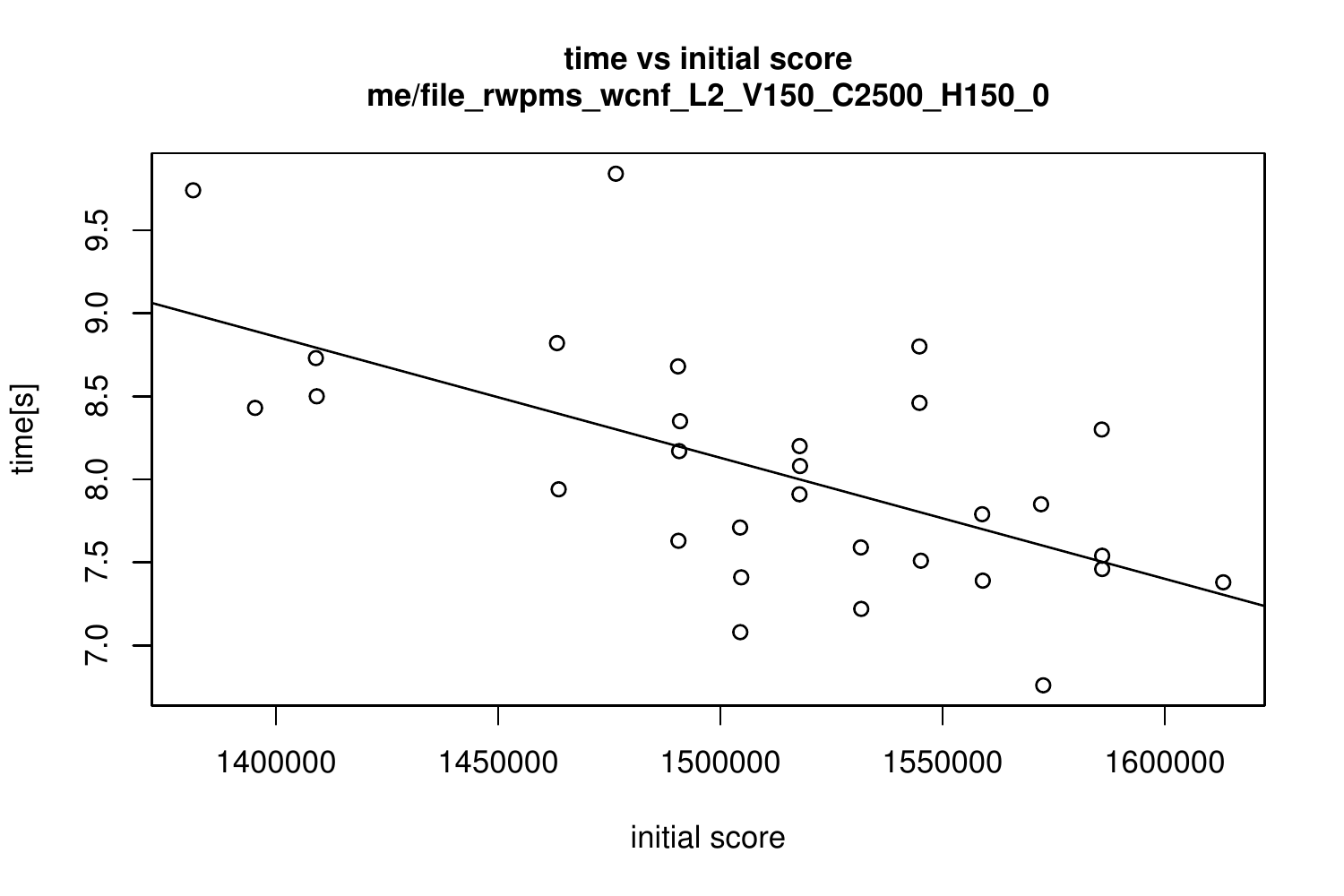}
    \label{fig_me/file_rwpms_wcnf_L2_V150_C2500_H150_0/file_rwpms_wcnf_L2_V150_C2500_H150_0-time_vs_initial_score}
\end{figure}

\begin{figure}[H]
    \centering
    \includegraphics[height=3.5in]{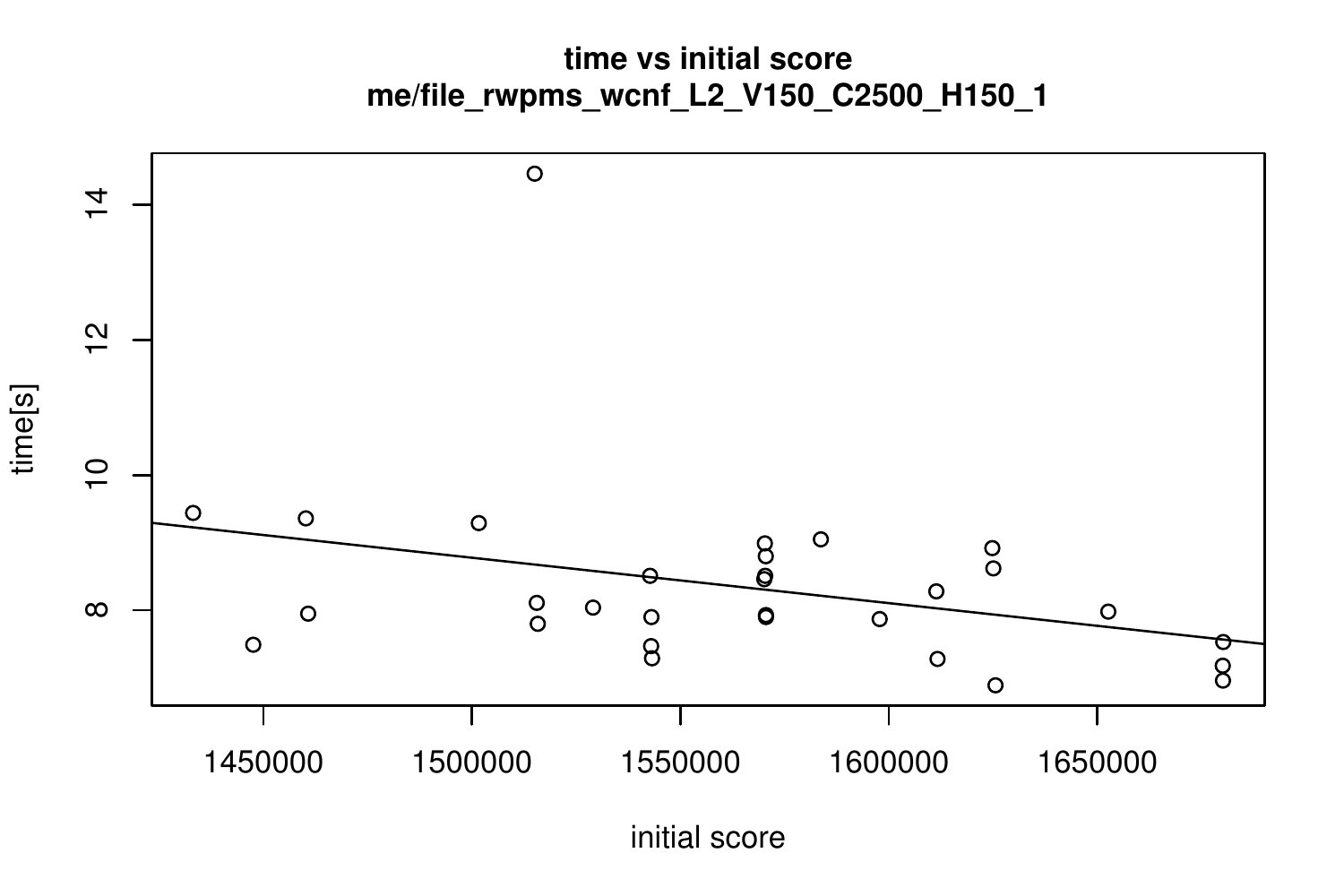}
    \label{fig_me/file_rwpms_wcnf_L2_V150_C2500_H150_1/file_rwpms_wcnf_L2_V150_C2500_H150_1-time_vs_initial_score}
\end{figure}

\begin{figure}[H]
    \centering
    \includegraphics[height=3.5in]{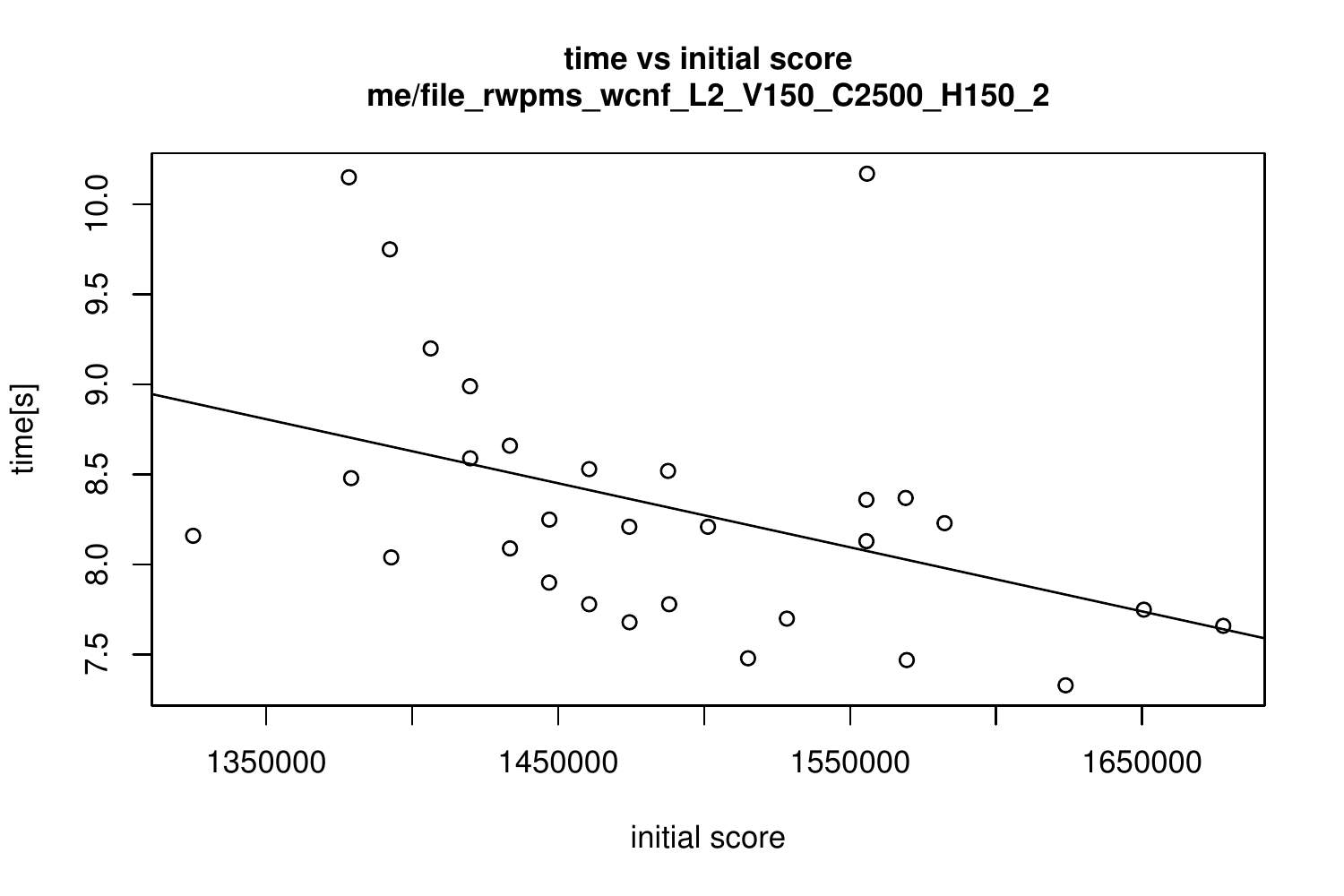}
    \label{fig_me/file_rwpms_wcnf_L2_V150_C2500_H150_2/file_rwpms_wcnf_L2_V150_C2500_H150_2-time_vs_initial_score}
\end{figure}

\begin{figure}[H]
    \centering
    \includegraphics[height=3.5in]{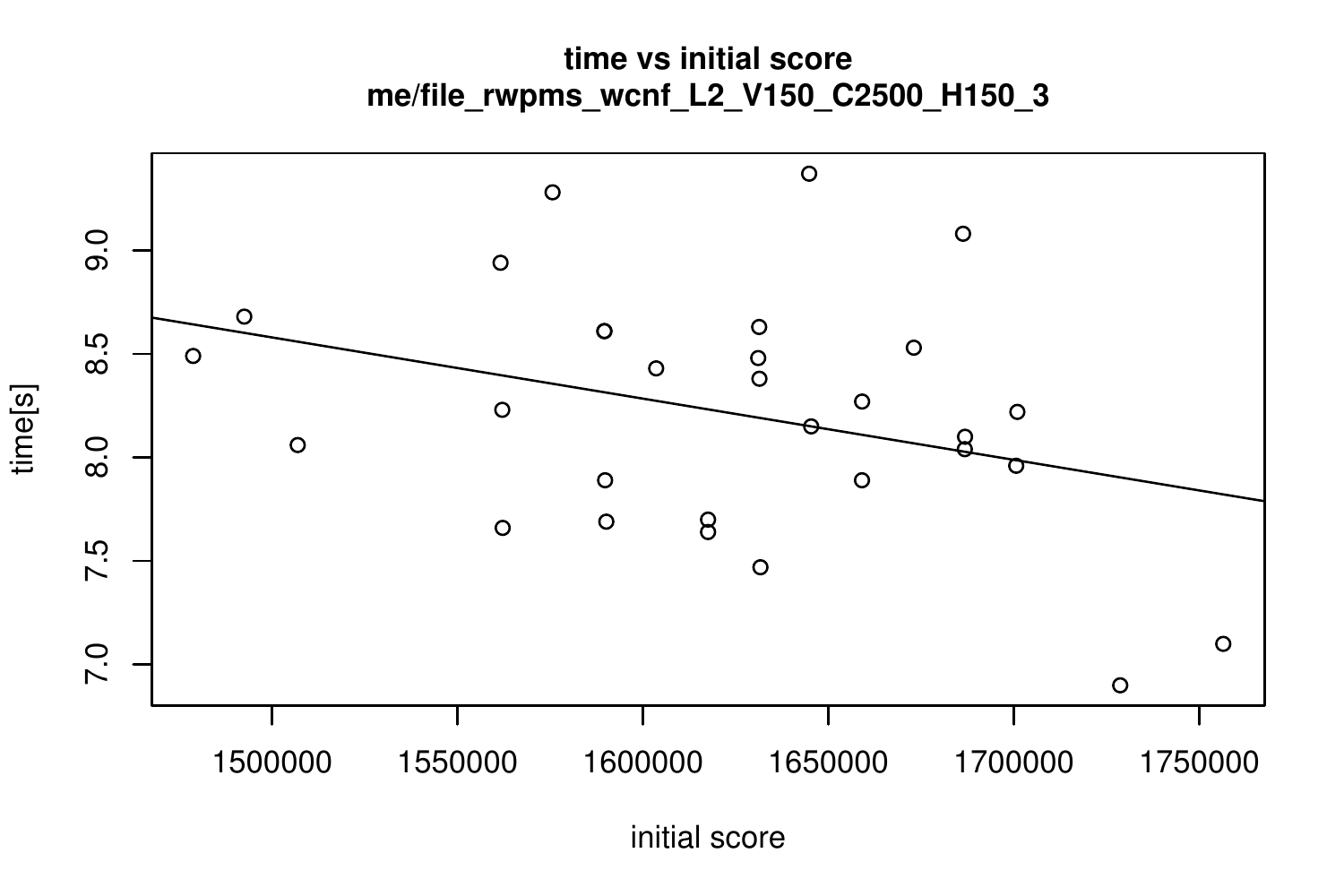}
    \label{fig_me/file_rwpms_wcnf_L2_V150_C2500_H150_3/file_rwpms_wcnf_L2_V150_C2500_H150_3-time_vs_initial_score}
\end{figure}

\begin{figure}[H]
    \centering
    \includegraphics[height=3.5in]{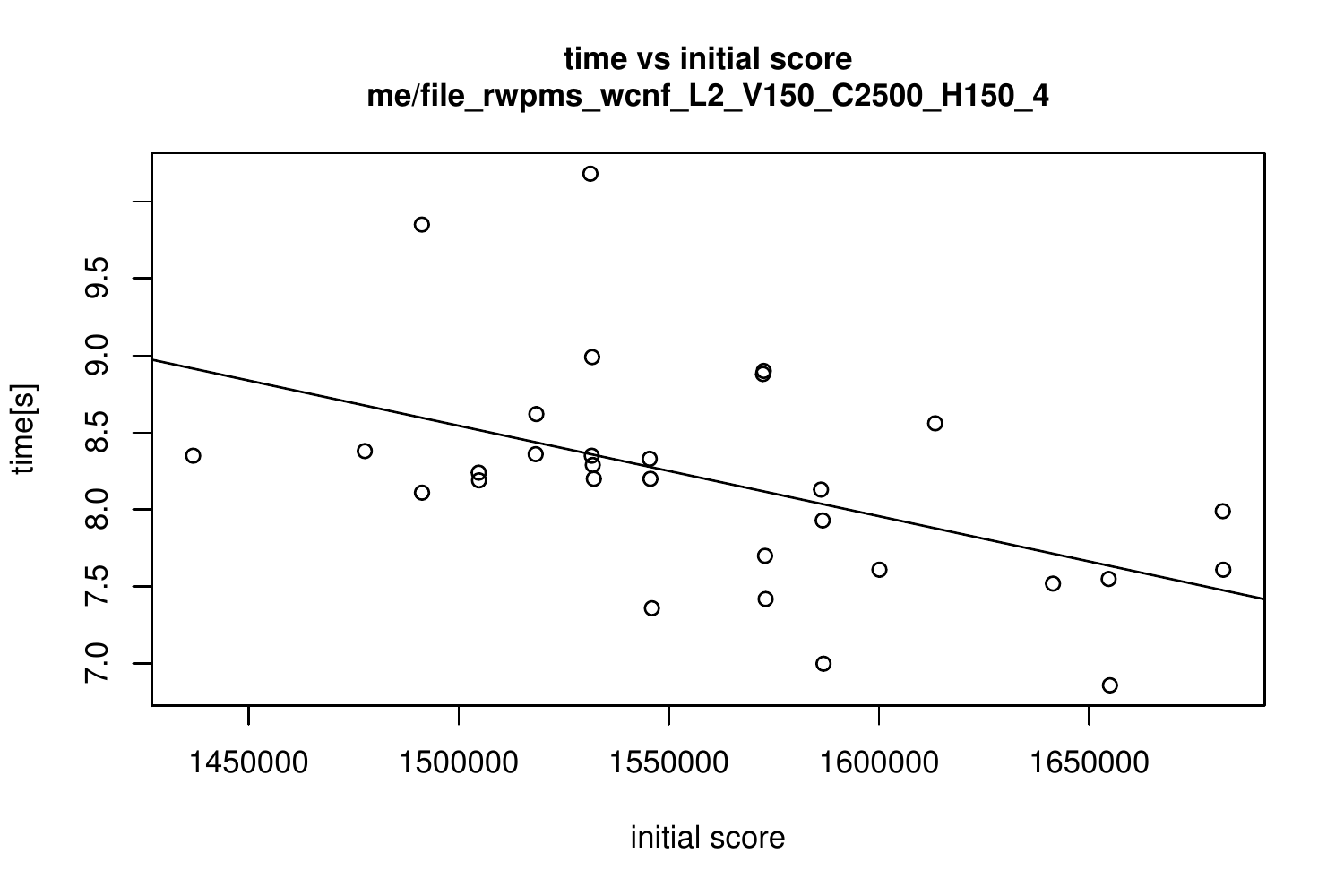}
    \label{fig_me/file_rwpms_wcnf_L2_V150_C2500_H150_4/file_rwpms_wcnf_L2_V150_C2500_H150_4-time_vs_initial_score}
\end{figure}

\begin{figure}[H]
    \centering
    \includegraphics[height=3.5in]{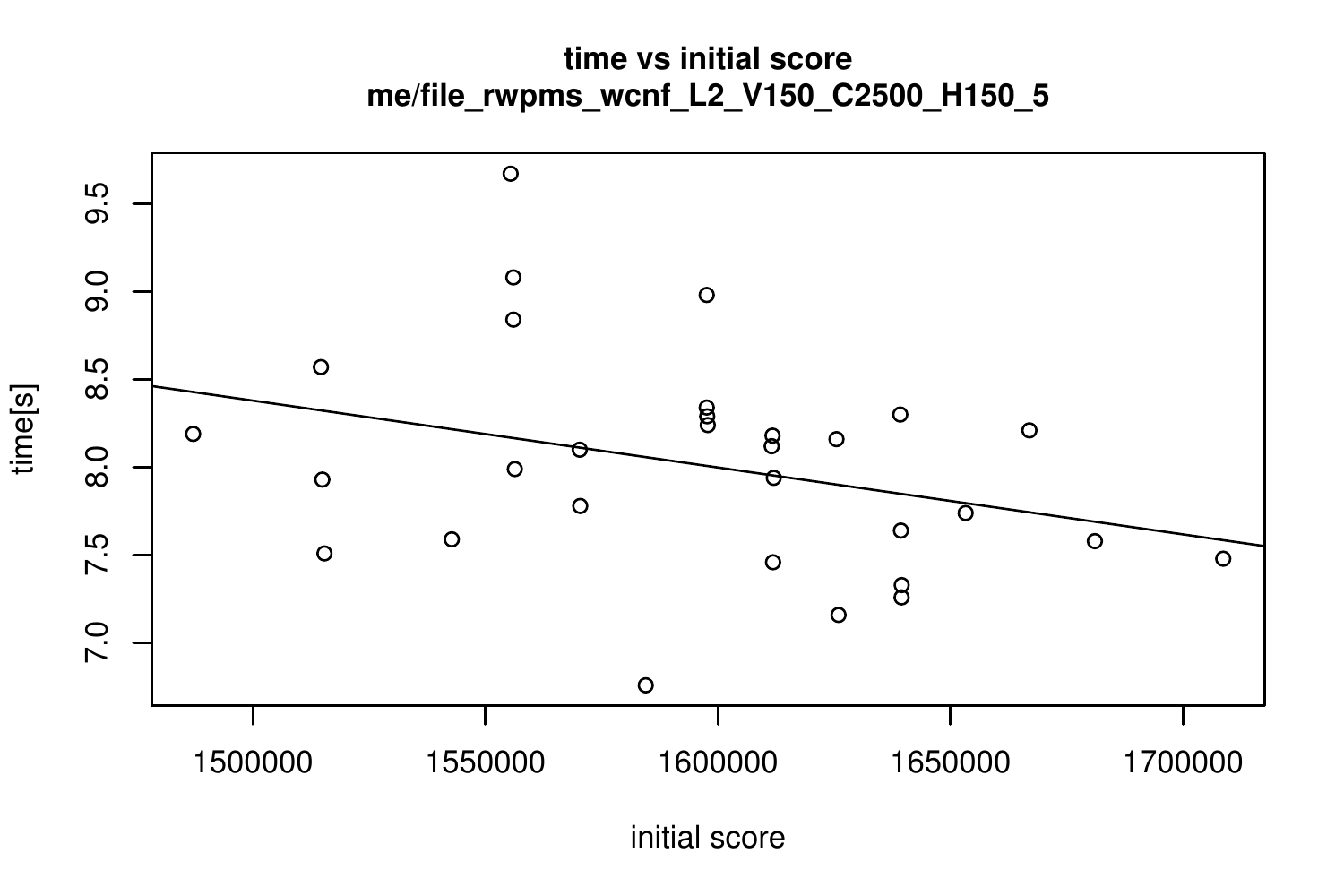}
    \label{fig_me/file_rwpms_wcnf_L2_V150_C2500_H150_5/file_rwpms_wcnf_L2_V150_C2500_H150_5-time_vs_initial_score}
\end{figure}

\begin{figure}[H]
    \centering
    \includegraphics[height=3.5in]{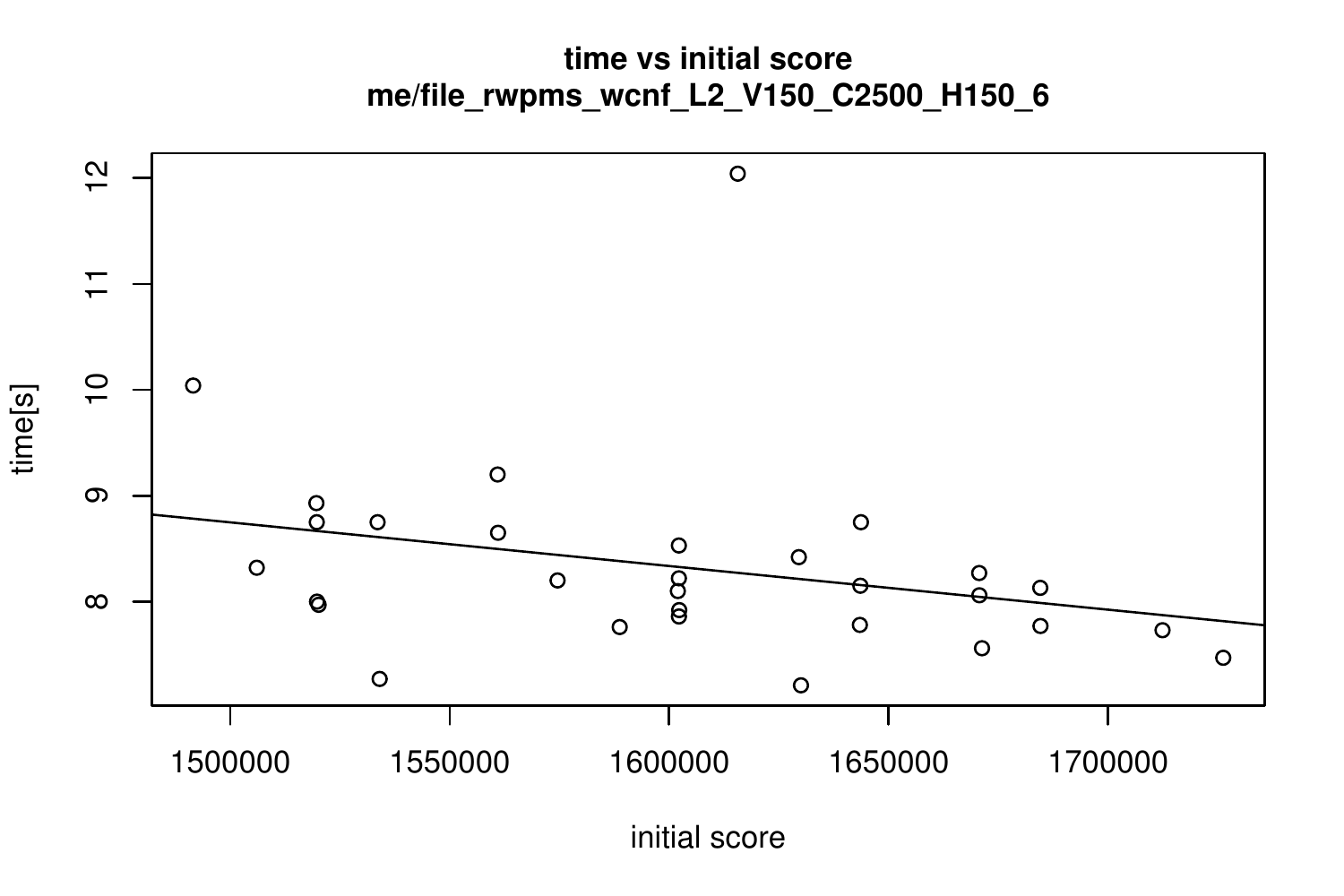}
    \label{fig_me/file_rwpms_wcnf_L2_V150_C2500_H150_6/file_rwpms_wcnf_L2_V150_C2500_H150_6-time_vs_initial_score}
\end{figure}

\begin{figure}[H]
    \centering
    \includegraphics[height=3.5in]{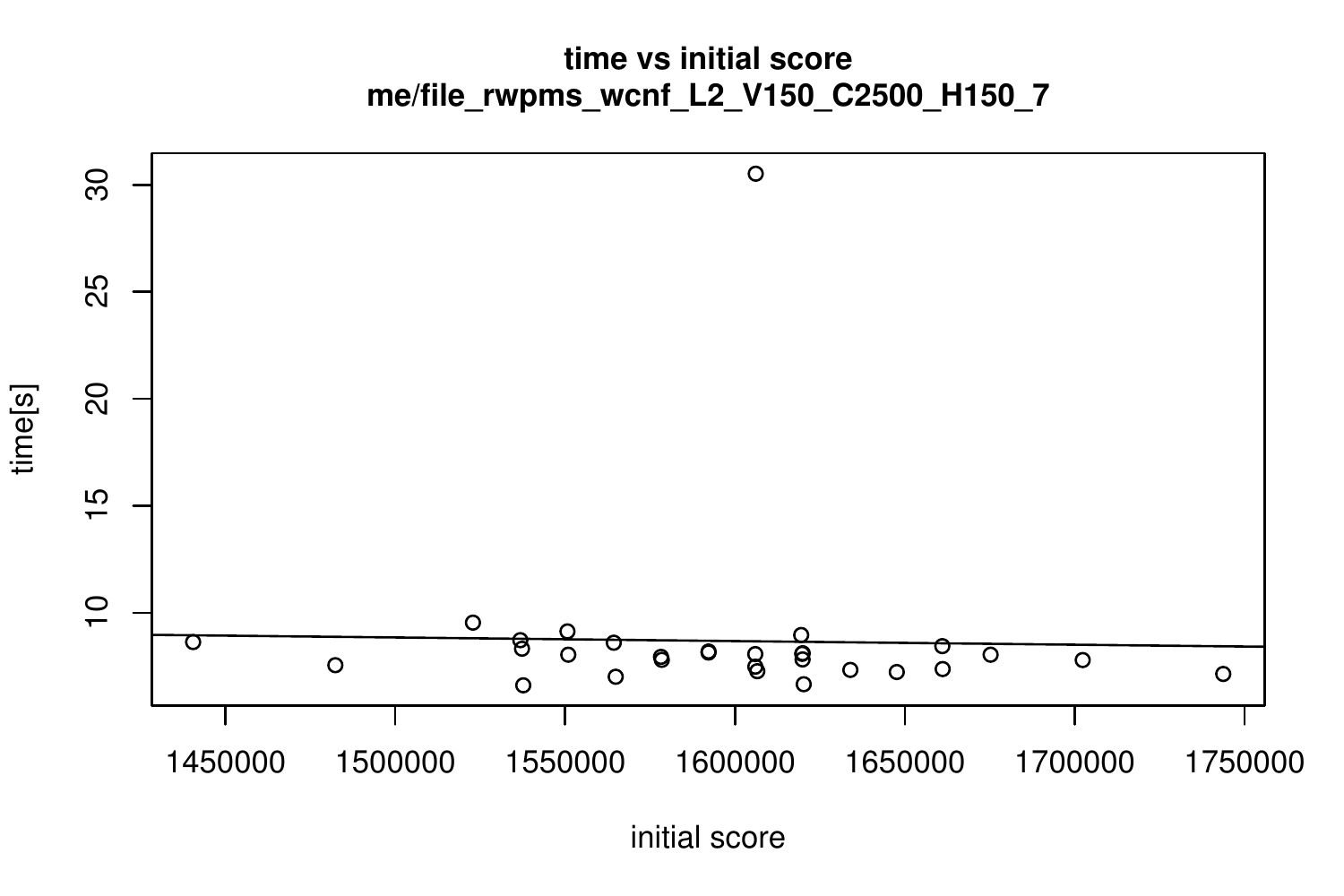}
    \label{fig_me/file_rwpms_wcnf_L2_V150_C2500_H150_7/file_rwpms_wcnf_L2_V150_C2500_H150_7-time_vs_initial_score}
\end{figure}

\begin{figure}[H]
    \centering
    \includegraphics[height=3.5in]{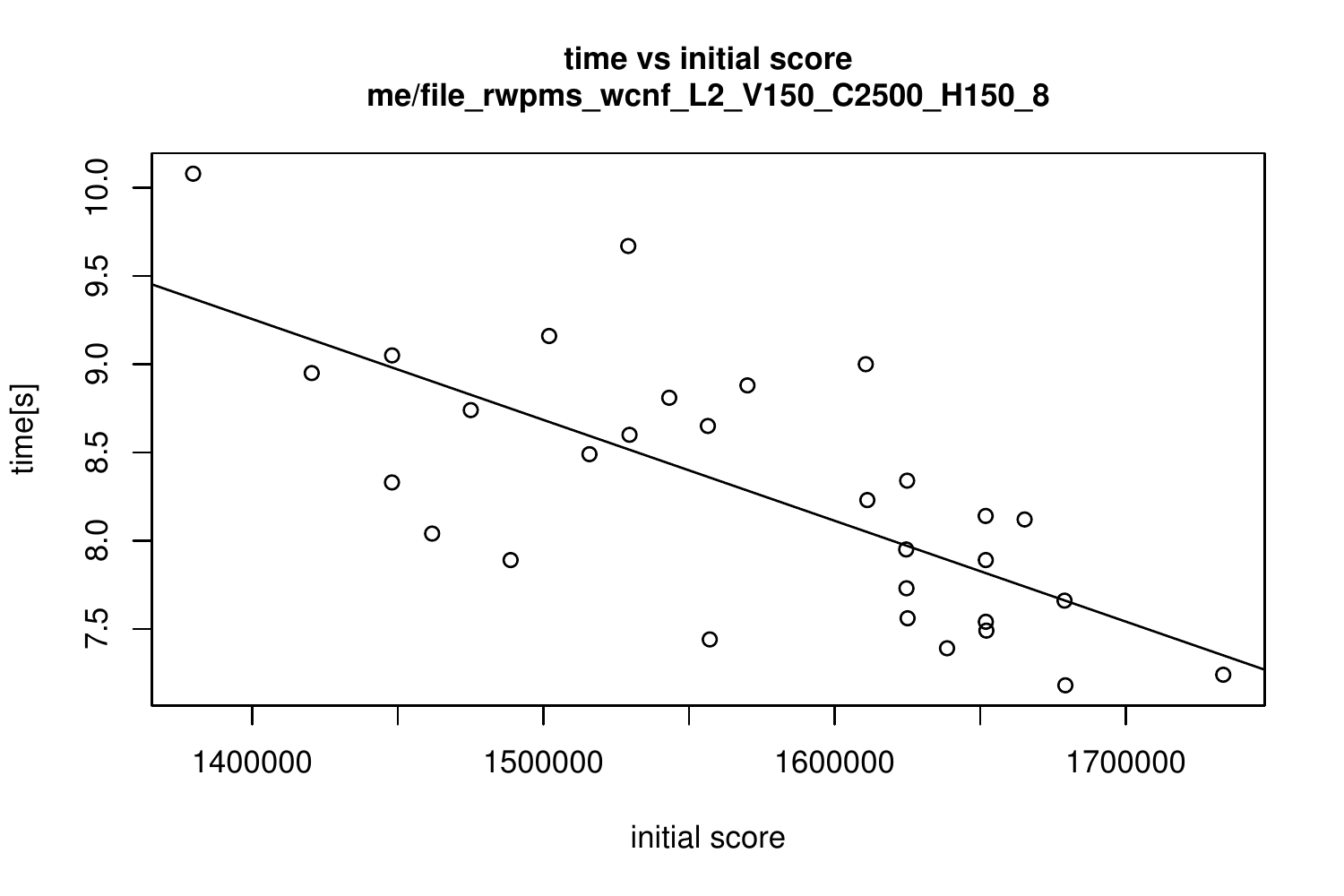}
    \label{fig_me/file_rwpms_wcnf_L2_V150_C2500_H150_8/file_rwpms_wcnf_L2_V150_C2500_H150_8-time_vs_initial_score}
\end{figure}

\begin{figure}[H]
    \centering
    \includegraphics[height=3.5in]{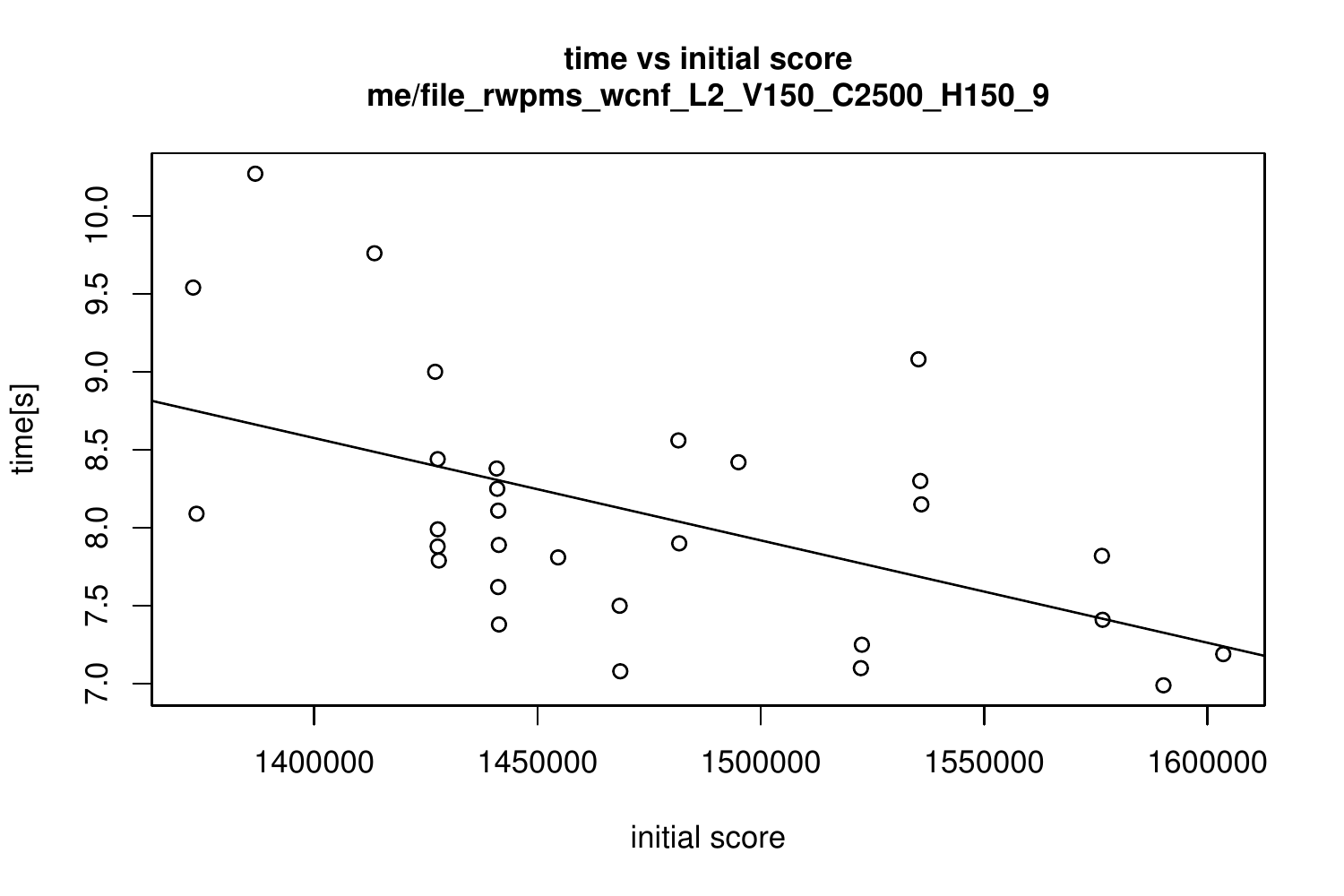}
    \label{fig_me/file_rwpms_wcnf_L2_V150_C2500_H150_9/file_rwpms_wcnf_L2_V150_C2500_H150_9-time_vs_initial_score}
\end{figure}

\begin{figure}[H]
    \centering
    \includegraphics[height=3.5in]{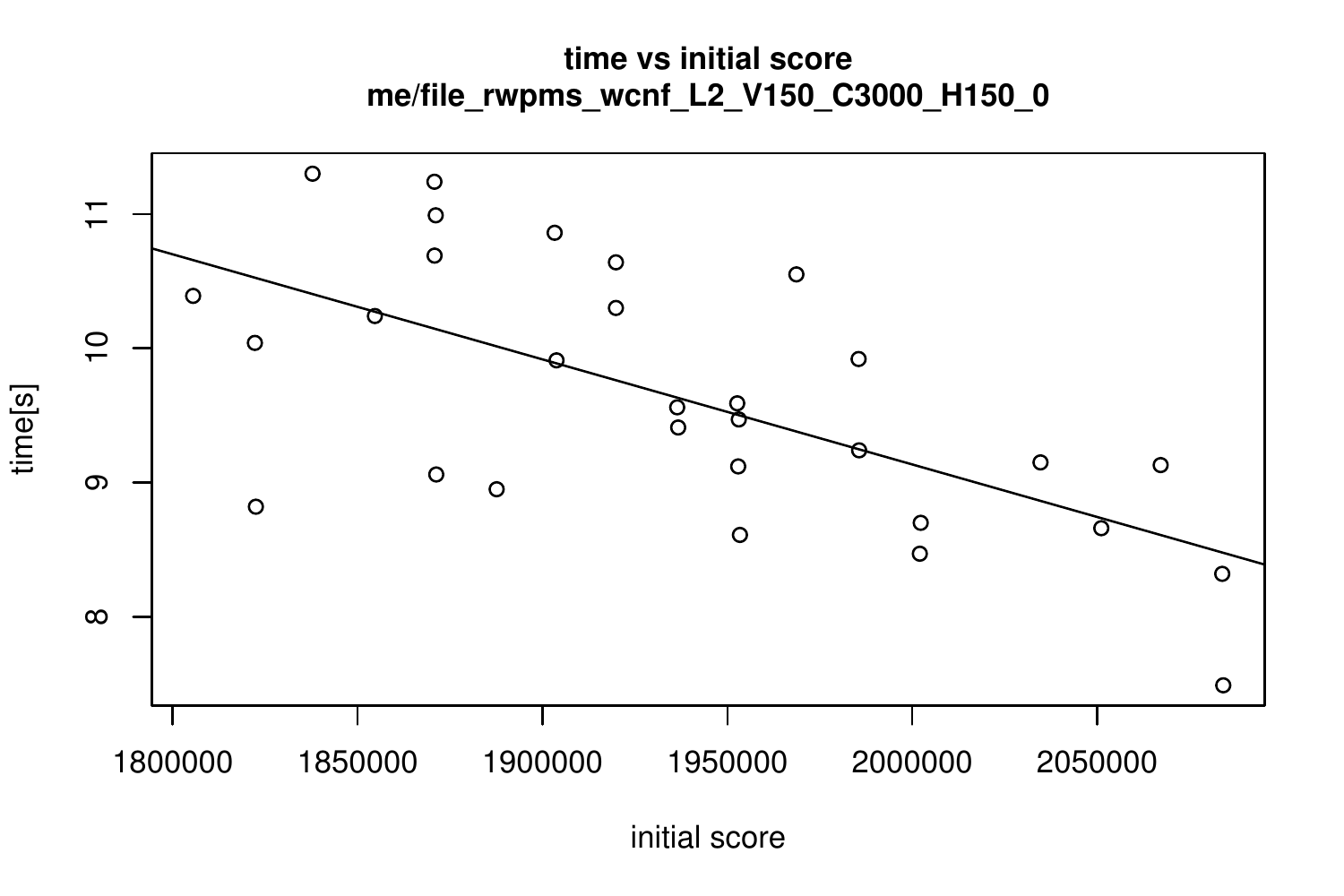}
    \label{fig_me/file_rwpms_wcnf_L2_V150_C3000_H150_0/file_rwpms_wcnf_L2_V150_C3000_H150_0-time_vs_initial_score}
\end{figure}

\begin{figure}[H]
    \centering
    \includegraphics[height=3.5in]{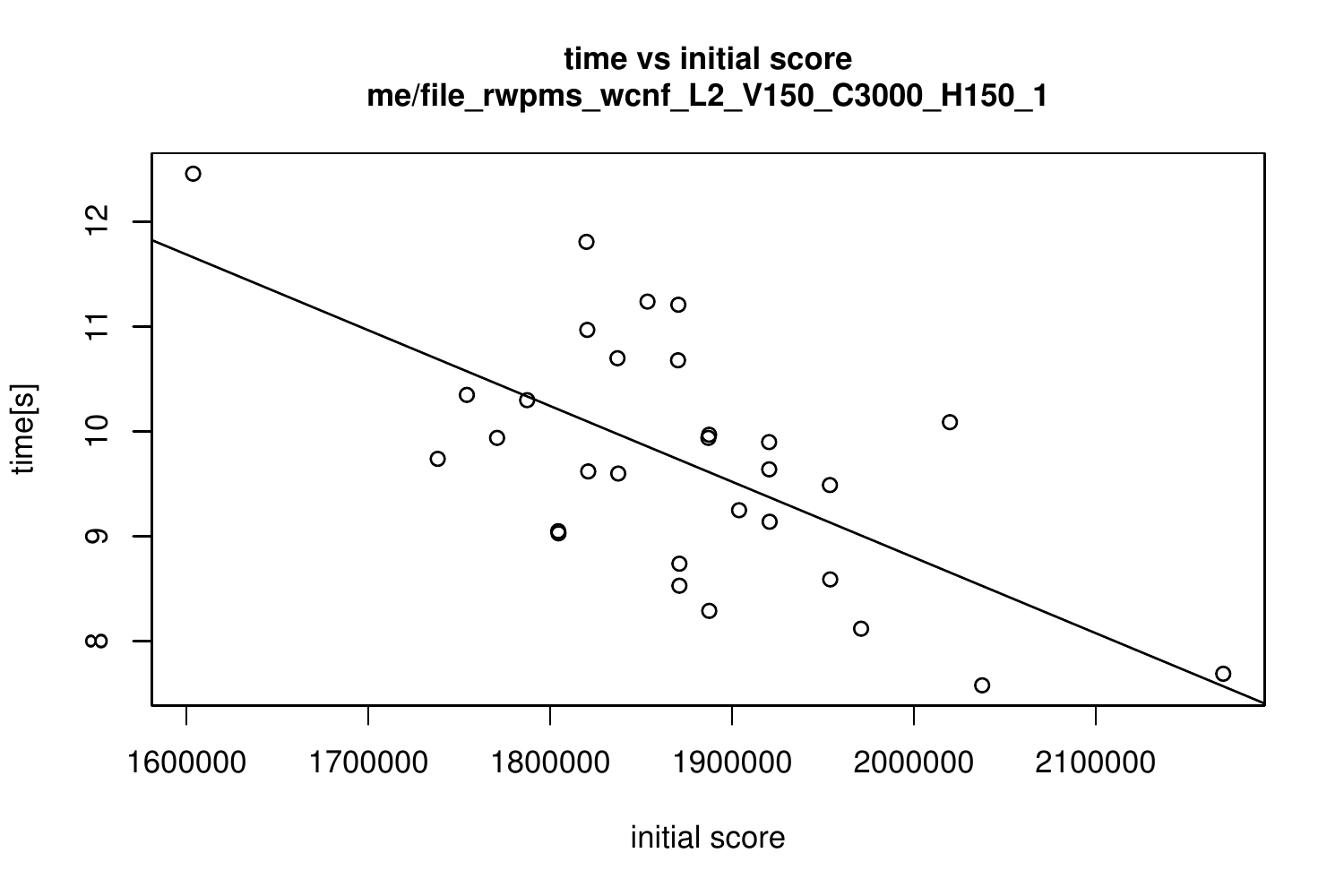}
    \label{fig_me/file_rwpms_wcnf_L2_V150_C3000_H150_1/file_rwpms_wcnf_L2_V150_C3000_H150_1-time_vs_initial_score}
\end{figure}

\begin{figure}[H]
    \centering
    \includegraphics[height=3.5in]{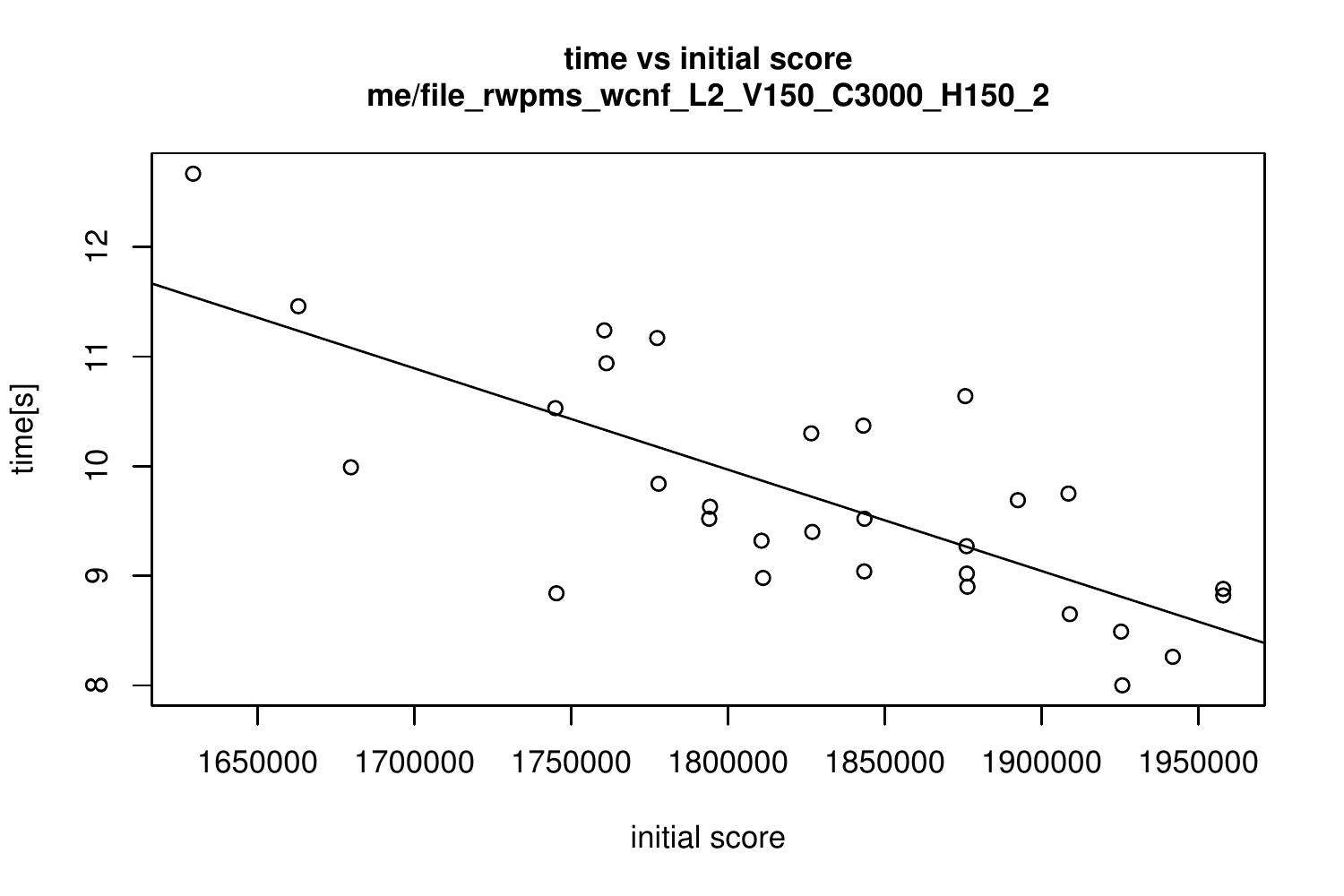}
    \label{fig_me/file_rwpms_wcnf_L2_V150_C3000_H150_2/file_rwpms_wcnf_L2_V150_C3000_H150_2-time_vs_initial_score}
\end{figure}

\begin{figure}[H]
    \centering
    \includegraphics[height=3.5in]{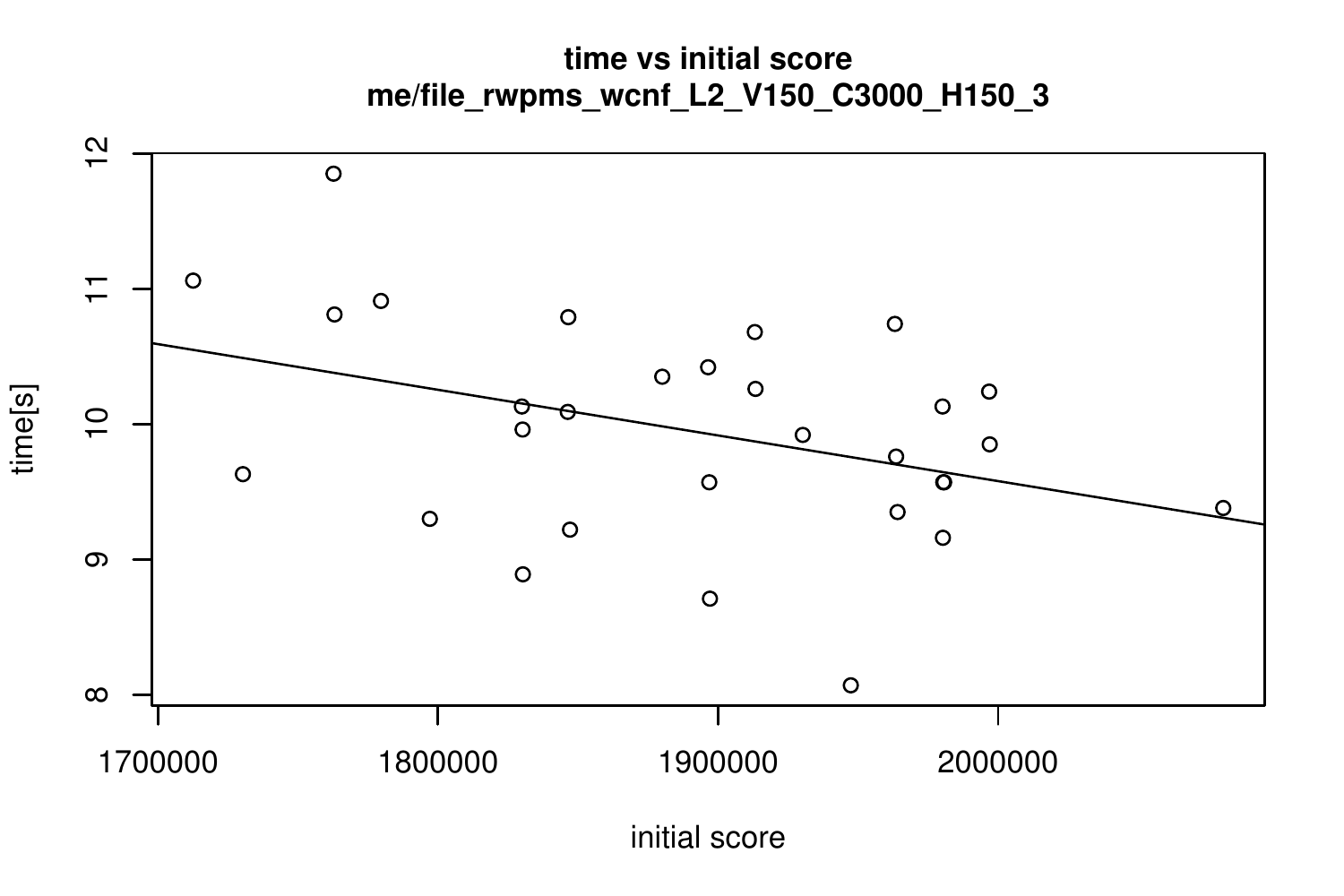}
    \label{fig_me/file_rwpms_wcnf_L2_V150_C3000_H150_3/file_rwpms_wcnf_L2_V150_C3000_H150_3-time_vs_initial_score}
\end{figure}

\begin{figure}[H]
    \centering
    \includegraphics[height=3.5in]{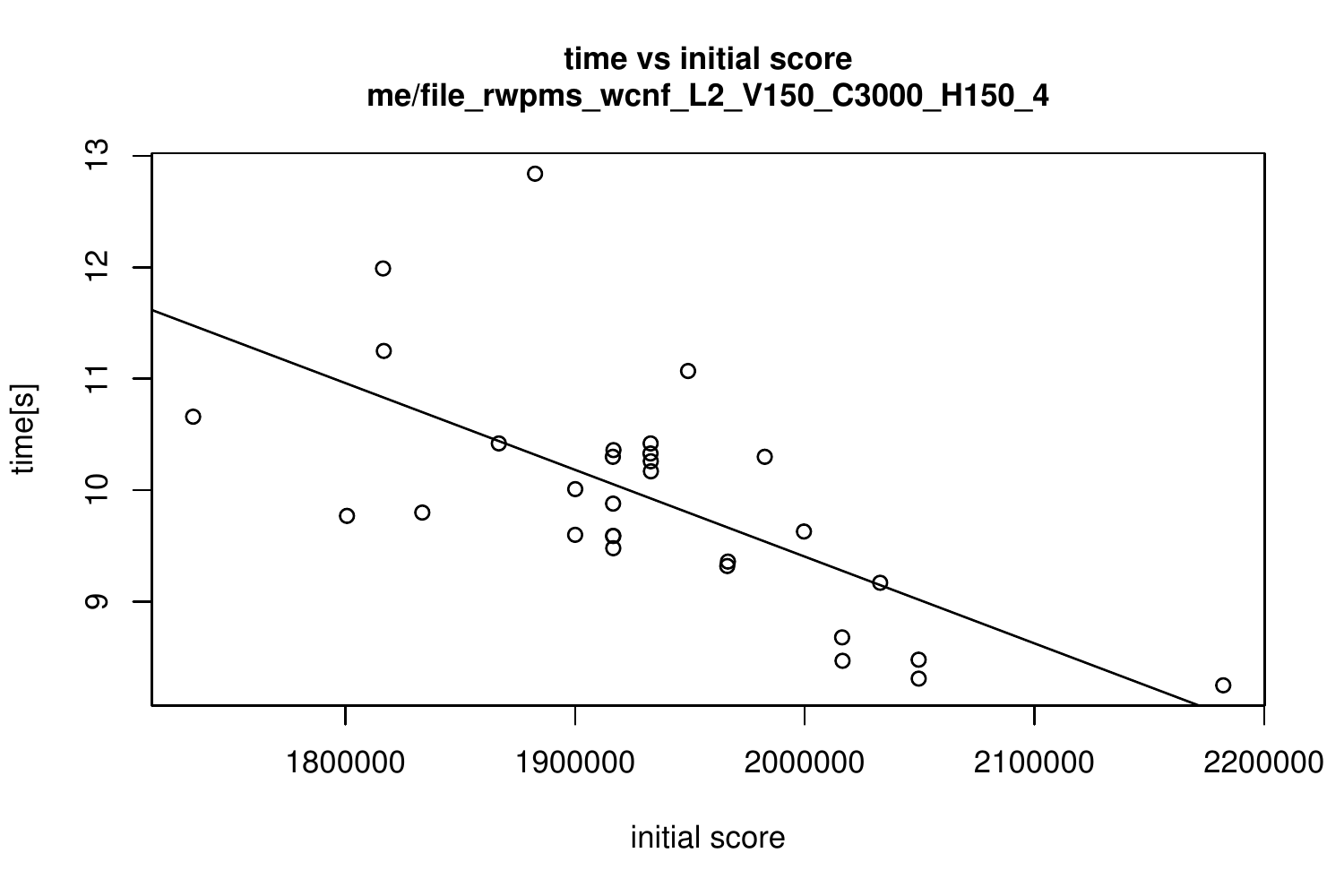}
    \label{fig_me/file_rwpms_wcnf_L2_V150_C3000_H150_4/file_rwpms_wcnf_L2_V150_C3000_H150_4-time_vs_initial_score}
\end{figure}

\begin{figure}[H]
    \centering
    \includegraphics[height=3.5in]{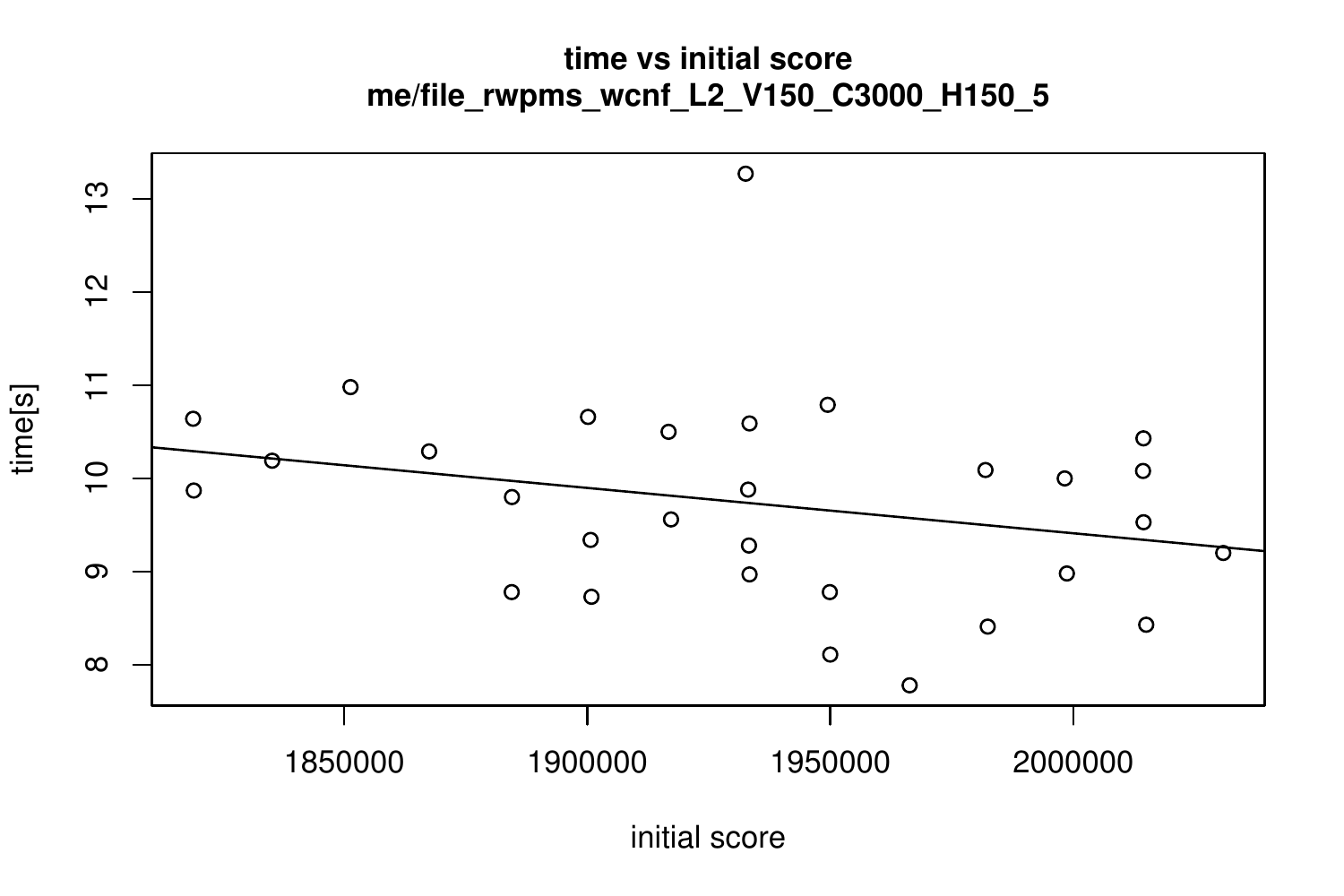}
    \label{fig_me/file_rwpms_wcnf_L2_V150_C3000_H150_5/file_rwpms_wcnf_L2_V150_C3000_H150_5-time_vs_initial_score}
\end{figure}

\begin{figure}[H]
    \centering
    \includegraphics[height=3.5in]{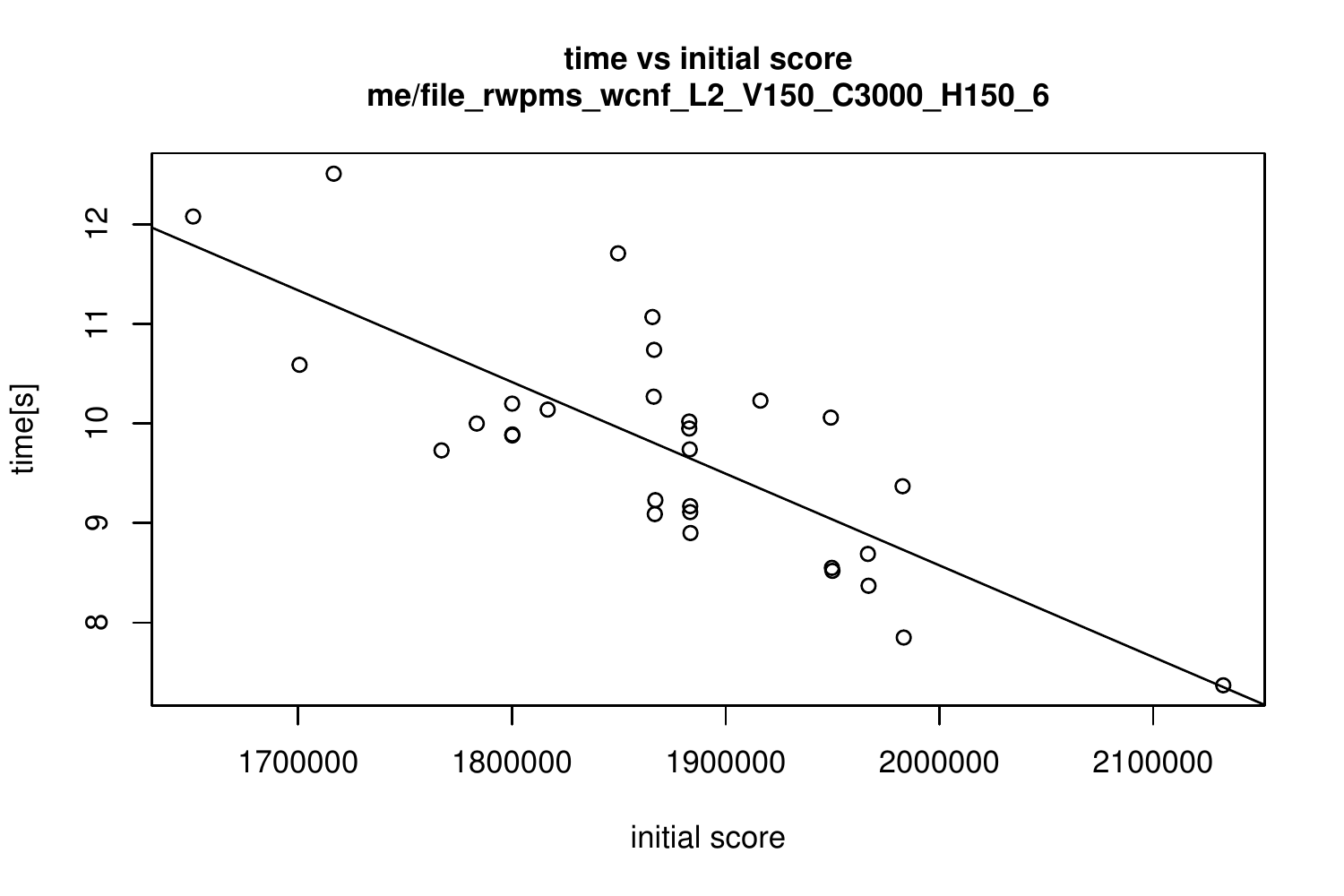}
    \label{fig_me/file_rwpms_wcnf_L2_V150_C3000_H150_6/file_rwpms_wcnf_L2_V150_C3000_H150_6-time_vs_initial_score}
\end{figure}

\begin{figure}[H]
    \centering
    \includegraphics[height=3.5in]{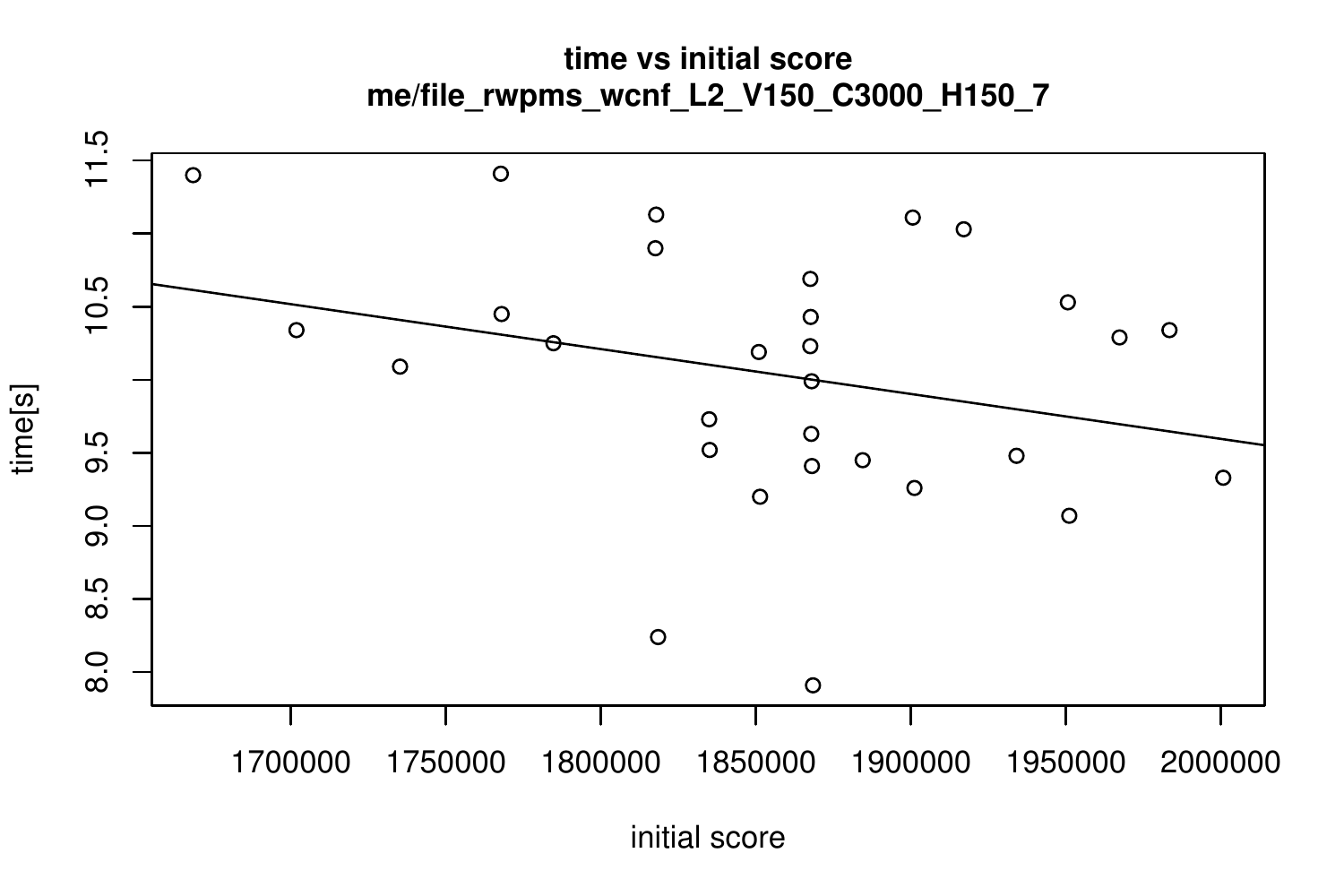}
    \label{fig_me/file_rwpms_wcnf_L2_V150_C3000_H150_7/file_rwpms_wcnf_L2_V150_C3000_H150_7-time_vs_initial_score}
\end{figure}

\begin{figure}[H]
    \centering
    \includegraphics[height=3.5in]{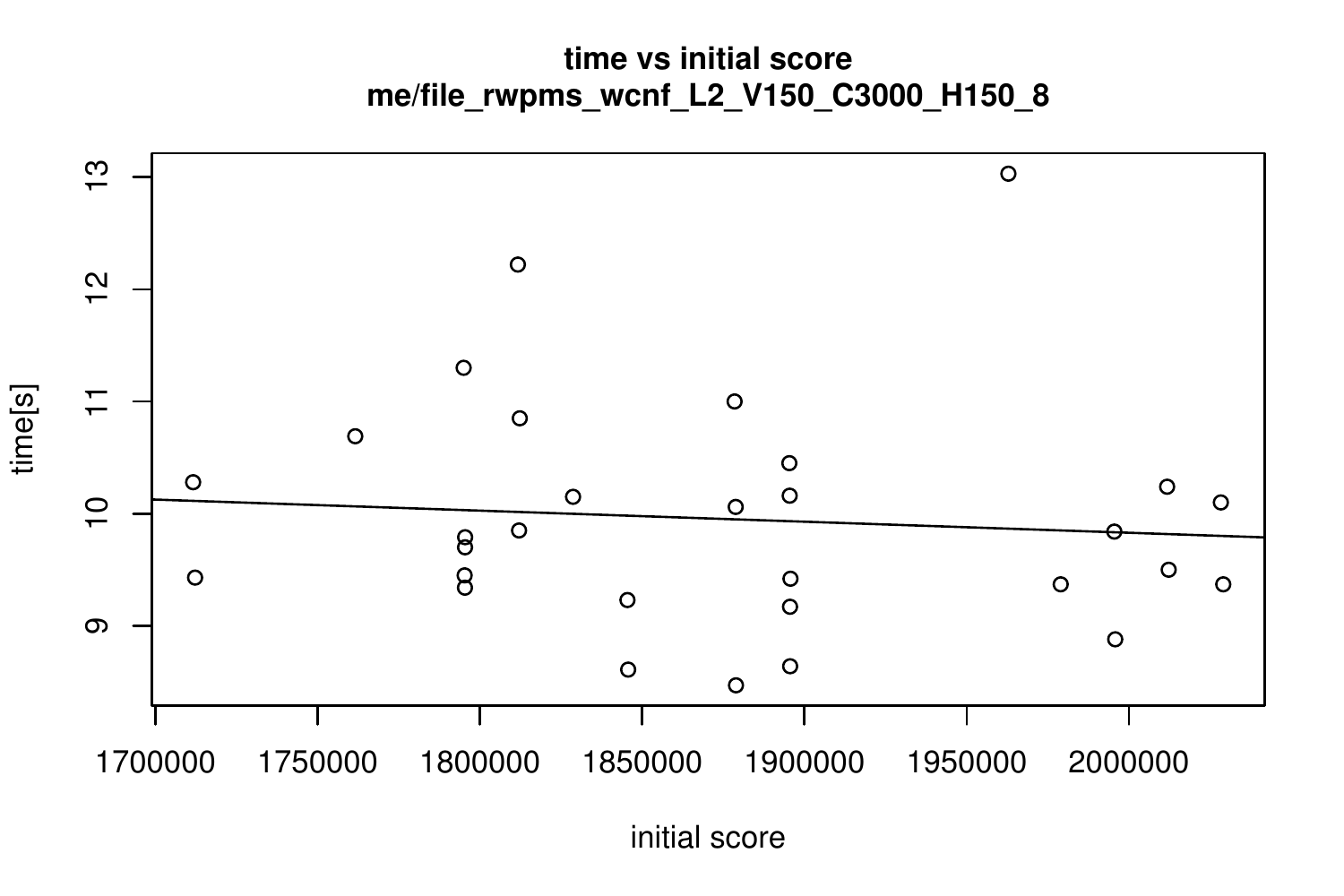}
    \label{fig_me/file_rwpms_wcnf_L2_V150_C3000_H150_8/file_rwpms_wcnf_L2_V150_C3000_H150_8-time_vs_initial_score}
\end{figure}

\begin{figure}[H]
    \centering
    \includegraphics[height=3.5in]{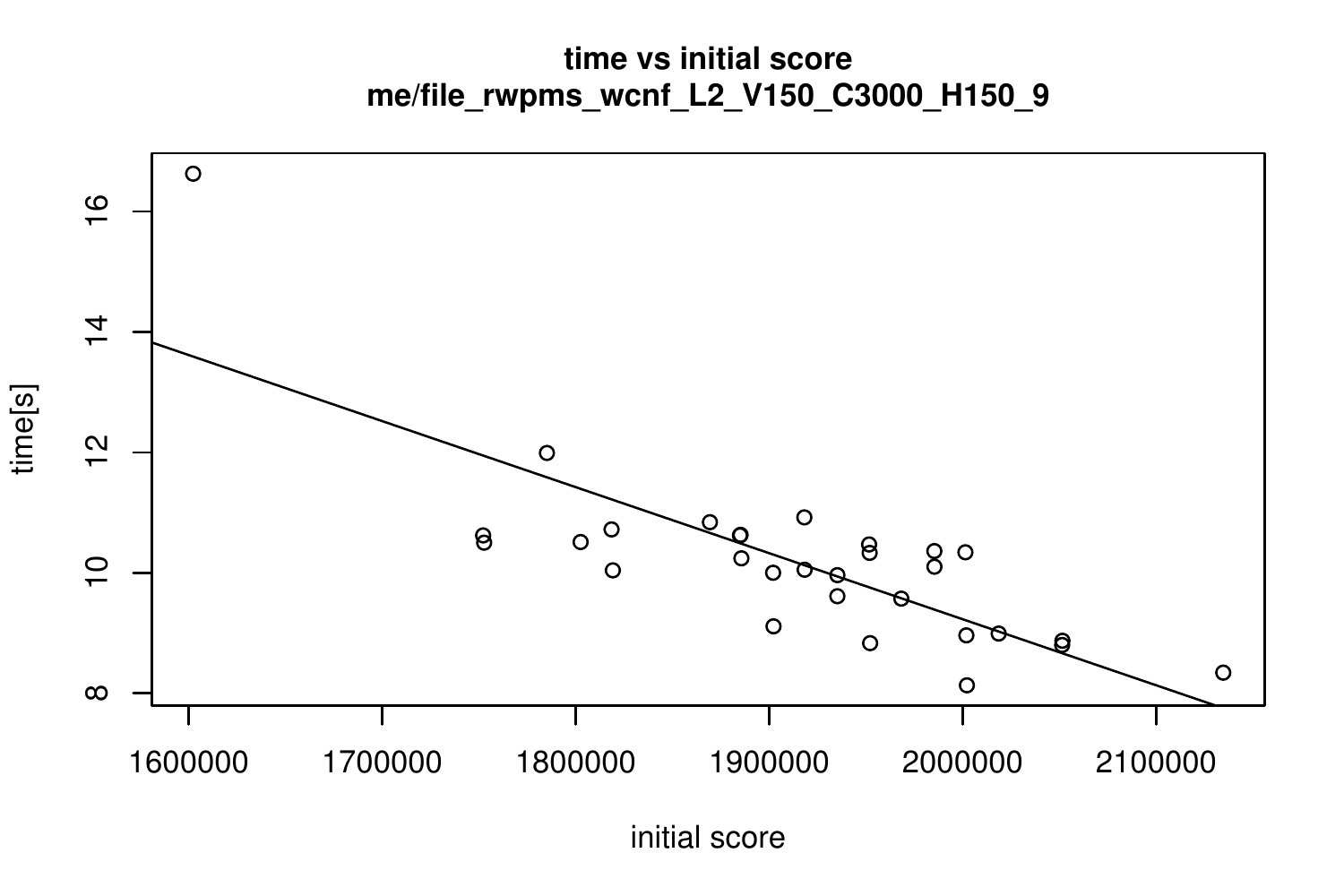}
    \label{fig_me/file_rwpms_wcnf_L2_V150_C3000_H150_9/file_rwpms_wcnf_L2_V150_C3000_H150_9-time_vs_initial_score}
\end{figure}

\begin{figure}[H]
    \centering
    \includegraphics[height=3.5in]{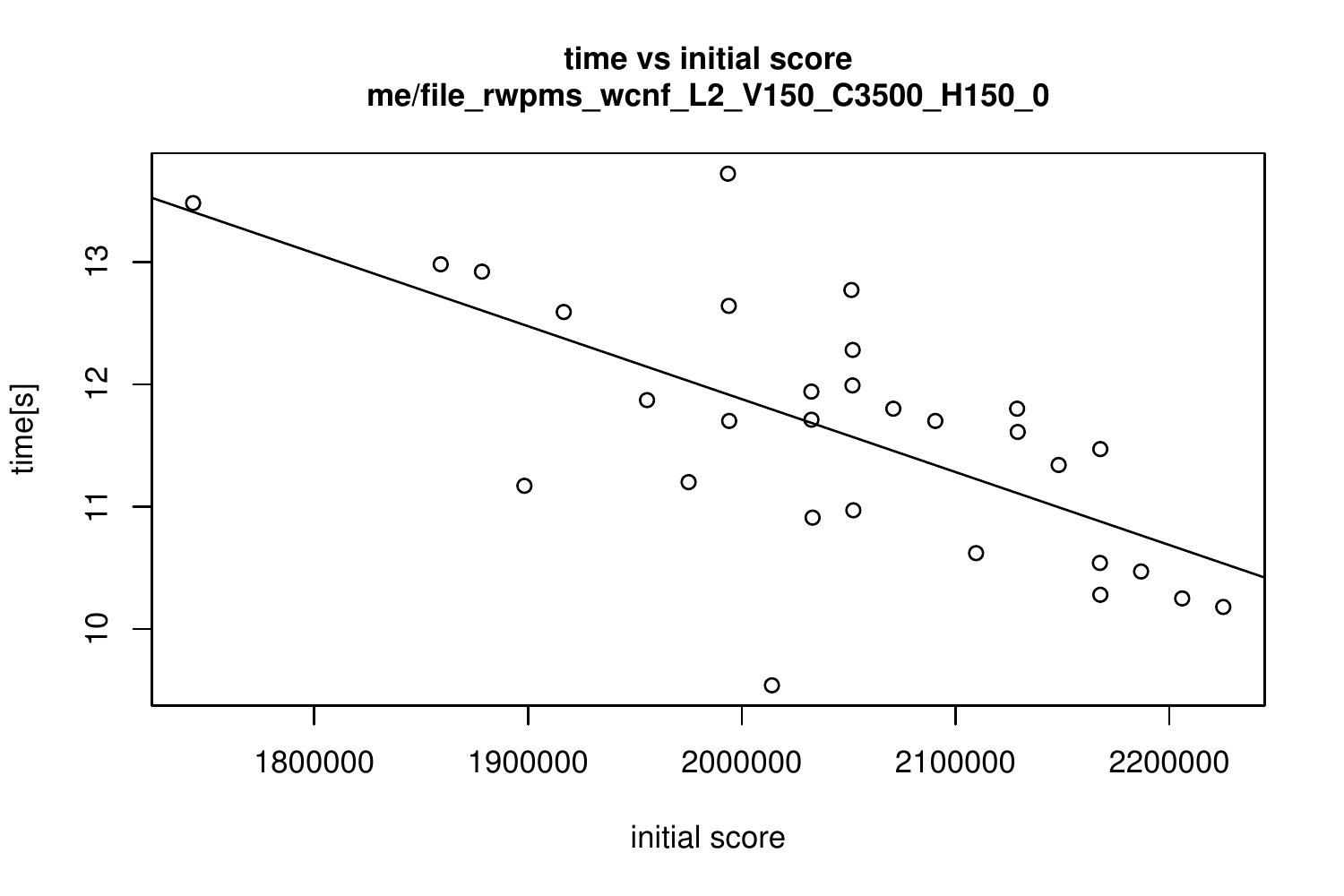}
    \label{fig_me/file_rwpms_wcnf_L2_V150_C3500_H150_0/file_rwpms_wcnf_L2_V150_C3500_H150_0-time_vs_initial_score}
\end{figure}

\begin{figure}[H]
    \centering
    \includegraphics[height=3.5in]{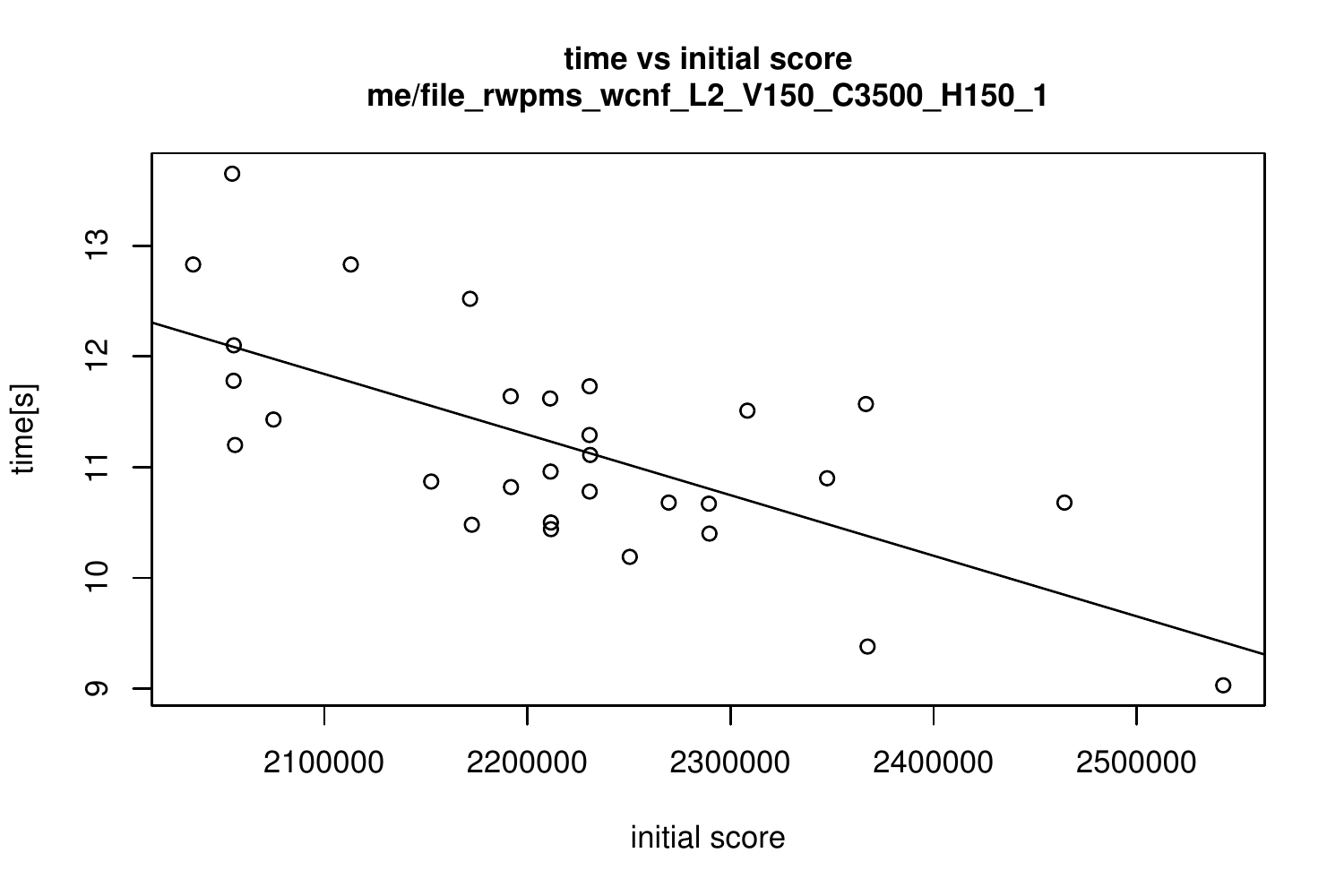}
    \label{fig_me/file_rwpms_wcnf_L2_V150_C3500_H150_1/file_rwpms_wcnf_L2_V150_C3500_H150_1-time_vs_initial_score}
\end{figure}

\begin{figure}[H]
    \centering
    \includegraphics[height=3.5in]{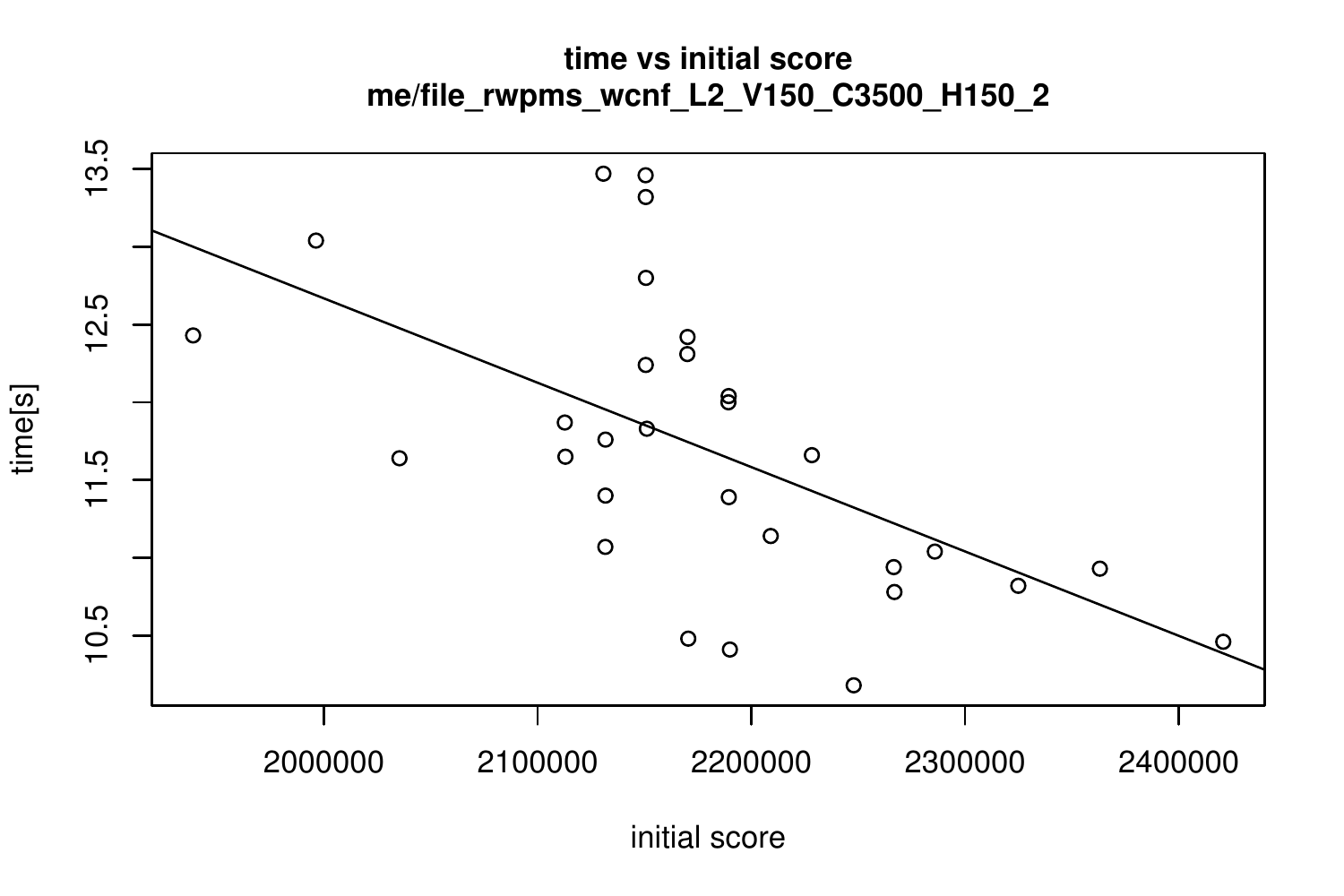}
    \label{fig_me/file_rwpms_wcnf_L2_V150_C3500_H150_2/file_rwpms_wcnf_L2_V150_C3500_H150_2-time_vs_initial_score}
\end{figure}

\begin{figure}[H]
    \centering
    \includegraphics[height=3.5in]{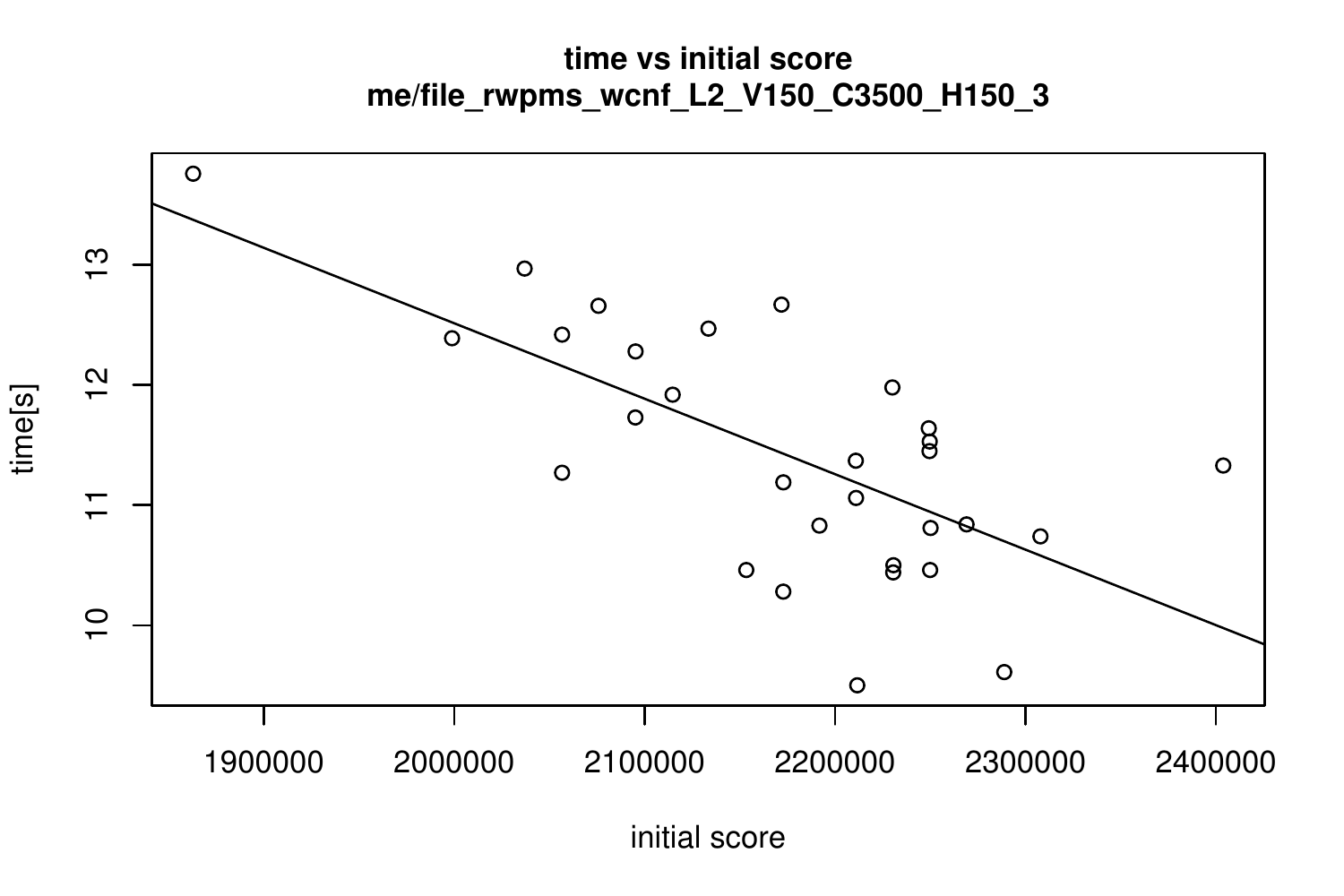}
    \label{fig_me/file_rwpms_wcnf_L2_V150_C3500_H150_3/file_rwpms_wcnf_L2_V150_C3500_H150_3-time_vs_initial_score}
\end{figure}

\begin{figure}[H]
    \centering
    \includegraphics[height=3.5in]{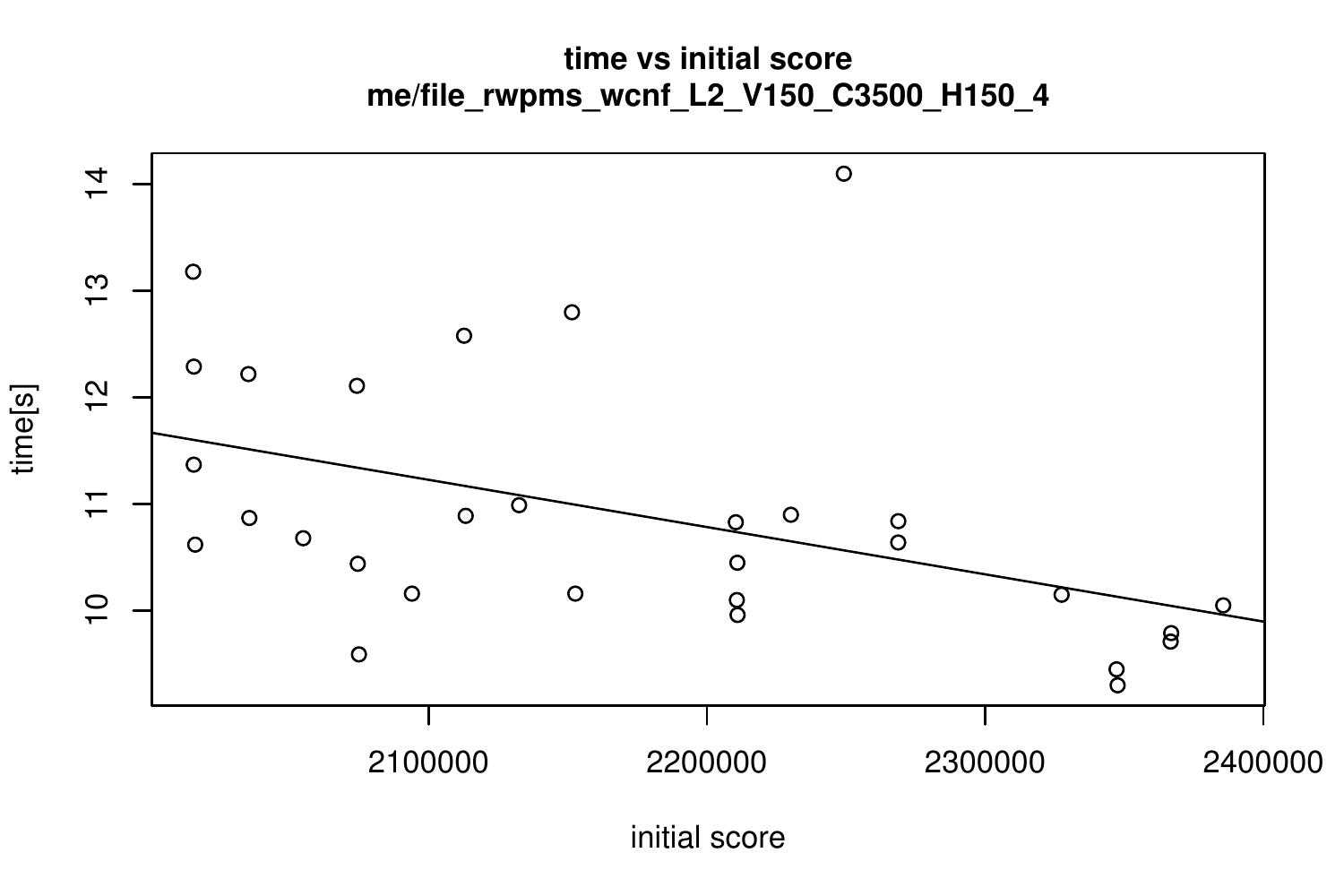}
    \label{fig_me/file_rwpms_wcnf_L2_V150_C3500_H150_4/file_rwpms_wcnf_L2_V150_C3500_H150_4-time_vs_initial_score}
\end{figure}

\begin{figure}[H]
    \centering
    \includegraphics[height=3.5in]{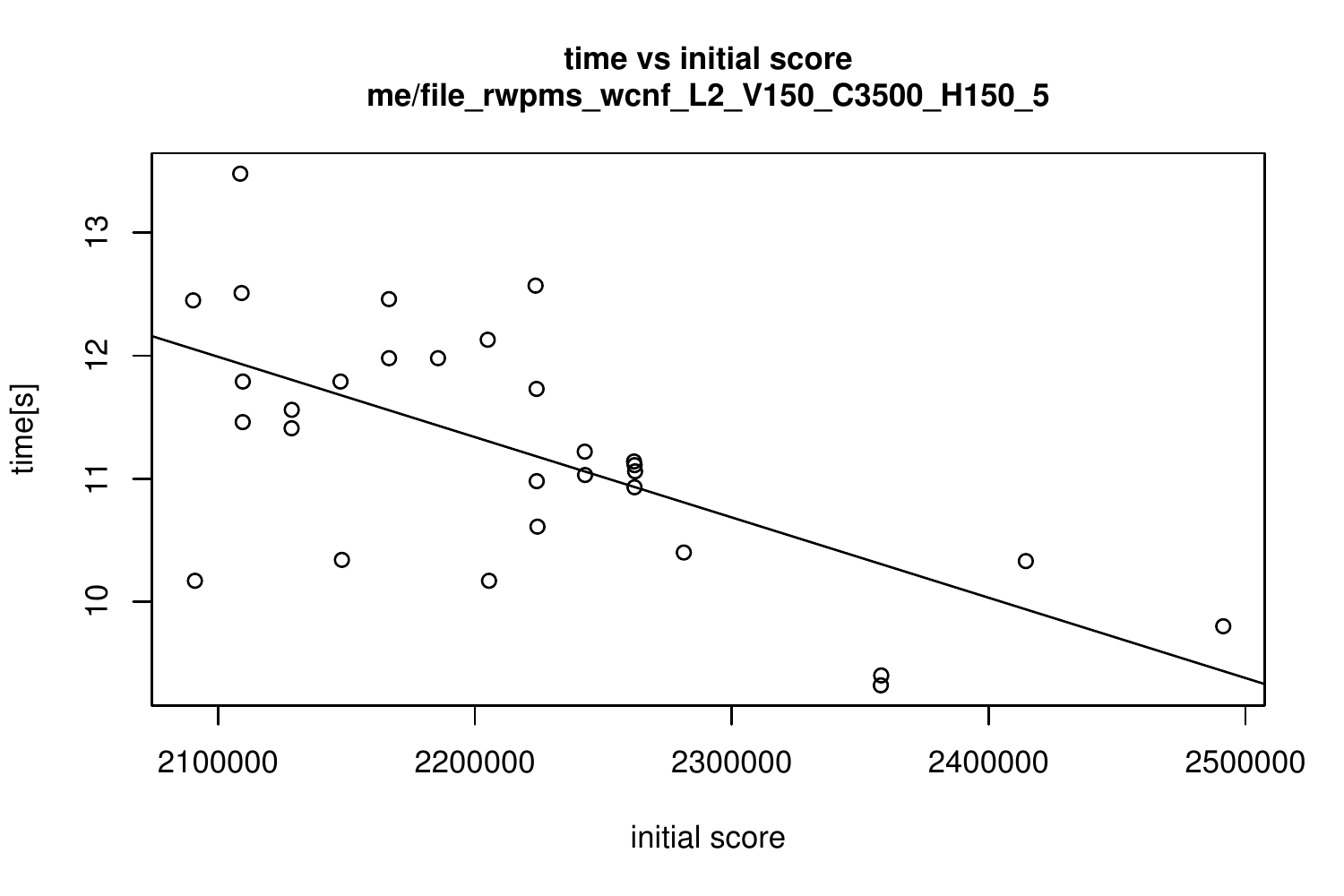}
    \label{fig_me/file_rwpms_wcnf_L2_V150_C3500_H150_5/file_rwpms_wcnf_L2_V150_C3500_H150_5-time_vs_initial_score}
\end{figure}

\begin{figure}[H]
    \centering
    \includegraphics[height=3.5in]{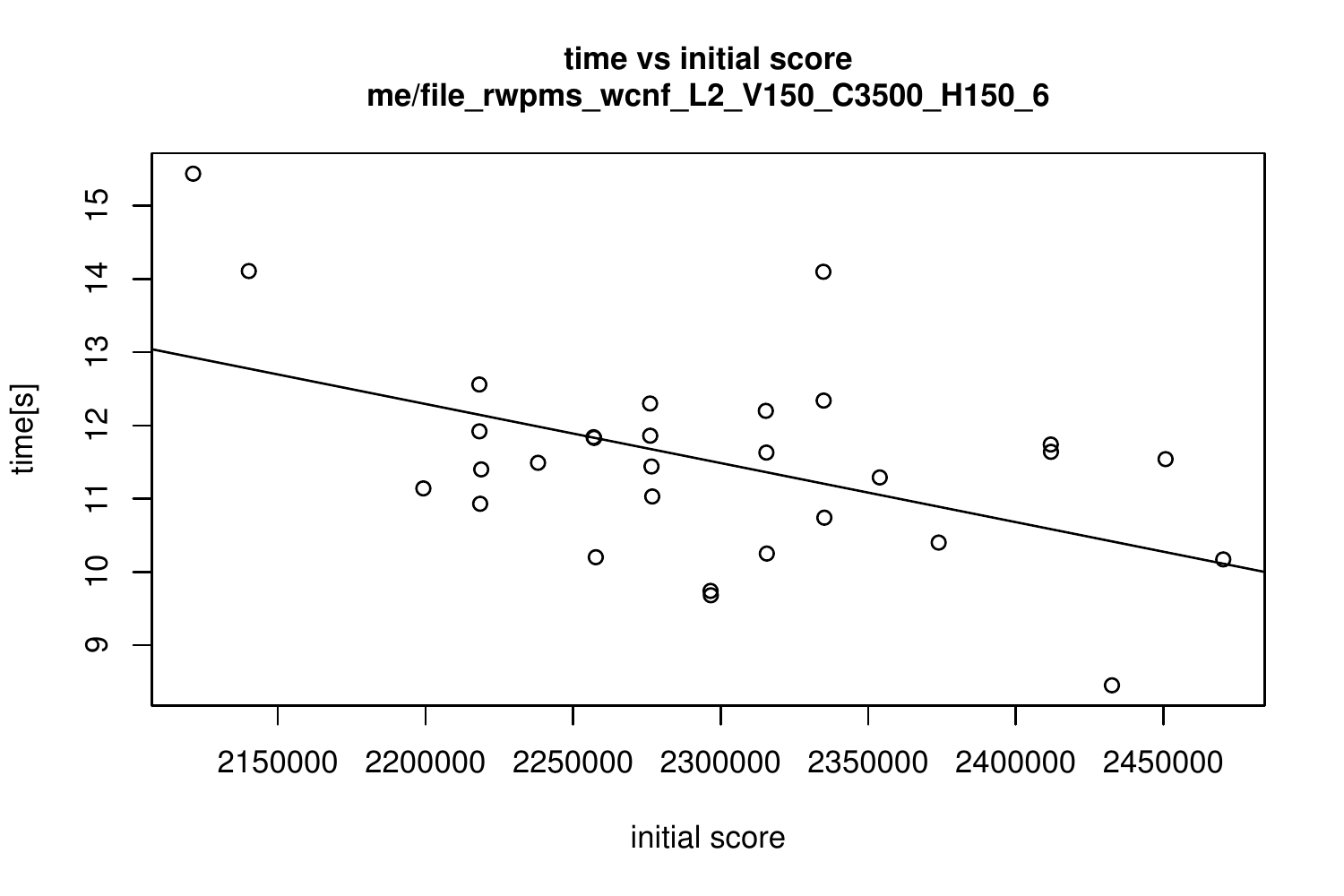}
    \label{fig_me/file_rwpms_wcnf_L2_V150_C3500_H150_6/file_rwpms_wcnf_L2_V150_C3500_H150_6-time_vs_initial_score}
\end{figure}

\begin{figure}[H]
    \centering
    \includegraphics[height=3.5in]{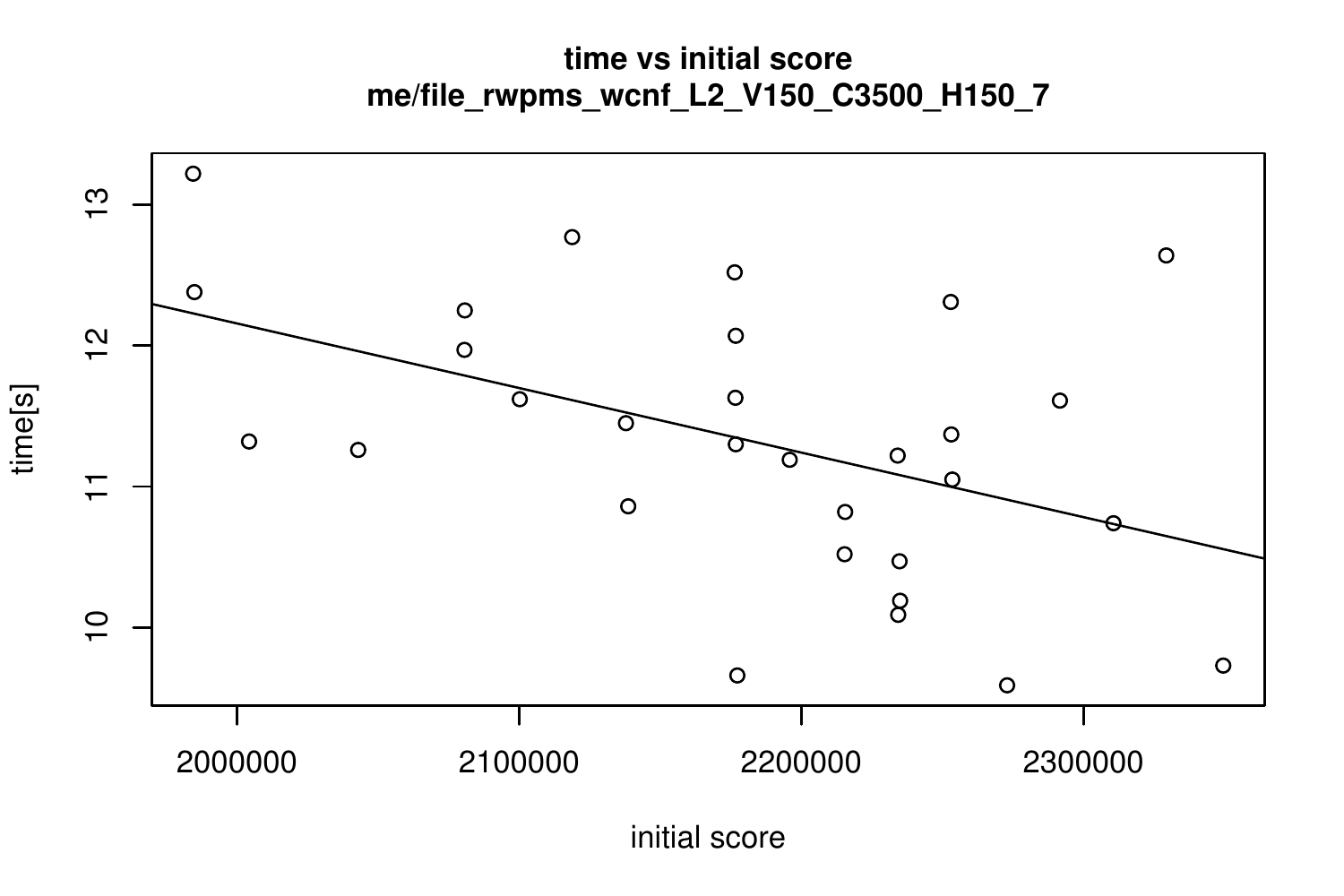}
    \label{fig_me/file_rwpms_wcnf_L2_V150_C3500_H150_7/file_rwpms_wcnf_L2_V150_C3500_H150_7-time_vs_initial_score}
\end{figure}

\begin{figure}[H]
    \centering
    \includegraphics[height=3.5in]{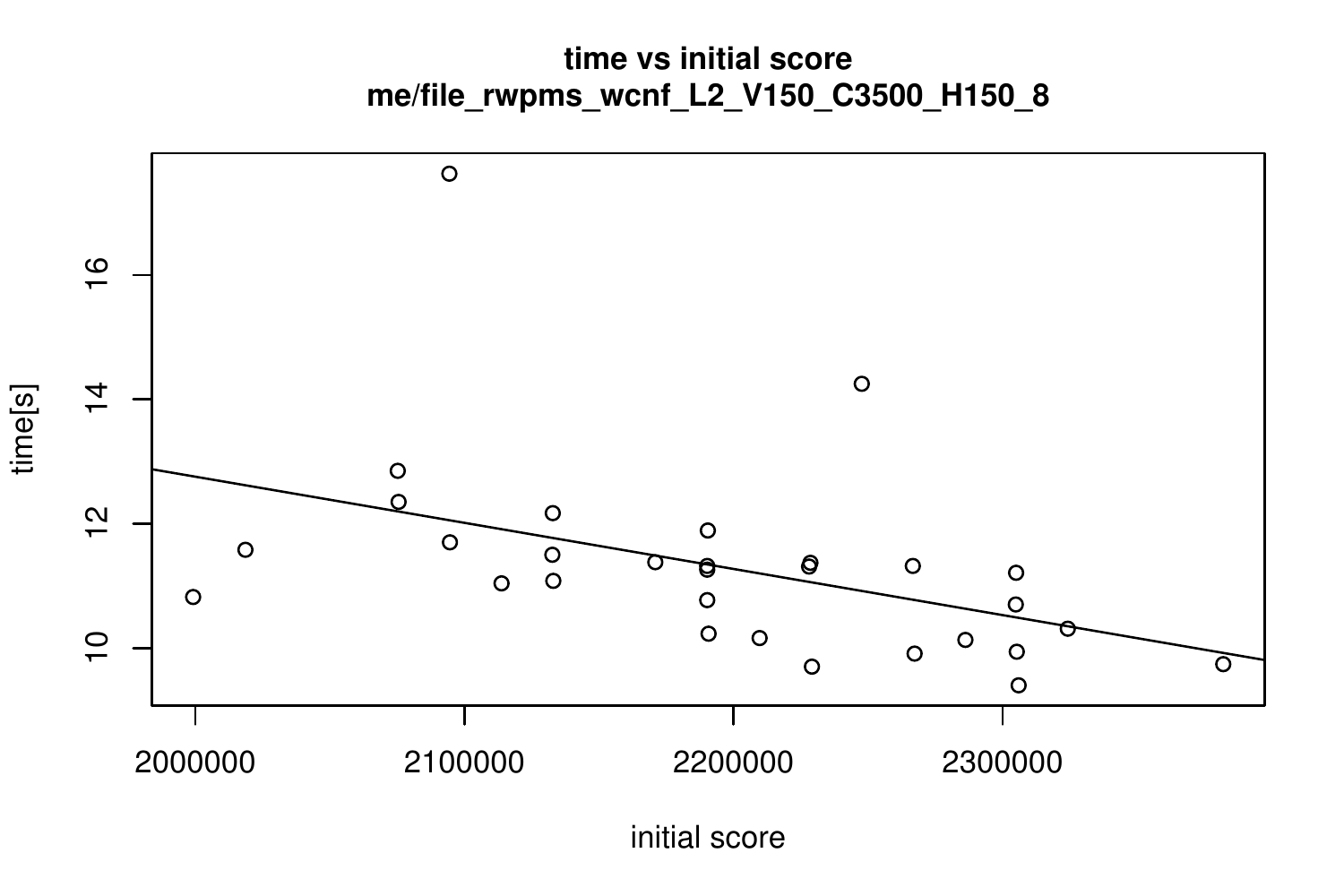}
    \label{fig_me/file_rwpms_wcnf_L2_V150_C3500_H150_8/file_rwpms_wcnf_L2_V150_C3500_H150_8-time_vs_initial_score}
\end{figure}

\begin{figure}[H]
    \centering
    \includegraphics[height=3.5in]{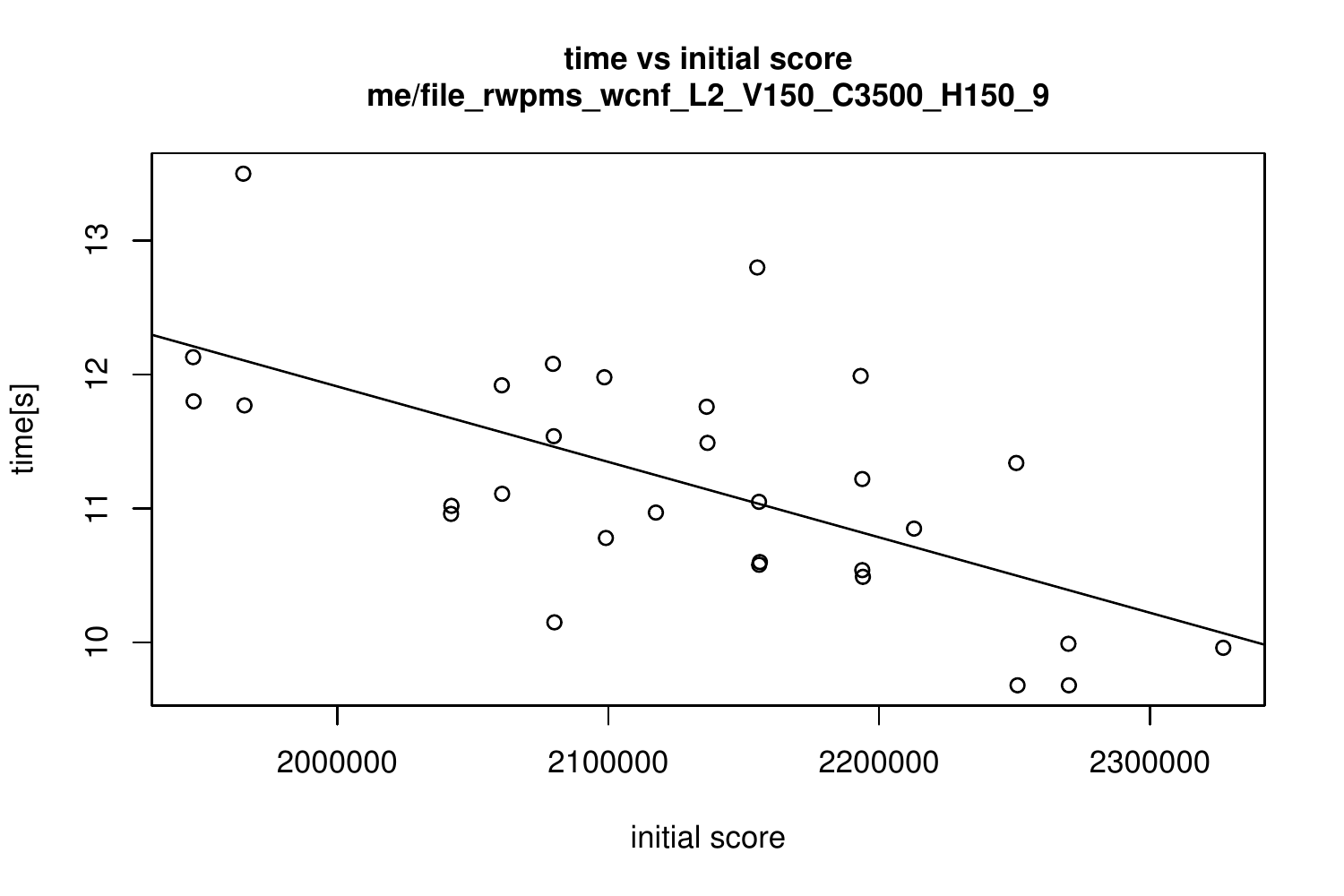}
    \label{fig_me/file_rwpms_wcnf_L2_V150_C3500_H150_9/file_rwpms_wcnf_L2_V150_C3500_H150_9-time_vs_initial_score}
\end{figure}

\begin{figure}[H]
    \centering
    \includegraphics[height=3.5in]{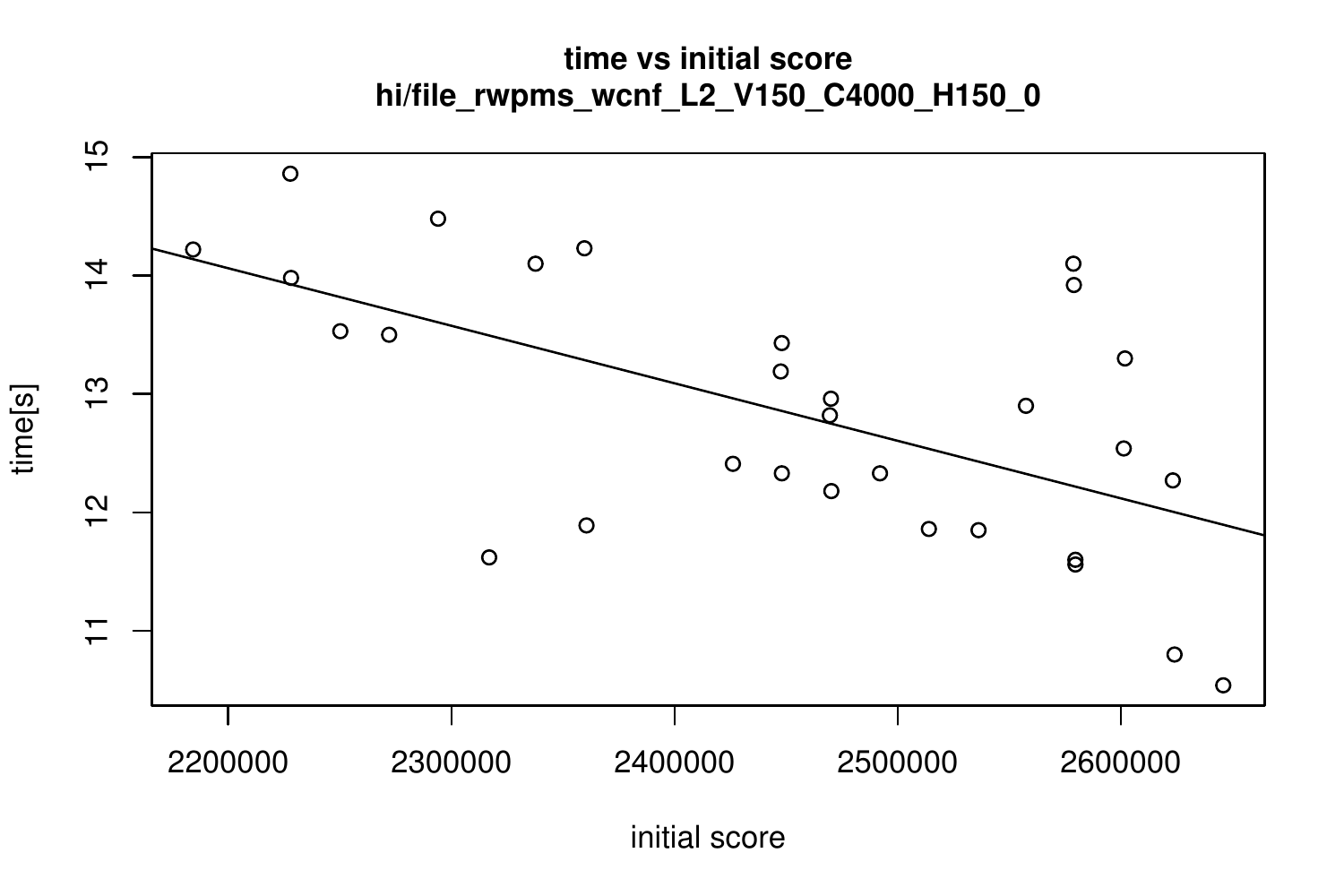}
    \label{fig_hi/file_rwpms_wcnf_L2_V150_C4000_H150_0/file_rwpms_wcnf_L2_V150_C4000_H150_0-time_vs_initial_score}
\end{figure}

\begin{figure}[H]
    \centering
    \includegraphics[height=3.5in]{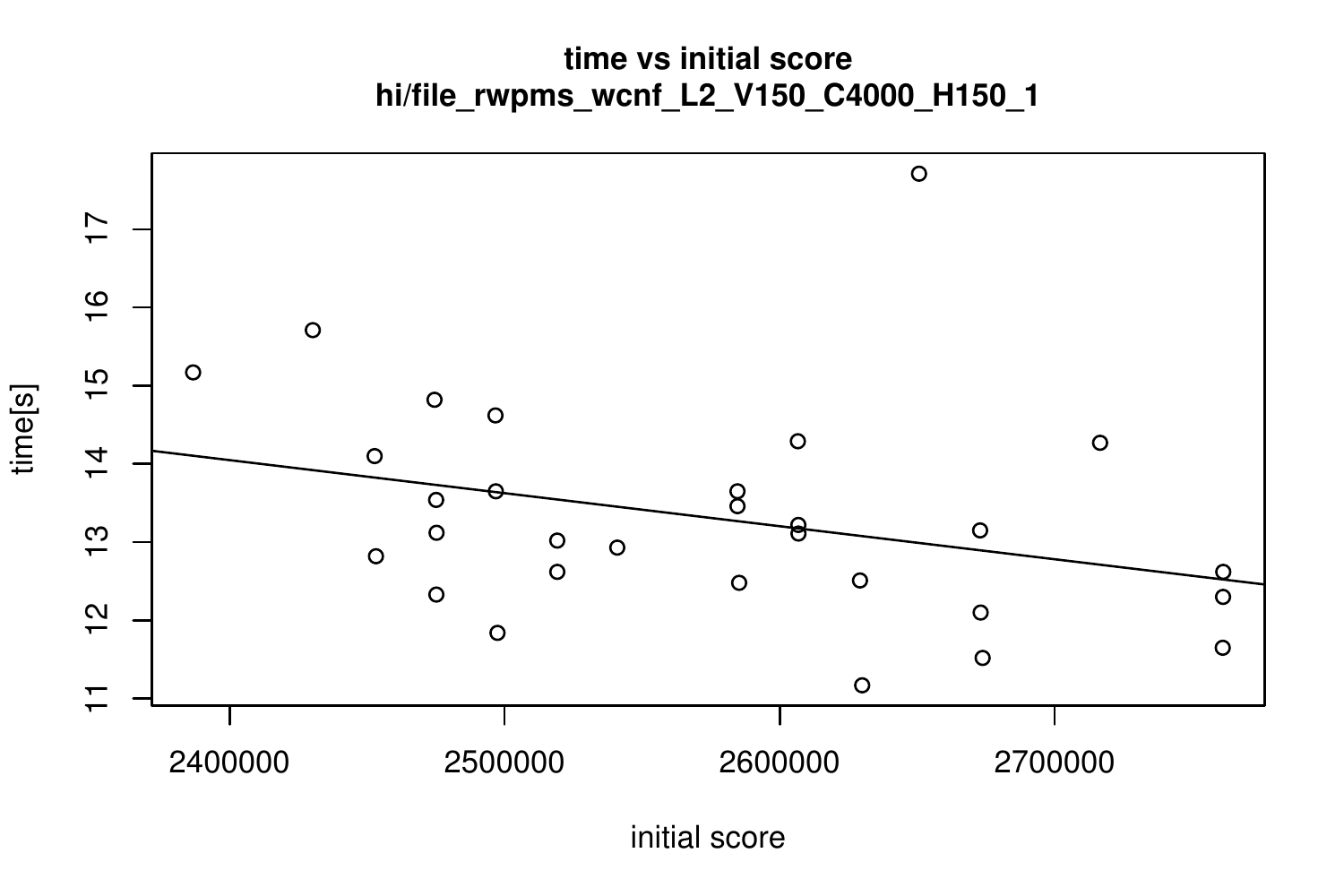}
    \label{fig_hi/file_rwpms_wcnf_L2_V150_C4000_H150_1/file_rwpms_wcnf_L2_V150_C4000_H150_1-time_vs_initial_score}
\end{figure}

\begin{figure}[H]
    \centering
    \includegraphics[height=3.5in]{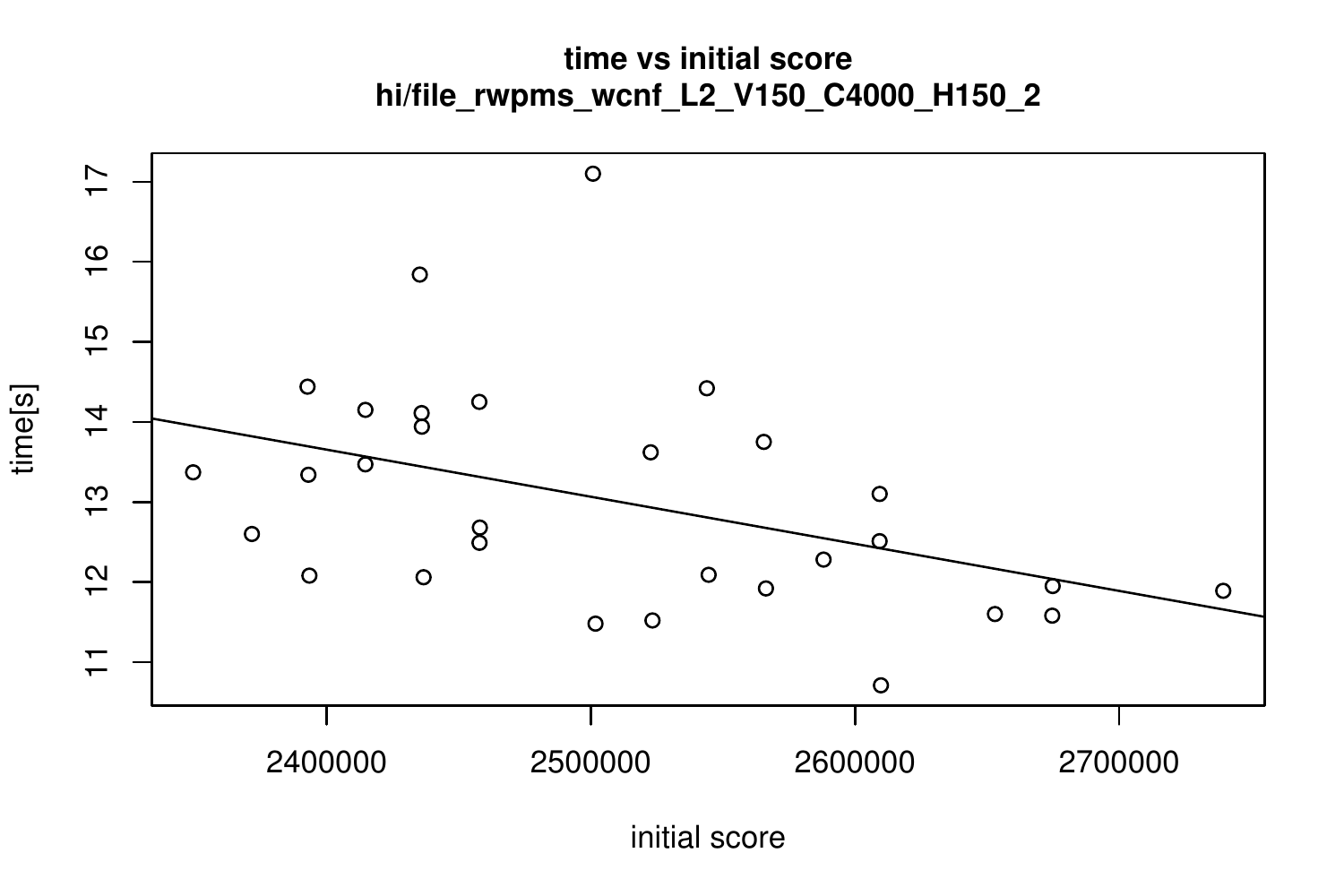}
    \label{fig_hi/file_rwpms_wcnf_L2_V150_C4000_H150_2/file_rwpms_wcnf_L2_V150_C4000_H150_2-time_vs_initial_score}
\end{figure}

\begin{figure}[H]
    \centering
    \includegraphics[height=3.5in]{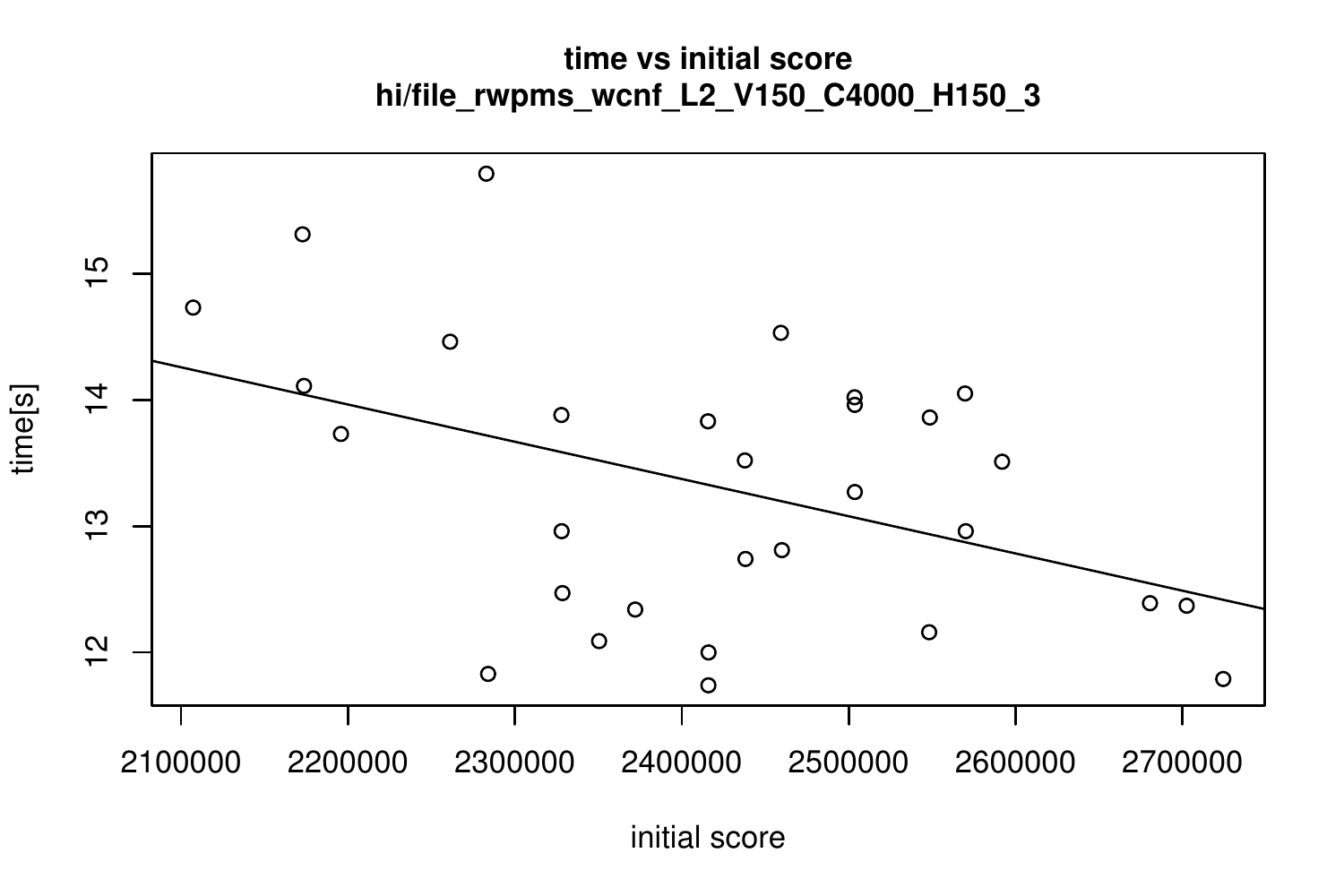}
    \label{fig_hi/file_rwpms_wcnf_L2_V150_C4000_H150_3/file_rwpms_wcnf_L2_V150_C4000_H150_3-time_vs_initial_score}
\end{figure}

\begin{figure}[H]
    \centering
    \includegraphics[height=3.5in]{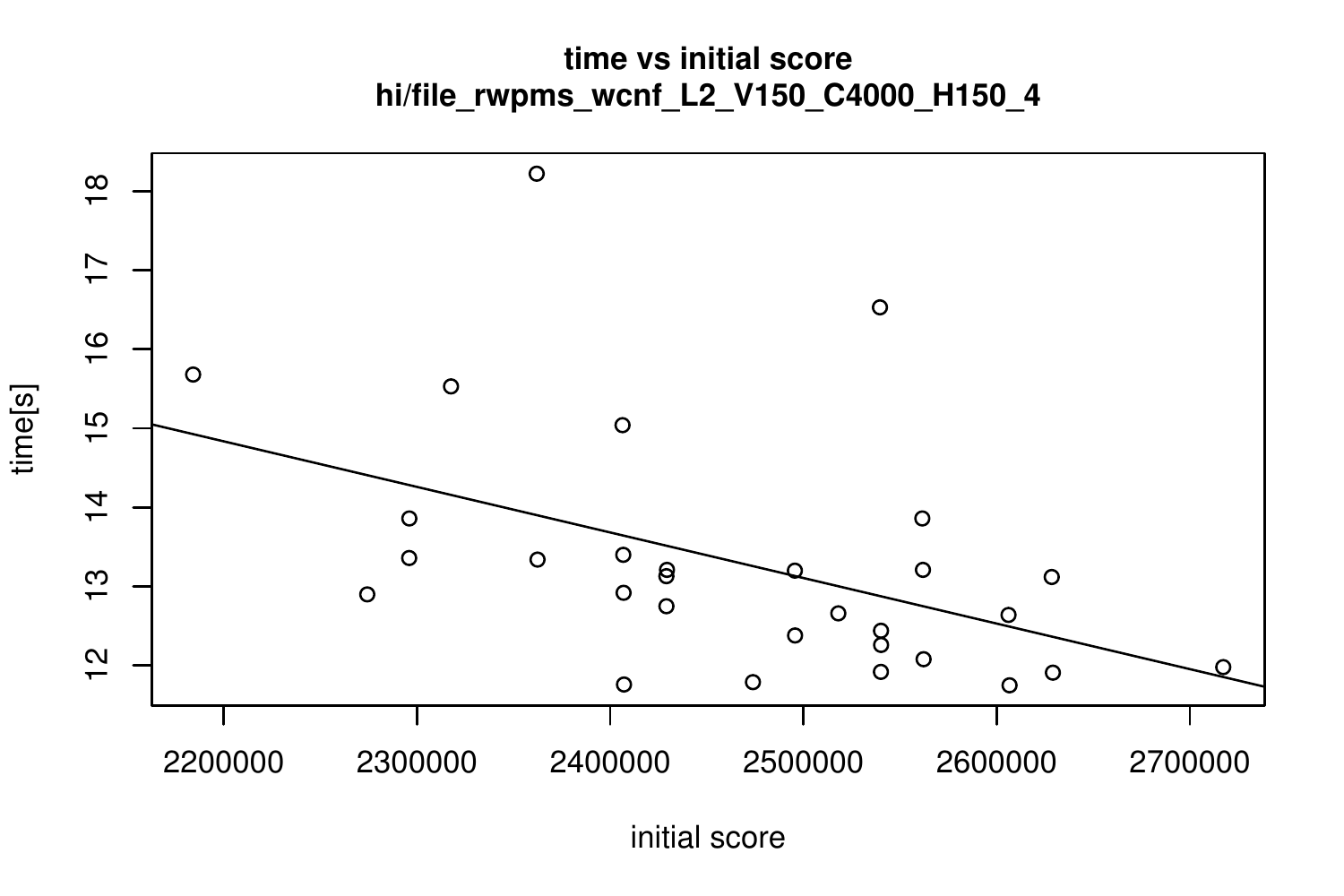}
    \label{fig_hi/file_rwpms_wcnf_L2_V150_C4000_H150_4/file_rwpms_wcnf_L2_V150_C4000_H150_4-time_vs_initial_score}
\end{figure}

\begin{figure}[H]
    \centering
    \includegraphics[height=3.5in]{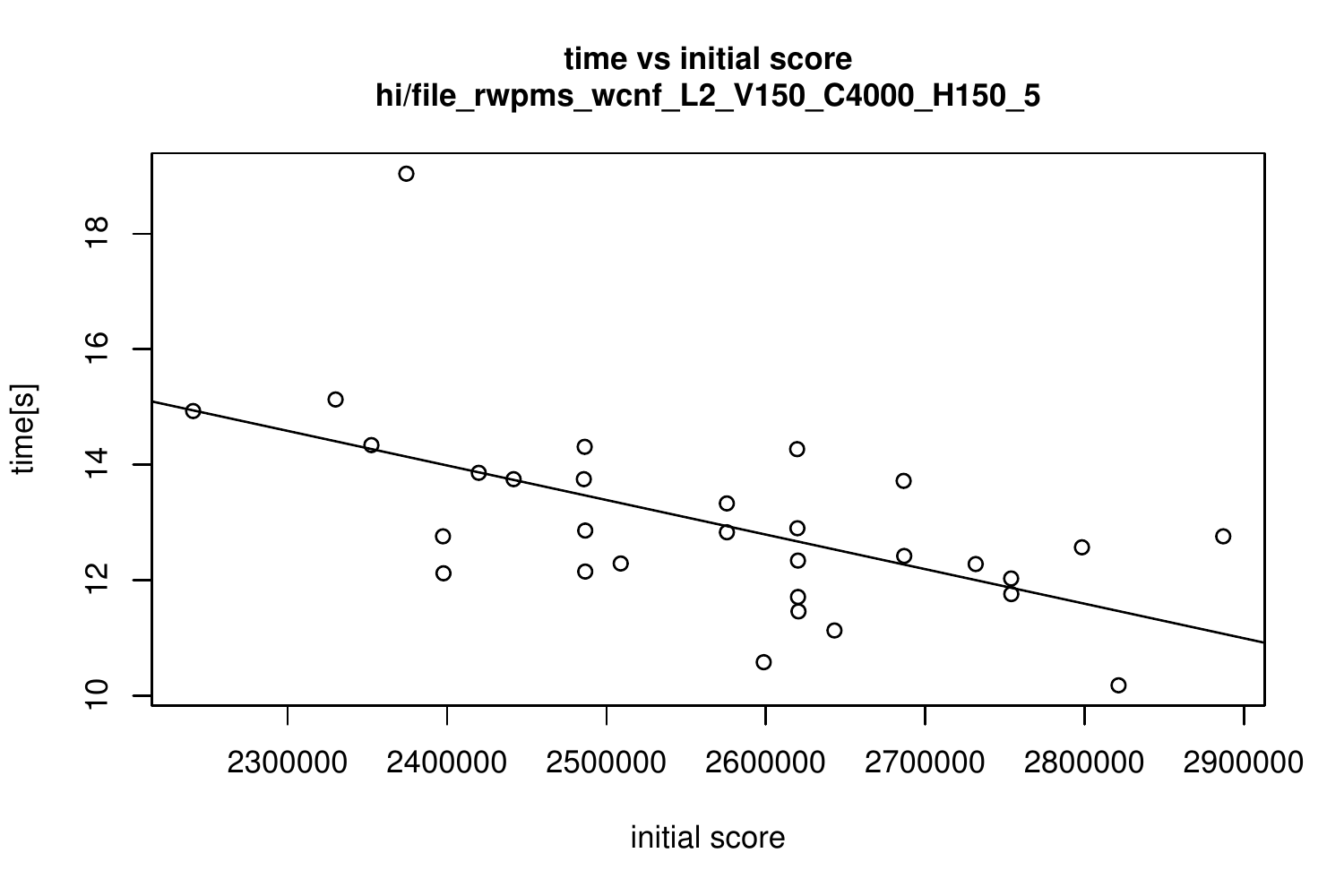}
    \label{fig_hi/file_rwpms_wcnf_L2_V150_C4000_H150_5/file_rwpms_wcnf_L2_V150_C4000_H150_5-time_vs_initial_score}
\end{figure}

\begin{figure}[H]
    \centering
    \includegraphics[height=3.5in]{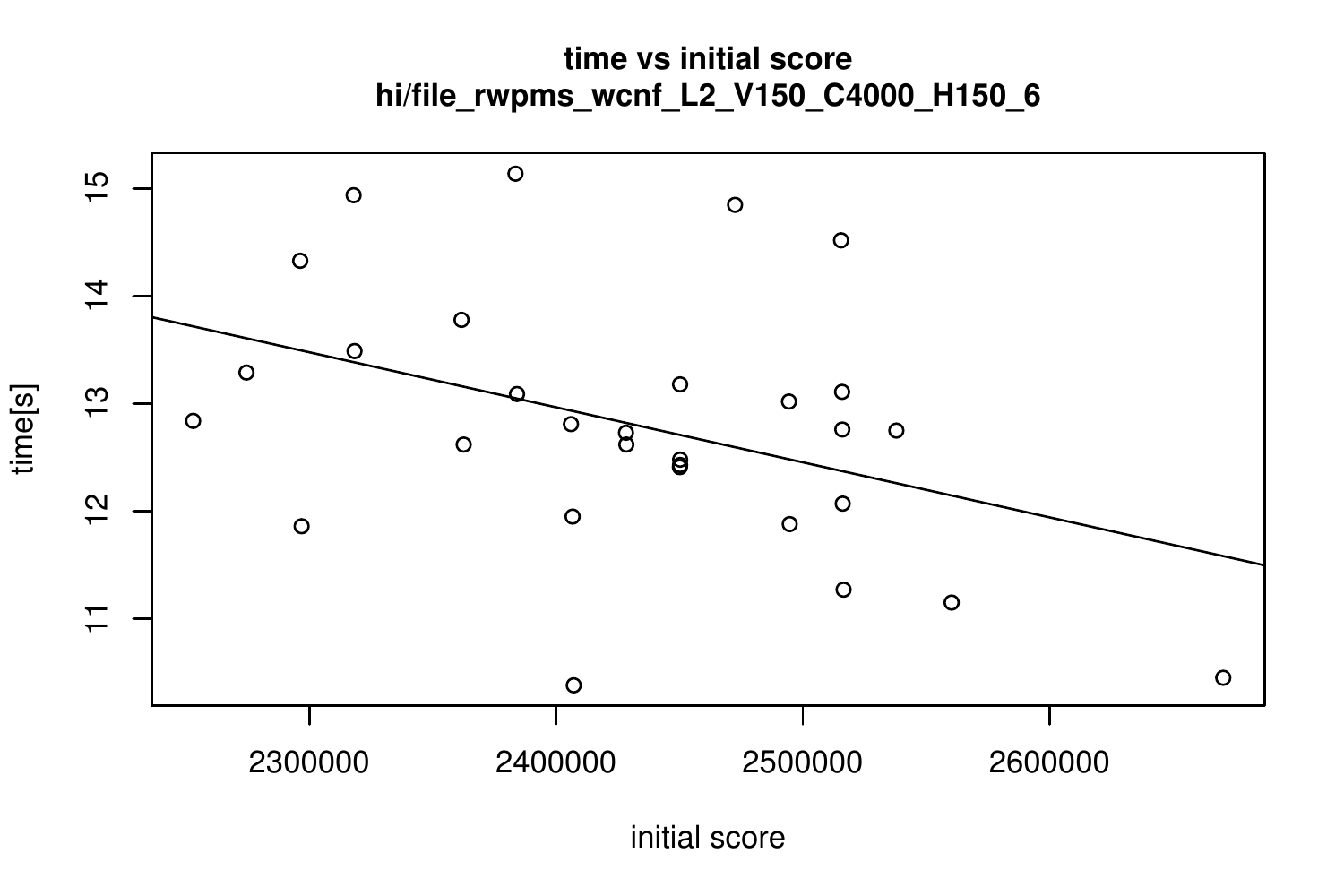}
    \label{fig_hi/file_rwpms_wcnf_L2_V150_C4000_H150_6/file_rwpms_wcnf_L2_V150_C4000_H150_6-time_vs_initial_score}
\end{figure}

\begin{figure}[H]
    \centering
    \includegraphics[height=3.5in]{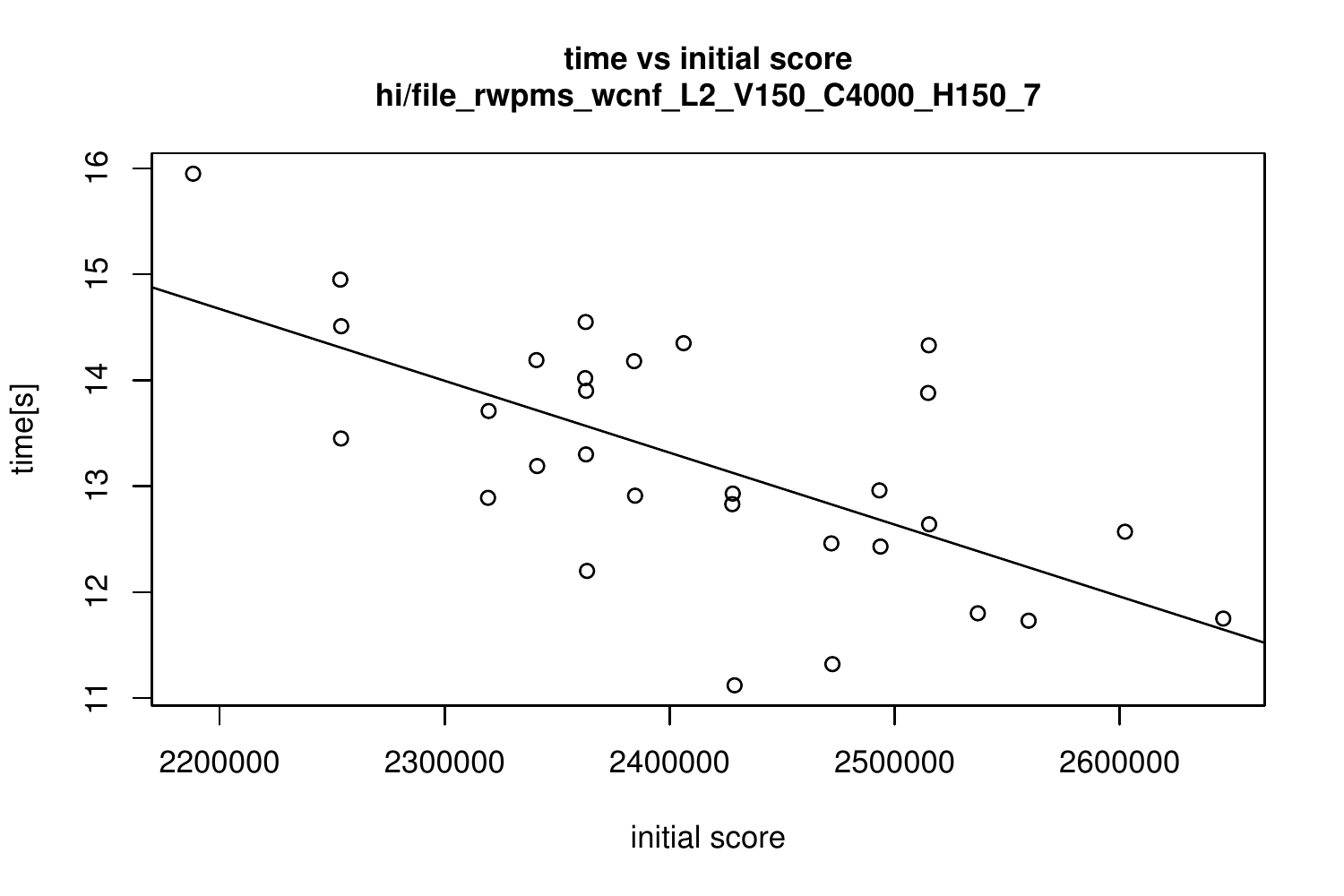}
    \label{fig_hi/file_rwpms_wcnf_L2_V150_C4000_H150_7/file_rwpms_wcnf_L2_V150_C4000_H150_7-time_vs_initial_score}
\end{figure}

\begin{figure}[H]
    \centering
    \includegraphics[height=3.5in]{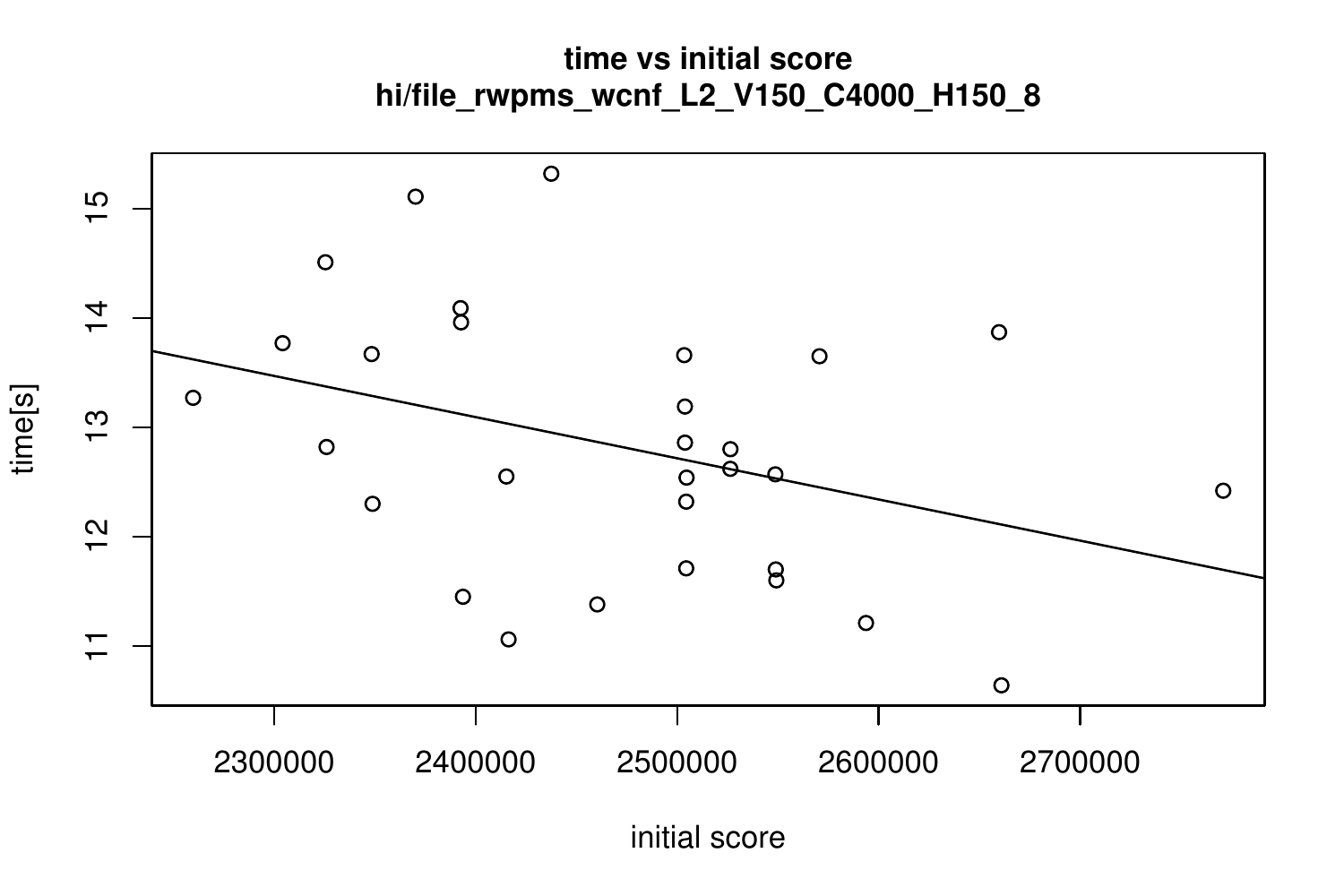}
    \label{fig_hi/file_rwpms_wcnf_L2_V150_C4000_H150_8/file_rwpms_wcnf_L2_V150_C4000_H150_8-time_vs_initial_score}
\end{figure}

\begin{figure}[H]
    \centering
    \includegraphics[height=3.5in]{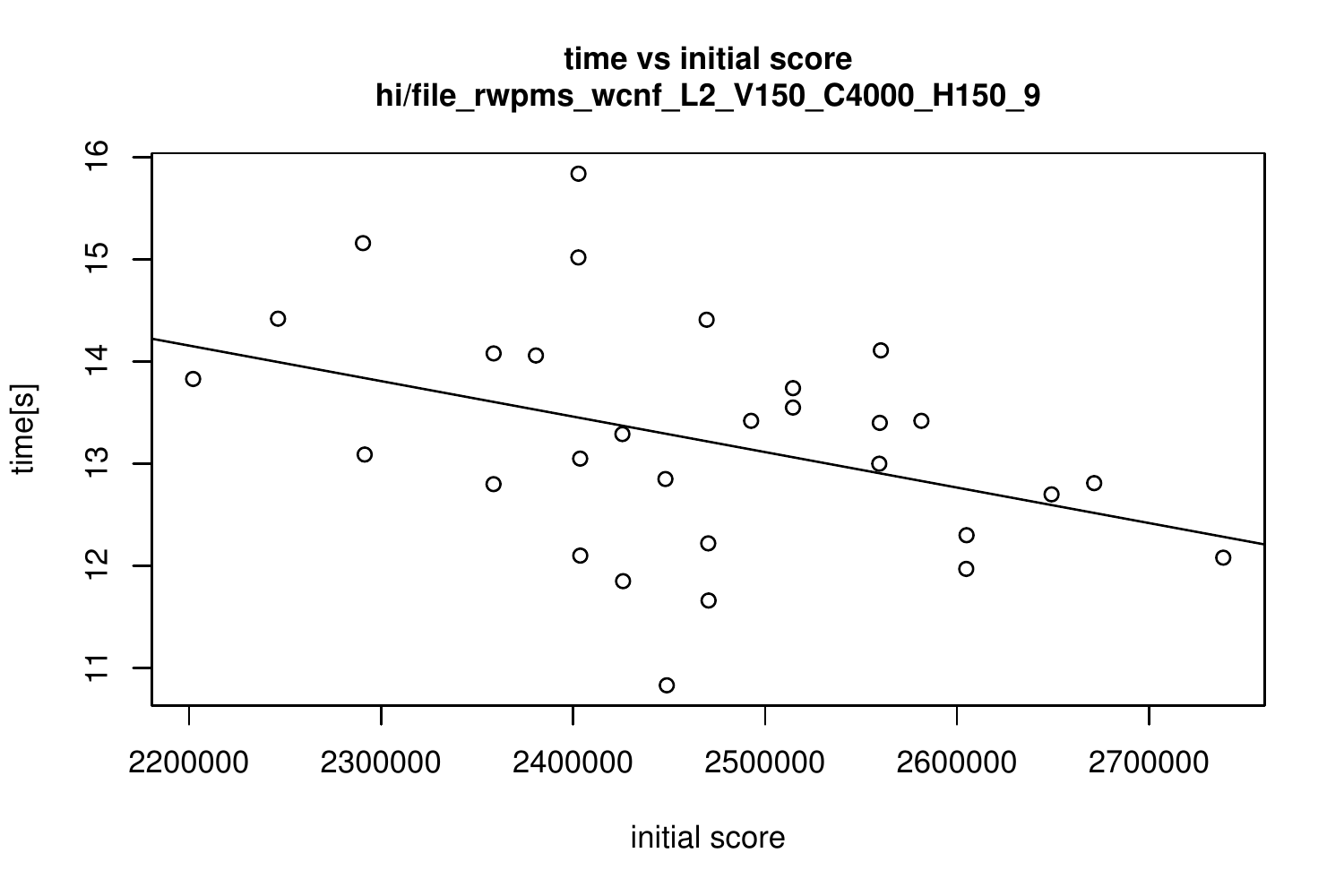}
    \label{fig_hi/file_rwpms_wcnf_L2_V150_C4000_H150_9/file_rwpms_wcnf_L2_V150_C4000_H150_9-time_vs_initial_score}
\end{figure}

\begin{figure}[H]
    \centering
    \includegraphics[height=3.5in]{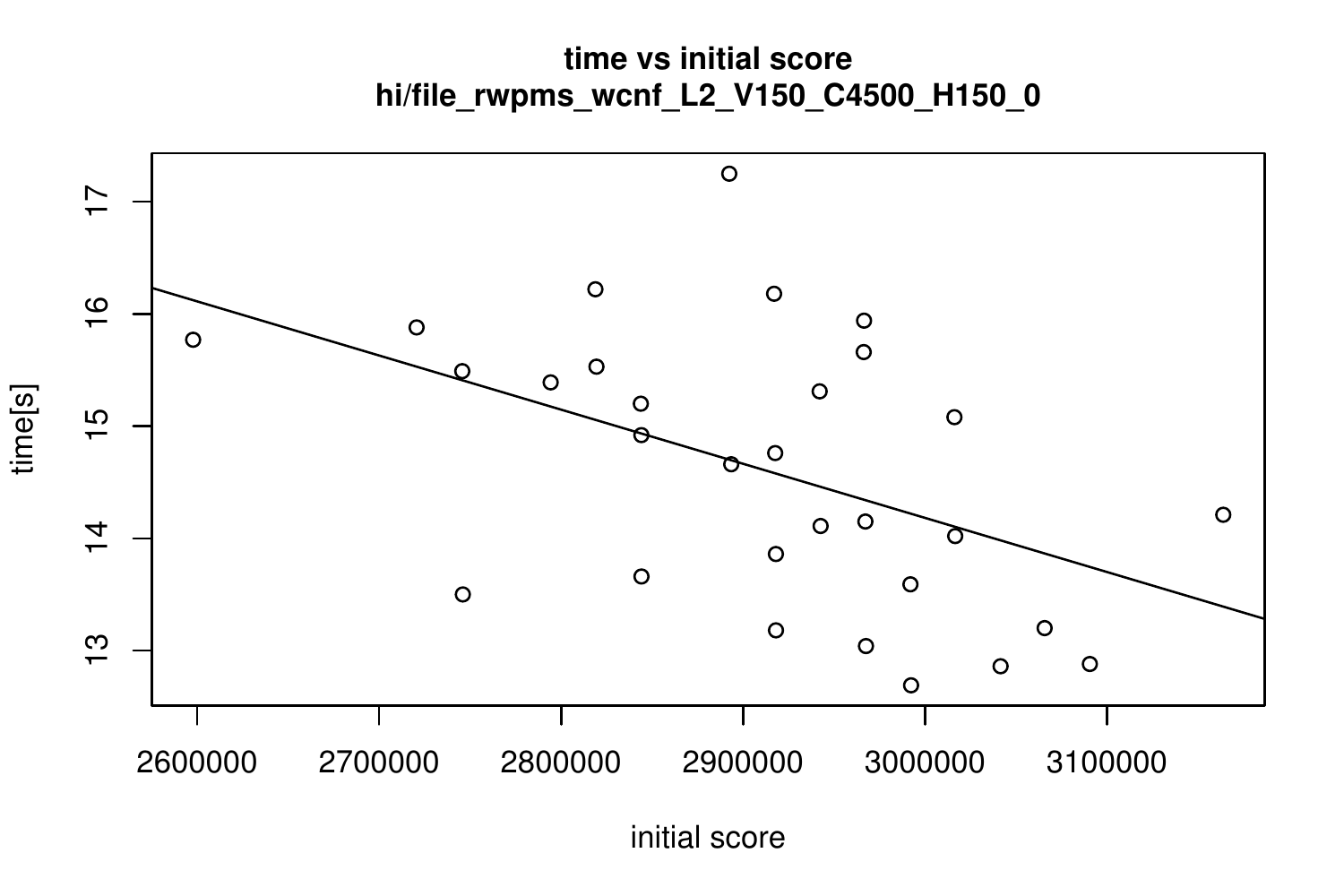}
    \label{fig_hi/file_rwpms_wcnf_L2_V150_C4500_H150_0/file_rwpms_wcnf_L2_V150_C4500_H150_0-time_vs_initial_score}
\end{figure}

\begin{figure}[H]
    \centering
    \includegraphics[height=3.5in]{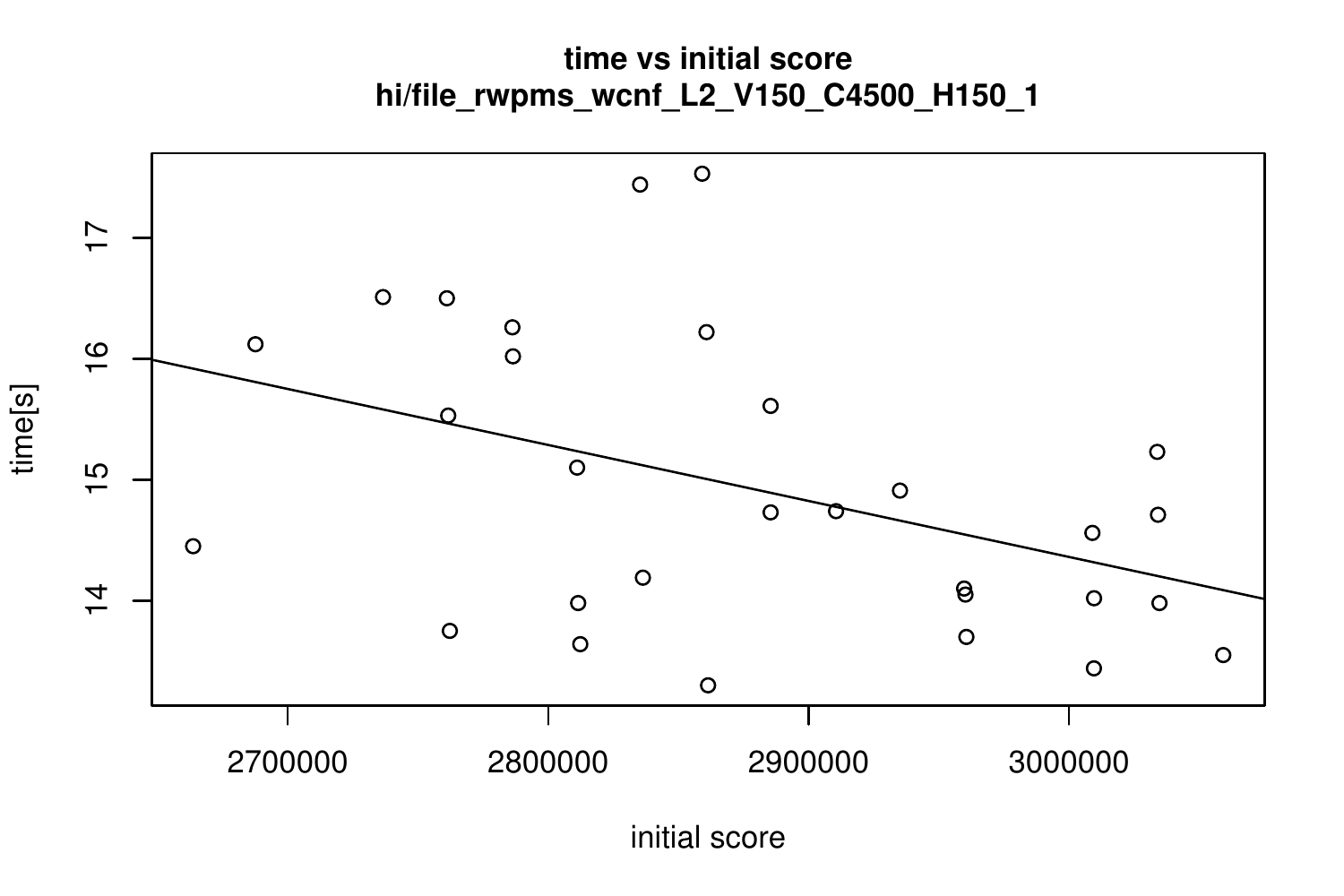}
    \label{fig_hi/file_rwpms_wcnf_L2_V150_C4500_H150_1/file_rwpms_wcnf_L2_V150_C4500_H150_1-time_vs_initial_score}
\end{figure}

\begin{figure}[H]
    \centering
    \includegraphics[height=3.5in]{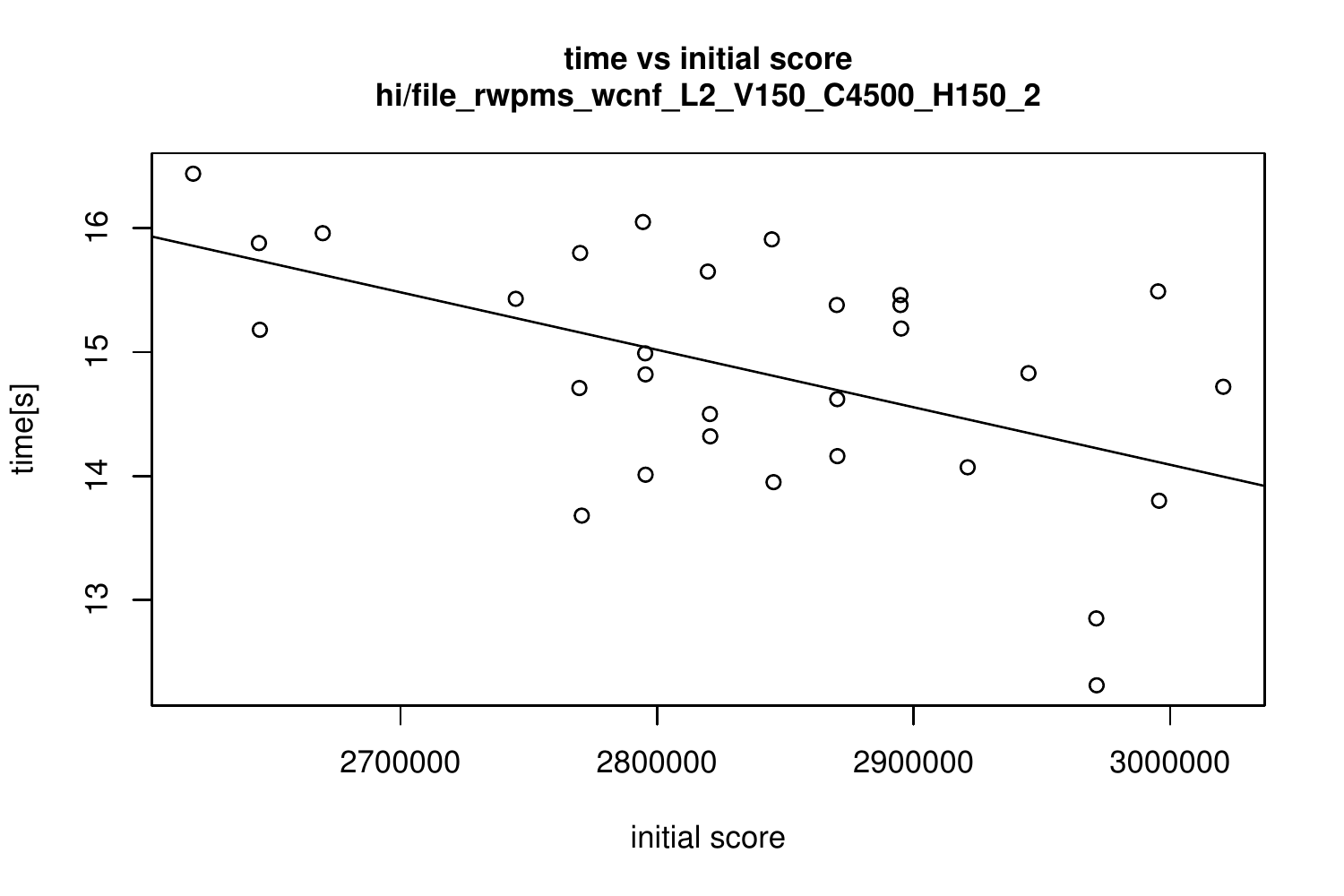}
    \label{fig_hi/file_rwpms_wcnf_L2_V150_C4500_H150_2/file_rwpms_wcnf_L2_V150_C4500_H150_2-time_vs_initial_score}
\end{figure}

\begin{figure}[H]
    \centering
    \includegraphics[height=3.5in]{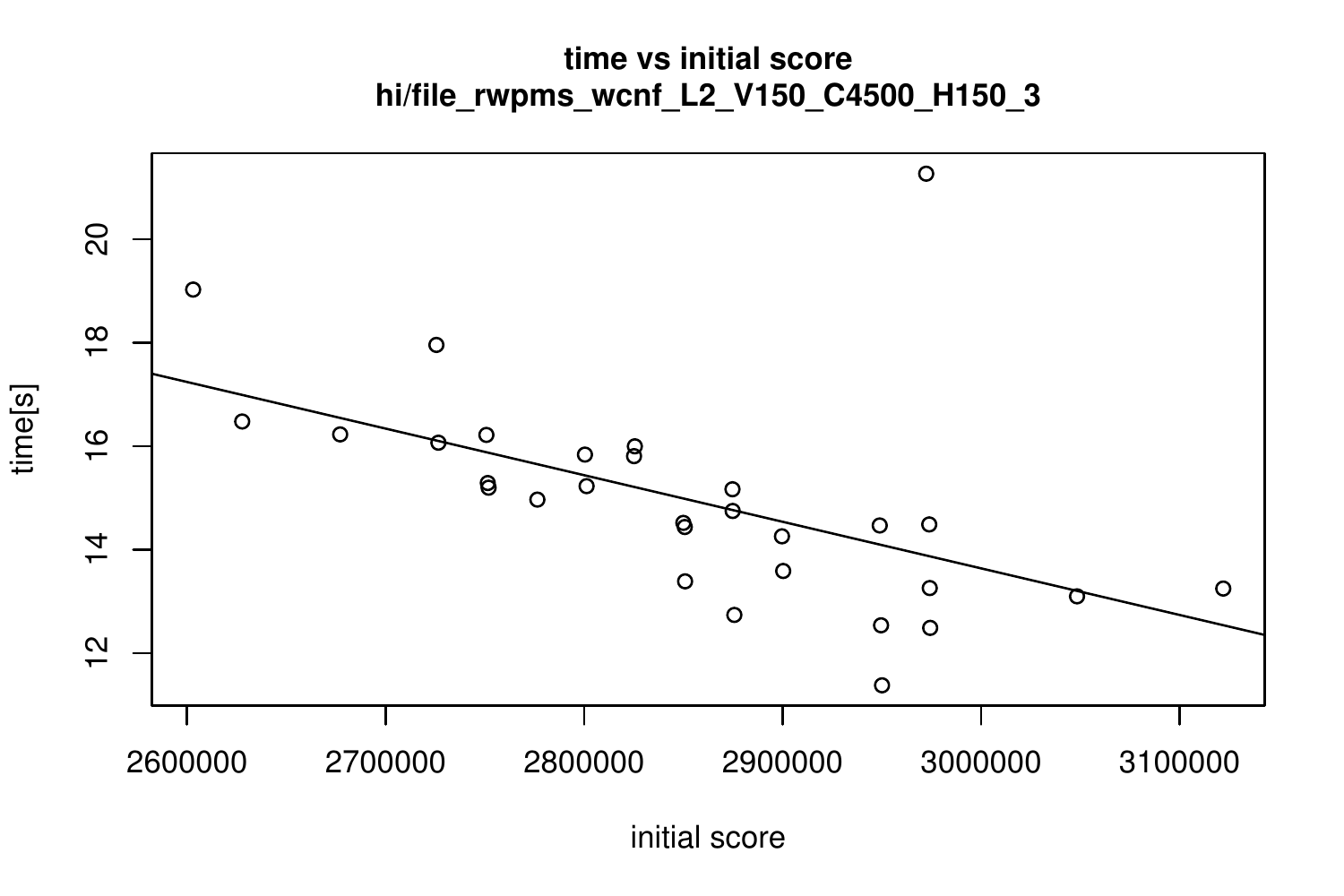}
    \label{fig_hi/file_rwpms_wcnf_L2_V150_C4500_H150_3/file_rwpms_wcnf_L2_V150_C4500_H150_3-time_vs_initial_score}
\end{figure}

\begin{figure}[H]
    \centering
    \includegraphics[height=3.5in]{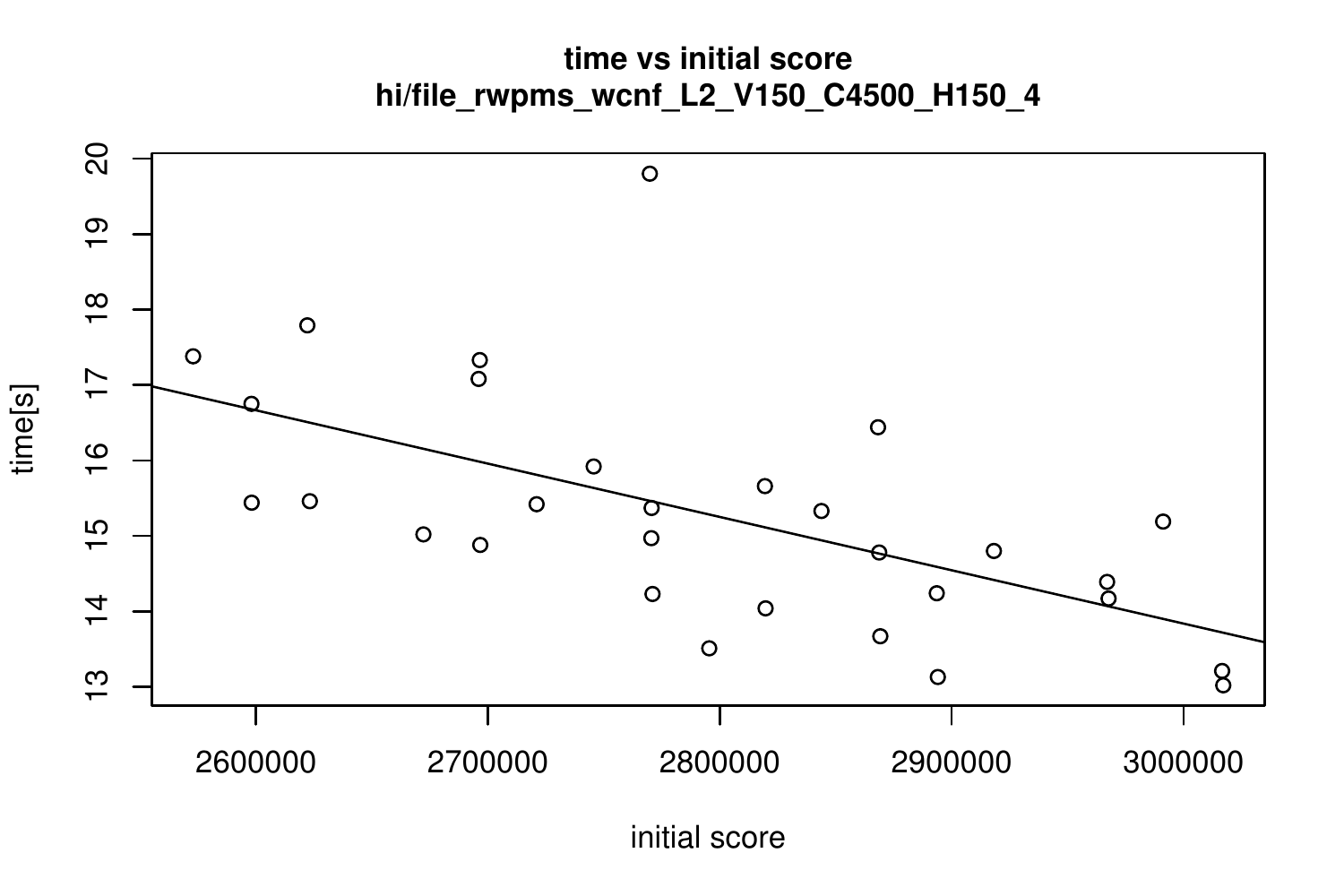}
    \label{fig_hi/file_rwpms_wcnf_L2_V150_C4500_H150_4/file_rwpms_wcnf_L2_V150_C4500_H150_4-time_vs_initial_score}
\end{figure}

\begin{figure}[H]
    \centering
    \includegraphics[height=3.5in]{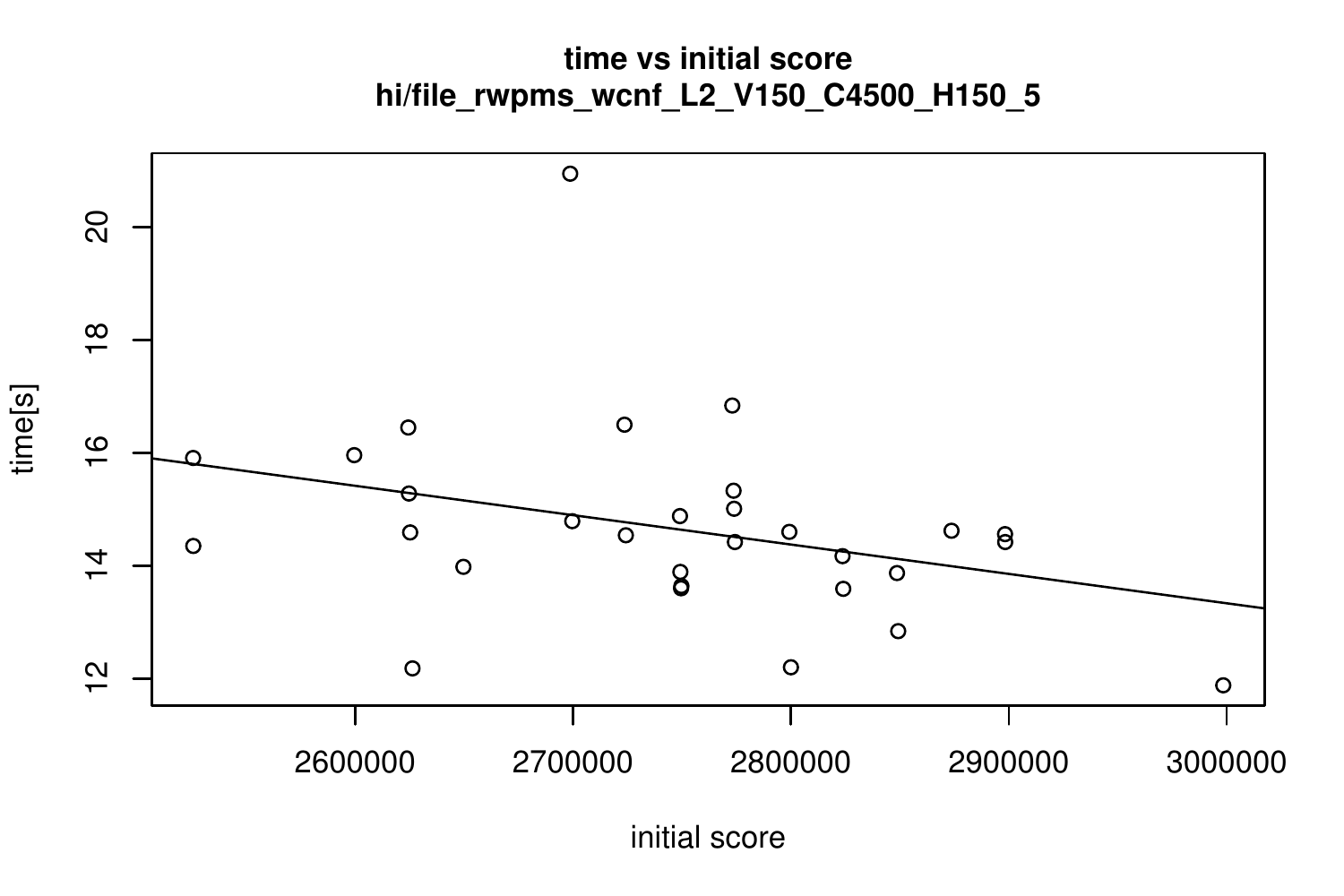}
    \label{fig_hi/file_rwpms_wcnf_L2_V150_C4500_H150_5/file_rwpms_wcnf_L2_V150_C4500_H150_5-time_vs_initial_score}
\end{figure}

\begin{figure}[H]
    \centering
    \includegraphics[height=3.5in]{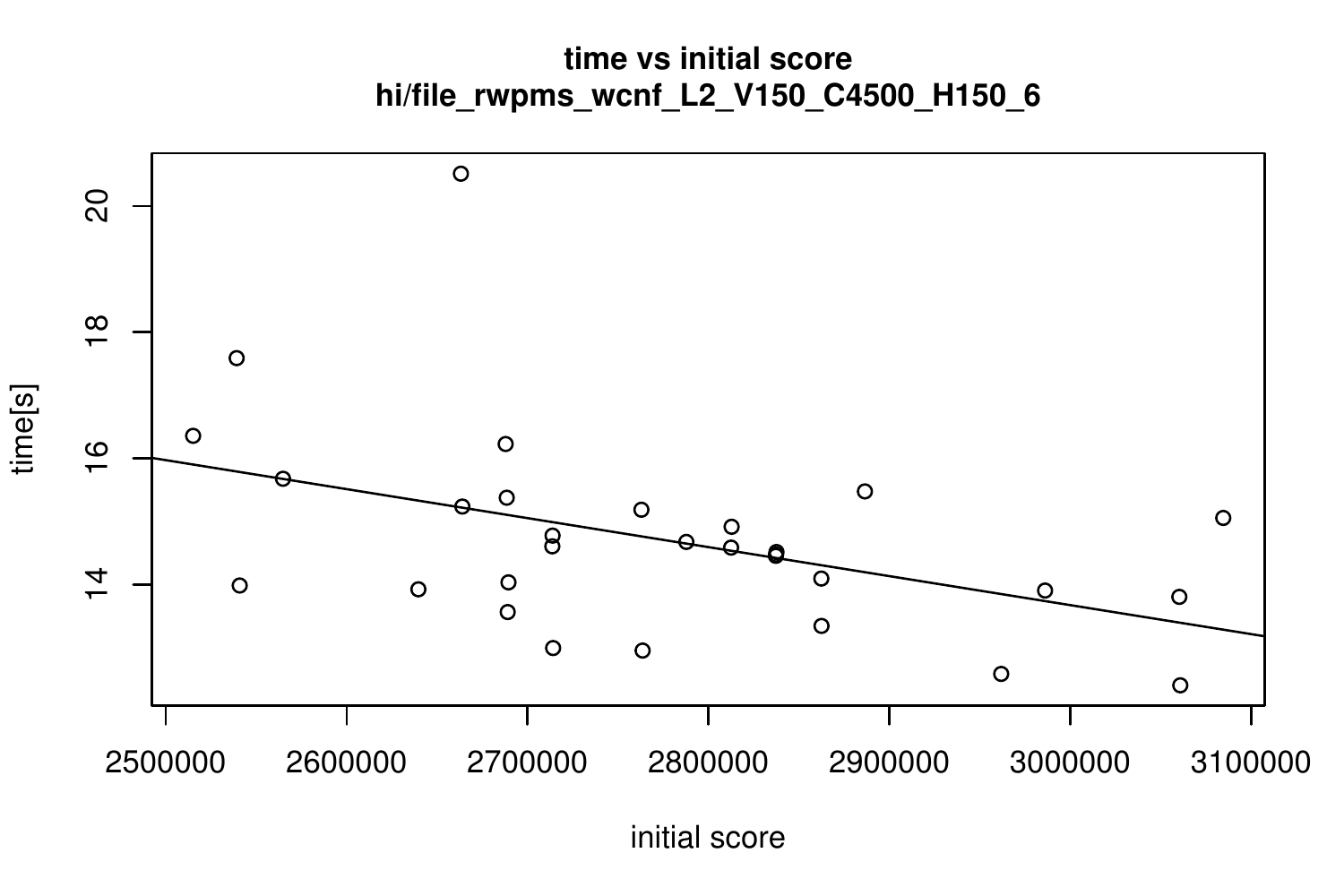}
    \label{fig_hi/file_rwpms_wcnf_L2_V150_C4500_H150_6/file_rwpms_wcnf_L2_V150_C4500_H150_6-time_vs_initial_score}
\end{figure}

\begin{figure}[H]
    \centering
    \includegraphics[height=3.5in]{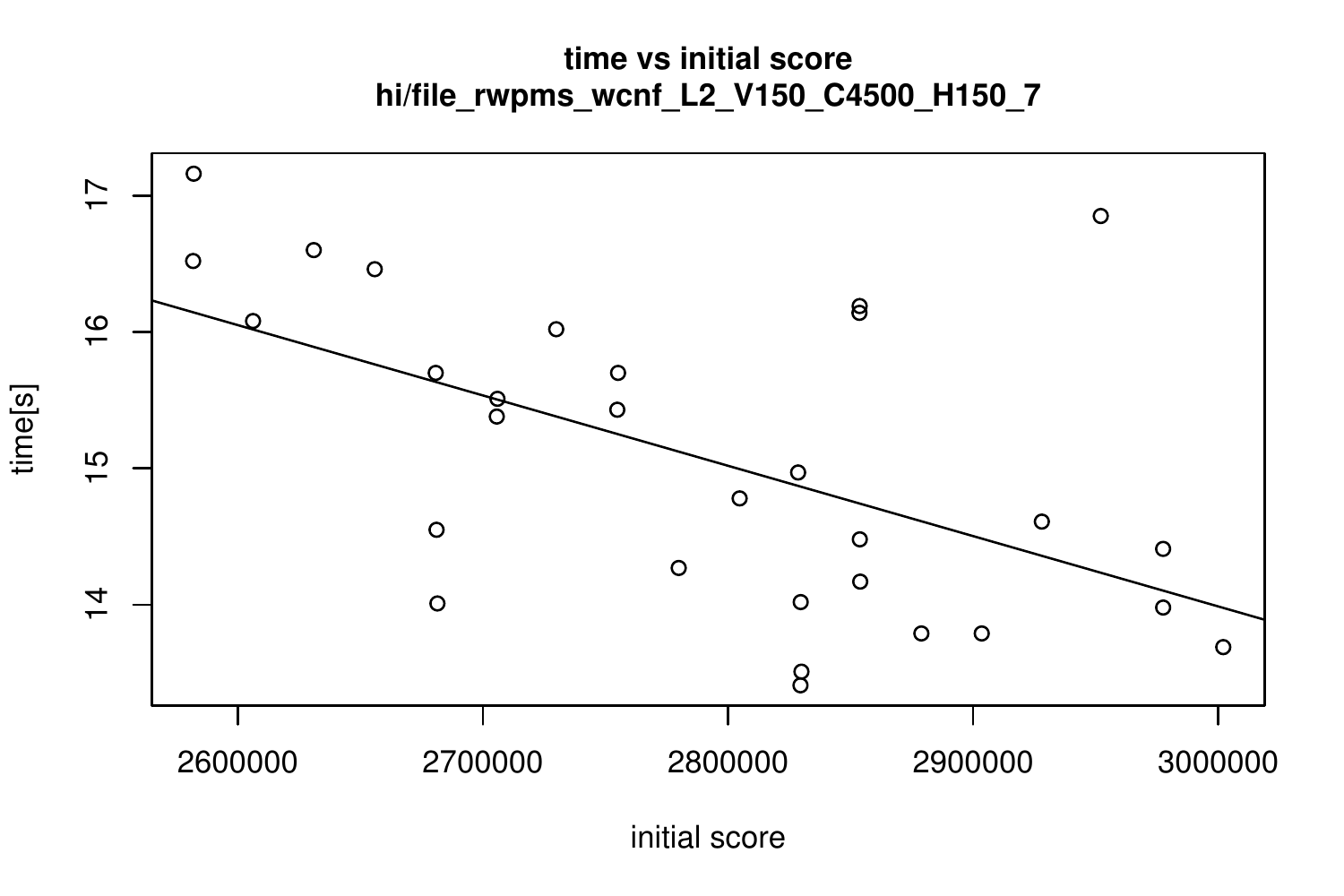}
    \label{fig_hi/file_rwpms_wcnf_L2_V150_C4500_H150_7/file_rwpms_wcnf_L2_V150_C4500_H150_7-time_vs_initial_score}
\end{figure}

\begin{figure}[H]
    \centering
    \includegraphics[height=3.5in]{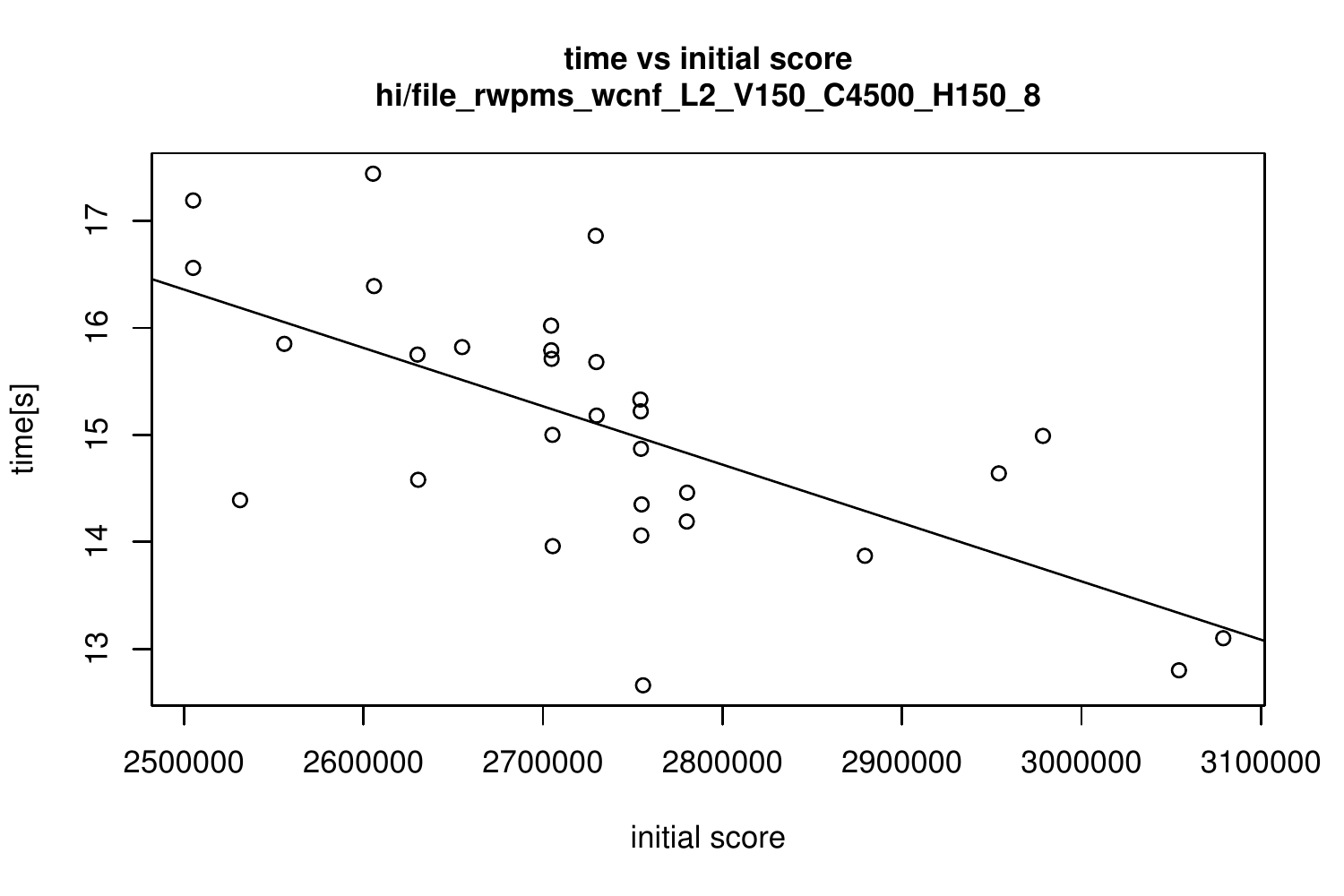}
    \label{fig_hi/file_rwpms_wcnf_L2_V150_C4500_H150_8/file_rwpms_wcnf_L2_V150_C4500_H150_8-time_vs_initial_score}
\end{figure}

\begin{figure}[H]
    \centering
    \includegraphics[height=3.5in]{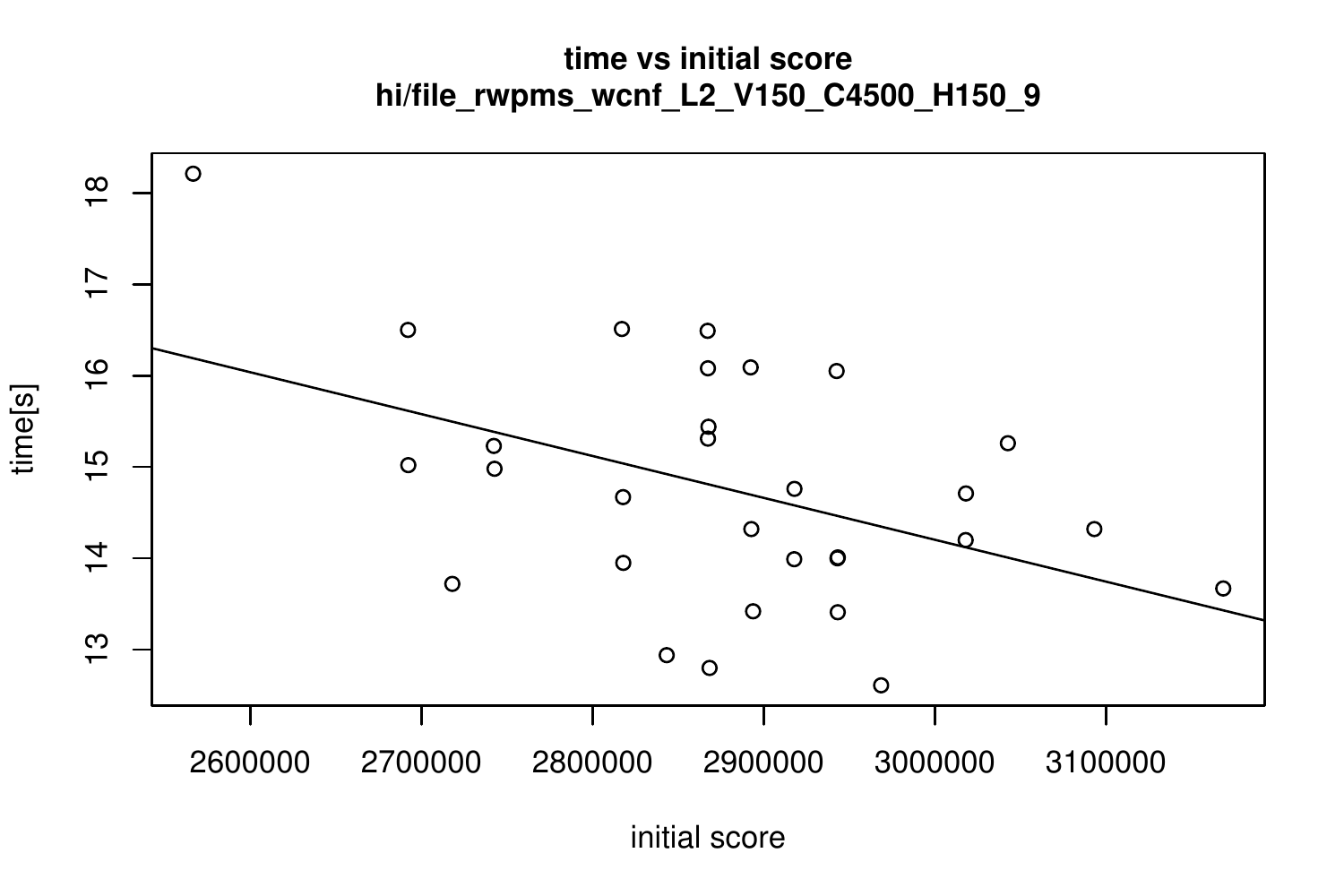}
    \label{fig_hi/file_rwpms_wcnf_L2_V150_C4500_H150_9/file_rwpms_wcnf_L2_V150_C4500_H150_9-time_vs_initial_score}
\end{figure}

\begin{figure}[H]
    \centering
    \includegraphics[height=3.5in]{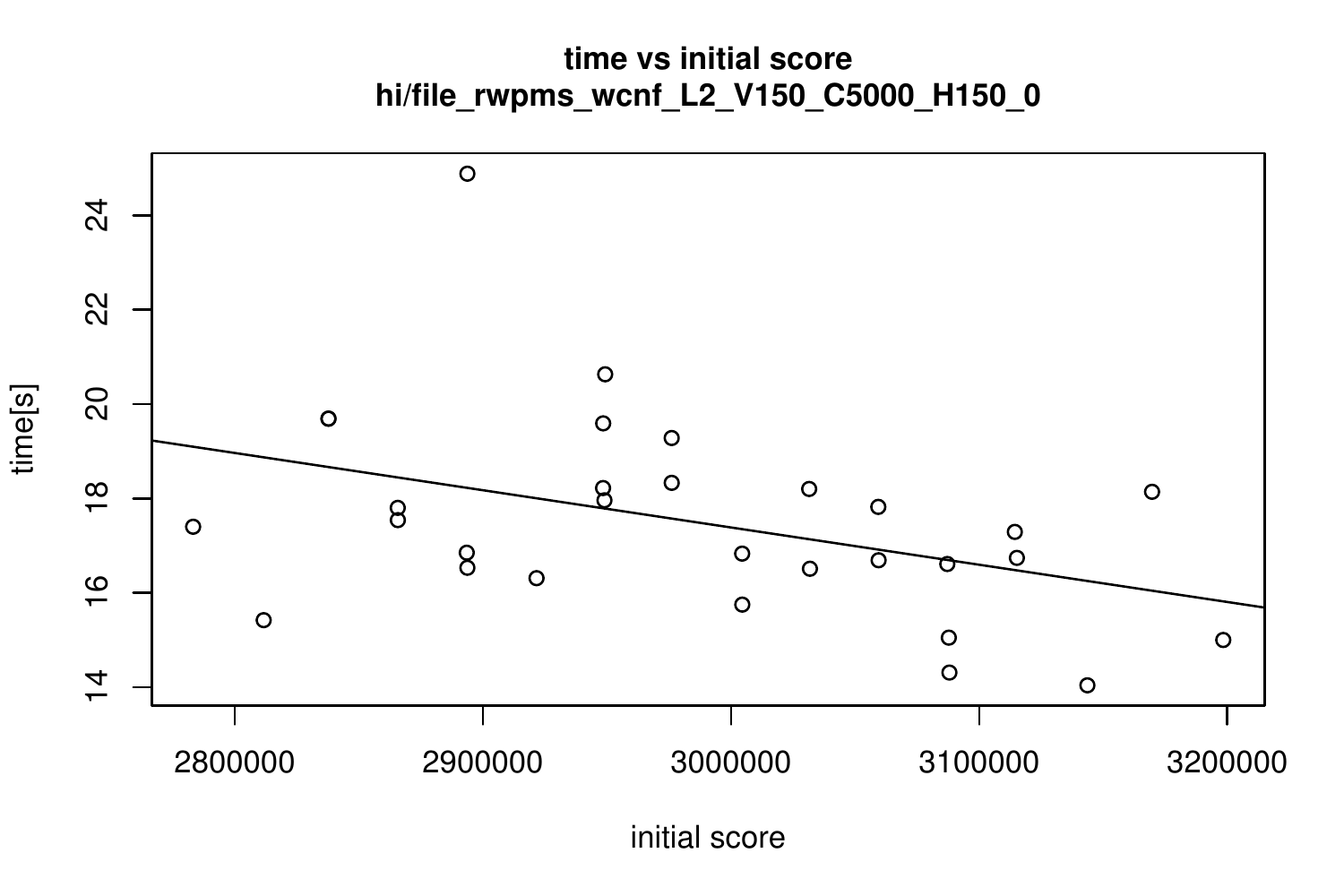}
    \label{fig_hi/file_rwpms_wcnf_L2_V150_C5000_H150_0/file_rwpms_wcnf_L2_V150_C5000_H150_0-time_vs_initial_score}
\end{figure}

\begin{figure}[H]
    \centering
    \includegraphics[height=3.5in]{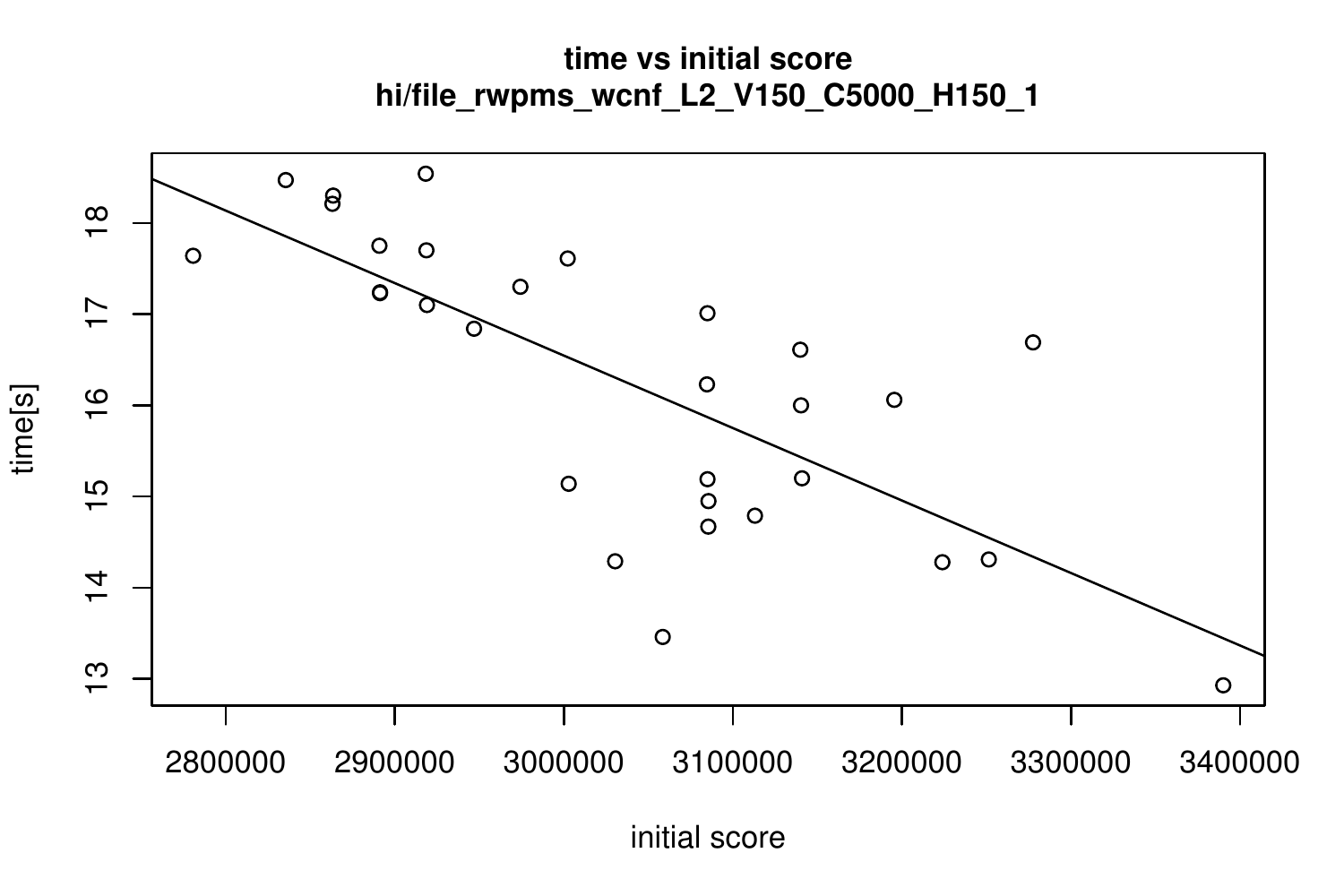}
    \label{fig_hi/file_rwpms_wcnf_L2_V150_C5000_H150_1/file_rwpms_wcnf_L2_V150_C5000_H150_1-time_vs_initial_score}
\end{figure}

\begin{figure}[H]
    \centering
    \includegraphics[height=3.5in]{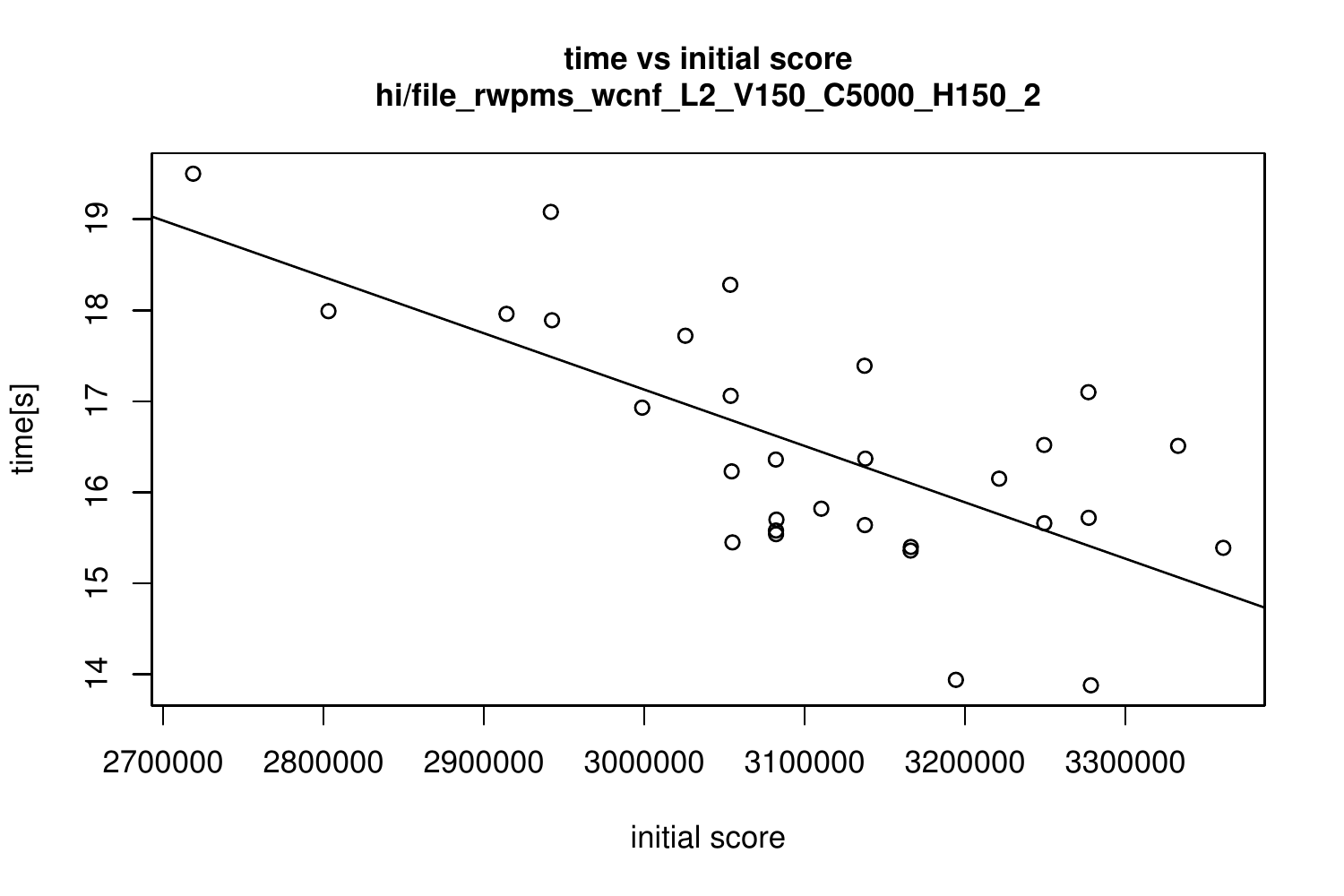}
    \label{fig_hi/file_rwpms_wcnf_L2_V150_C5000_H150_2/file_rwpms_wcnf_L2_V150_C5000_H150_2-time_vs_initial_score}
\end{figure}

\begin{figure}[H]
    \centering
    \includegraphics[height=3.5in]{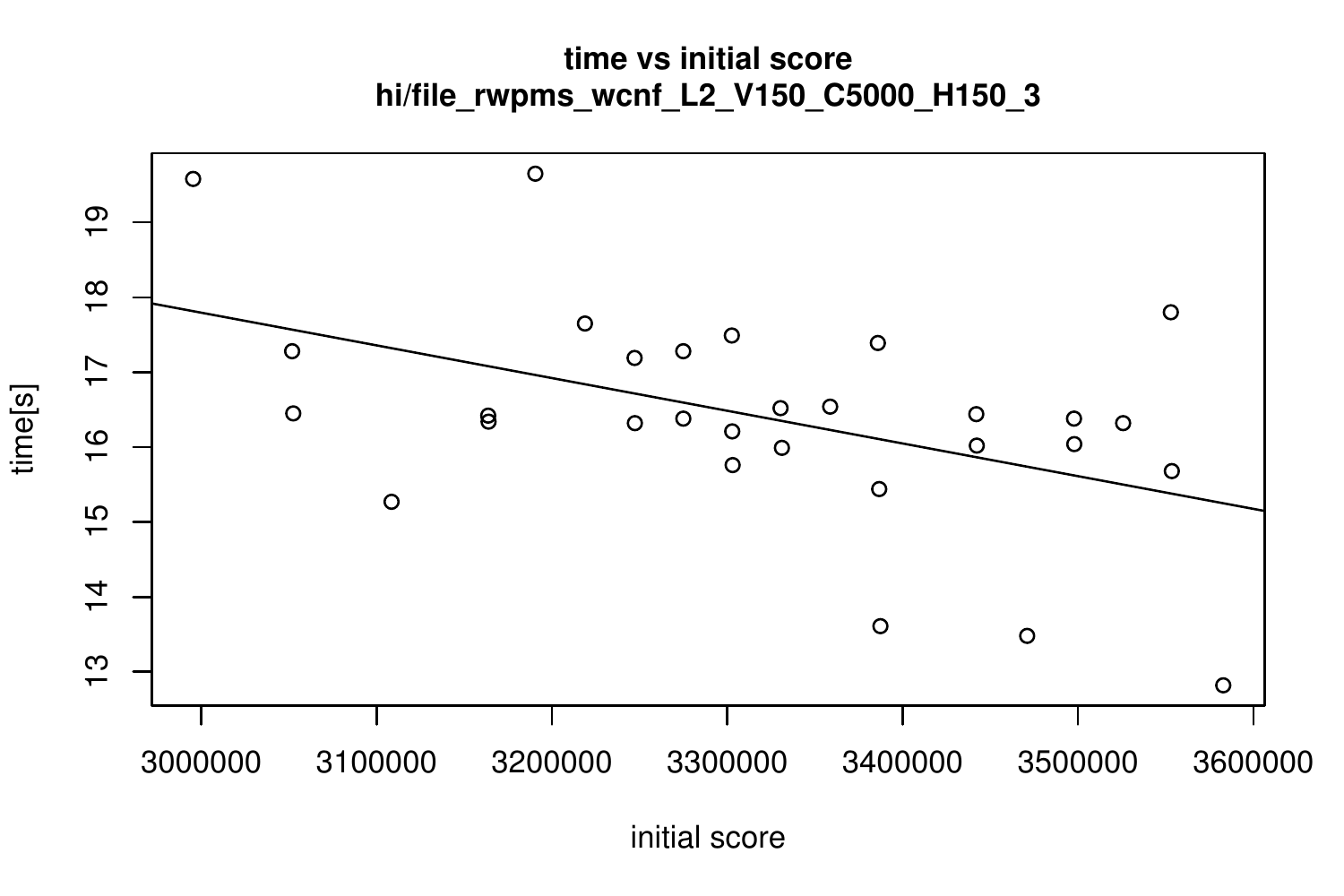}
    \label{fig_hi/file_rwpms_wcnf_L2_V150_C5000_H150_3/file_rwpms_wcnf_L2_V150_C5000_H150_3-time_vs_initial_score}
\end{figure}

\begin{figure}[H]
    \centering
    \includegraphics[height=3.5in]{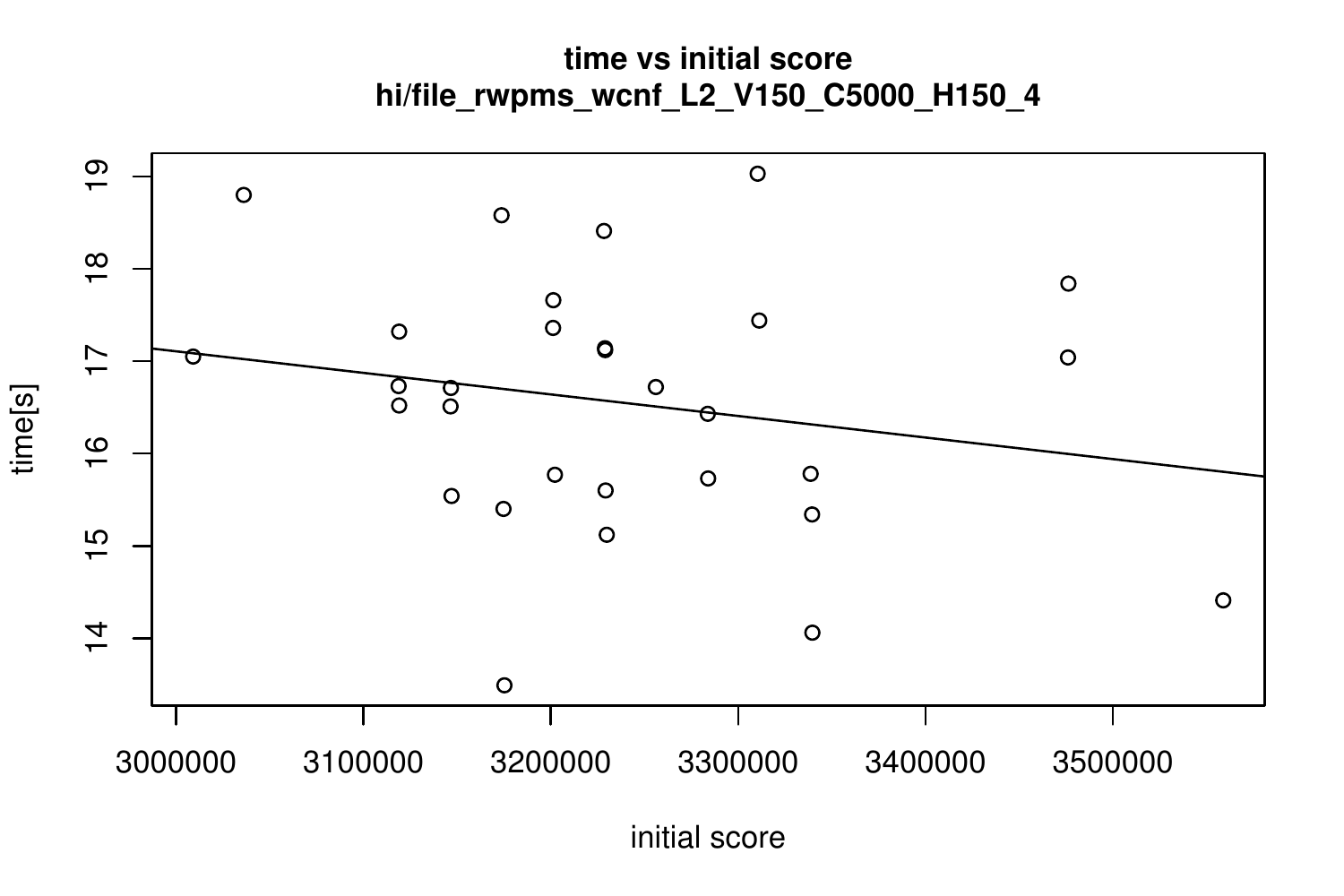}
    \label{fig_hi/file_rwpms_wcnf_L2_V150_C5000_H150_4/file_rwpms_wcnf_L2_V150_C5000_H150_4-time_vs_initial_score}
\end{figure}

\begin{figure}[H]
    \centering
    \includegraphics[height=3.5in]{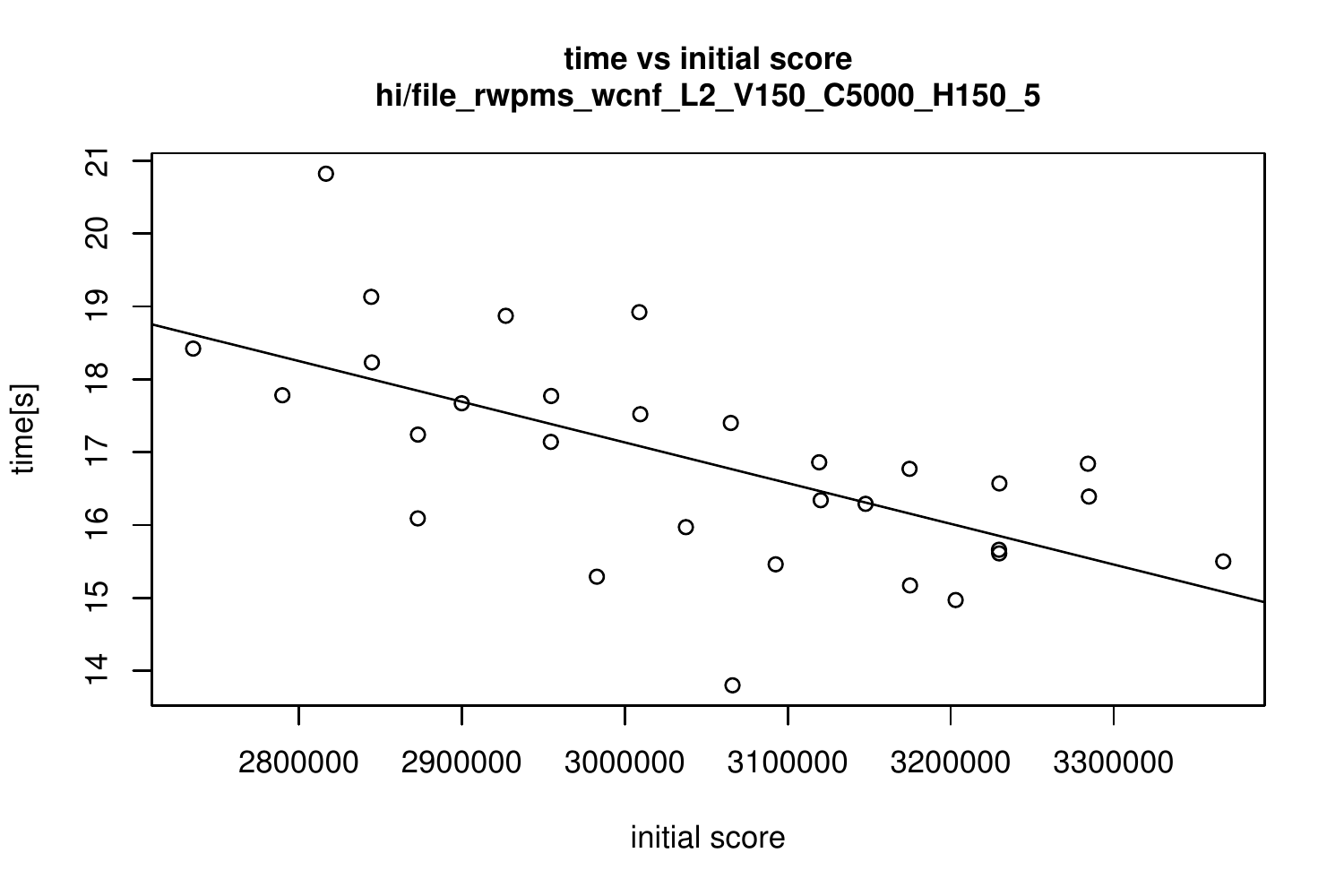}
    \label{fig_hi/file_rwpms_wcnf_L2_V150_C5000_H150_5/file_rwpms_wcnf_L2_V150_C5000_H150_5-time_vs_initial_score}
\end{figure}

\begin{figure}[H]
    \centering
    \includegraphics[height=3.5in]{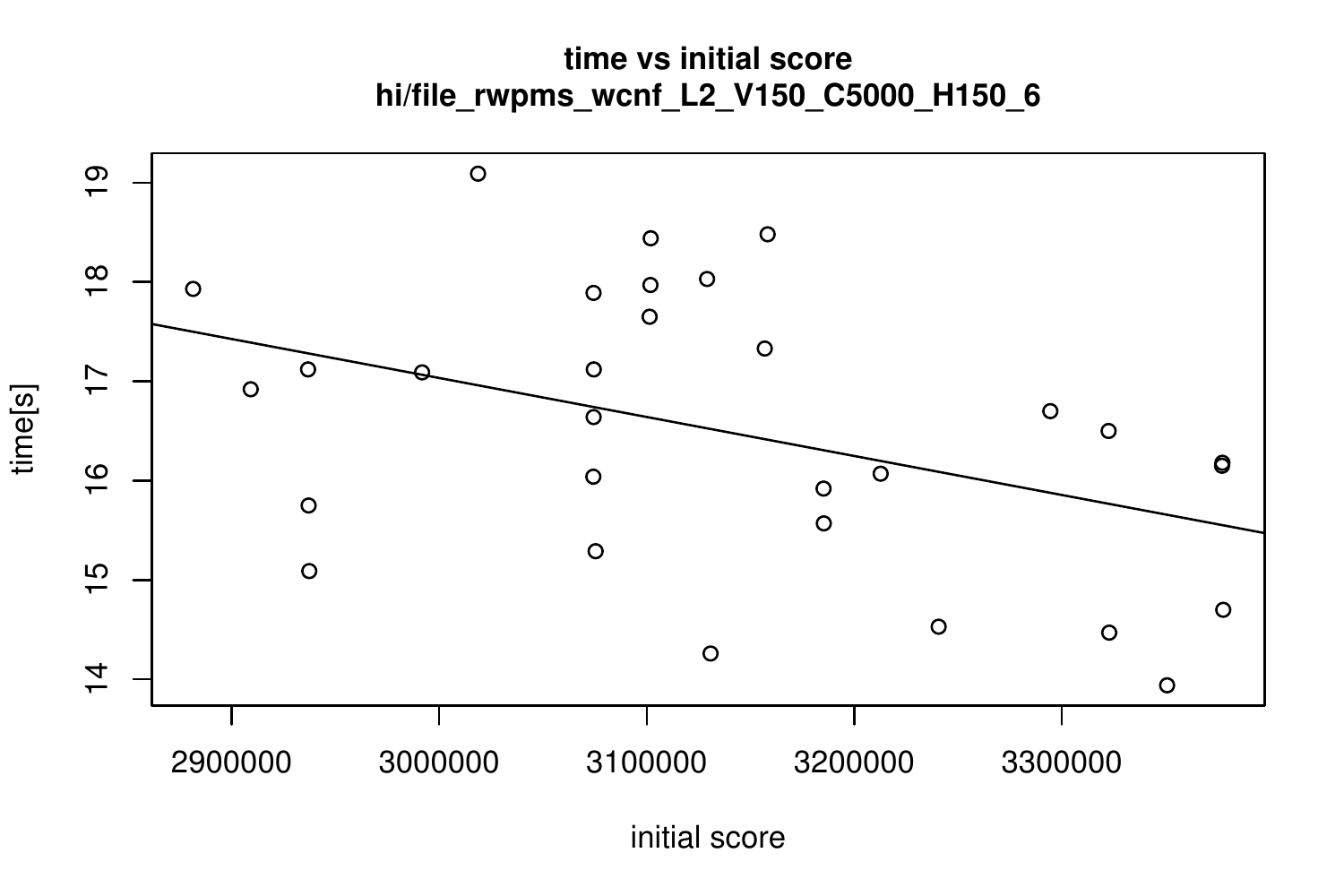}
    \label{fig_hi/file_rwpms_wcnf_L2_V150_C5000_H150_6/file_rwpms_wcnf_L2_V150_C5000_H150_6-time_vs_initial_score}
\end{figure}

\begin{figure}[H]
    \centering
    \includegraphics[height=3.5in]{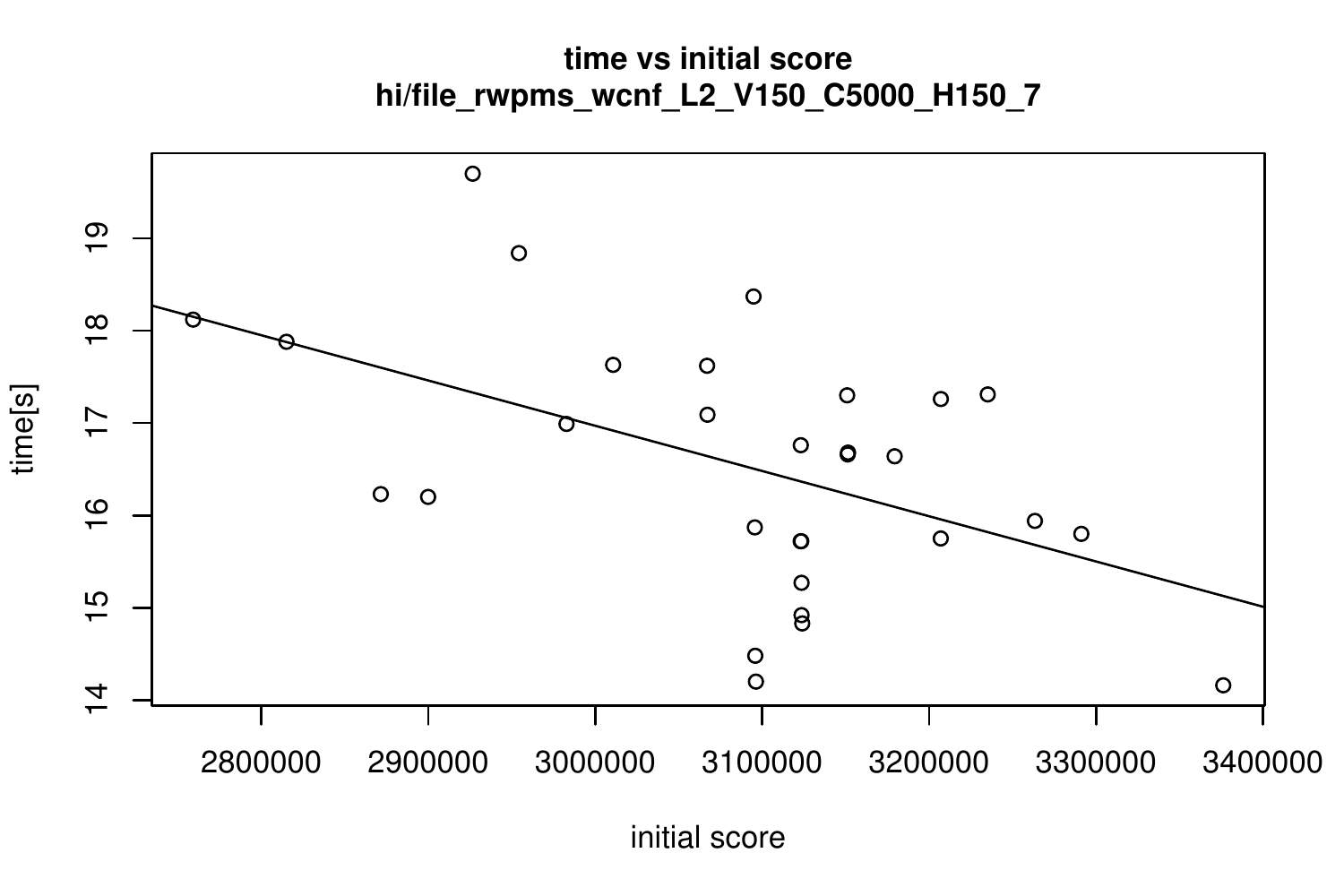}
    \label{fig_hi/file_rwpms_wcnf_L2_V150_C5000_H150_7/file_rwpms_wcnf_L2_V150_C5000_H150_7-time_vs_initial_score}
\end{figure}

\begin{figure}[H]
    \centering
    \includegraphics[height=3.5in]{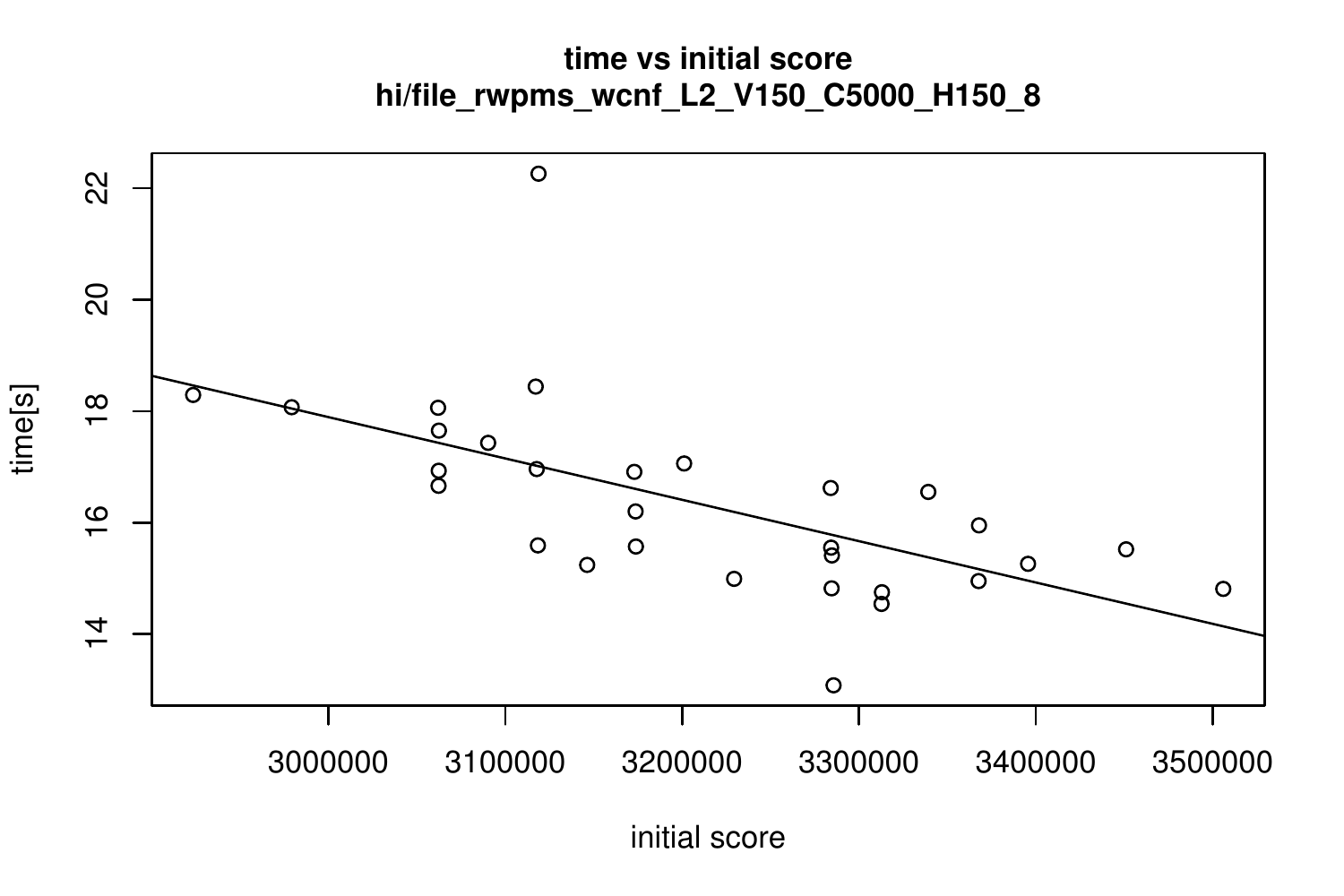}
    \label{fig_hi/file_rwpms_wcnf_L2_V150_C5000_H150_8/file_rwpms_wcnf_L2_V150_C5000_H150_8-time_vs_initial_score}
\end{figure}

\begin{figure}[H]
    \centering
    \includegraphics[height=3.5in]{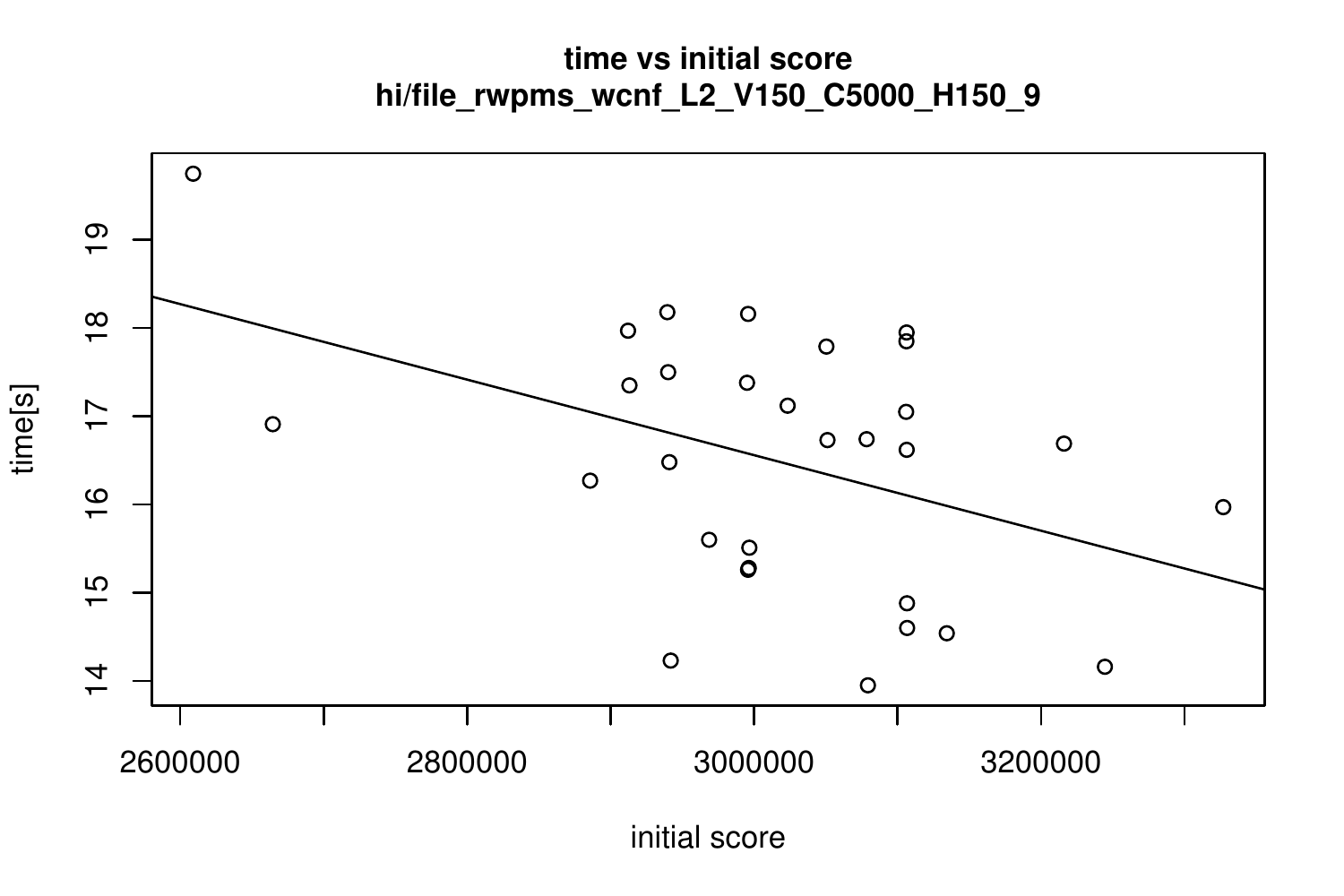}
    \label{fig_hi/file_rwpms_wcnf_L2_V150_C5000_H150_9/file_rwpms_wcnf_L2_V150_C5000_H150_9-time_vs_initial_score}
\end{figure}

\chapter{MaxWalkSat Runtimes\label{maxwalksat_runtimes}}
\begin{center}
    \begin{longtable}{|l|c|}
        \caption{Mean run times for MaxWalkSat over 30 runs} \\
        \hline
        \textbf{Problem} & \textbf{Time[s]} \\
        \hline
        \hline
        lo/file\_rwpms\_wcnf\_L2\_V150\_C1000\_H150\_0 & 2.975 \\
        lo/file\_rwpms\_wcnf\_L2\_V150\_C1000\_H150\_1 & 3.074 \\
        lo/file\_rwpms\_wcnf\_L2\_V150\_C1000\_H150\_2 & 2.876 \\
        lo/file\_rwpms\_wcnf\_L2\_V150\_C1000\_H150\_3 & 3.101 \\
        lo/file\_rwpms\_wcnf\_L2\_V150\_C1000\_H150\_4 & 2.999 \\
        lo/file\_rwpms\_wcnf\_L2\_V150\_C1000\_H150\_5 & 2.958 \\
        lo/file\_rwpms\_wcnf\_L2\_V150\_C1000\_H150\_6 & 2.967 \\
        lo/file\_rwpms\_wcnf\_L2\_V150\_C1000\_H150\_7 & 2.998 \\
        lo/file\_rwpms\_wcnf\_L2\_V150\_C1000\_H150\_8 & 3.072 \\
        lo/file\_rwpms\_wcnf\_L2\_V150\_C1000\_H150\_9 & 2.935 \\
        lo/file\_rwpms\_wcnf\_L2\_V150\_C1500\_H150\_0 & 4.567 \\
        lo/file\_rwpms\_wcnf\_L2\_V150\_C1500\_H150\_1 & 4.511 \\
        lo/file\_rwpms\_wcnf\_L2\_V150\_C1500\_H150\_2 & 4.540 \\
        lo/file\_rwpms\_wcnf\_L2\_V150\_C1500\_H150\_3 & 4.482 \\
        lo/file\_rwpms\_wcnf\_L2\_V150\_C1500\_H150\_4 & 4.747 \\
        lo/file\_rwpms\_wcnf\_L2\_V150\_C1500\_H150\_5 & 4.567 \\
        lo/file\_rwpms\_wcnf\_L2\_V150\_C1500\_H150\_6 & 4.699 \\
        lo/file\_rwpms\_wcnf\_L2\_V150\_C1500\_H150\_7 & 4.544 \\
        lo/file\_rwpms\_wcnf\_L2\_V150\_C1500\_H150\_8 & 4.561 \\
        lo/file\_rwpms\_wcnf\_L2\_V150\_C1500\_H150\_9 & 4.770 \\
        lo/file\_rwpms\_wcnf\_L2\_V150\_C2000\_H150\_0 & 6.072 \\
        lo/file\_rwpms\_wcnf\_L2\_V150\_C2000\_H150\_1 & 5.814 \\
        lo/file\_rwpms\_wcnf\_L2\_V150\_C2000\_H150\_2 & 5.984 \\
        lo/file\_rwpms\_wcnf\_L2\_V150\_C2000\_H150\_3 & 5.966 \\
        lo/file\_rwpms\_wcnf\_L2\_V150\_C2000\_H150\_4 & 5.977 \\
        lo/file\_rwpms\_wcnf\_L2\_V150\_C2000\_H150\_5 & 5.910 \\
        lo/file\_rwpms\_wcnf\_L2\_V150\_C2000\_H150\_6 & 6.205 \\
        lo/file\_rwpms\_wcnf\_L2\_V150\_C2000\_H150\_7 & 6.008 \\
        lo/file\_rwpms\_wcnf\_L2\_V150\_C2000\_H150\_8 & 6.044 \\
        lo/file\_rwpms\_wcnf\_L2\_V150\_C2000\_H150\_9 & 5.979 \\
        me/file\_rwpms\_wcnf\_L2\_V150\_C2500\_H150\_0 & 8.042 \\
        me/file\_rwpms\_wcnf\_L2\_V150\_C2500\_H150\_1 & 8.342 \\
        me/file\_rwpms\_wcnf\_L2\_V150\_C2500\_H150\_2 & 8.321 \\
        me/file\_rwpms\_wcnf\_L2\_V150\_C2500\_H150\_3 & 8.216 \\
        me/file\_rwpms\_wcnf\_L2\_V150\_C2500\_H150\_4 & 8.189 \\
        me/file\_rwpms\_wcnf\_L2\_V150\_C2500\_H150\_5 & 8.014 \\
        me/file\_rwpms\_wcnf\_L2\_V150\_C2500\_H150\_6 & 8.327 \\
        me/file\_rwpms\_wcnf\_L2\_V150\_C2500\_H150\_7 & 8.684 \\
        me/file\_rwpms\_wcnf\_L2\_V150\_C2500\_H150\_8 & 8.275 \\
        me/file\_rwpms\_wcnf\_L2\_V150\_C2500\_H150\_9 & 8.098 \\
        me/file\_rwpms\_wcnf\_L2\_V150\_C3000\_H150\_0 & 9.627 \\
        me/file\_rwpms\_wcnf\_L2\_V150\_C3000\_H150\_1 & 9.722 \\
        me/file\_rwpms\_wcnf\_L2\_V150\_C3000\_H150\_2 & 9.704 \\
        me/file\_rwpms\_wcnf\_L2\_V150\_C3000\_H150\_3 & 9.946 \\
        me/file\_rwpms\_wcnf\_L2\_V150\_C3000\_H150\_4 & 9.925 \\
        me/file\_rwpms\_wcnf\_L2\_V150\_C3000\_H150\_5 & 9.731 \\
        me/file\_rwpms\_wcnf\_L2\_V150\_C3000\_H150\_6 & 9.768 \\
        me/file\_rwpms\_wcnf\_L2\_V150\_C3000\_H150\_7 & 10.034 \\
        me/file\_rwpms\_wcnf\_L2\_V150\_C3000\_H150\_8 & 9.953 \\
        me/file\_rwpms\_wcnf\_L2\_V150\_C3000\_H150\_9 & 10.169 \\
        me/file\_rwpms\_wcnf\_L2\_V150\_C3500\_H150\_0 & 11.615 \\
        me/file\_rwpms\_wcnf\_L2\_V150\_C3500\_H150\_1 & 11.186 \\
        me/file\_rwpms\_wcnf\_L2\_V150\_C3500\_H150\_2 & 11.699 \\
        me/file\_rwpms\_wcnf\_L2\_V150\_C3500\_H150\_3 & 11.419 \\
        me/file\_rwpms\_wcnf\_L2\_V150\_C3500\_H150\_4 & 10.907 \\
        me/file\_rwpms\_wcnf\_L2\_V150\_C3500\_H150\_5 & 11.244 \\
        me/file\_rwpms\_wcnf\_L2\_V150\_C3500\_H150\_6 & 11.513 \\
        me/file\_rwpms\_wcnf\_L2\_V150\_C3500\_H150\_7 & 11.327 \\
        me/file\_rwpms\_wcnf\_L2\_V150\_C3500\_H150\_8 & 11.301 \\
        me/file\_rwpms\_wcnf\_L2\_V150\_C3500\_H150\_9 & 11.191 \\
        hi/file\_rwpms\_wcnf\_L2\_V150\_C4000\_H150\_0 & 12.843 \\
        hi/file\_rwpms\_wcnf\_L2\_V150\_C4000\_H150\_1 & 13.317 \\
        hi/file\_rwpms\_wcnf\_L2\_V150\_C4000\_H150\_2 & 13.011 \\
        hi/file\_rwpms\_wcnf\_L2\_V150\_C4000\_H150\_3 & 13.307 \\
        hi/file\_rwpms\_wcnf\_L2\_V150\_C4000\_H150\_4 & 13.294 \\
        hi/file\_rwpms\_wcnf\_L2\_V150\_C4000\_H150\_5 & 12.985 \\
        hi/file\_rwpms\_wcnf\_L2\_V150\_C4000\_H150\_6 & 12.807 \\
        hi/file\_rwpms\_wcnf\_L2\_V150\_C4000\_H150\_7 & 13.233 \\
        hi/file\_rwpms\_wcnf\_L2\_V150\_C4000\_H150\_8 & 12.821 \\
        hi/file\_rwpms\_wcnf\_L2\_V150\_C4000\_H150\_9 & 13.235 \\
        hi/file\_rwpms\_wcnf\_L2\_V150\_C4500\_H150\_0 & 14.606 \\
        hi/file\_rwpms\_wcnf\_L2\_V150\_C4500\_H150\_1 & 14.929 \\
        hi/file\_rwpms\_wcnf\_L2\_V150\_C4500\_H150\_2 & 14.851 \\
        hi/file\_rwpms\_wcnf\_L2\_V150\_C4500\_H150\_3 & 14.981 \\
        hi/file\_rwpms\_wcnf\_L2\_V150\_C4500\_H150\_4 & 15.281 \\
        hi/file\_rwpms\_wcnf\_L2\_V150\_C4500\_H150\_5 & 14.661 \\
        hi/file\_rwpms\_wcnf\_L2\_V150\_C4500\_H150\_6 & 14.714 \\
        hi/file\_rwpms\_wcnf\_L2\_V150\_C4500\_H150\_7 & 15.073 \\
        hi/file\_rwpms\_wcnf\_L2\_V150\_C4500\_H150\_8 & 15.090 \\
        hi/file\_rwpms\_wcnf\_L2\_V150\_C4500\_H150\_9 & 14.756 \\
        hi/file\_rwpms\_wcnf\_L2\_V150\_C5000\_H150\_0 & 17.503 \\
        hi/file\_rwpms\_wcnf\_L2\_V150\_C5000\_H150\_1 & 16.258 \\
        hi/file\_rwpms\_wcnf\_L2\_V150\_C5000\_H150\_2 & 16.471 \\
        hi/file\_rwpms\_wcnf\_L2\_V150\_C5000\_H150\_3 & 16.391 \\
        hi/file\_rwpms\_wcnf\_L2\_V150\_C5000\_H150\_4 & 16.555 \\
        hi/file\_rwpms\_wcnf\_L2\_V150\_C5000\_H150\_5 & 16.883 \\
        hi/file\_rwpms\_wcnf\_L2\_V150\_C5000\_H150\_6 & 16.495 \\
        hi/file\_rwpms\_wcnf\_L2\_V150\_C5000\_H150\_7 & 16.531 \\
        hi/file\_rwpms\_wcnf\_L2\_V150\_C5000\_H150\_8 & 16.337 \\
        hi/file\_rwpms\_wcnf\_L2\_V150\_C5000\_H150\_9 & 16.482 \\
        \hline
    \end{longtable}
\end{center}

\chapter{DMaxWalkSat vs. MaxWalkSat for DWCSP Result Graphs\label{dmaxwalksat_graphs}}
\clearpage

\begin{figure}[H]
    \setcounter{subfigure}{0}
    \centering
        \subfloat[Constraint addition]
        {
            \includegraphics[width=2.7in]{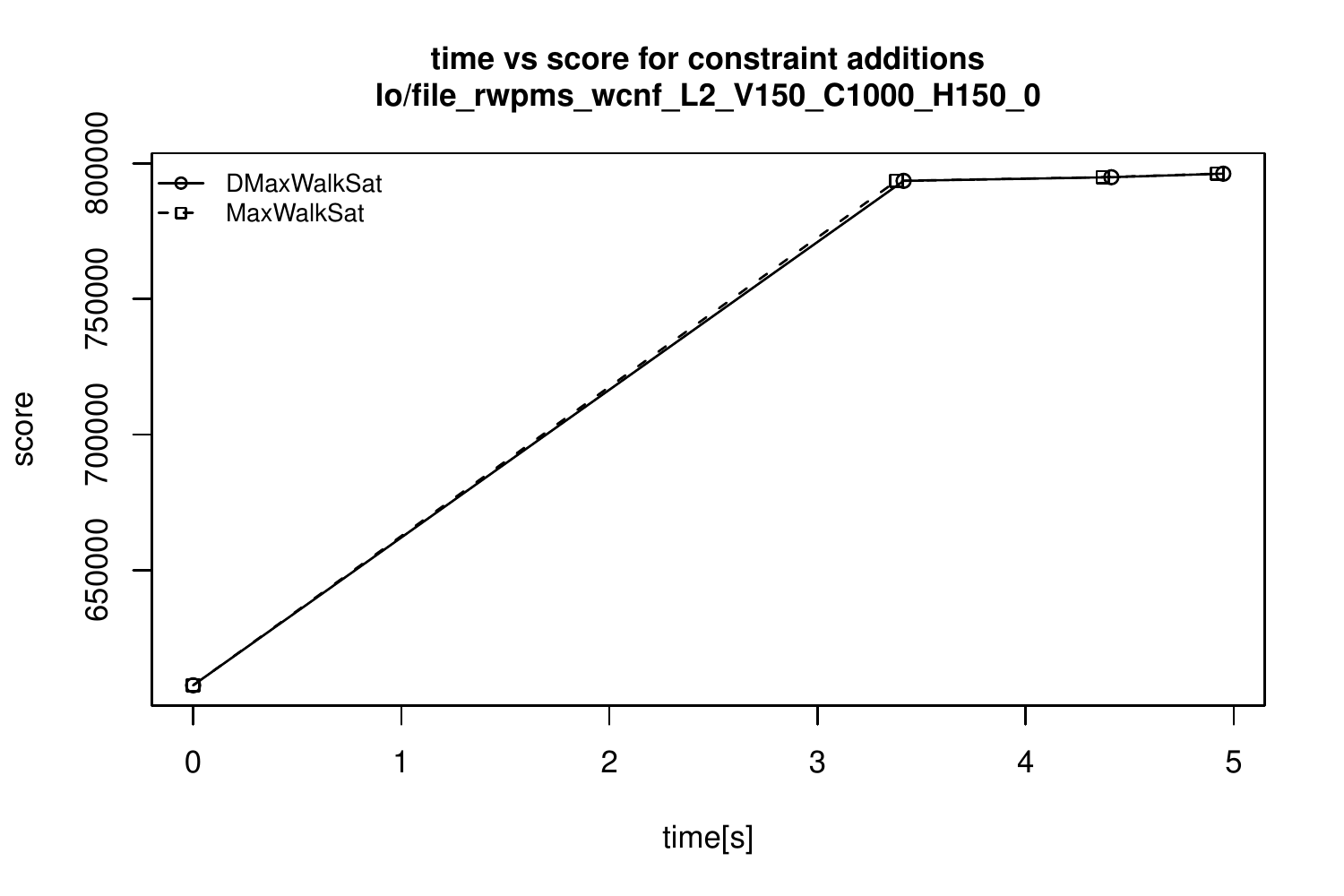}
        }
        \qquad
        \subfloat[Constraint removal]
        {
            \includegraphics[width=2.7in]{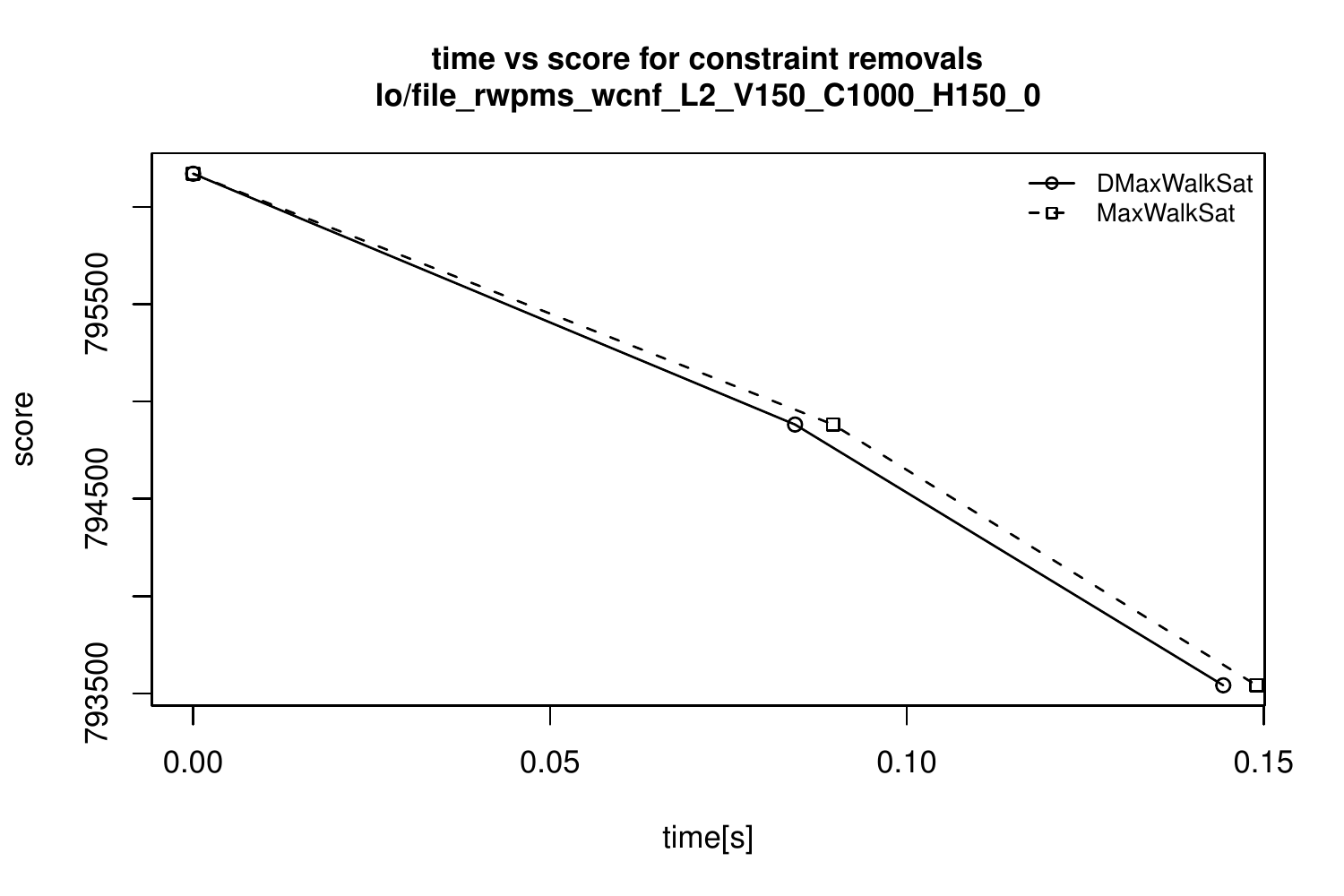}
        }

    \caption*{lo/file\_rwpms\_wcnf\_L2\_V150\_C1000\_H150\_0}
    \label{fig_lo/file_rwpms_wcnf_L2_V150_C1000_H150_0}
\end{figure}

\begin{figure}[H]
    \setcounter{subfigure}{0}
    \centering
        \subfloat[Constraint addition]
        {
            \includegraphics[width=2.7in]{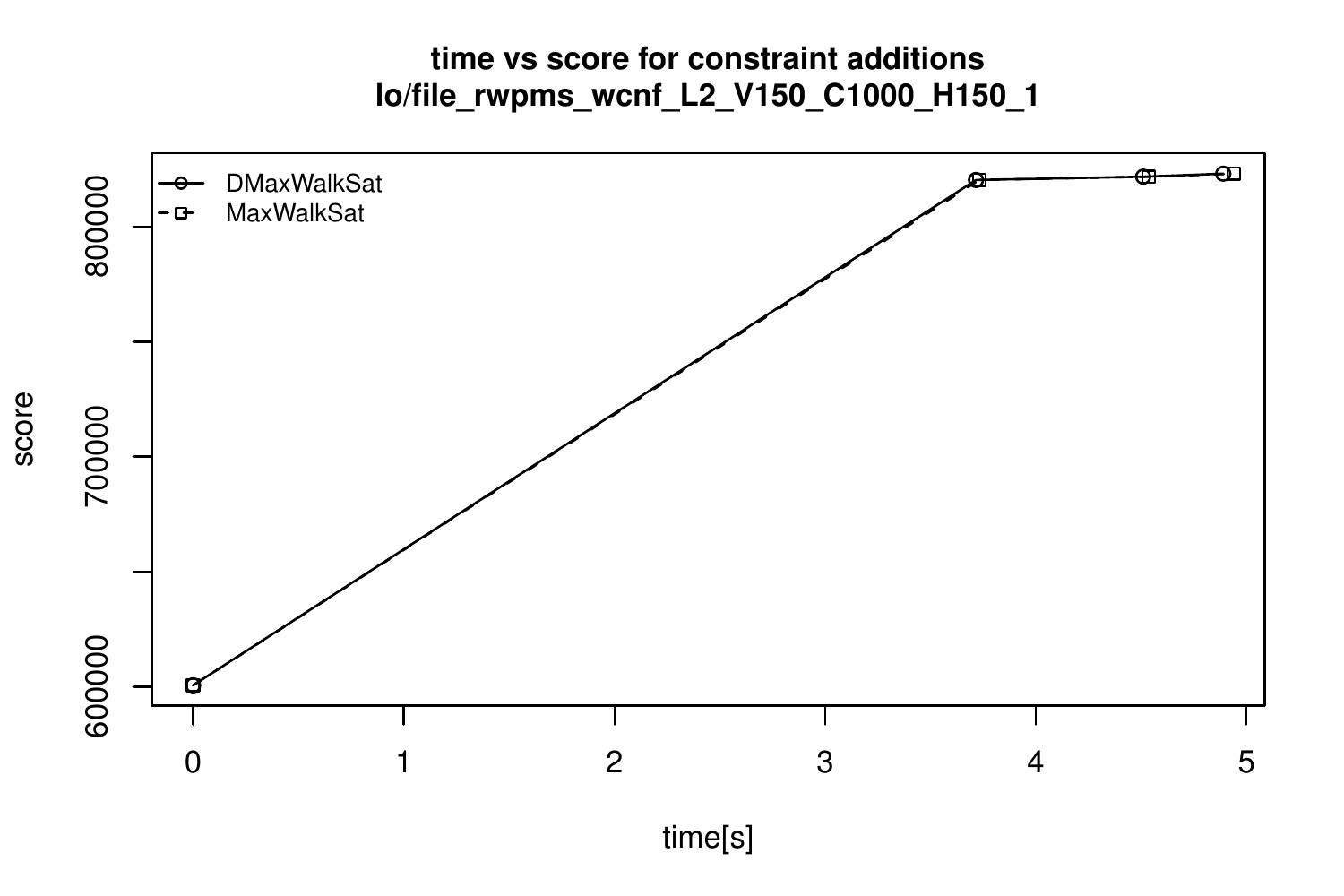}
        }
        \qquad
        \subfloat[Constraint removal]
        {
            \includegraphics[width=2.7in]{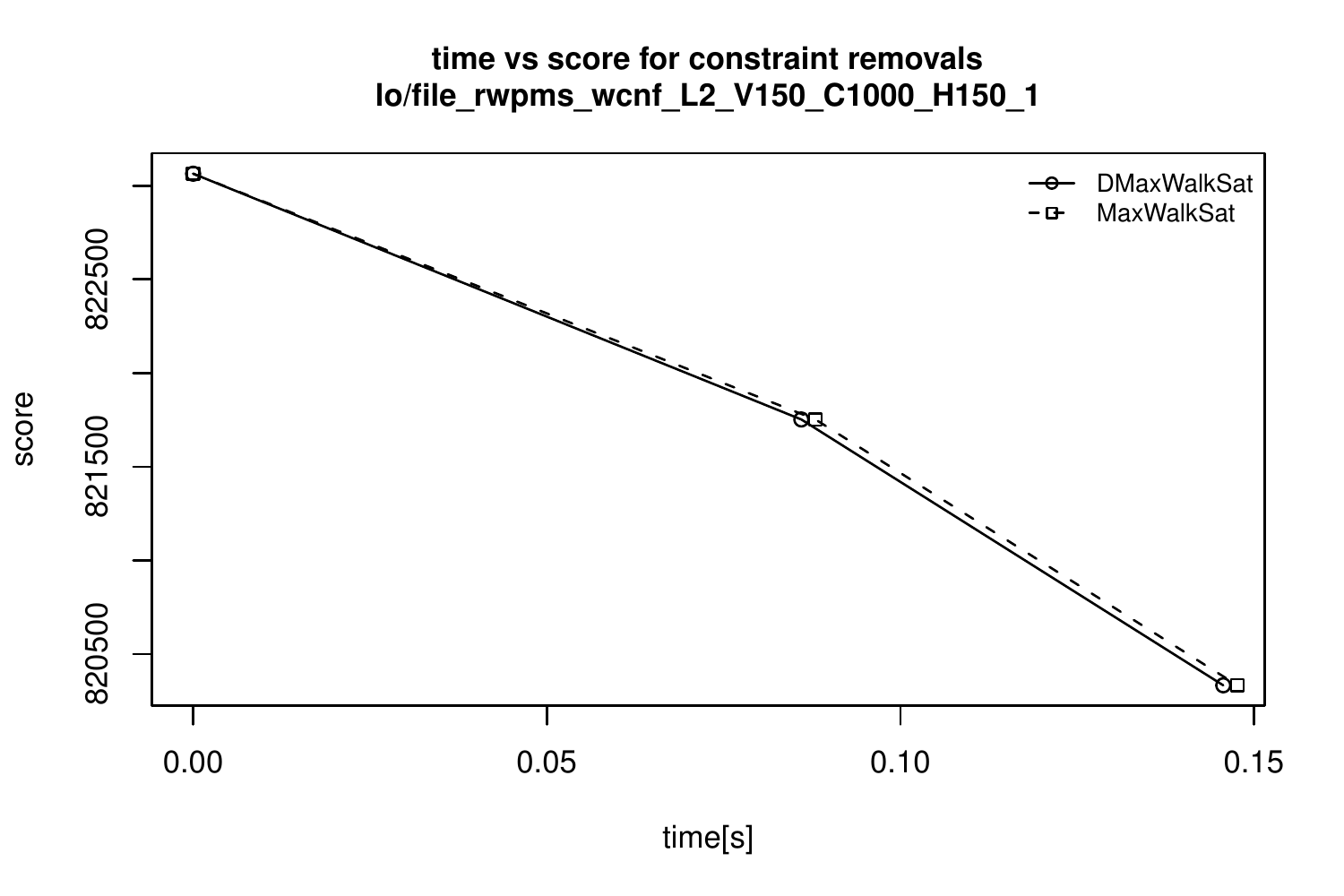}
        }

    \caption*{lo/file\_rwpms\_wcnf\_L2\_V150\_C1000\_H150\_1}
    \label{fig_lo/file_rwpms_wcnf_L2_V150_C1000_H150_1}
\end{figure}

\begin{figure}[H]
    \setcounter{subfigure}{0}
    \centering
        \subfloat[Constraint addition]
        {
            \includegraphics[width=2.7in]{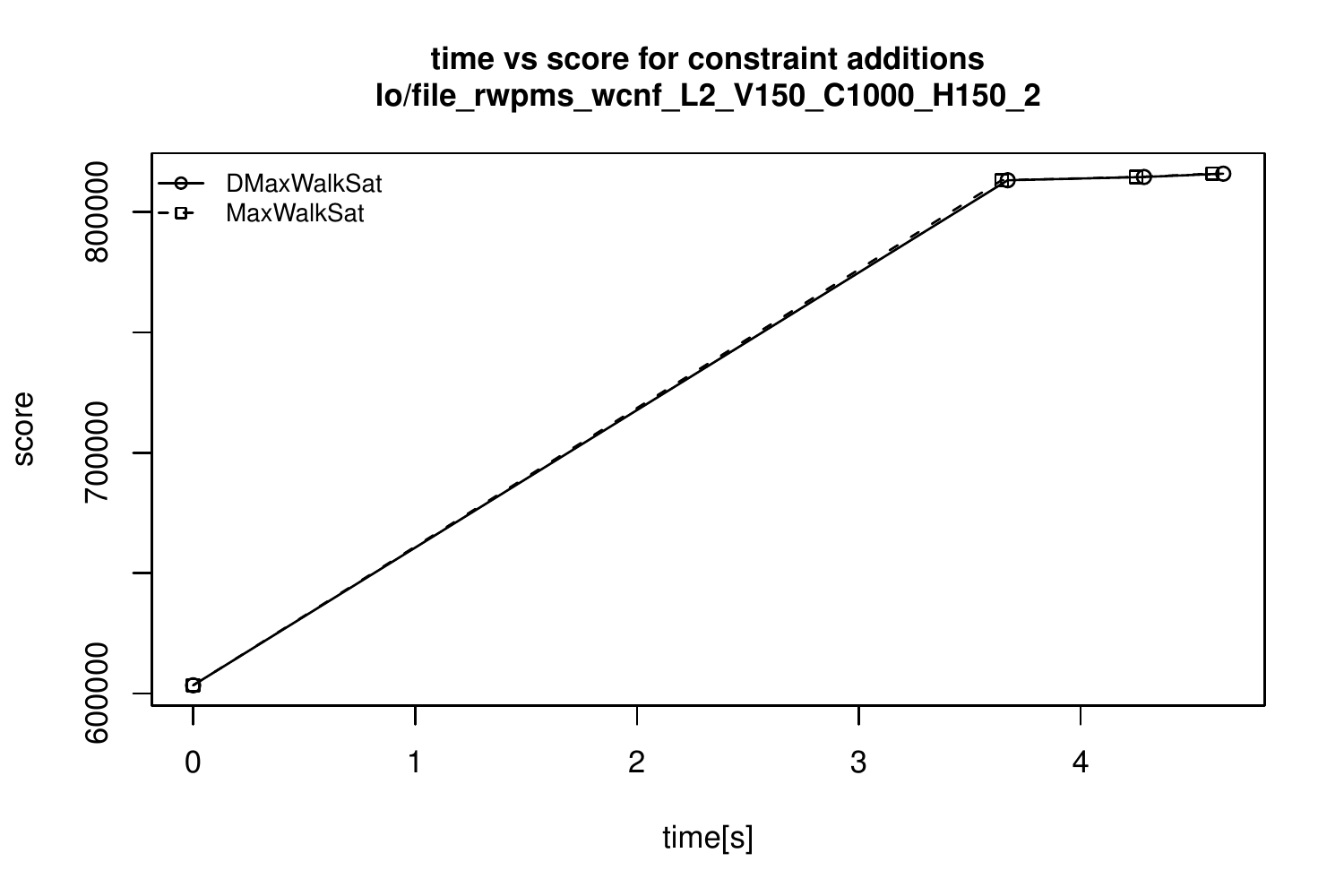}
        }
        \qquad
        \subfloat[Constraint removal]
        {
            \includegraphics[width=2.7in]{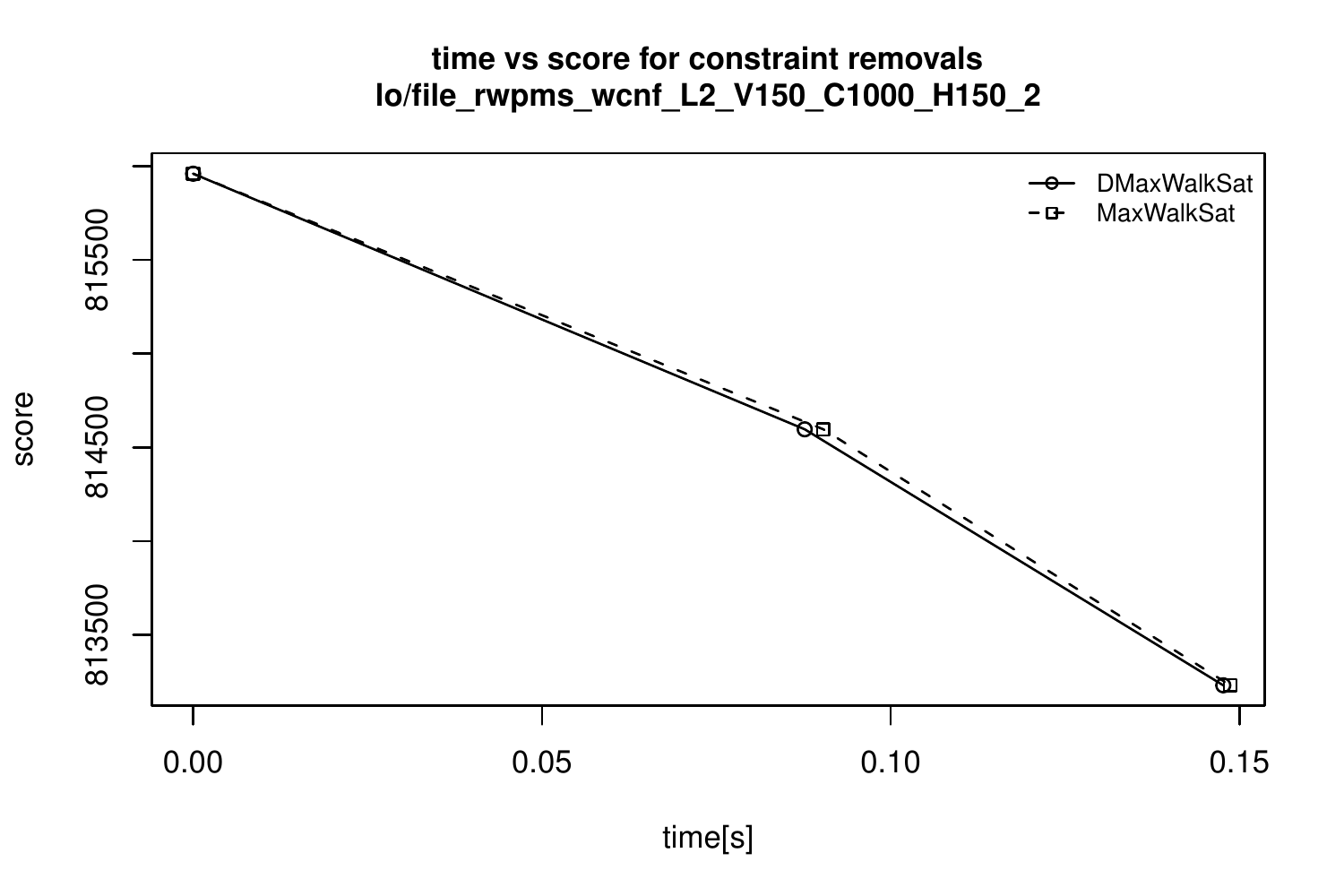}
        }

    \caption*{lo/file\_rwpms\_wcnf\_L2\_V150\_C1000\_H150\_2}
    \label{fig_lo/file_rwpms_wcnf_L2_V150_C1000_H150_2}
\end{figure}

\begin{figure}[H]
    \setcounter{subfigure}{0}
    \centering
        \subfloat[Constraint addition]
        {
            \includegraphics[width=2.7in]{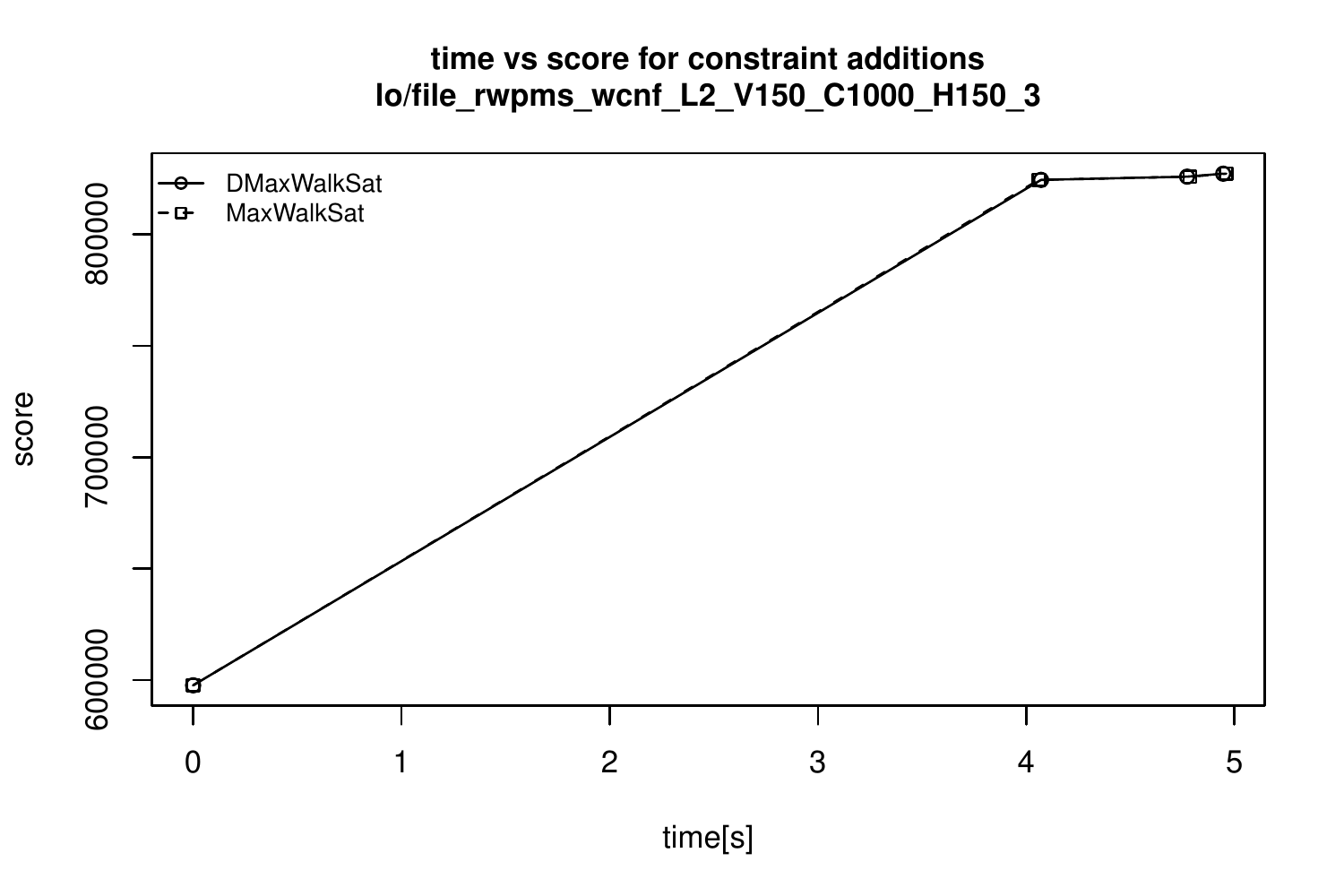}
        }
        \qquad
        \subfloat[Constraint removal]
        {
            \includegraphics[width=2.7in]{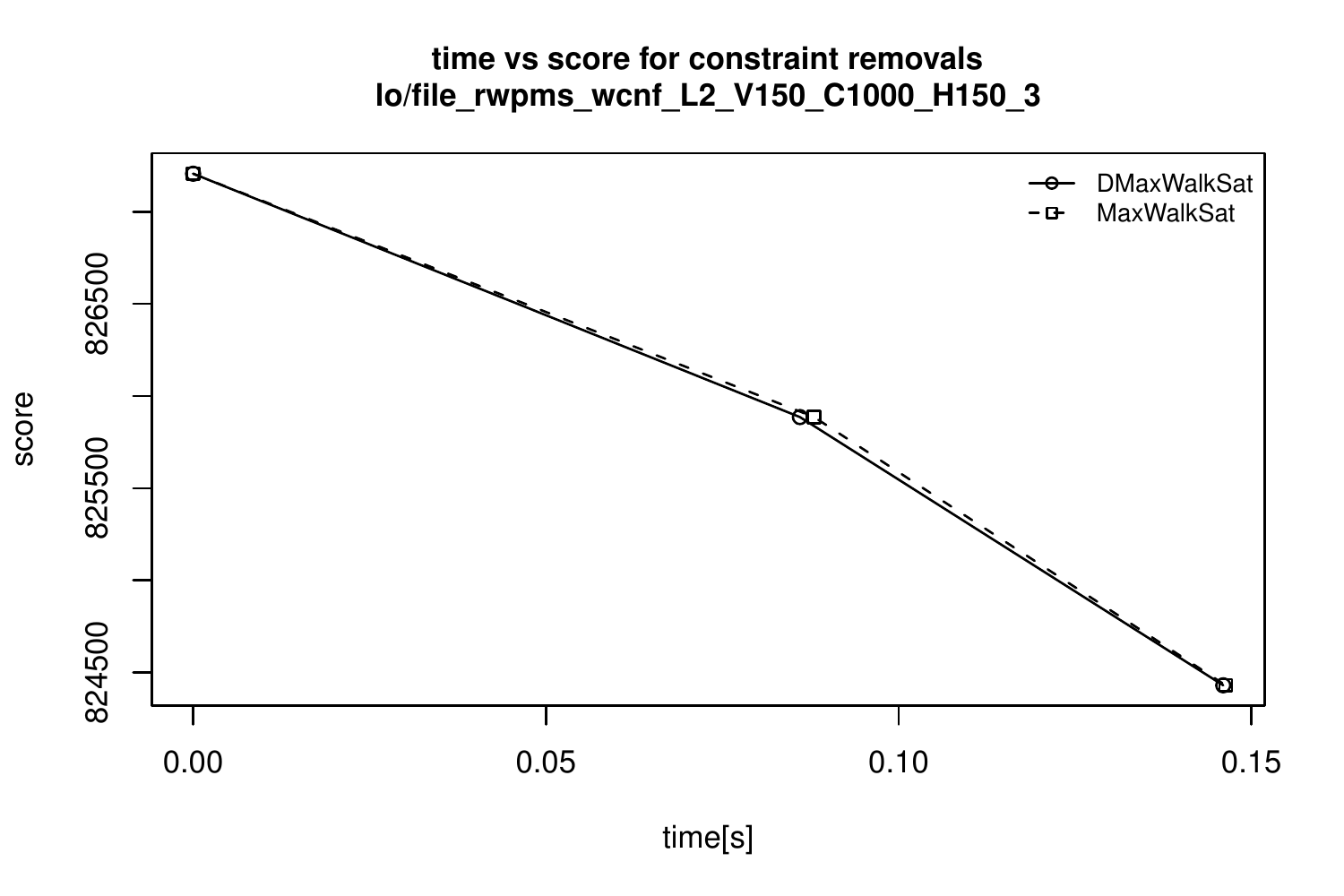}
        }

    \caption*{lo/file\_rwpms\_wcnf\_L2\_V150\_C1000\_H150\_3}
    \label{fig_lo/file_rwpms_wcnf_L2_V150_C1000_H150_3}
\end{figure}

\begin{figure}[H]
    \setcounter{subfigure}{0}
    \centering
        \subfloat[Constraint addition]
        {
            \includegraphics[width=2.7in]{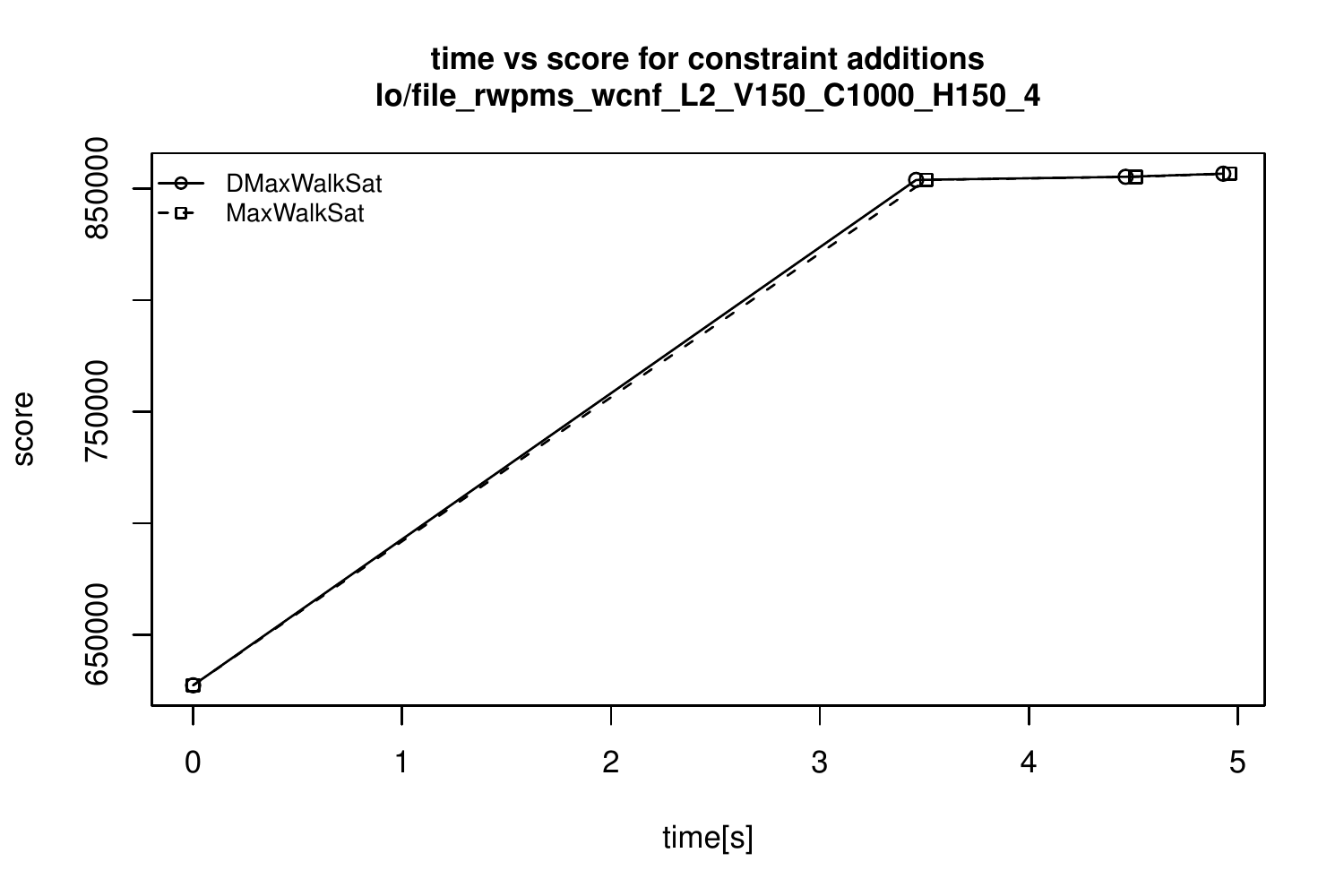}
        }
        \qquad
        \subfloat[Constraint removal]
        {
            \includegraphics[width=2.7in]{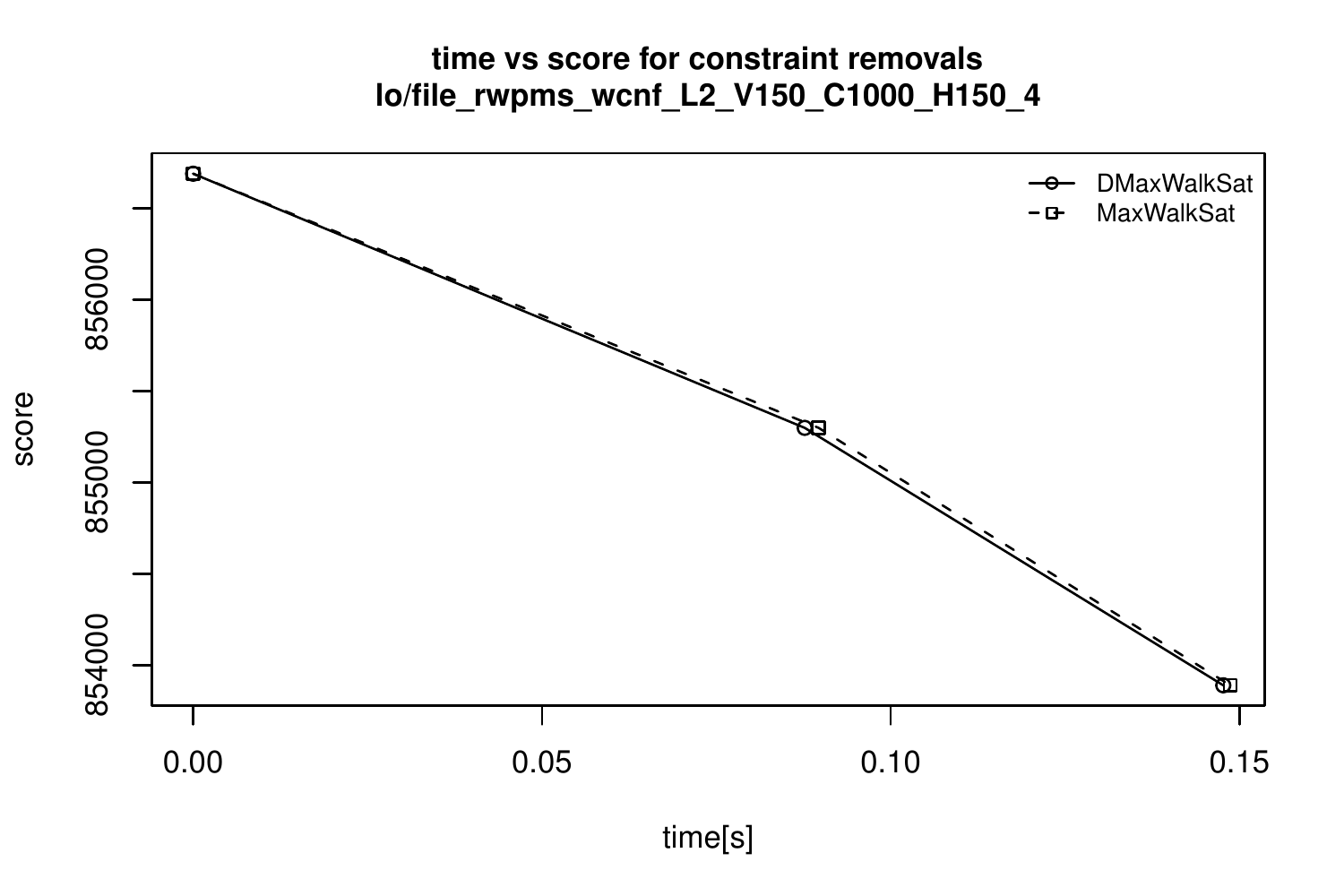}
        }

    \caption*{lo/file\_rwpms\_wcnf\_L2\_V150\_C1000\_H150\_4}
    \label{fig_lo/file_rwpms_wcnf_L2_V150_C1000_H150_4}
\end{figure}

\begin{figure}[H]
    \setcounter{subfigure}{0}
    \centering
        \subfloat[Constraint addition]
        {
            \includegraphics[width=2.7in]{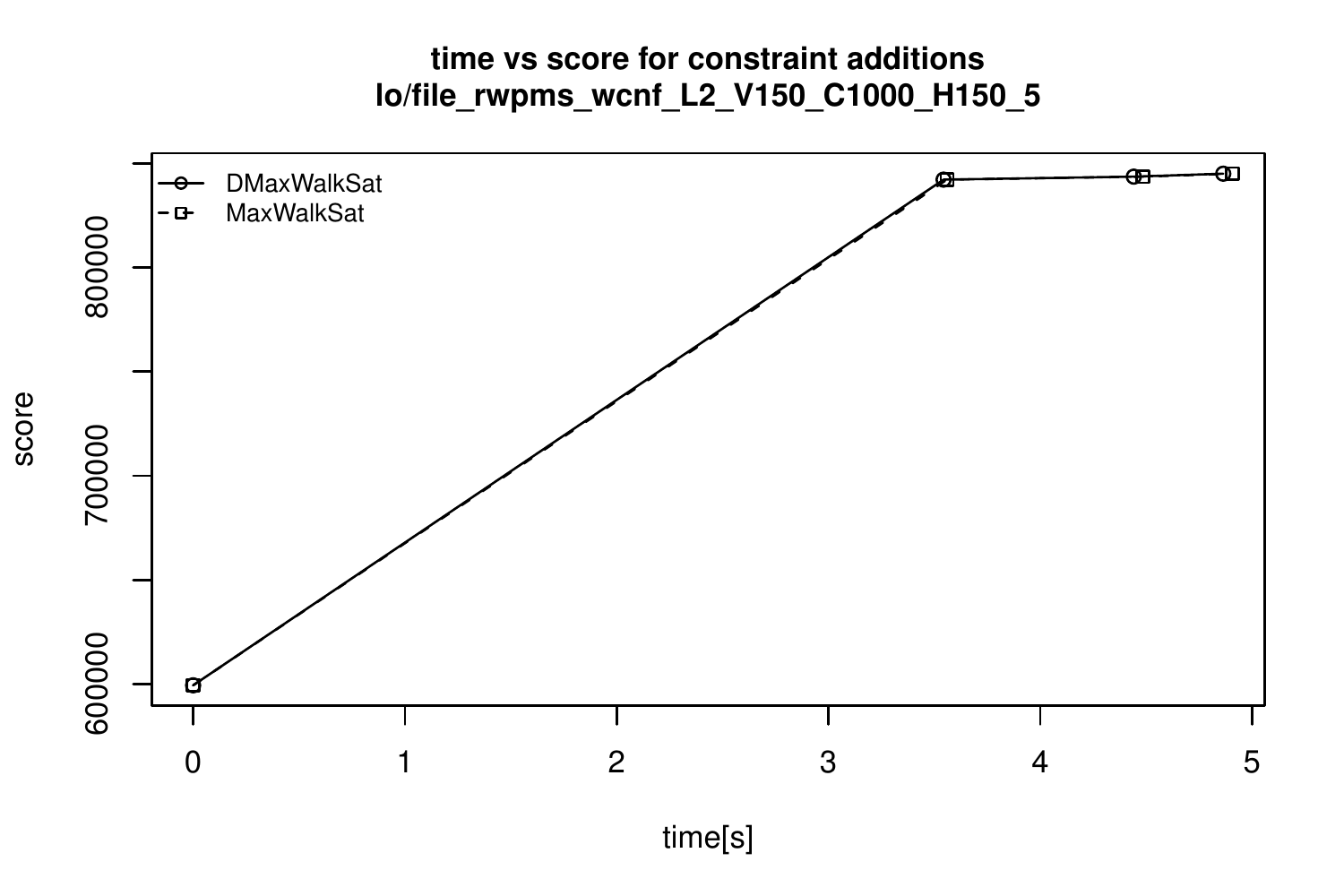}
        }
        \qquad
        \subfloat[Constraint removal]
        {
            \includegraphics[width=2.7in]{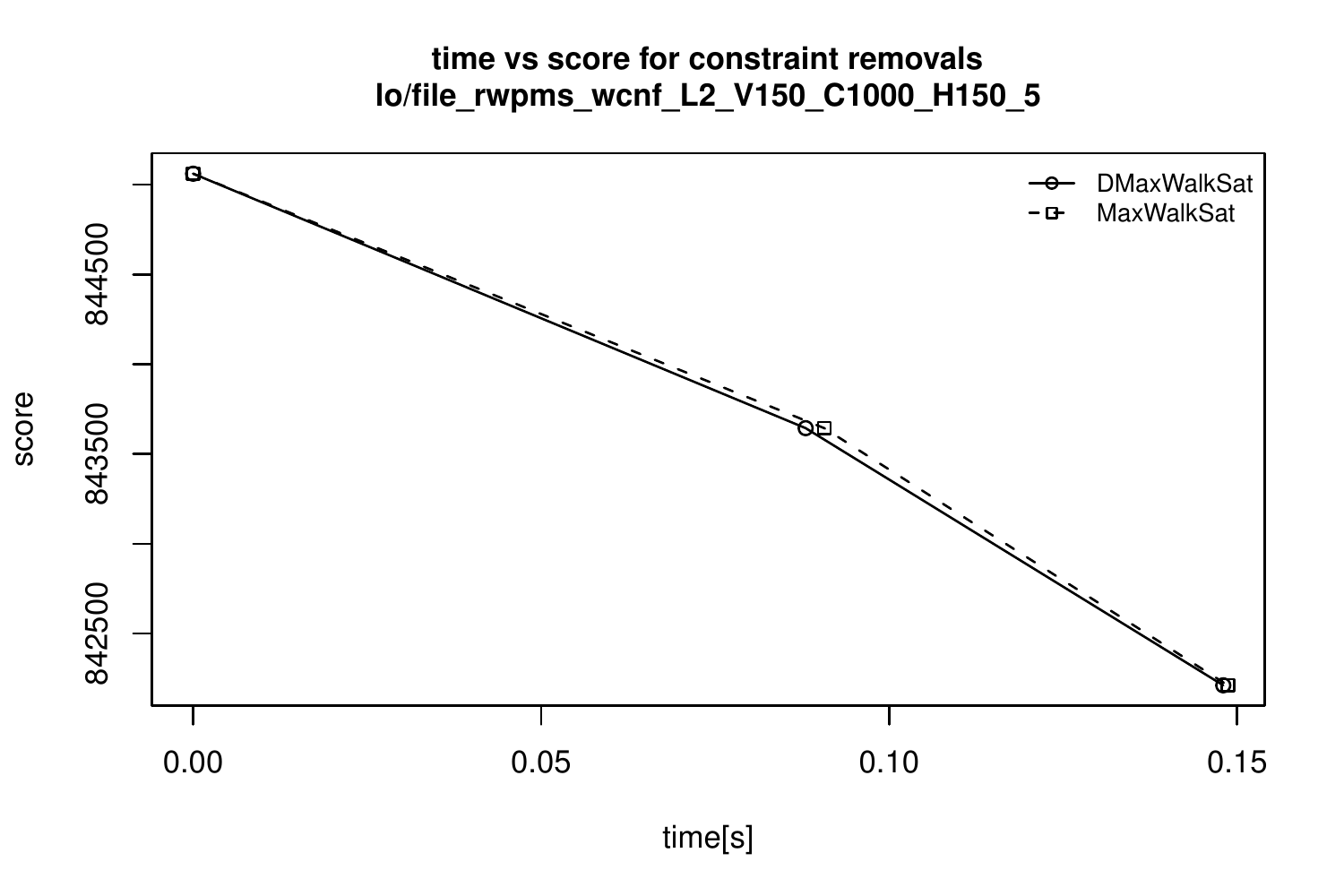}
        }

    \caption*{lo/file\_rwpms\_wcnf\_L2\_V150\_C1000\_H150\_5}
    \label{fig_lo/file_rwpms_wcnf_L2_V150_C1000_H150_5}
\end{figure}

\begin{figure}[H]
    \setcounter{subfigure}{0}
    \centering
        \subfloat[Constraint addition]
        {
            \includegraphics[width=2.7in]{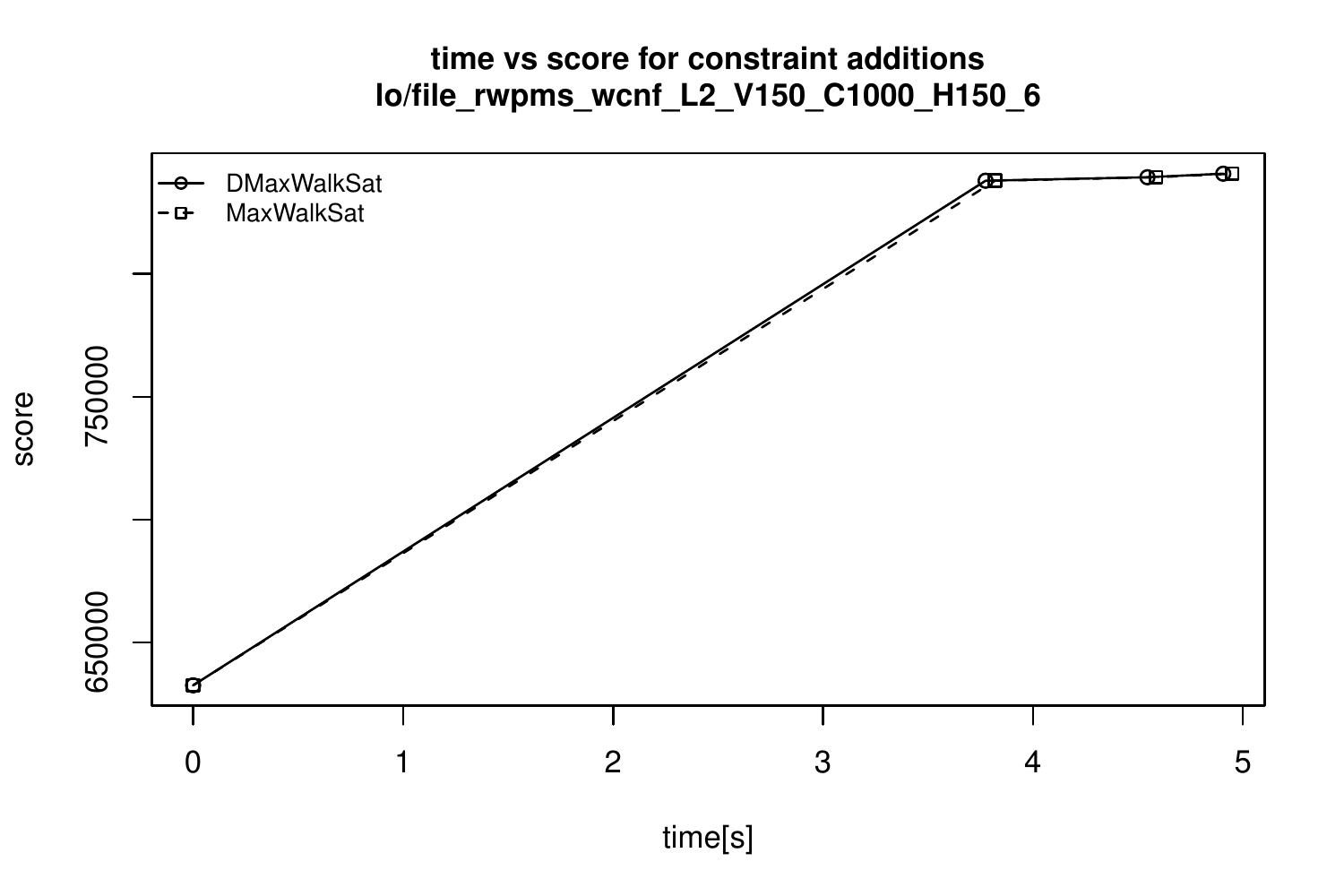}
        }
        \qquad
        \subfloat[Constraint removal]
        {
            \includegraphics[width=2.7in]{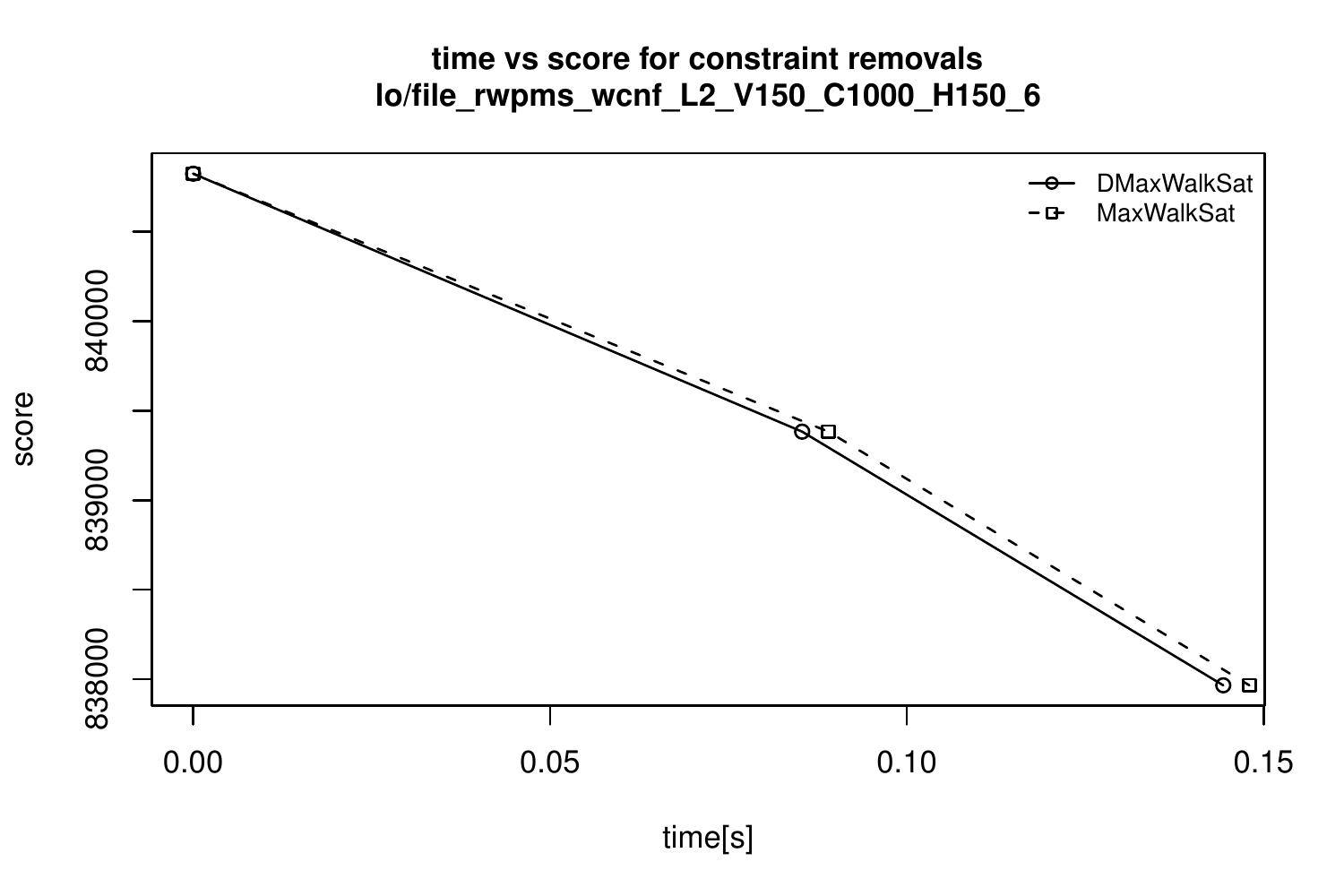}
        }

    \caption*{lo/file\_rwpms\_wcnf\_L2\_V150\_C1000\_H150\_6}
    \label{fig_lo/file_rwpms_wcnf_L2_V150_C1000_H150_6}
\end{figure}

\begin{figure}[H]
    \setcounter{subfigure}{0}
    \centering
        \subfloat[Constraint addition]
        {
            \includegraphics[width=2.7in]{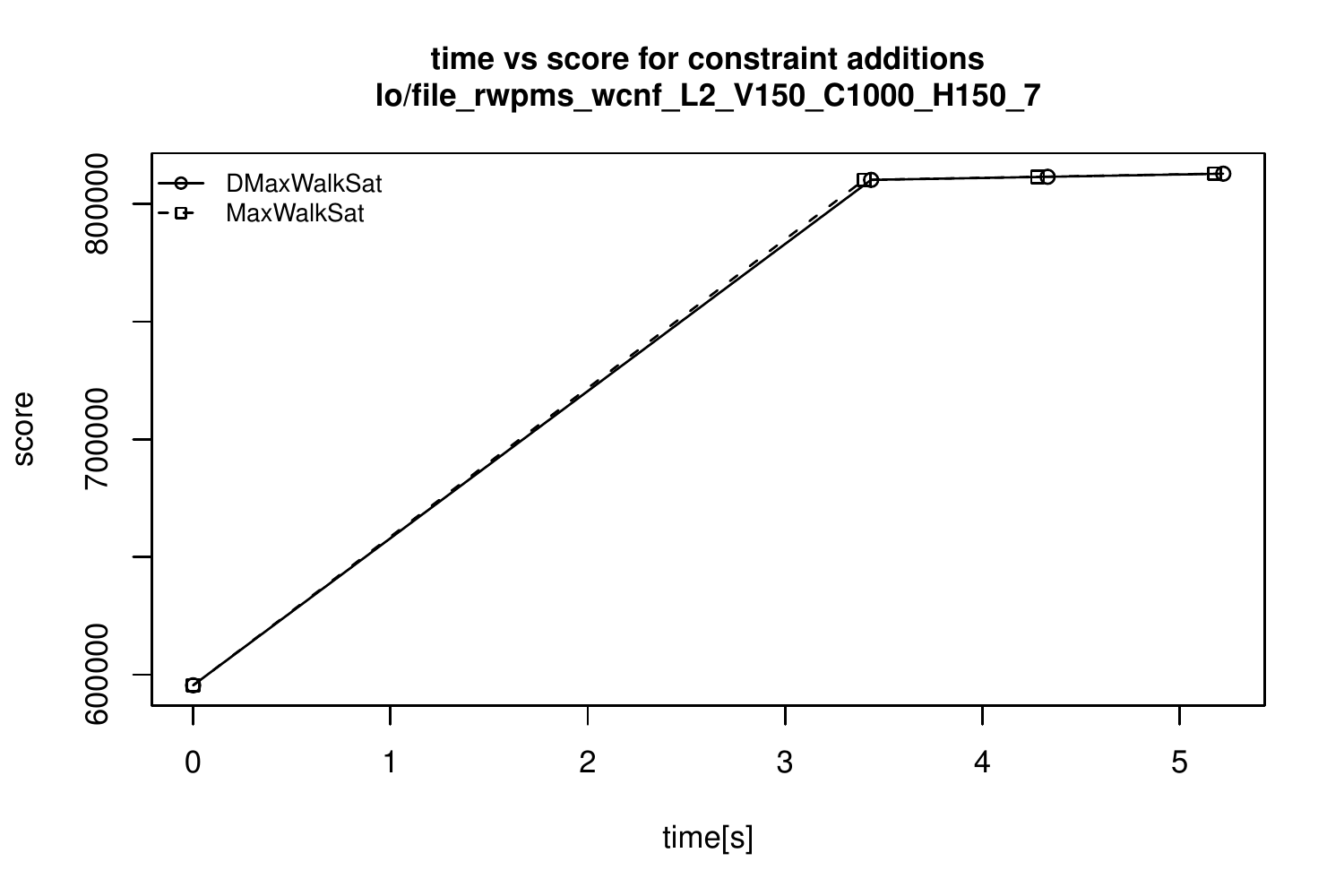}
        }
        \qquad
        \subfloat[Constraint removal]
        {
            \includegraphics[width=2.7in]{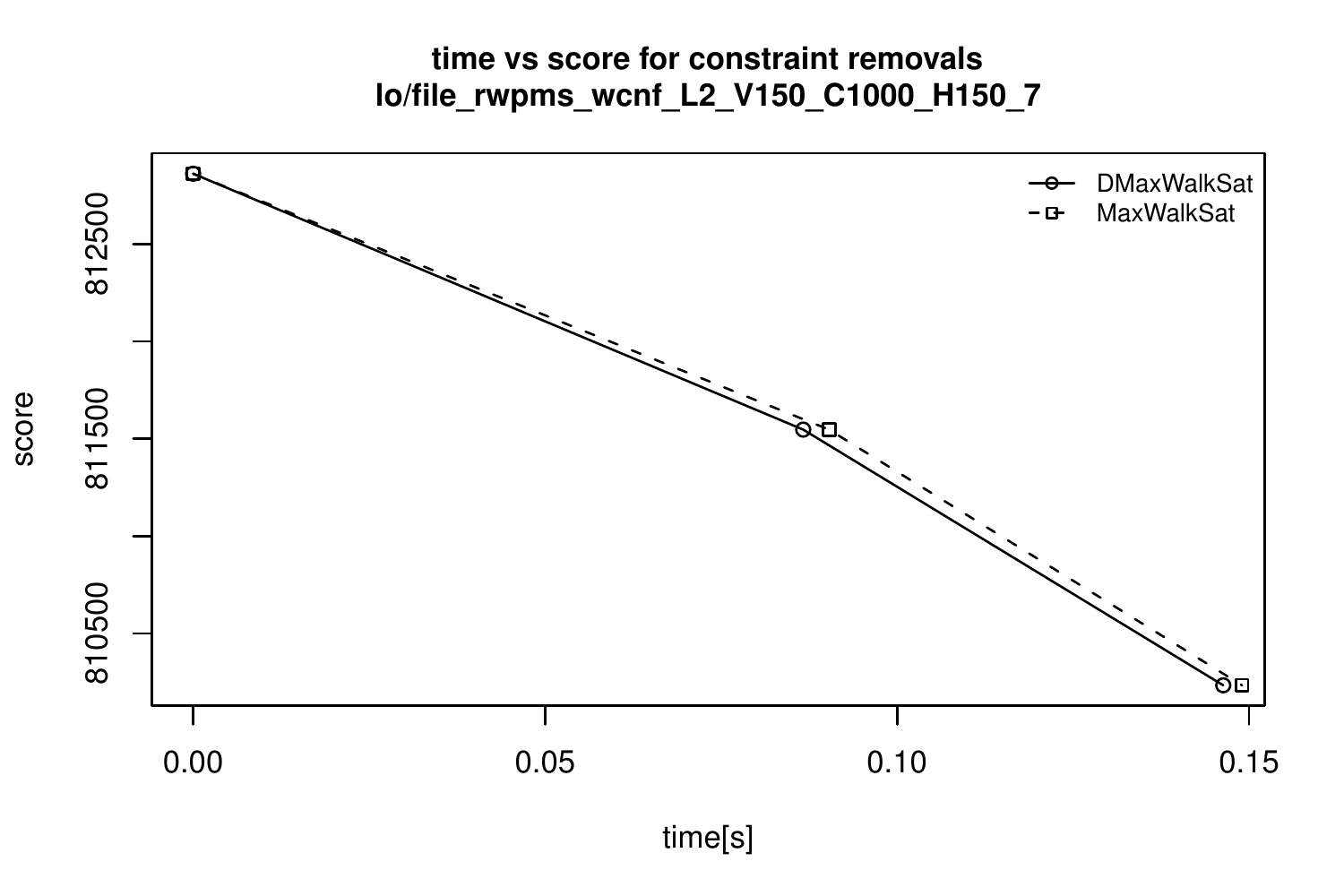}
        }

    \caption*{lo/file\_rwpms\_wcnf\_L2\_V150\_C1000\_H150\_7}
    \label{fig_lo/file_rwpms_wcnf_L2_V150_C1000_H150_7}
\end{figure}

\begin{figure}[H]
    \setcounter{subfigure}{0}
    \centering
        \subfloat[Constraint addition]
        {
            \includegraphics[width=2.7in]{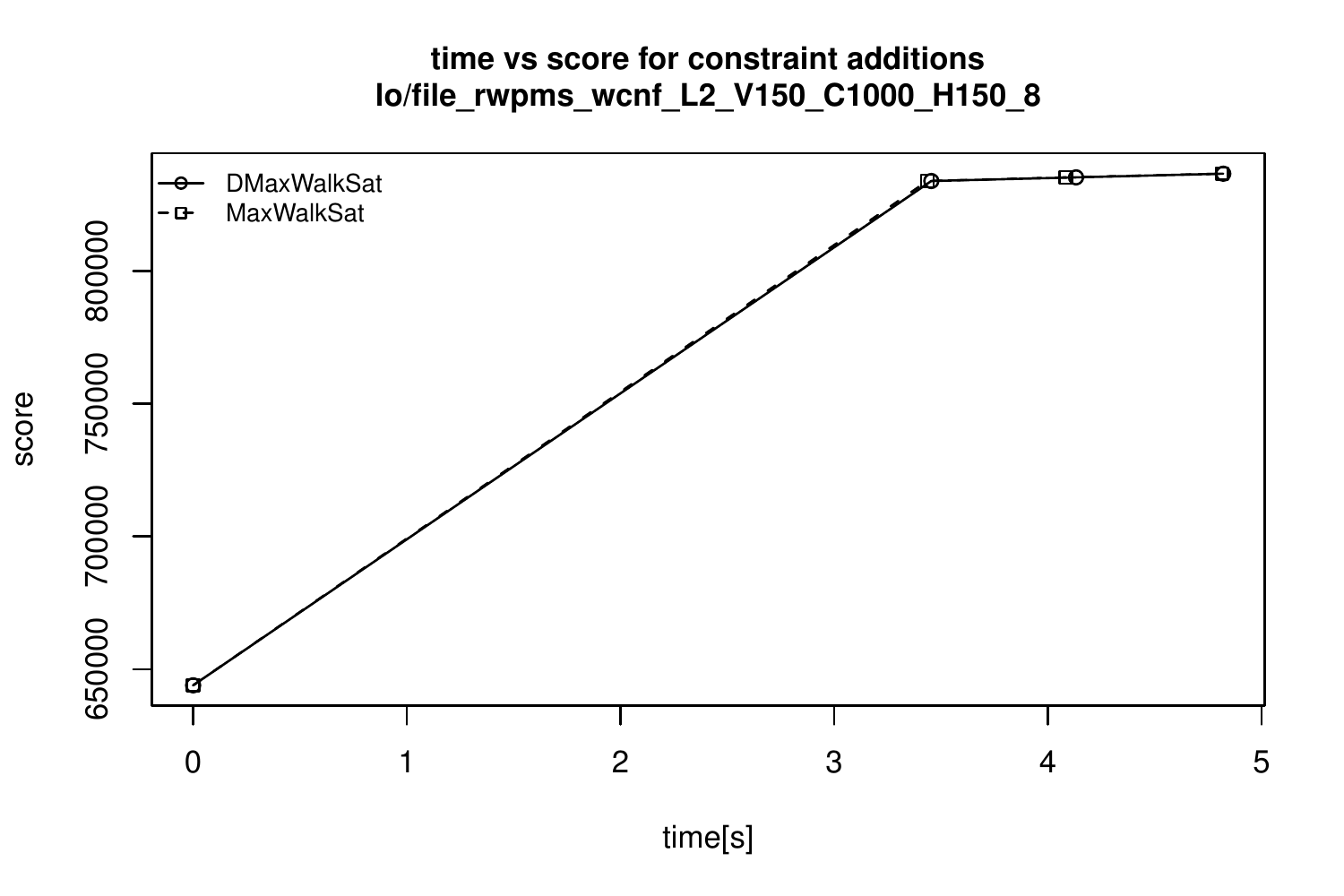}
        }
        \qquad
        \subfloat[Constraint removal]
        {
            \includegraphics[width=2.7in]{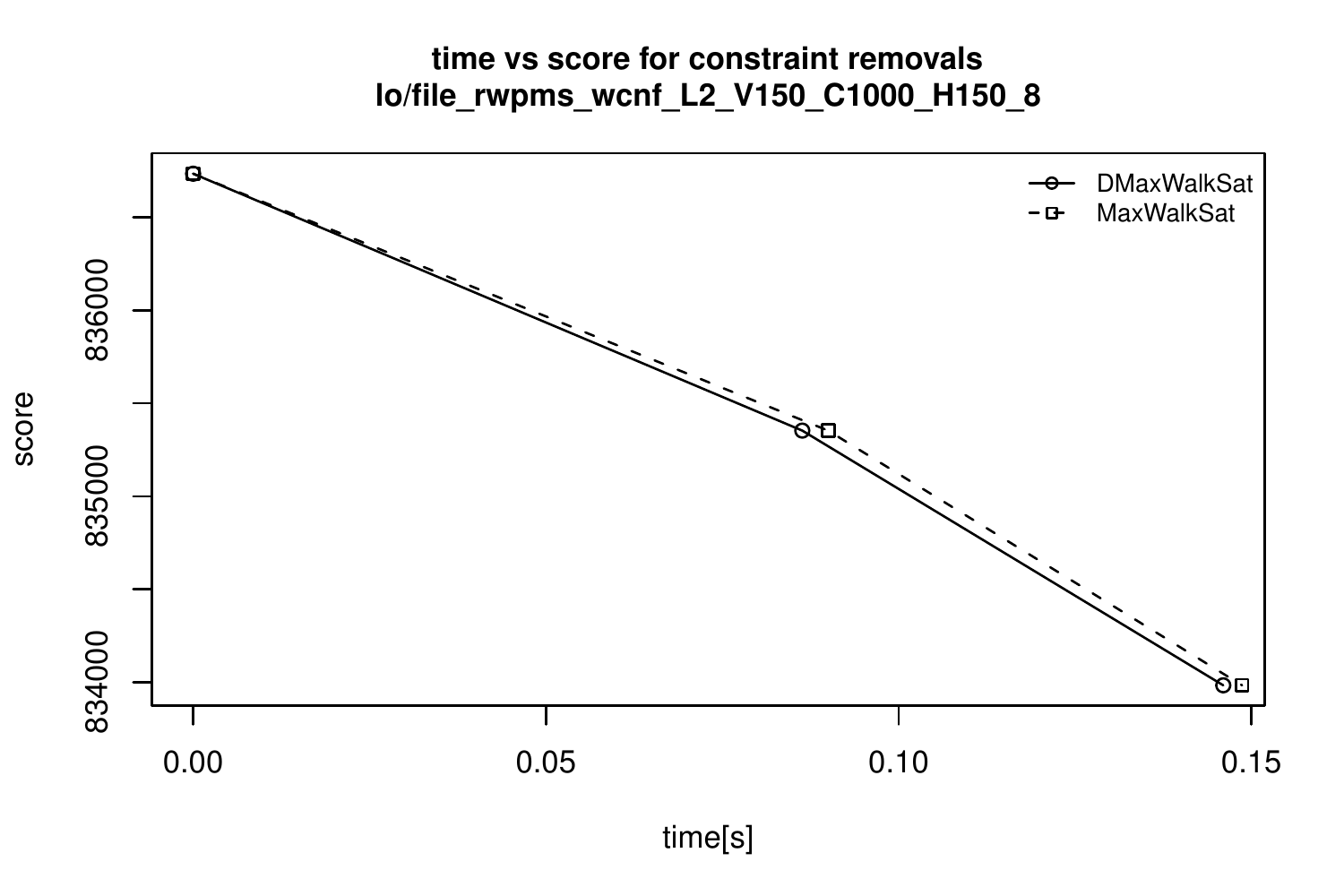}
        }

    \caption*{lo/file\_rwpms\_wcnf\_L2\_V150\_C1000\_H150\_8}
    \label{fig_lo/file_rwpms_wcnf_L2_V150_C1000_H150_8}
\end{figure}

\begin{figure}[H]
    \setcounter{subfigure}{0}
    \centering
        \subfloat[Constraint addition]
        {
            \includegraphics[width=2.7in]{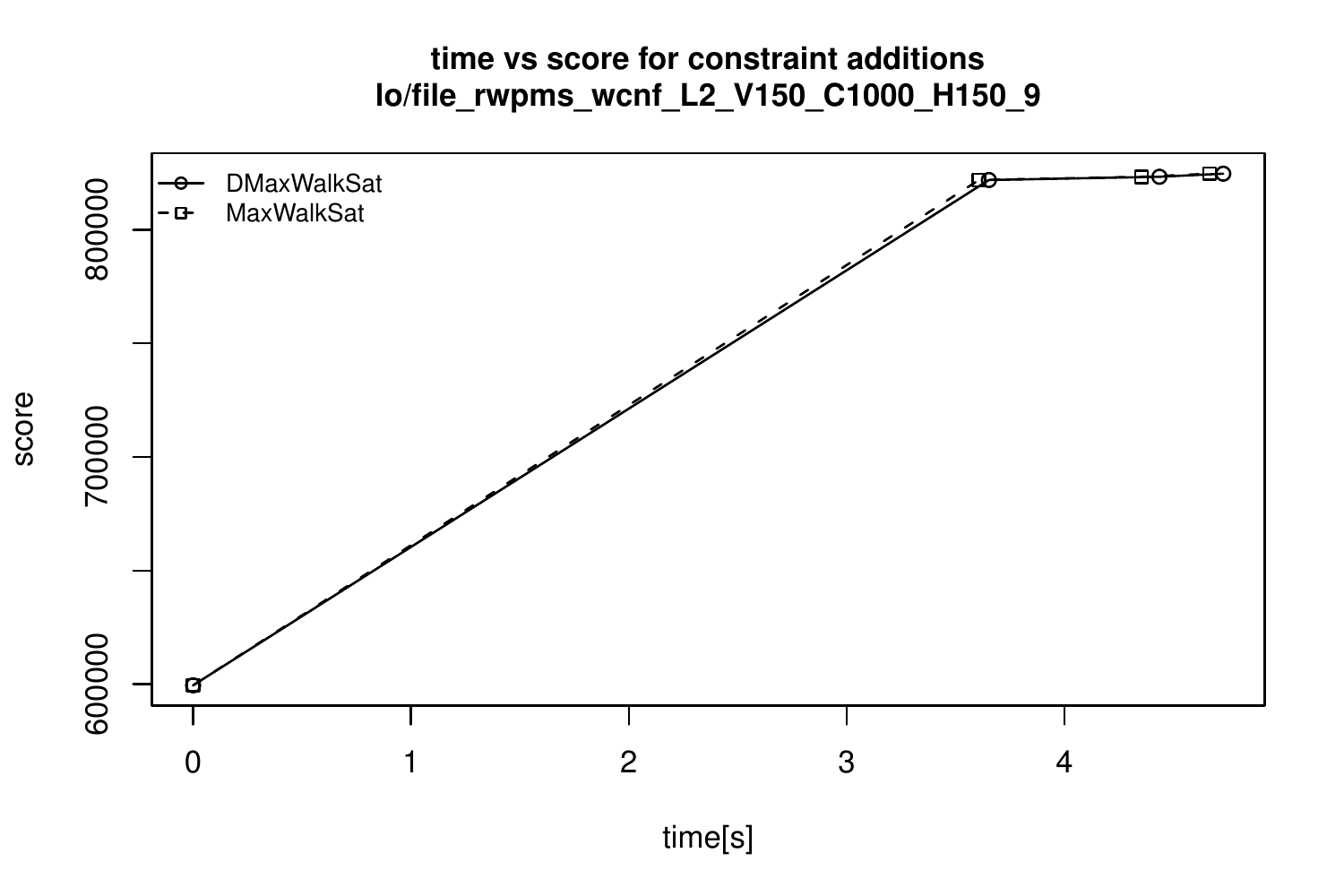}
        }
        \qquad
        \subfloat[Constraint removal]
        {
            \includegraphics[width=2.7in]{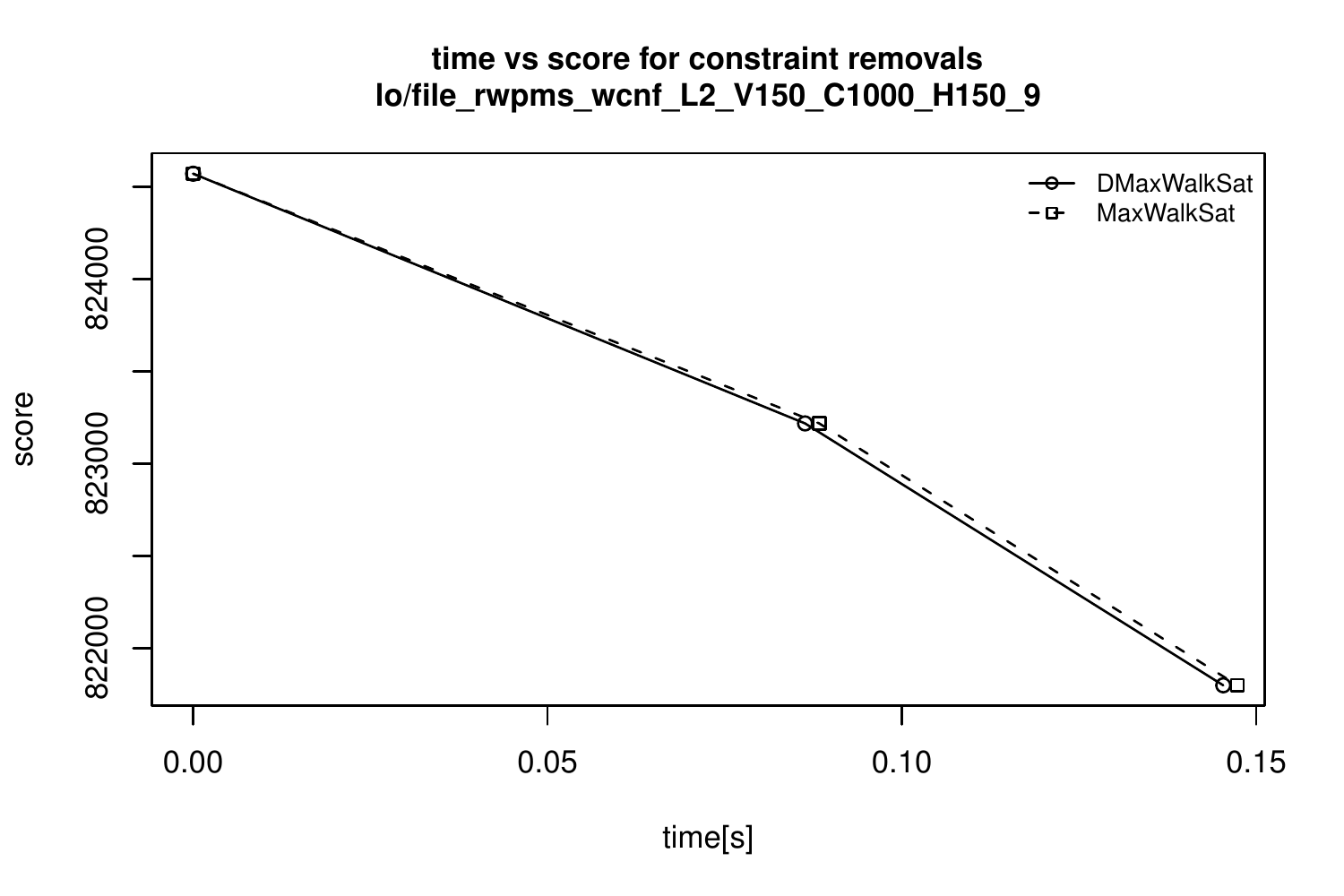}
        }

    \caption*{lo/file\_rwpms\_wcnf\_L2\_V150\_C1000\_H150\_9}
    \label{fig_lo/file_rwpms_wcnf_L2_V150_C1000_H150_9}
\end{figure}

\begin{figure}[H]
    \setcounter{subfigure}{0}
    \centering
        \subfloat[Constraint addition]
        {
            \includegraphics[width=2.7in]{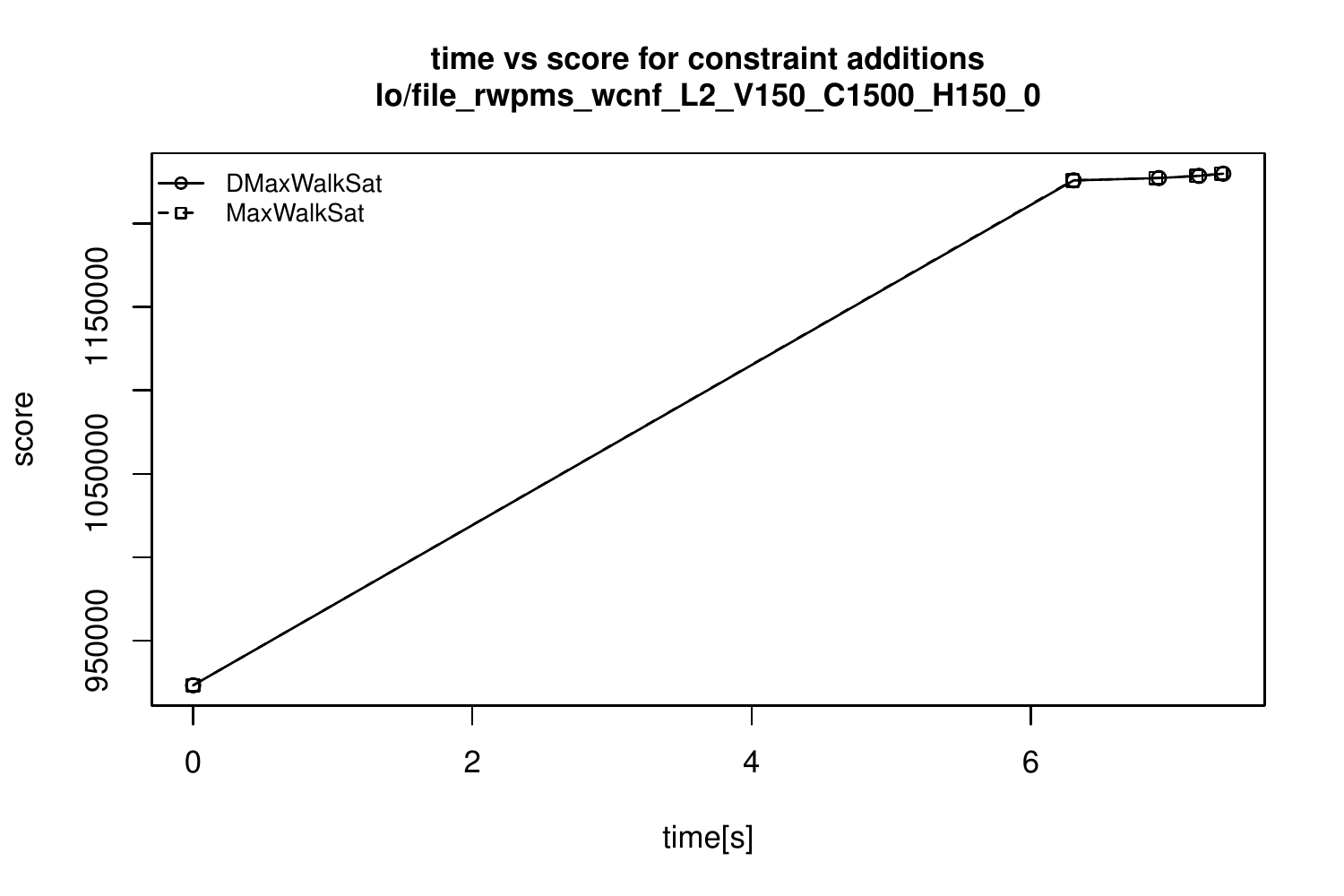}
        }
        \qquad
        \subfloat[Constraint removal]
        {
            \includegraphics[width=2.7in]{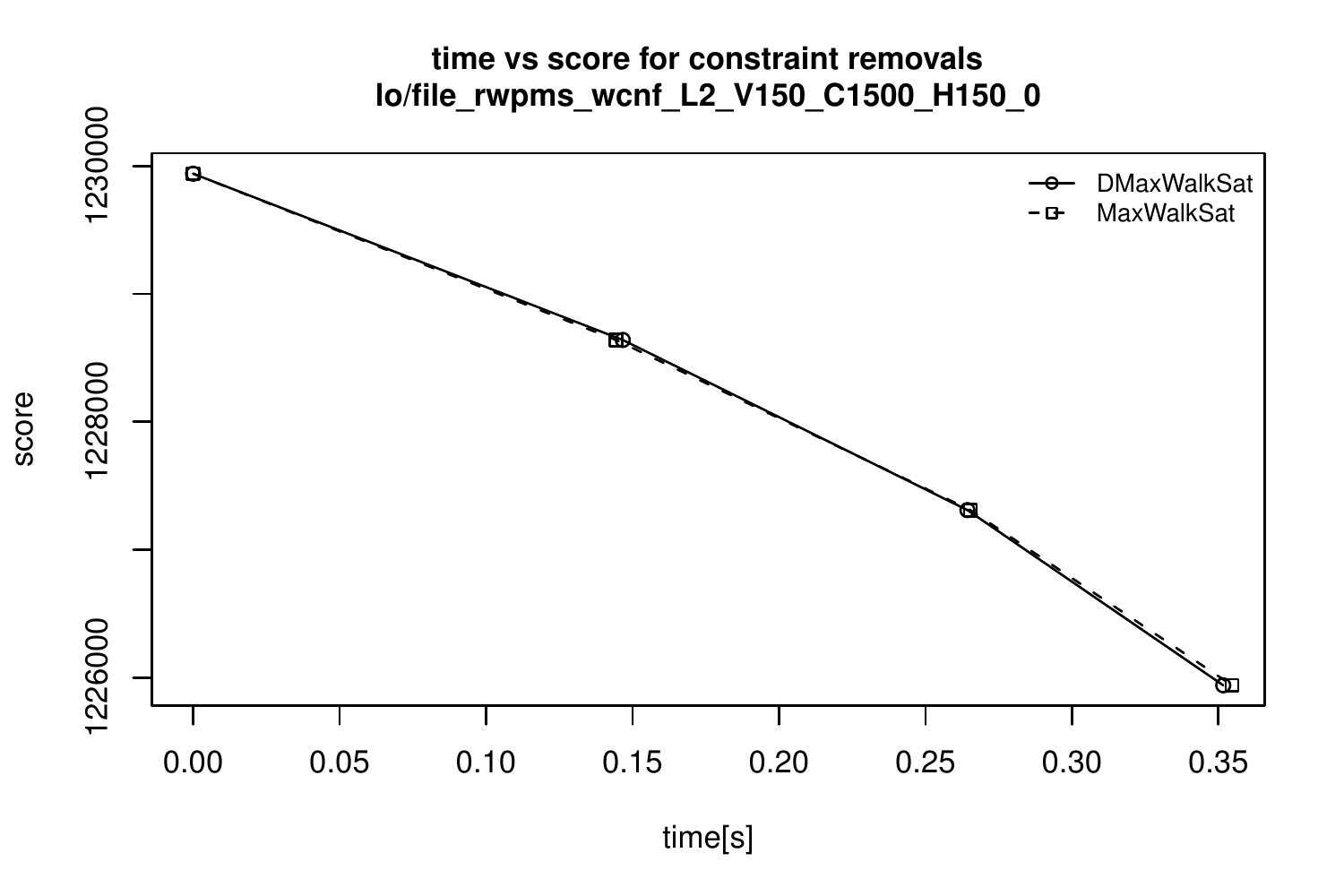}
        }

    \caption*{lo/file\_rwpms\_wcnf\_L2\_V150\_C1500\_H150\_0}
    \label{fig_lo/file_rwpms_wcnf_L2_V150_C1500_H150_0}
\end{figure}

\begin{figure}[H]
    \setcounter{subfigure}{0}
    \centering
        \subfloat[Constraint addition]
        {
            \includegraphics[width=2.7in]{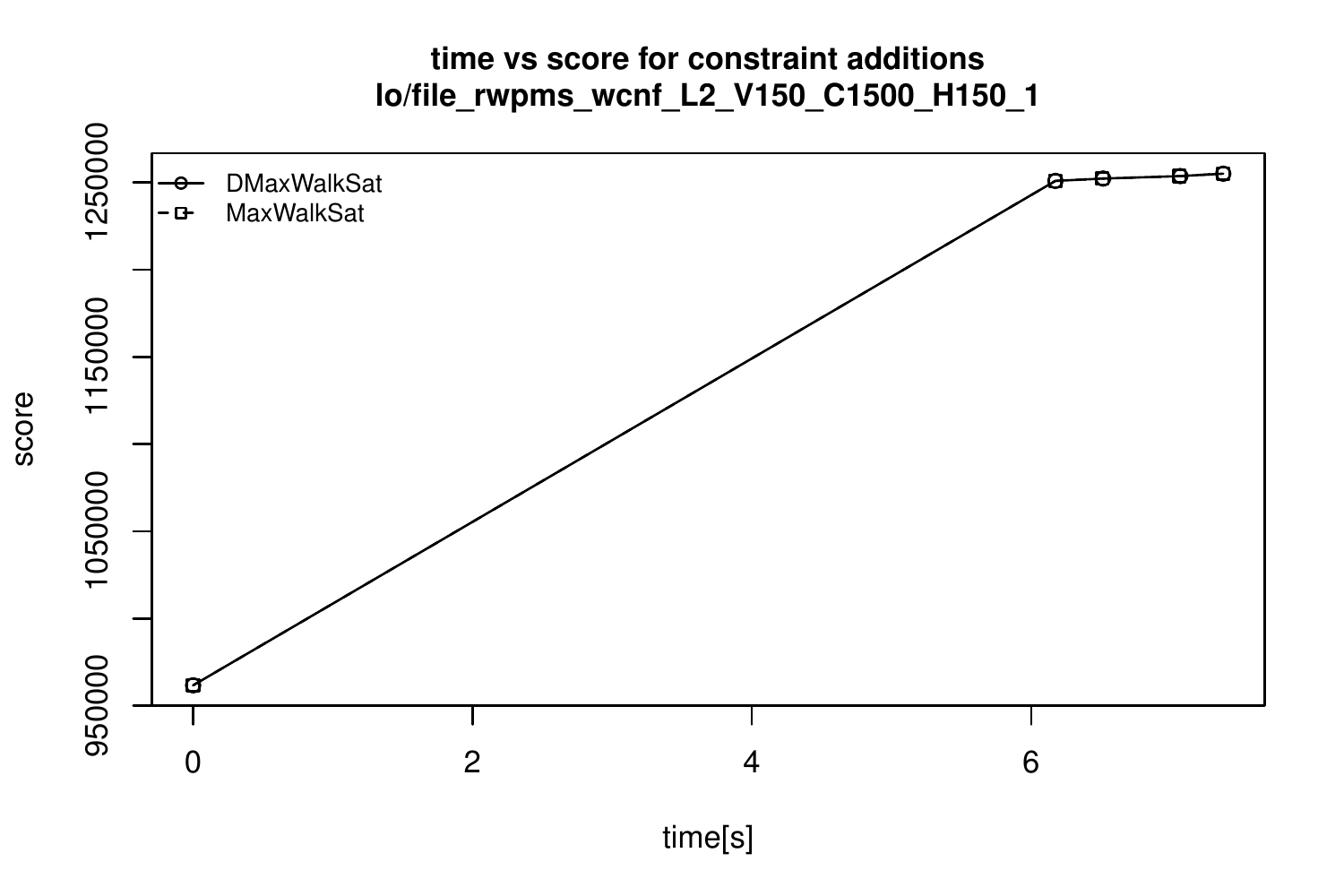}
        }
        \qquad
        \subfloat[Constraint removal]
        {
            \includegraphics[width=2.7in]{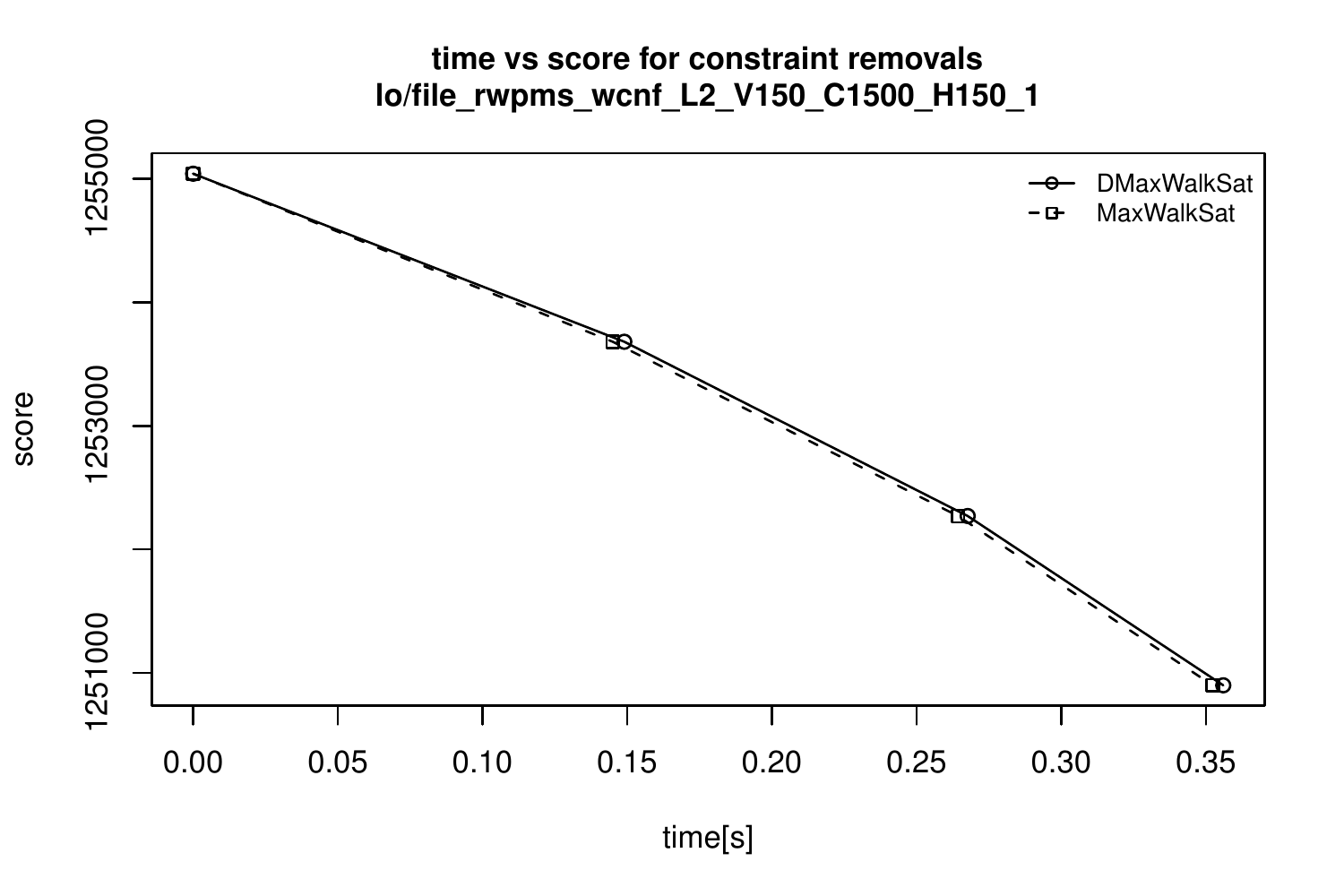}
        }

    \caption*{lo/file\_rwpms\_wcnf\_L2\_V150\_C1500\_H150\_1}
    \label{fig_lo/file_rwpms_wcnf_L2_V150_C1500_H150_1}
\end{figure}

\begin{figure}[H]
    \setcounter{subfigure}{0}
    \centering
        \subfloat[Constraint addition]
        {
            \includegraphics[width=2.7in]{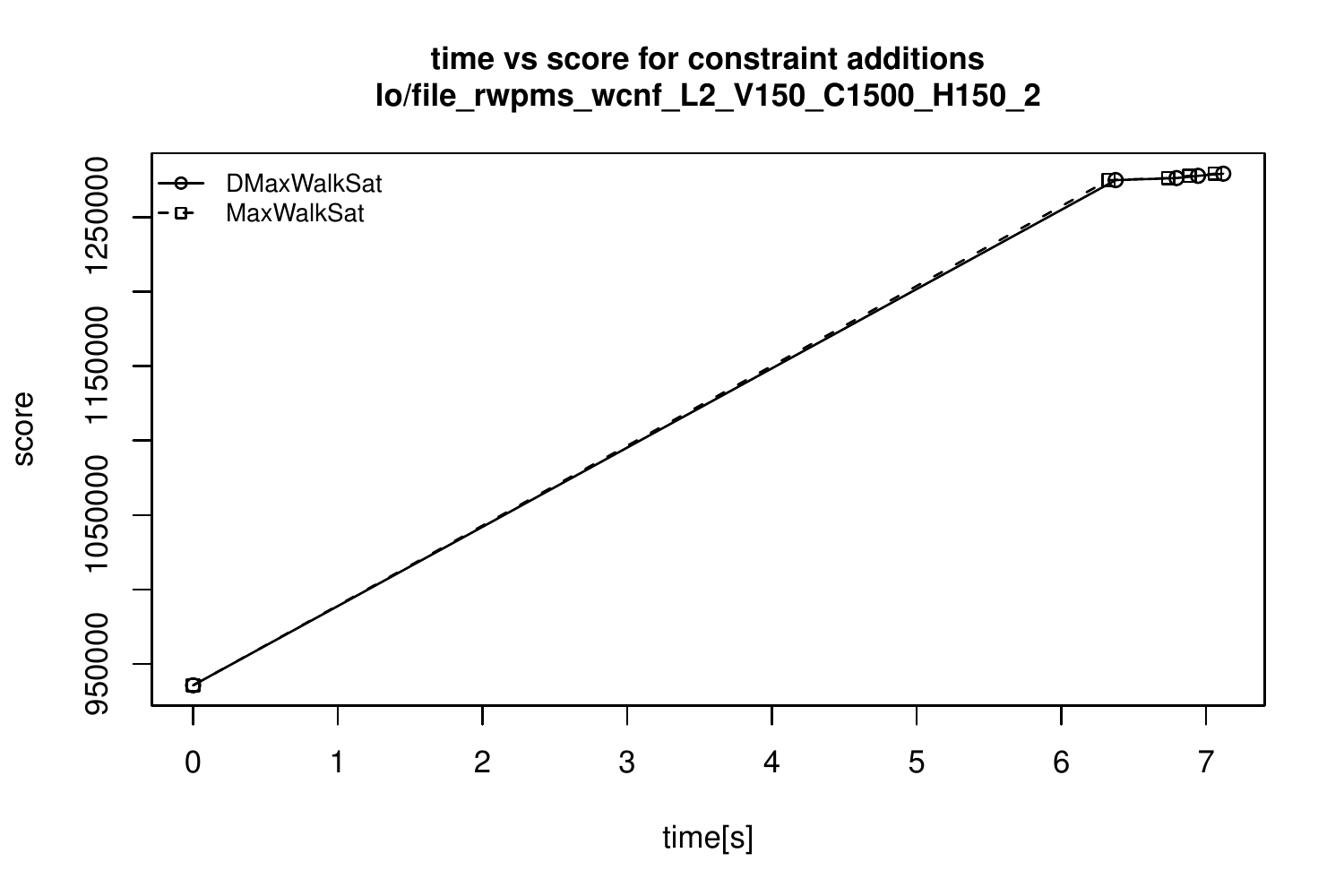}
        }
        \qquad
        \subfloat[Constraint removal]
        {
            \includegraphics[width=2.7in]{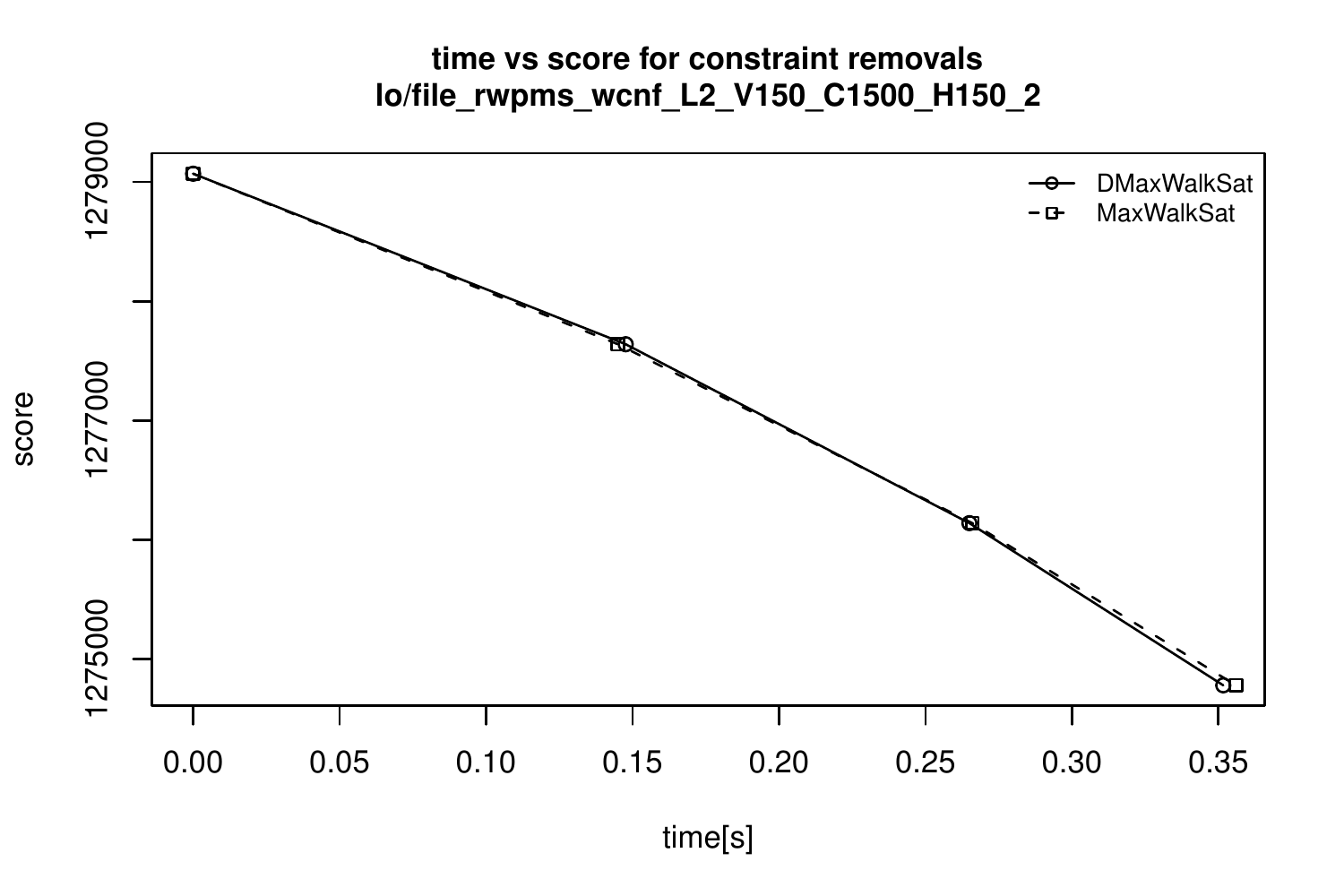}
        }

    \caption*{lo/file\_rwpms\_wcnf\_L2\_V150\_C1500\_H150\_2}
    \label{fig_lo/file_rwpms_wcnf_L2_V150_C1500_H150_2}
\end{figure}

\begin{figure}[H]
    \setcounter{subfigure}{0}
    \centering
        \subfloat[Constraint addition]
        {
            \includegraphics[width=2.7in]{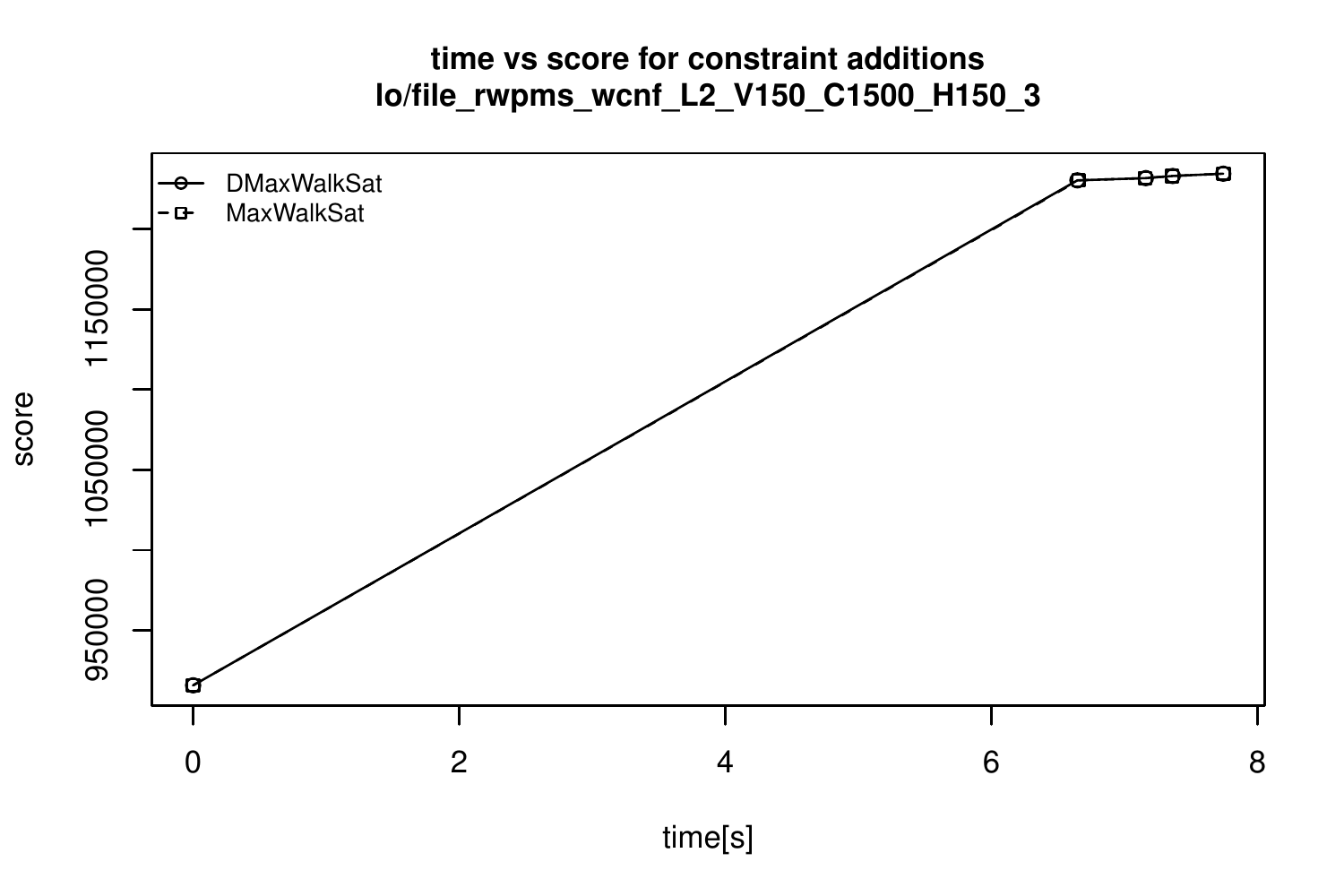}
        }
        \qquad
        \subfloat[Constraint removal]
        {
            \includegraphics[width=2.7in]{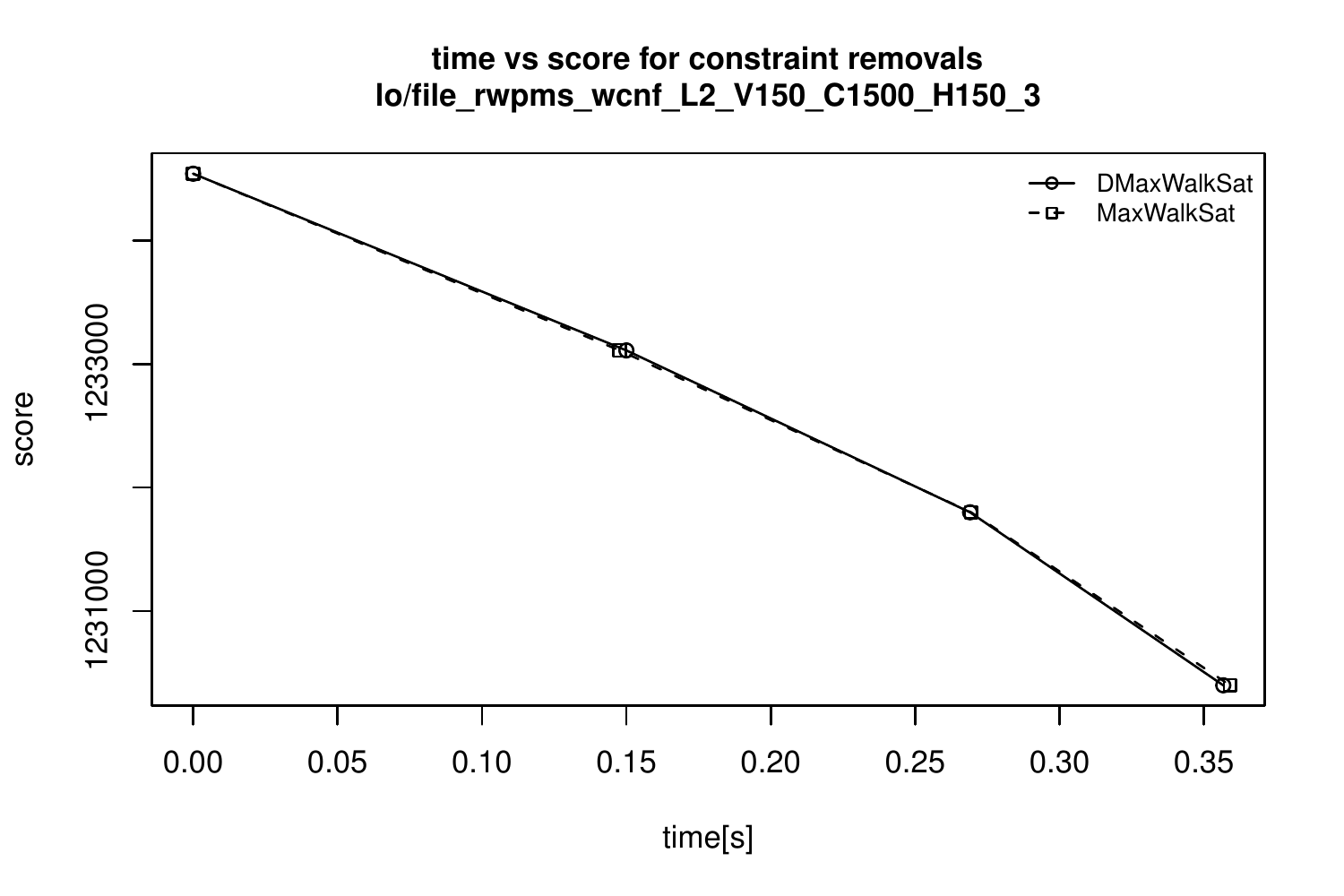}
        }

    \caption*{lo/file\_rwpms\_wcnf\_L2\_V150\_C1500\_H150\_3}
    \label{fig_lo/file_rwpms_wcnf_L2_V150_C1500_H150_3}
\end{figure}

\begin{figure}[H]
    \setcounter{subfigure}{0}
    \centering
        \subfloat[Constraint addition]
        {
            \includegraphics[width=2.7in]{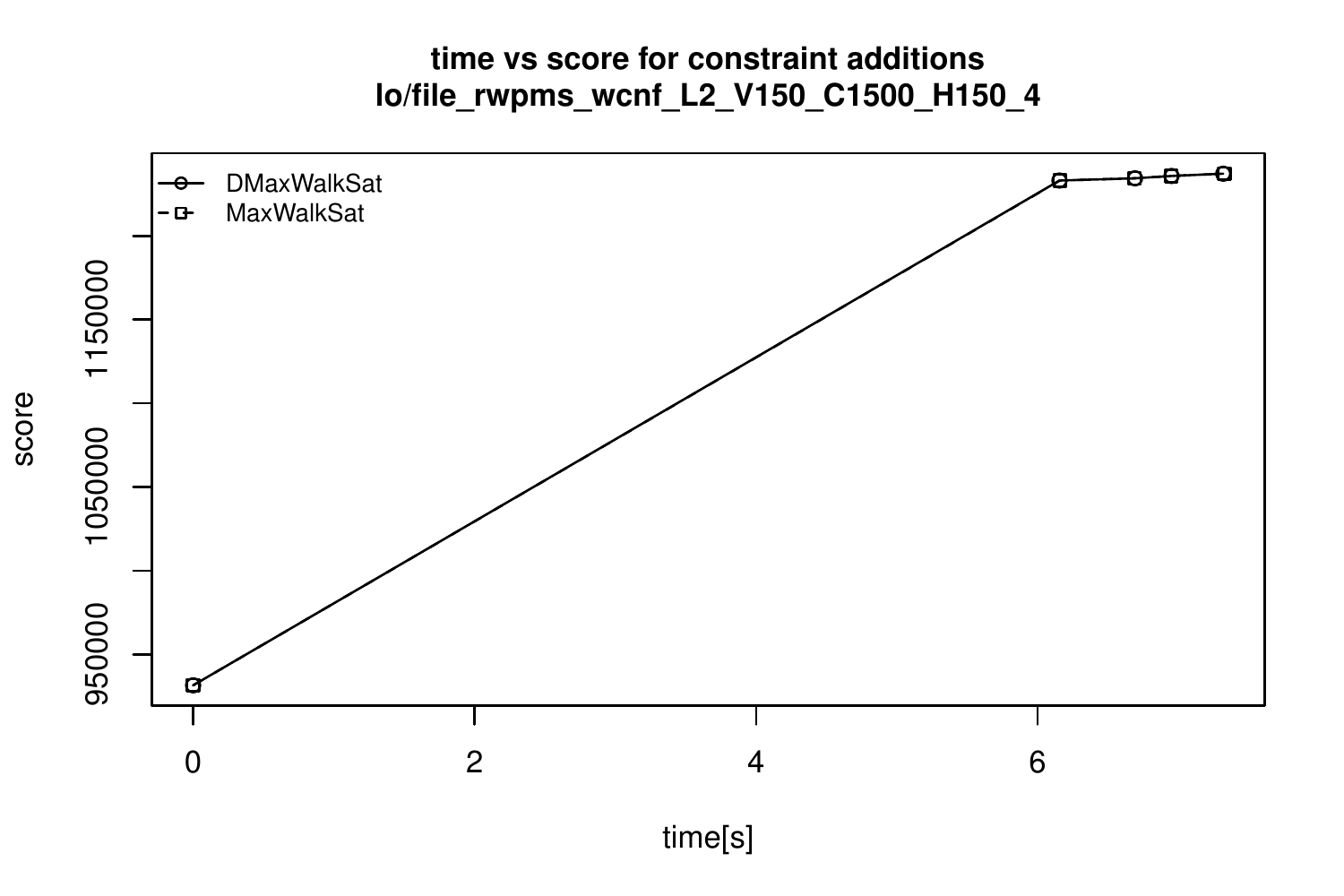}
        }
        \qquad
        \subfloat[Constraint removal]
        {
            \includegraphics[width=2.7in]{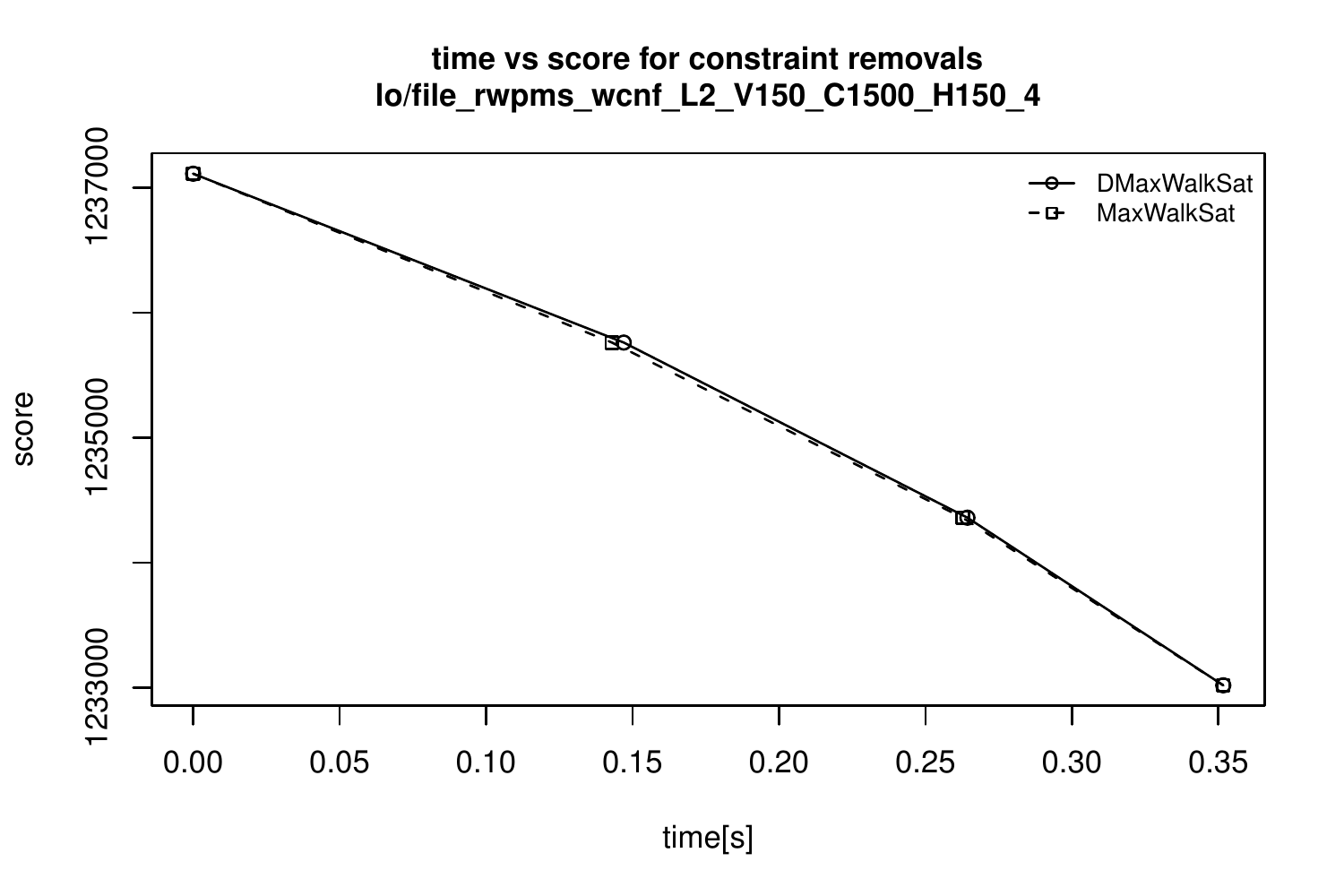}
        }

    \caption*{lo/file\_rwpms\_wcnf\_L2\_V150\_C1500\_H150\_4}
    \label{fig_lo/file_rwpms_wcnf_L2_V150_C1500_H150_4}
\end{figure}

\begin{figure}[H]
    \setcounter{subfigure}{0}
    \centering
        \subfloat[Constraint addition]
        {
            \includegraphics[width=2.7in]{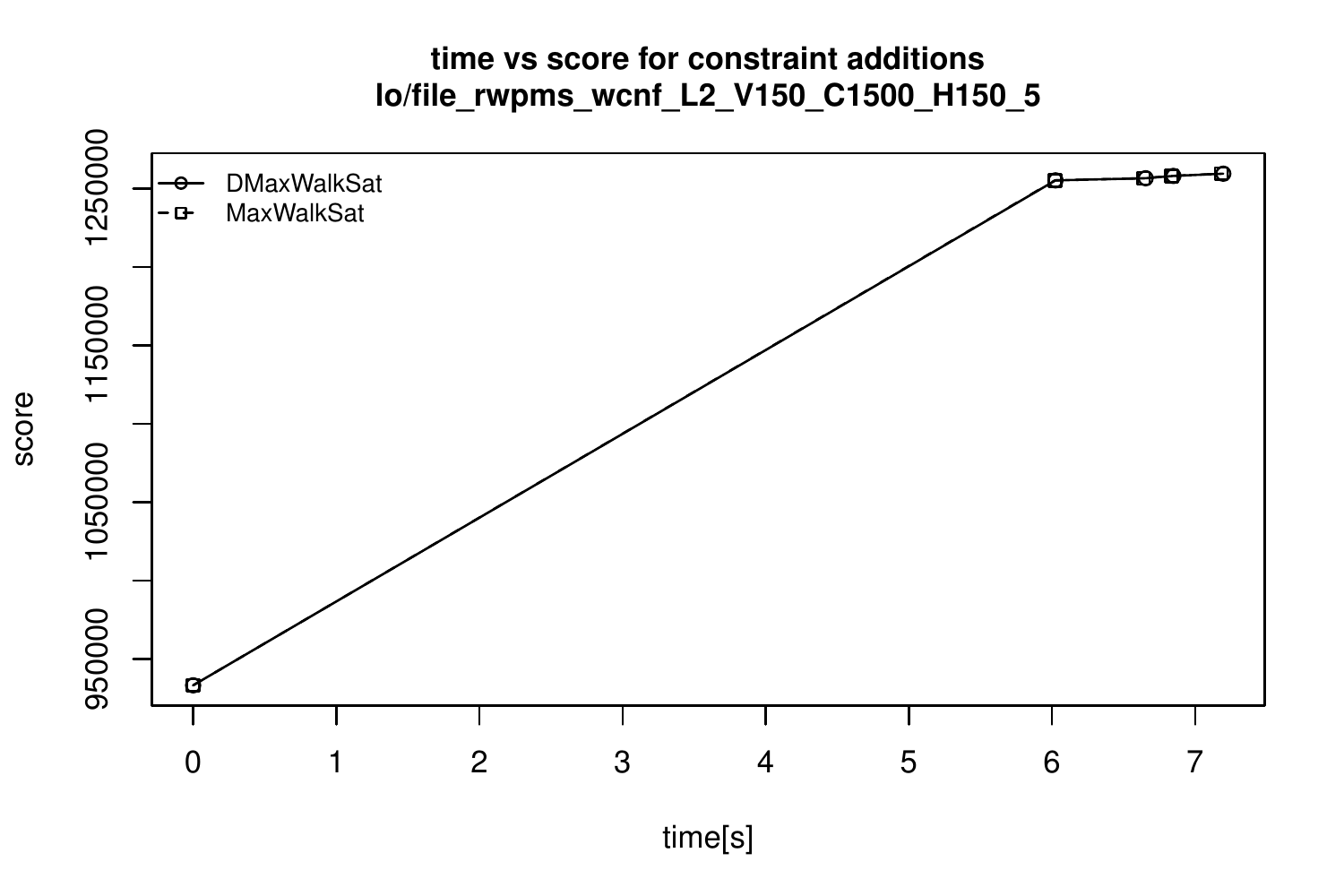}
        }
        \qquad
        \subfloat[Constraint removal]
        {
            \includegraphics[width=2.7in]{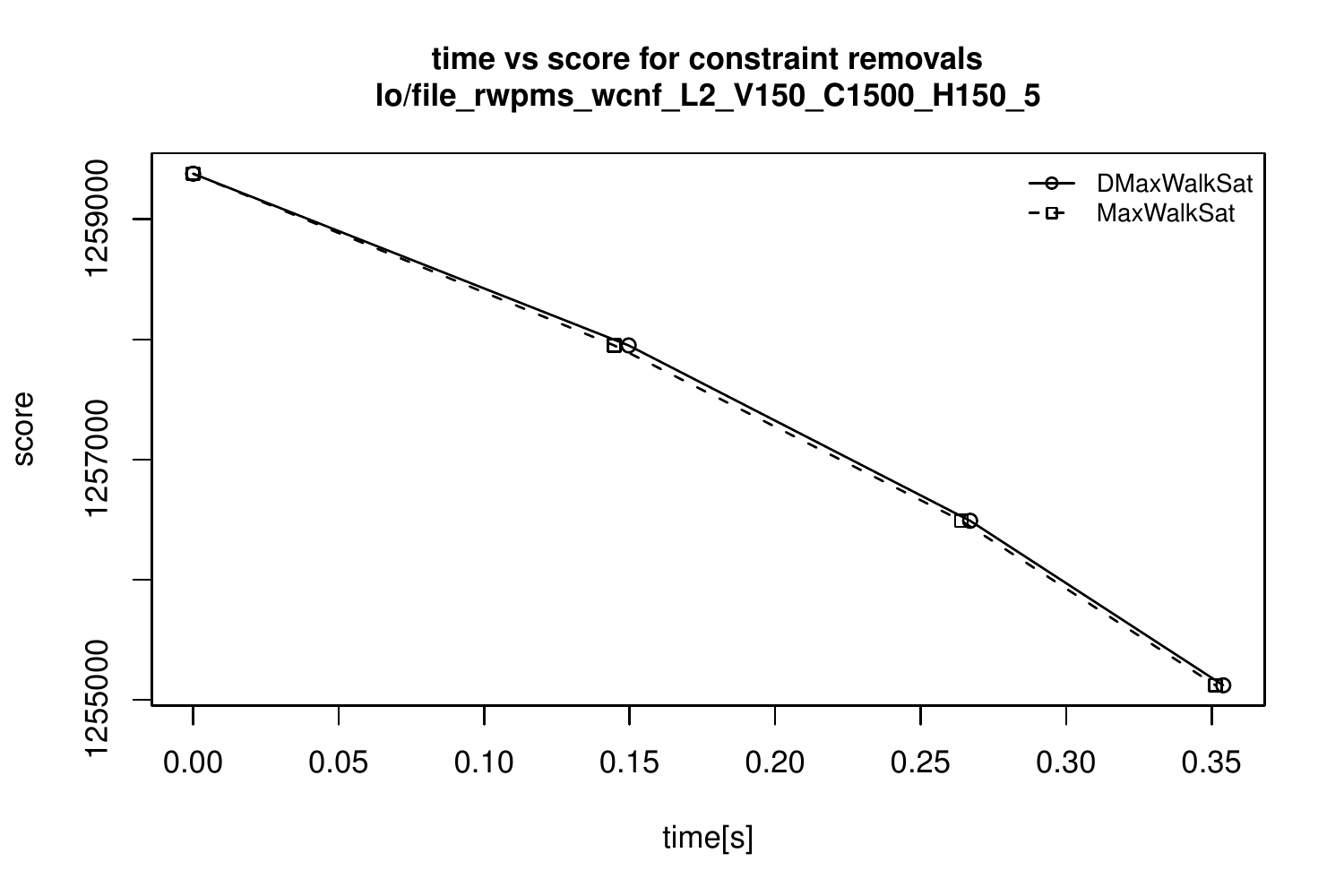}
        }

    \caption*{lo/file\_rwpms\_wcnf\_L2\_V150\_C1500\_H150\_5}
    \label{fig_lo/file_rwpms_wcnf_L2_V150_C1500_H150_5}
\end{figure}

\begin{figure}[H]
    \setcounter{subfigure}{0}
    \centering
        \subfloat[Constraint addition]
        {
            \includegraphics[width=2.7in]{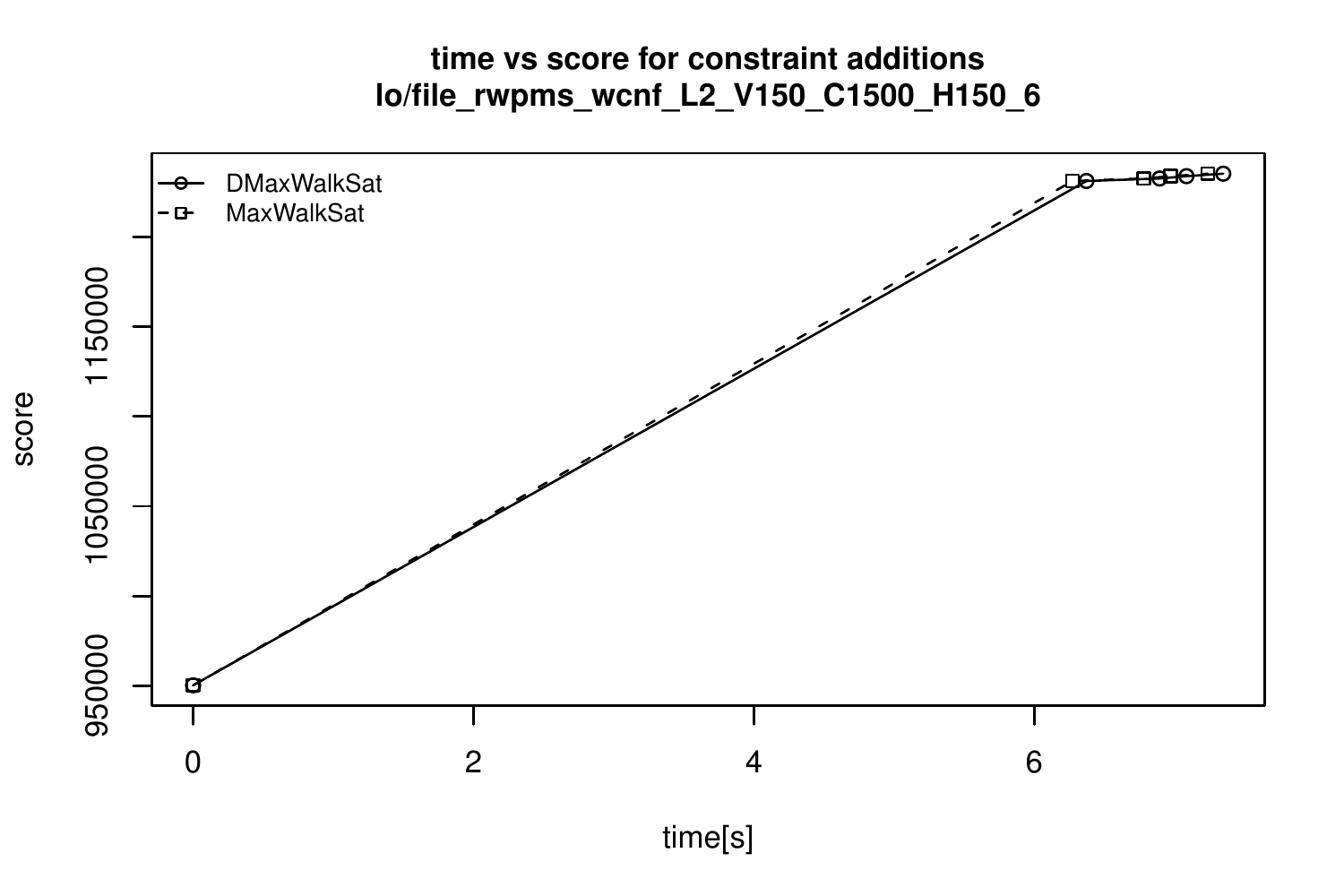}
        }
        \qquad
        \subfloat[Constraint removal]
        {
            \includegraphics[width=2.7in]{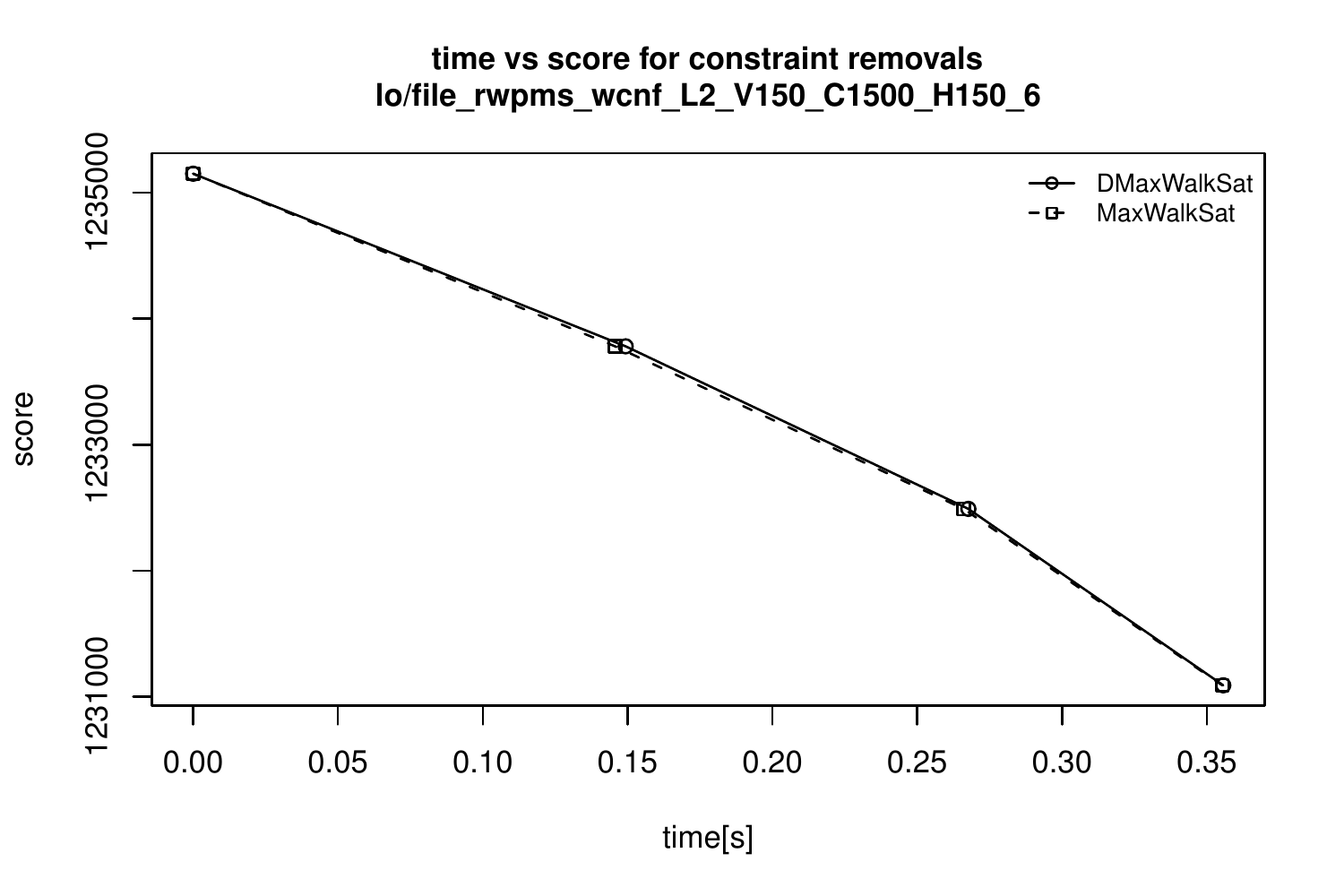}
        }

    \caption*{lo/file\_rwpms\_wcnf\_L2\_V150\_C1500\_H150\_6}
    \label{fig_lo/file_rwpms_wcnf_L2_V150_C1500_H150_6}
\end{figure}

\begin{figure}[H]
    \setcounter{subfigure}{0}
    \centering
        \subfloat[Constraint addition]
        {
            \includegraphics[width=2.7in]{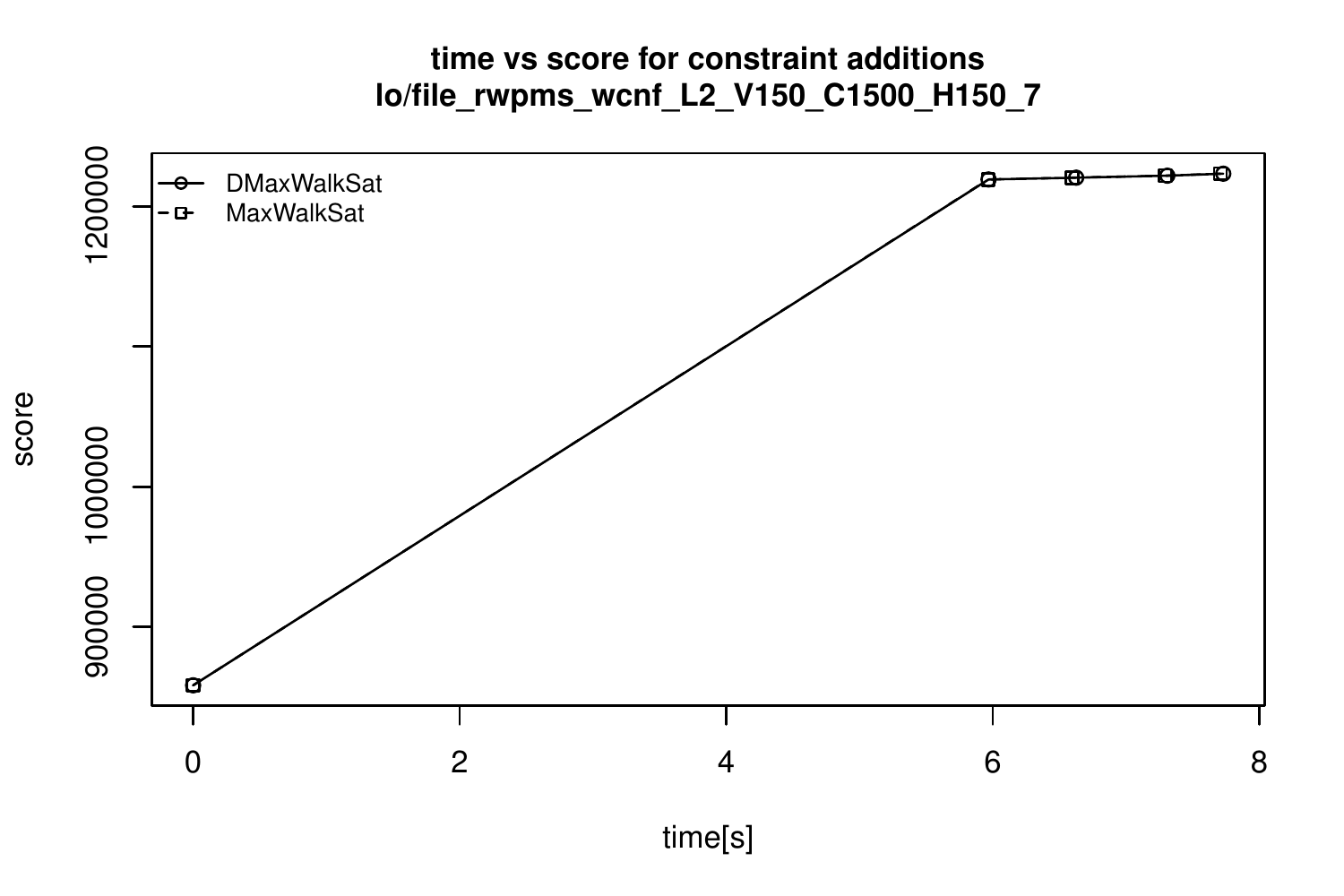}
        }
        \qquad
        \subfloat[Constraint removal]
        {
            \includegraphics[width=2.7in]{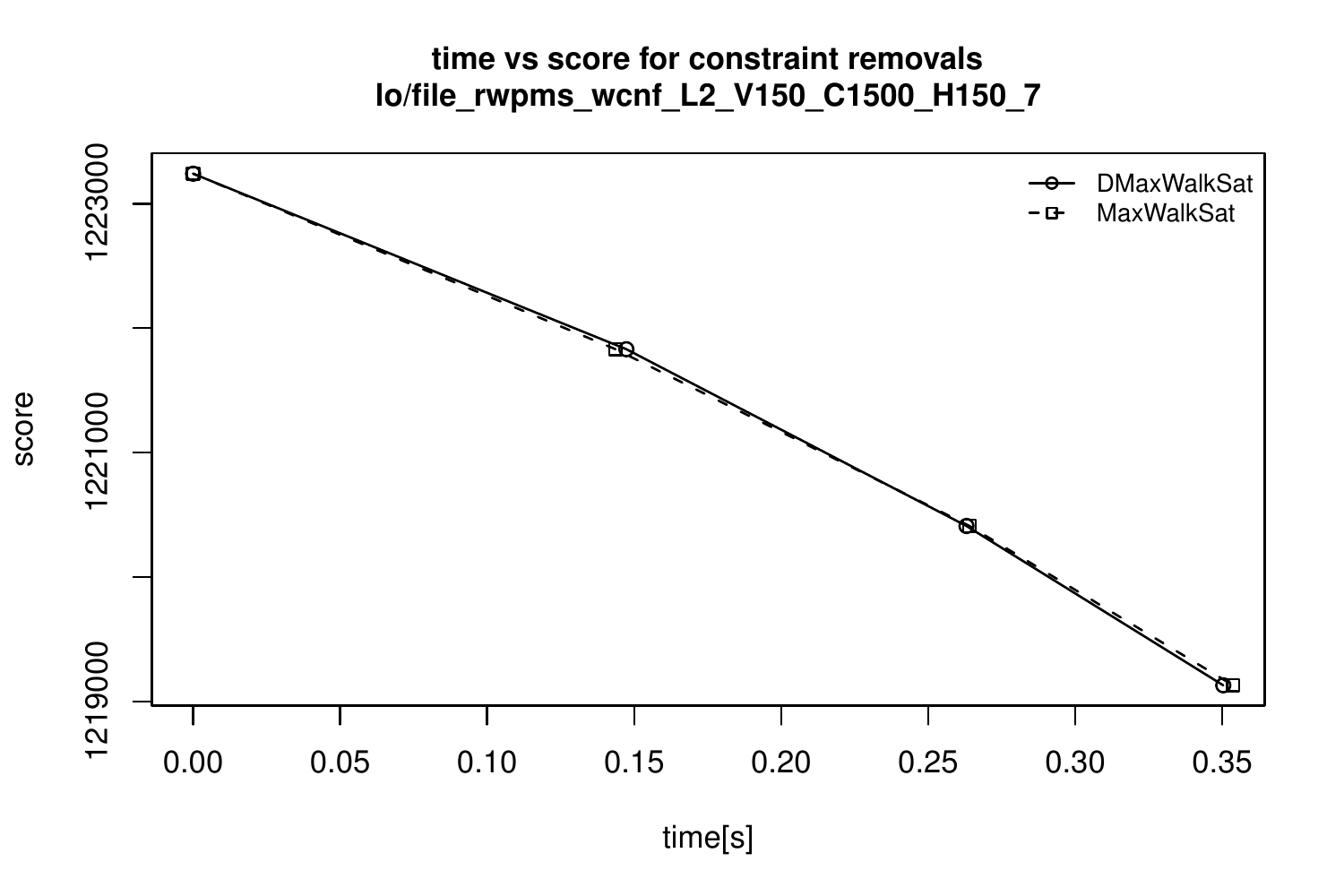}
        }

    \caption*{lo/file\_rwpms\_wcnf\_L2\_V150\_C1500\_H150\_7}
    \label{fig_lo/file_rwpms_wcnf_L2_V150_C1500_H150_7}
\end{figure}

\begin{figure}[H]
    \setcounter{subfigure}{0}
    \centering
        \subfloat[Constraint addition]
        {
            \includegraphics[width=2.7in]{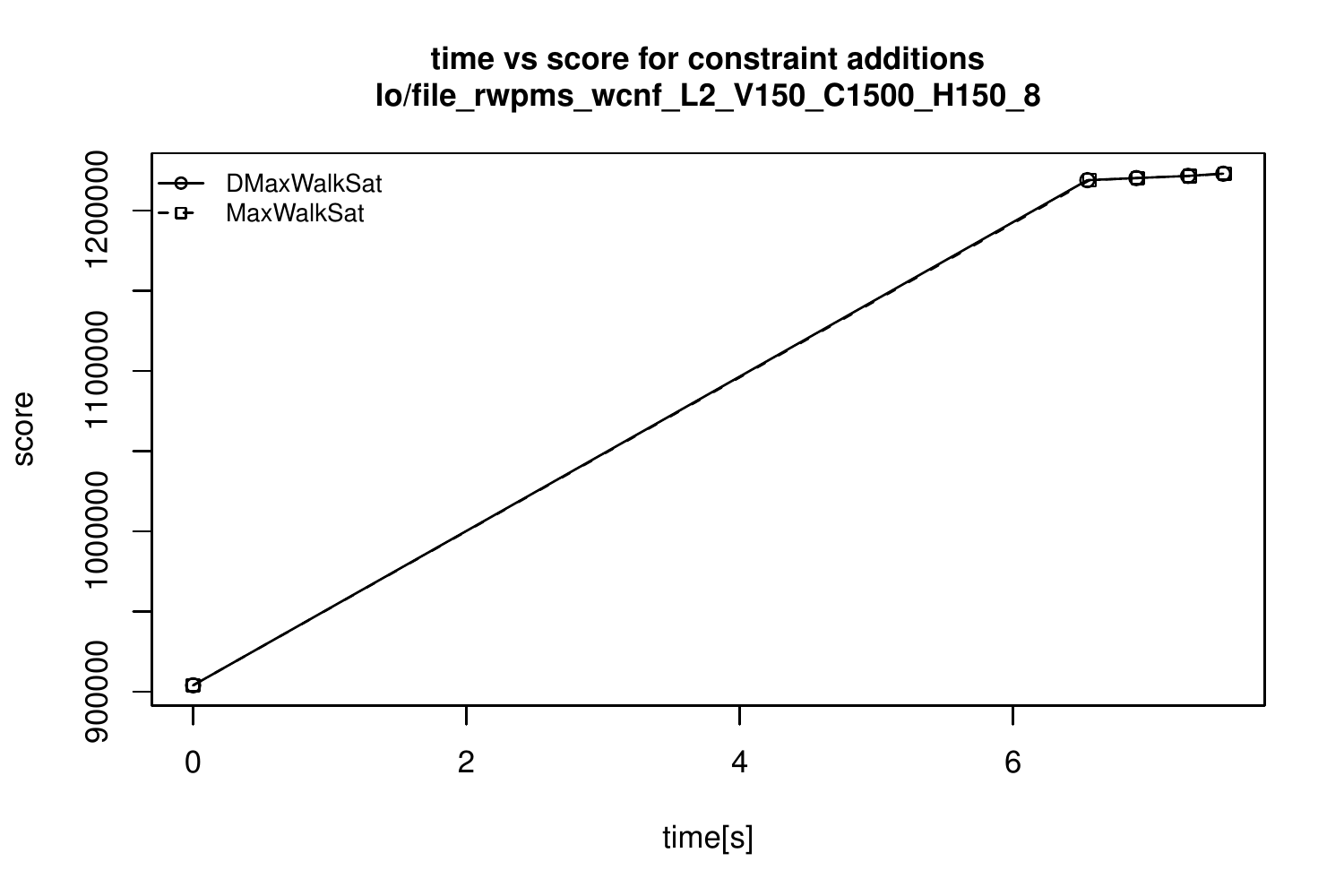}
        }
        \qquad
        \subfloat[Constraint removal]
        {
            \includegraphics[width=2.7in]{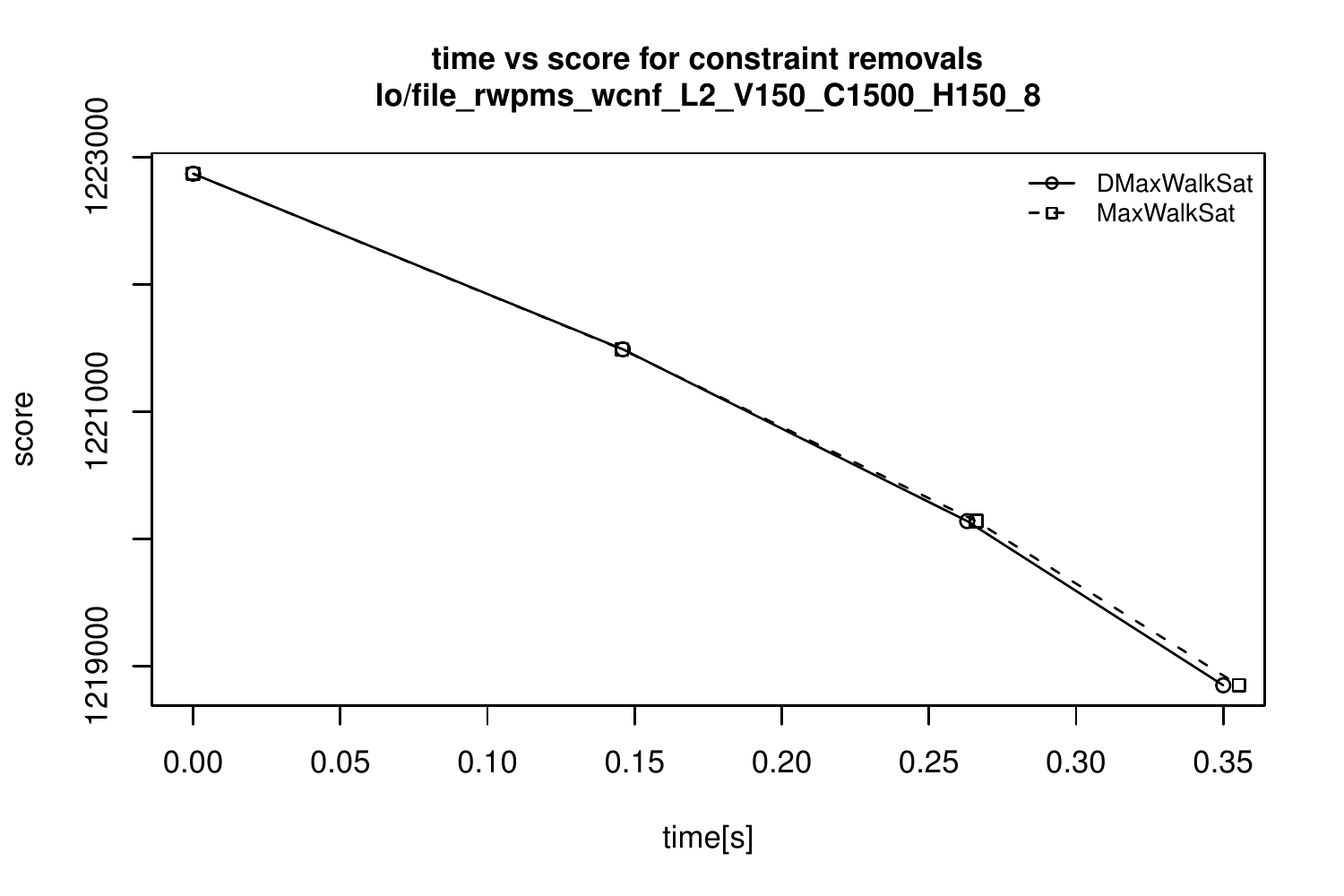}
        }

    \caption*{lo/file\_rwpms\_wcnf\_L2\_V150\_C1500\_H150\_8}
    \label{fig_lo/file_rwpms_wcnf_L2_V150_C1500_H150_8}
\end{figure}

\begin{figure}[H]
    \setcounter{subfigure}{0}
    \centering
        \subfloat[Constraint addition]
        {
            \includegraphics[width=2.7in]{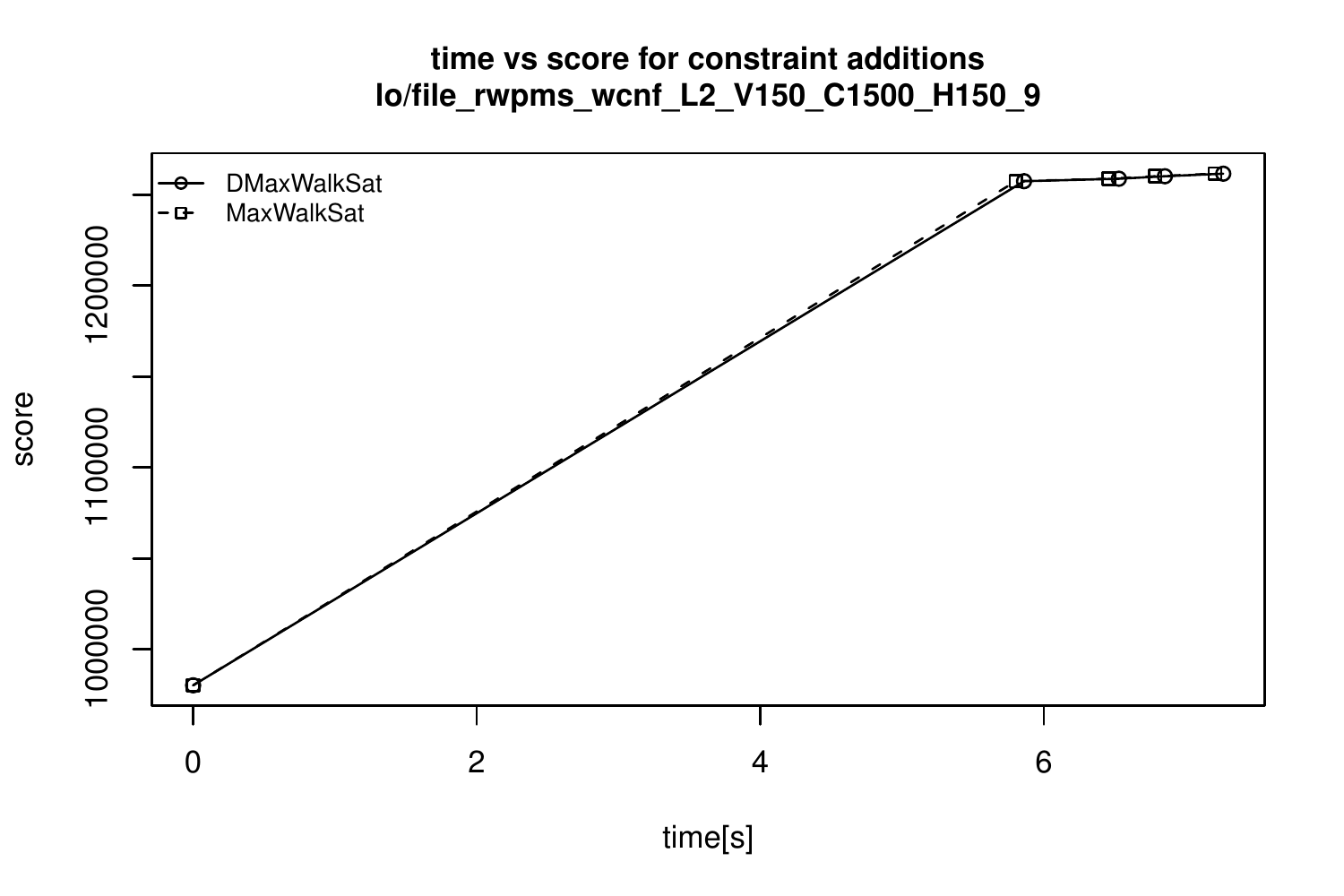}
        }
        \qquad
        \subfloat[Constraint removal]
        {
            \includegraphics[width=2.7in]{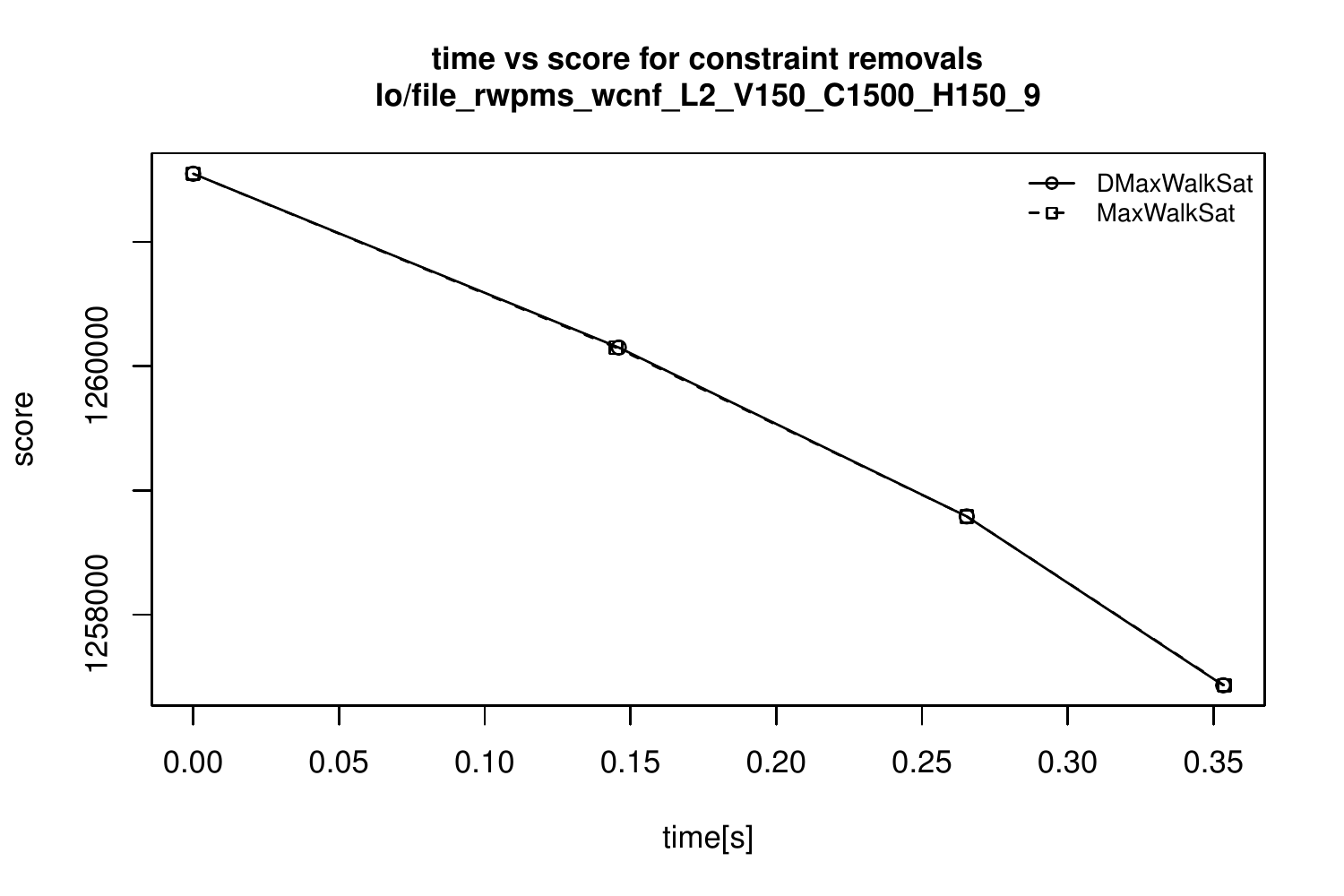}
        }

    \caption*{lo/file\_rwpms\_wcnf\_L2\_V150\_C1500\_H150\_9}
    \label{fig_lo/file_rwpms_wcnf_L2_V150_C1500_H150_9}
\end{figure}

\begin{figure}[H]
    \setcounter{subfigure}{0}
    \centering
        \subfloat[Constraint addition]
        {
            \includegraphics[width=2.7in]{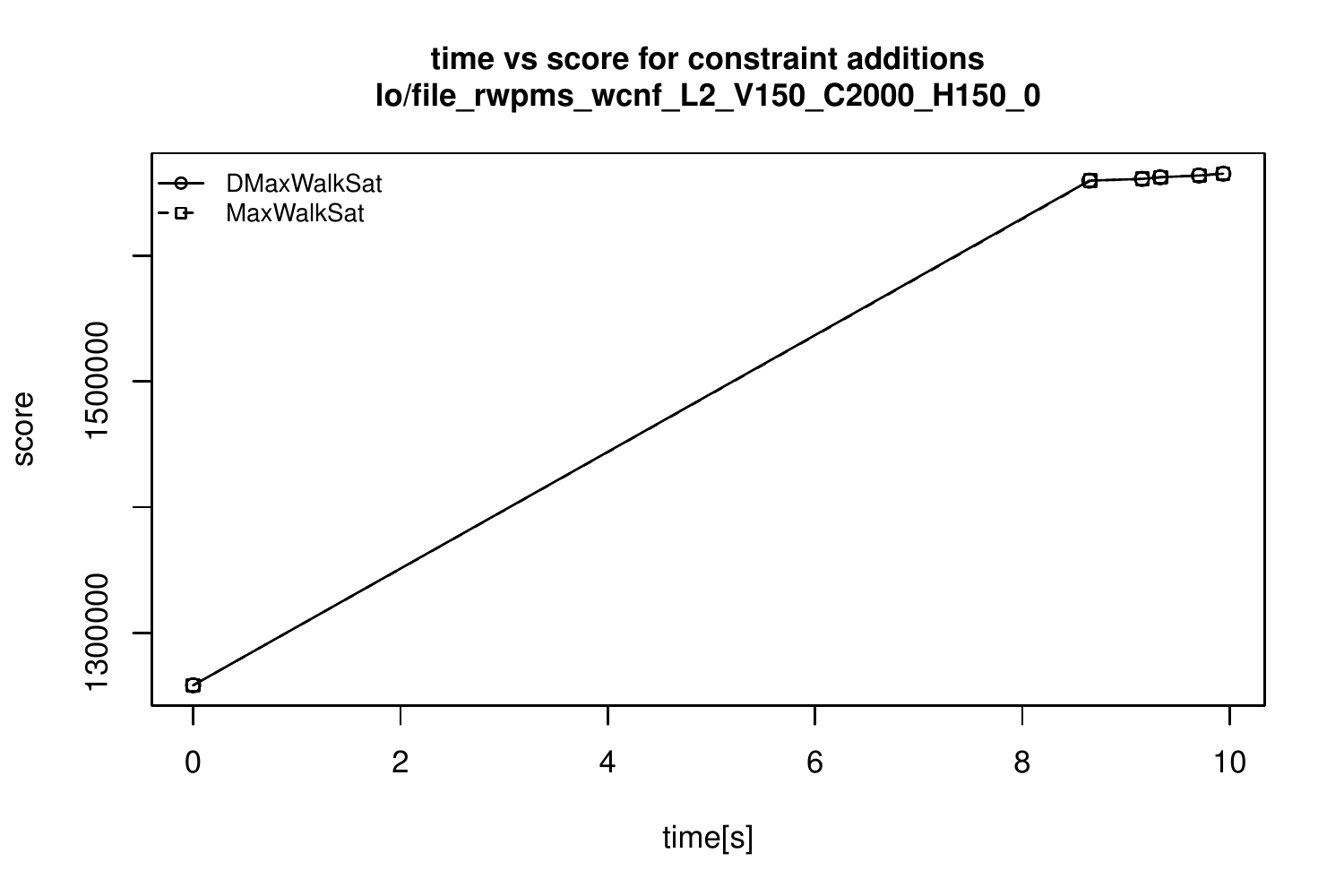}
        }
        \qquad
        \subfloat[Constraint removal]
        {
            \includegraphics[width=2.7in]{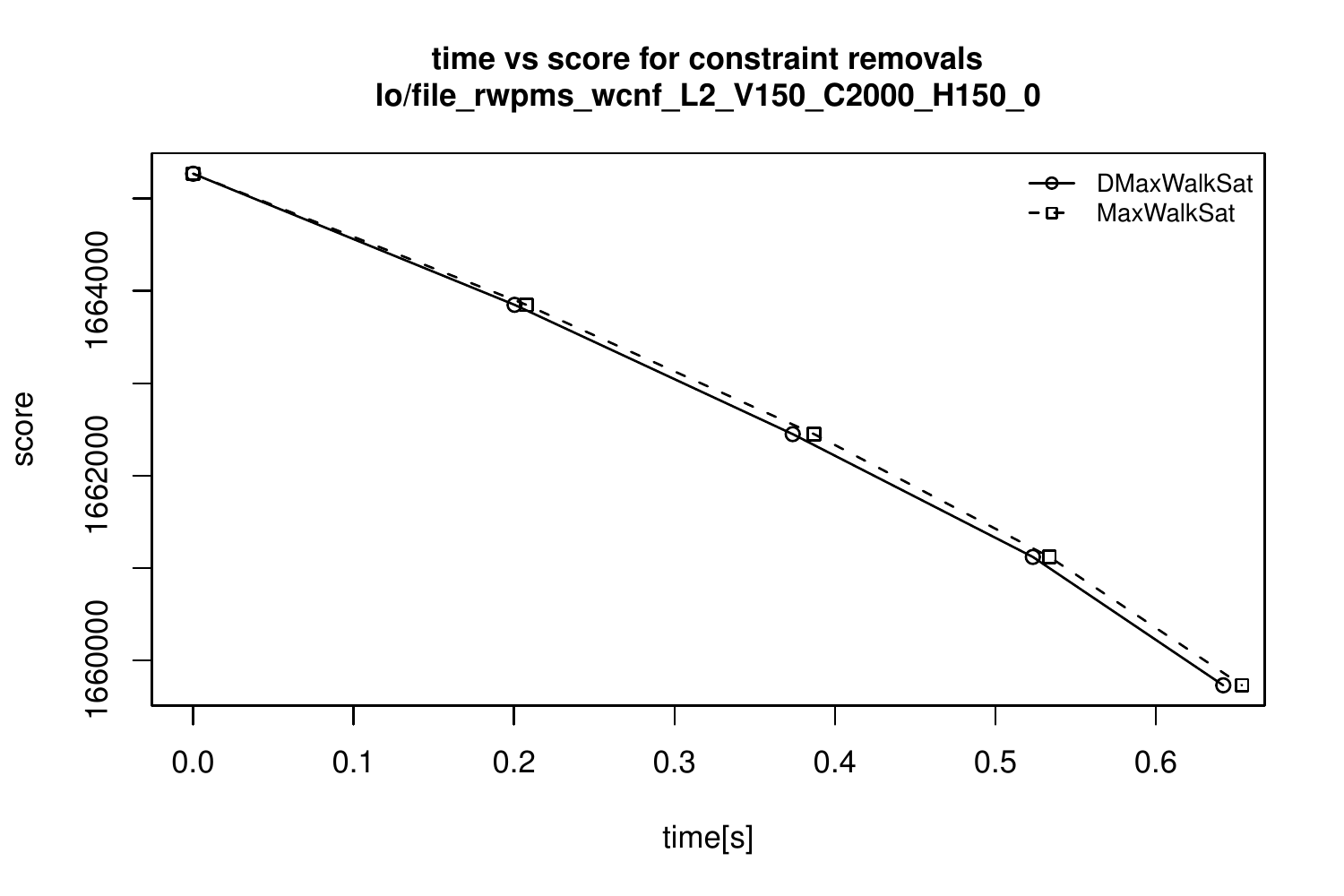}
        }

    \caption*{lo/file\_rwpms\_wcnf\_L2\_V150\_C2000\_H150\_0}
    \label{fig_lo/file_rwpms_wcnf_L2_V150_C2000_H150_0}
\end{figure}

\begin{figure}[H]
    \setcounter{subfigure}{0}
    \centering
        \subfloat[Constraint addition]
        {
            \includegraphics[width=2.7in]{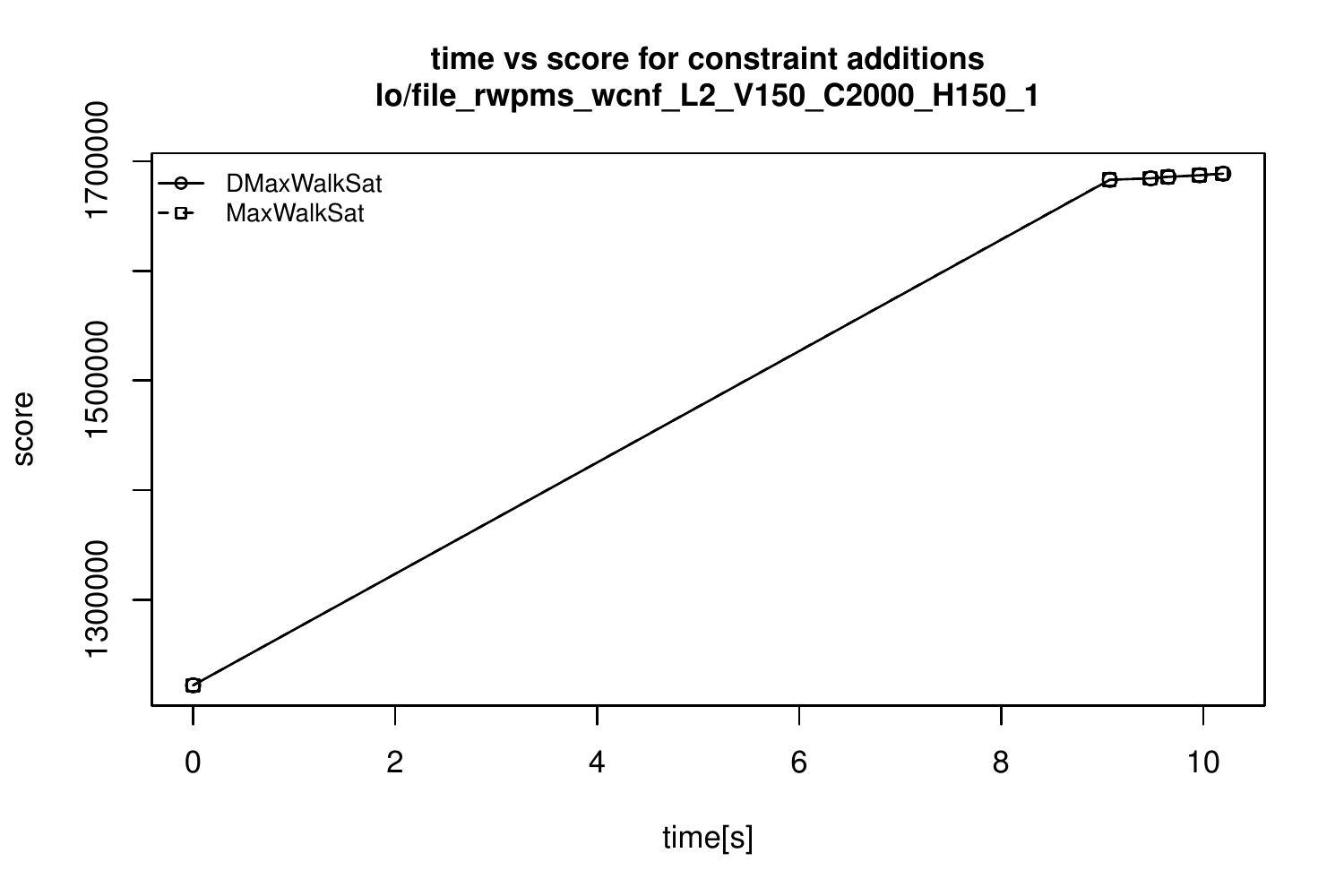}
        }
        \qquad
        \subfloat[Constraint removal]
        {
            \includegraphics[width=2.7in]{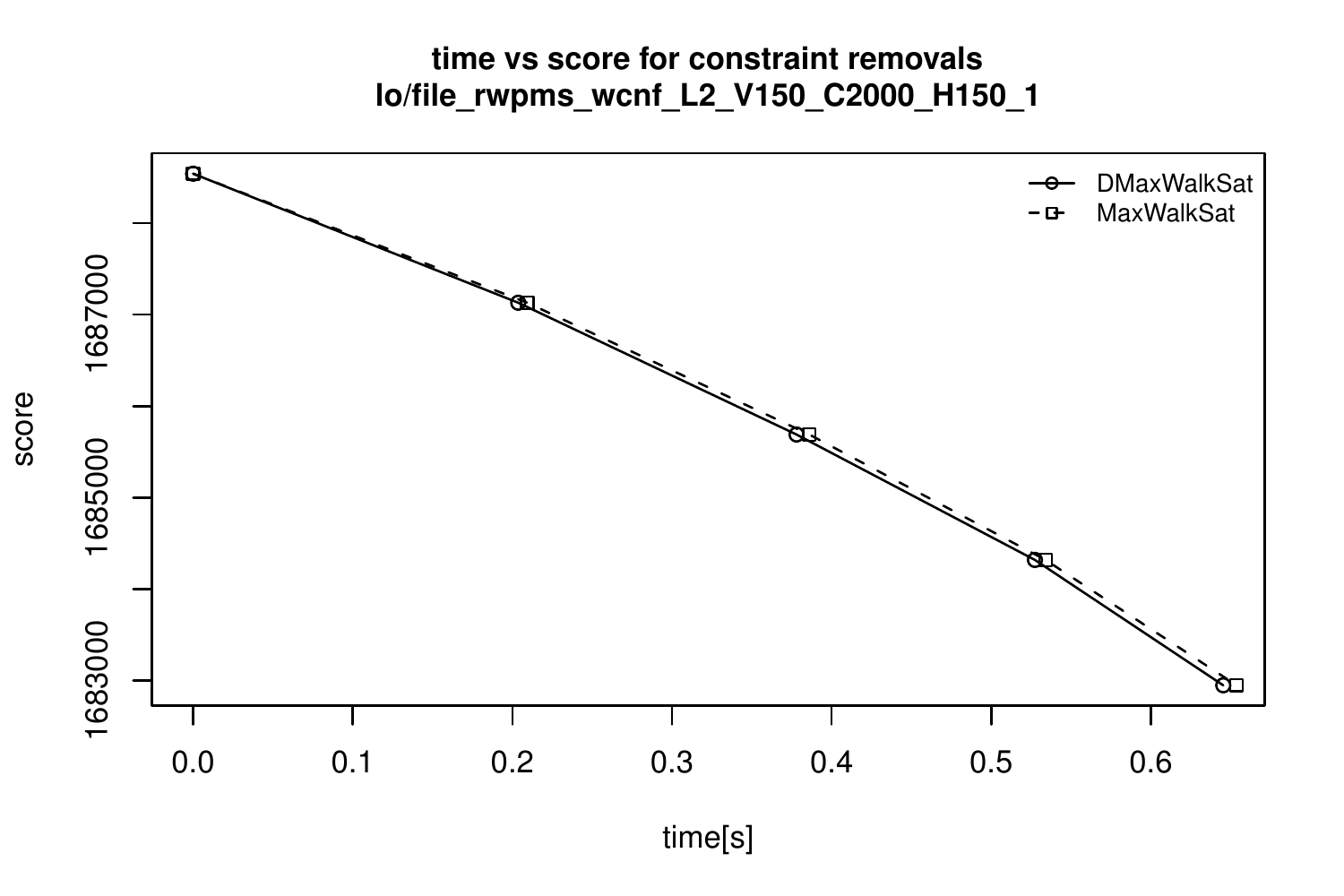}
        }

    \caption*{lo/file\_rwpms\_wcnf\_L2\_V150\_C2000\_H150\_1}
    \label{fig_lo/file_rwpms_wcnf_L2_V150_C2000_H150_1}
\end{figure}

\begin{figure}[H]
    \setcounter{subfigure}{0}
    \centering
        \subfloat[Constraint addition]
        {
            \includegraphics[width=2.7in]{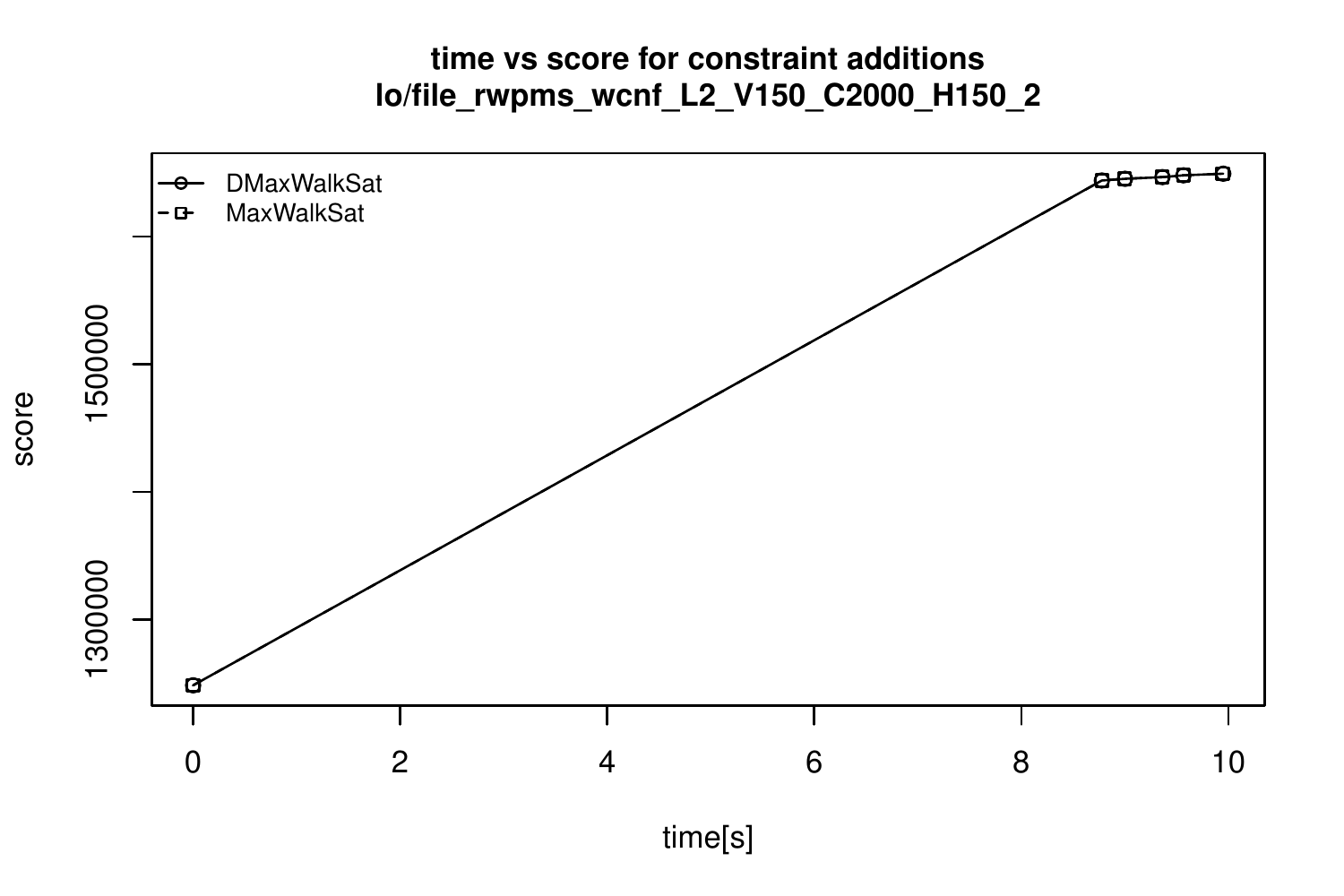}
        }
        \qquad
        \subfloat[Constraint removal]
        {
            \includegraphics[width=2.7in]{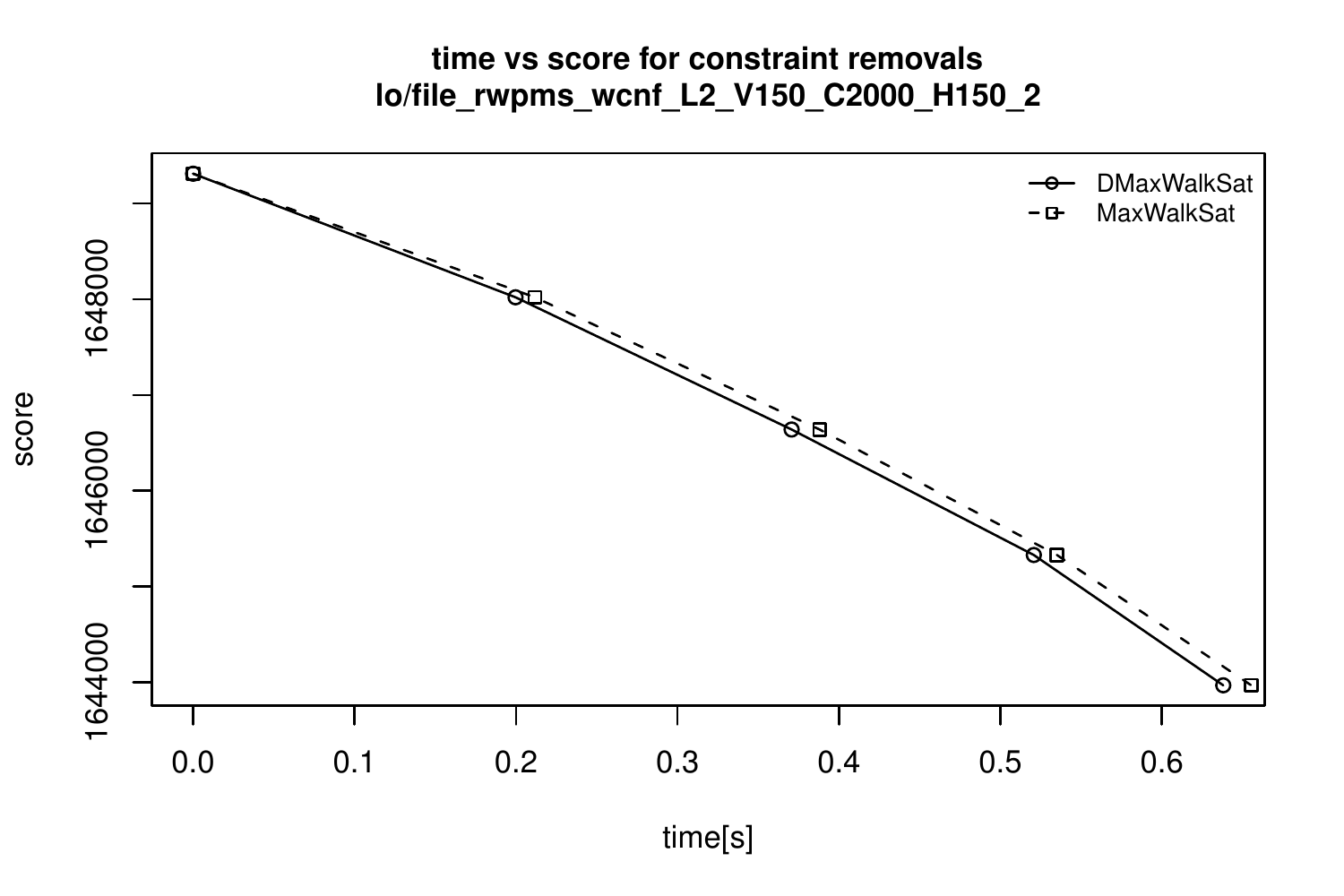}
        }

    \caption*{lo/file\_rwpms\_wcnf\_L2\_V150\_C2000\_H150\_2}
    \label{fig_lo/file_rwpms_wcnf_L2_V150_C2000_H150_2}
\end{figure}

\begin{figure}[H]
    \setcounter{subfigure}{0}
    \centering
        \subfloat[Constraint addition]
        {
            \includegraphics[width=2.7in]{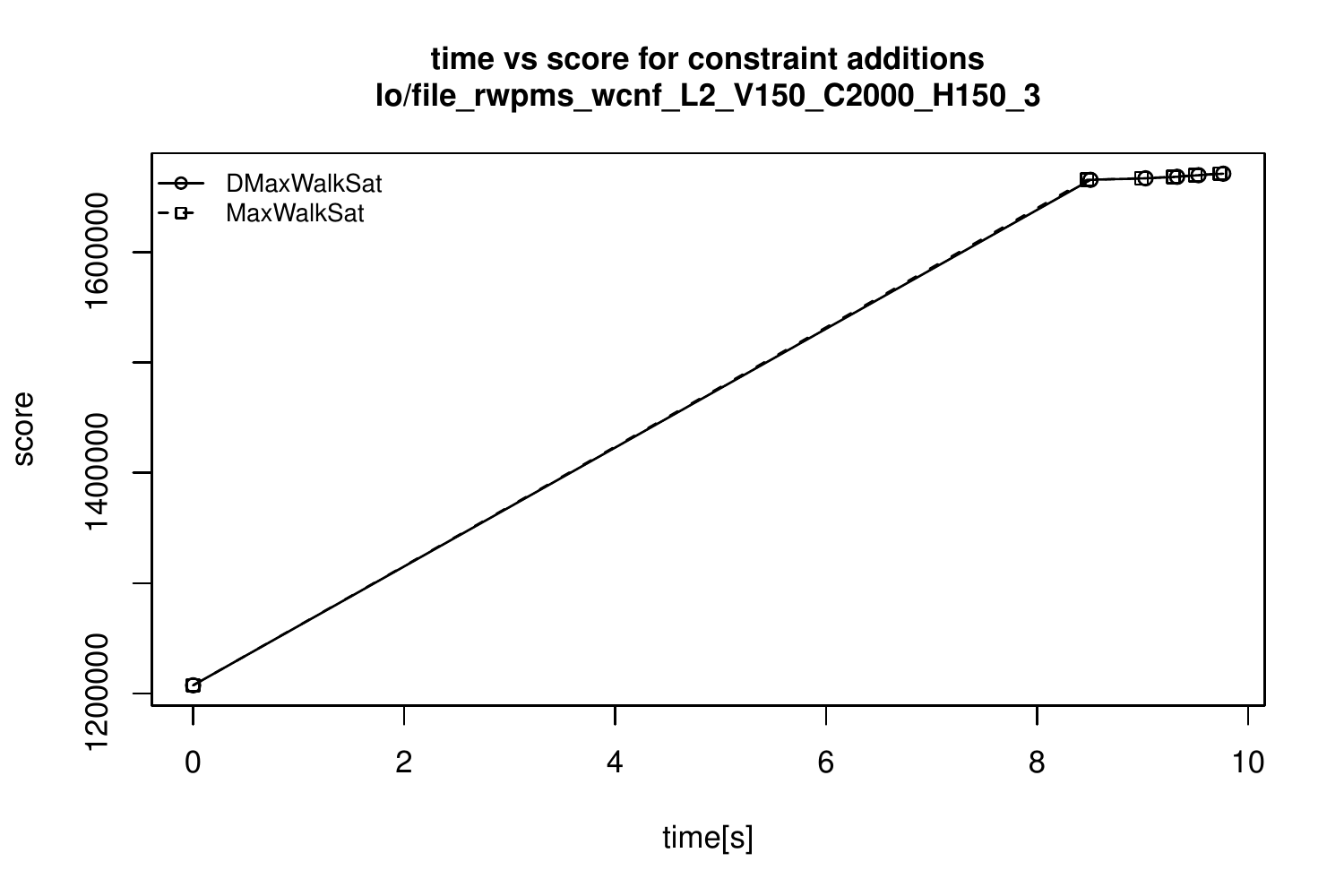}
        }
        \qquad
        \subfloat[Constraint removal]
        {
            \includegraphics[width=2.7in]{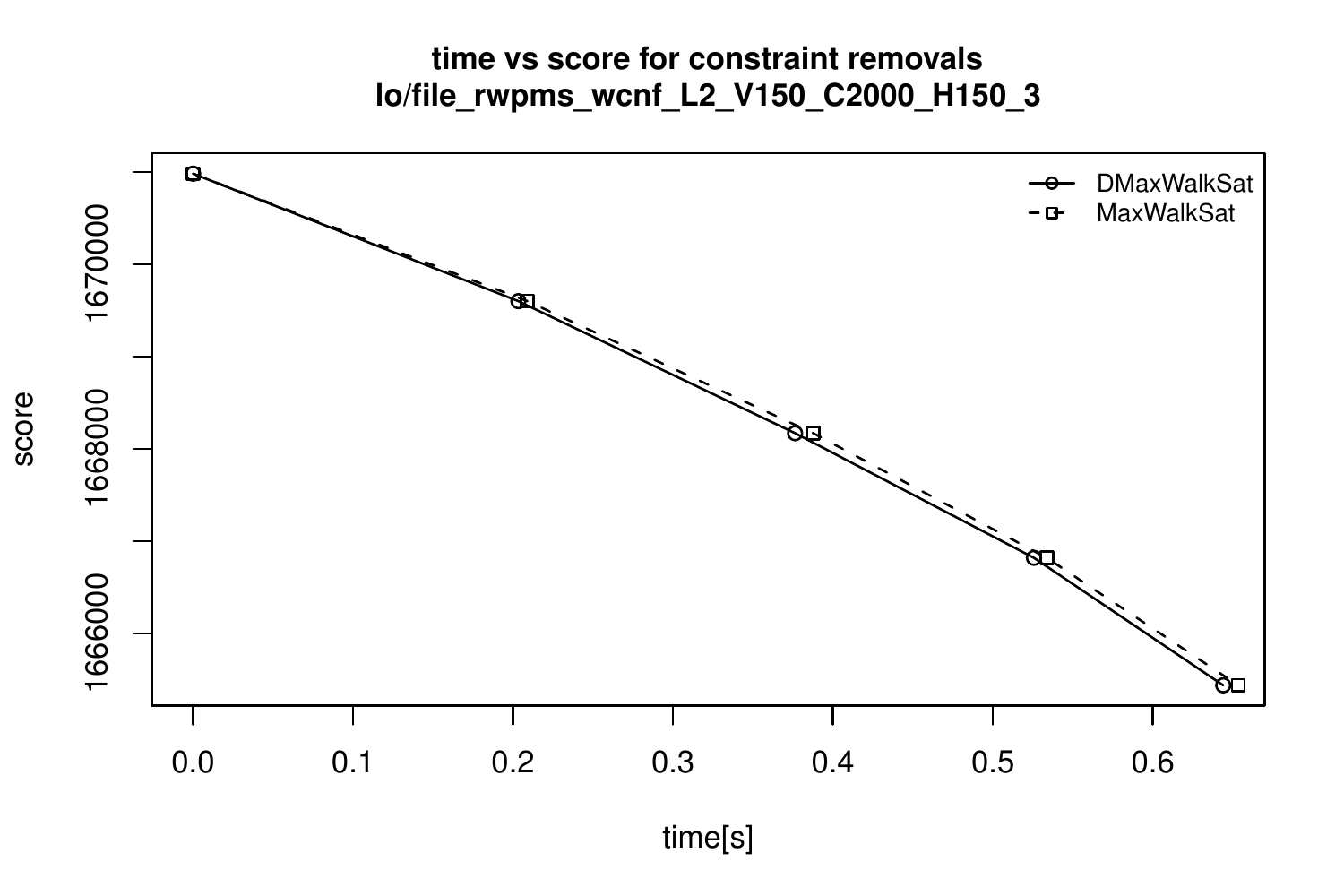}
        }

    \caption*{lo/file\_rwpms\_wcnf\_L2\_V150\_C2000\_H150\_3}
    \label{fig_lo/file_rwpms_wcnf_L2_V150_C2000_H150_3}
\end{figure}

\begin{figure}[H]
    \setcounter{subfigure}{0}
    \centering
        \subfloat[Constraint addition]
        {
            \includegraphics[width=2.7in]{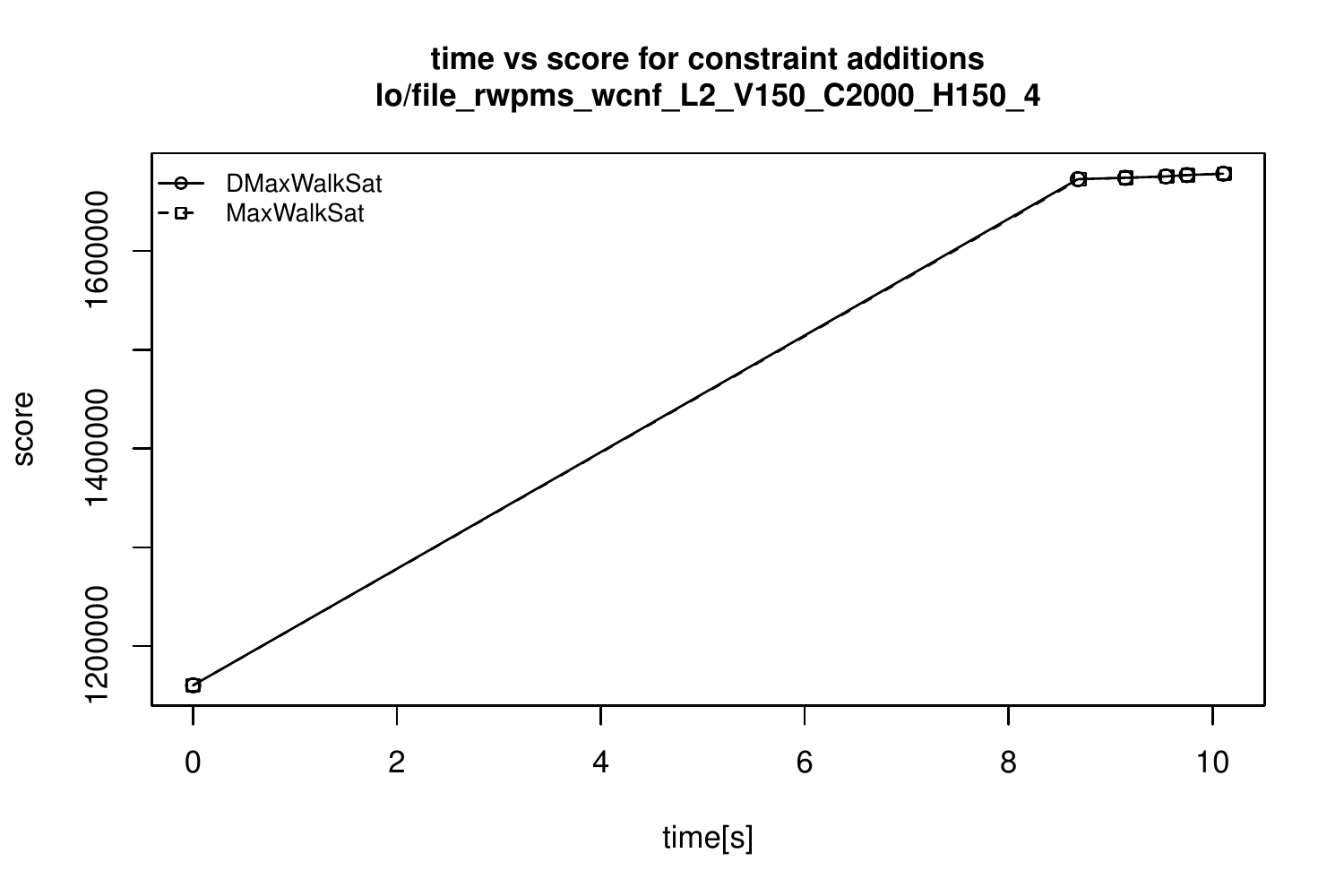}
        }
        \qquad
        \subfloat[Constraint removal]
        {
            \includegraphics[width=2.7in]{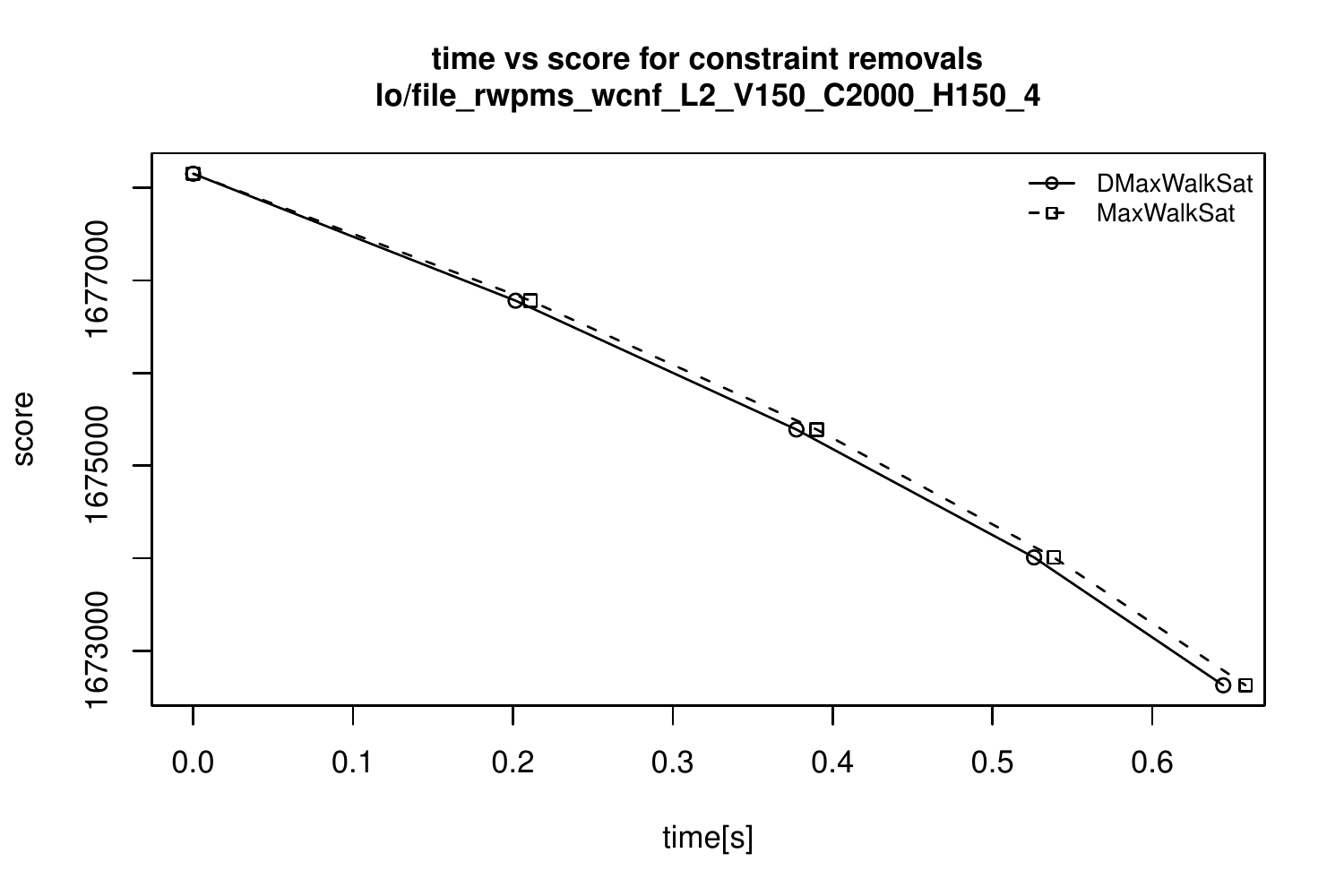}
        }

    \caption*{lo/file\_rwpms\_wcnf\_L2\_V150\_C2000\_H150\_4}
    \label{fig_lo/file_rwpms_wcnf_L2_V150_C2000_H150_4}
\end{figure}

\begin{figure}[H]
    \setcounter{subfigure}{0}
    \centering
        \subfloat[Constraint addition]
        {
            \includegraphics[width=2.7in]{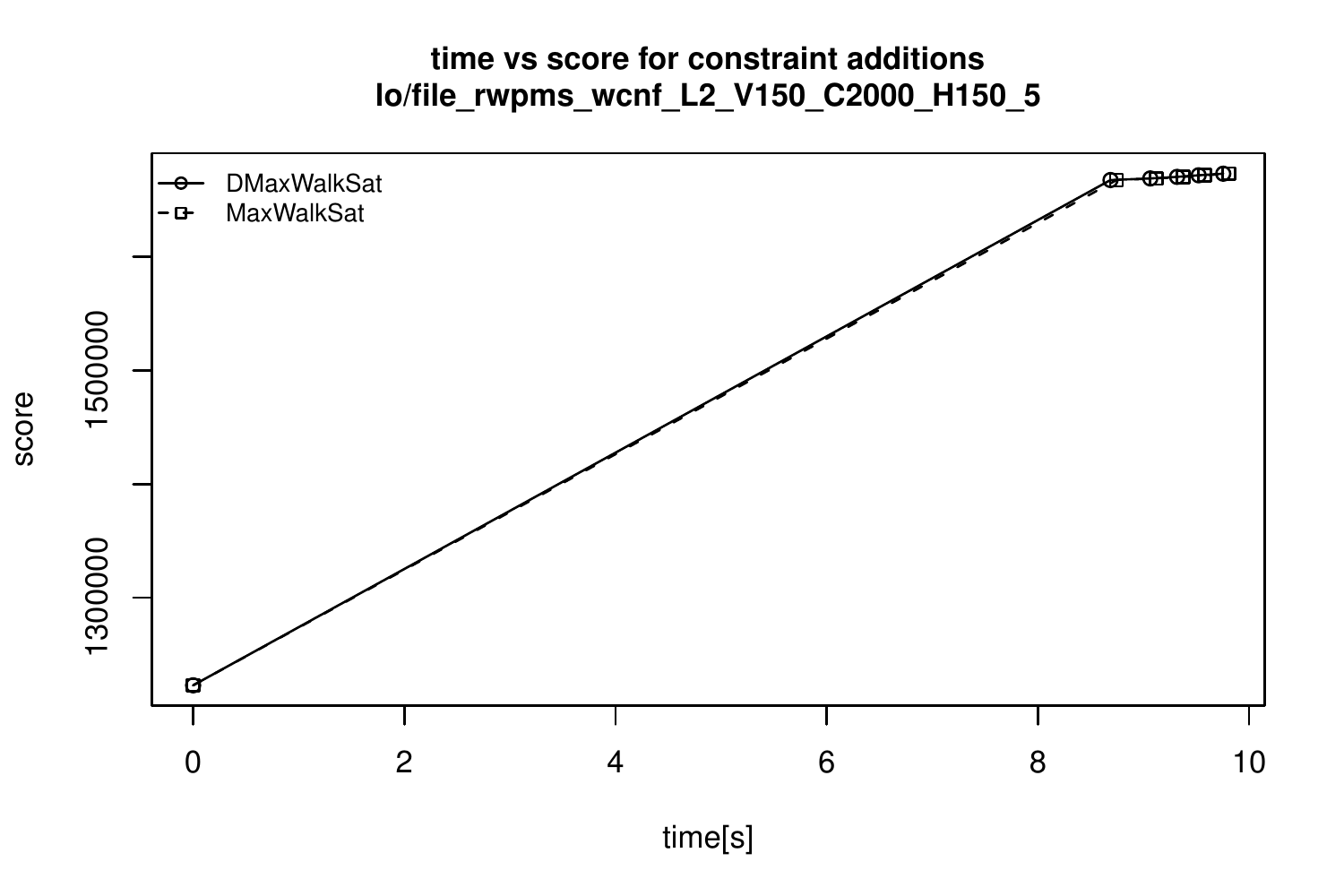}
        }
        \qquad
        \subfloat[Constraint removal]
        {
            \includegraphics[width=2.7in]{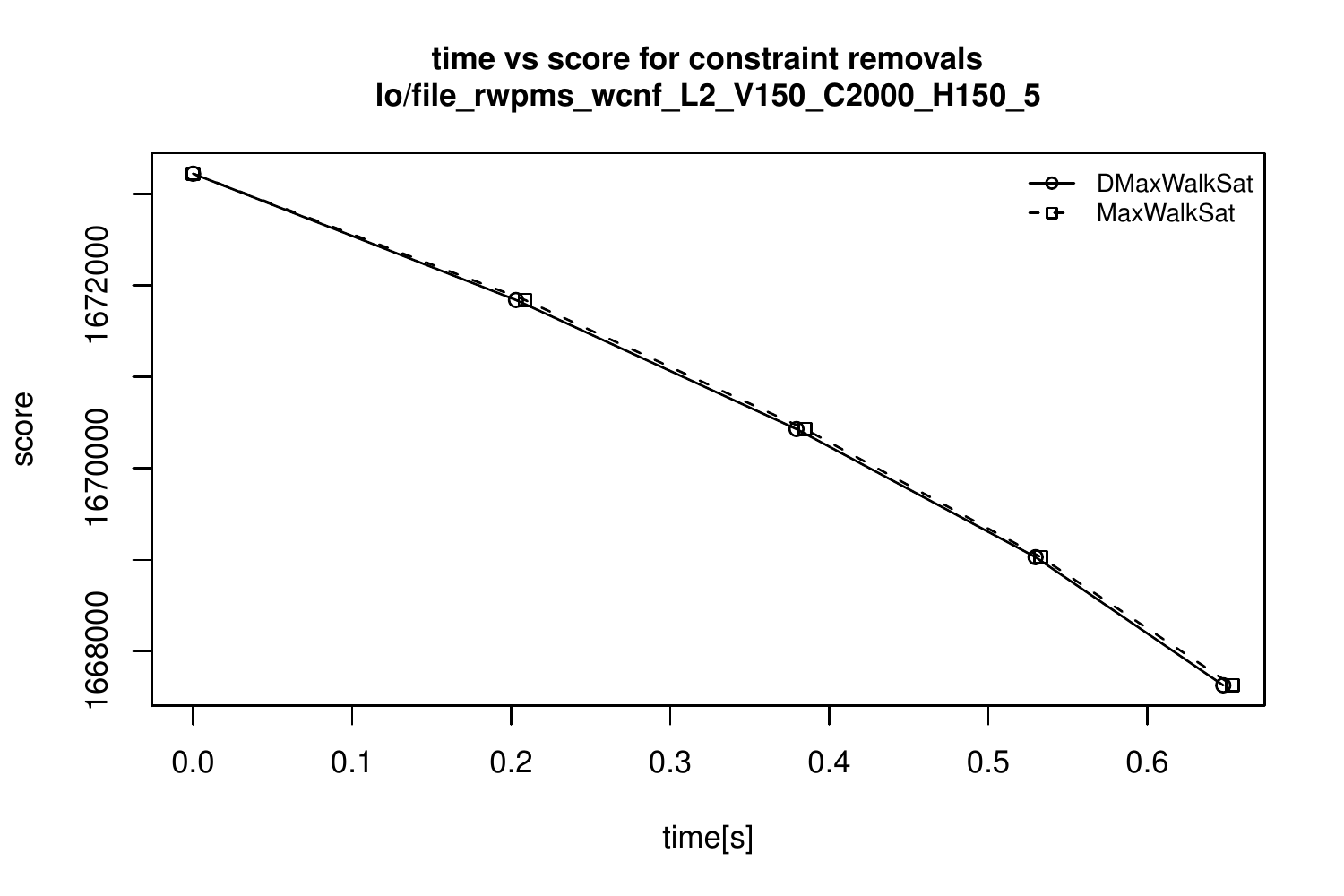}
        }

    \caption*{lo/file\_rwpms\_wcnf\_L2\_V150\_C2000\_H150\_5}
    \label{fig_lo/file_rwpms_wcnf_L2_V150_C2000_H150_5}
\end{figure}

\begin{figure}[H]
    \setcounter{subfigure}{0}
    \centering
        \subfloat[Constraint addition]
        {
            \includegraphics[width=2.7in]{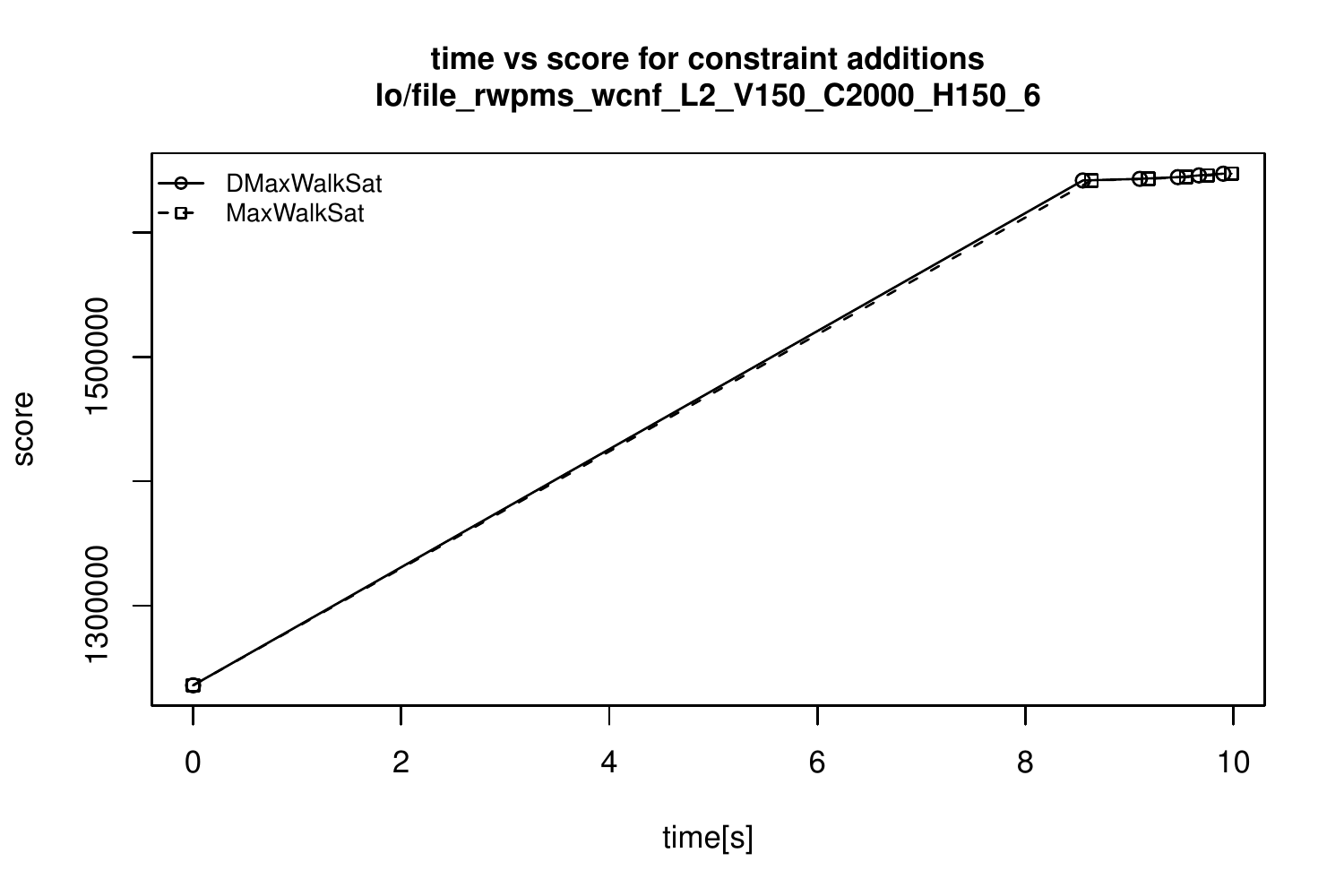}
        }
        \qquad
        \subfloat[Constraint removal]
        {
            \includegraphics[width=2.7in]{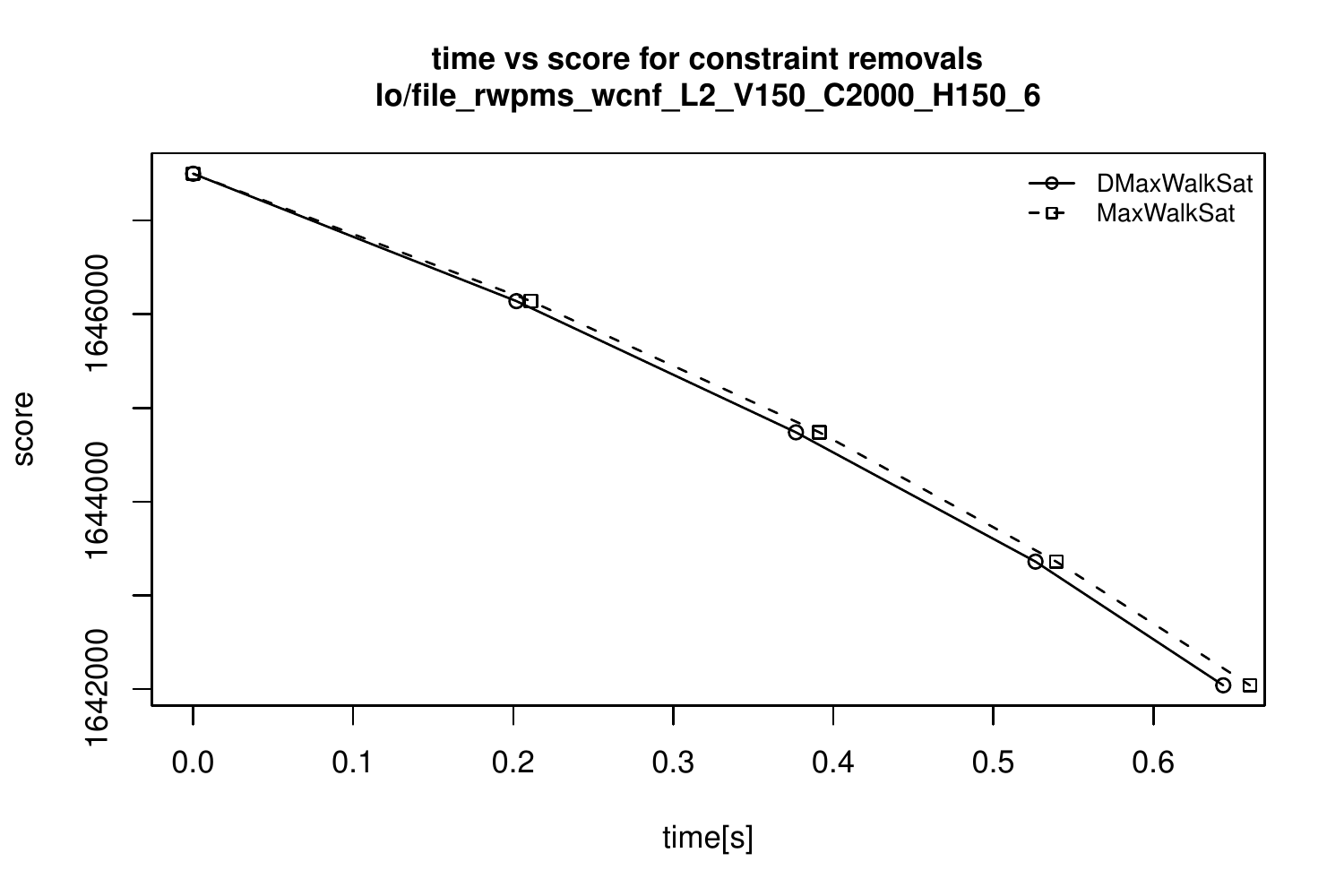}
        }

    \caption*{lo/file\_rwpms\_wcnf\_L2\_V150\_C2000\_H150\_6}
    \label{fig_lo/file_rwpms_wcnf_L2_V150_C2000_H150_6}
\end{figure}

\begin{figure}[H]
    \setcounter{subfigure}{0}
    \centering
        \subfloat[Constraint addition]
        {
            \includegraphics[width=2.7in]{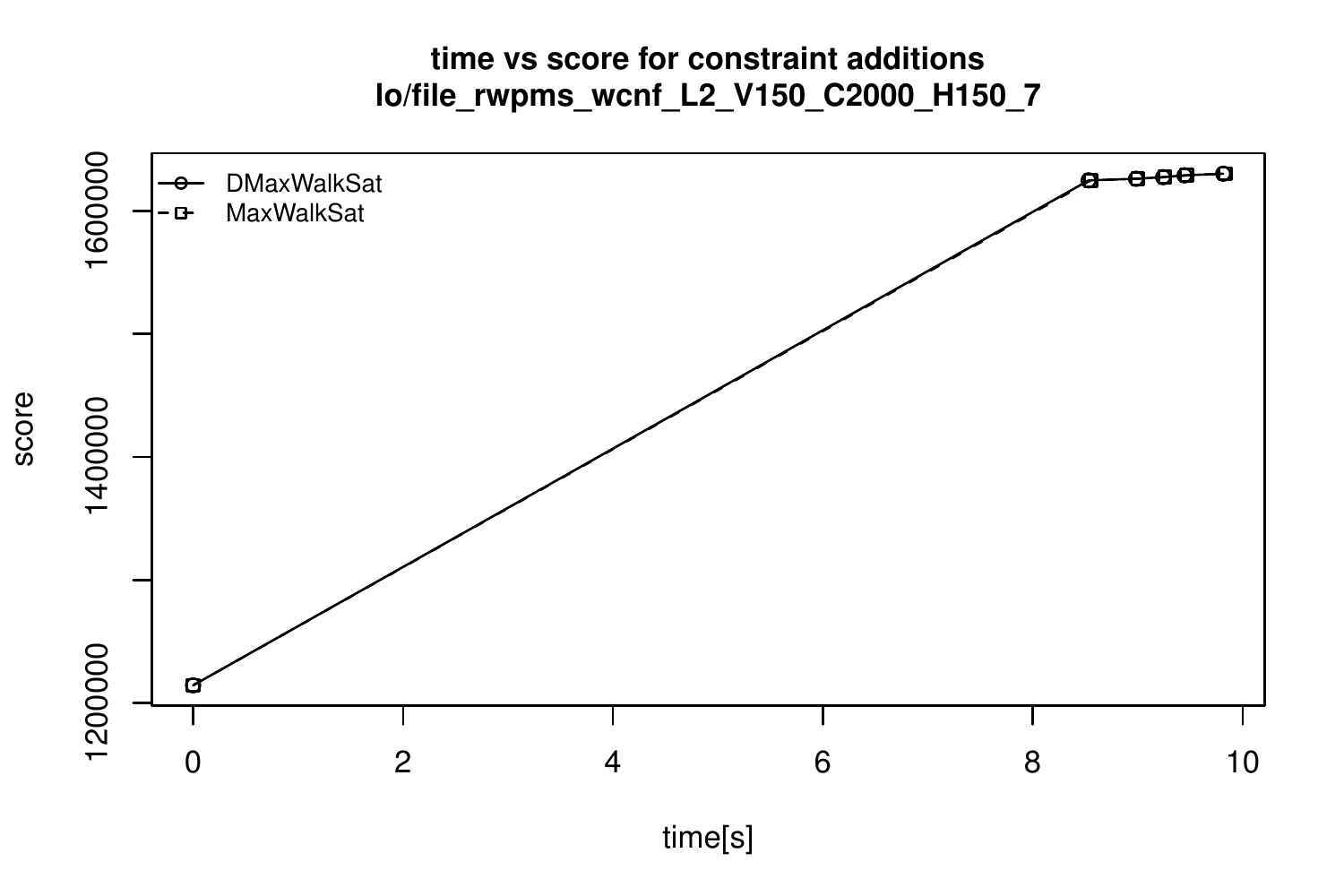}
        }
        \qquad
        \subfloat[Constraint removal]
        {
            \includegraphics[width=2.7in]{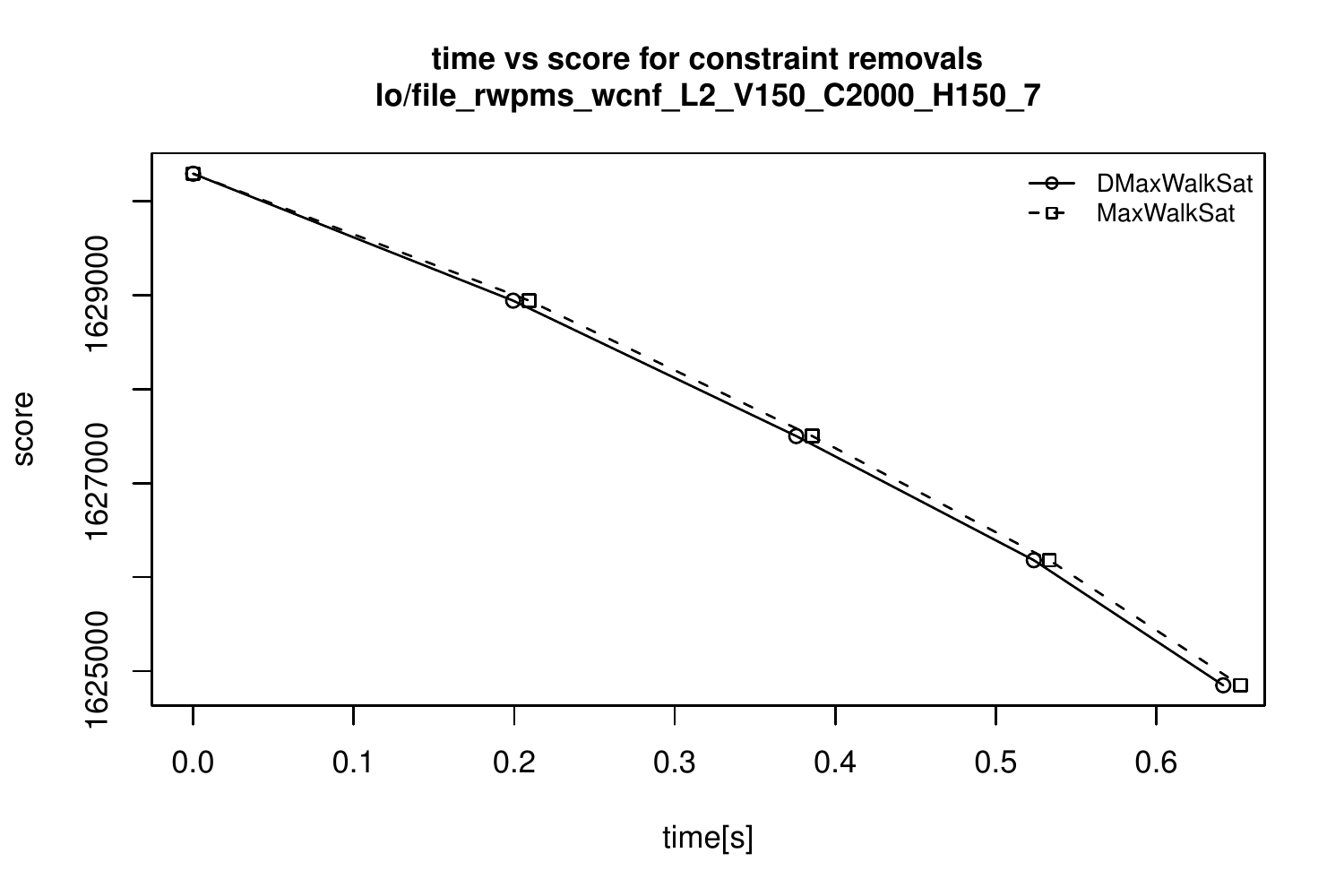}
        }

    \caption*{lo/file\_rwpms\_wcnf\_L2\_V150\_C2000\_H150\_7}
    \label{fig_lo/file_rwpms_wcnf_L2_V150_C2000_H150_7}
\end{figure}

\begin{figure}[H]
    \setcounter{subfigure}{0}
    \centering
        \subfloat[Constraint addition]
        {
            \includegraphics[width=2.7in]{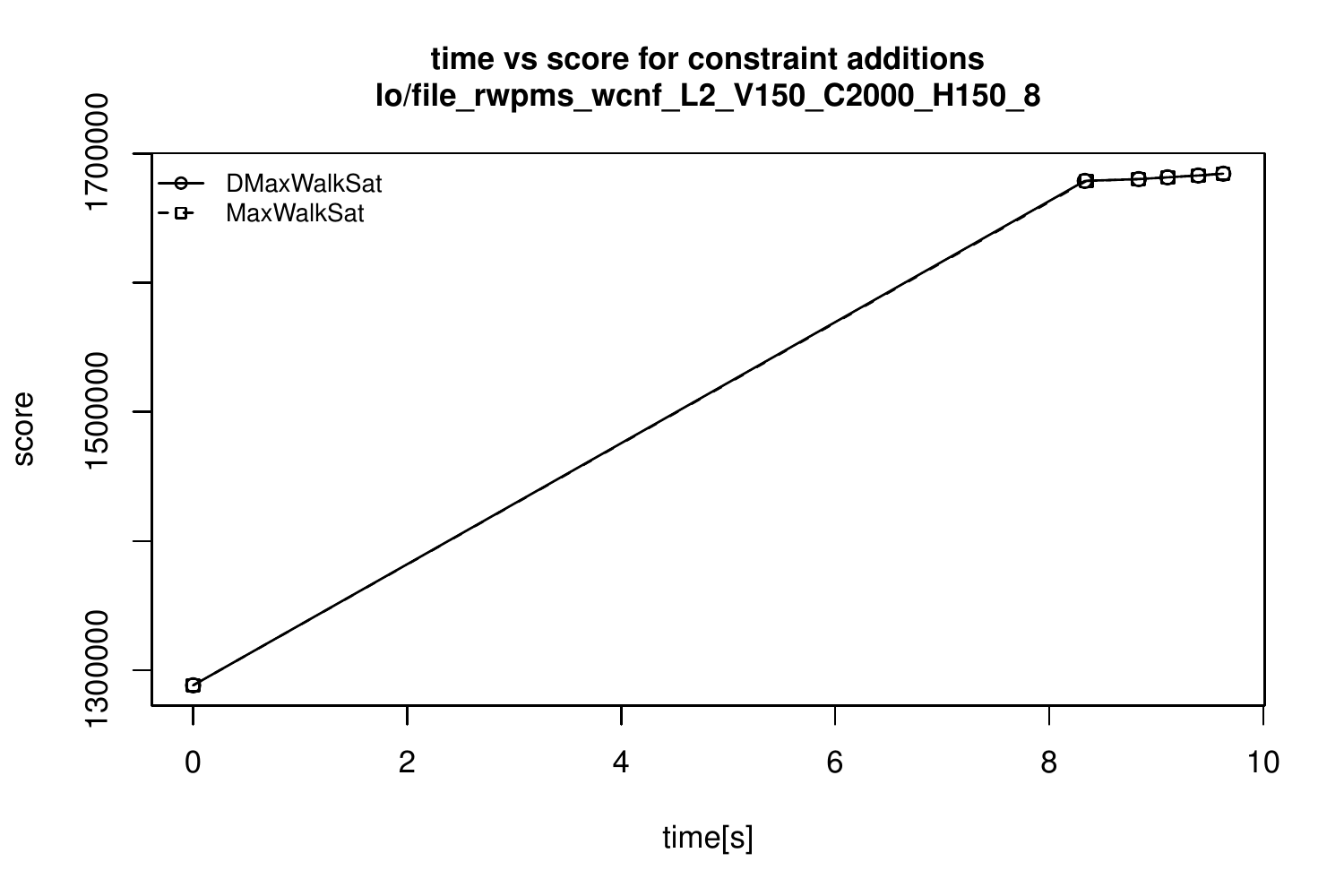}
        }
        \qquad
        \subfloat[Constraint removal]
        {
            \includegraphics[width=2.7in]{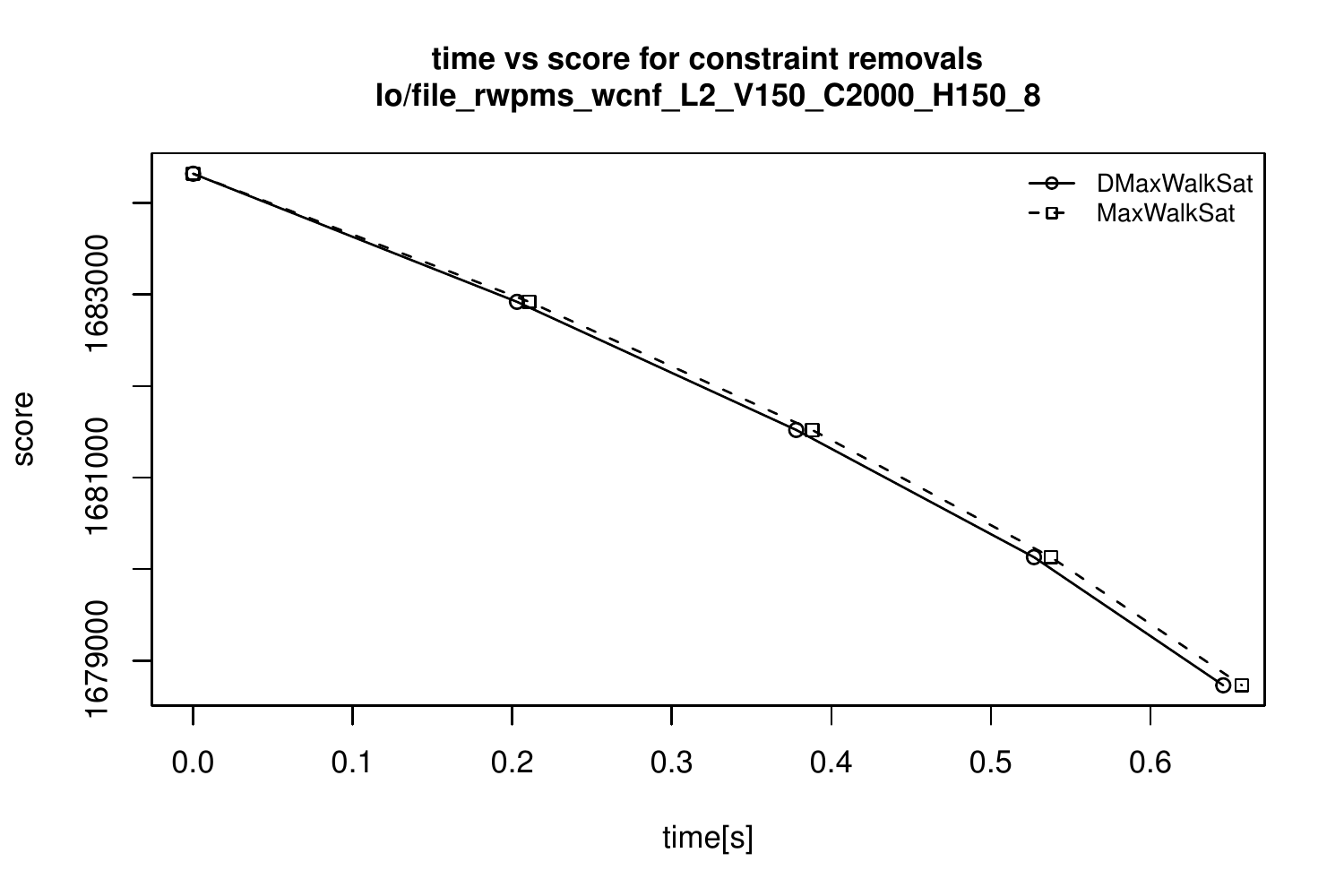}
        }

    \caption*{lo/file\_rwpms\_wcnf\_L2\_V150\_C2000\_H150\_8}
    \label{fig_lo/file_rwpms_wcnf_L2_V150_C2000_H150_8}
\end{figure}

\begin{figure}[H]
    \setcounter{subfigure}{0}
    \centering
        \subfloat[Constraint addition]
        {
            \includegraphics[width=2.7in]{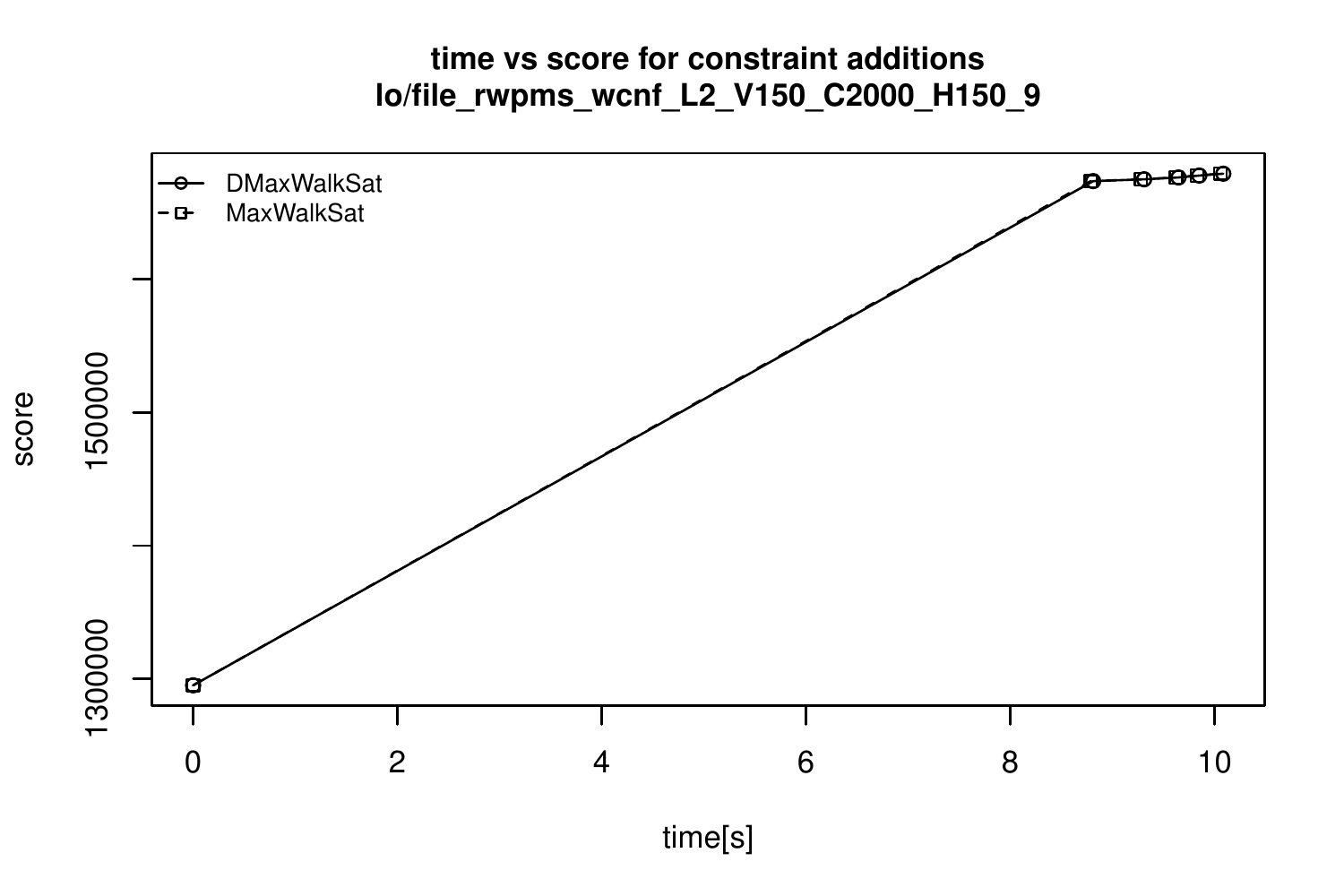}
        }
        \qquad
        \subfloat[Constraint removal]
        {
            \includegraphics[width=2.7in]{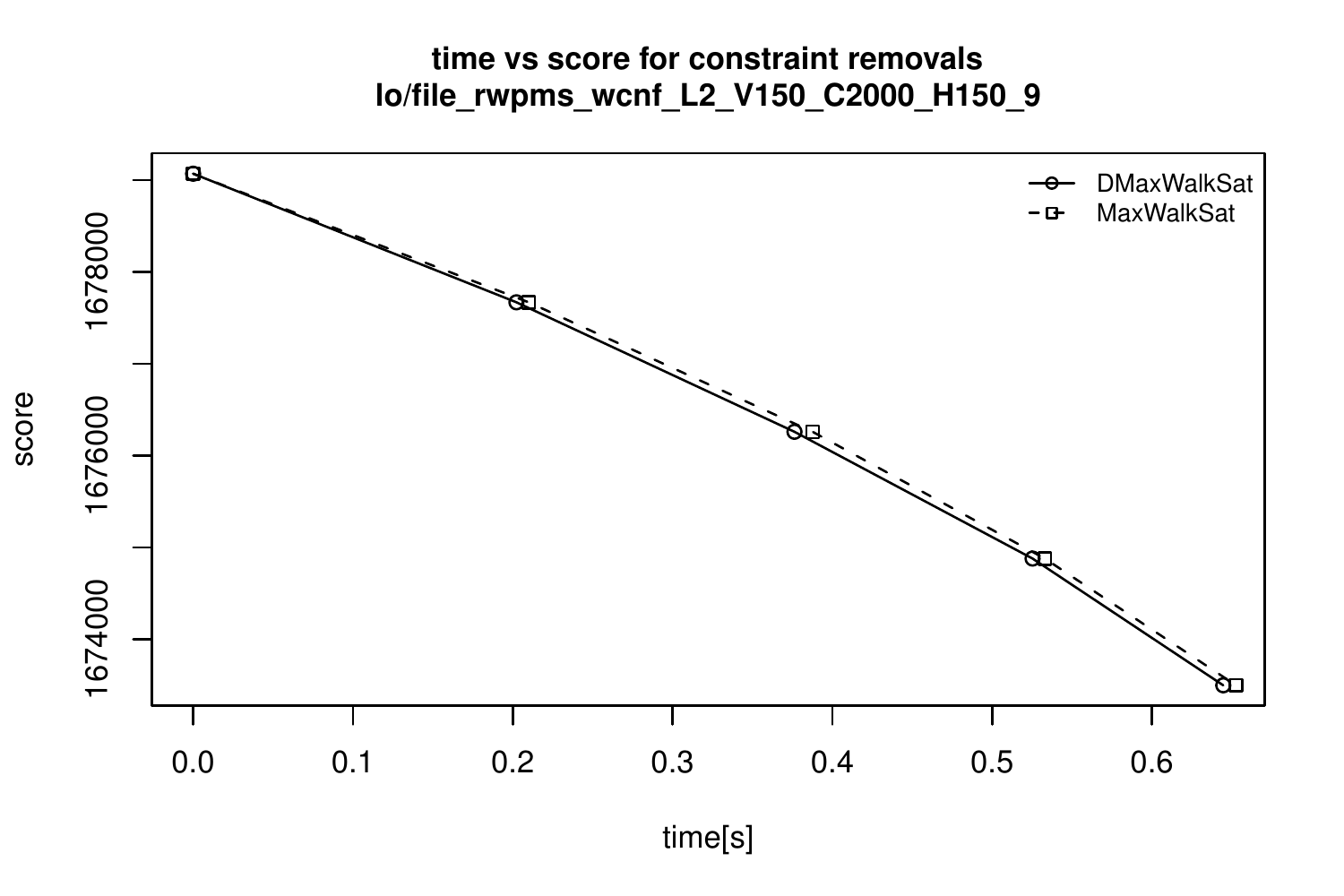}
        }

    \caption*{lo/file\_rwpms\_wcnf\_L2\_V150\_C2000\_H150\_9}
    \label{fig_lo/file_rwpms_wcnf_L2_V150_C2000_H150_9}
\end{figure}

\begin{figure}[H]
    \setcounter{subfigure}{0}
    \centering
        \subfloat[Constraint addition]
        {
            \includegraphics[width=2.7in]{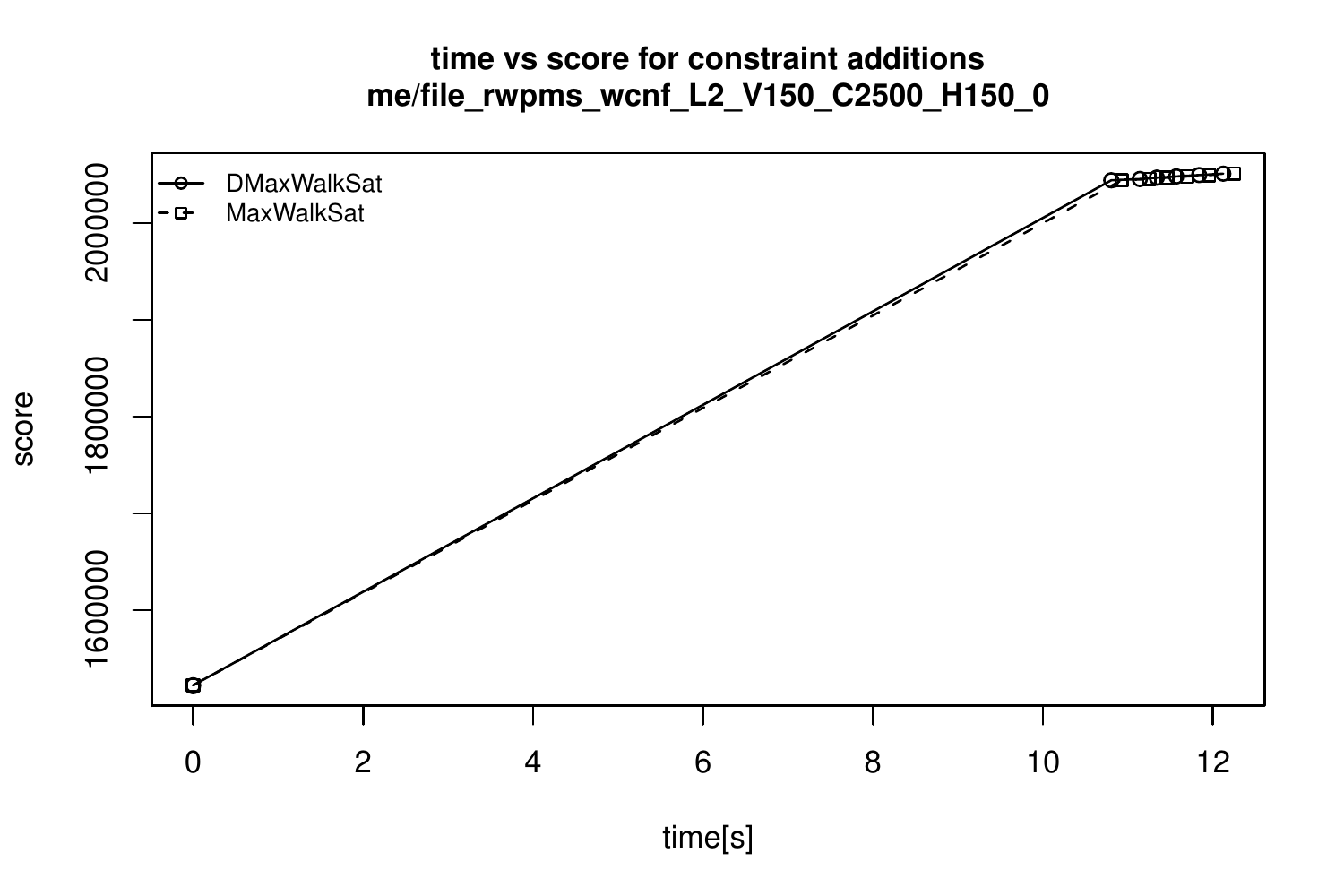}
        }
        \qquad
        \subfloat[Constraint removal]
        {
            \includegraphics[width=2.7in]{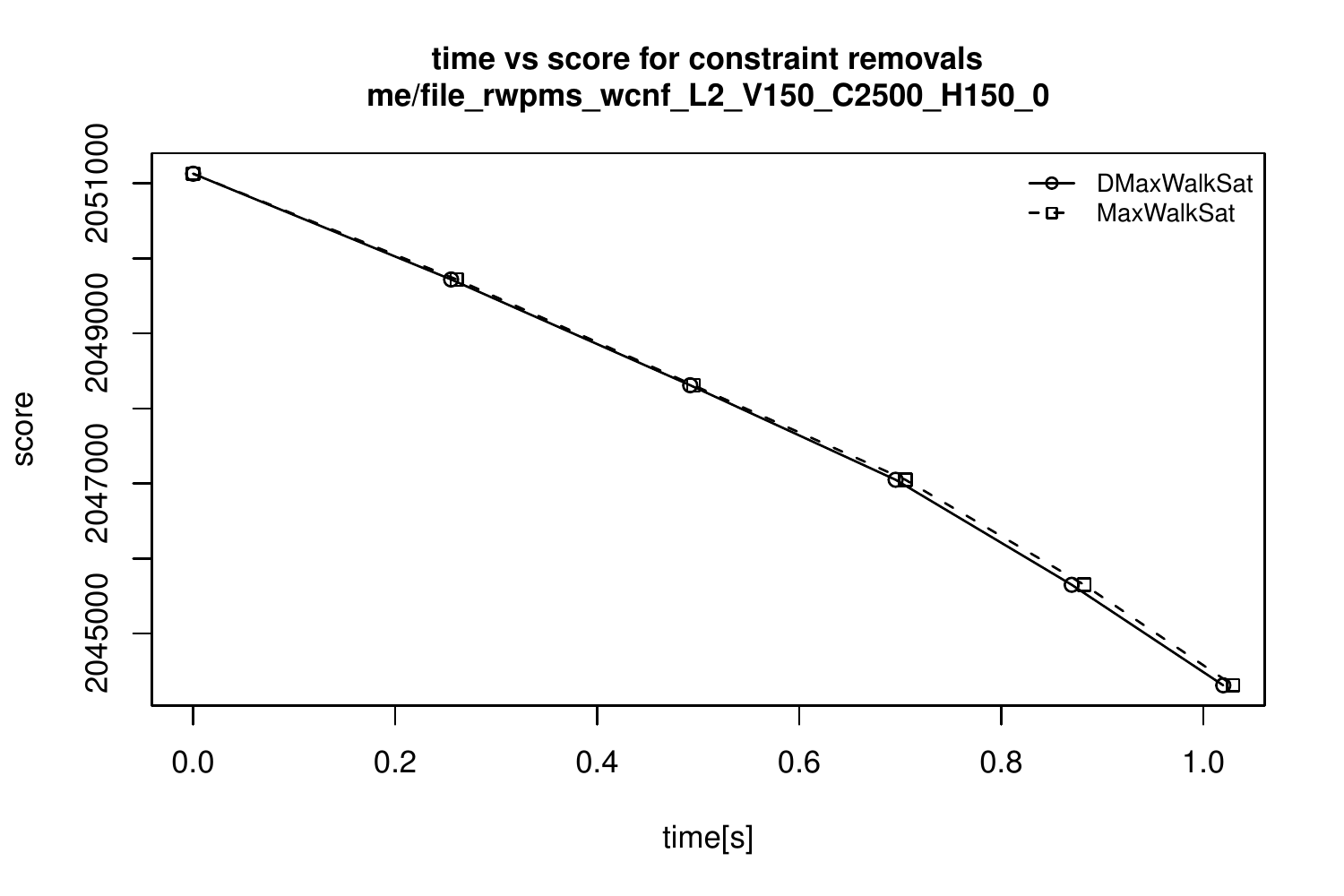}
        }

    \caption*{me/file\_rwpms\_wcnf\_L2\_V150\_C2500\_H150\_0}
    \label{fig_me/file_rwpms_wcnf_L2_V150_C2500_H150_0}
\end{figure}

\begin{figure}[H]
    \setcounter{subfigure}{0}
    \centering
        \subfloat[Constraint addition]
        {
            \includegraphics[width=2.7in]{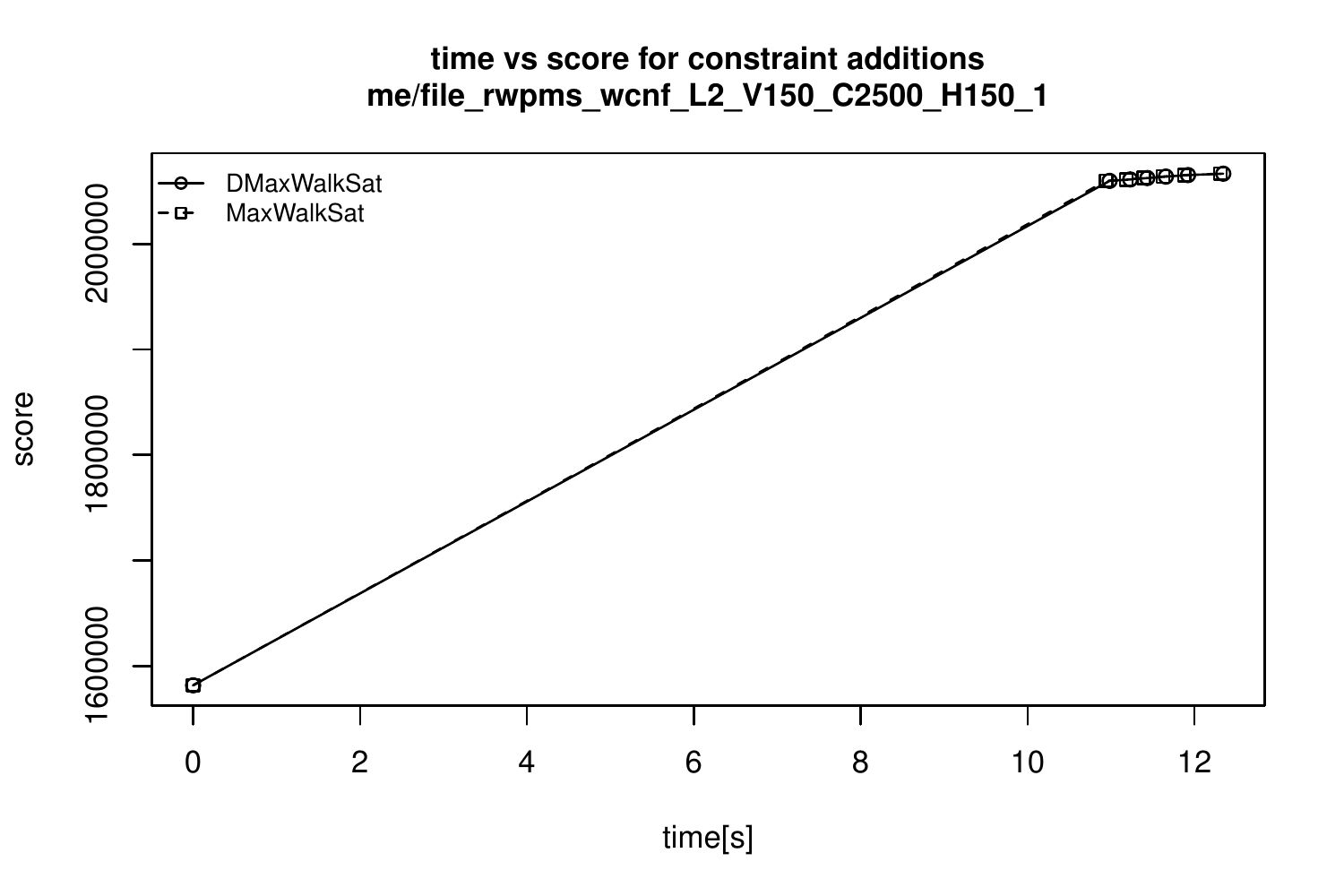}
        }
        \qquad
        \subfloat[Constraint removal]
        {
            \includegraphics[width=2.7in]{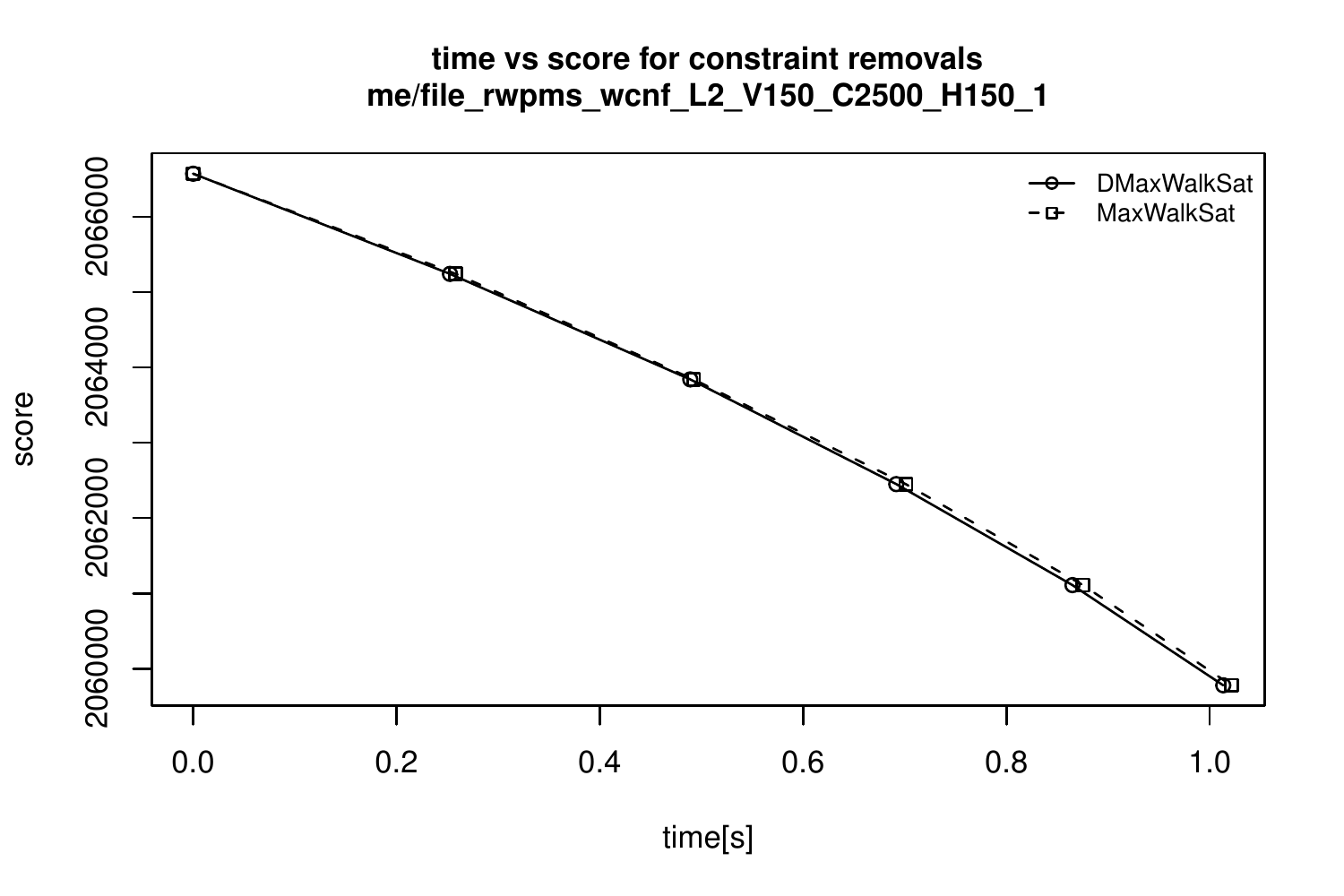}
        }

    \caption*{me/file\_rwpms\_wcnf\_L2\_V150\_C2500\_H150\_1}
    \label{fig_me/file_rwpms_wcnf_L2_V150_C2500_H150_1}
\end{figure}

\begin{figure}[H]
    \setcounter{subfigure}{0}
    \centering
        \subfloat[Constraint addition]
        {
            \includegraphics[width=2.7in]{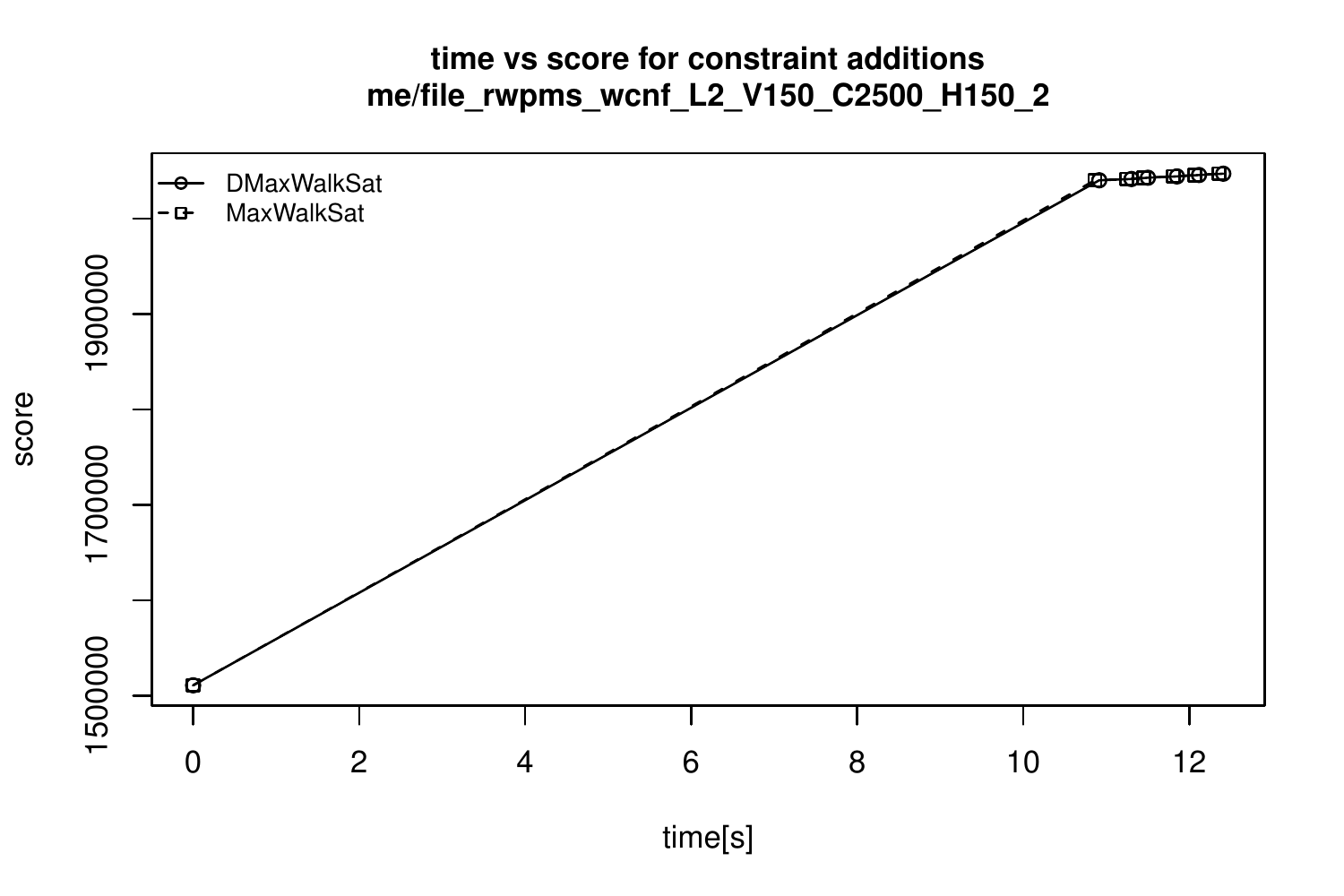}
        }
        \qquad
        \subfloat[Constraint removal]
        {
            \includegraphics[width=2.7in]{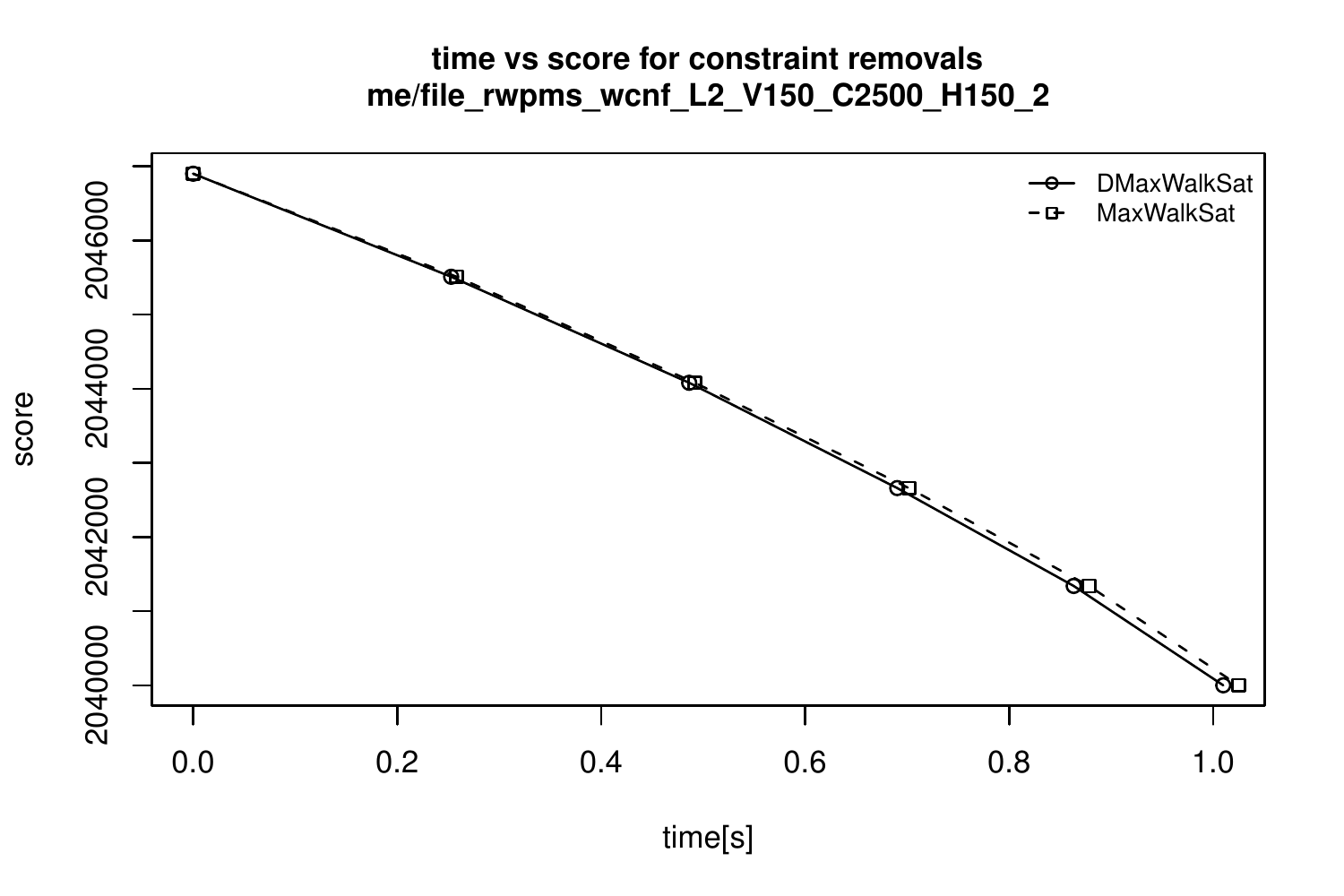}
        }

    \caption*{me/file\_rwpms\_wcnf\_L2\_V150\_C2500\_H150\_2}
    \label{fig_me/file_rwpms_wcnf_L2_V150_C2500_H150_2}
\end{figure}

\begin{figure}[H]
    \setcounter{subfigure}{0}
    \centering
        \subfloat[Constraint addition]
        {
            \includegraphics[width=2.7in]{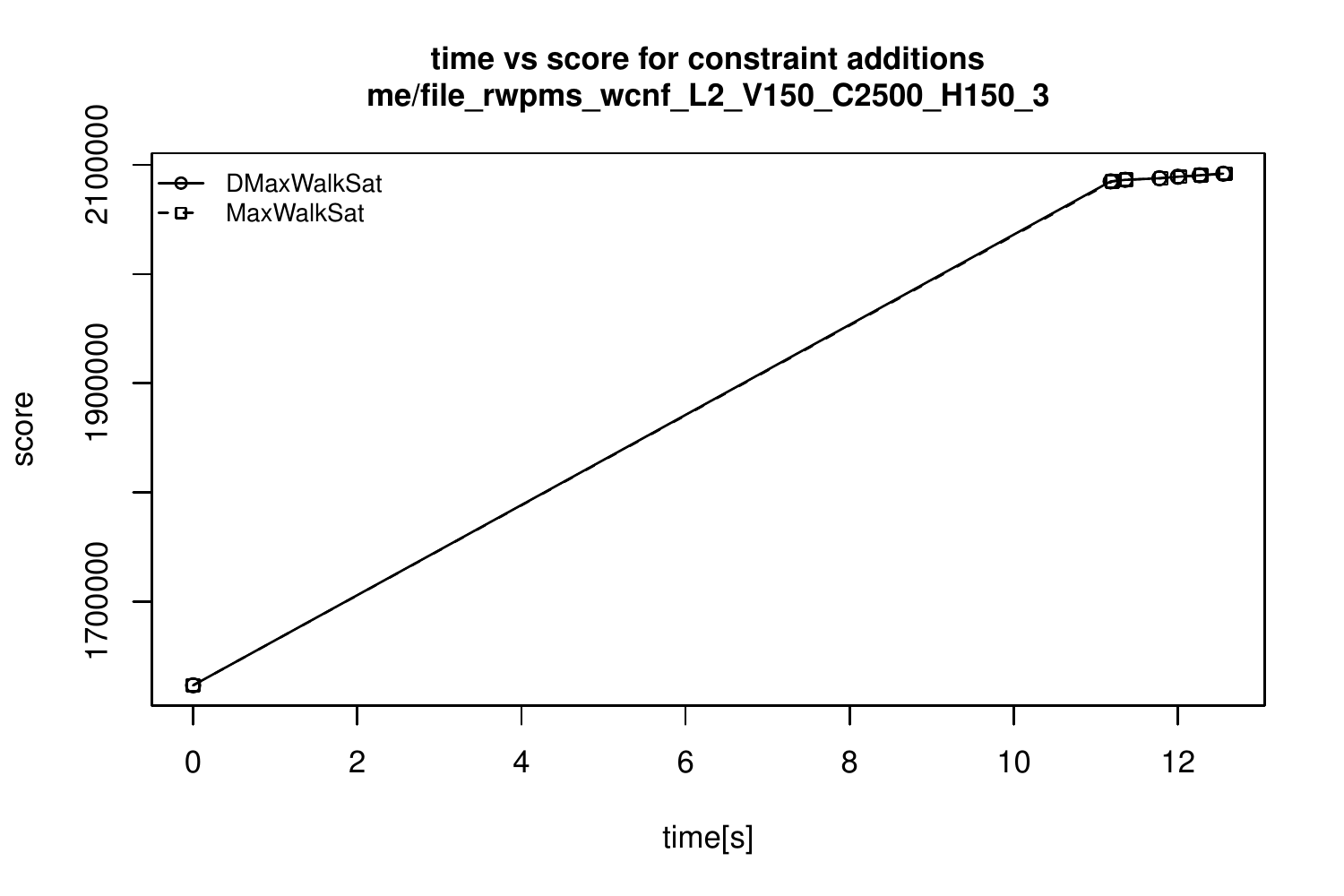}
        }
        \qquad
        \subfloat[Constraint removal]
        {
            \includegraphics[width=2.7in]{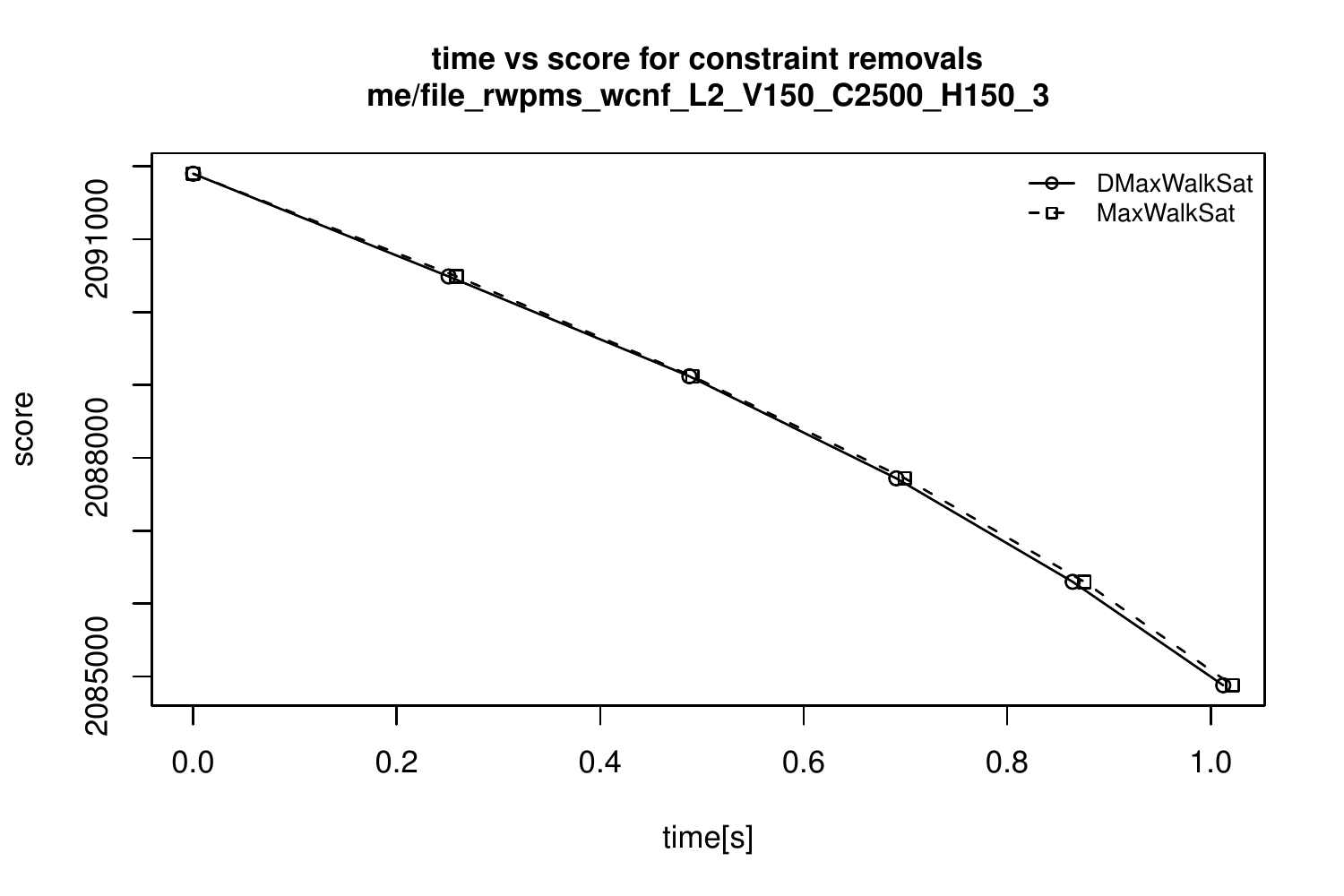}
        }

    \caption*{me/file\_rwpms\_wcnf\_L2\_V150\_C2500\_H150\_3}
    \label{fig_me/file_rwpms_wcnf_L2_V150_C2500_H150_3}
\end{figure}

\begin{figure}[H]
    \setcounter{subfigure}{0}
    \centering
        \subfloat[Constraint addition]
        {
            \includegraphics[width=2.7in]{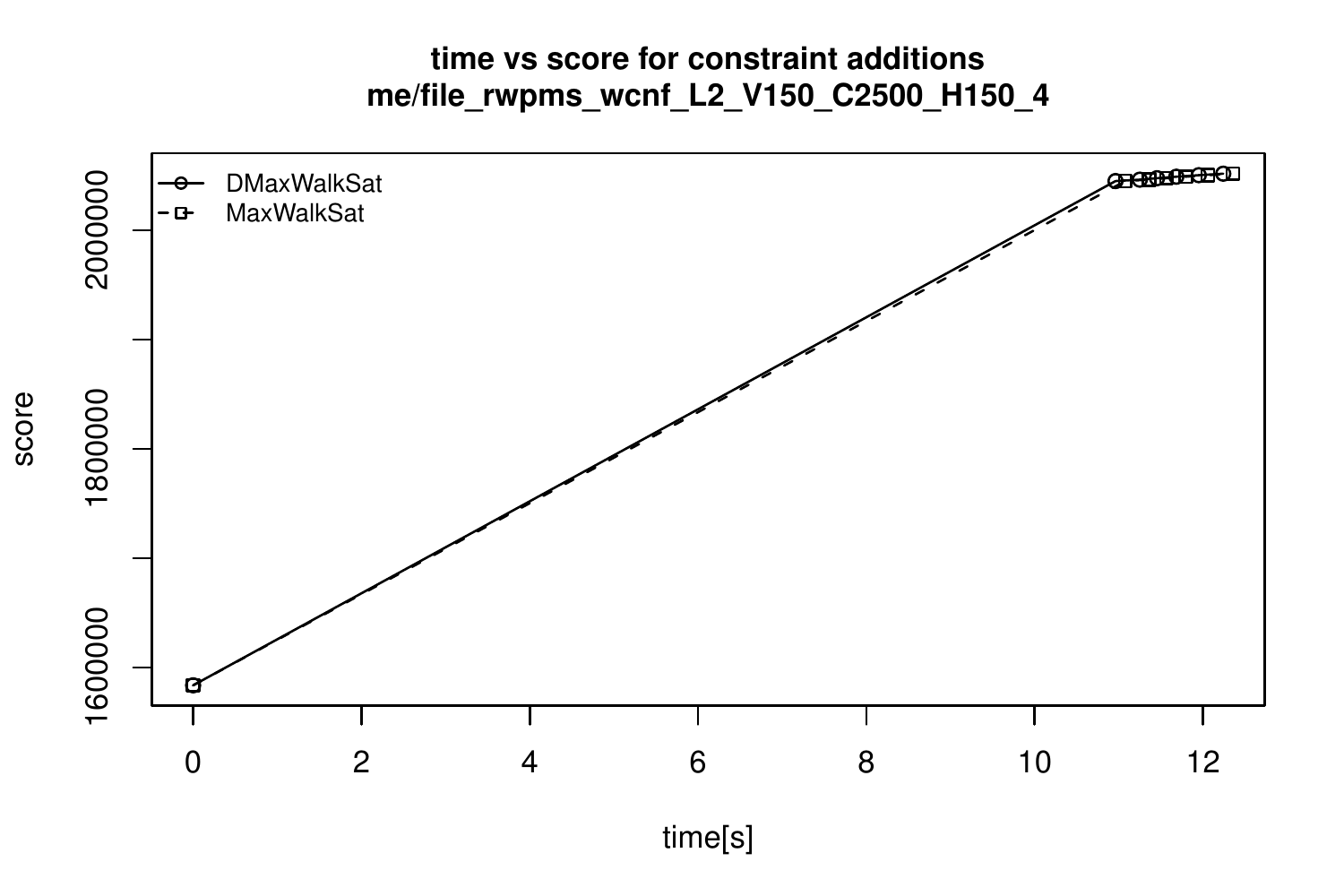}
        }
        \qquad
        \subfloat[Constraint removal]
        {
            \includegraphics[width=2.7in]{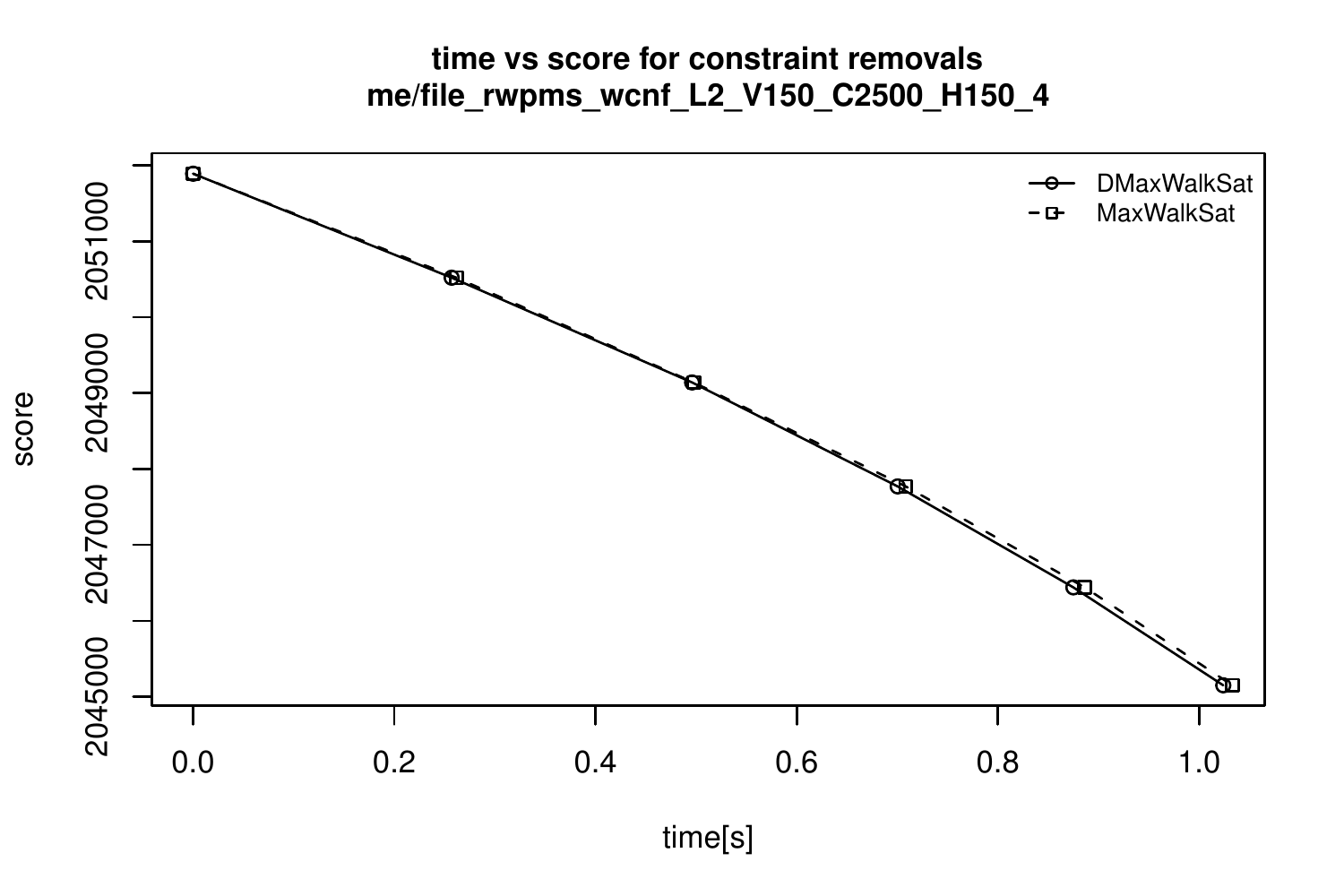}
        }

    \caption*{me/file\_rwpms\_wcnf\_L2\_V150\_C2500\_H150\_4}
    \label{fig_me/file_rwpms_wcnf_L2_V150_C2500_H150_4}
\end{figure}

\begin{figure}[H]
    \setcounter{subfigure}{0}
    \centering
        \subfloat[Constraint addition]
        {
            \includegraphics[width=2.7in]{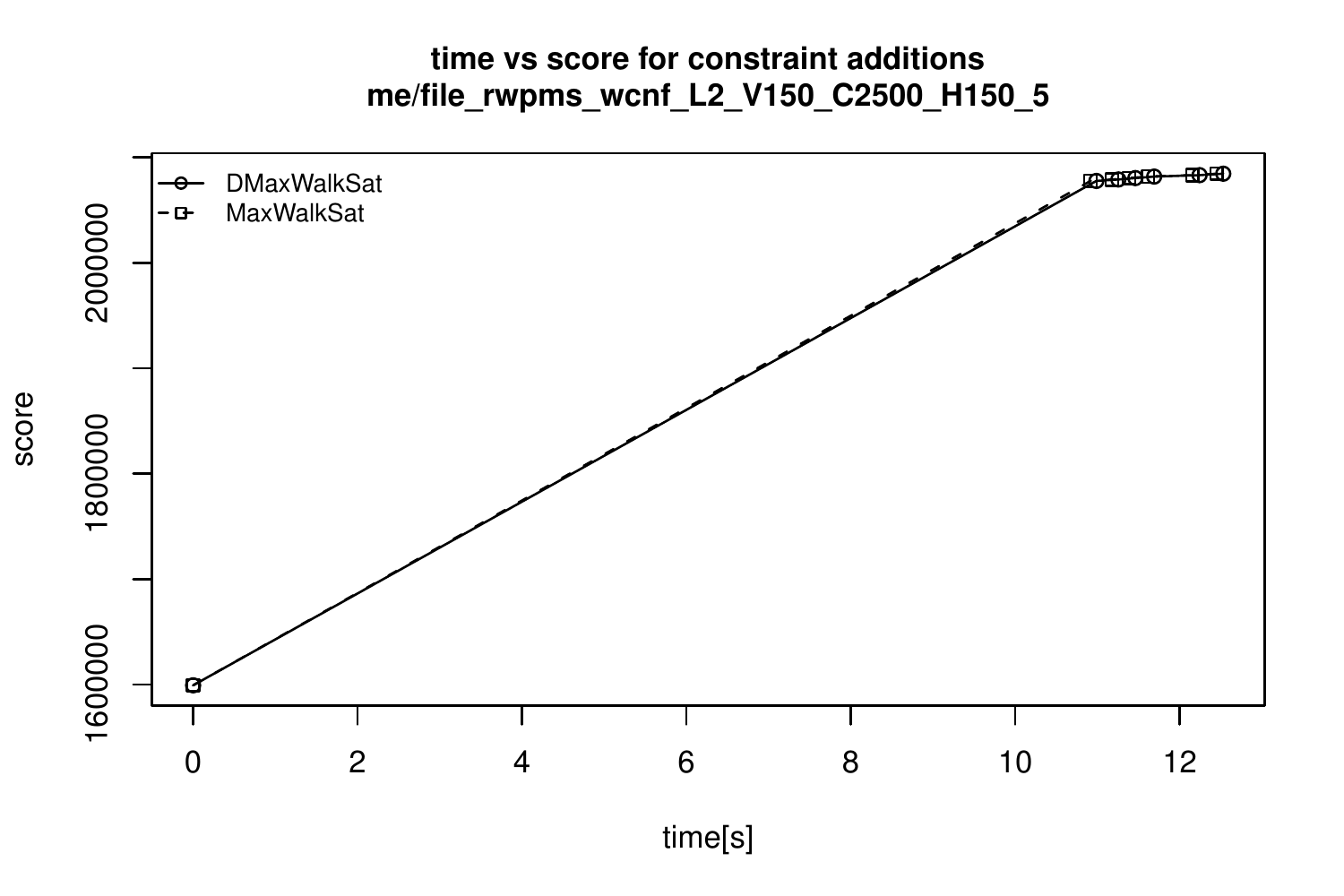}
        }
        \qquad
        \subfloat[Constraint removal]
        {
            \includegraphics[width=2.7in]{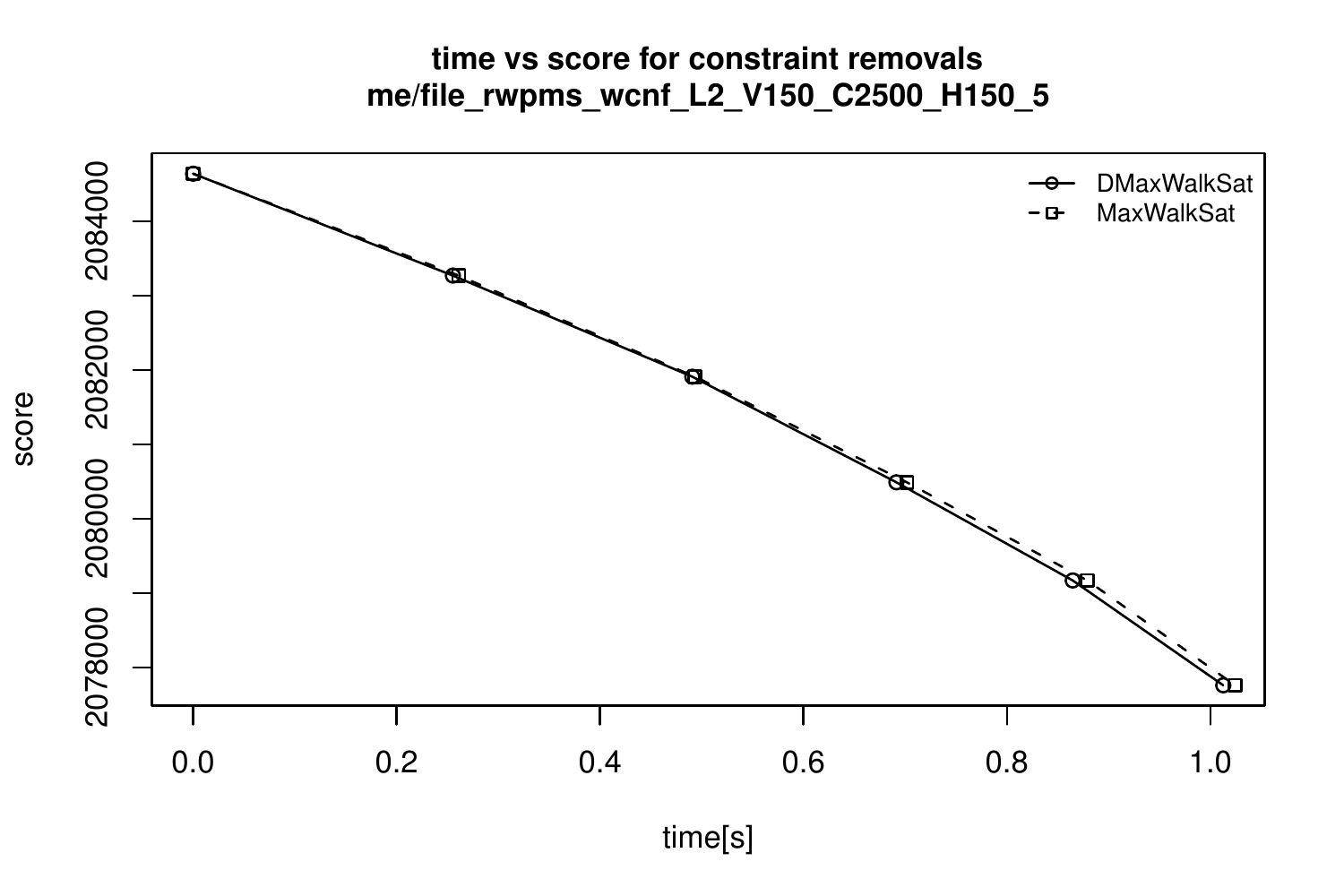}
        }

    \caption*{me/file\_rwpms\_wcnf\_L2\_V150\_C2500\_H150\_5}
    \label{fig_me/file_rwpms_wcnf_L2_V150_C2500_H150_5}
\end{figure}

\begin{figure}[H]
    \setcounter{subfigure}{0}
    \centering
        \subfloat[Constraint addition]
        {
            \includegraphics[width=2.7in]{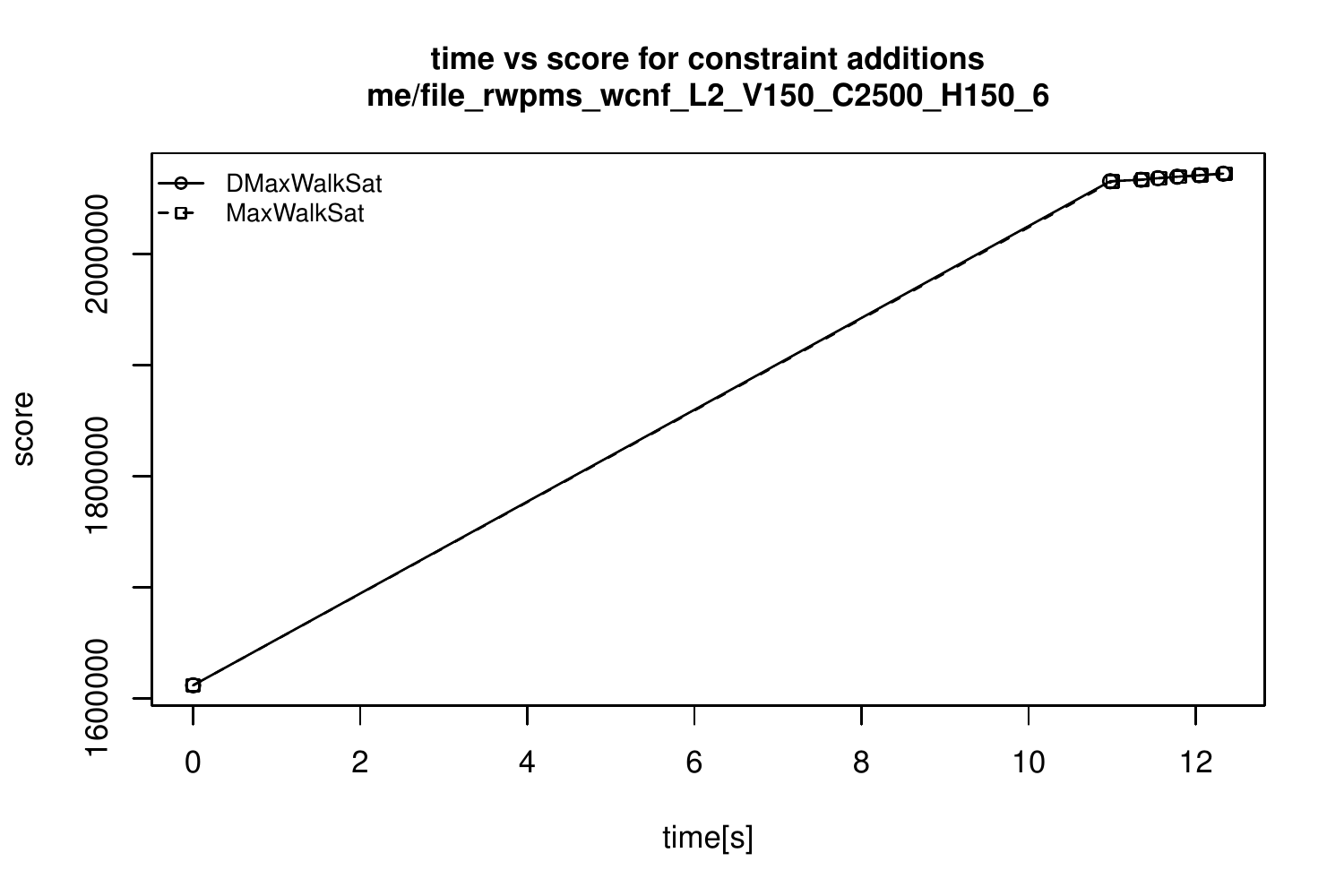}
        }
        \qquad
        \subfloat[Constraint removal]
        {
            \includegraphics[width=2.7in]{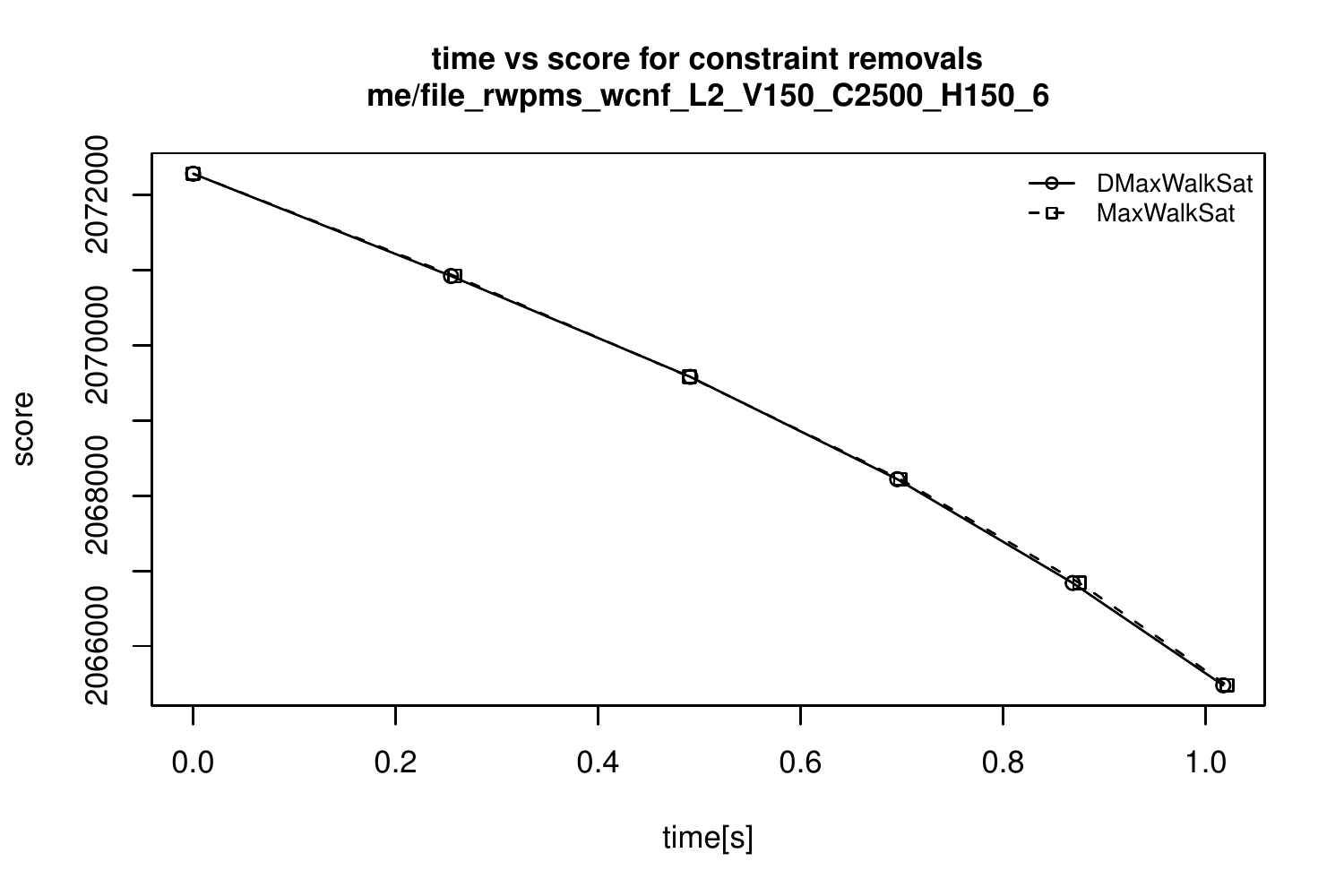}
        }

    \caption*{me/file\_rwpms\_wcnf\_L2\_V150\_C2500\_H150\_6}
    \label{fig_me/file_rwpms_wcnf_L2_V150_C2500_H150_6}
\end{figure}

\begin{figure}[H]
    \setcounter{subfigure}{0}
    \centering
        \subfloat[Constraint addition]
        {
            \includegraphics[width=2.7in]{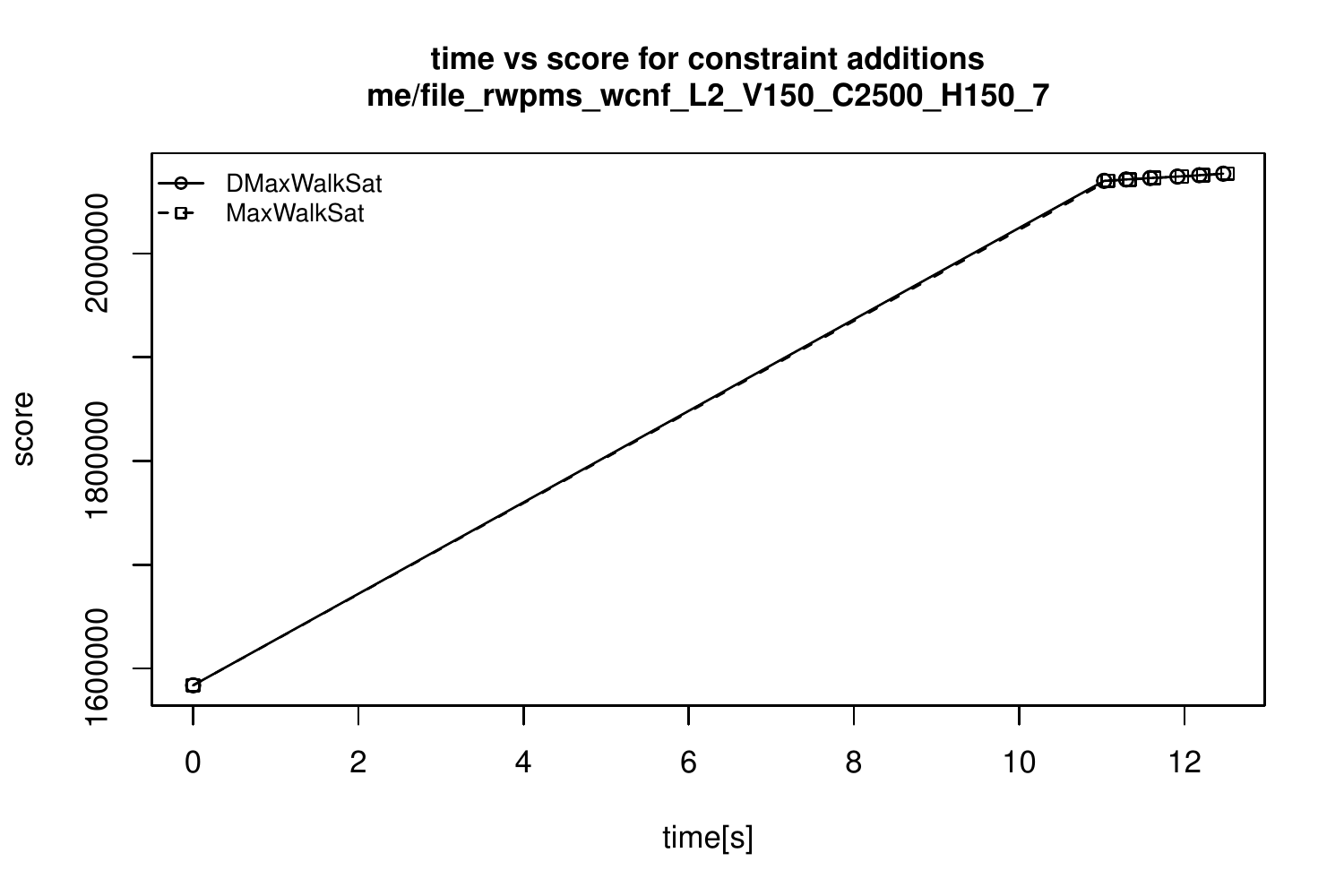}
        }
        \qquad
        \subfloat[Constraint removal]
        {
            \includegraphics[width=2.7in]{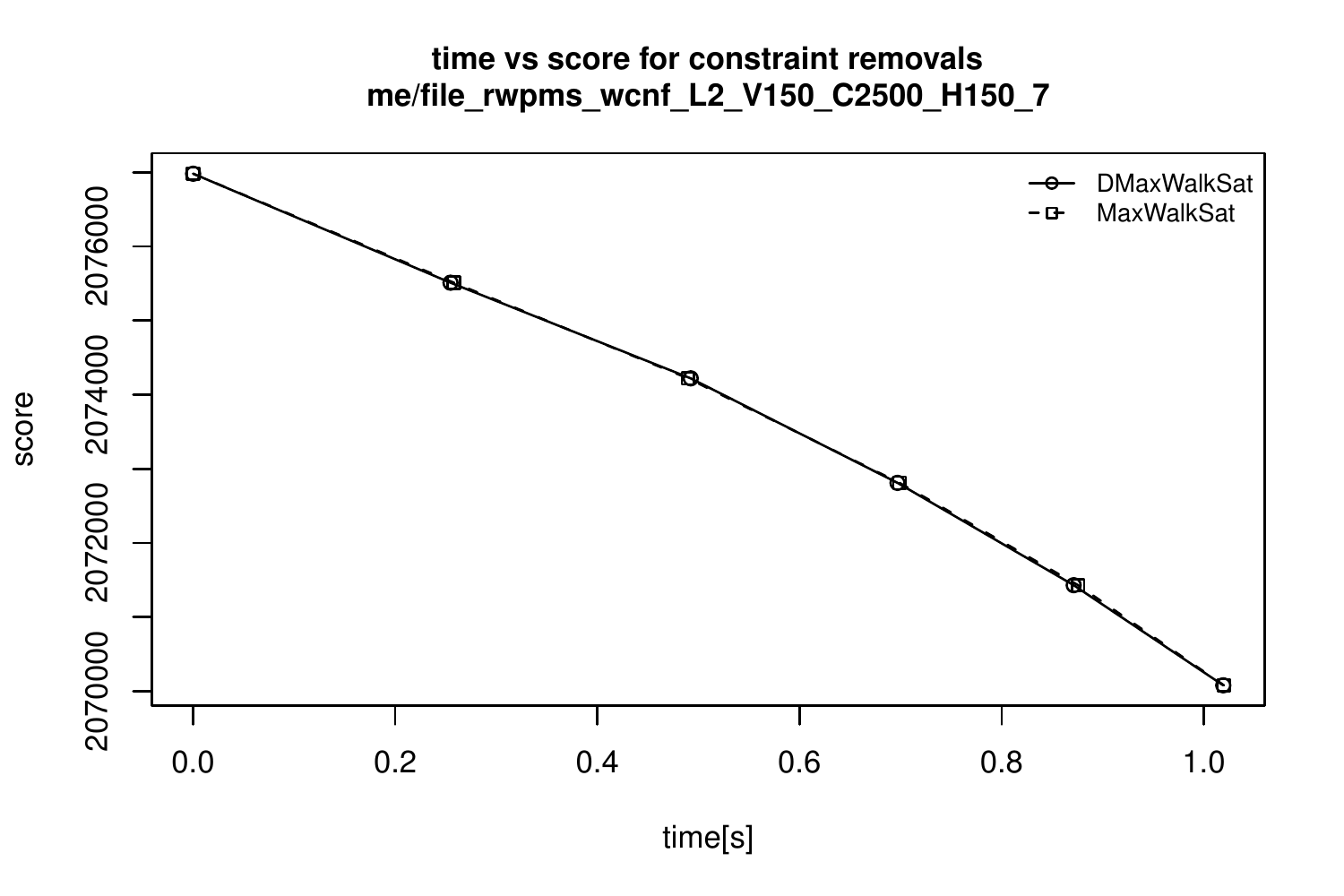}
        }

    \caption*{me/file\_rwpms\_wcnf\_L2\_V150\_C2500\_H150\_7}
    \label{fig_me/file_rwpms_wcnf_L2_V150_C2500_H150_7}
\end{figure}

\begin{figure}[H]
    \setcounter{subfigure}{0}
    \centering
        \subfloat[Constraint addition]
        {
            \includegraphics[width=2.7in]{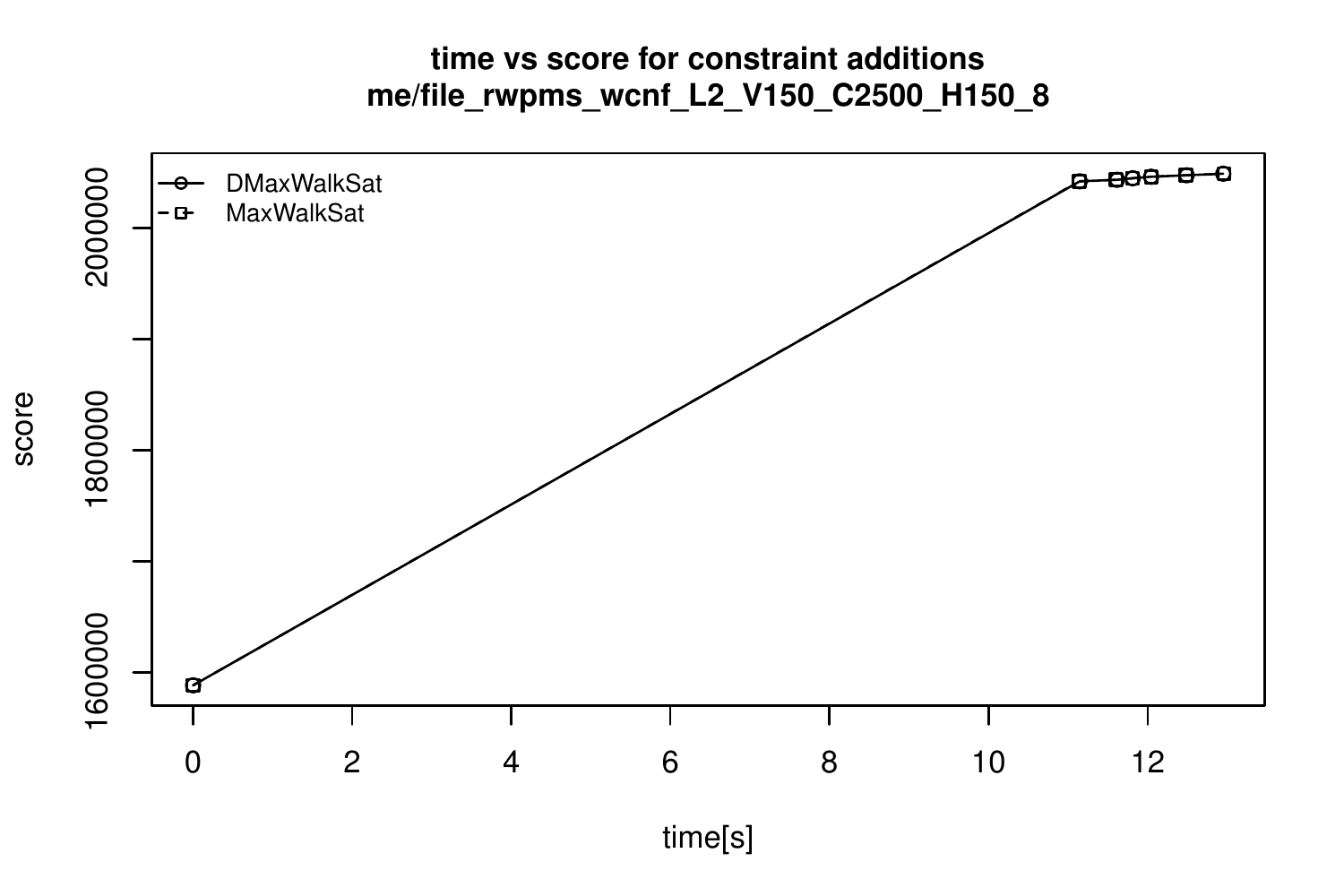}
        }
        \qquad
        \subfloat[Constraint removal]
        {
            \includegraphics[width=2.7in]{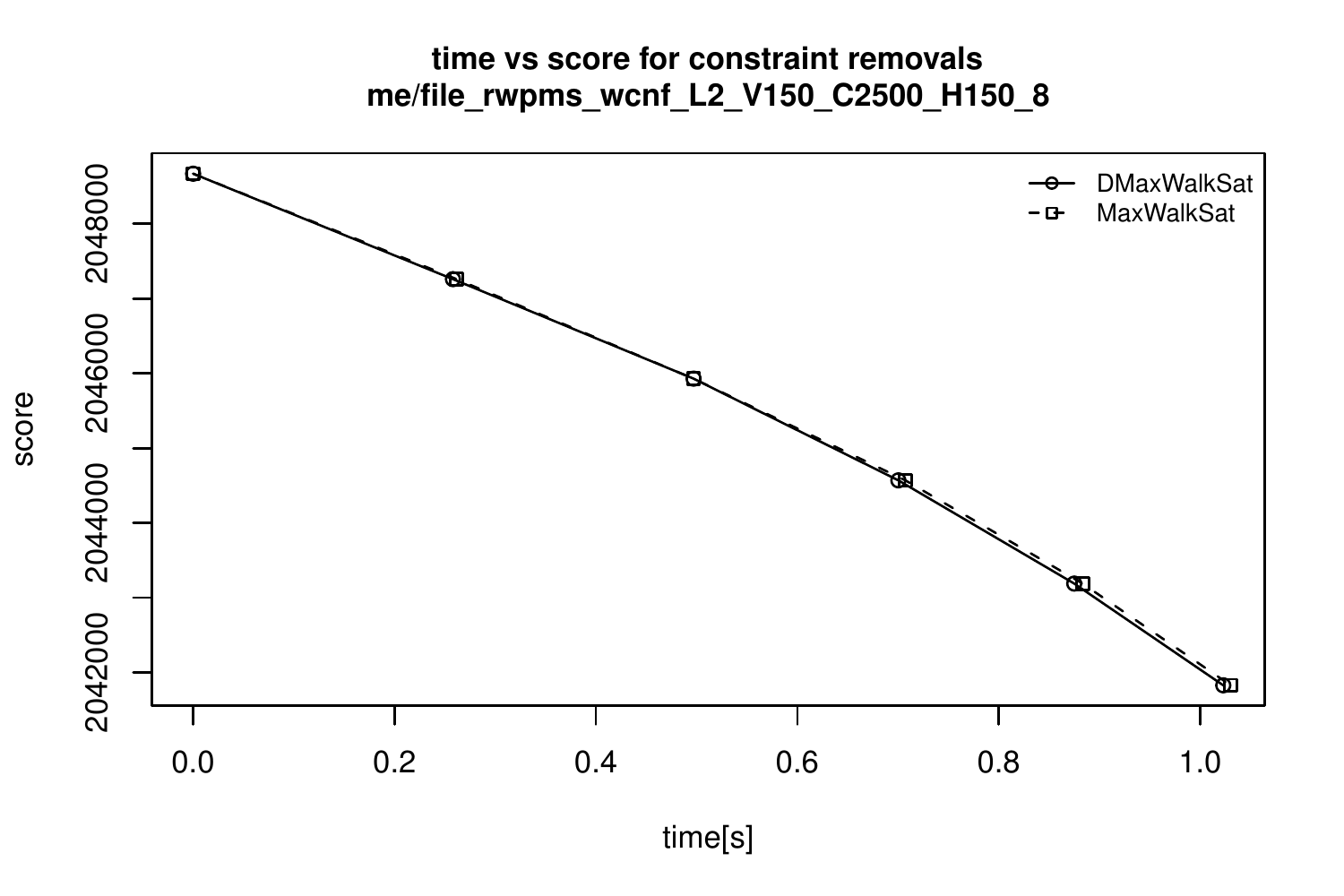}
        }

    \caption*{me/file\_rwpms\_wcnf\_L2\_V150\_C2500\_H150\_8}
    \label{fig_me/file_rwpms_wcnf_L2_V150_C2500_H150_8}
\end{figure}

\begin{figure}[H]
    \setcounter{subfigure}{0}
    \centering
        \subfloat[Constraint addition]
        {
            \includegraphics[width=2.7in]{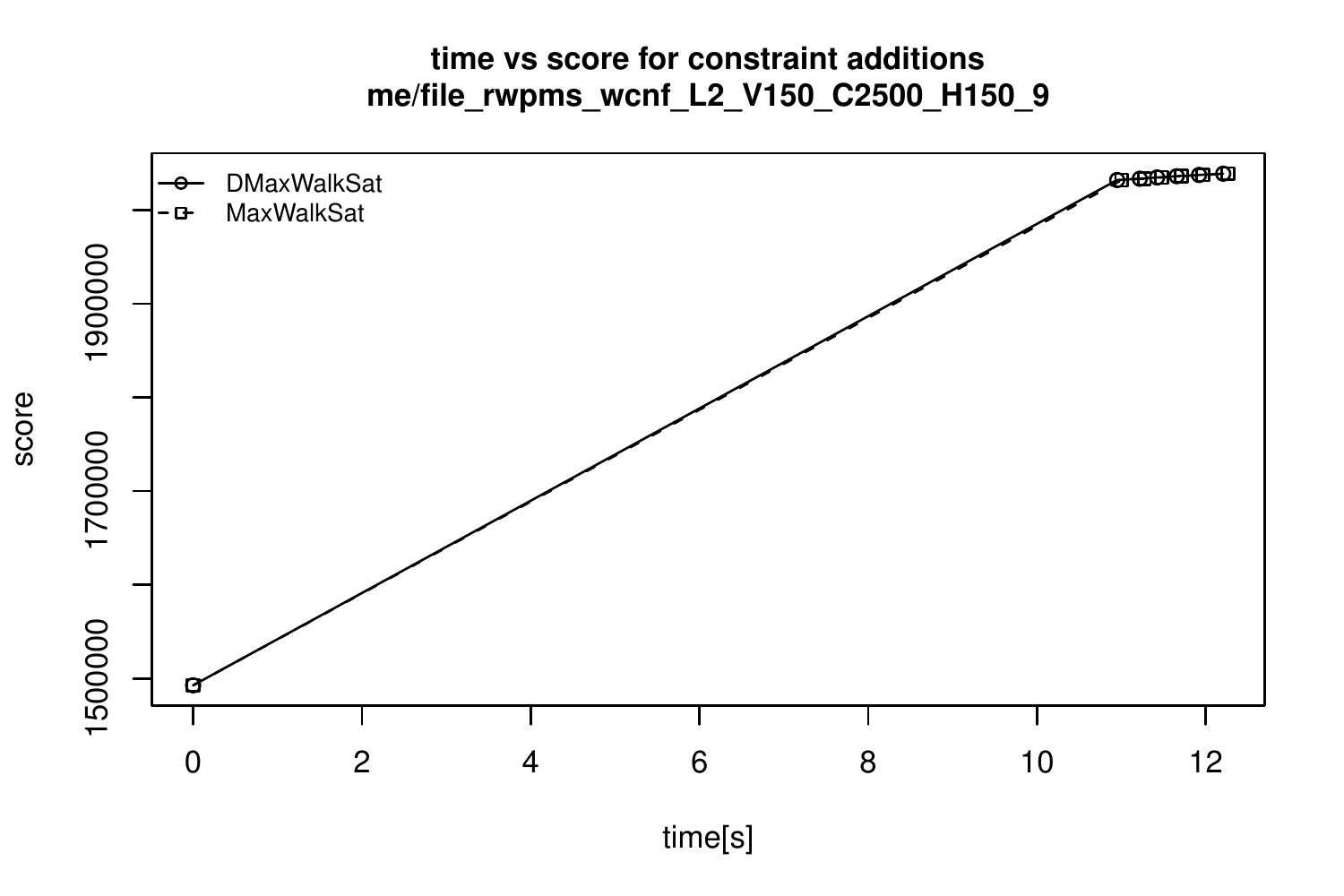}
        }
        \qquad
        \subfloat[Constraint removal]
        {
            \includegraphics[width=2.7in]{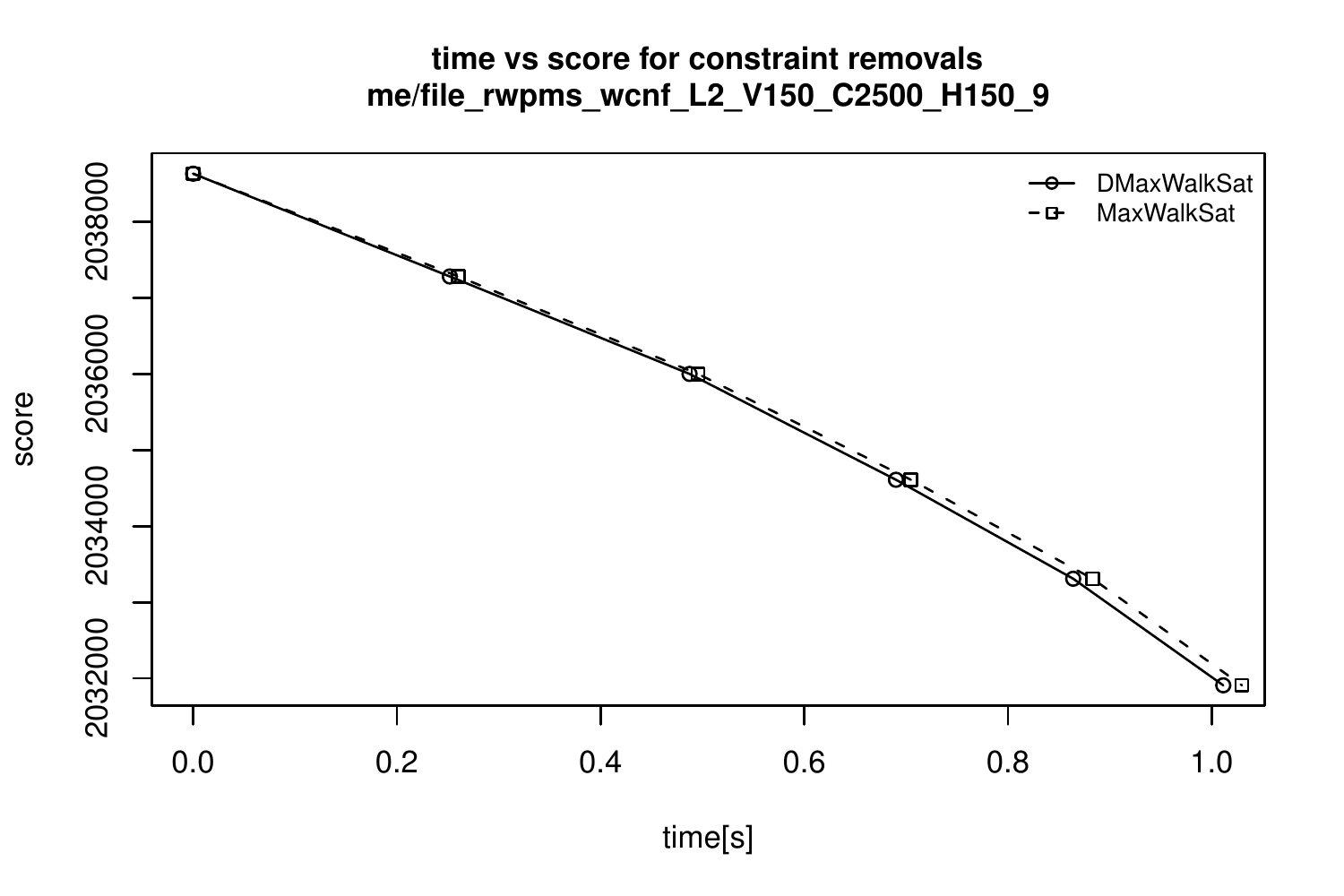}
        }

    \caption*{me/file\_rwpms\_wcnf\_L2\_V150\_C2500\_H150\_9}
    \label{fig_me/file_rwpms_wcnf_L2_V150_C2500_H150_9}
\end{figure}

\begin{figure}[H]
    \setcounter{subfigure}{0}
    \centering
        \subfloat[Constraint addition]
        {
            \includegraphics[width=2.7in]{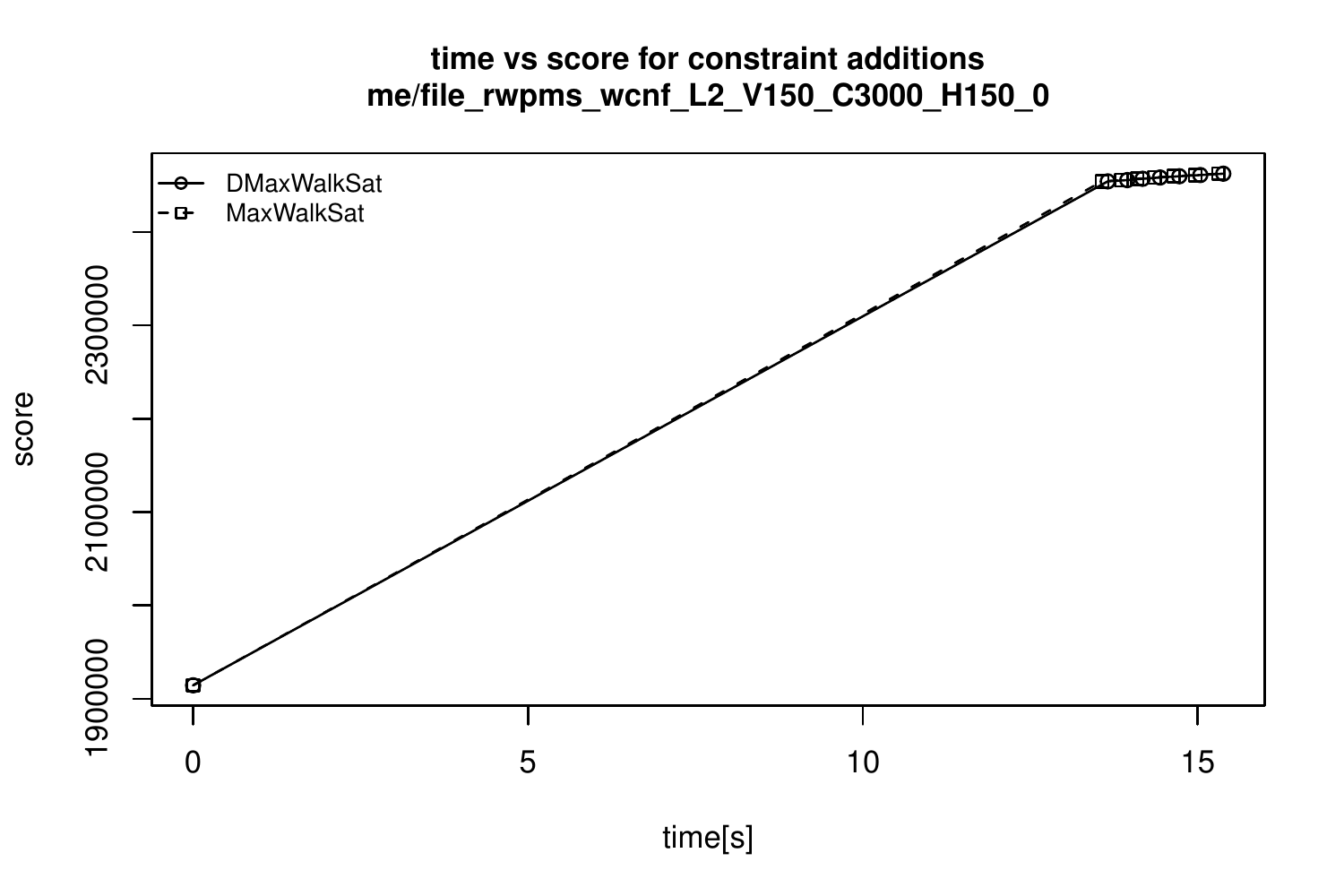}
        }
        \qquad
        \subfloat[Constraint removal]
        {
            \includegraphics[width=2.7in]{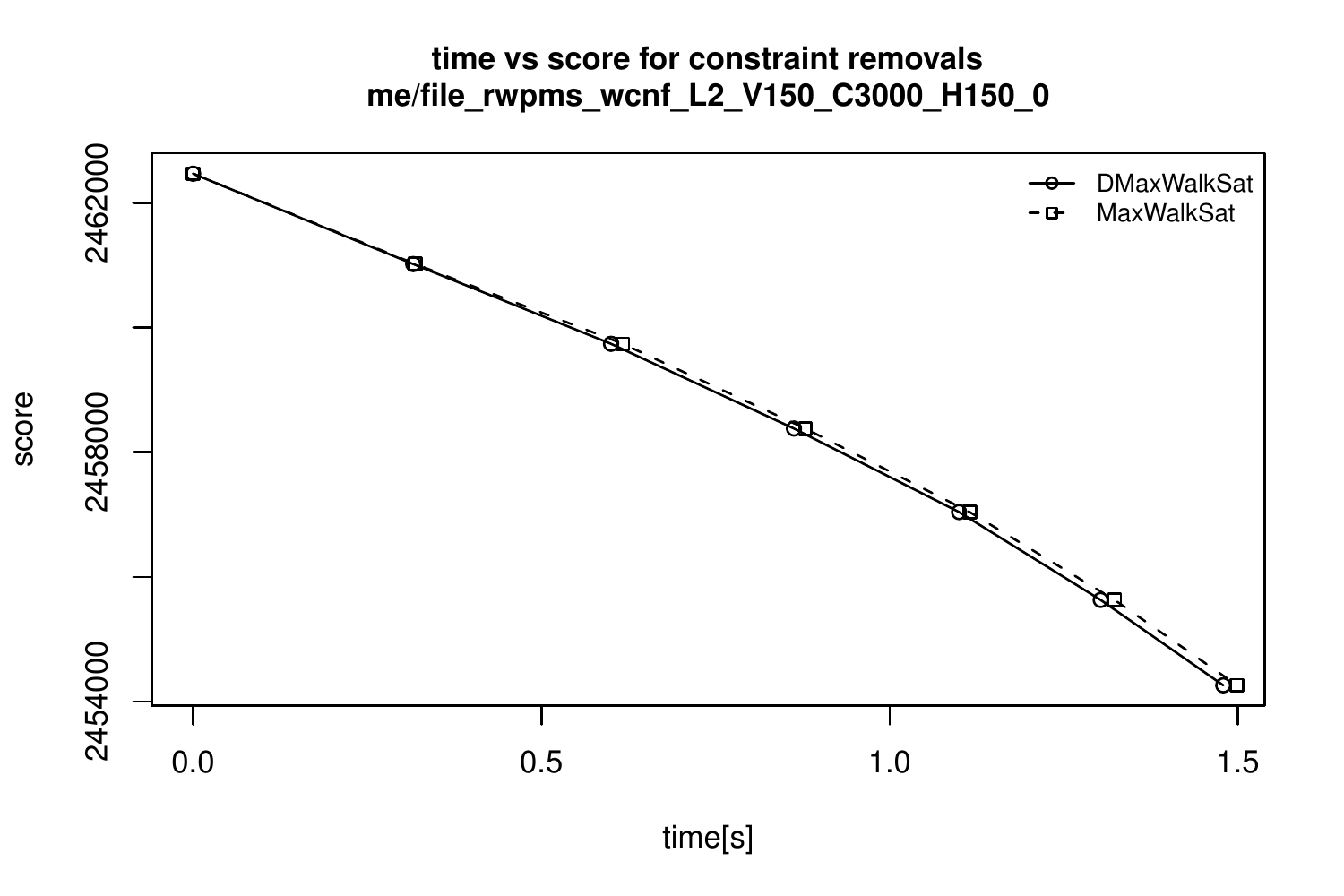}
        }

    \caption*{me/file\_rwpms\_wcnf\_L2\_V150\_C3000\_H150\_0}
    \label{fig_me/file_rwpms_wcnf_L2_V150_C3000_H150_0}
\end{figure}

\begin{figure}[H]
    \setcounter{subfigure}{0}
    \centering
        \subfloat[Constraint addition]
        {
            \includegraphics[width=2.7in]{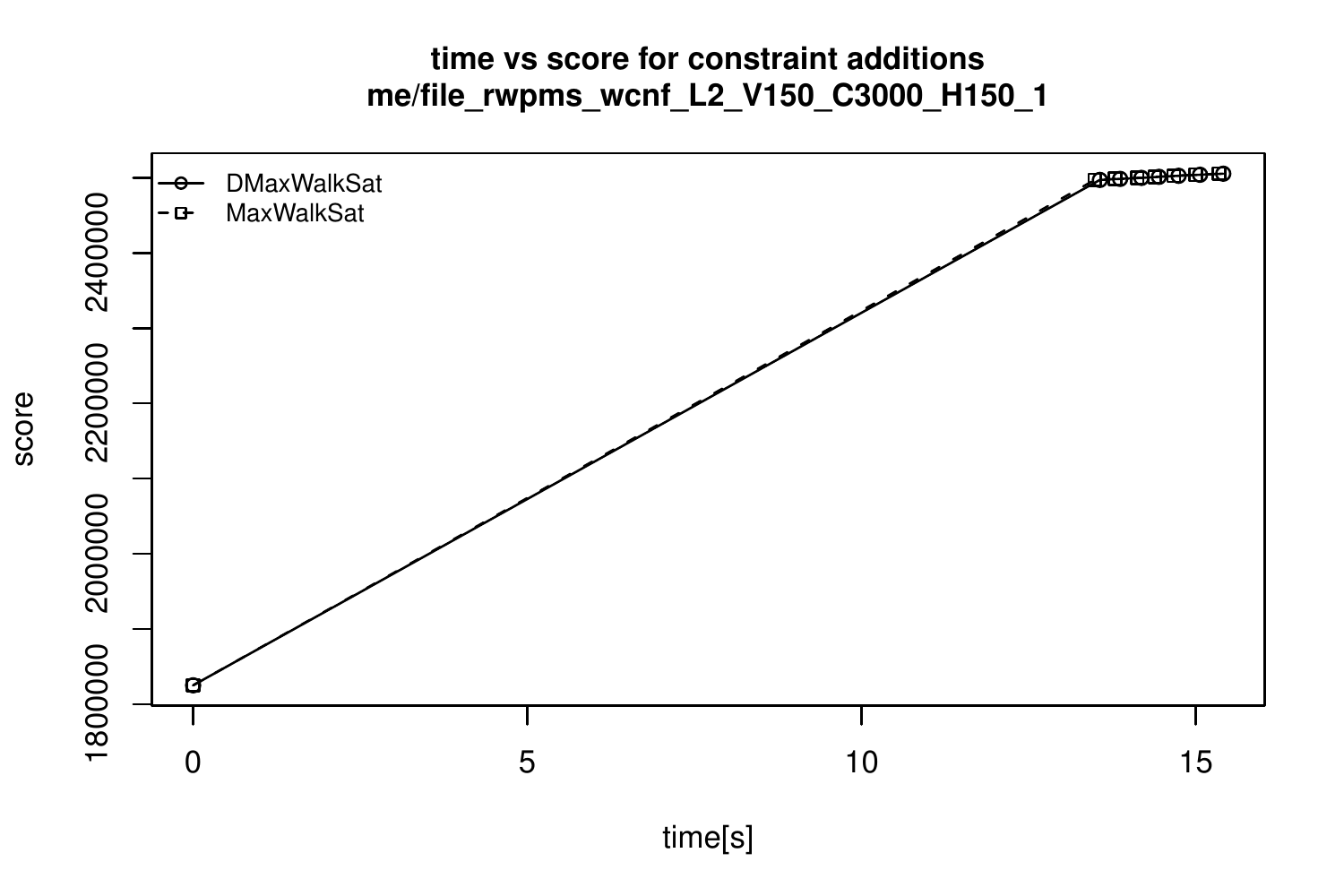}
        }
        \qquad
        \subfloat[Constraint removal]
        {
            \includegraphics[width=2.7in]{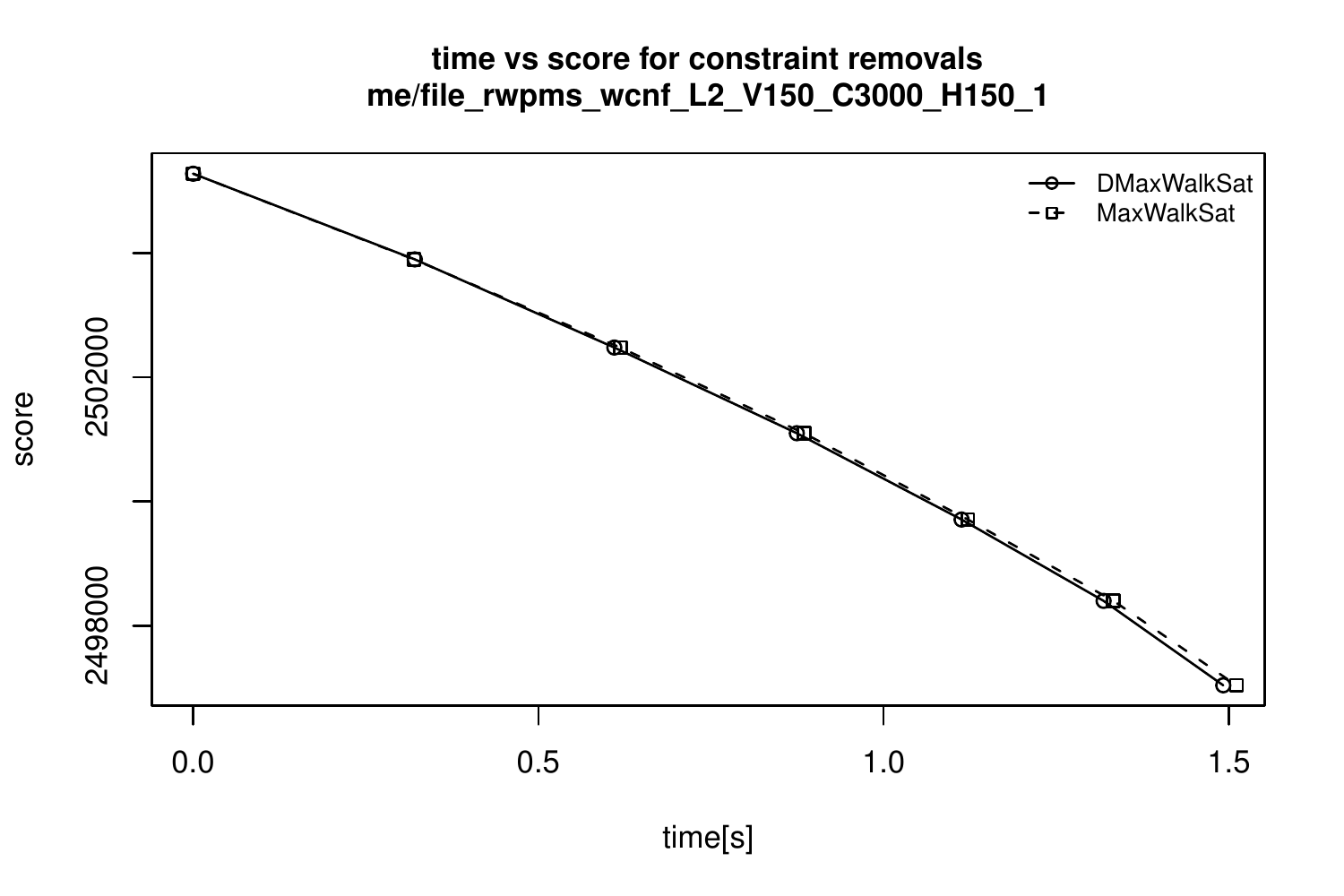}
        }

    \caption*{me/file\_rwpms\_wcnf\_L2\_V150\_C3000\_H150\_1}
    \label{fig_me/file_rwpms_wcnf_L2_V150_C3000_H150_1}
\end{figure}

\begin{figure}[H]
    \setcounter{subfigure}{0}
    \centering
        \subfloat[Constraint addition]
        {
            \includegraphics[width=2.7in]{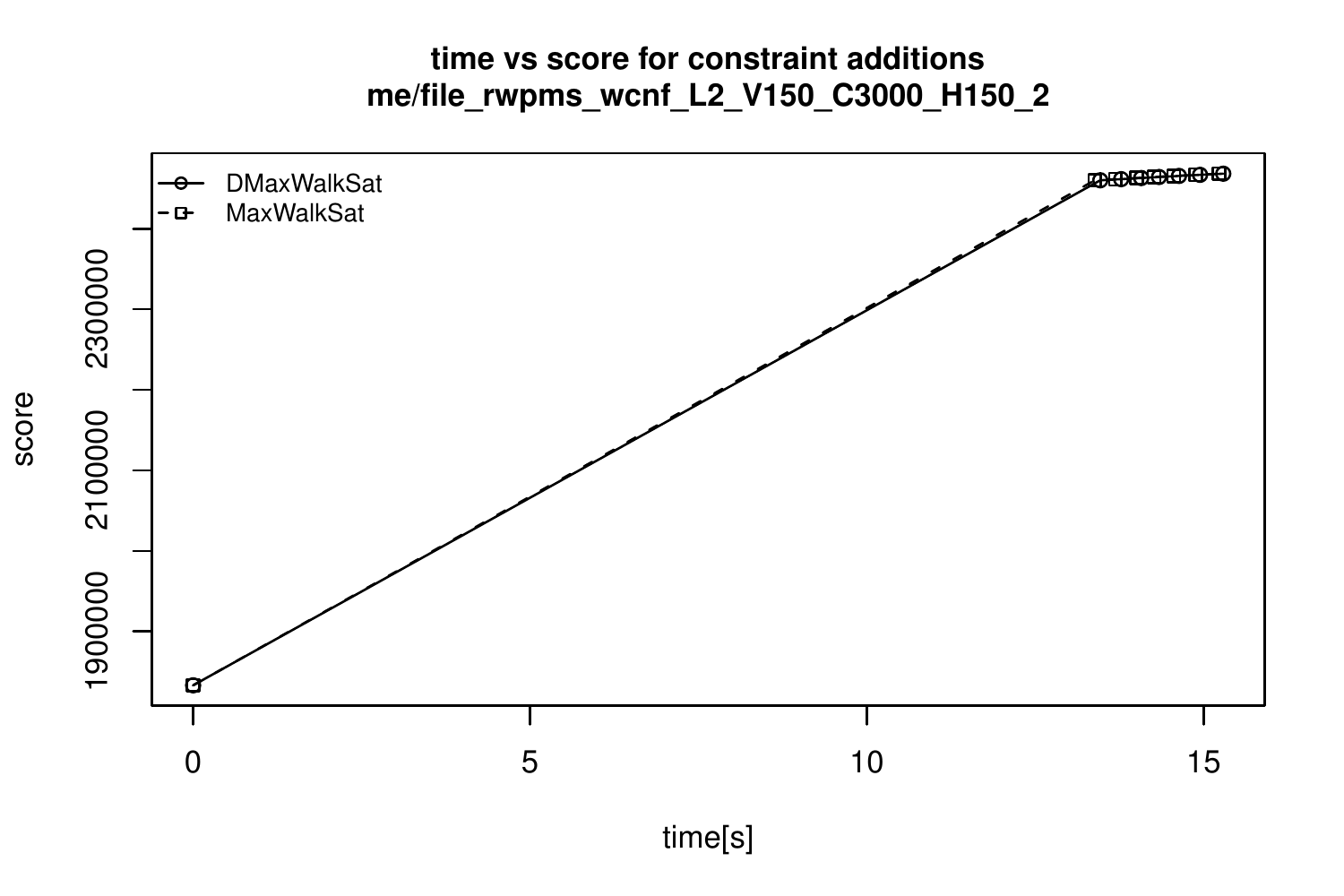}
        }
        \qquad
        \subfloat[Constraint removal]
        {
            \includegraphics[width=2.7in]{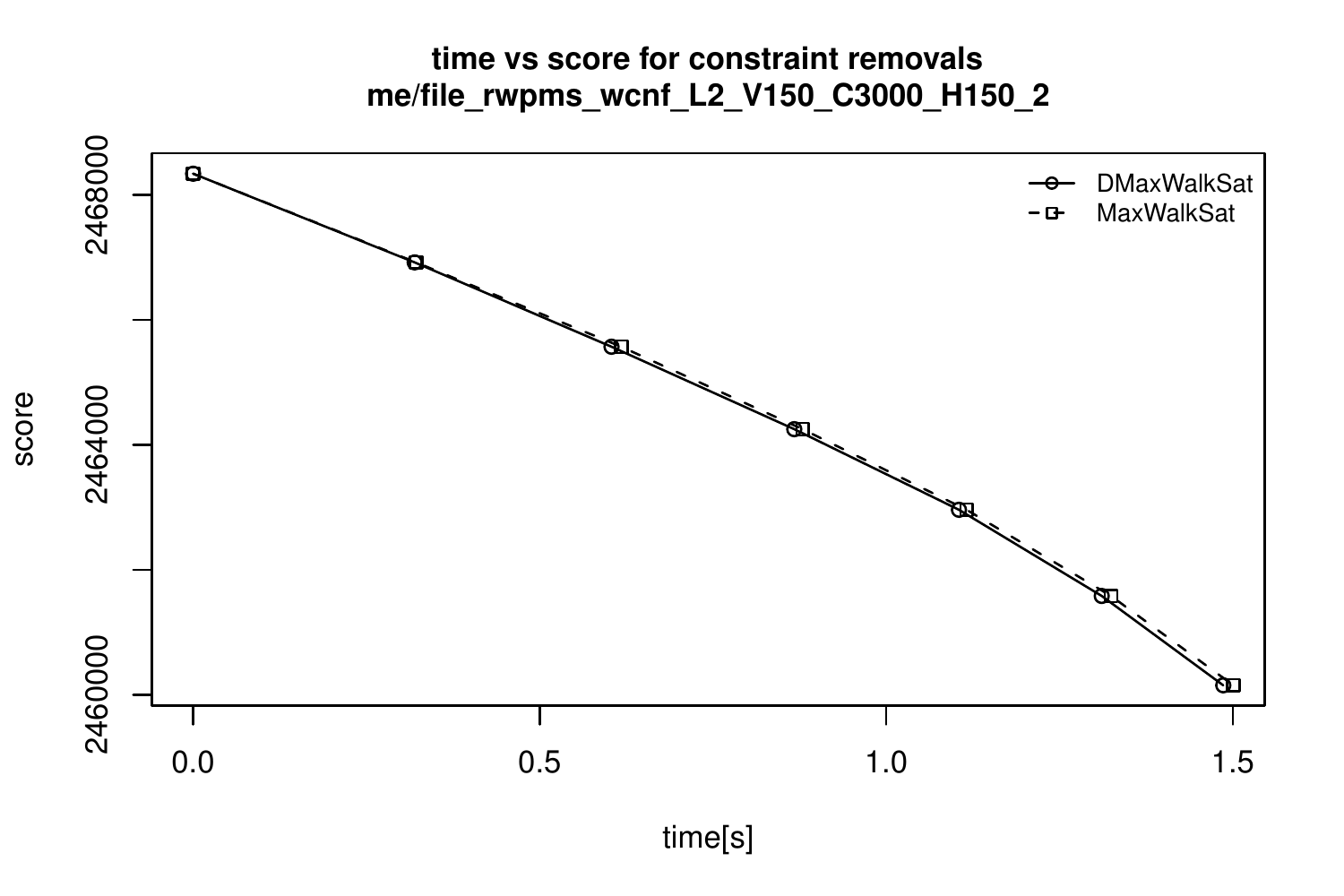}
        }

    \caption*{me/file\_rwpms\_wcnf\_L2\_V150\_C3000\_H150\_2}
    \label{fig_me/file_rwpms_wcnf_L2_V150_C3000_H150_2}
\end{figure}

\begin{figure}[H]
    \setcounter{subfigure}{0}
    \centering
        \subfloat[Constraint addition]
        {
            \includegraphics[width=2.7in]{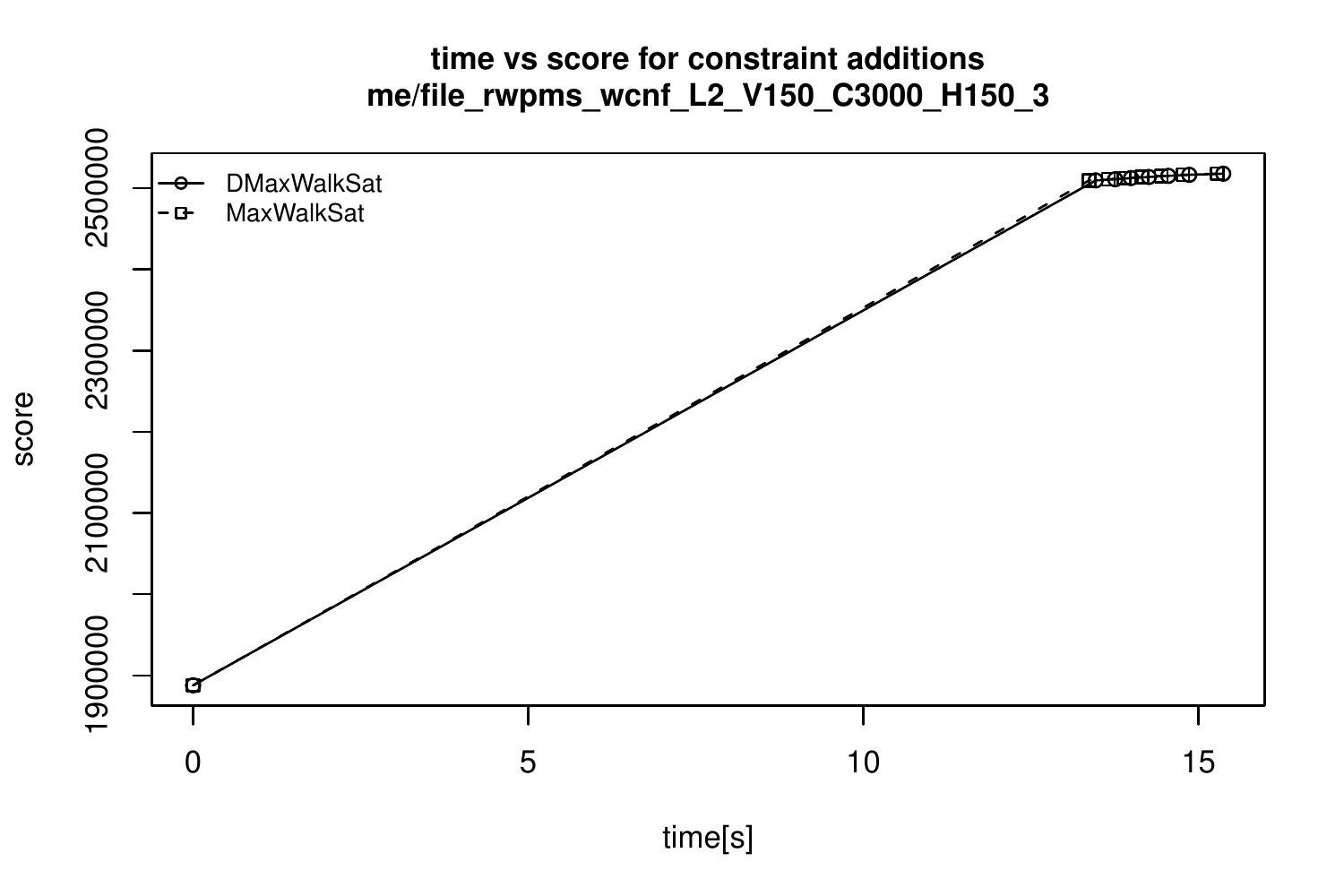}
        }
        \qquad
        \subfloat[Constraint removal]
        {
            \includegraphics[width=2.7in]{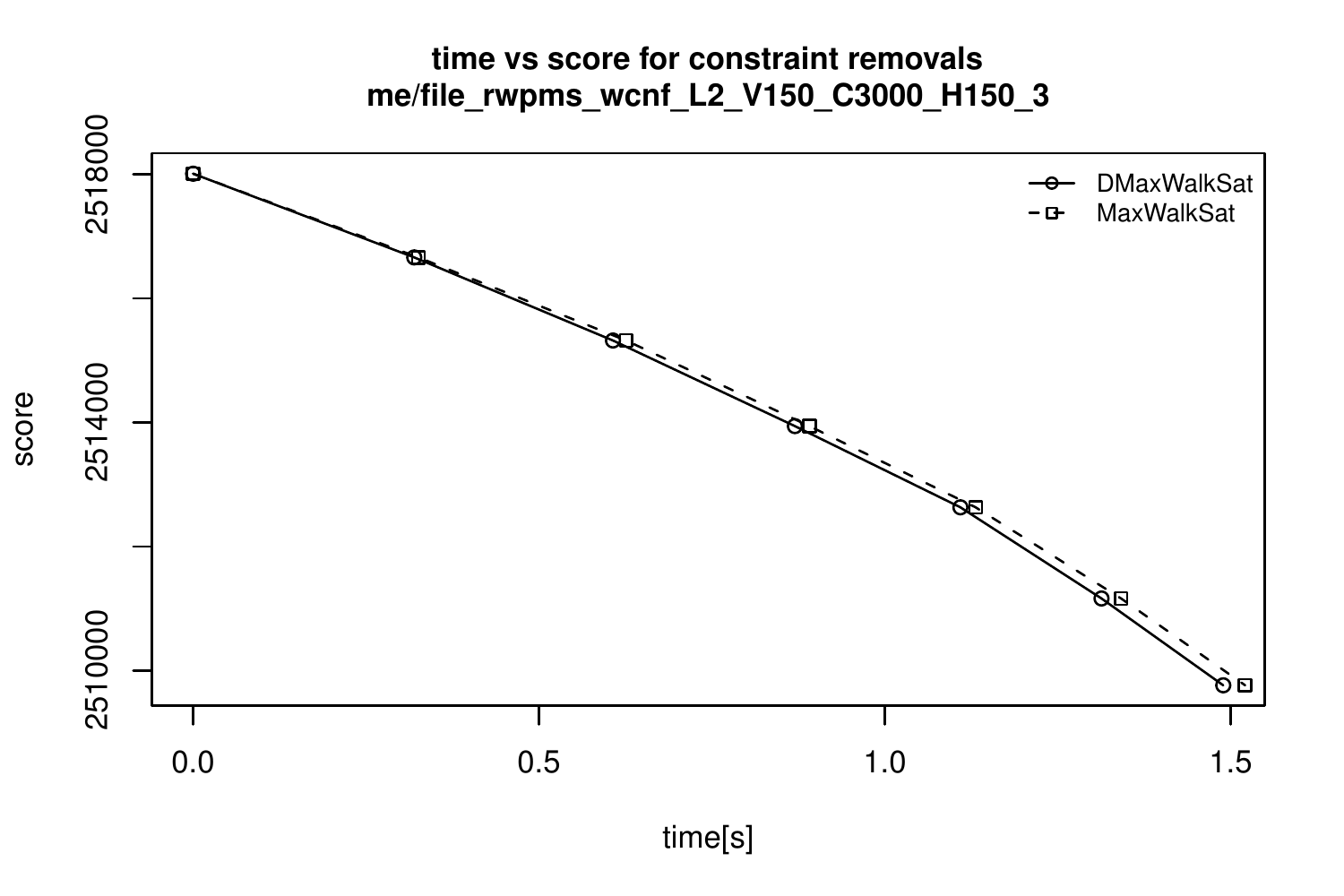}
        }

    \caption*{me/file\_rwpms\_wcnf\_L2\_V150\_C3000\_H150\_3}
    \label{fig_me/file_rwpms_wcnf_L2_V150_C3000_H150_3}
\end{figure}

\begin{figure}[H]
    \setcounter{subfigure}{0}
    \centering
        \subfloat[Constraint addition]
        {
            \includegraphics[width=2.7in]{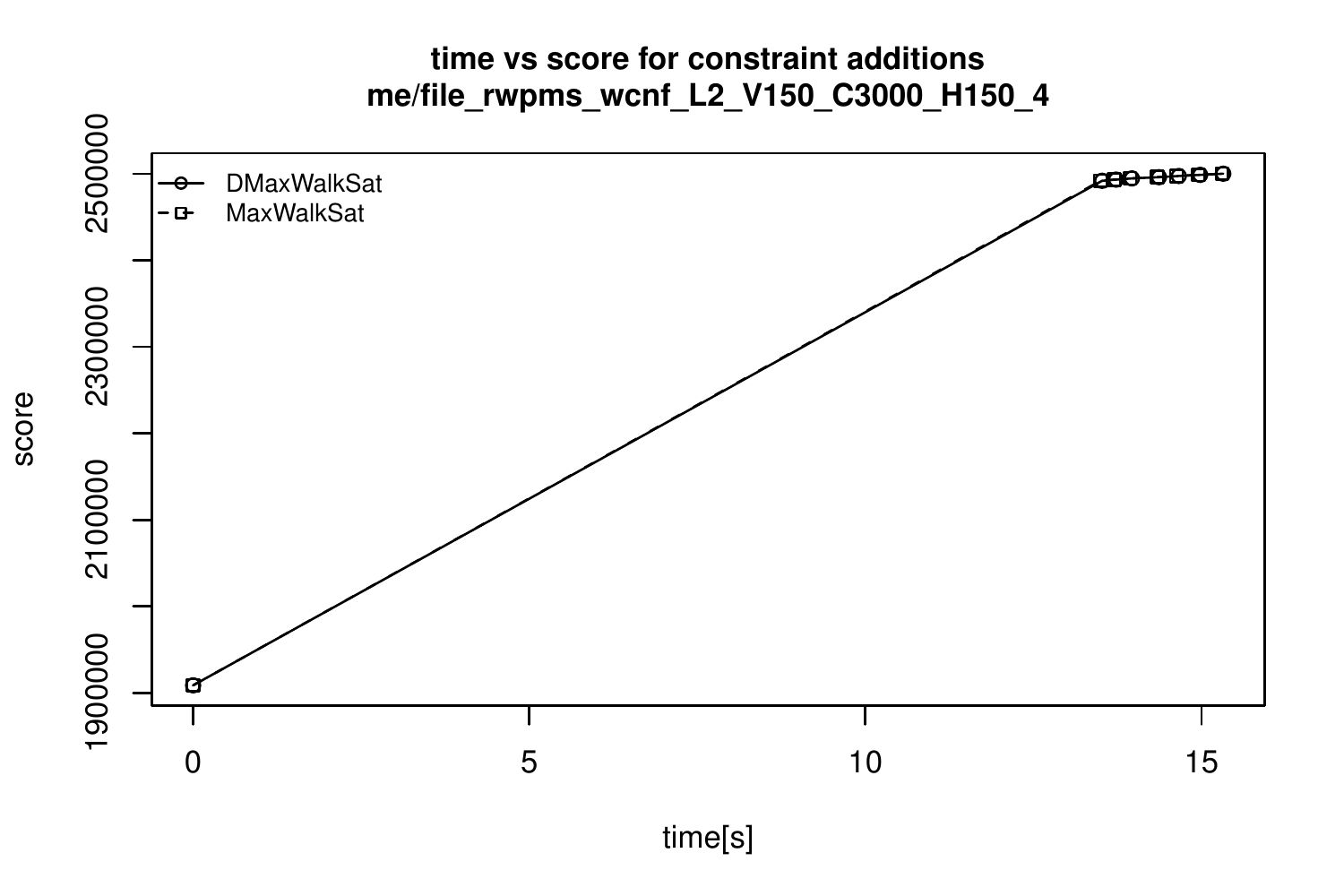}
        }
        \qquad
        \subfloat[Constraint removal]
        {
            \includegraphics[width=2.7in]{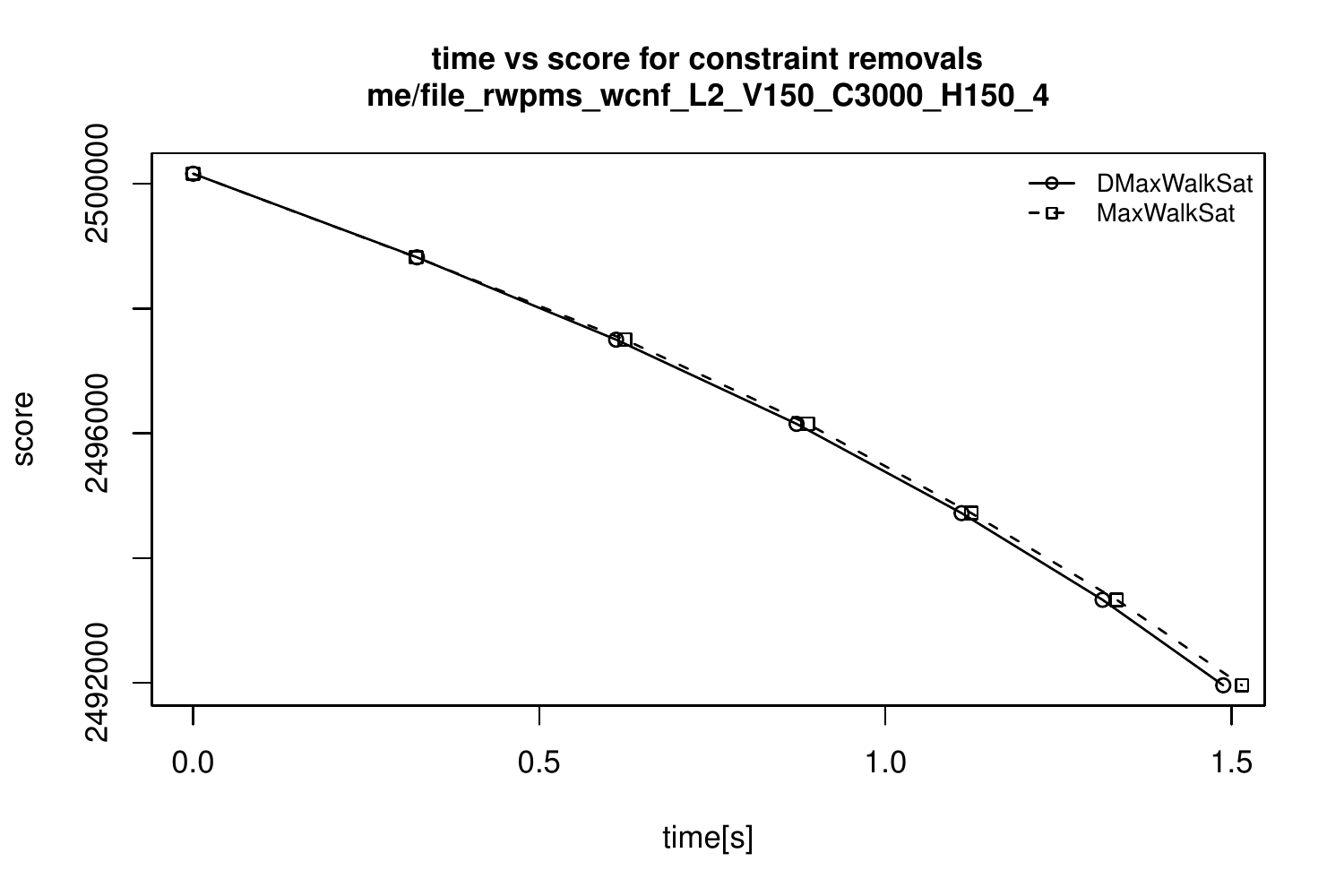}
        }

    \caption*{me/file\_rwpms\_wcnf\_L2\_V150\_C3000\_H150\_4}
    \label{fig_me/file_rwpms_wcnf_L2_V150_C3000_H150_4}
\end{figure}

\begin{figure}[H]
    \setcounter{subfigure}{0}
    \centering
        \subfloat[Constraint addition]
        {
            \includegraphics[width=2.7in]{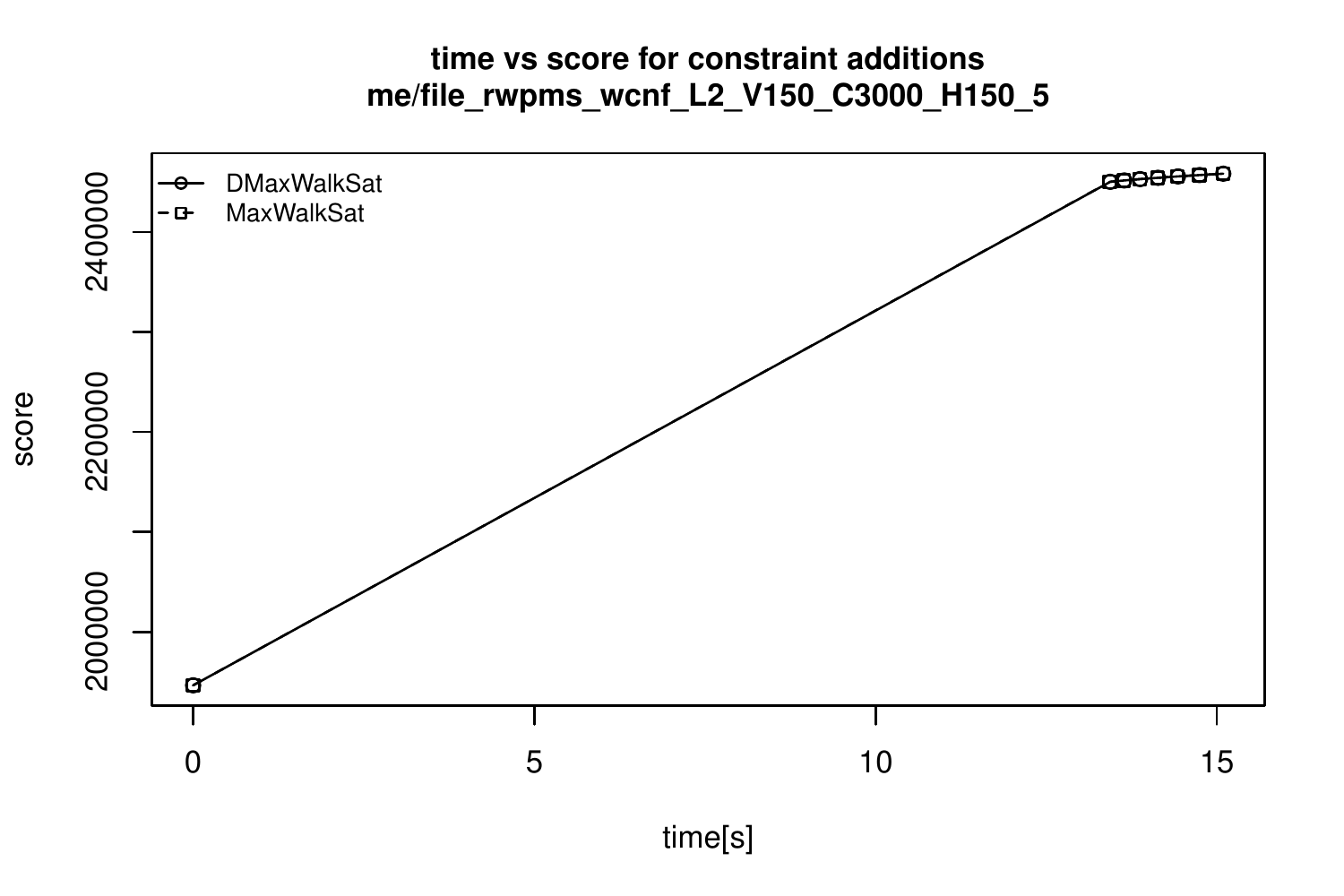}
        }
        \qquad
        \subfloat[Constraint removal]
        {
            \includegraphics[width=2.7in]{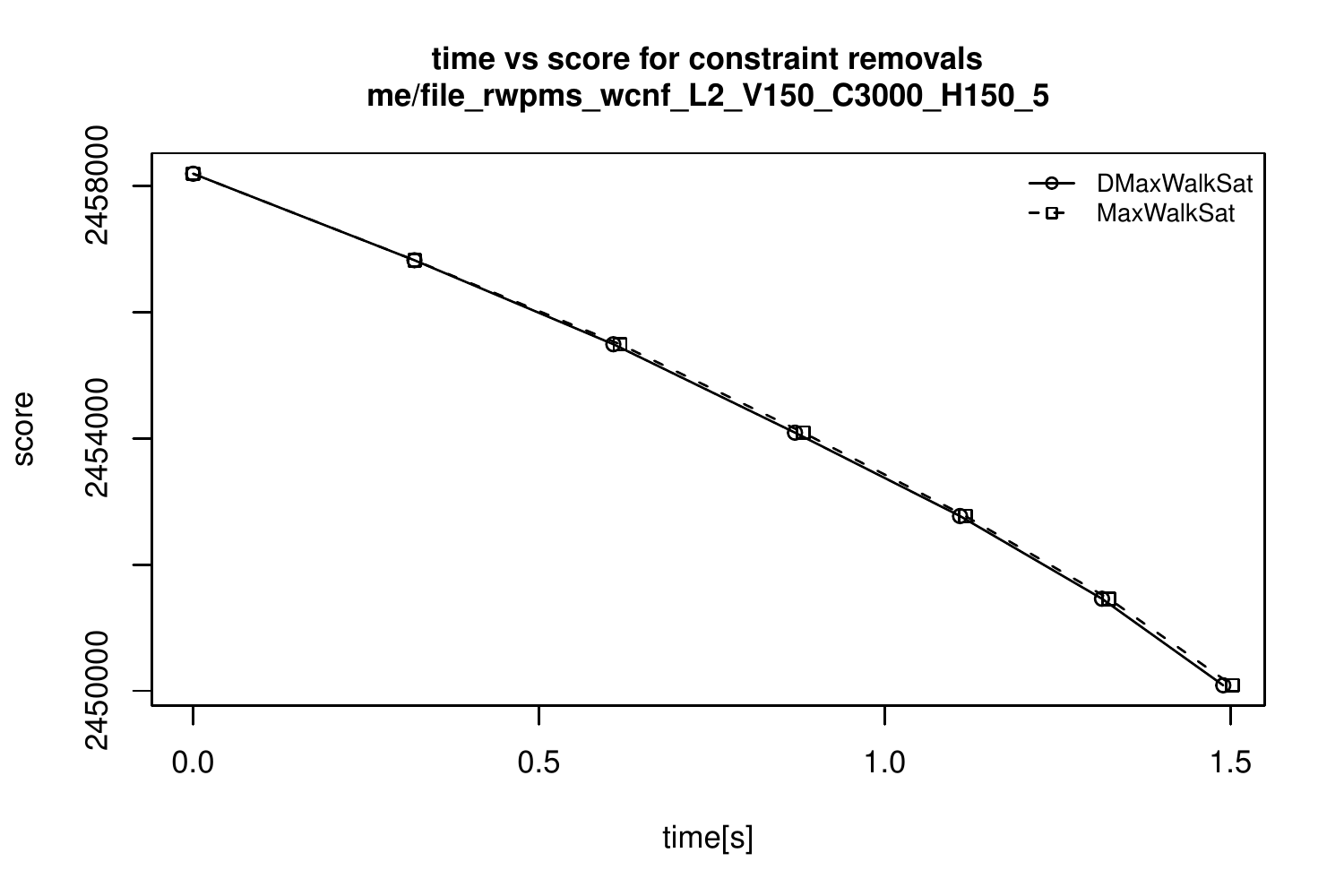}
        }

    \caption*{me/file\_rwpms\_wcnf\_L2\_V150\_C3000\_H150\_5}
    \label{fig_me/file_rwpms_wcnf_L2_V150_C3000_H150_5}
\end{figure}

\begin{figure}[H]
    \setcounter{subfigure}{0}
    \centering
        \subfloat[Constraint addition]
        {
            \includegraphics[width=2.7in]{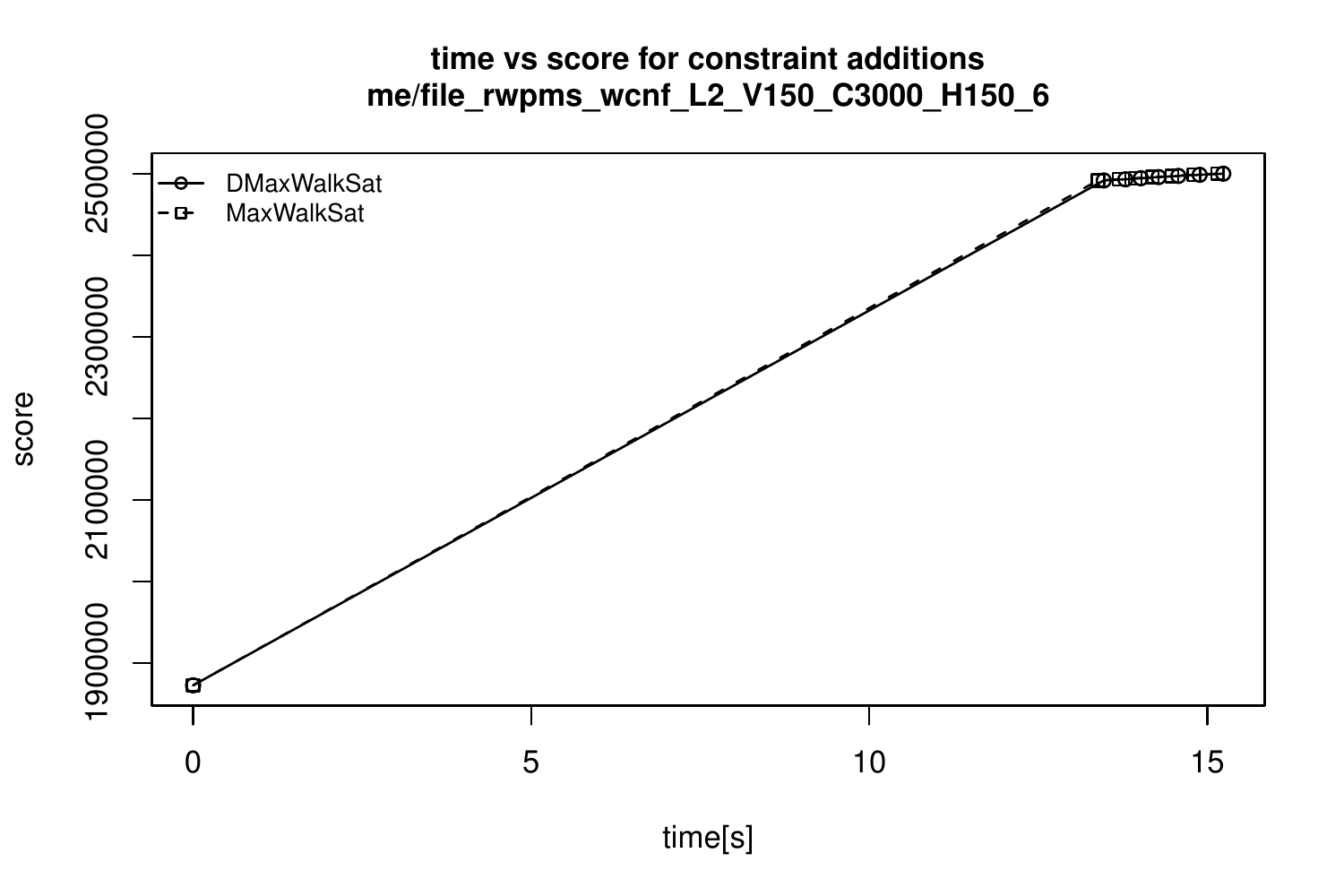}
        }
        \qquad
        \subfloat[Constraint removal]
        {
            \includegraphics[width=2.7in]{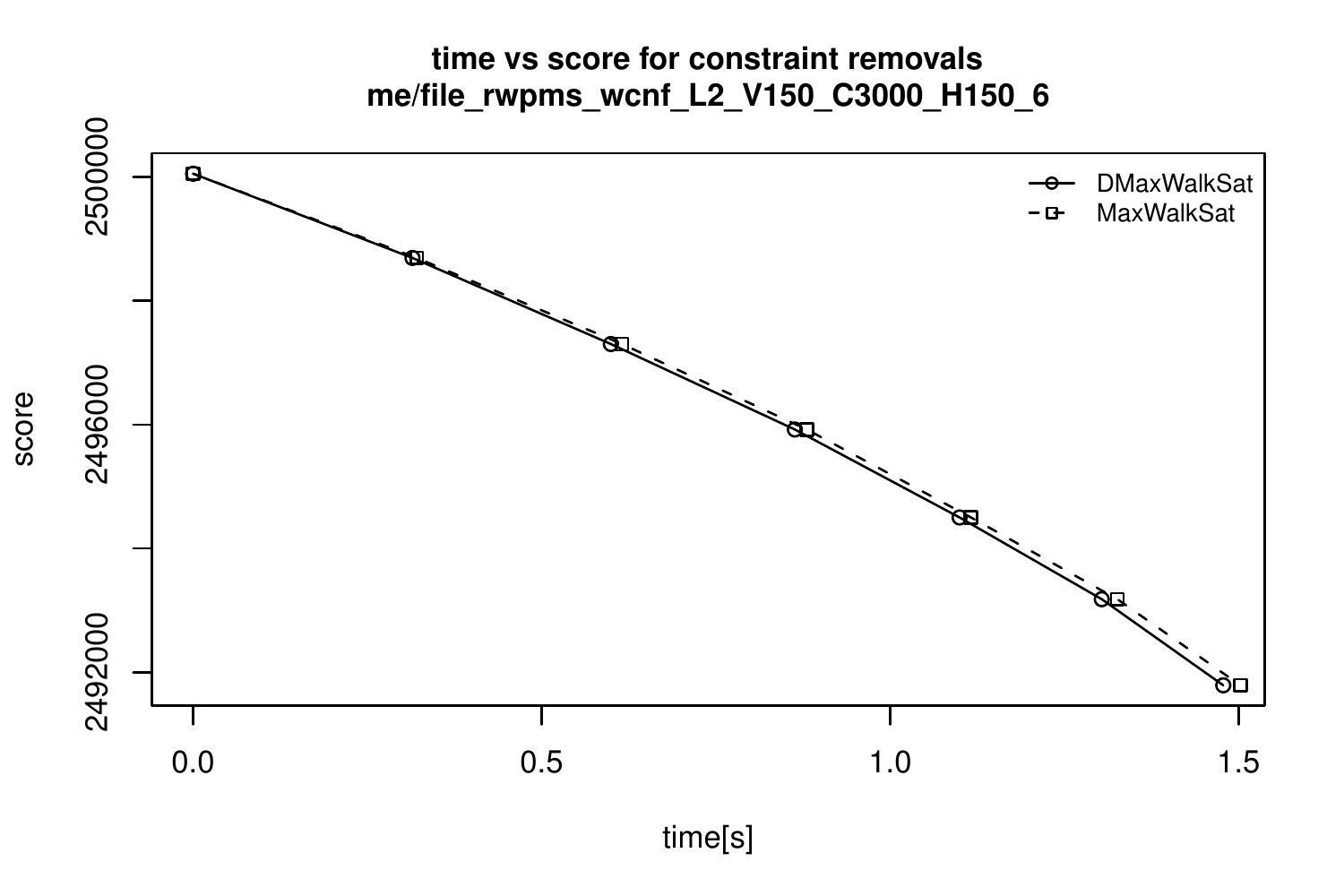}
        }

    \caption*{me/file\_rwpms\_wcnf\_L2\_V150\_C3000\_H150\_6}
    \label{fig_me/file_rwpms_wcnf_L2_V150_C3000_H150_6}
\end{figure}

\begin{figure}[H]
    \setcounter{subfigure}{0}
    \centering
        \subfloat[Constraint addition]
        {
            \includegraphics[width=2.7in]{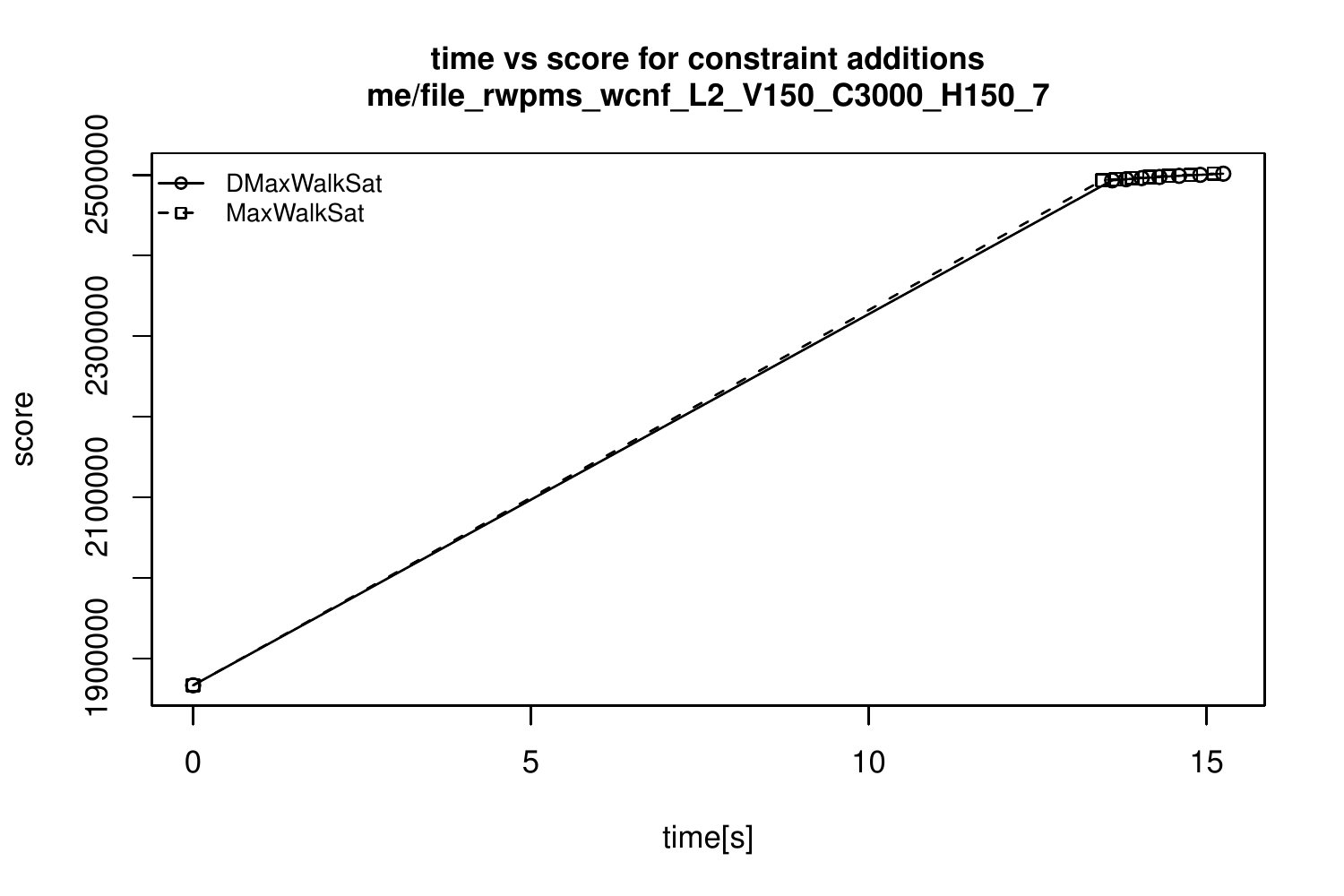}
        }
        \qquad
        \subfloat[Constraint removal]
        {
            \includegraphics[width=2.7in]{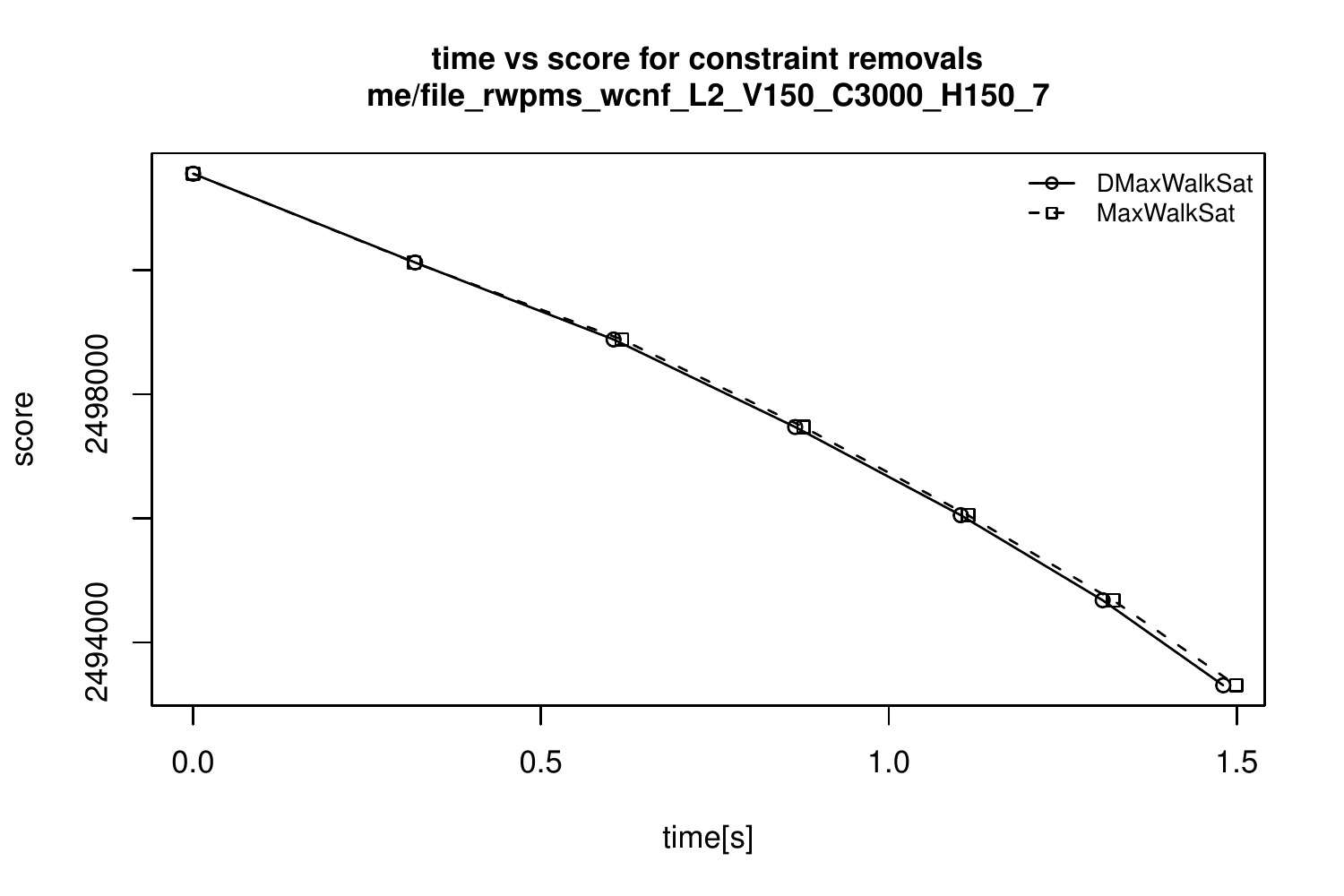}
        }

    \caption*{me/file\_rwpms\_wcnf\_L2\_V150\_C3000\_H150\_7}
    \label{fig_me/file_rwpms_wcnf_L2_V150_C3000_H150_7}
\end{figure}

\begin{figure}[H]
    \setcounter{subfigure}{0}
    \centering
        \subfloat[Constraint addition]
        {
            \includegraphics[width=2.7in]{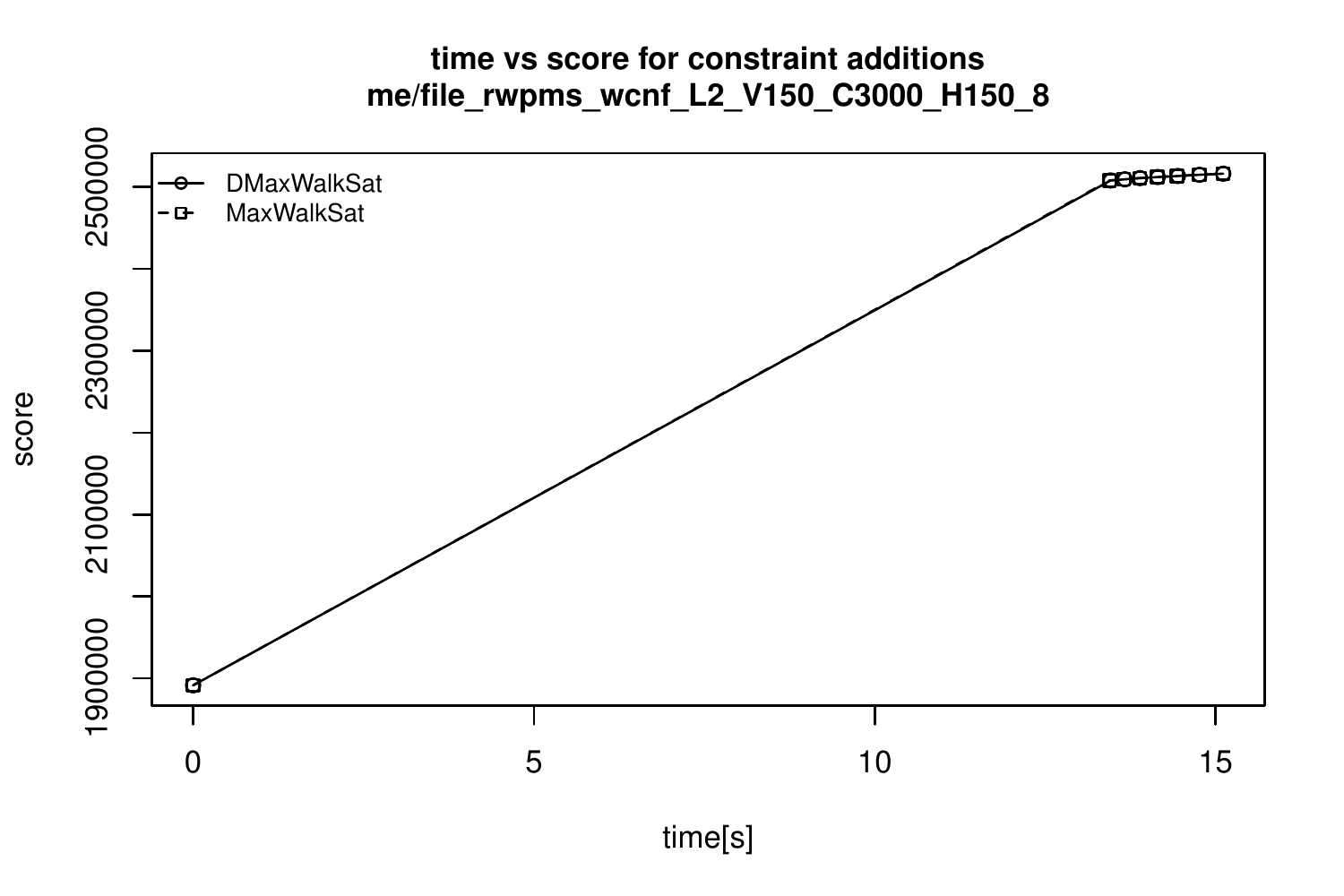}
        }
        \qquad
        \subfloat[Constraint removal]
        {
            \includegraphics[width=2.7in]{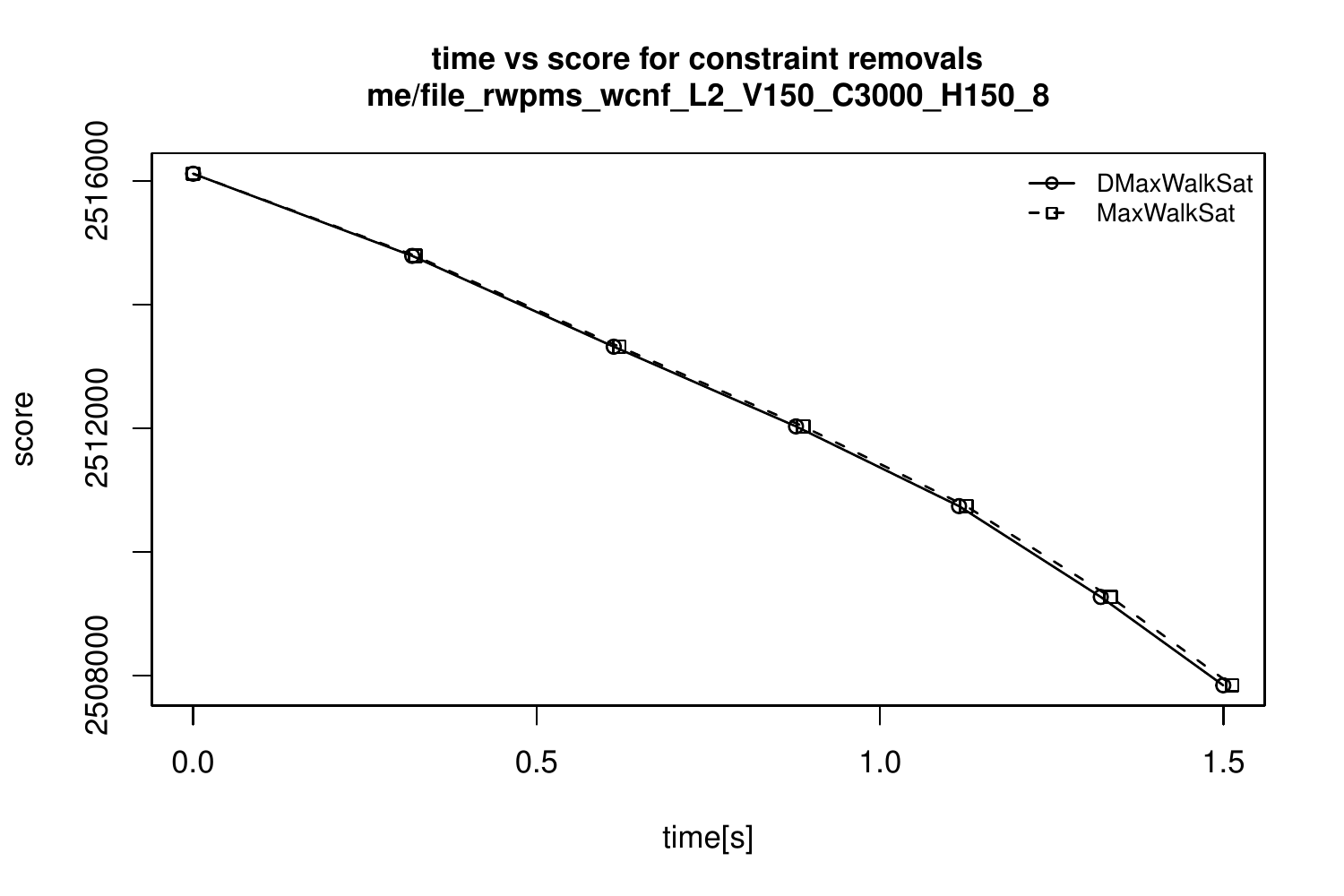}
        }

    \caption*{me/file\_rwpms\_wcnf\_L2\_V150\_C3000\_H150\_8}
    \label{fig_me/file_rwpms_wcnf_L2_V150_C3000_H150_8}
\end{figure}

\begin{figure}[H]
    \setcounter{subfigure}{0}
    \centering
        \subfloat[Constraint addition]
        {
            \includegraphics[width=2.7in]{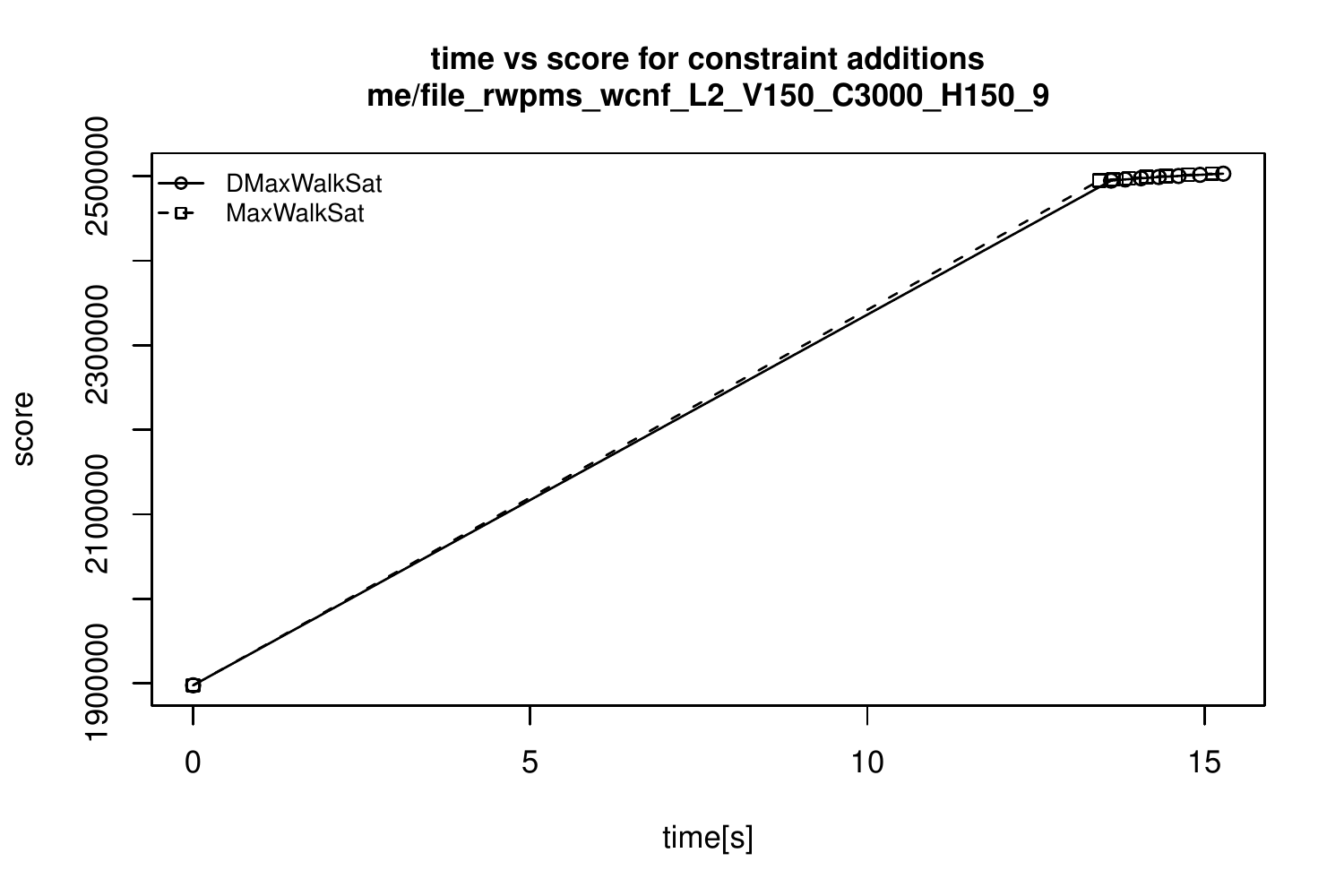}
        }
        \qquad
        \subfloat[Constraint removal]
        {
            \includegraphics[width=2.7in]{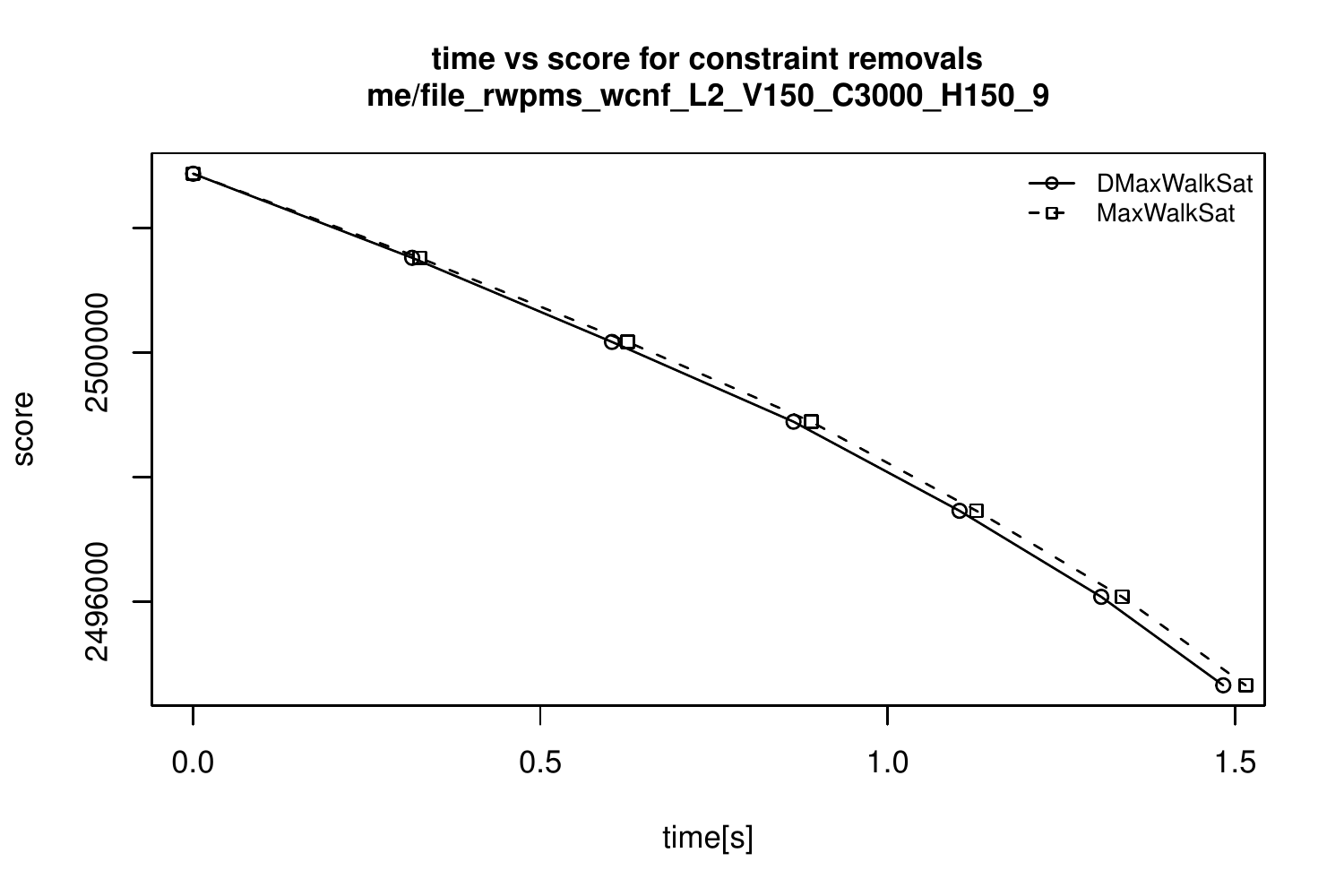}
        }

    \caption*{me/file\_rwpms\_wcnf\_L2\_V150\_C3000\_H150\_9}
    \label{fig_me/file_rwpms_wcnf_L2_V150_C3000_H150_9}
\end{figure}

\begin{figure}[H]
    \setcounter{subfigure}{0}
    \centering
        \subfloat[Constraint addition]
        {
            \includegraphics[width=2.7in]{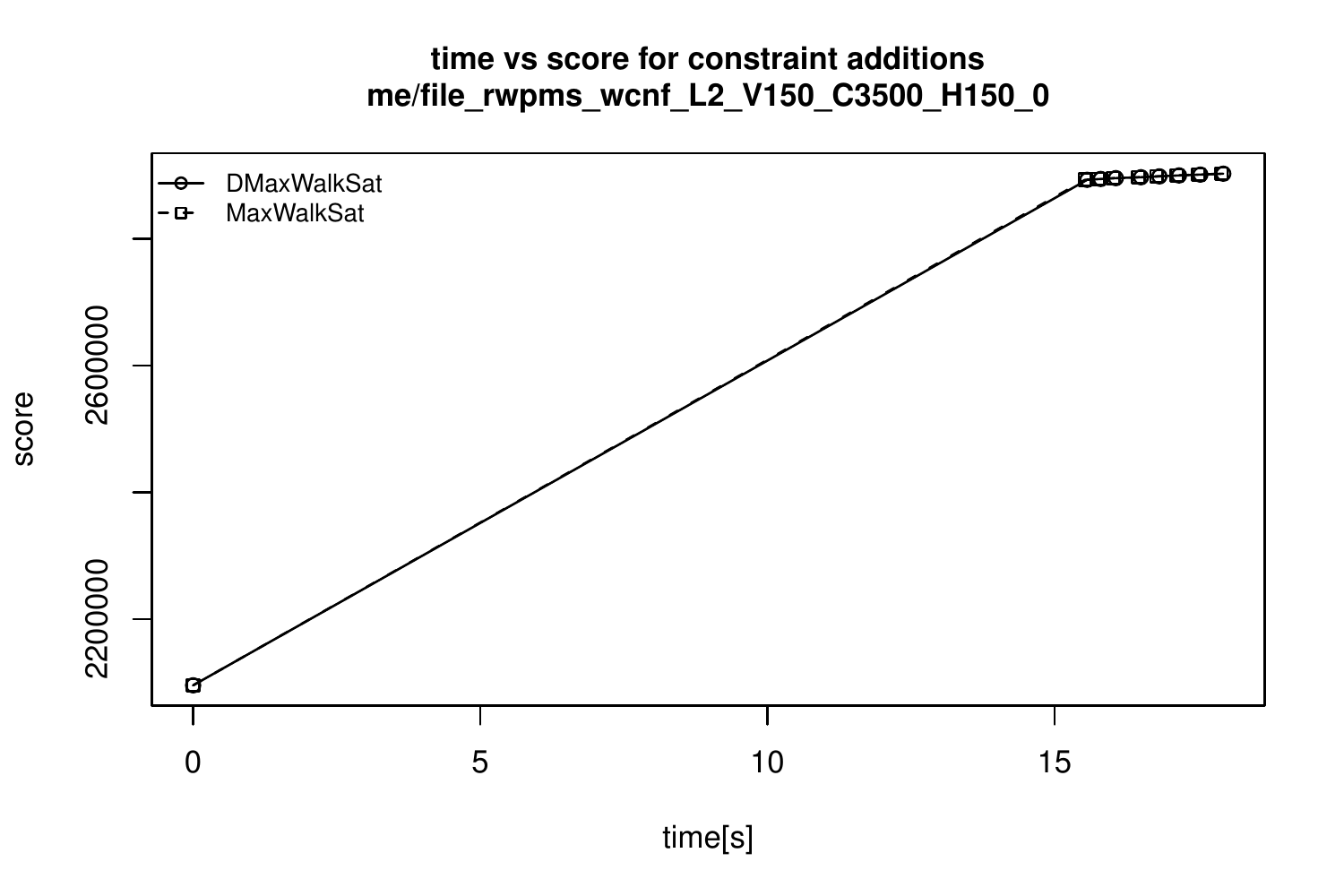}
        }
        \qquad
        \subfloat[Constraint removal]
        {
            \includegraphics[width=2.7in]{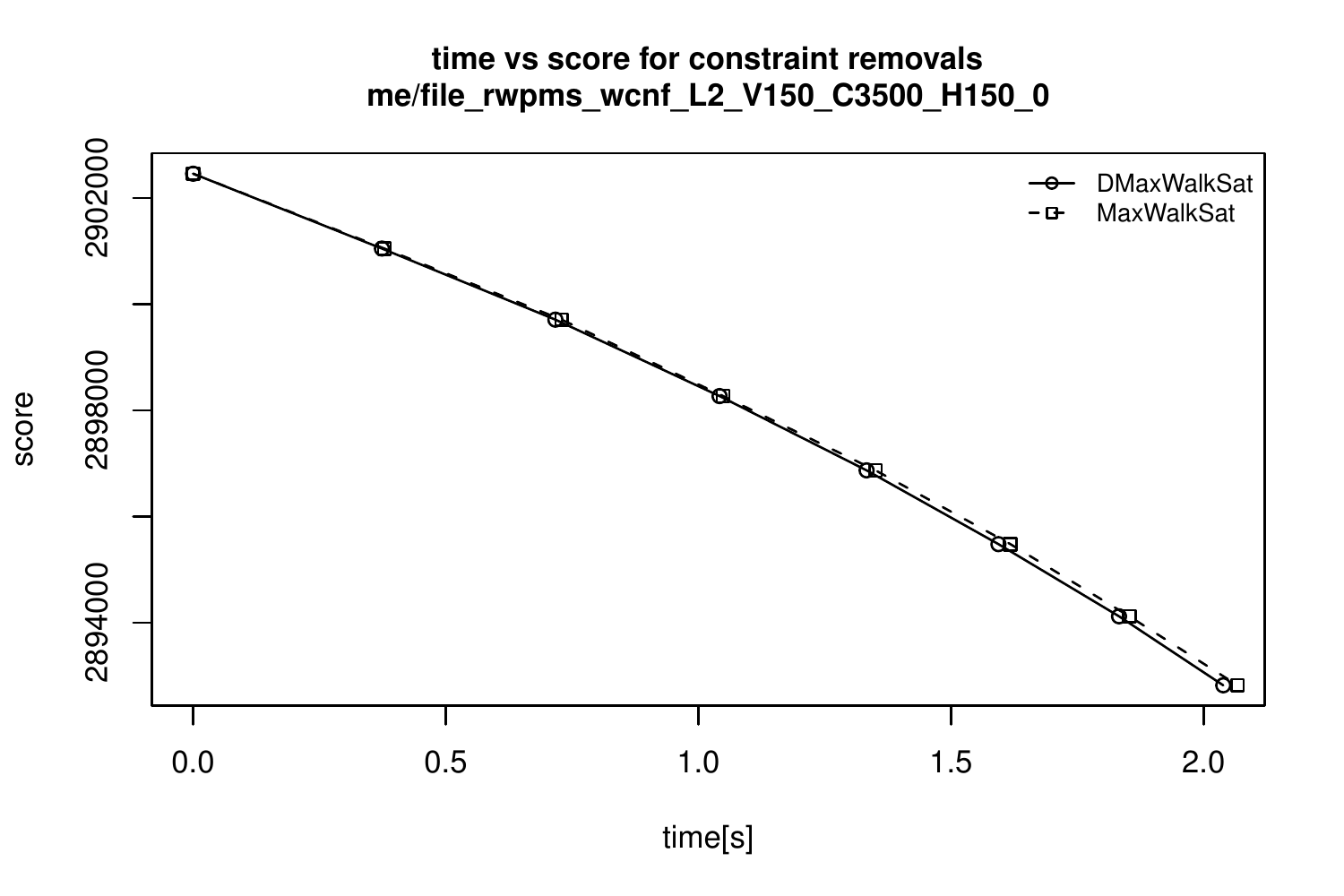}
        }

    \caption*{me/file\_rwpms\_wcnf\_L2\_V150\_C3500\_H150\_0}
    \label{fig_me/file_rwpms_wcnf_L2_V150_C3500_H150_0}
\end{figure}

\begin{figure}[H]
    \setcounter{subfigure}{0}
    \centering
        \subfloat[Constraint addition]
        {
            \includegraphics[width=2.7in]{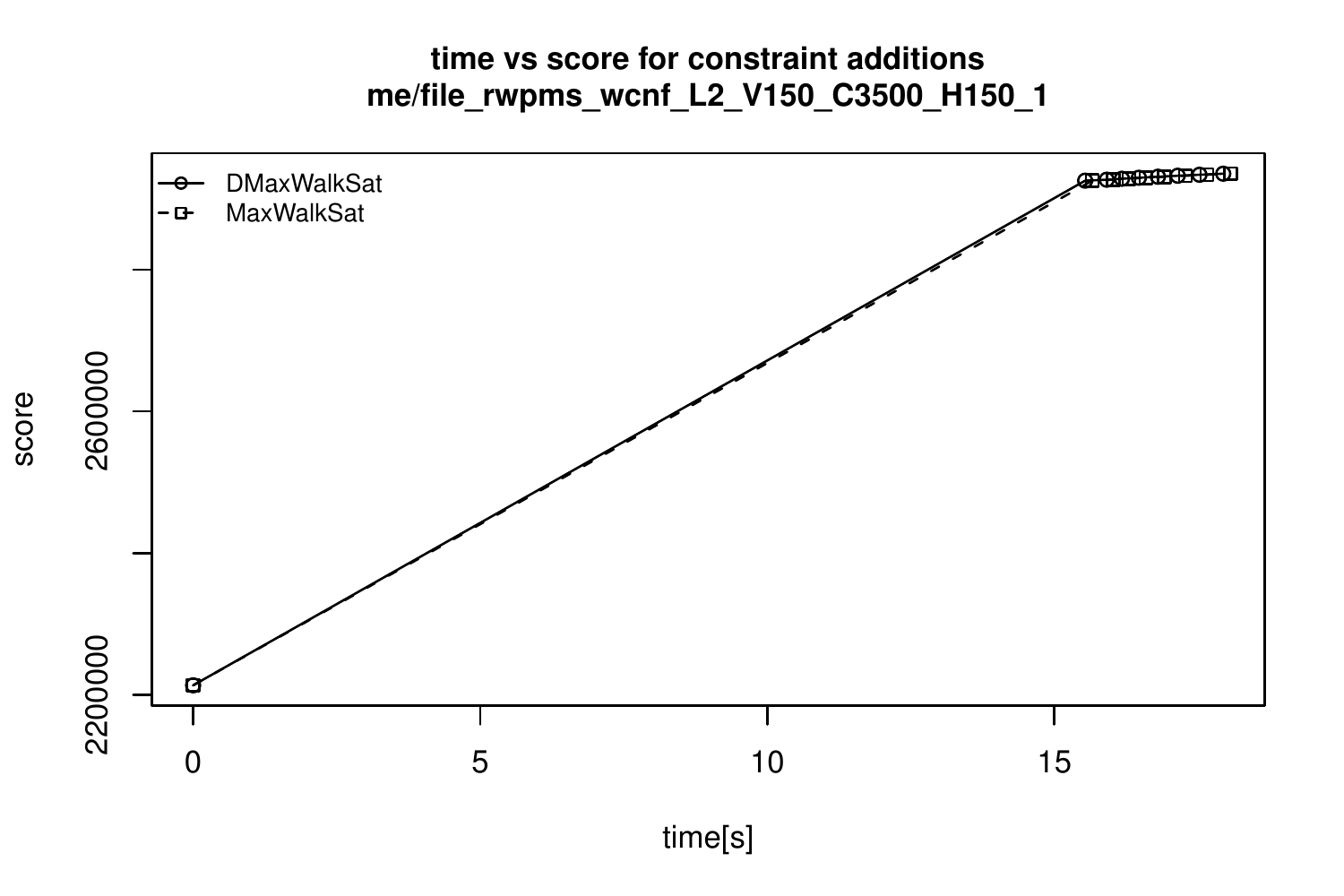}
        }
        \qquad
        \subfloat[Constraint removal]
        {
            \includegraphics[width=2.7in]{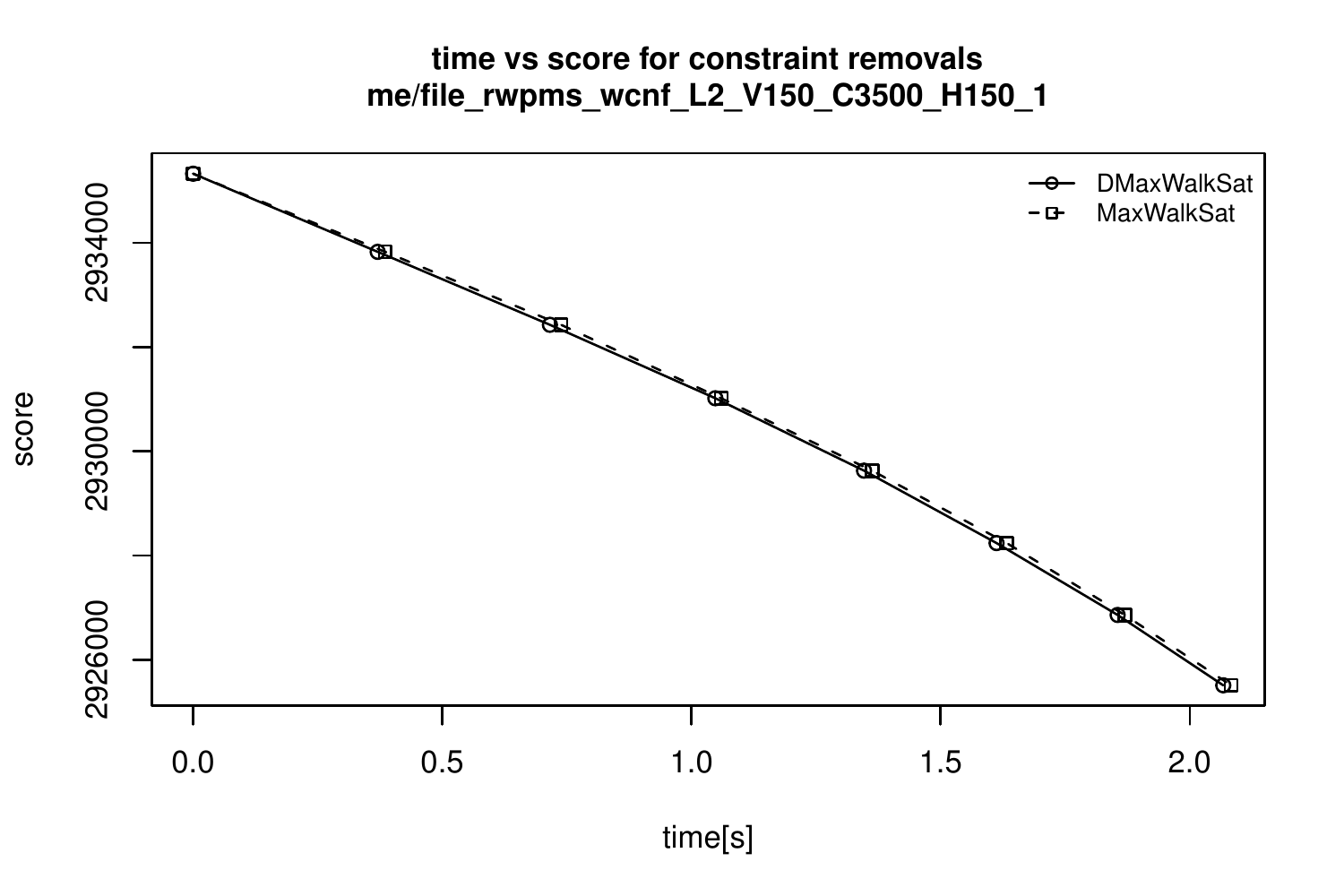}
        }

    \caption*{me/file\_rwpms\_wcnf\_L2\_V150\_C3500\_H150\_1}
    \label{fig_me/file_rwpms_wcnf_L2_V150_C3500_H150_1}
\end{figure}

\begin{figure}[H]
    \setcounter{subfigure}{0}
    \centering
        \subfloat[Constraint addition]
        {
            \includegraphics[width=2.7in]{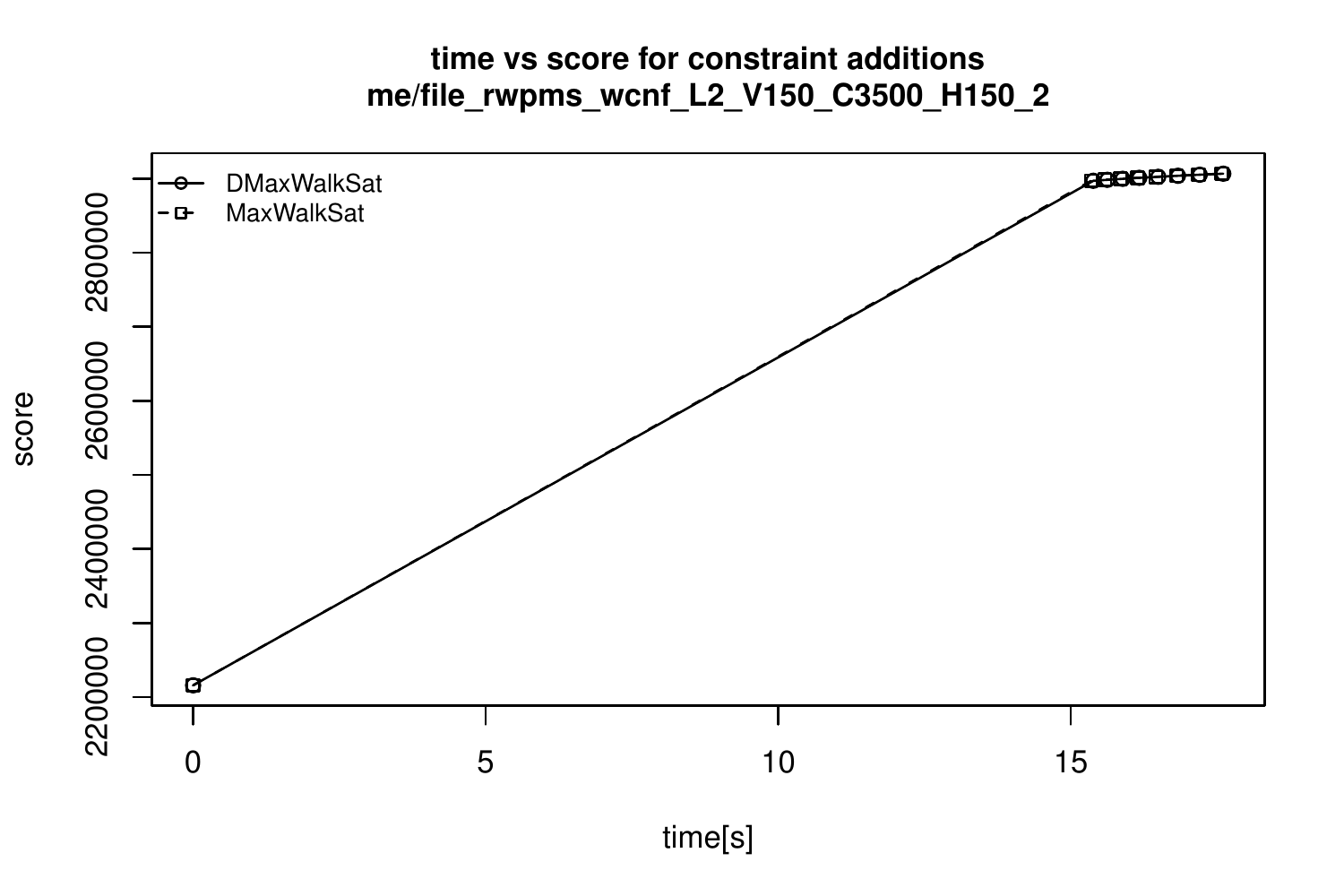}
        }
        \qquad
        \subfloat[Constraint removal]
        {
            \includegraphics[width=2.7in]{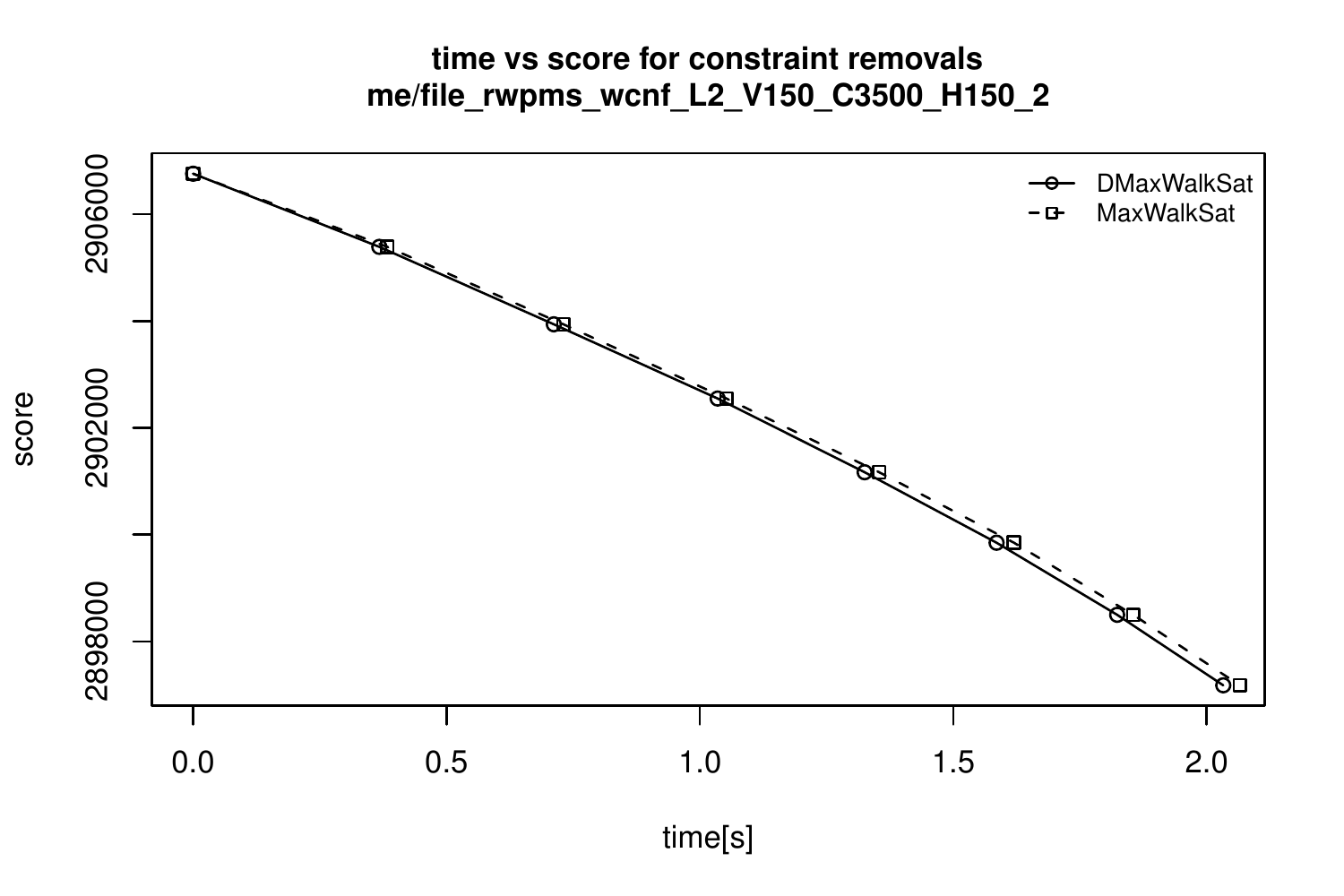}
        }

    \caption*{me/file\_rwpms\_wcnf\_L2\_V150\_C3500\_H150\_2}
    \label{fig_me/file_rwpms_wcnf_L2_V150_C3500_H150_2}
\end{figure}

\begin{figure}[H]
    \setcounter{subfigure}{0}
    \centering
        \subfloat[Constraint addition]
        {
            \includegraphics[width=2.7in]{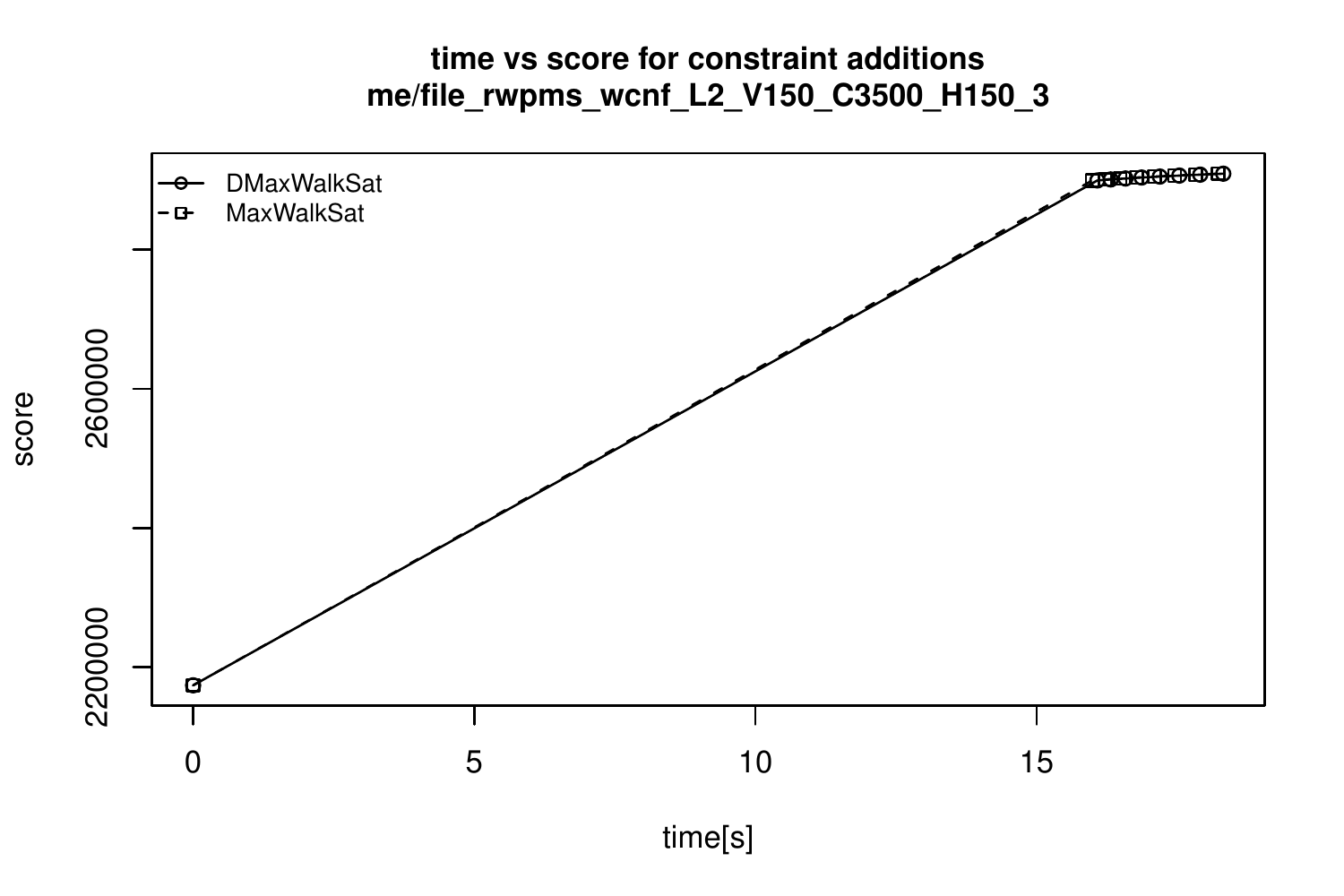}
        }
        \qquad
        \subfloat[Constraint removal]
        {
            \includegraphics[width=2.7in]{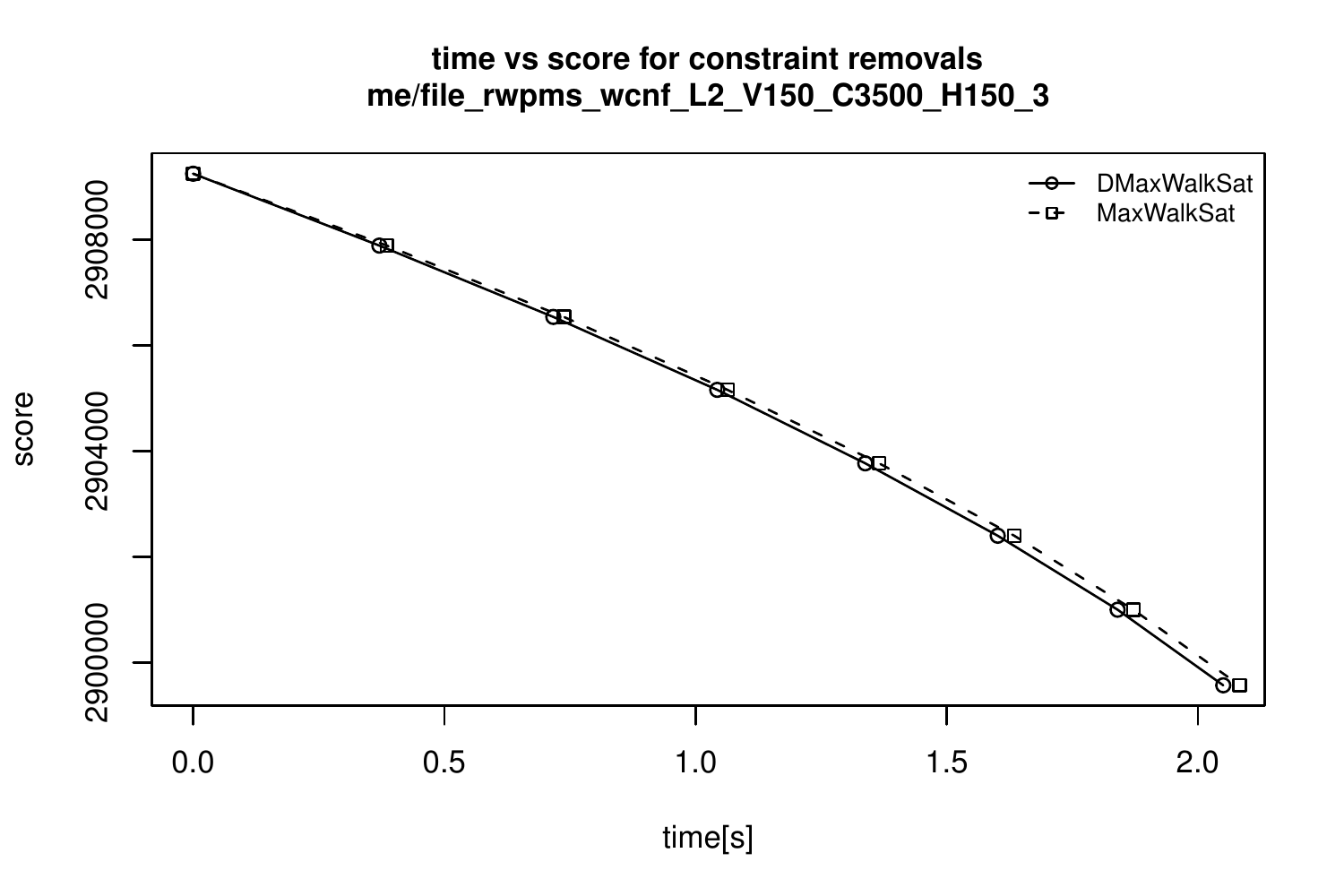}
        }

    \caption*{me/file\_rwpms\_wcnf\_L2\_V150\_C3500\_H150\_3}
    \label{fig_me/file_rwpms_wcnf_L2_V150_C3500_H150_3}
\end{figure}

\begin{figure}[H]
    \setcounter{subfigure}{0}
    \centering
        \subfloat[Constraint addition]
        {
            \includegraphics[width=2.7in]{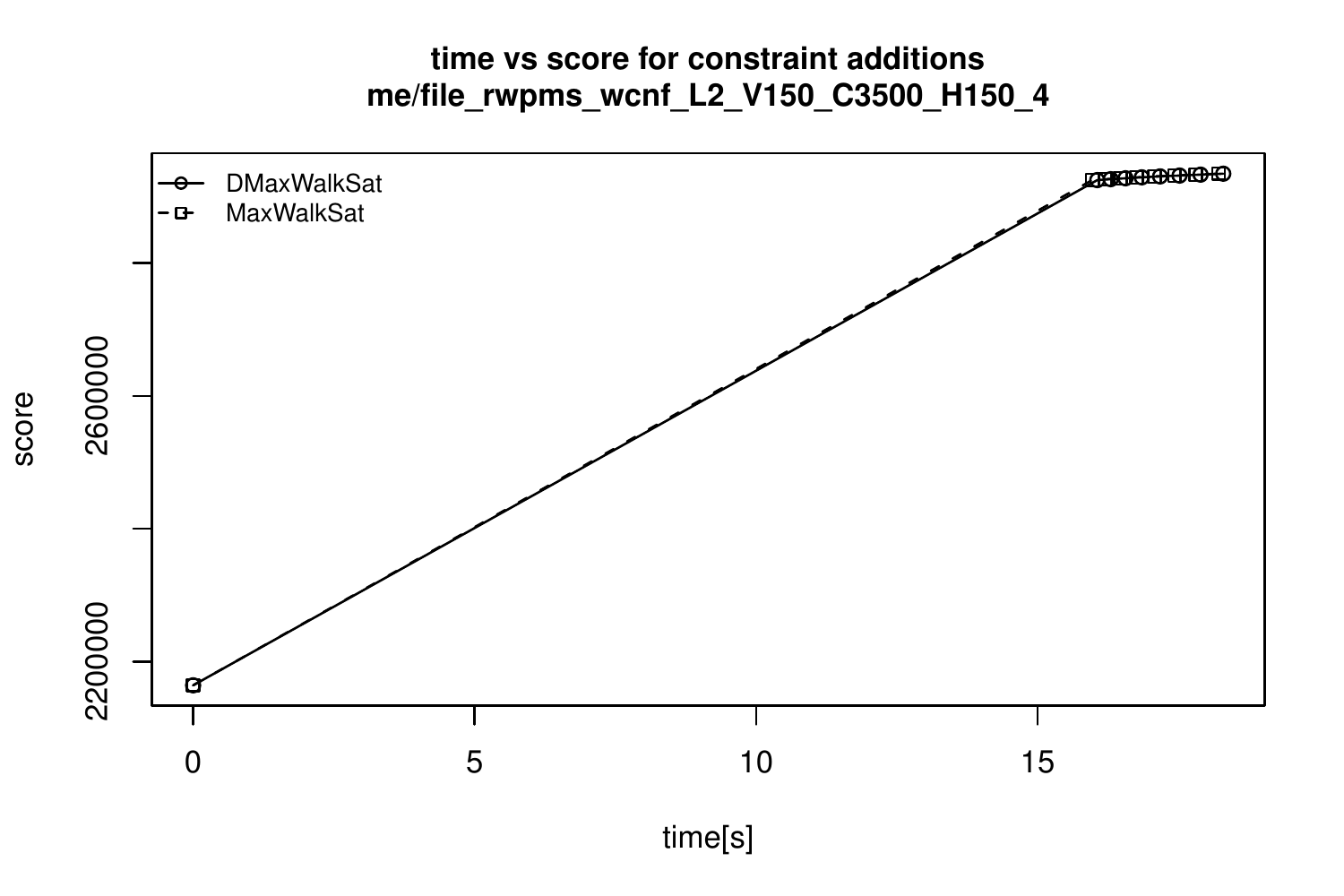}
        }
        \qquad
        \subfloat[Constraint removal]
        {
            \includegraphics[width=2.7in]{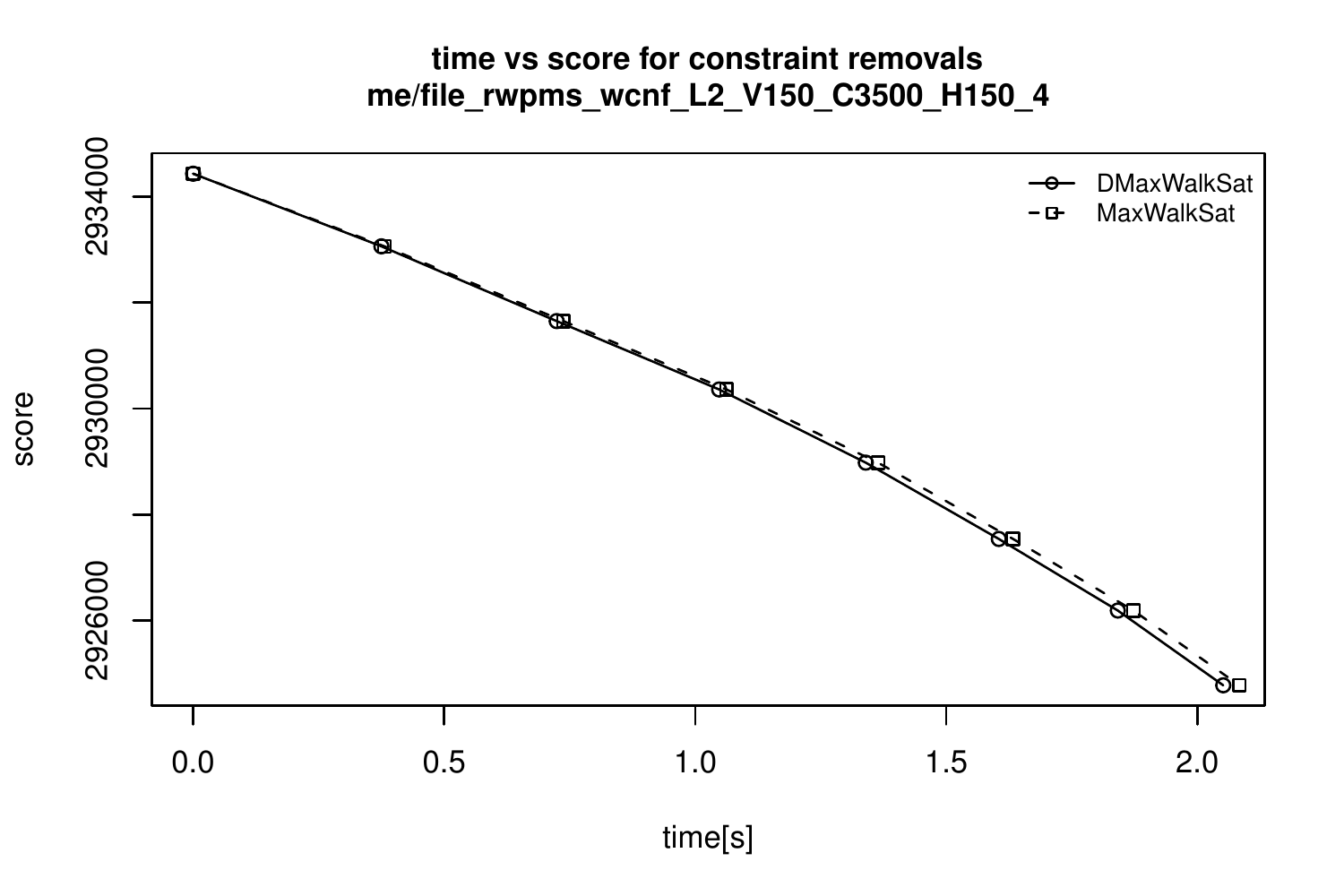}
        }

    \caption*{me/file\_rwpms\_wcnf\_L2\_V150\_C3500\_H150\_4}
    \label{fig_me/file_rwpms_wcnf_L2_V150_C3500_H150_4}
\end{figure}

\begin{figure}[H]
    \setcounter{subfigure}{0}
    \centering
        \subfloat[Constraint addition]
        {
            \includegraphics[width=2.7in]{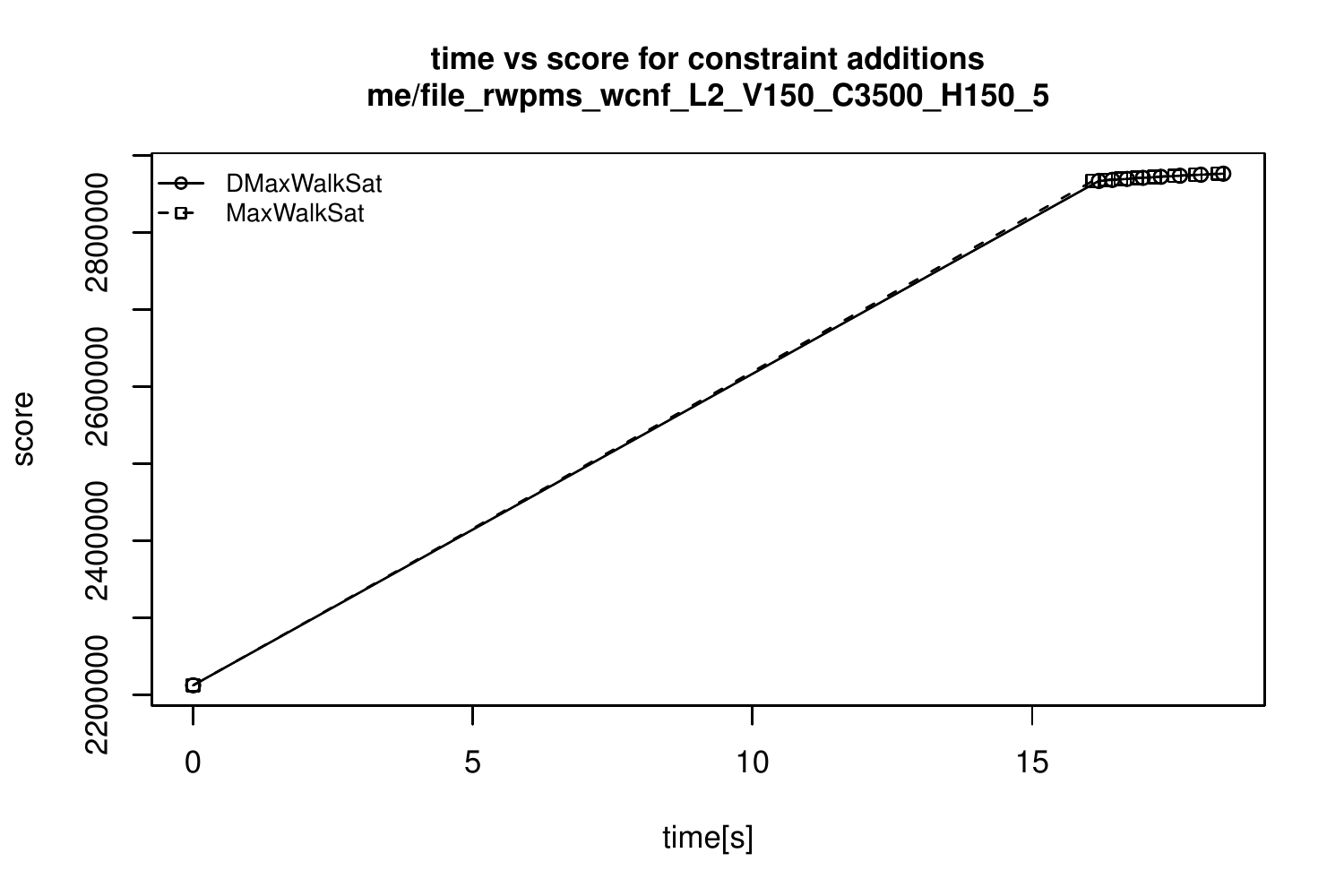}
        }
        \qquad
        \subfloat[Constraint removal]
        {
            \includegraphics[width=2.7in]{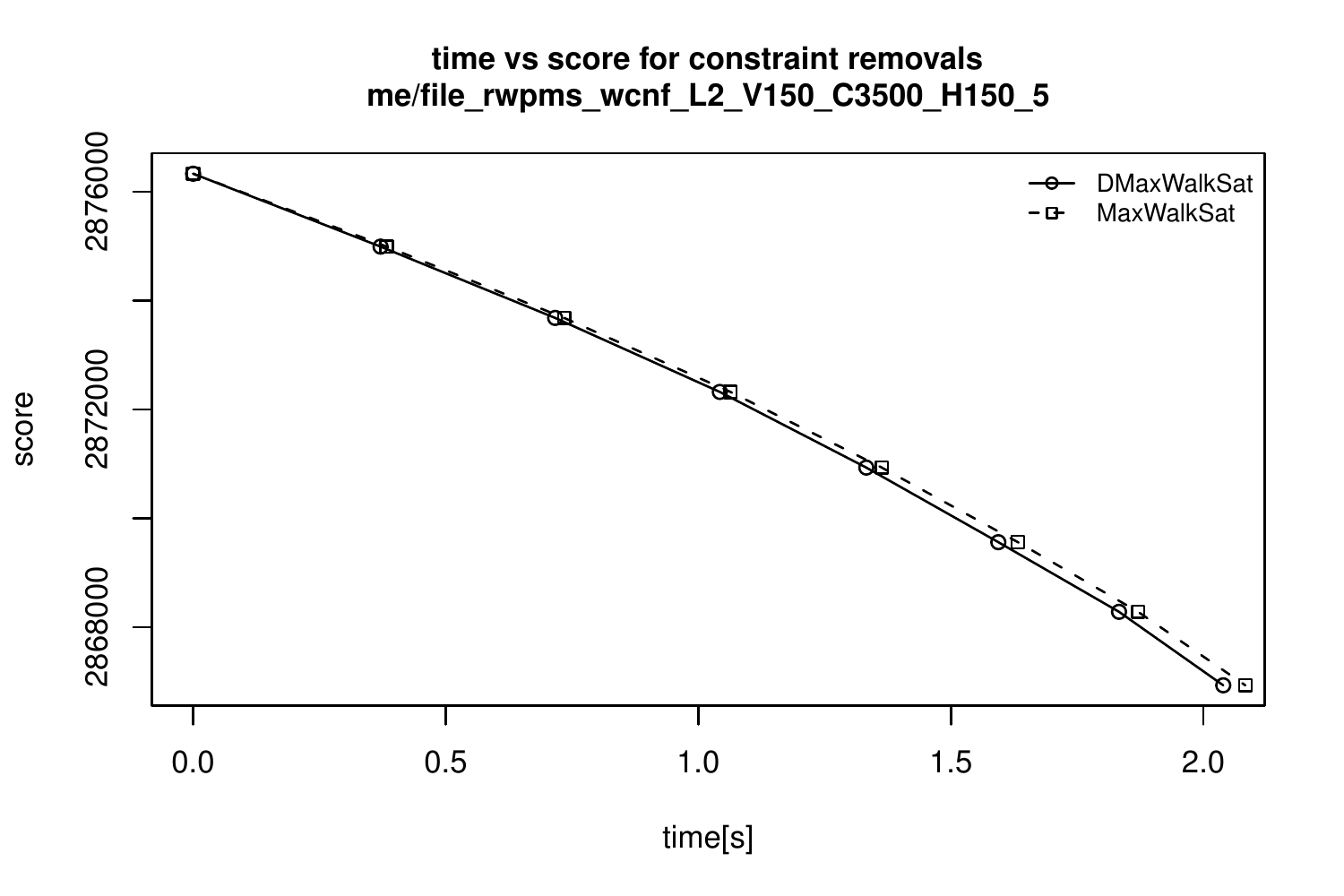}
        }

    \caption*{me/file\_rwpms\_wcnf\_L2\_V150\_C3500\_H150\_5}
    \label{fig_me/file_rwpms_wcnf_L2_V150_C3500_H150_5}
\end{figure}

\begin{figure}[H]
    \setcounter{subfigure}{0}
    \centering
        \subfloat[Constraint addition]
        {
            \includegraphics[width=2.7in]{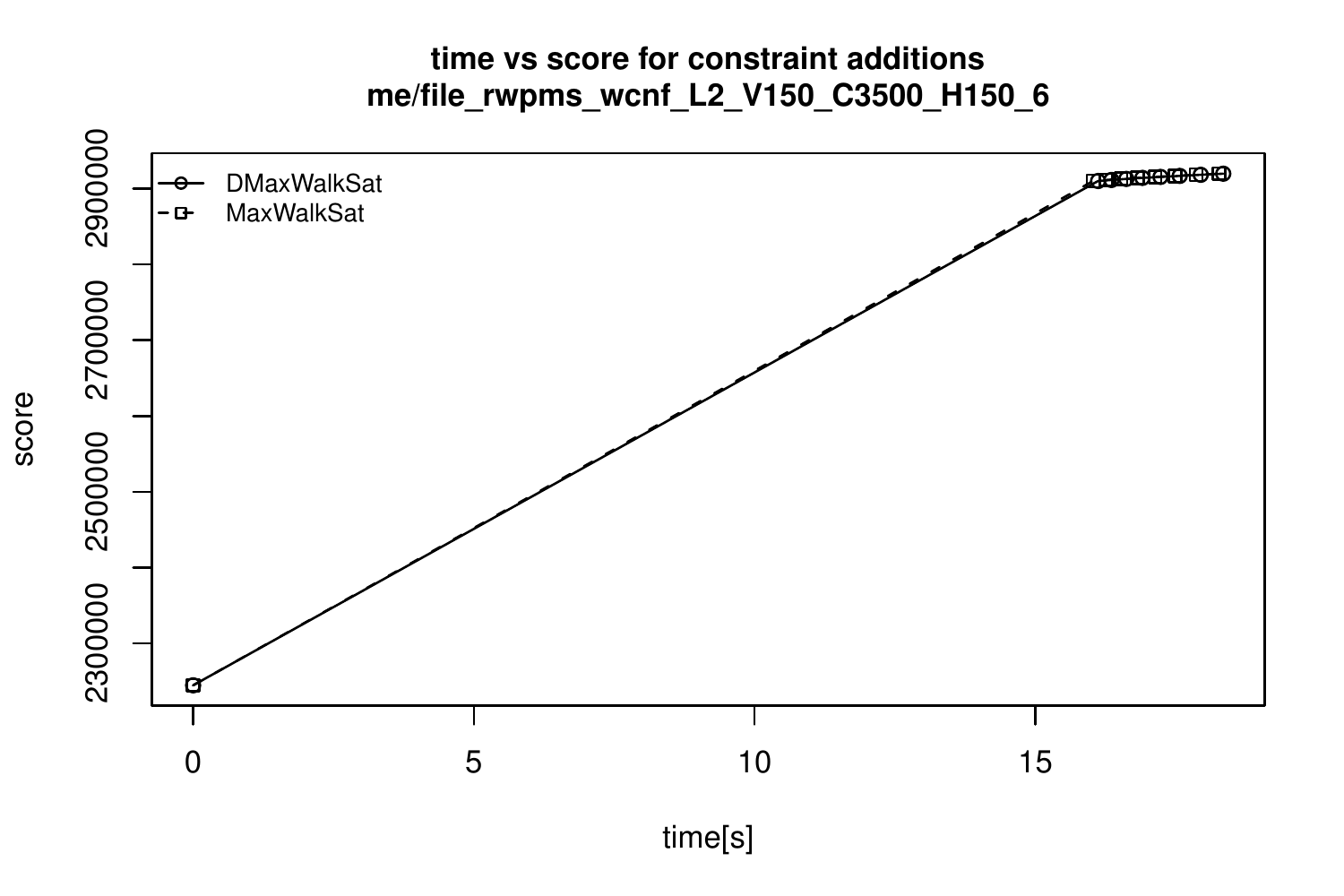}
        }
        \qquad
        \subfloat[Constraint removal]
        {
            \includegraphics[width=2.7in]{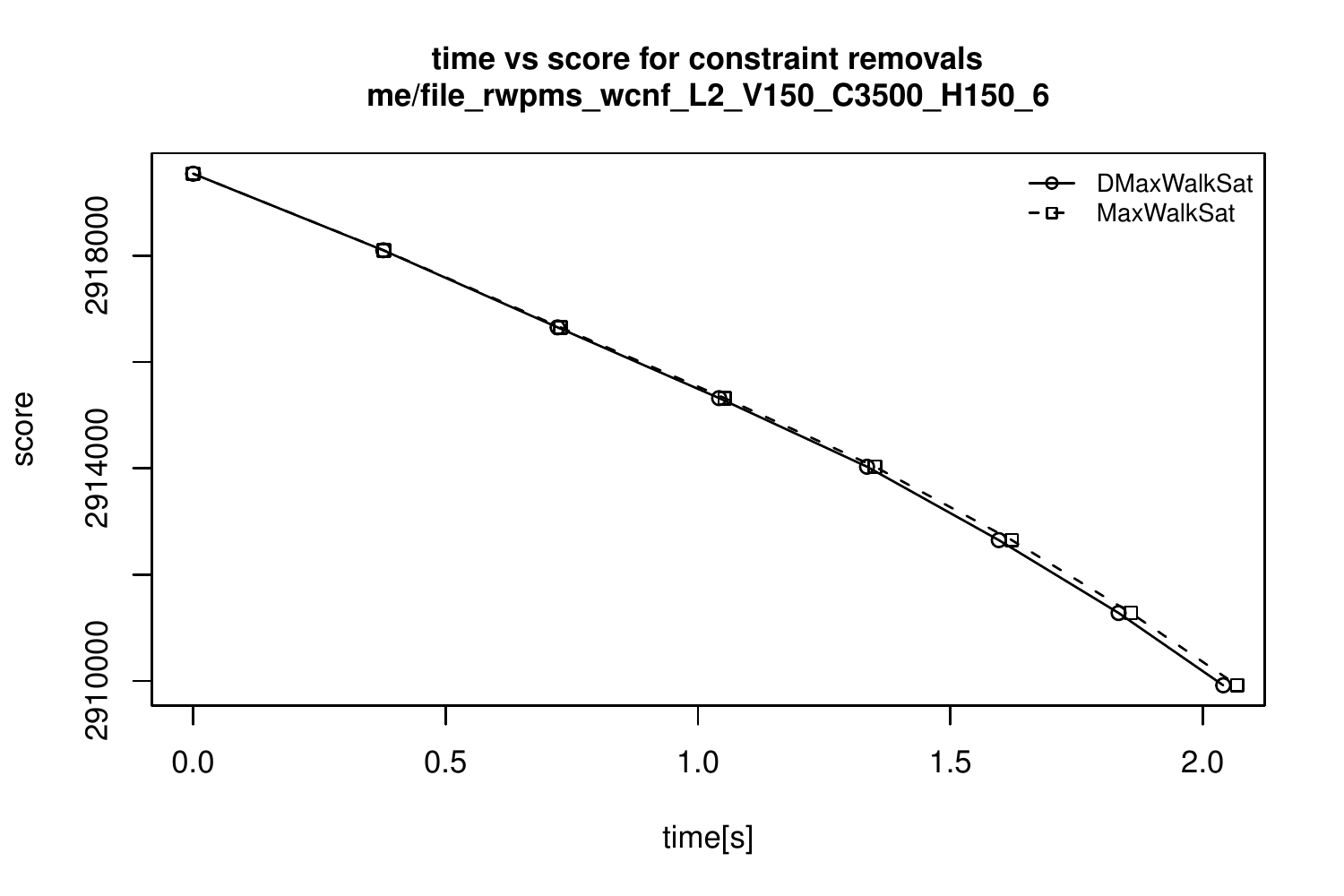}
        }

    \caption*{me/file\_rwpms\_wcnf\_L2\_V150\_C3500\_H150\_6}
    \label{fig_me/file_rwpms_wcnf_L2_V150_C3500_H150_6}
\end{figure}

\begin{figure}[H]
    \setcounter{subfigure}{0}
    \centering
        \subfloat[Constraint addition]
        {
            \includegraphics[width=2.7in]{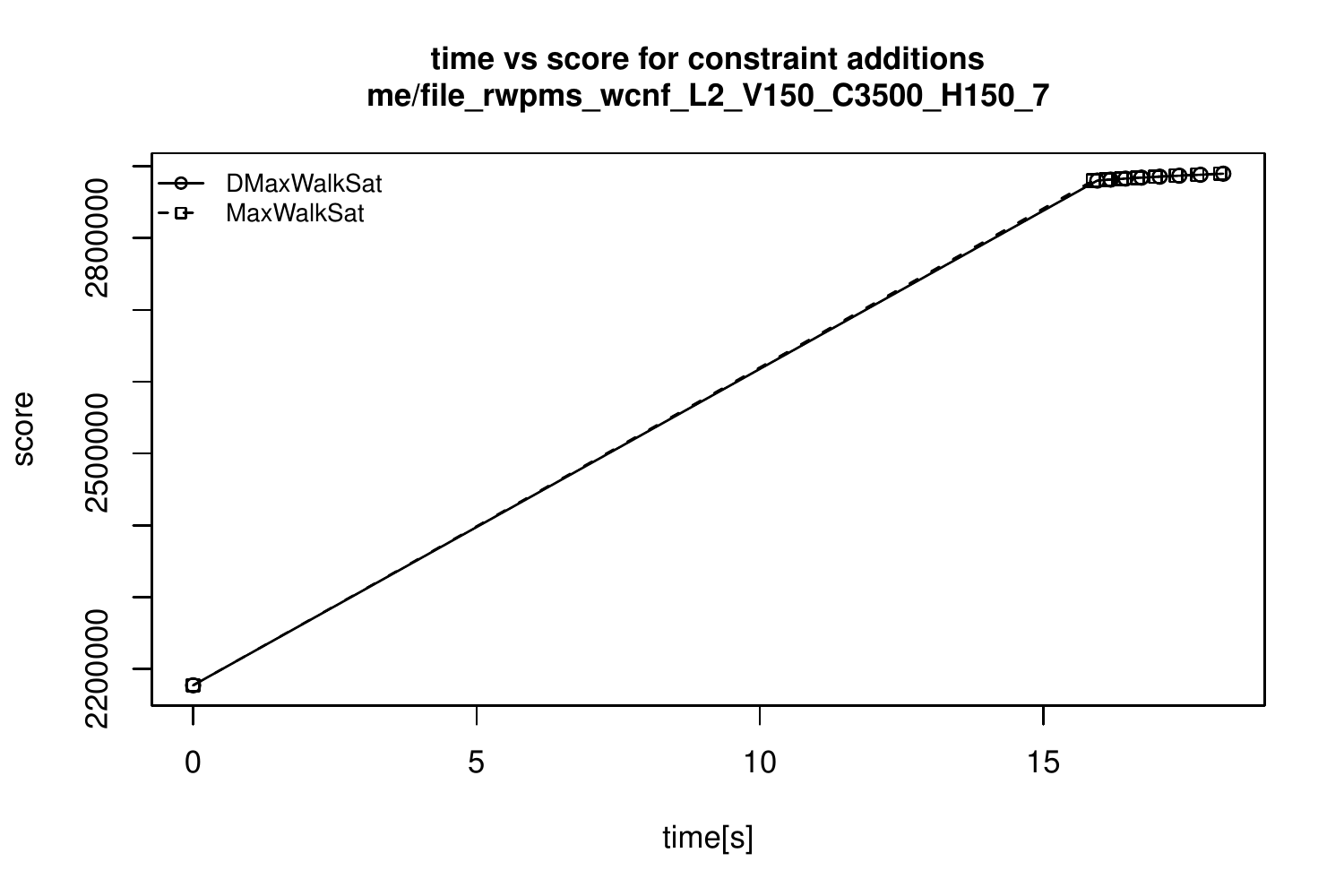}
        }
        \qquad
        \subfloat[Constraint removal]
        {
            \includegraphics[width=2.7in]{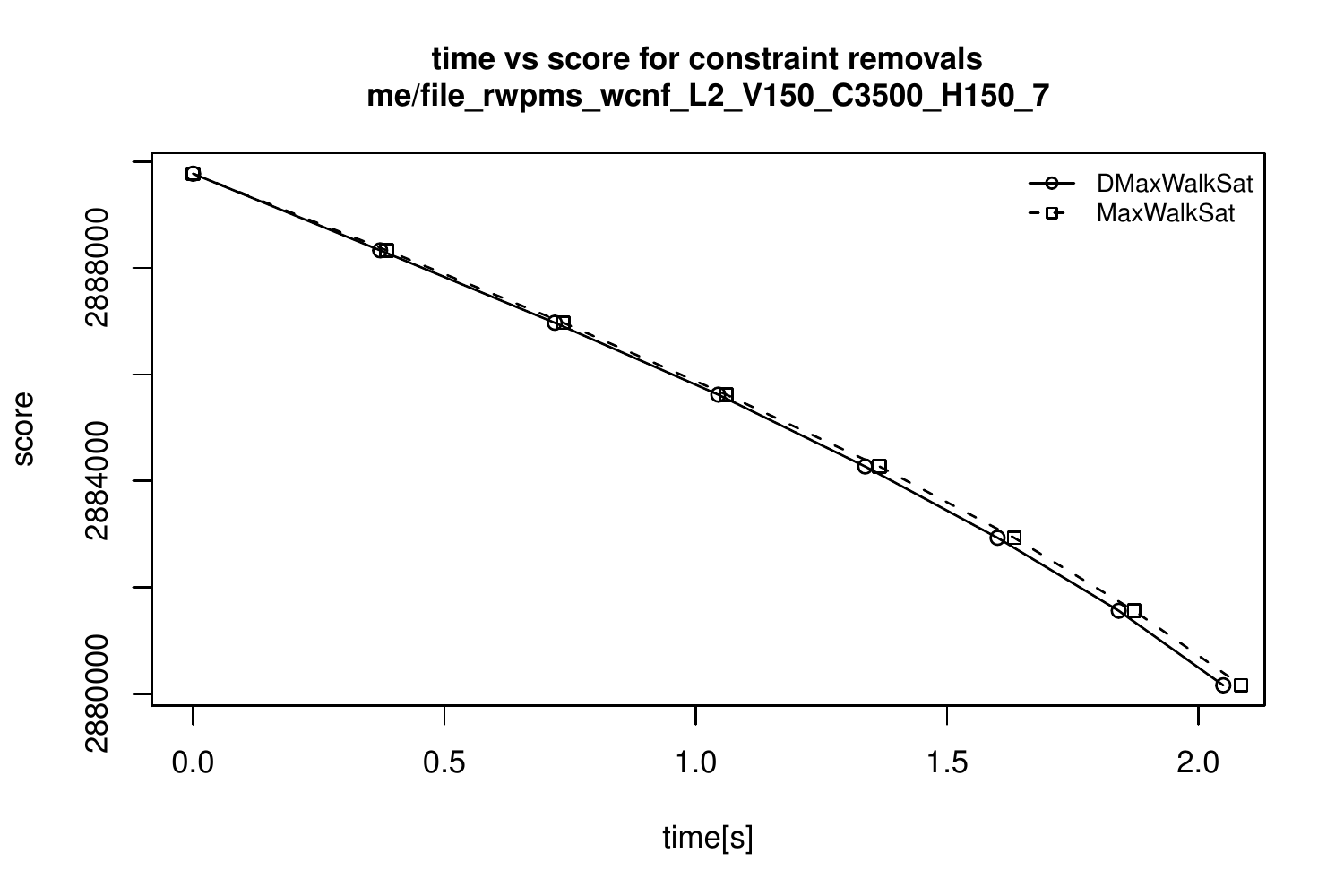}
        }

    \caption*{me/file\_rwpms\_wcnf\_L2\_V150\_C3500\_H150\_7}
    \label{fig_me/file_rwpms_wcnf_L2_V150_C3500_H150_7}
\end{figure}

\begin{figure}[H]
    \setcounter{subfigure}{0}
    \centering
        \subfloat[Constraint addition]
        {
            \includegraphics[width=2.7in]{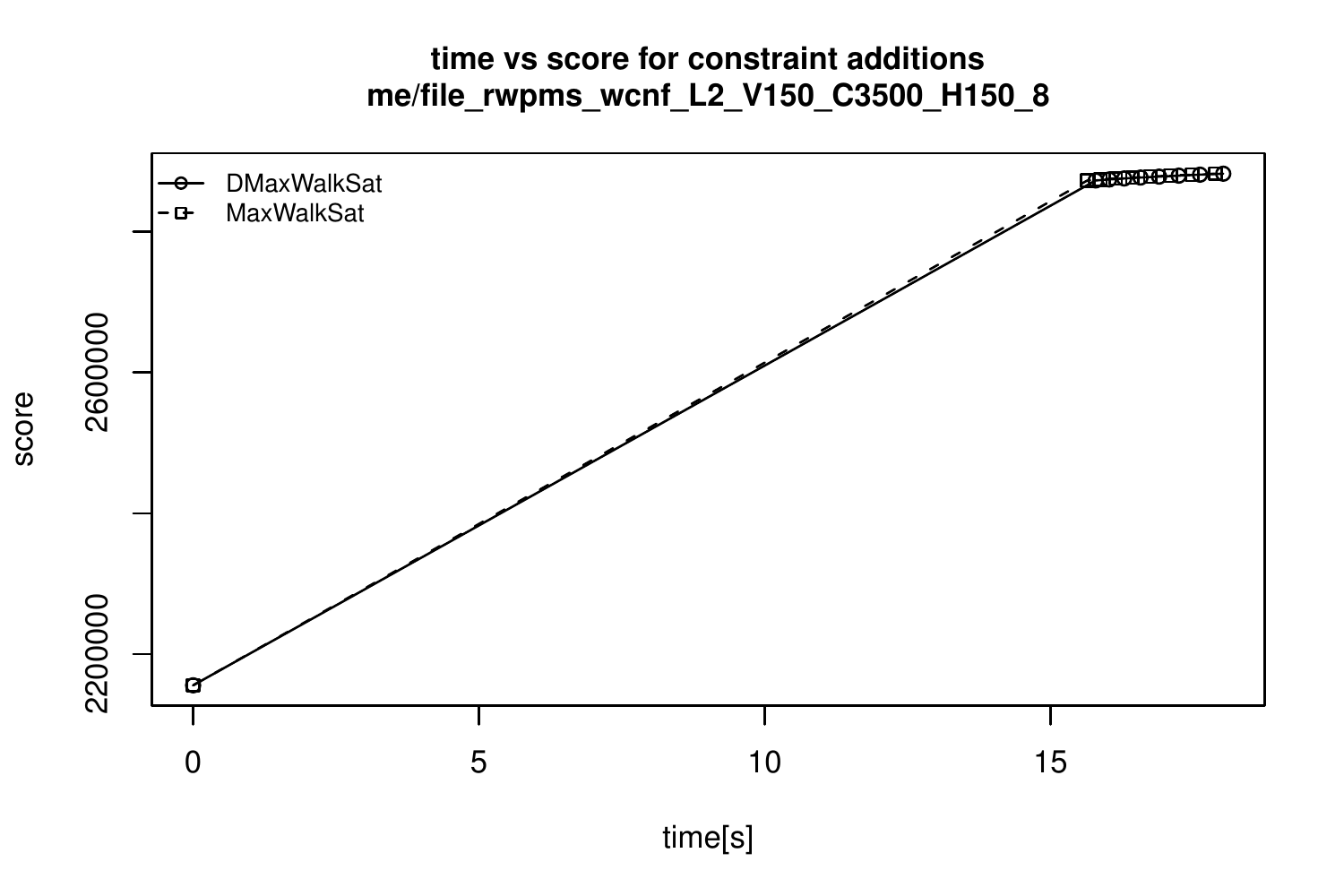}
        }
        \qquad
        \subfloat[Constraint removal]
        {
            \includegraphics[width=2.7in]{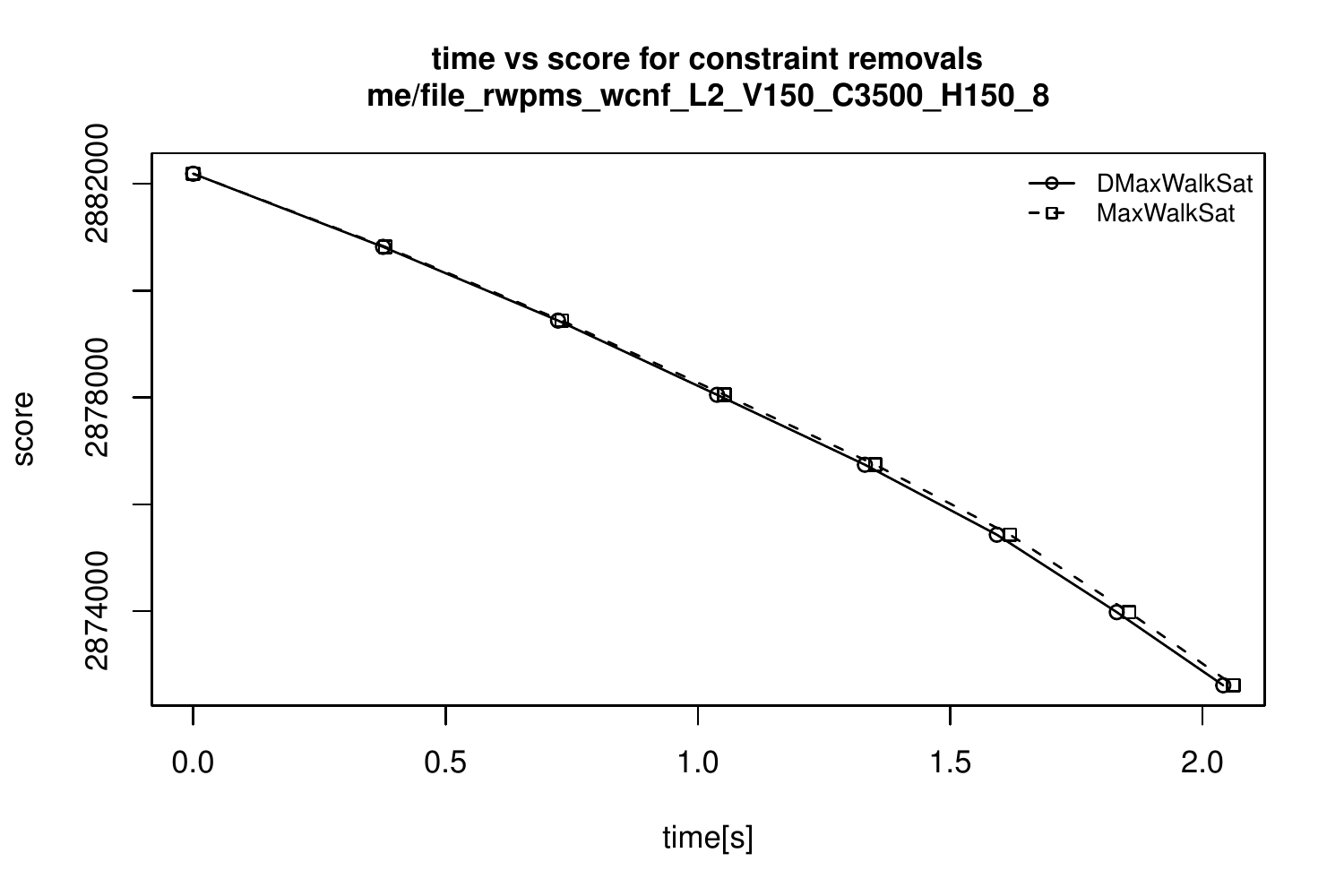}
        }

    \caption*{me/file\_rwpms\_wcnf\_L2\_V150\_C3500\_H150\_8}
    \label{fig_me/file_rwpms_wcnf_L2_V150_C3500_H150_8}
\end{figure}

\begin{figure}[H]
    \setcounter{subfigure}{0}
    \centering
        \subfloat[Constraint addition]
        {
            \includegraphics[width=2.7in]{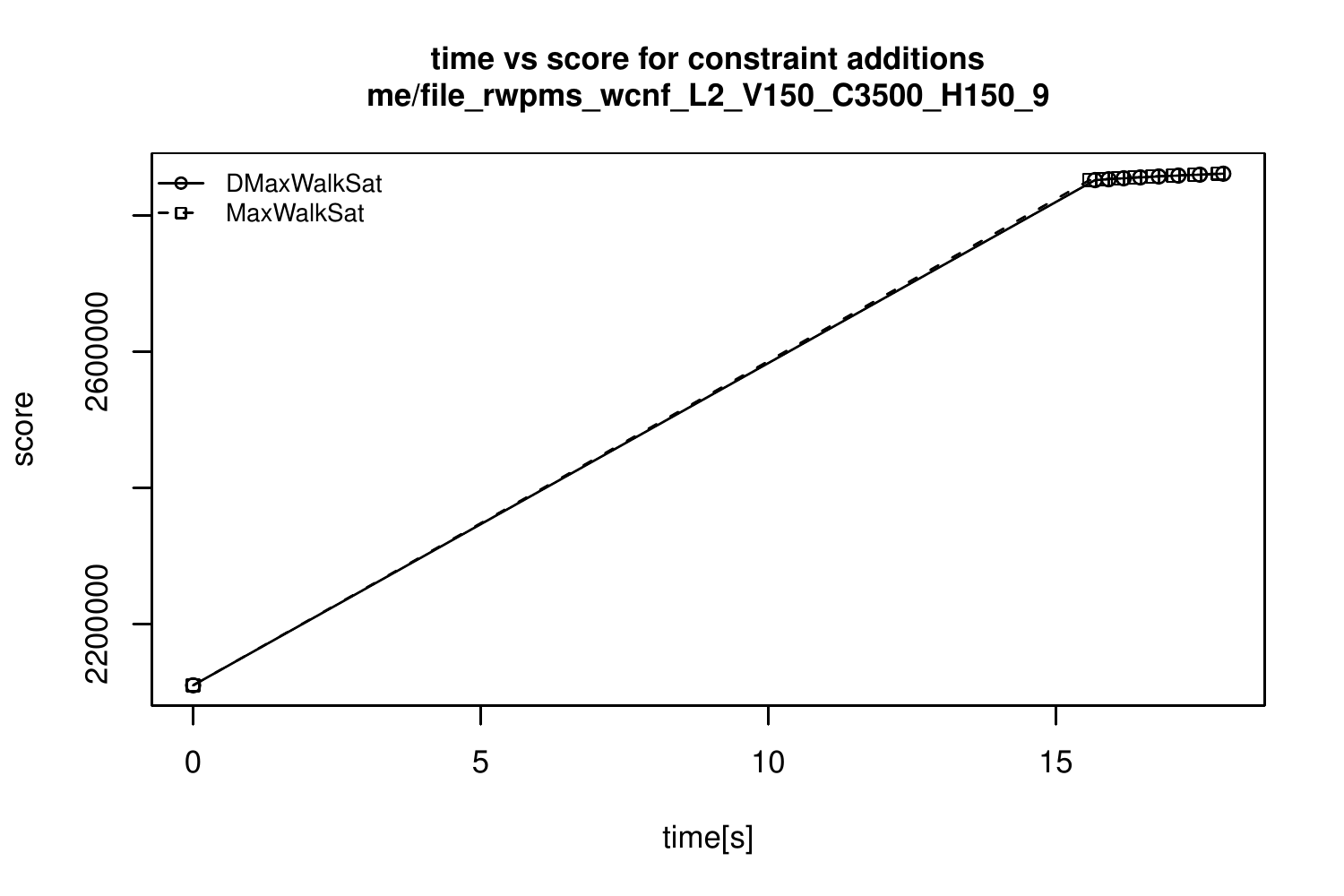}
        }
        \qquad
        \subfloat[Constraint removal]
        {
            \includegraphics[width=2.7in]{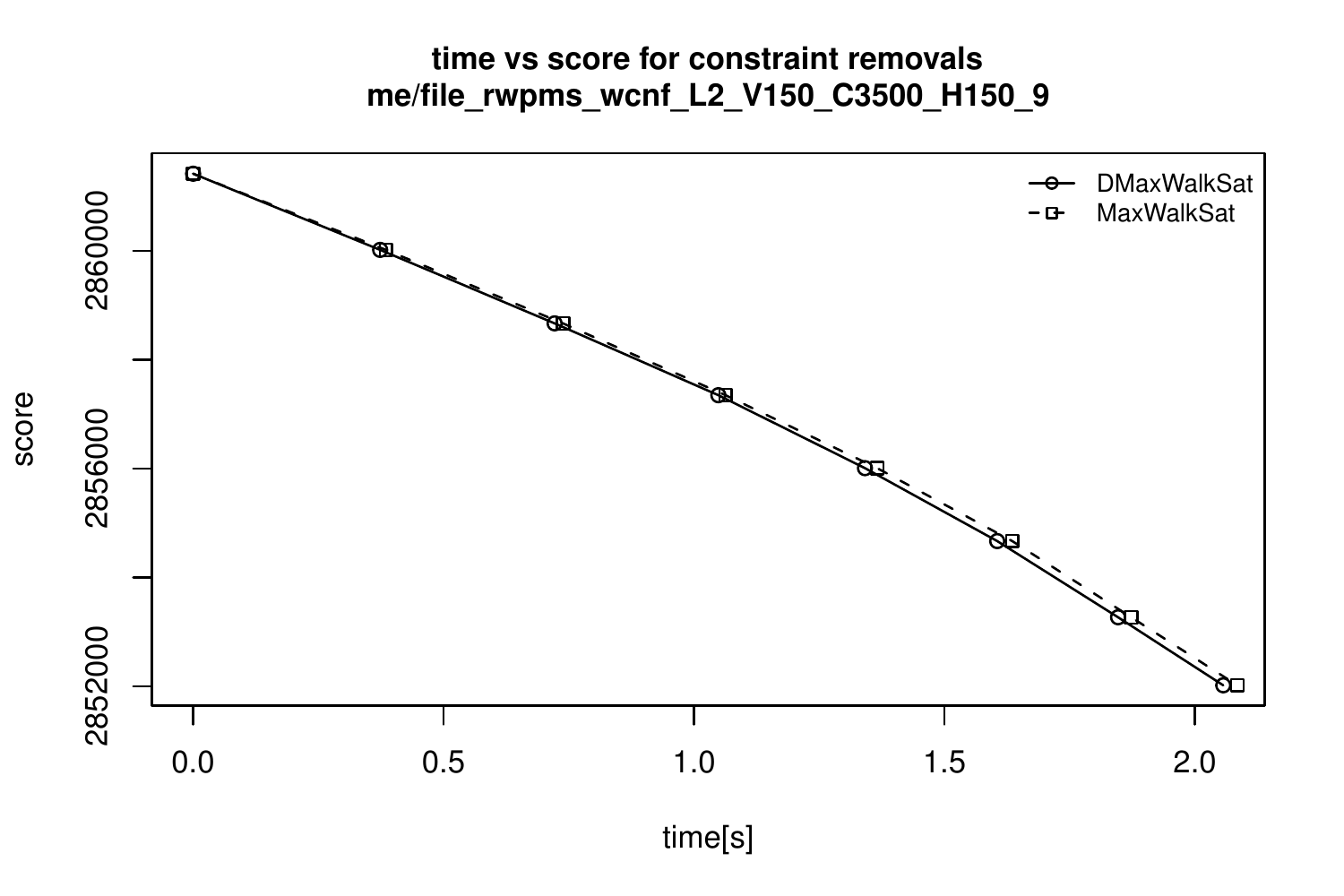}
        }

    \caption*{me/file\_rwpms\_wcnf\_L2\_V150\_C3500\_H150\_9}
    \label{fig_me/file_rwpms_wcnf_L2_V150_C3500_H150_9}
\end{figure}

\begin{figure}[H]
    \setcounter{subfigure}{0}
    \centering
        \subfloat[Constraint addition]
        {
            \includegraphics[width=2.7in]{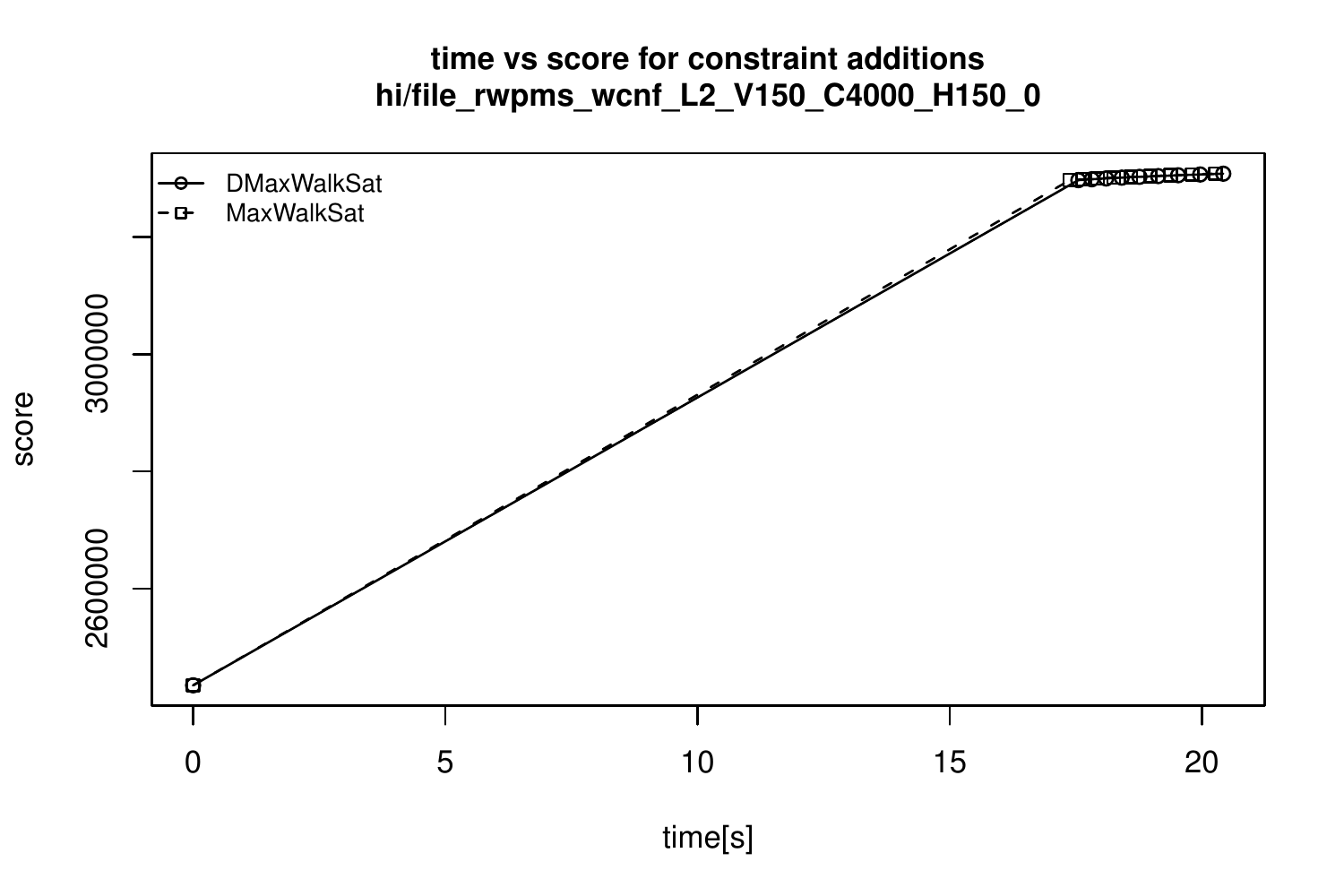}
        }
        \qquad
        \subfloat[Constraint removal]
        {
            \includegraphics[width=2.7in]{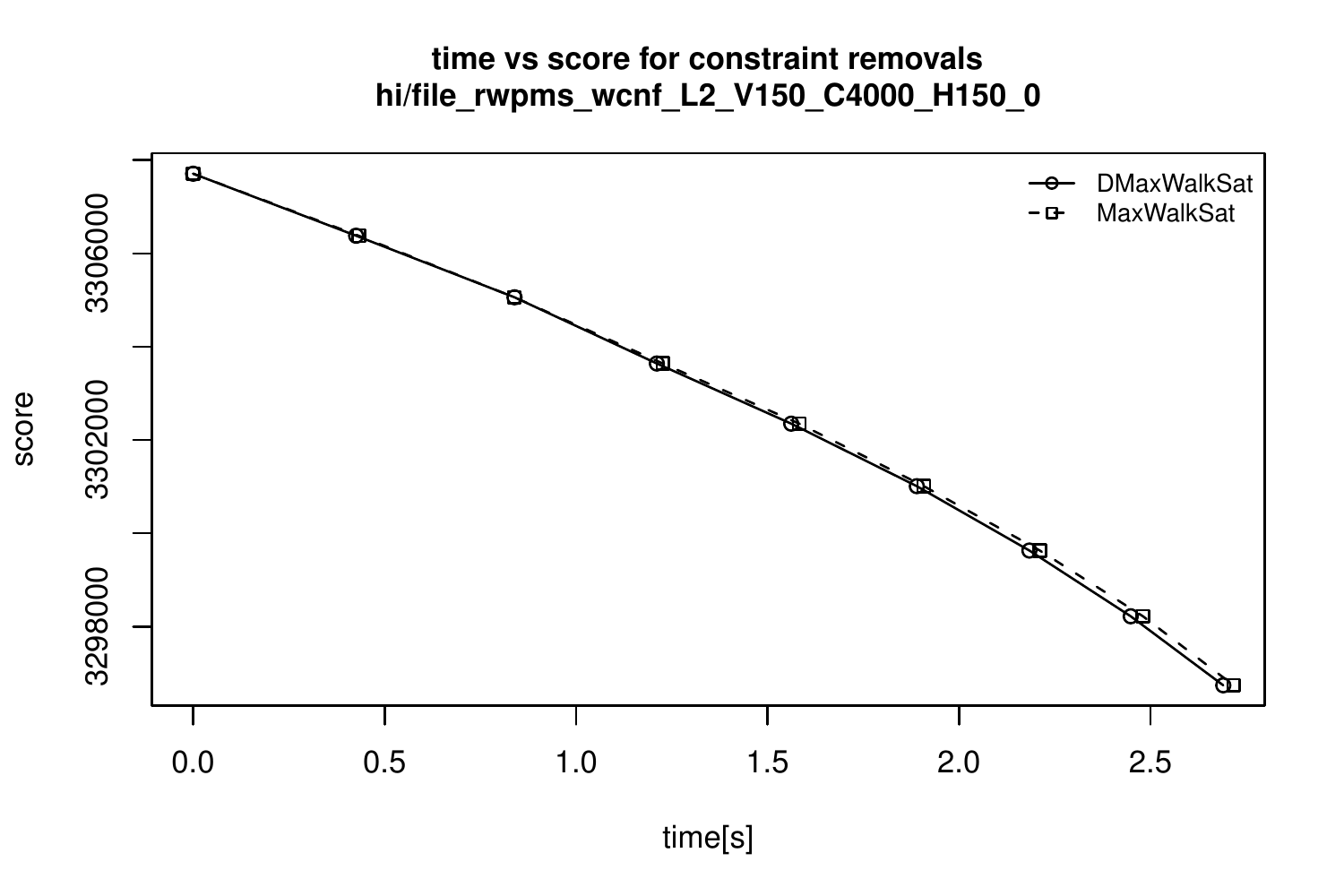}
        }

    \caption*{hi/file\_rwpms\_wcnf\_L2\_V150\_C4000\_H150\_0}
    \label{fig_hi/file_rwpms_wcnf_L2_V150_C4000_H150_0}
\end{figure}

\begin{figure}[H]
    \setcounter{subfigure}{0}
    \centering
        \subfloat[Constraint addition]
        {
            \includegraphics[width=2.7in]{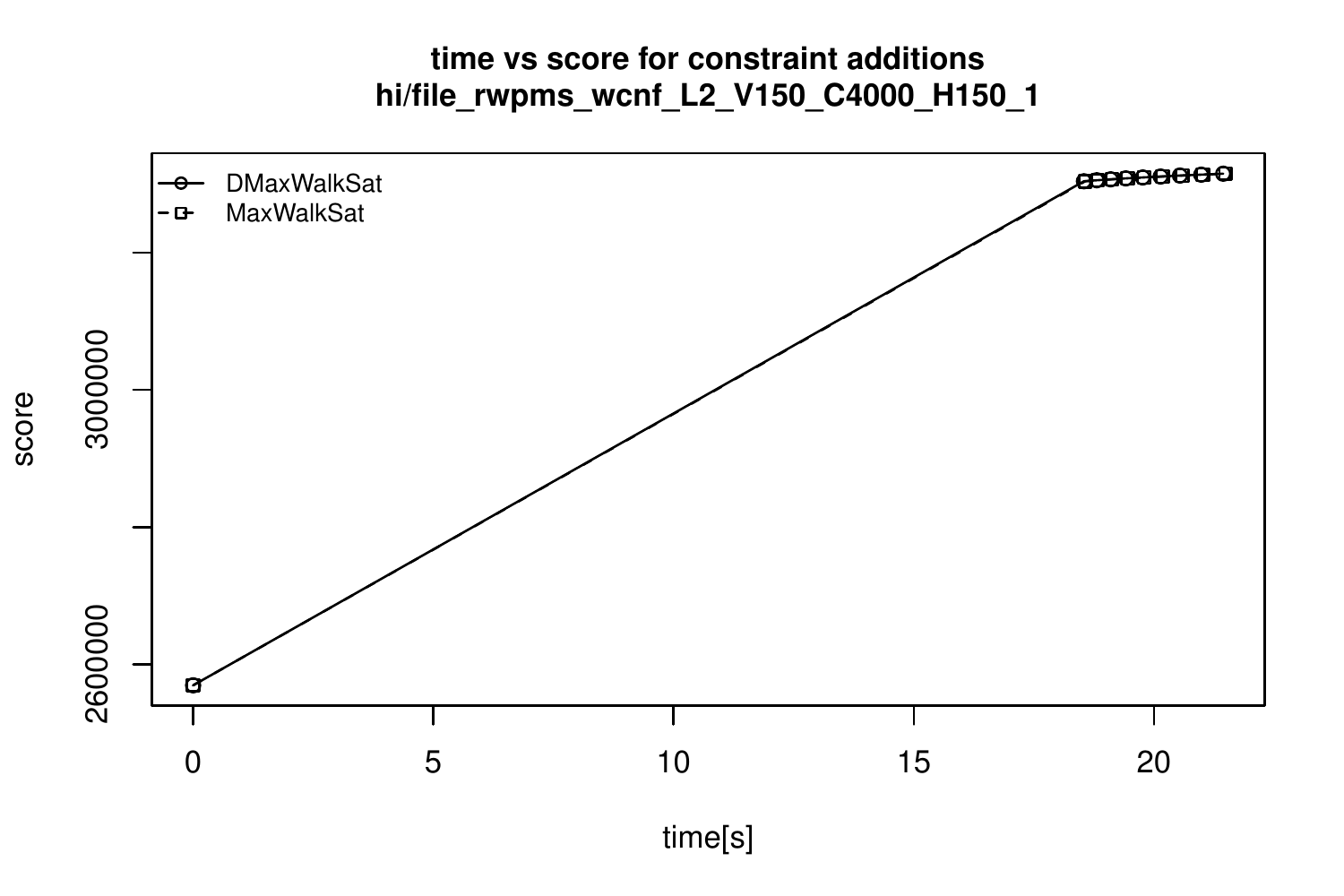}
        }
        \qquad
        \subfloat[Constraint removal]
        {
            \includegraphics[width=2.7in]{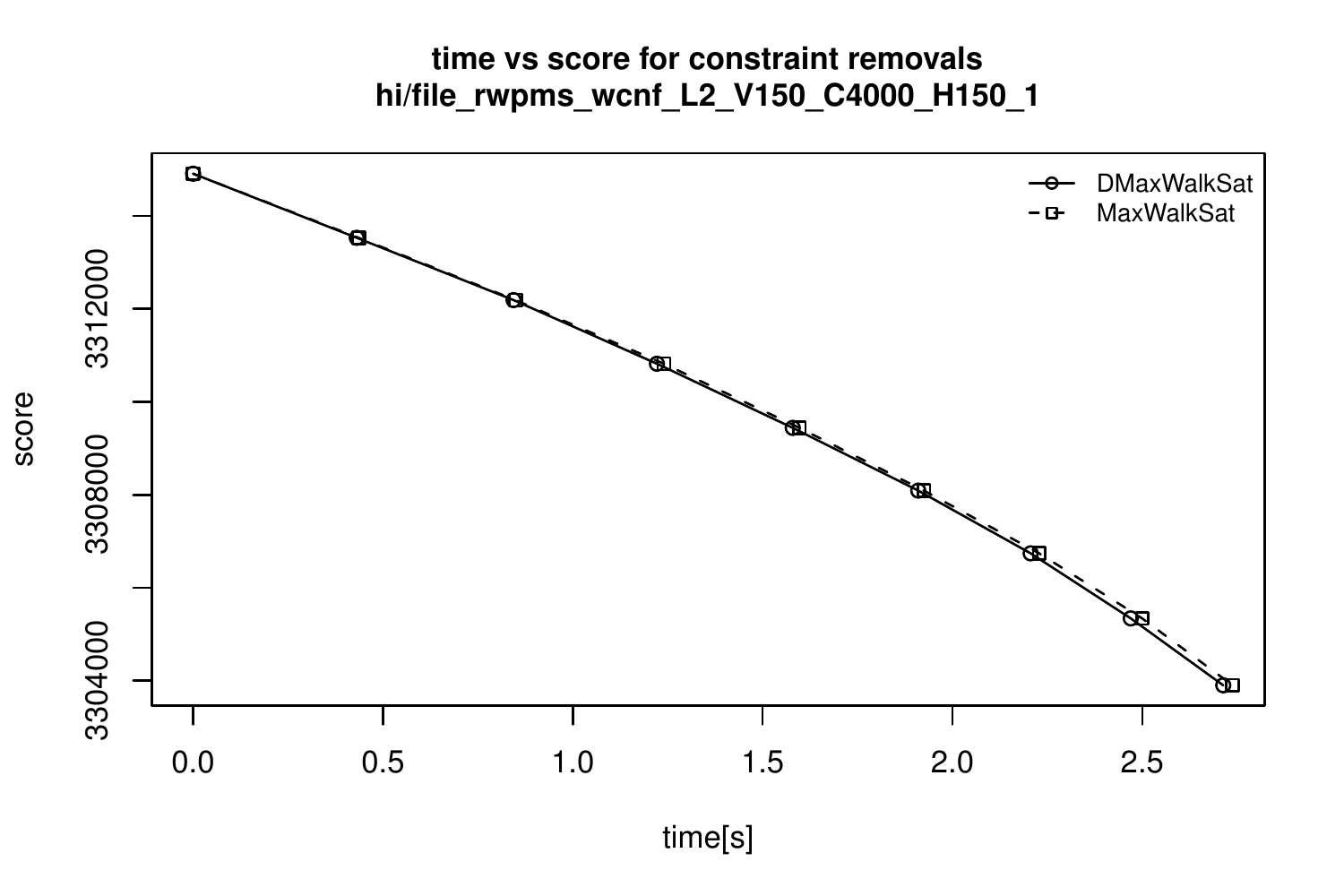}
        }

    \caption*{hi/file\_rwpms\_wcnf\_L2\_V150\_C4000\_H150\_1}
    \label{fig_hi/file_rwpms_wcnf_L2_V150_C4000_H150_1}
\end{figure}

\begin{figure}[H]
    \setcounter{subfigure}{0}
    \centering
        \subfloat[Constraint addition]
        {
            \includegraphics[width=2.7in]{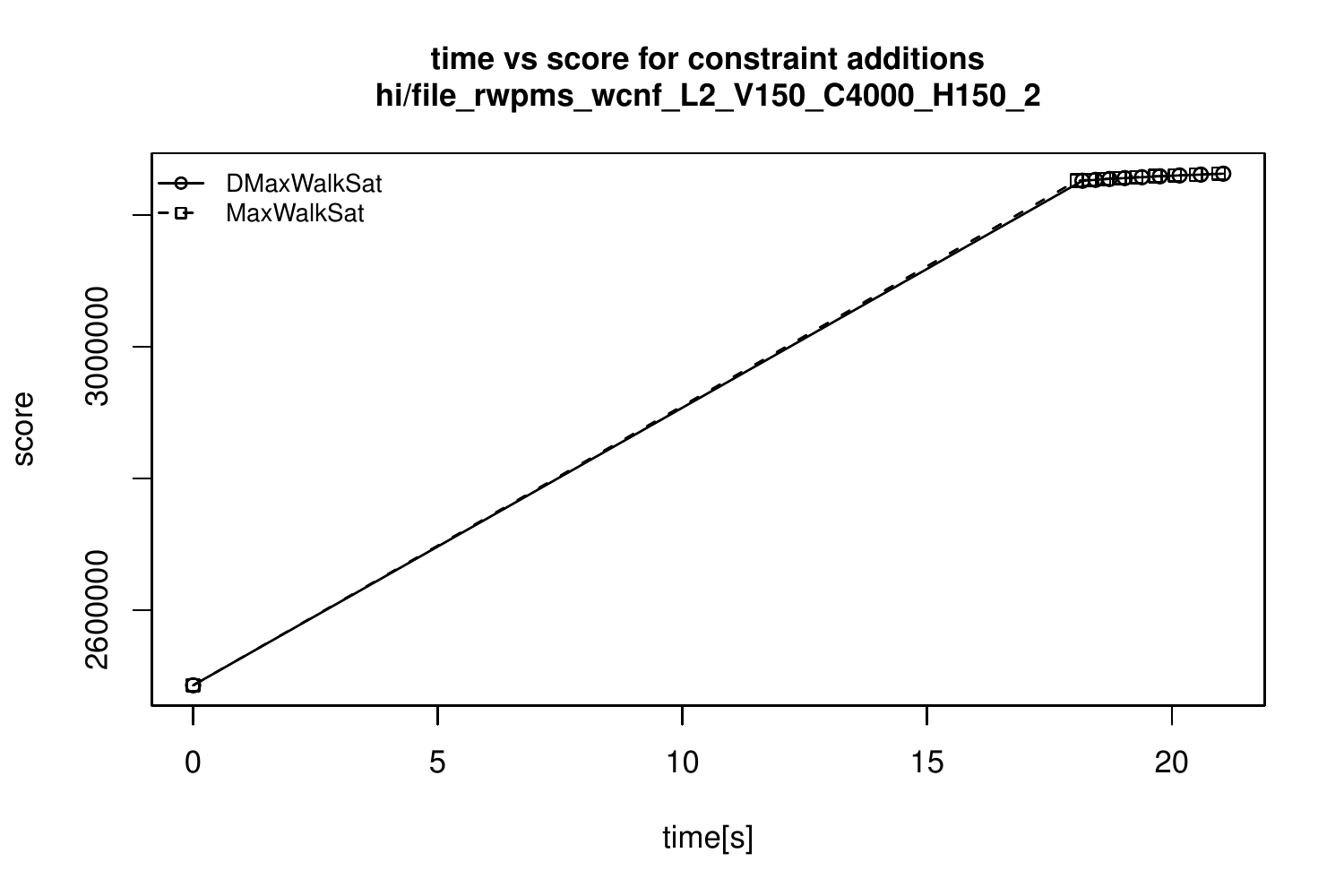}
        }
        \qquad
        \subfloat[Constraint removal]
        {
            \includegraphics[width=2.7in]{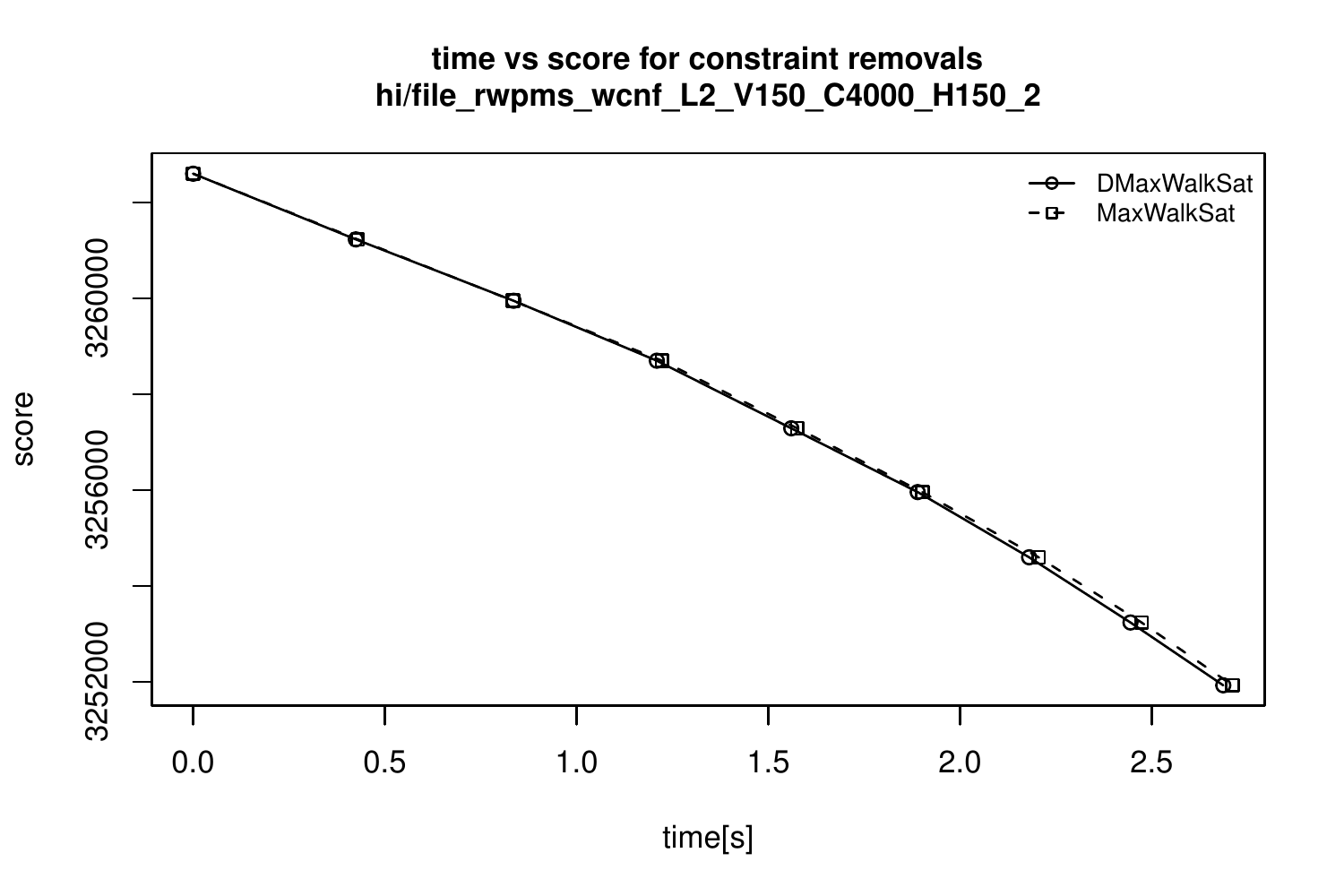}
        }

    \caption*{hi/file\_rwpms\_wcnf\_L2\_V150\_C4000\_H150\_2}
    \label{fig_hi/file_rwpms_wcnf_L2_V150_C4000_H150_2}
\end{figure}

\begin{figure}[H]
    \setcounter{subfigure}{0}
    \centering
        \subfloat[Constraint addition]
        {
            \includegraphics[width=2.7in]{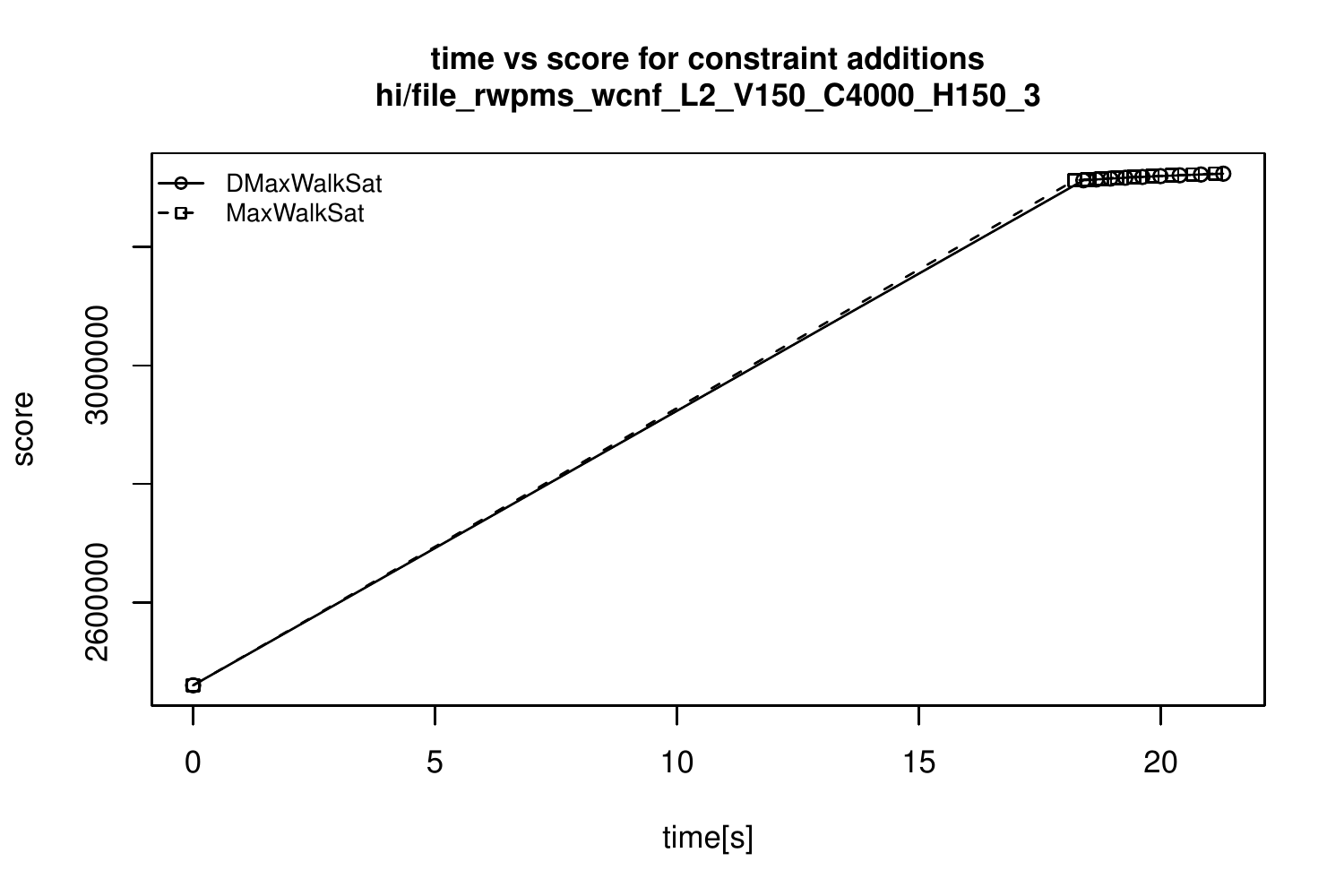}
        }
        \qquad
        \subfloat[Constraint removal]
        {
            \includegraphics[width=2.7in]{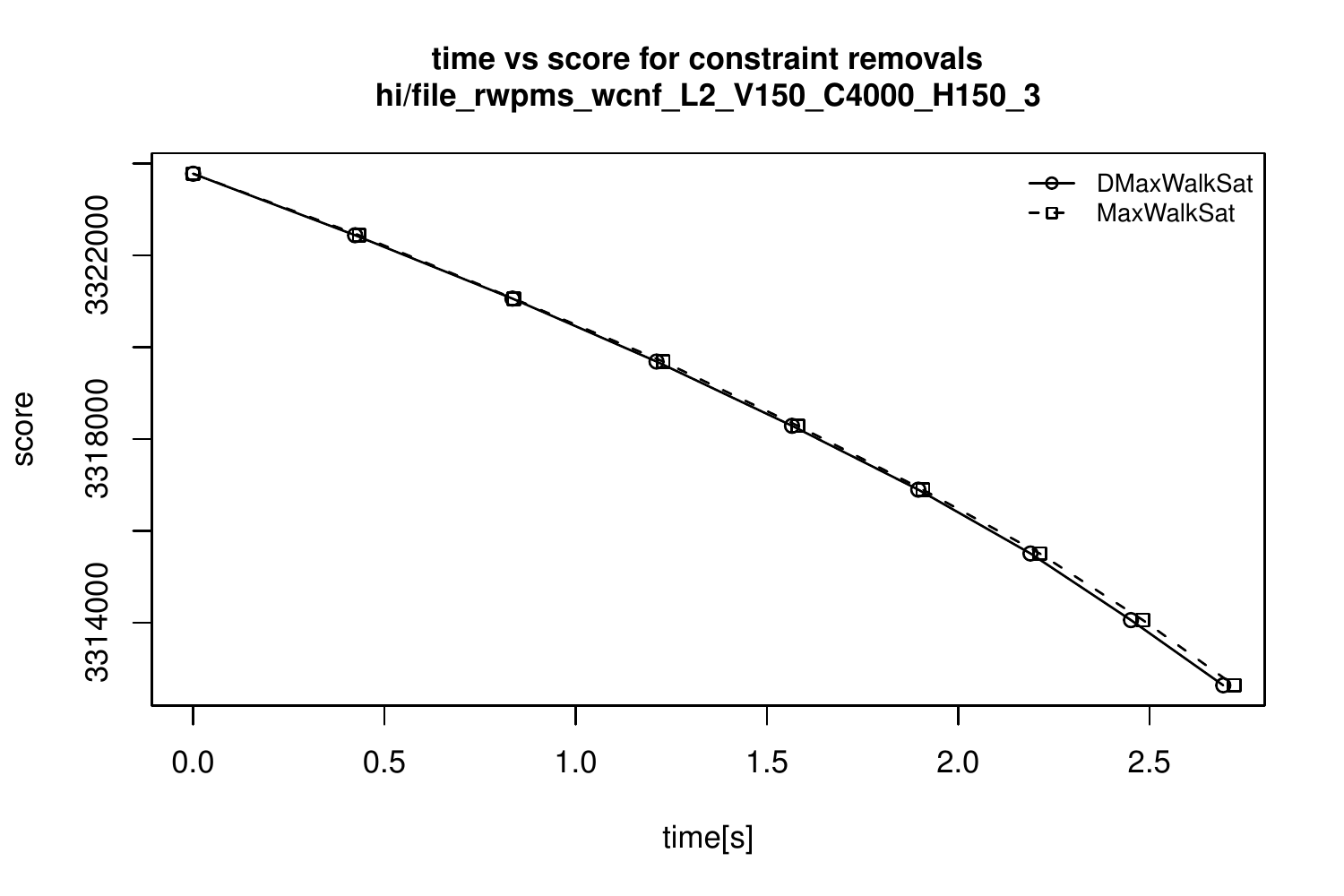}
        }

    \caption*{hi/file\_rwpms\_wcnf\_L2\_V150\_C4000\_H150\_3}
    \label{fig_hi/file_rwpms_wcnf_L2_V150_C4000_H150_3}
\end{figure}

\begin{figure}[H]
    \setcounter{subfigure}{0}
    \centering
        \subfloat[Constraint addition]
        {
            \includegraphics[width=2.7in]{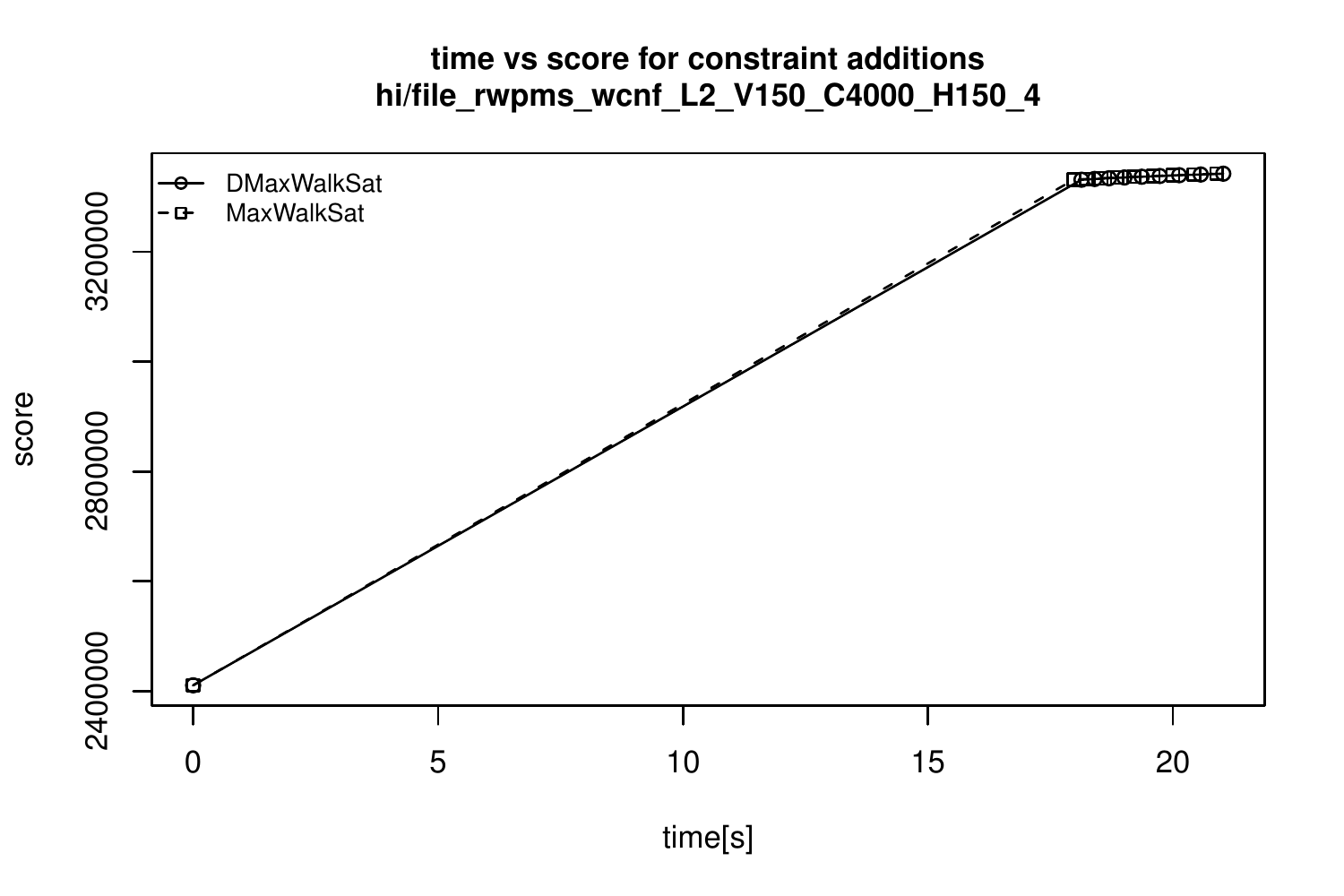}
        }
        \qquad
        \subfloat[Constraint removal]
        {
            \includegraphics[width=2.7in]{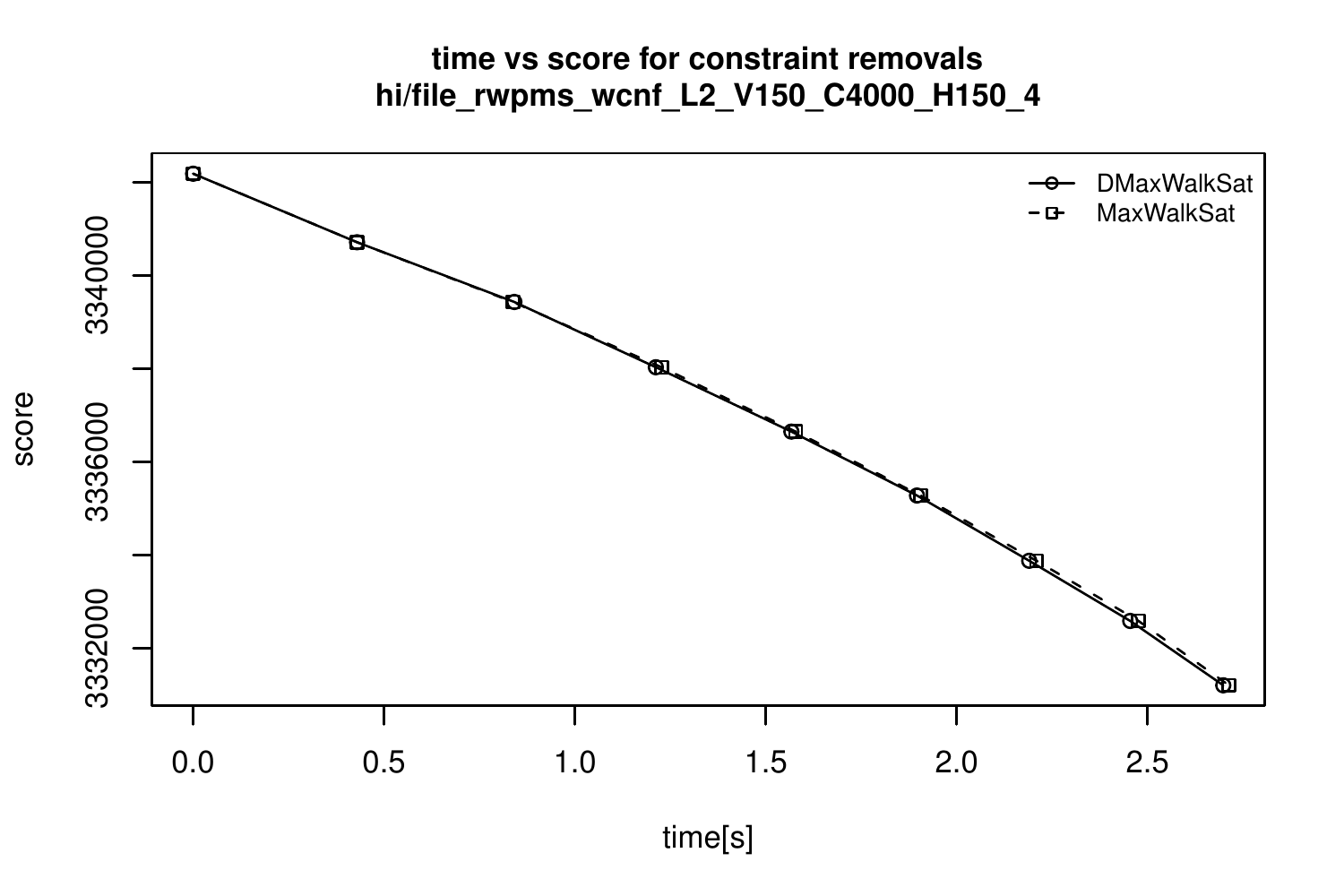}
        }

    \caption*{hi/file\_rwpms\_wcnf\_L2\_V150\_C4000\_H150\_4}
    \label{fig_hi/file_rwpms_wcnf_L2_V150_C4000_H150_4}
\end{figure}

\begin{figure}[H]
    \setcounter{subfigure}{0}
    \centering
        \subfloat[Constraint addition]
        {
            \includegraphics[width=2.7in]{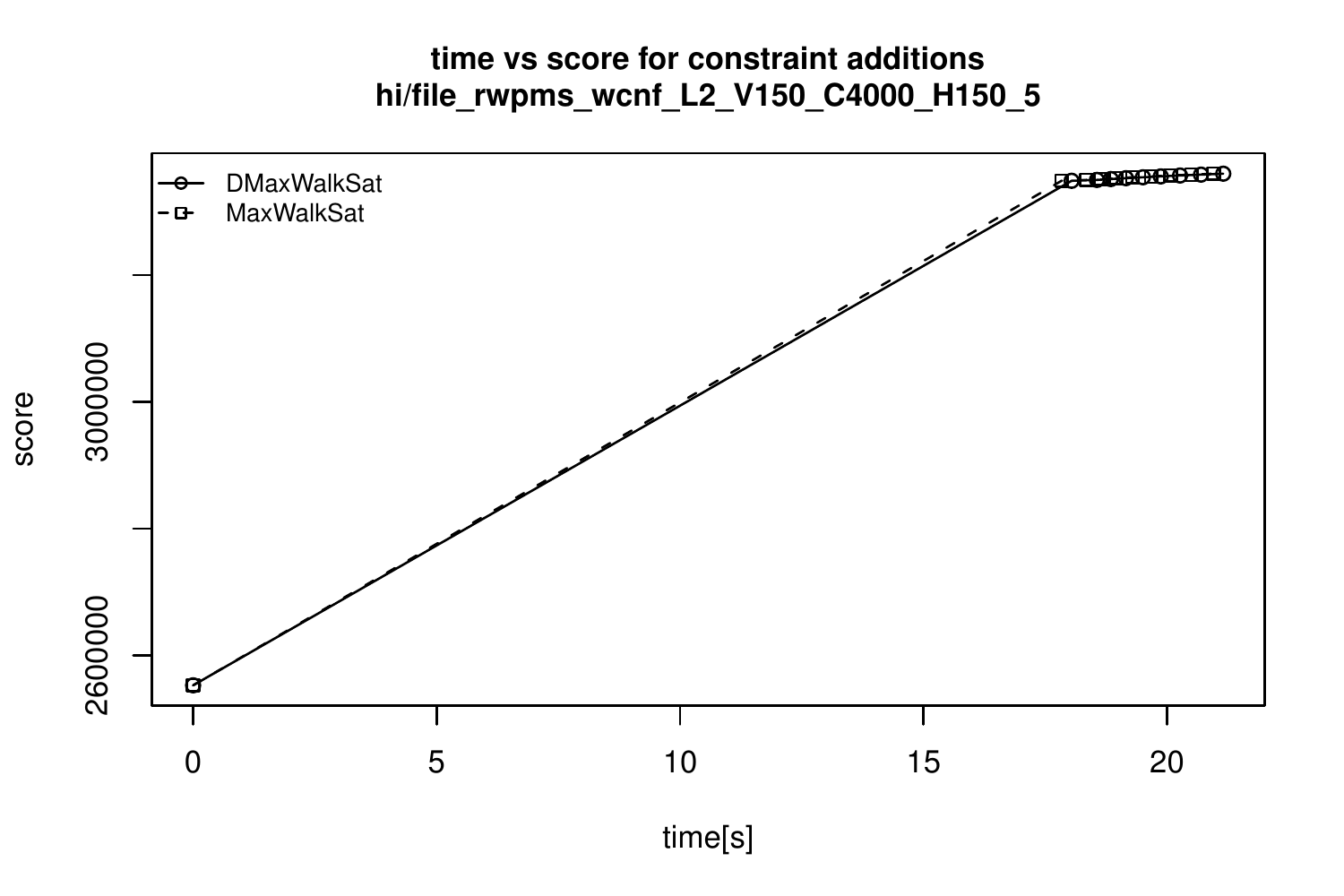}
        }
        \qquad
        \subfloat[Constraint removal]
        {
            \includegraphics[width=2.7in]{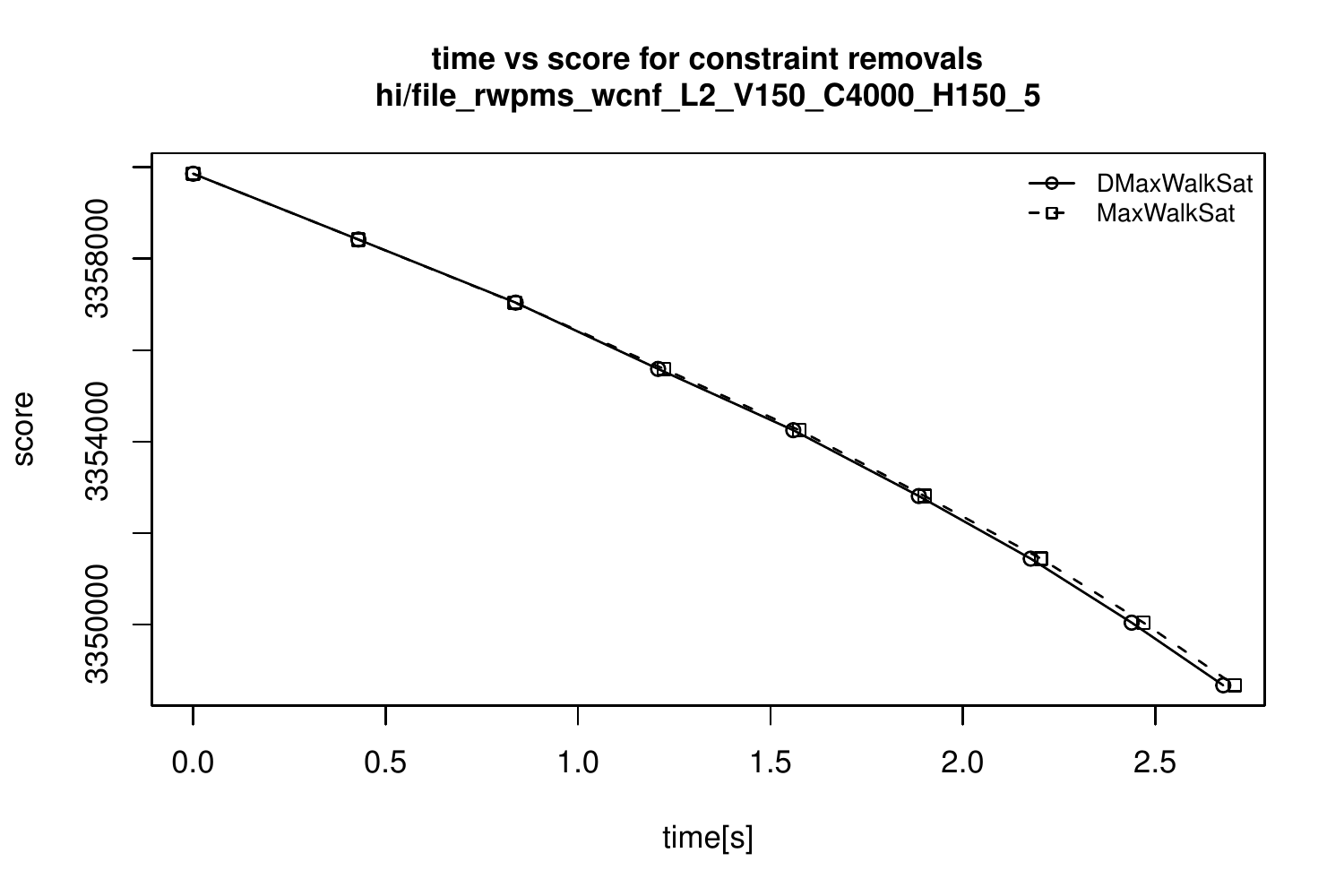}
        }

    \caption*{hi/file\_rwpms\_wcnf\_L2\_V150\_C4000\_H150\_5}
    \label{fig_hi/file_rwpms_wcnf_L2_V150_C4000_H150_5}
\end{figure}

\begin{figure}[H]
    \setcounter{subfigure}{0}
    \centering
        \subfloat[Constraint addition]
        {
            \includegraphics[width=2.7in]{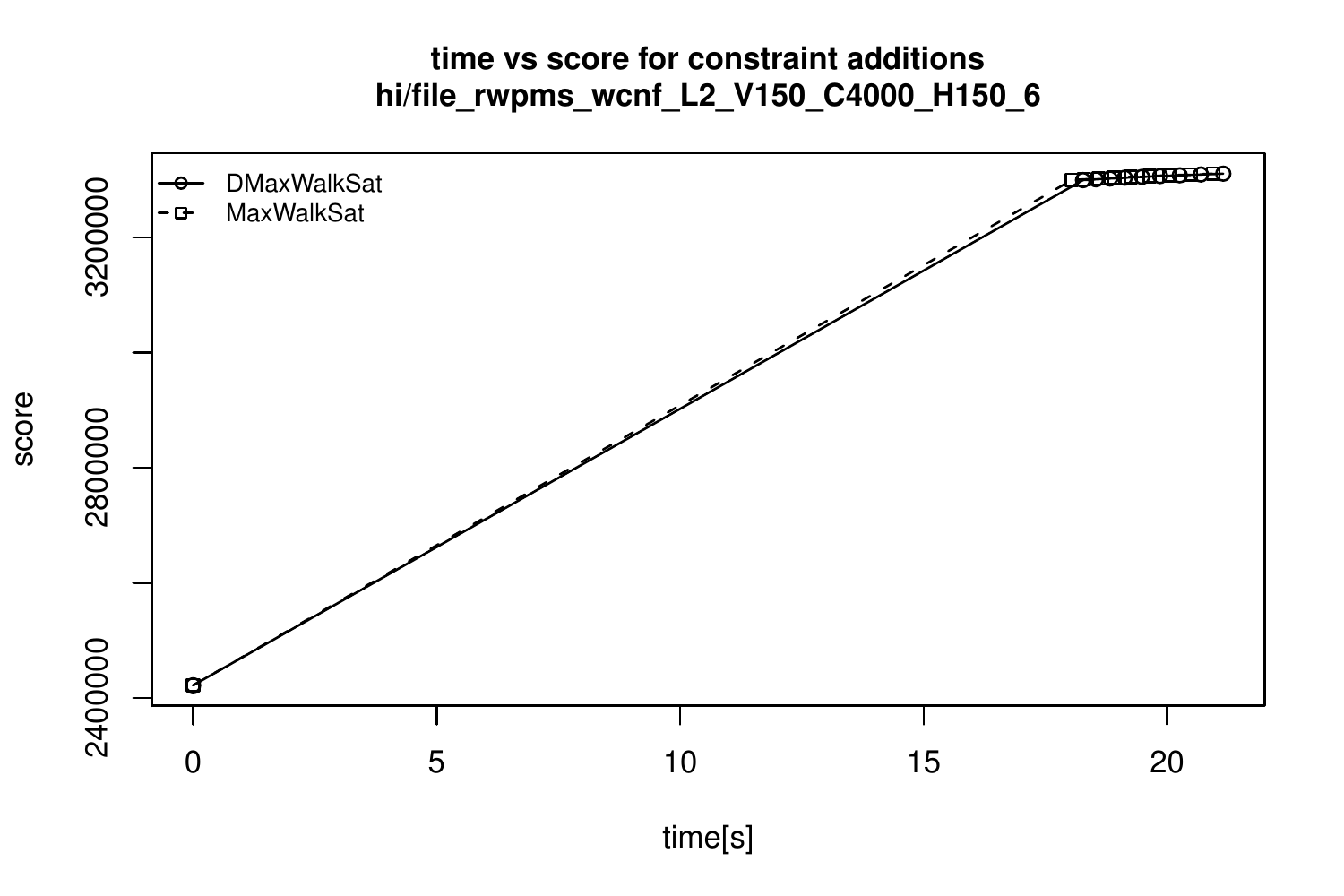}
        }
        \qquad
        \subfloat[Constraint removal]
        {
            \includegraphics[width=2.7in]{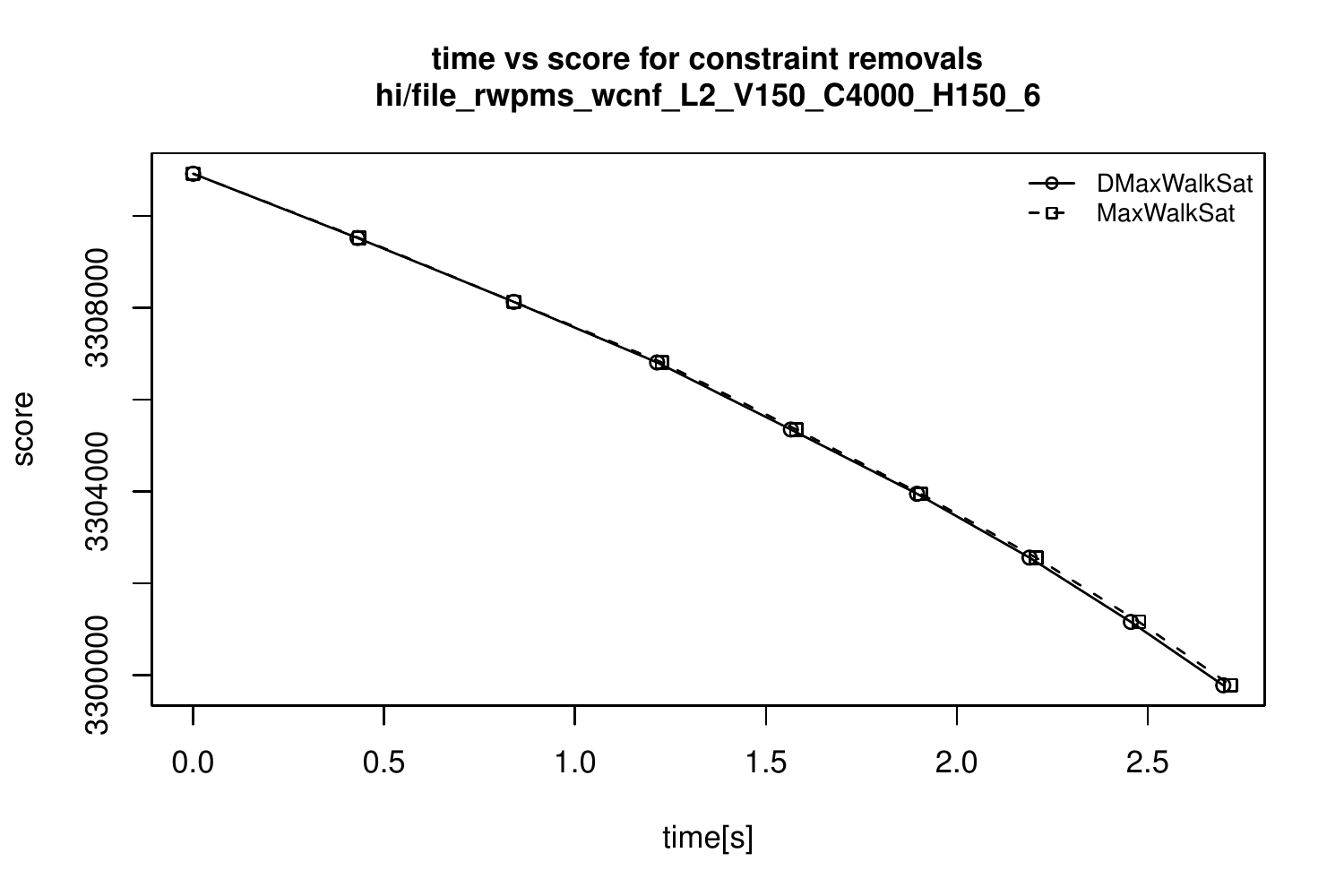}
        }

    \caption*{hi/file\_rwpms\_wcnf\_L2\_V150\_C4000\_H150\_6}
    \label{fig_hi/file_rwpms_wcnf_L2_V150_C4000_H150_6}
\end{figure}

\begin{figure}[H]
    \setcounter{subfigure}{0}
    \centering
        \subfloat[Constraint addition]
        {
            \includegraphics[width=2.7in]{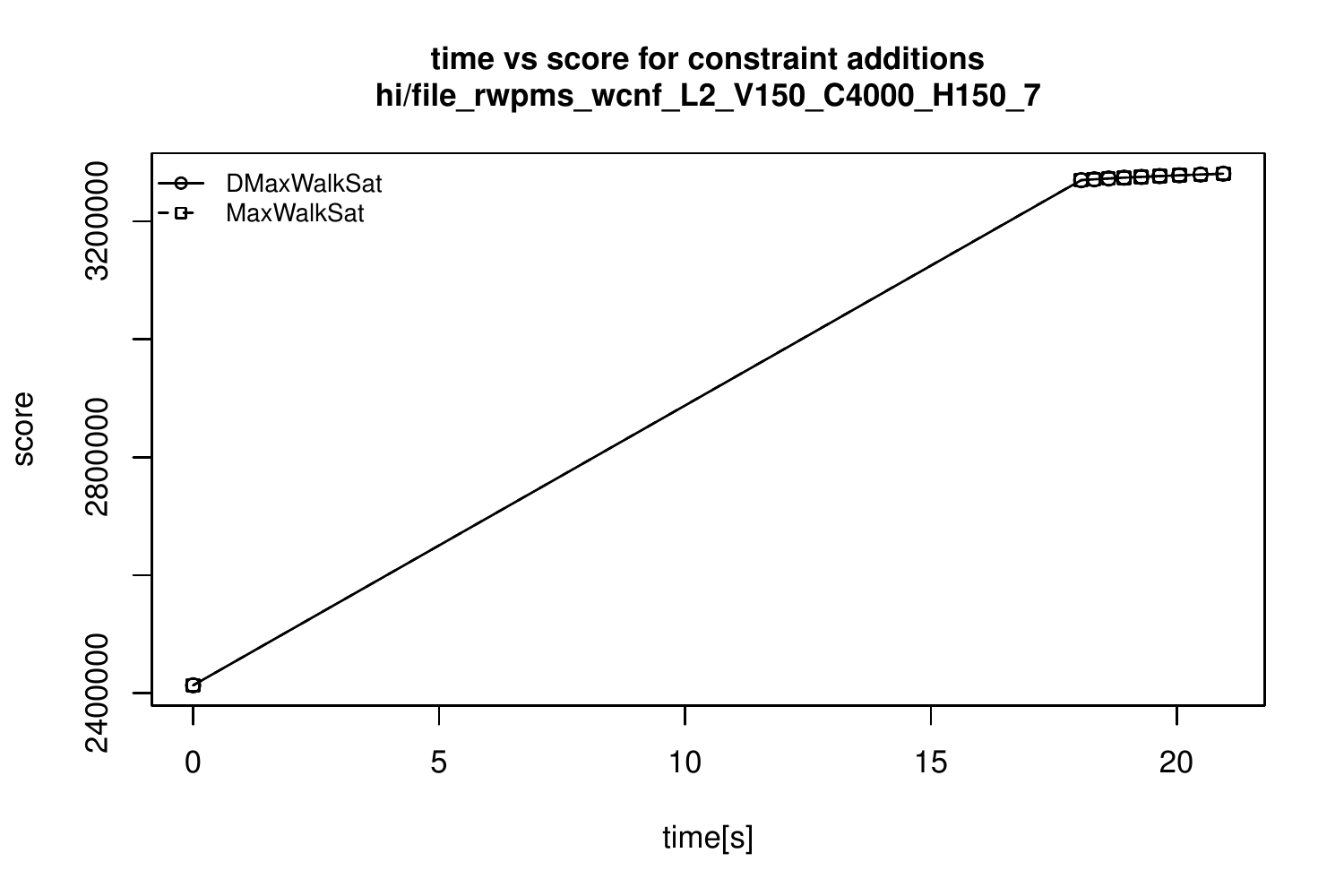}
        }
        \qquad
        \subfloat[Constraint removal]
        {
            \includegraphics[width=2.7in]{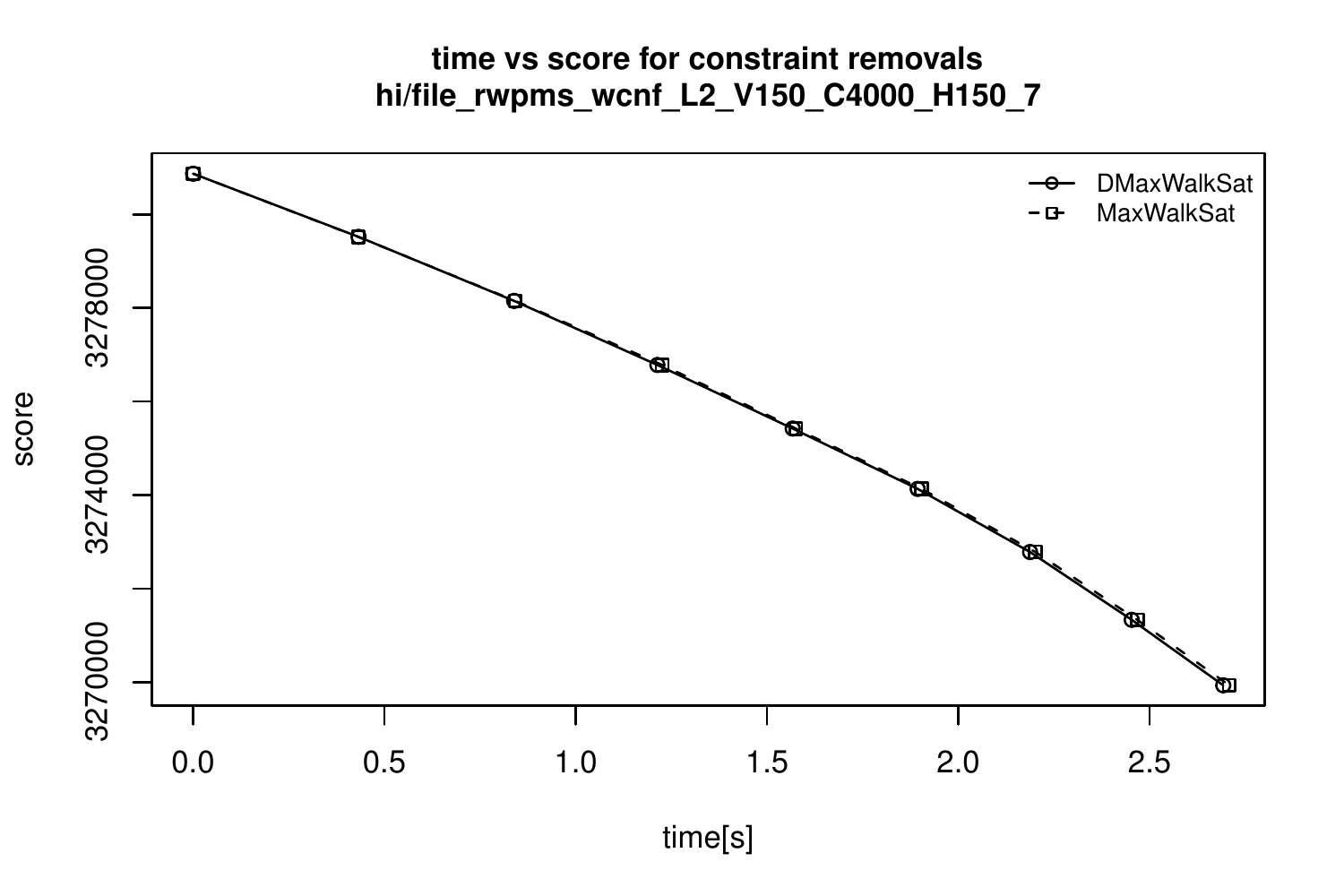}
        }

    \caption*{hi/file\_rwpms\_wcnf\_L2\_V150\_C4000\_H150\_7}
    \label{fig_hi/file_rwpms_wcnf_L2_V150_C4000_H150_7}
\end{figure}

\begin{figure}[H]
    \setcounter{subfigure}{0}
    \centering
        \subfloat[Constraint addition]
        {
            \includegraphics[width=2.7in]{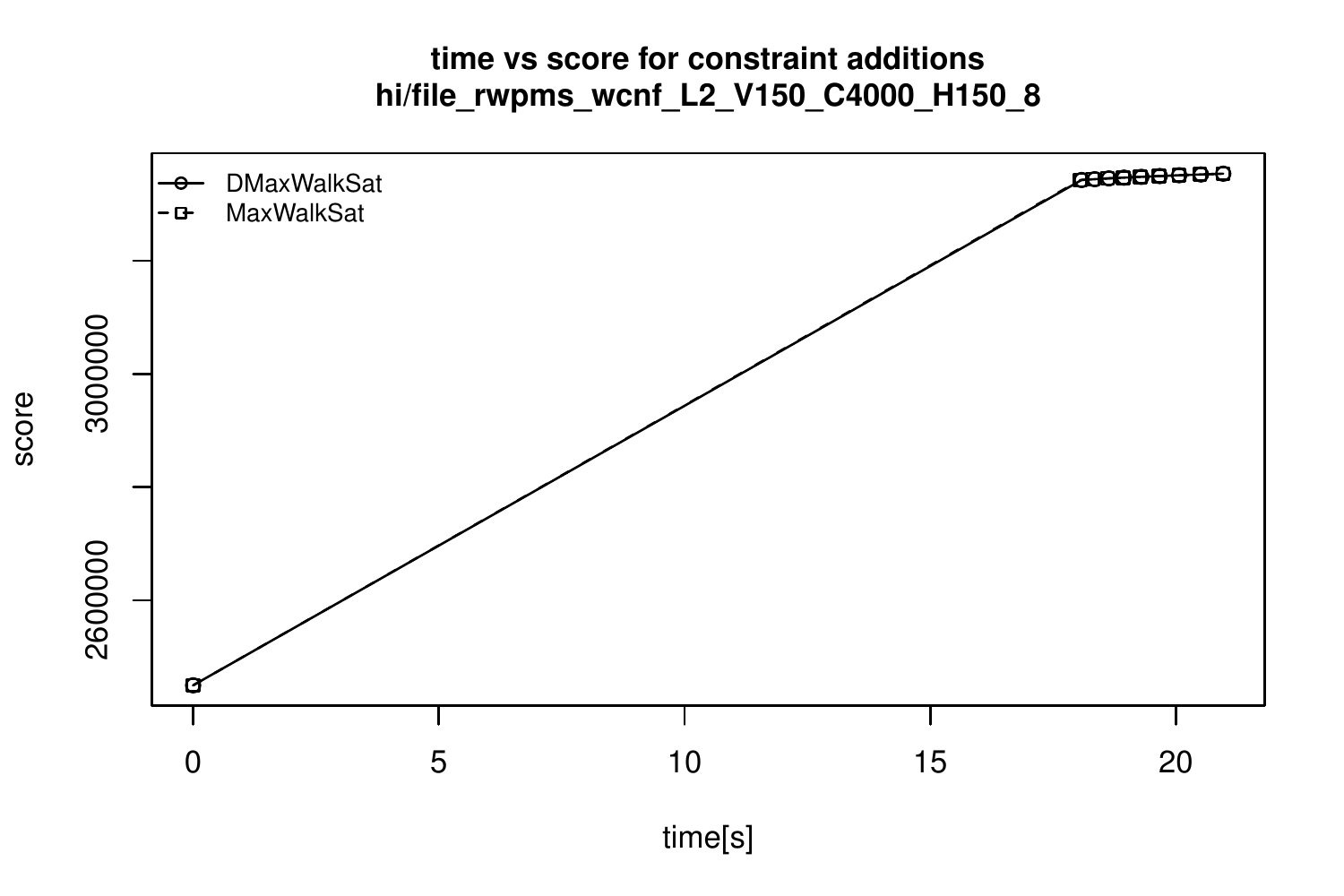}
        }
        \qquad
        \subfloat[Constraint removal]
        {
            \includegraphics[width=2.7in]{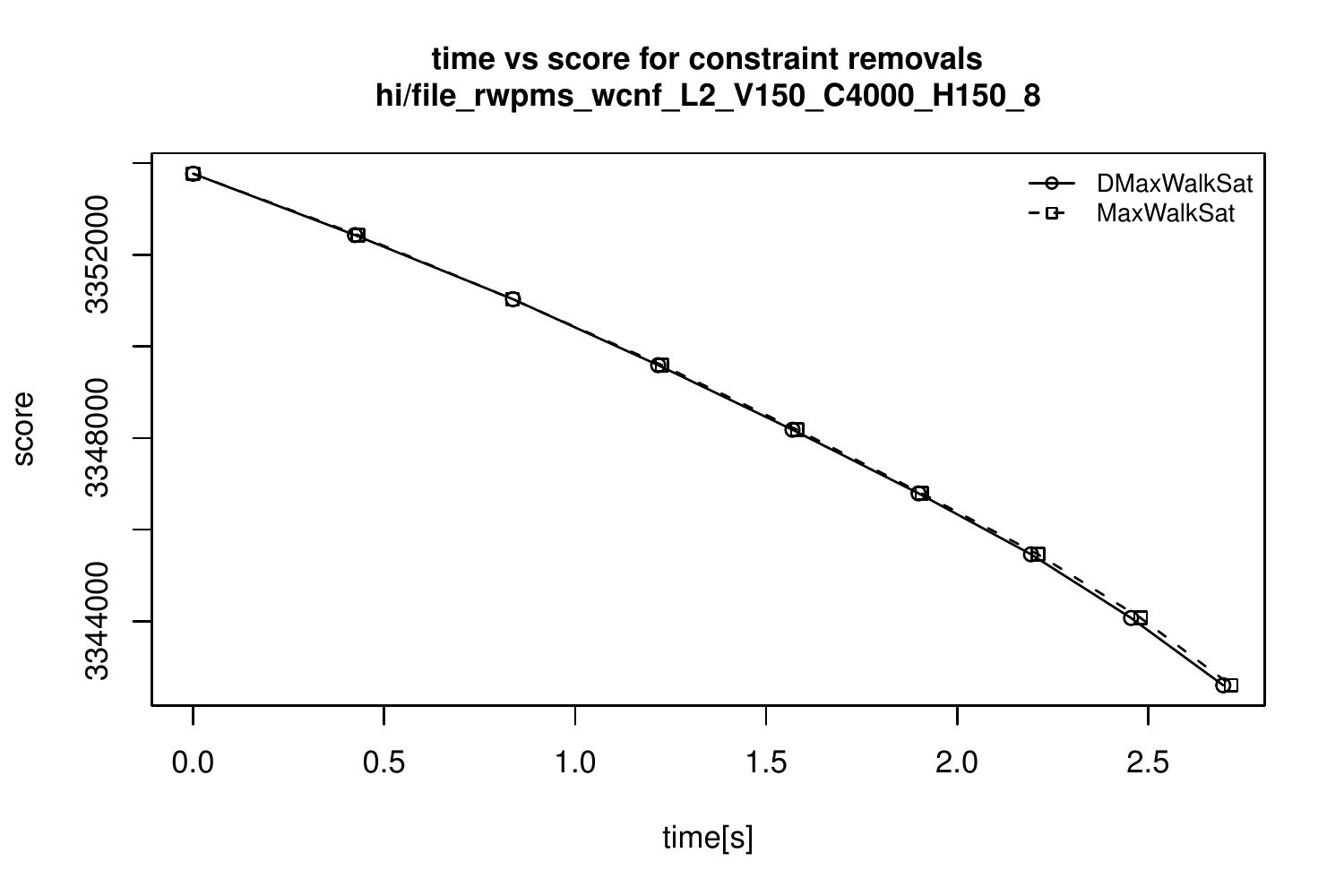}
        }

    \caption*{hi/file\_rwpms\_wcnf\_L2\_V150\_C4000\_H150\_8}
    \label{fig_hi/file_rwpms_wcnf_L2_V150_C4000_H150_8}
\end{figure}

\begin{figure}[H]
    \setcounter{subfigure}{0}
    \centering
        \subfloat[Constraint addition]
        {
            \includegraphics[width=2.7in]{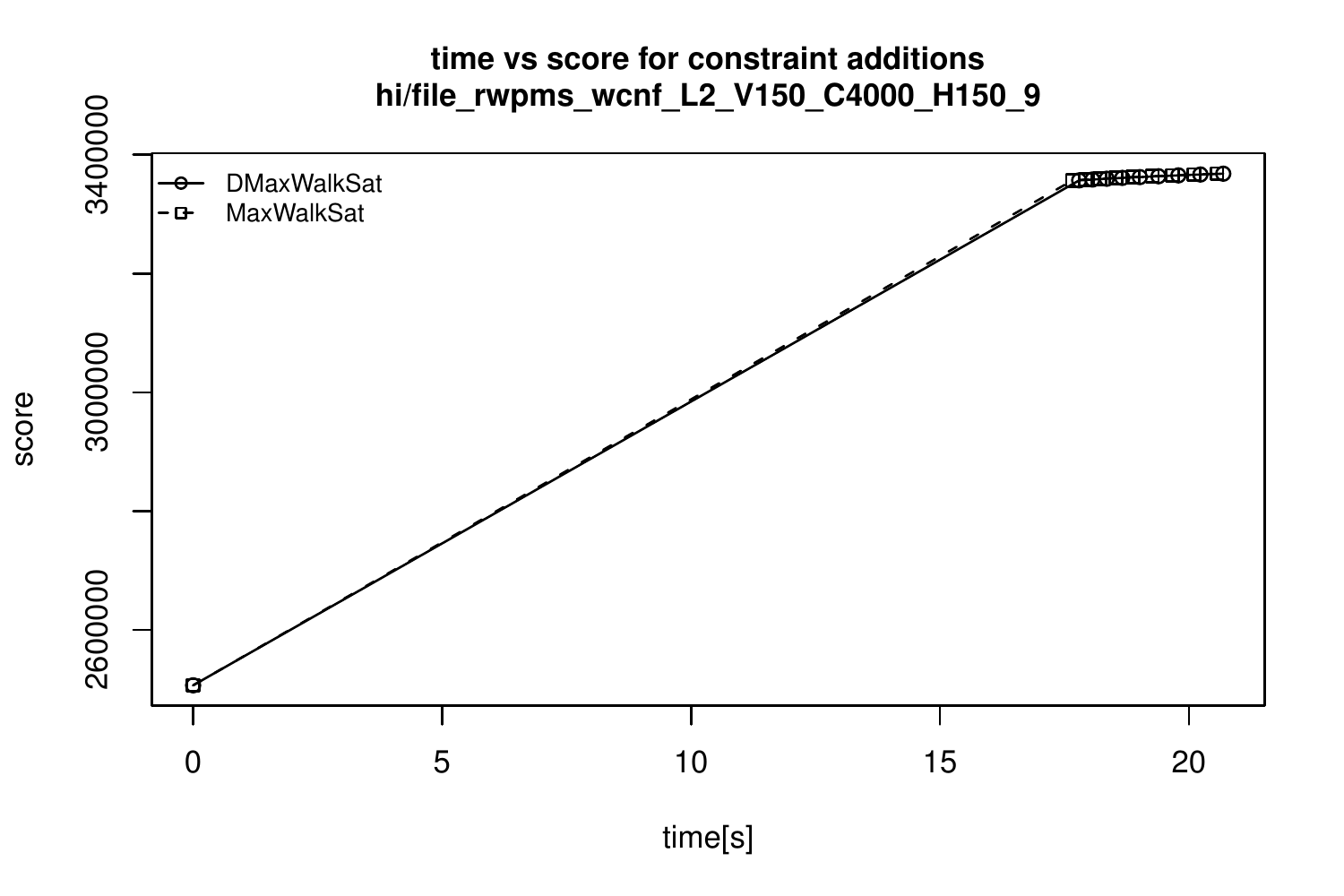}
        }
        \qquad
        \subfloat[Constraint removal]
        {
            \includegraphics[width=2.7in]{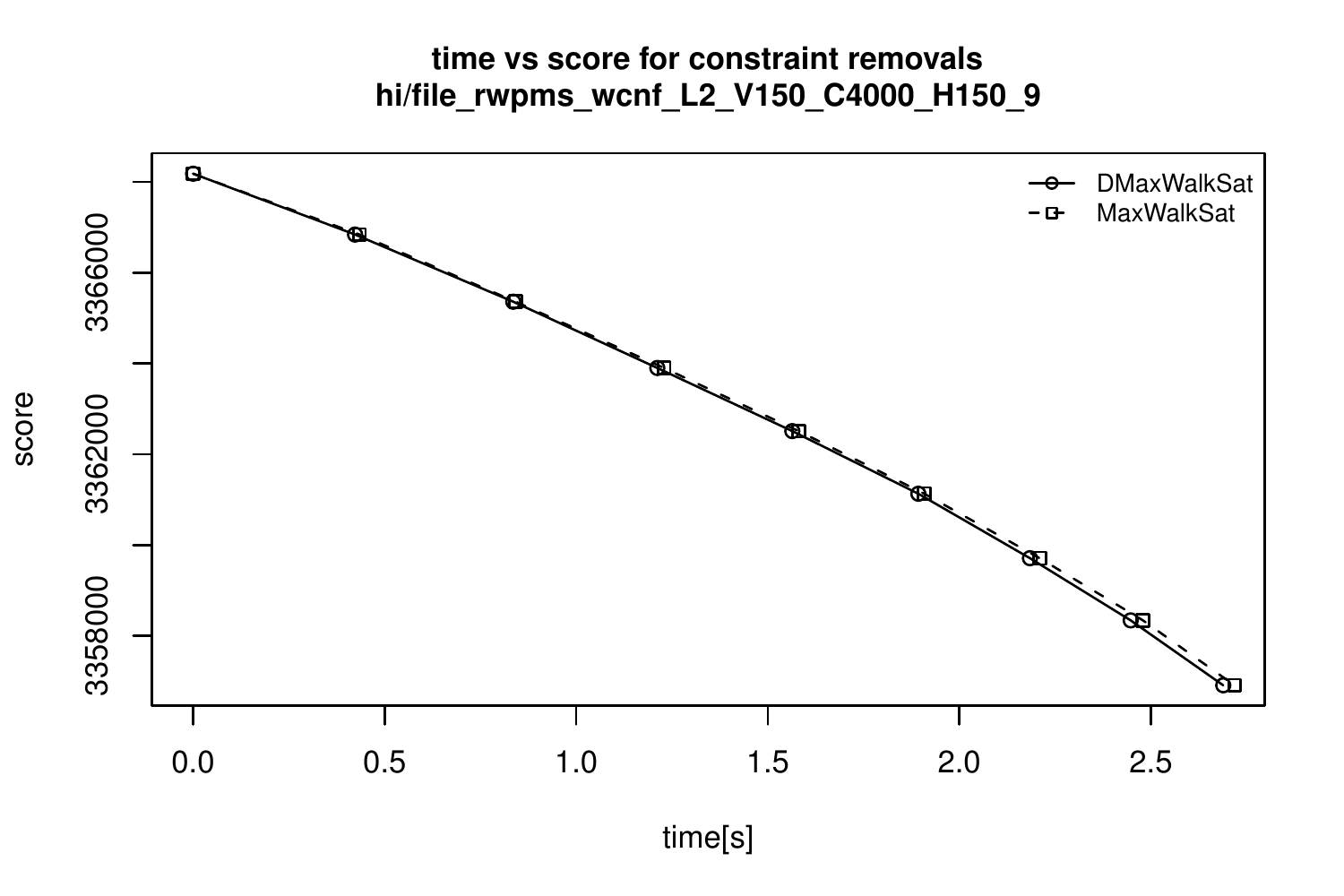}
        }

    \caption*{hi/file\_rwpms\_wcnf\_L2\_V150\_C4000\_H150\_9}
    \label{fig_hi/file_rwpms_wcnf_L2_V150_C4000_H150_9}
\end{figure}

\begin{figure}[H]
    \setcounter{subfigure}{0}
    \centering
        \subfloat[Constraint addition]
        {
            \includegraphics[width=2.7in]{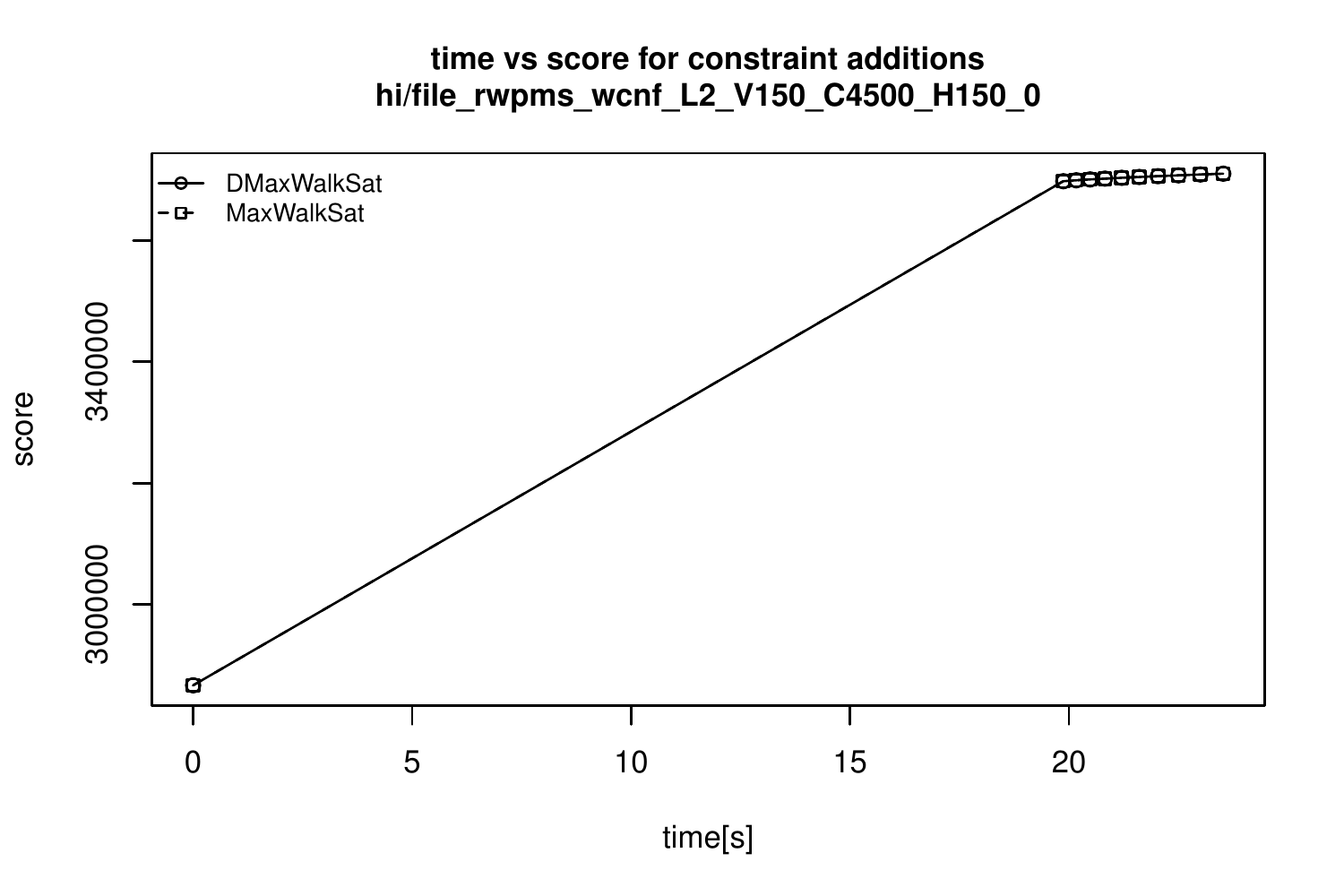}
        }
        \qquad
        \subfloat[Constraint removal]
        {
            \includegraphics[width=2.7in]{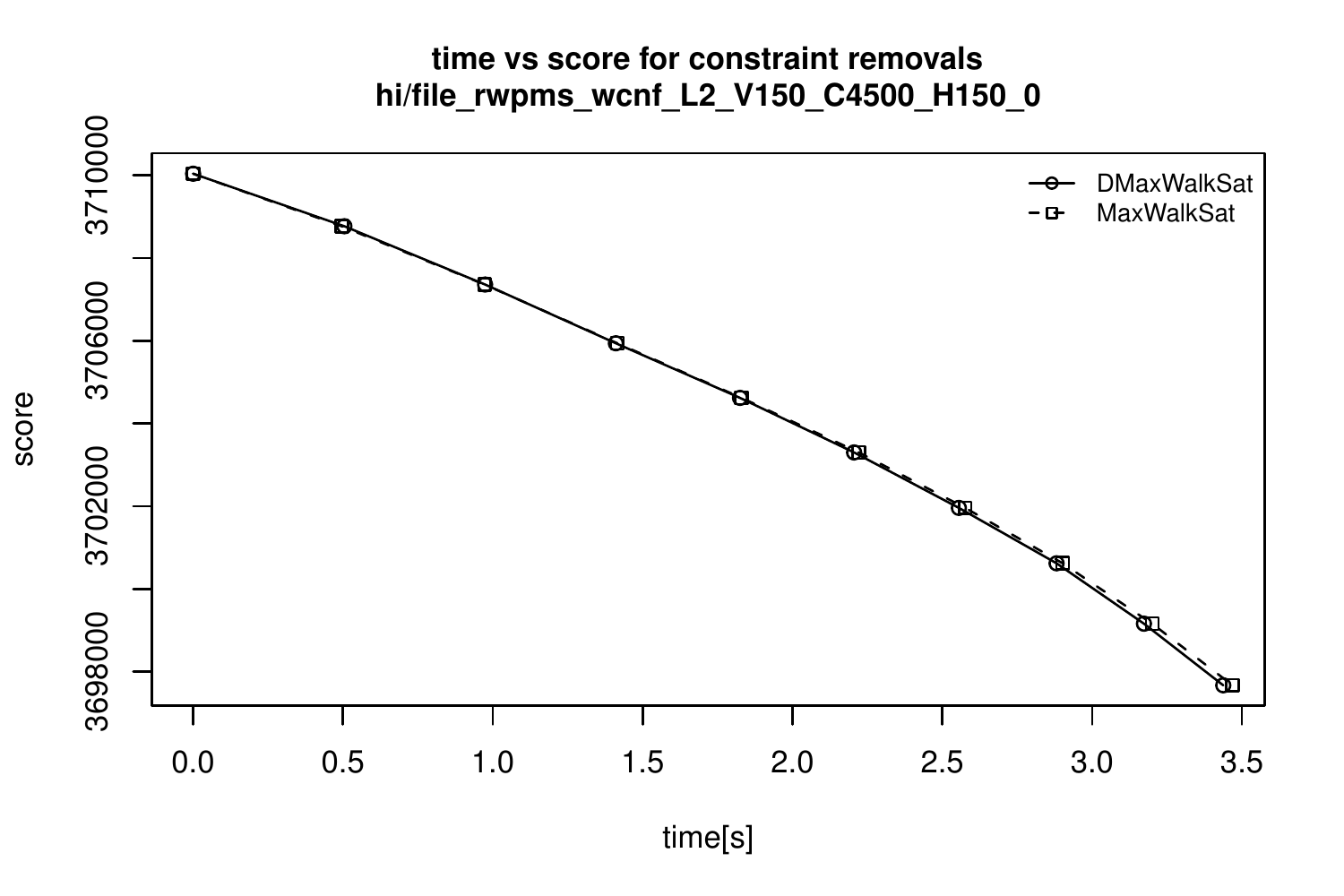}
        }

    \caption*{hi/file\_rwpms\_wcnf\_L2\_V150\_C4500\_H150\_0}
    \label{fig_hi/file_rwpms_wcnf_L2_V150_C4500_H150_0}
\end{figure}

\begin{figure}[H]
    \setcounter{subfigure}{0}
    \centering
        \subfloat[Constraint addition]
        {
            \includegraphics[width=2.7in]{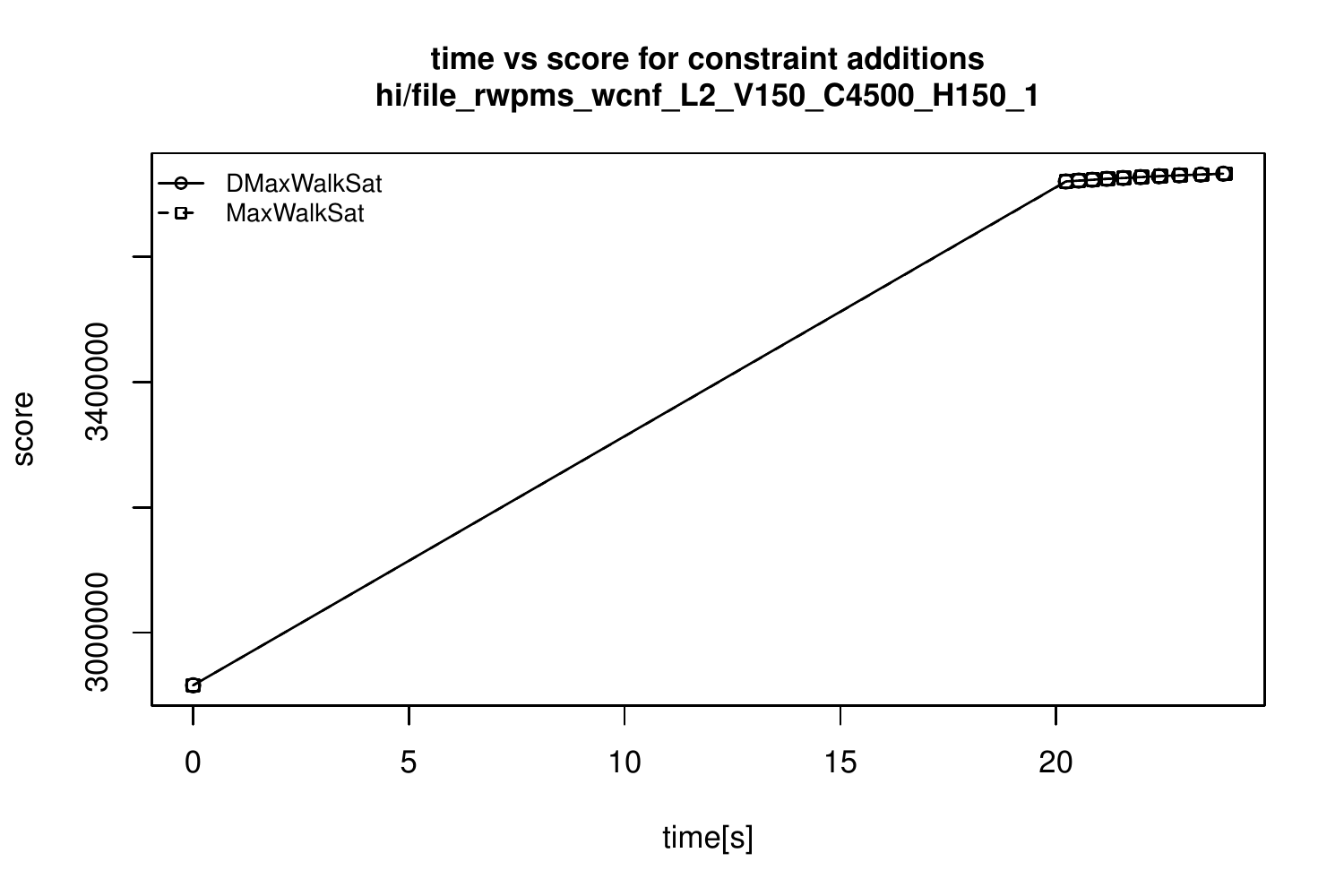}
        }
        \qquad
        \subfloat[Constraint removal]
        {
            \includegraphics[width=2.7in]{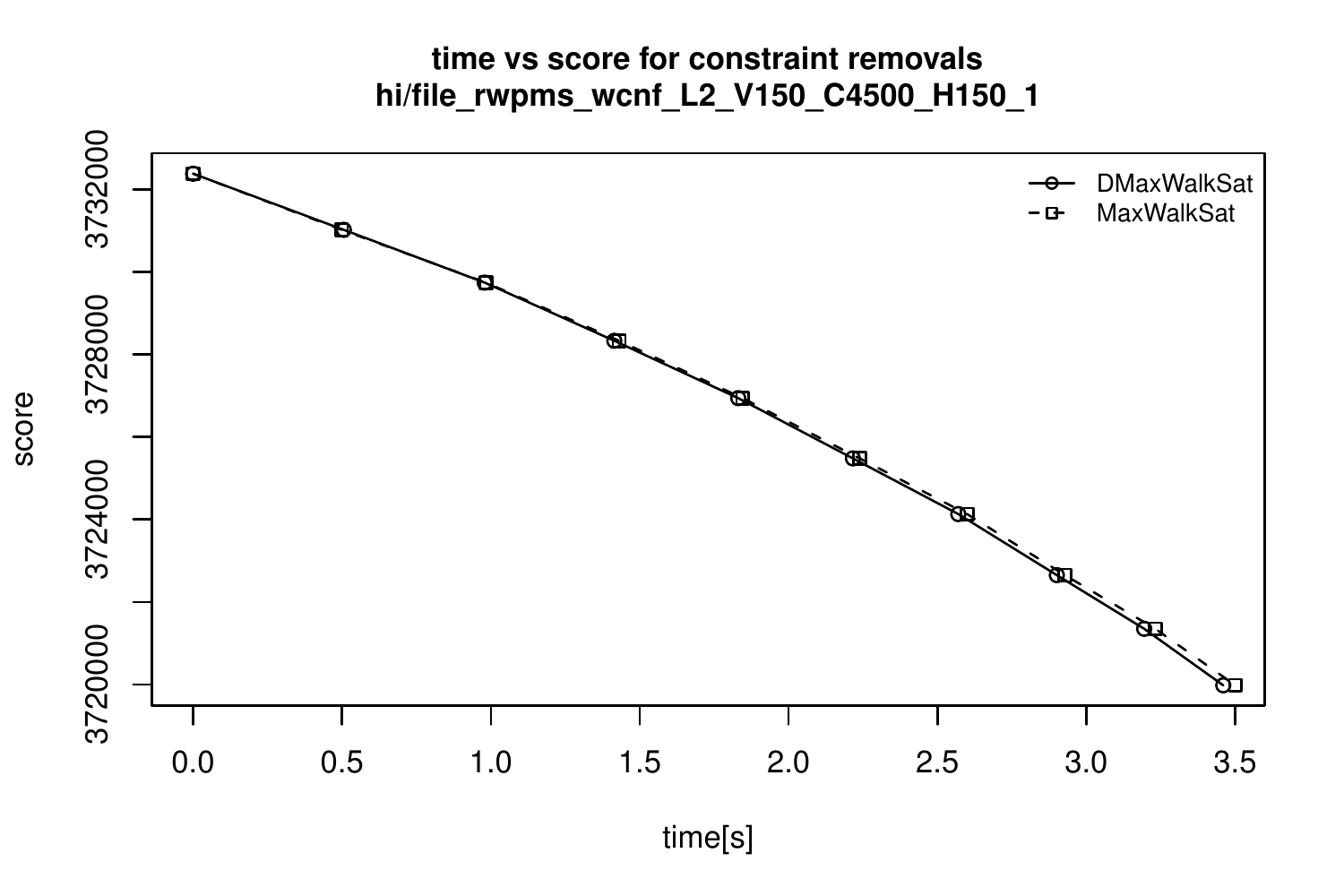}
        }

    \caption*{hi/file\_rwpms\_wcnf\_L2\_V150\_C4500\_H150\_1}
    \label{fig_hi/file_rwpms_wcnf_L2_V150_C4500_H150_1}
\end{figure}

\begin{figure}[H]
    \setcounter{subfigure}{0}
    \centering
        \subfloat[Constraint addition]
        {
            \includegraphics[width=2.7in]{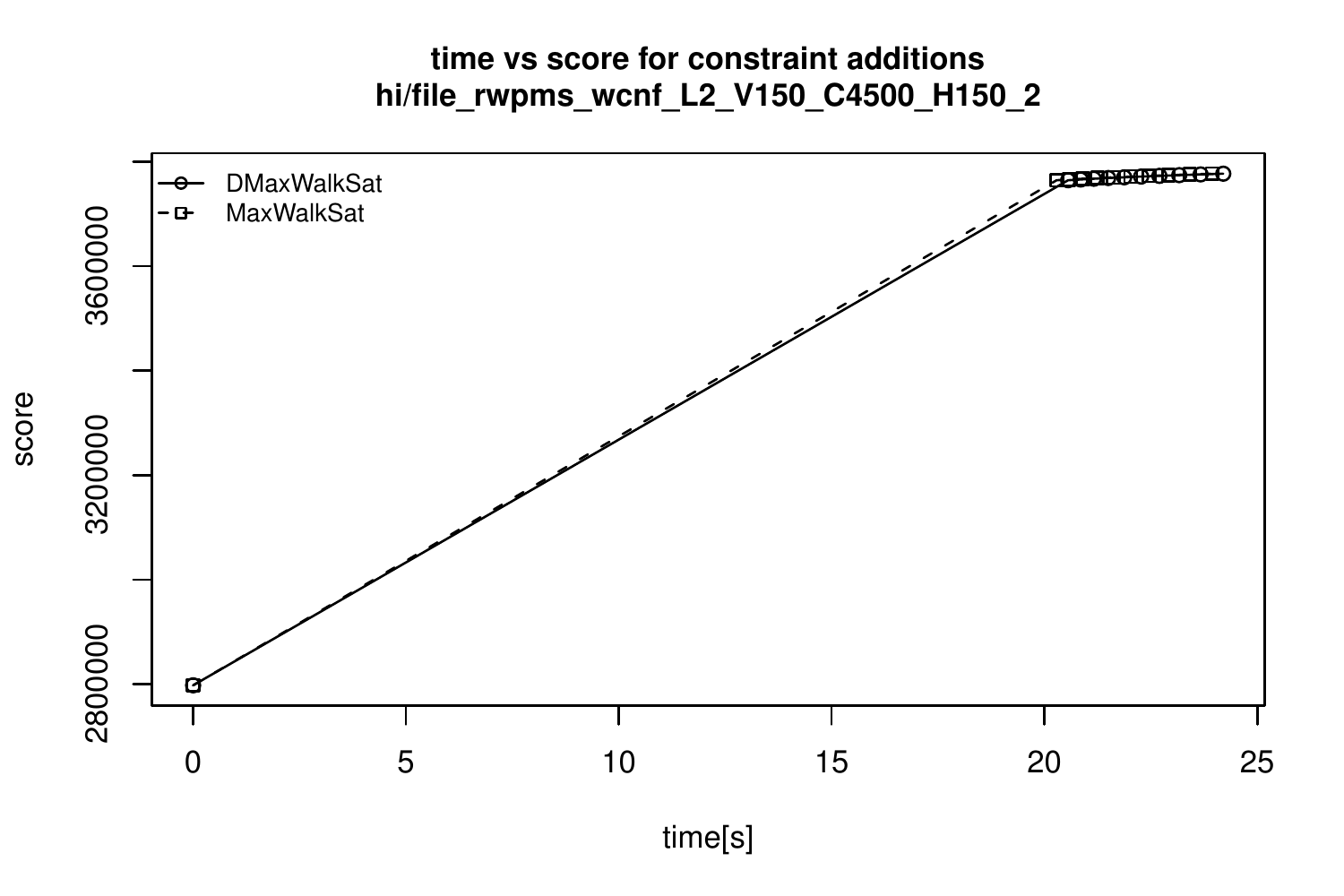}
        }
        \qquad
        \subfloat[Constraint removal]
        {
            \includegraphics[width=2.7in]{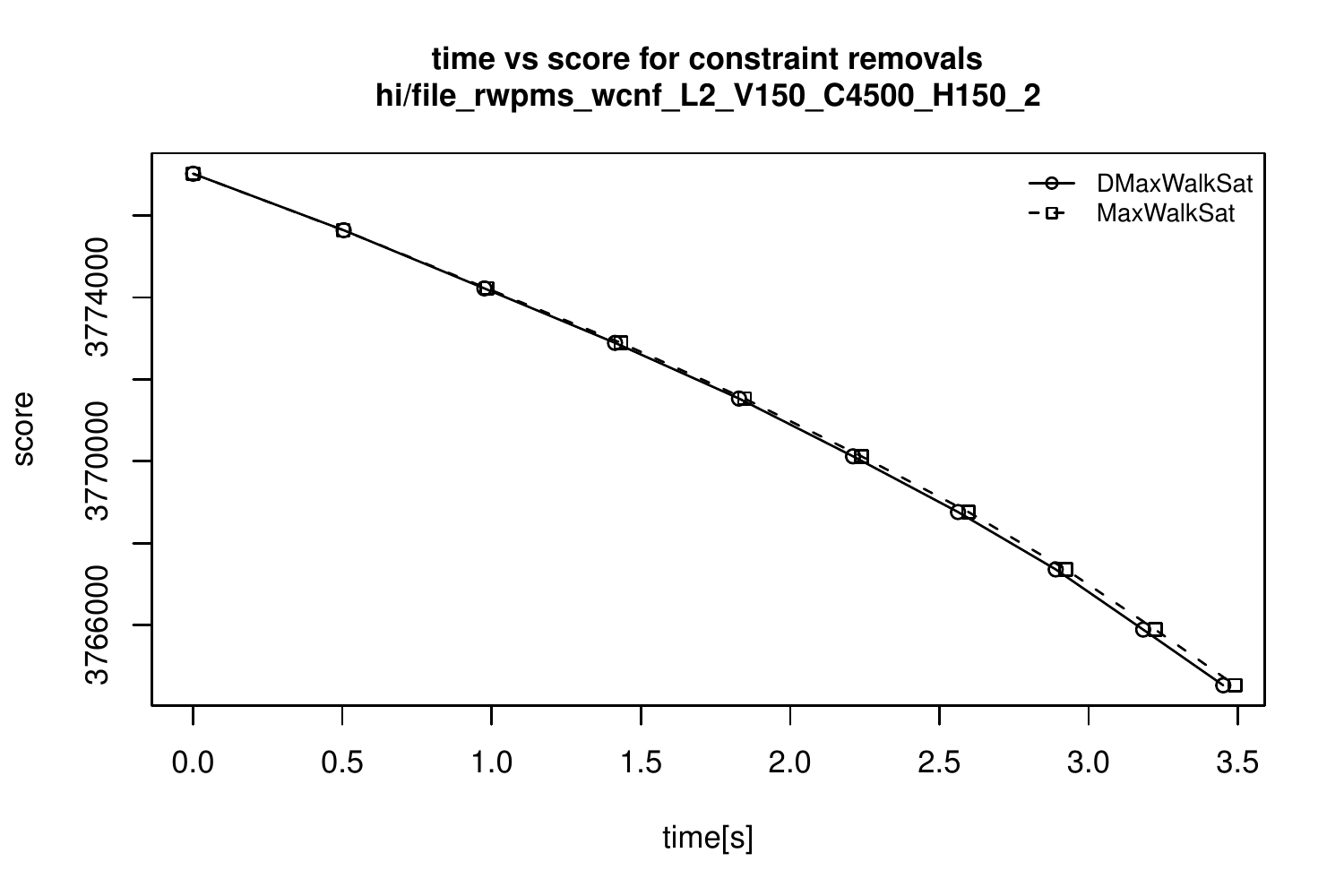}
        }

    \caption*{hi/file\_rwpms\_wcnf\_L2\_V150\_C4500\_H150\_2}
    \label{fig_hi/file_rwpms_wcnf_L2_V150_C4500_H150_2}
\end{figure}

\begin{figure}[H]
    \setcounter{subfigure}{0}
    \centering
        \subfloat[Constraint addition]
        {
            \includegraphics[width=2.7in]{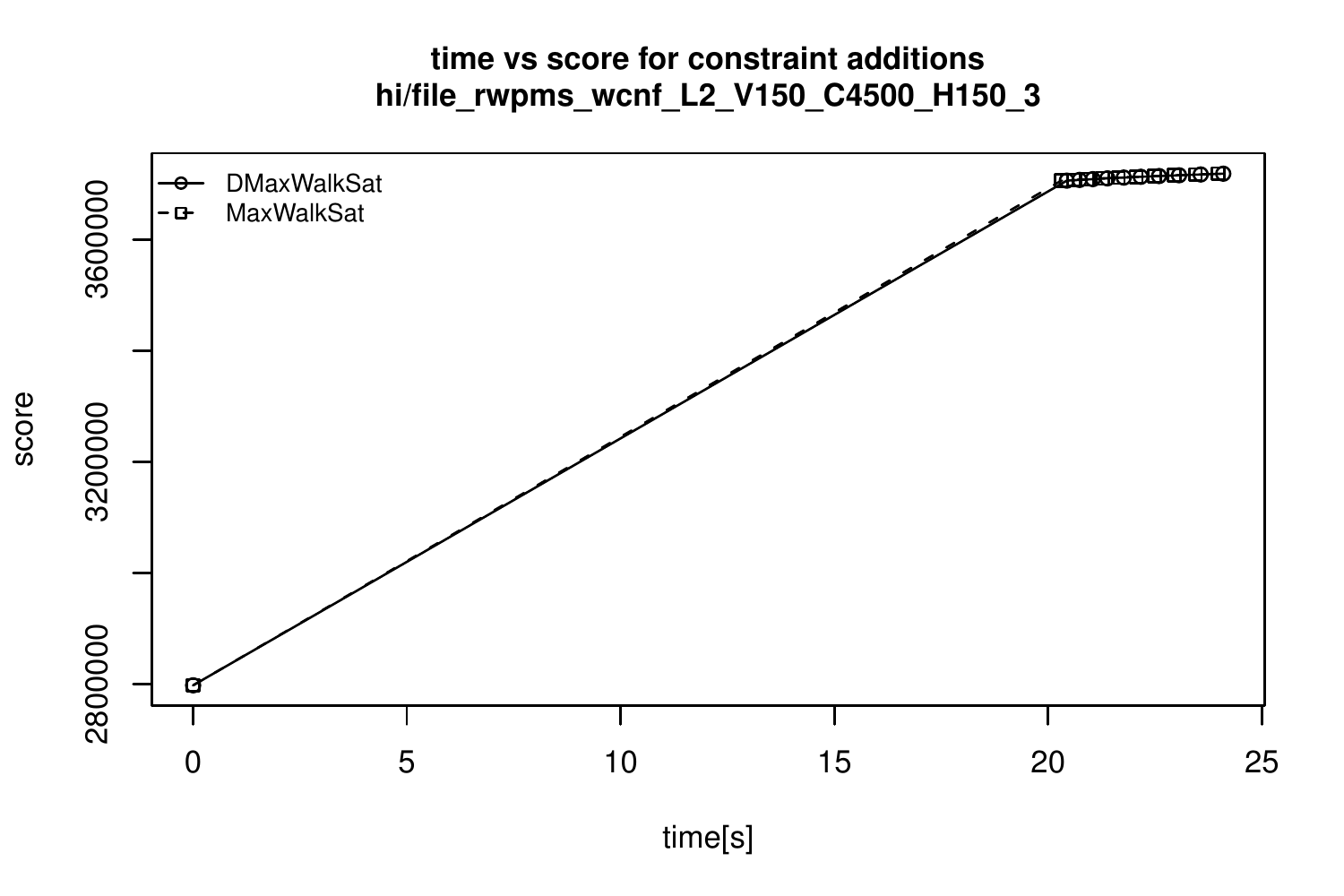}
        }
        \qquad
        \subfloat[Constraint removal]
        {
            \includegraphics[width=2.7in]{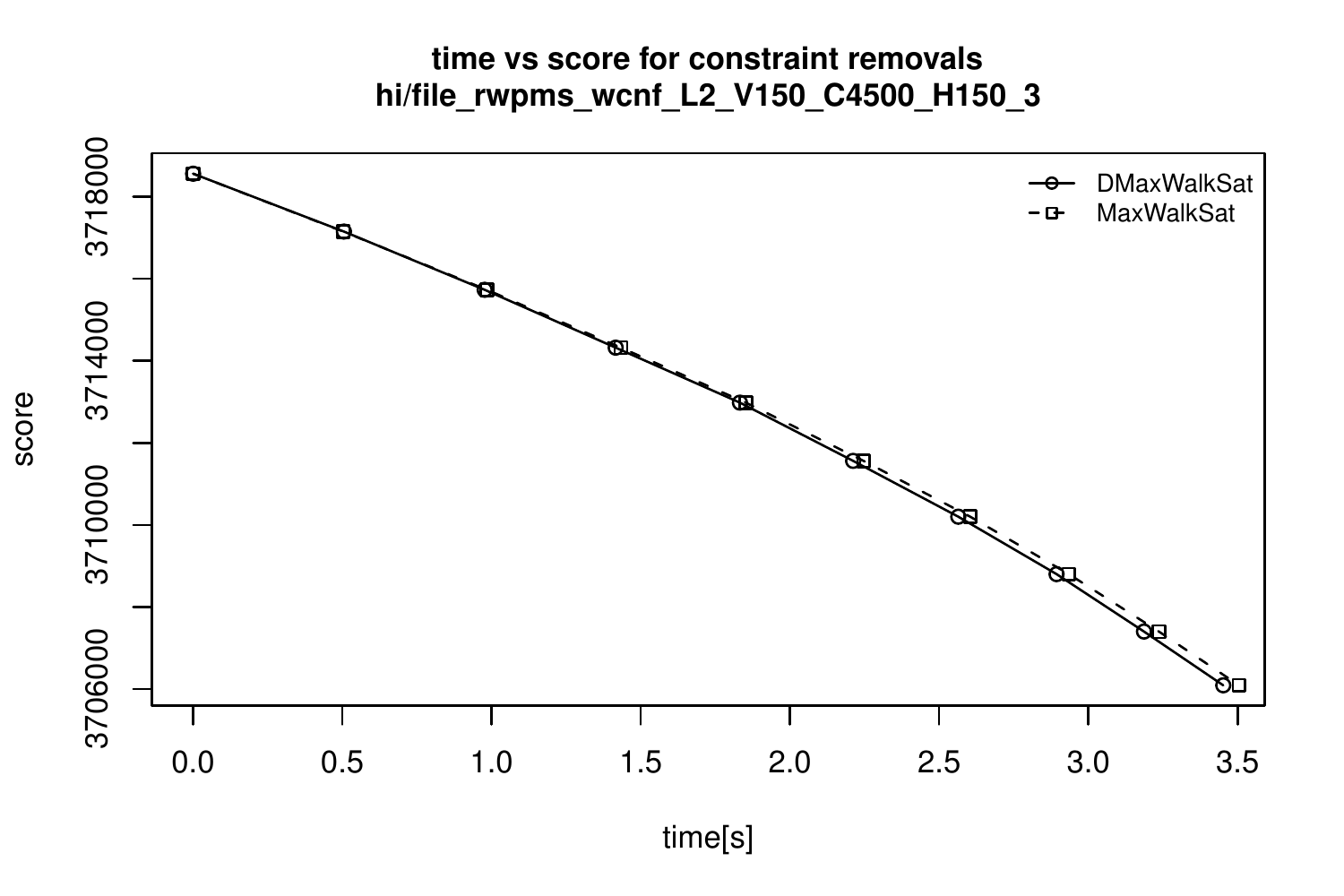}
        }

    \caption*{hi/file\_rwpms\_wcnf\_L2\_V150\_C4500\_H150\_3}
    \label{fig_hi/file_rwpms_wcnf_L2_V150_C4500_H150_3}
\end{figure}

\begin{figure}[H]
    \setcounter{subfigure}{0}
    \centering
        \subfloat[Constraint addition]
        {
            \includegraphics[width=2.7in]{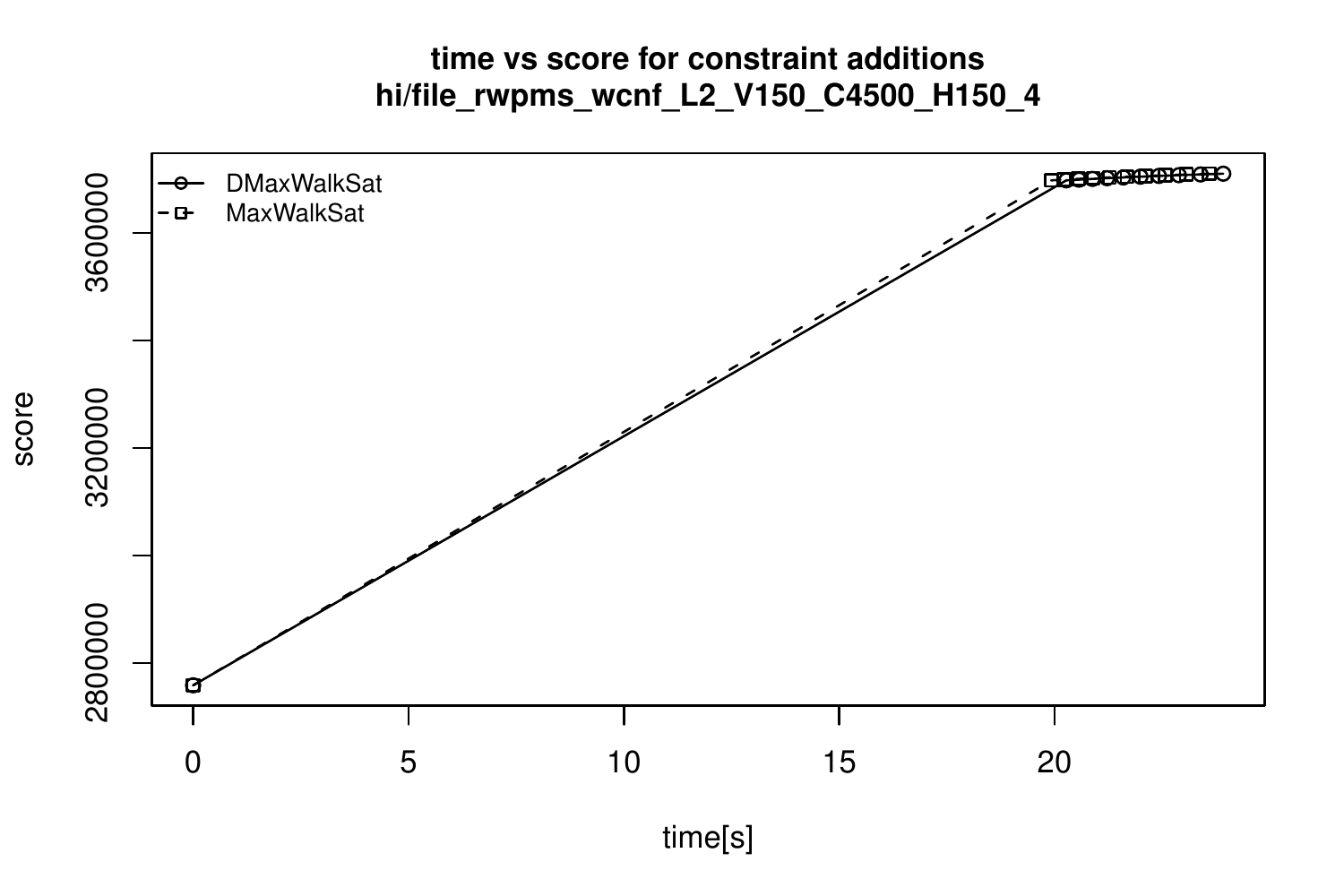}
        }
        \qquad
        \subfloat[Constraint removal]
        {
            \includegraphics[width=2.7in]{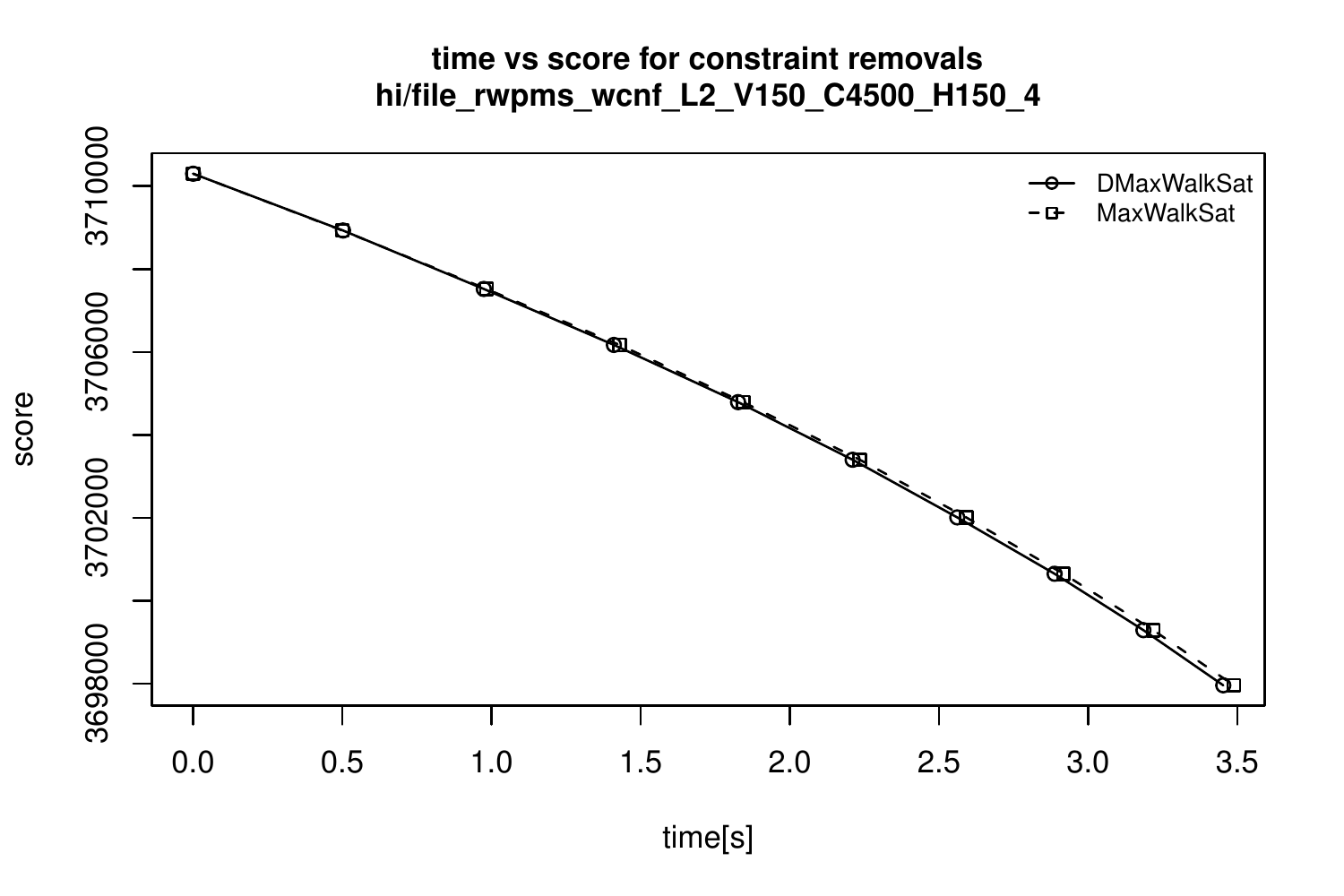}
        }

    \caption*{hi/file\_rwpms\_wcnf\_L2\_V150\_C4500\_H150\_4}
    \label{fig_hi/file_rwpms_wcnf_L2_V150_C4500_H150_4}
\end{figure}

\begin{figure}[H]
    \setcounter{subfigure}{0}
    \centering
        \subfloat[Constraint addition]
        {
            \includegraphics[width=2.7in]{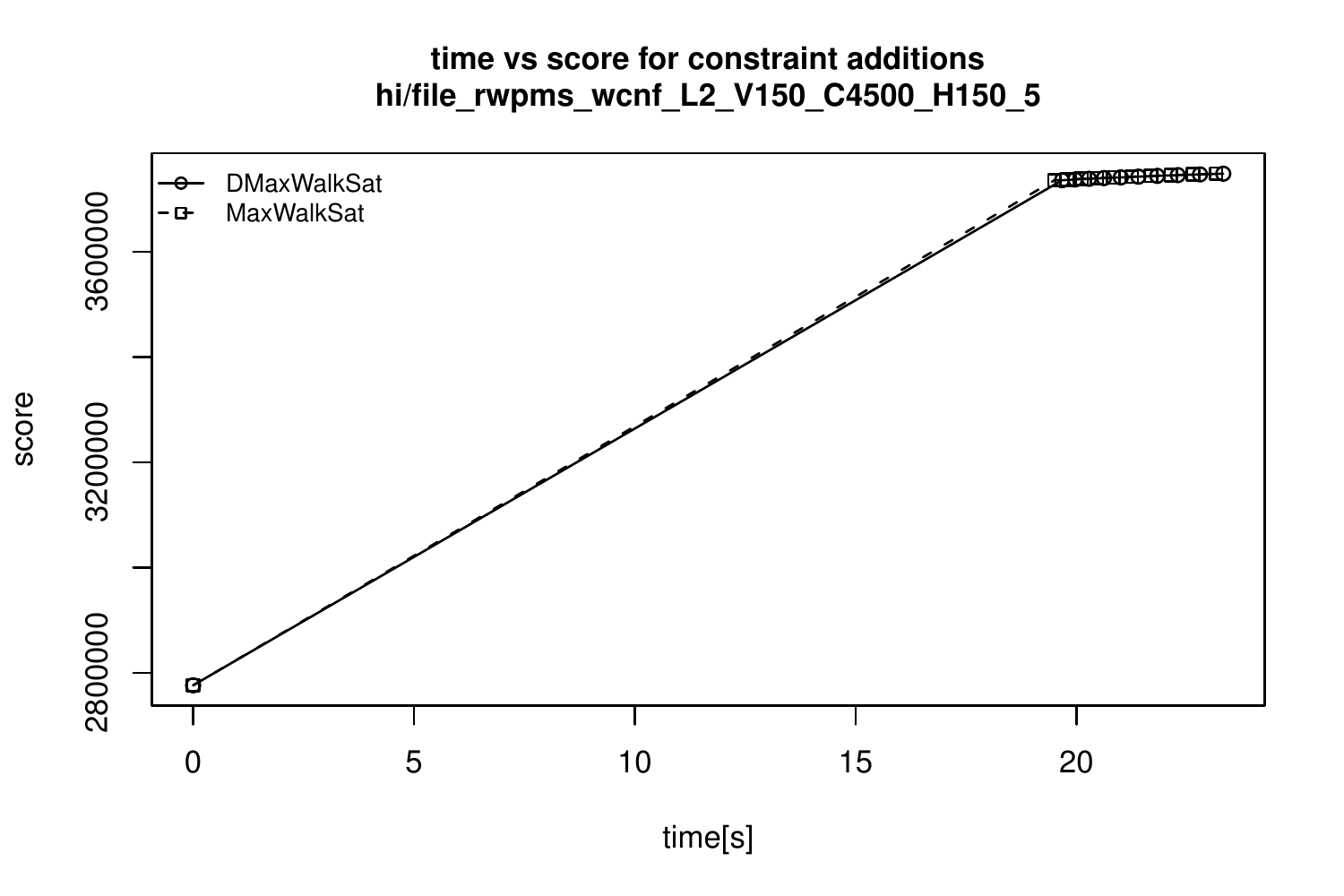}
        }
        \qquad
        \subfloat[Constraint removal]
        {
            \includegraphics[width=2.7in]{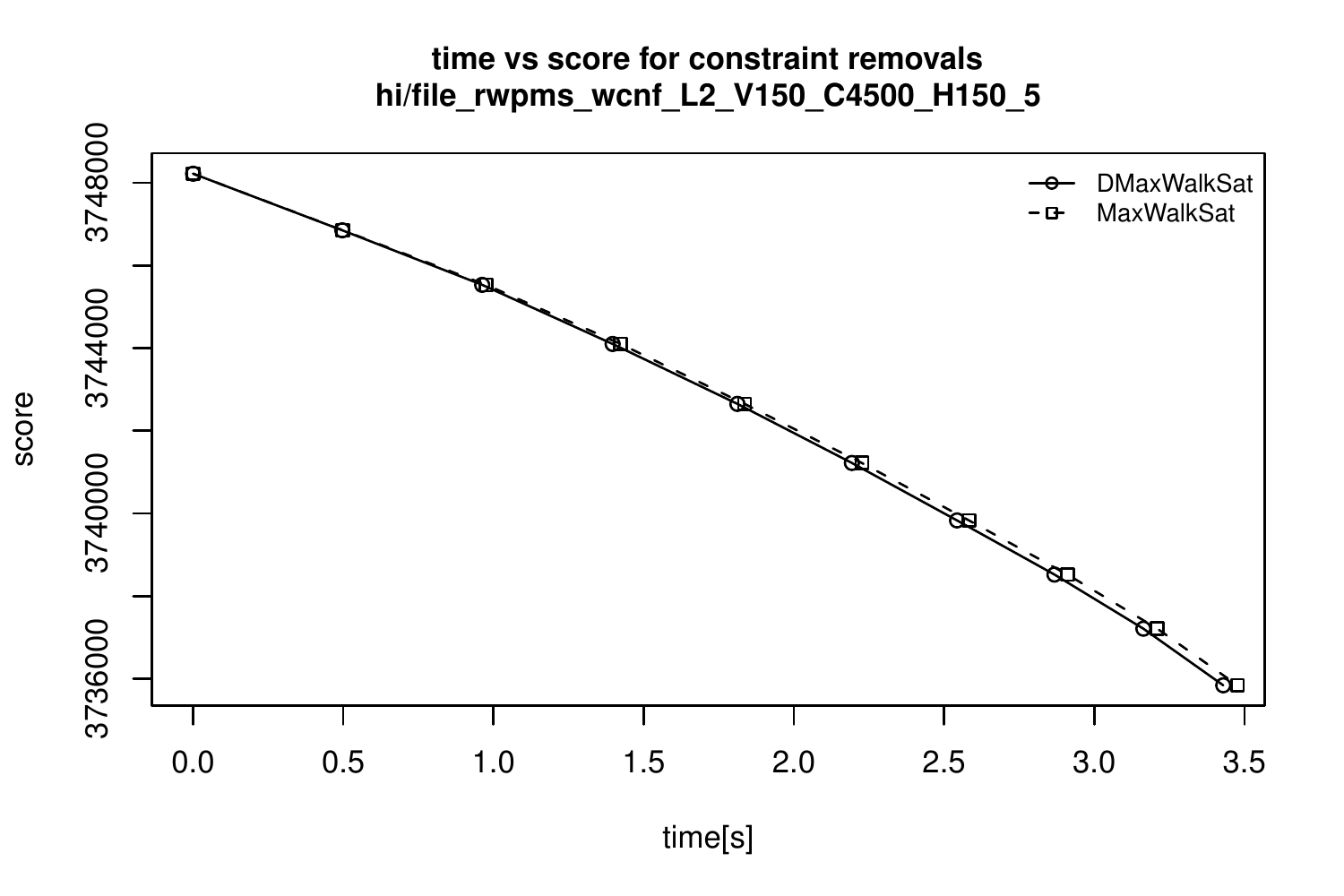}
        }

    \caption*{hi/file\_rwpms\_wcnf\_L2\_V150\_C4500\_H150\_5}
    \label{fig_hi/file_rwpms_wcnf_L2_V150_C4500_H150_5}
\end{figure}

\begin{figure}[H]
    \setcounter{subfigure}{0}
    \centering
        \subfloat[Constraint addition]
        {
            \includegraphics[width=2.7in]{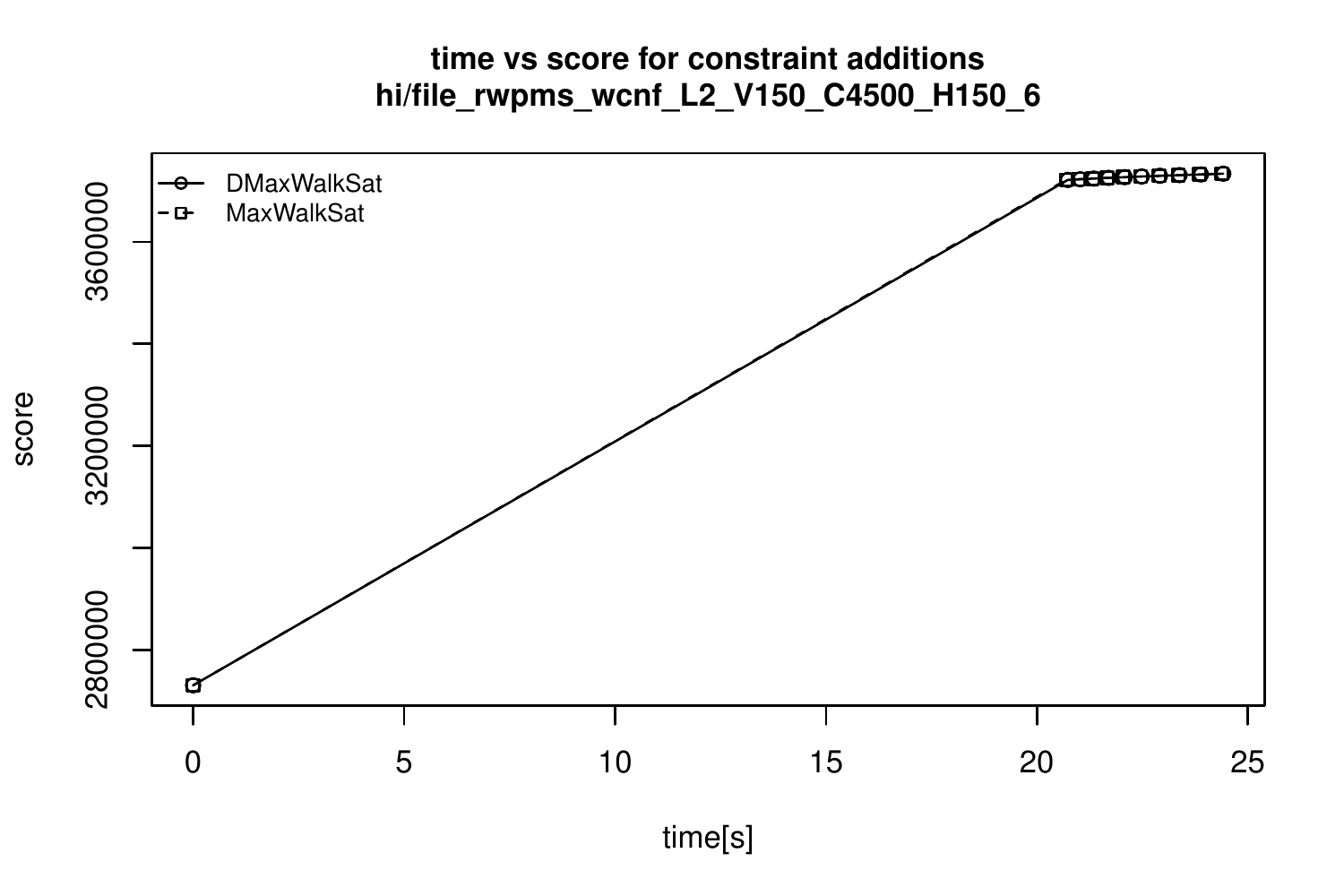}
        }
        \qquad
        \subfloat[Constraint removal]
        {
            \includegraphics[width=2.7in]{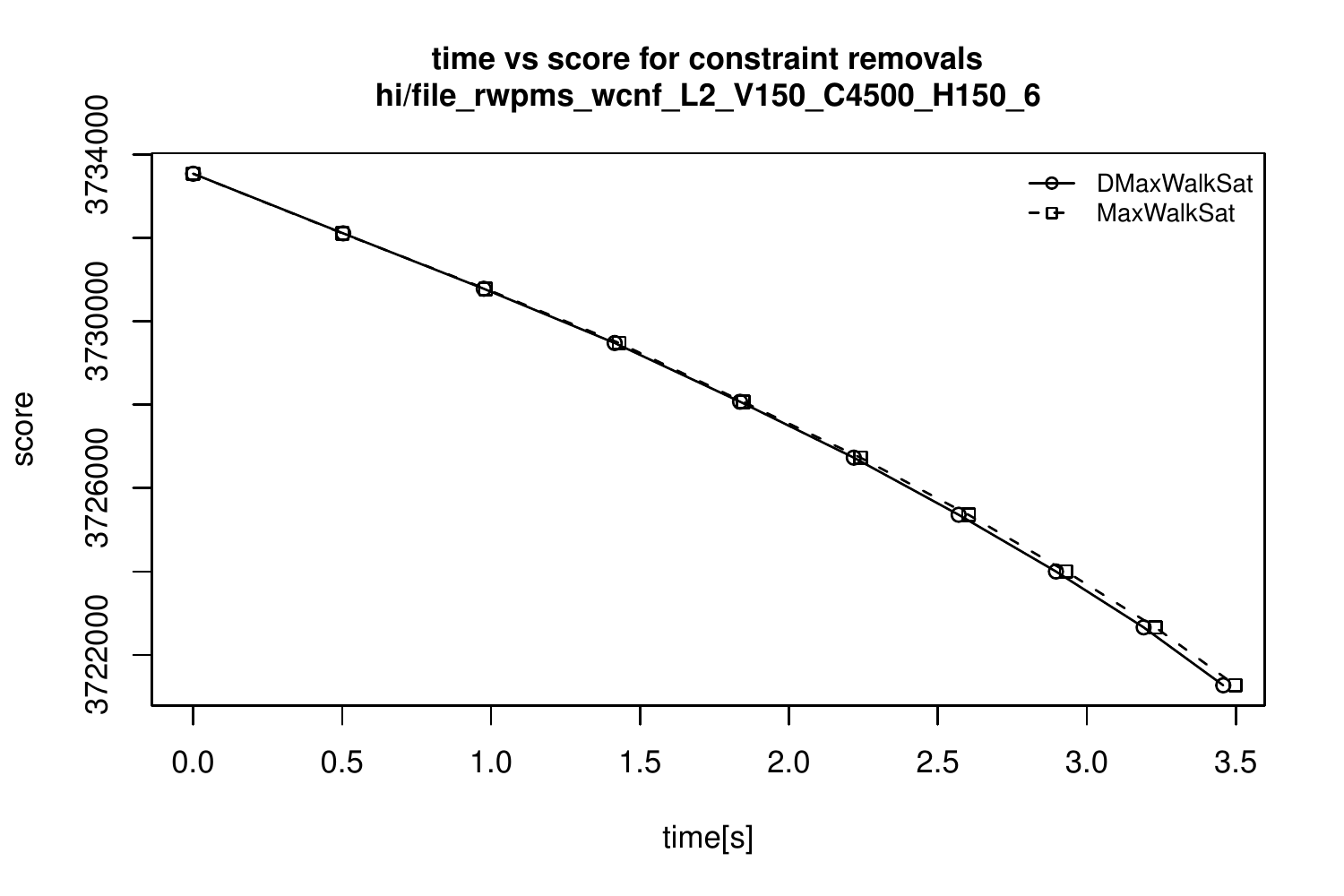}
        }

    \caption*{hi/file\_rwpms\_wcnf\_L2\_V150\_C4500\_H150\_6}
    \label{fig_hi/file_rwpms_wcnf_L2_V150_C4500_H150_6}
\end{figure}

\begin{figure}[H]
    \setcounter{subfigure}{0}
    \centering
        \subfloat[Constraint addition]
        {
            \includegraphics[width=2.7in]{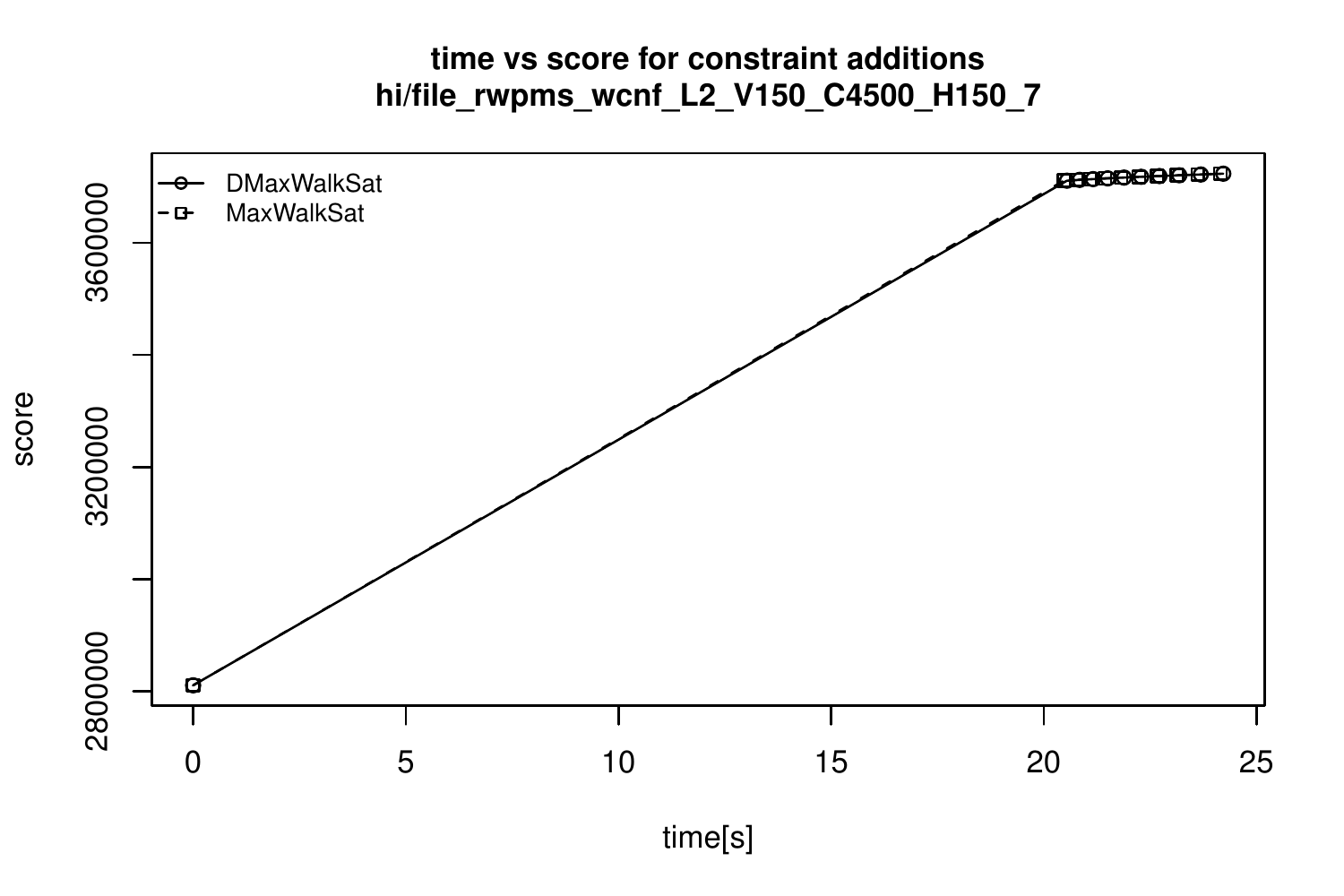}
        }
        \qquad
        \subfloat[Constraint removal]
        {
            \includegraphics[width=2.7in]{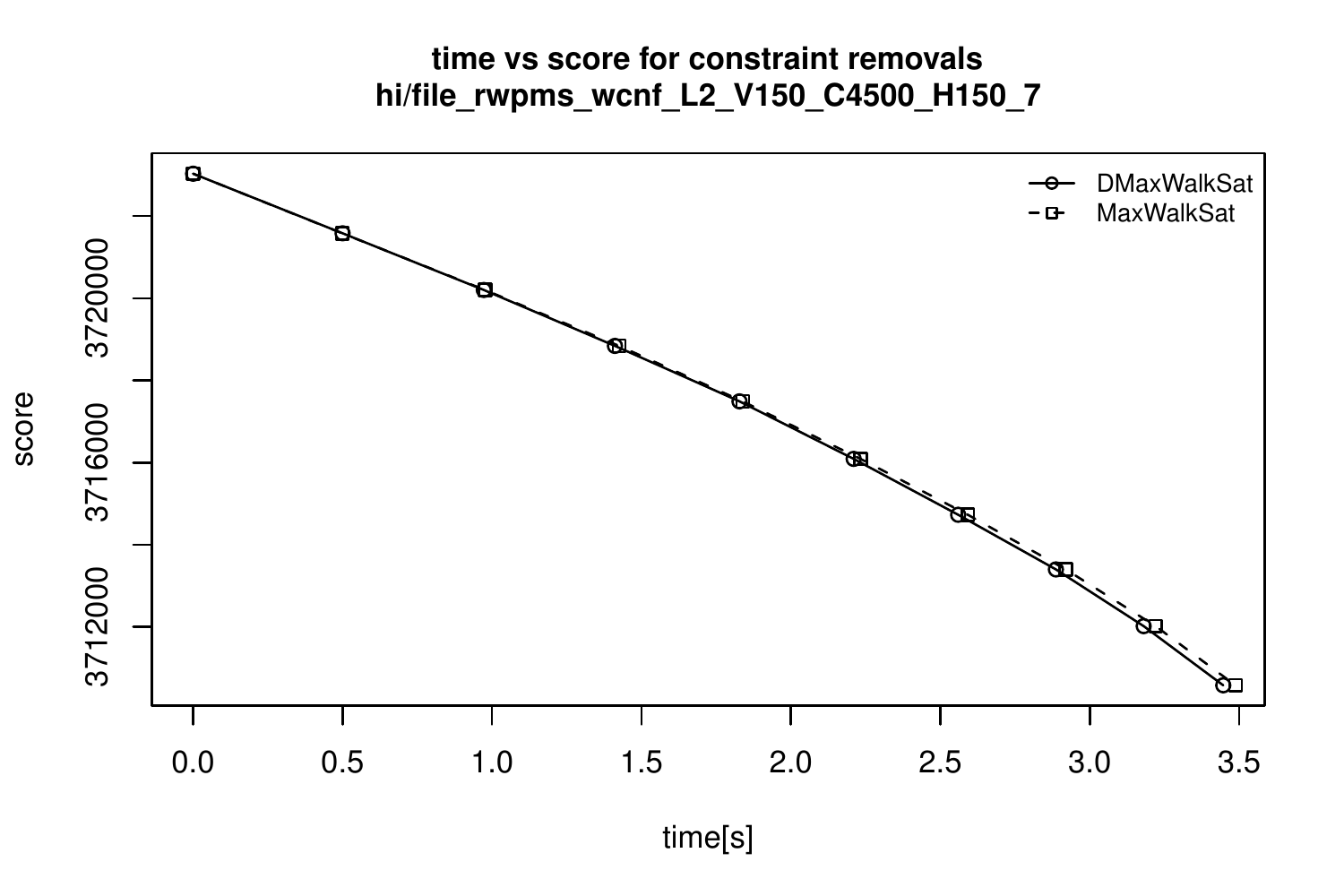}
        }

    \caption*{hi/file\_rwpms\_wcnf\_L2\_V150\_C4500\_H150\_7}
    \label{fig_hi/file_rwpms_wcnf_L2_V150_C4500_H150_7}
\end{figure}

\begin{figure}[H]
    \setcounter{subfigure}{0}
    \centering
        \subfloat[Constraint addition]
        {
            \includegraphics[width=2.7in]{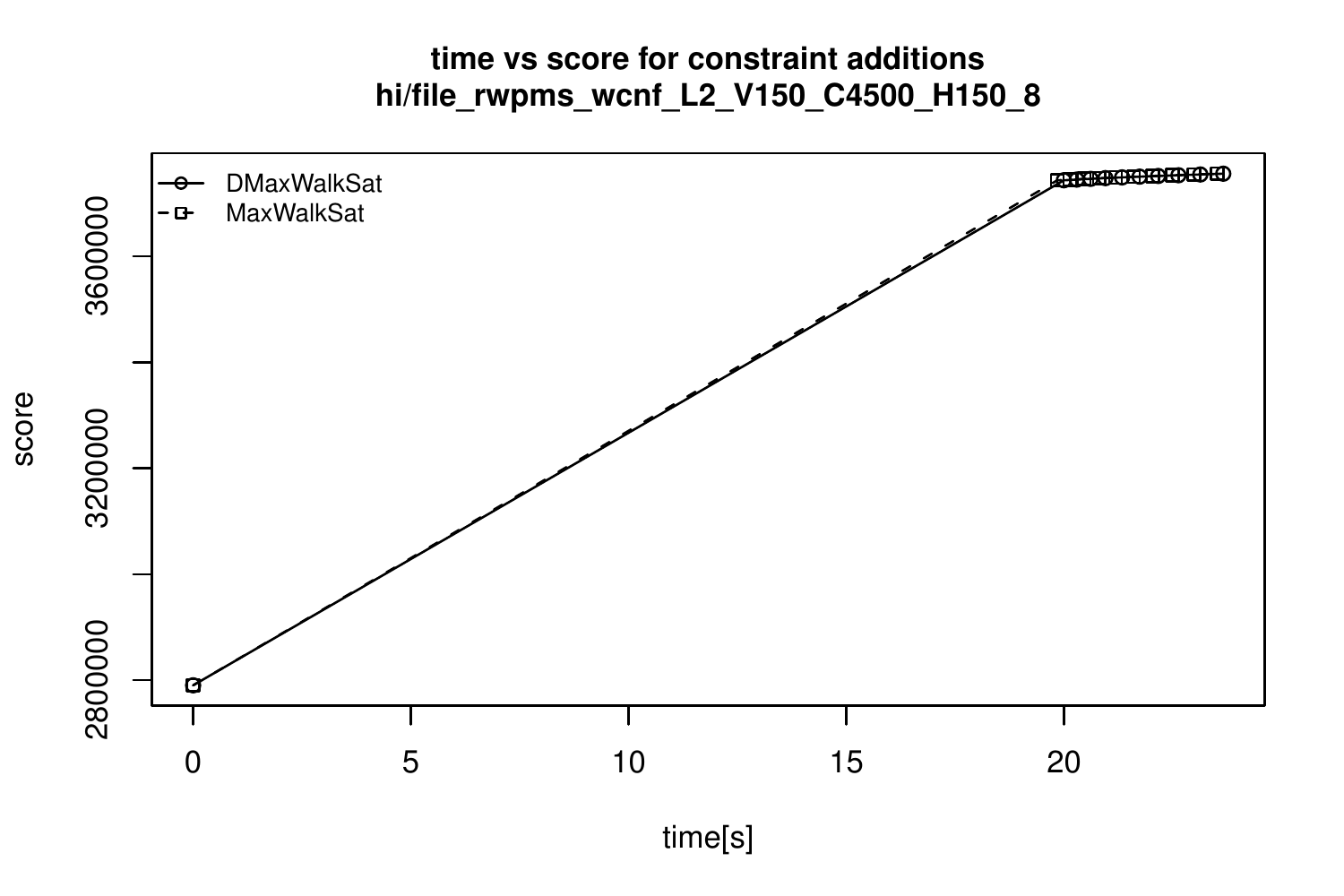}
        }
        \qquad
        \subfloat[Constraint removal]
        {
            \includegraphics[width=2.7in]{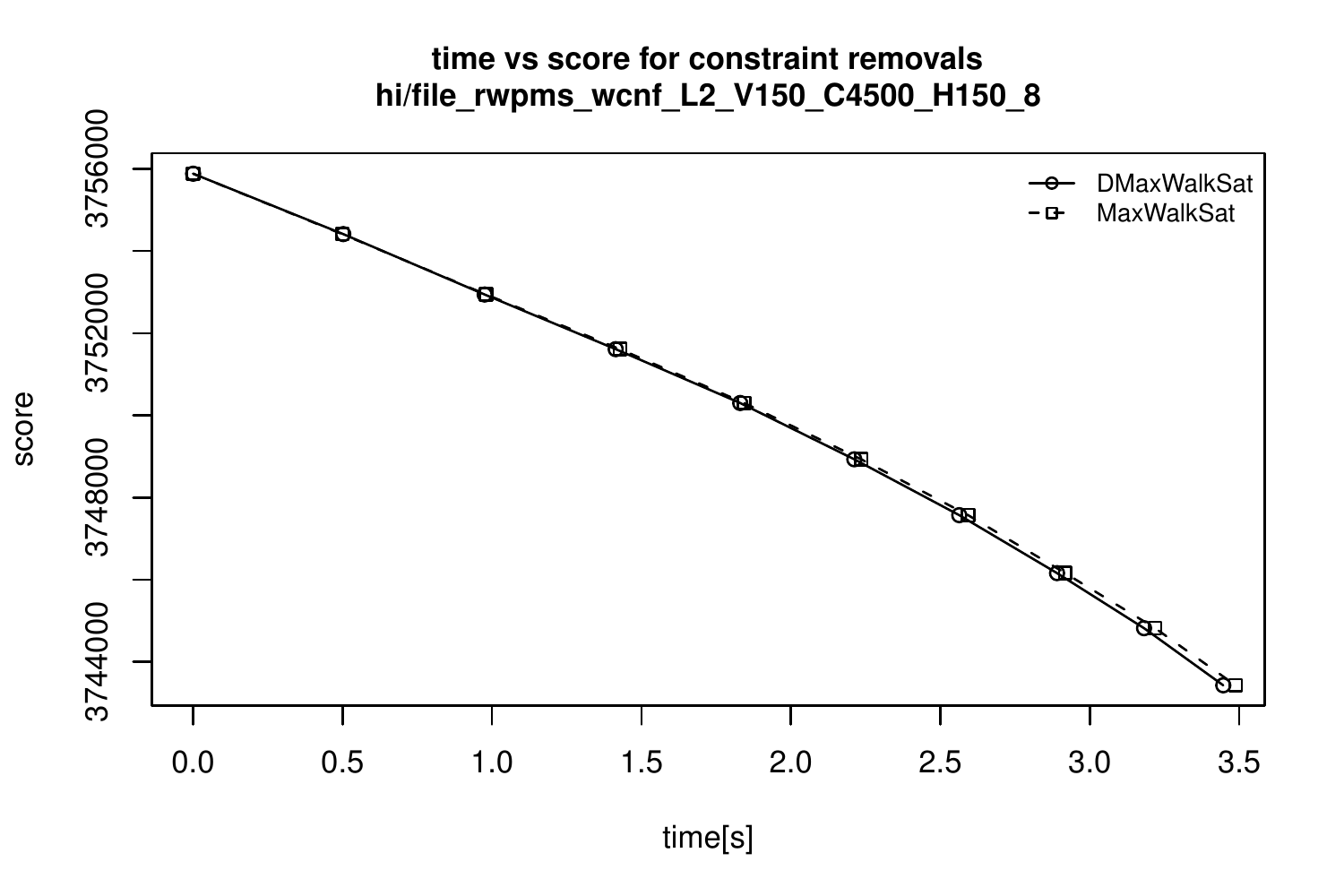}
        }

    \caption*{hi/file\_rwpms\_wcnf\_L2\_V150\_C4500\_H150\_8}
    \label{fig_hi/file_rwpms_wcnf_L2_V150_C4500_H150_8}
\end{figure}

\begin{figure}[H]
    \setcounter{subfigure}{0}
    \centering
        \subfloat[Constraint addition]
        {
            \includegraphics[width=2.7in]{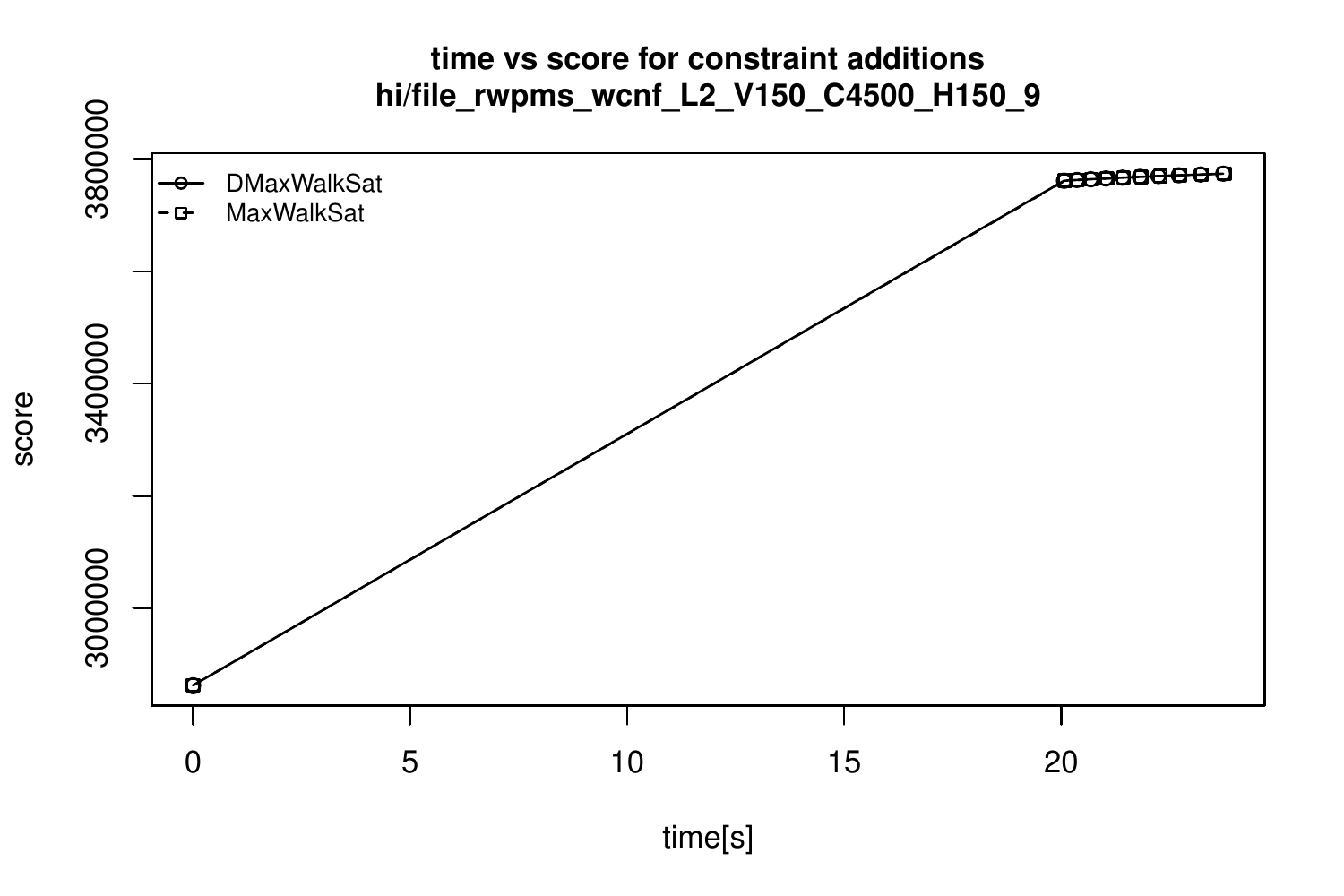}
        }
        \qquad
        \subfloat[Constraint removal]
        {
            \includegraphics[width=2.7in]{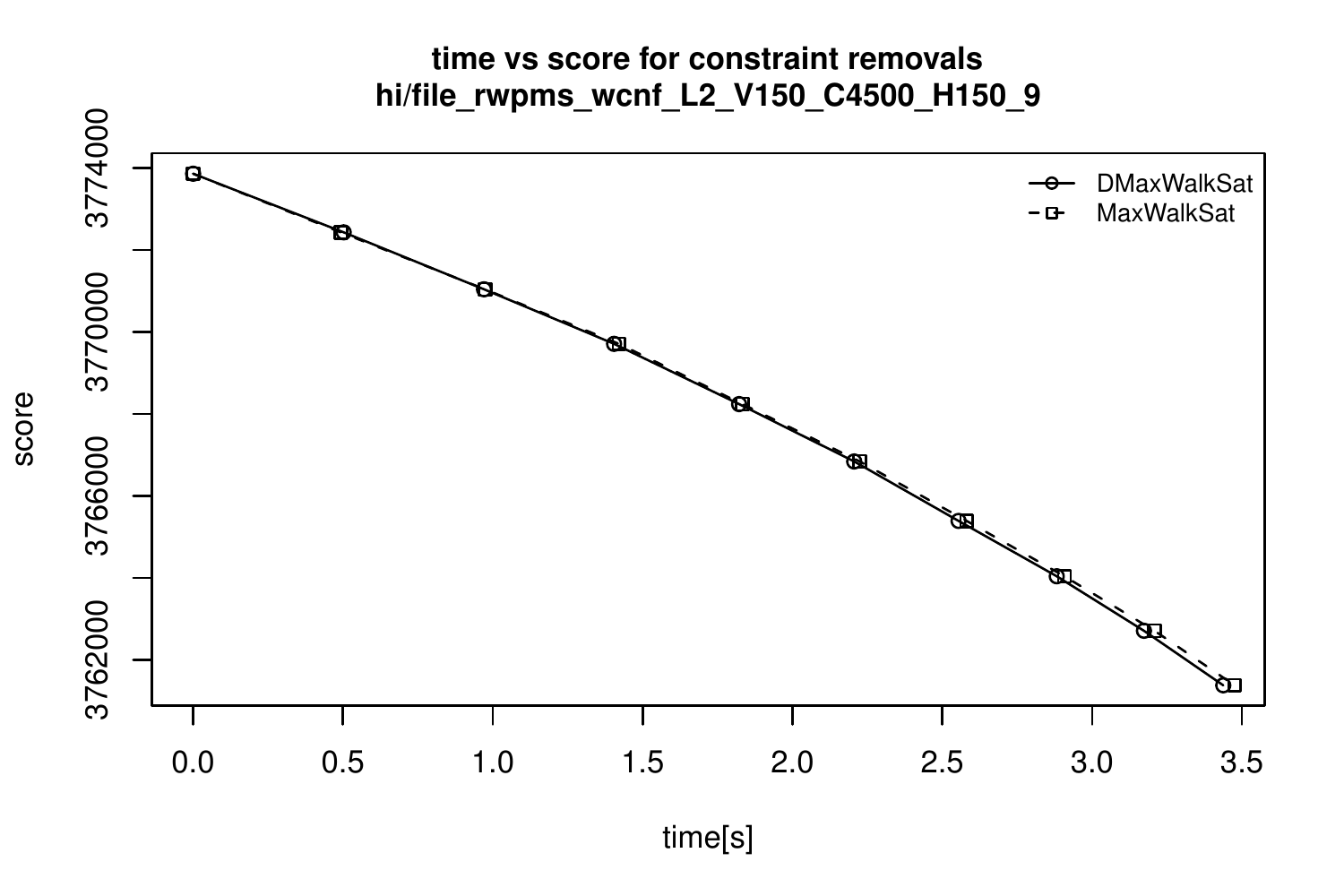}
        }

    \caption*{hi/file\_rwpms\_wcnf\_L2\_V150\_C4500\_H150\_9}
    \label{fig_hi/file_rwpms_wcnf_L2_V150_C4500_H150_9}
\end{figure}

\begin{figure}[H]
    \setcounter{subfigure}{0}
    \centering
        \subfloat[Constraint addition]
        {
            \includegraphics[width=2.7in]{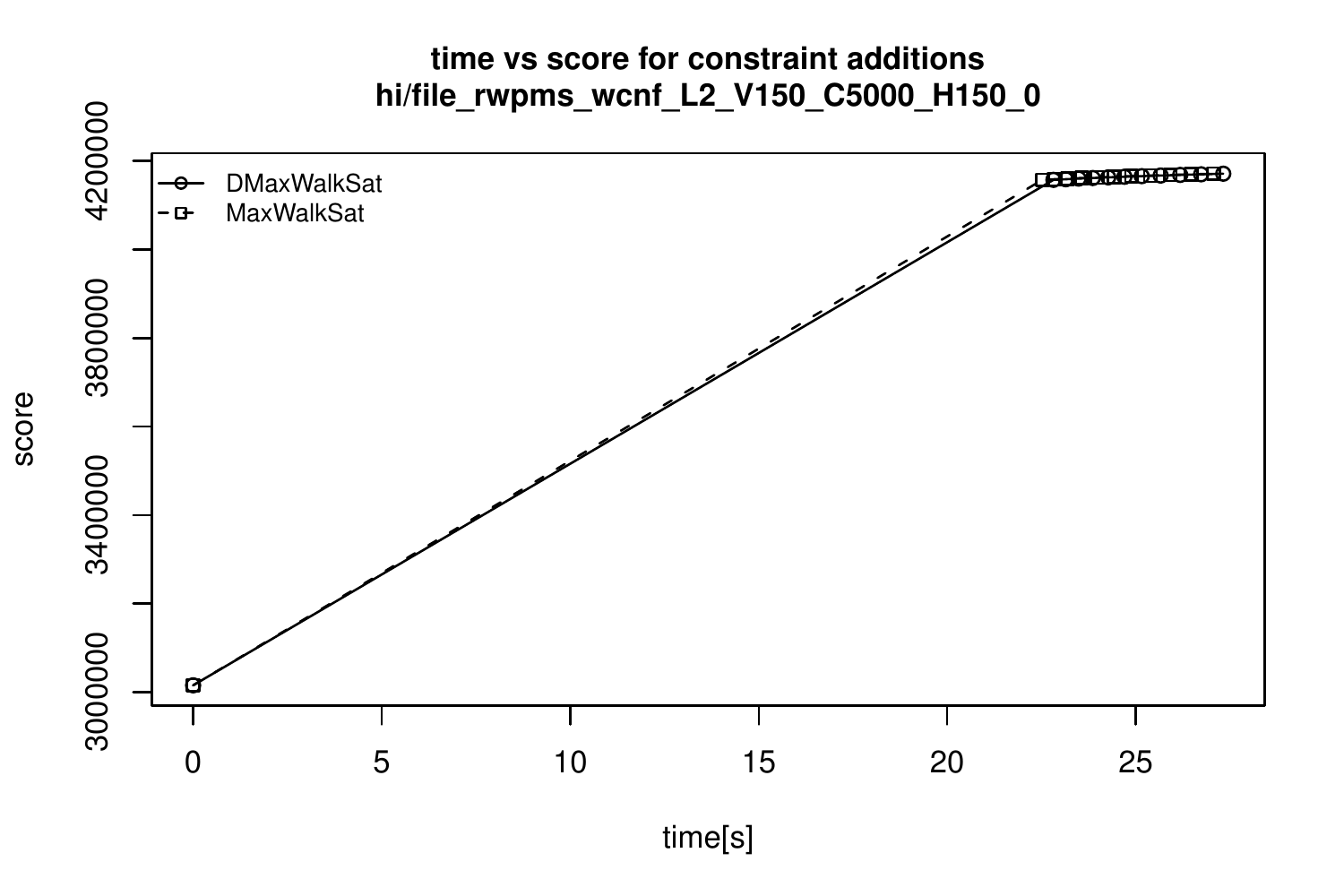}
        }
        \qquad
        \subfloat[Constraint removal]
        {
            \includegraphics[width=2.7in]{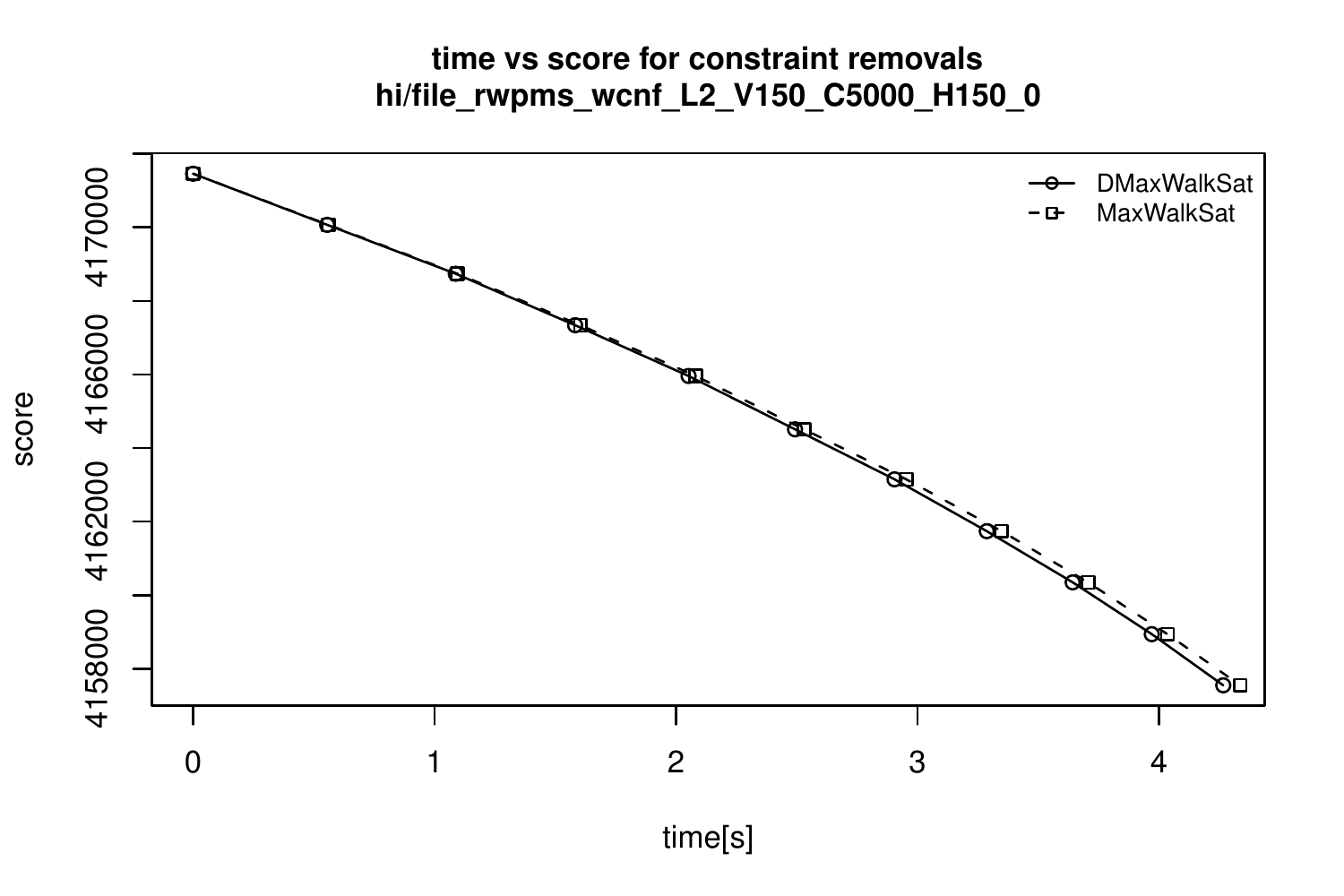}
        }

    \caption*{hi/file\_rwpms\_wcnf\_L2\_V150\_C5000\_H150\_0}
    \label{fig_hi/file_rwpms_wcnf_L2_V150_C5000_H150_0}
\end{figure}

\begin{figure}[H]
    \setcounter{subfigure}{0}
    \centering
        \subfloat[Constraint addition]
        {
            \includegraphics[width=2.7in]{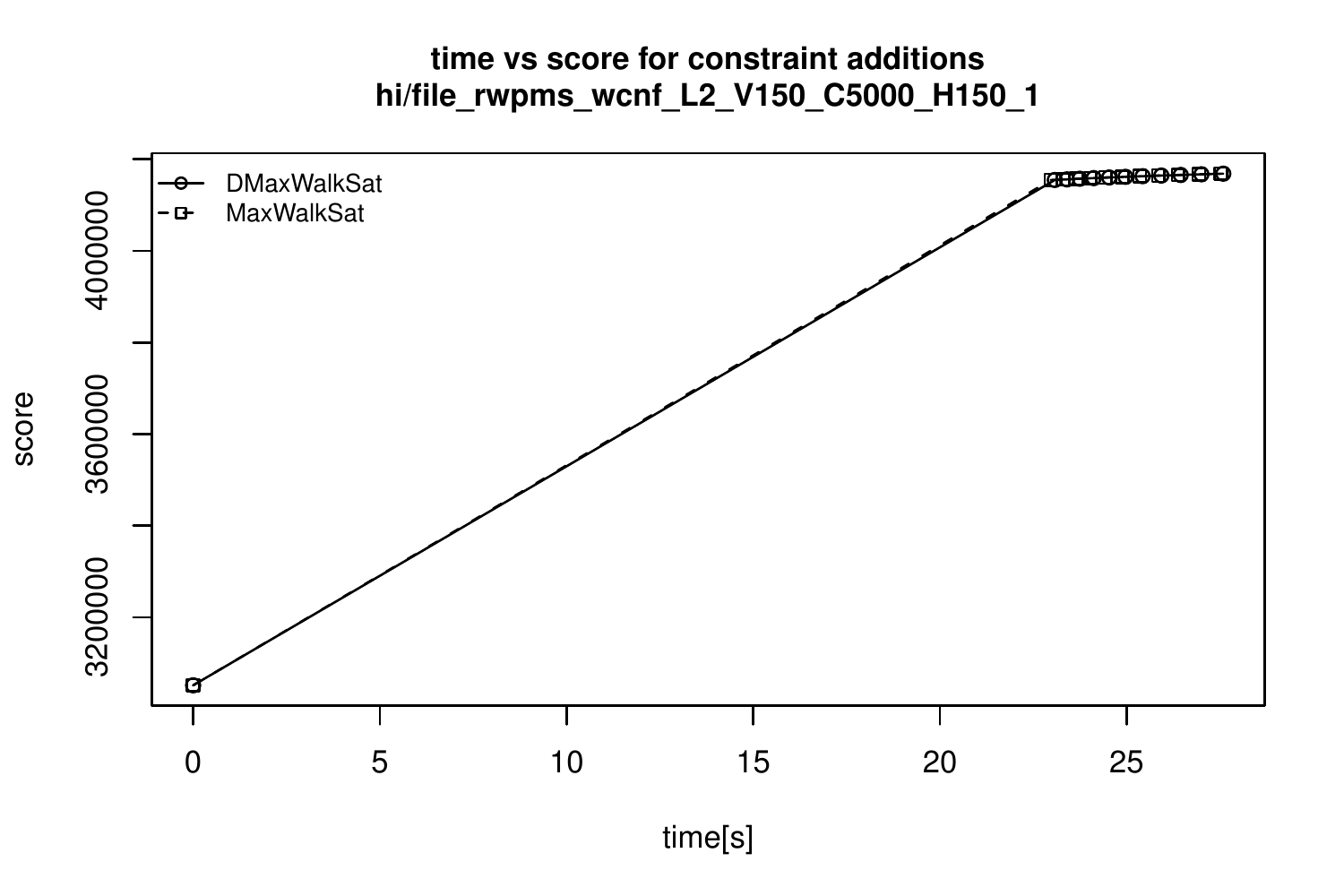}
        }
        \qquad
        \subfloat[Constraint removal]
        {
            \includegraphics[width=2.7in]{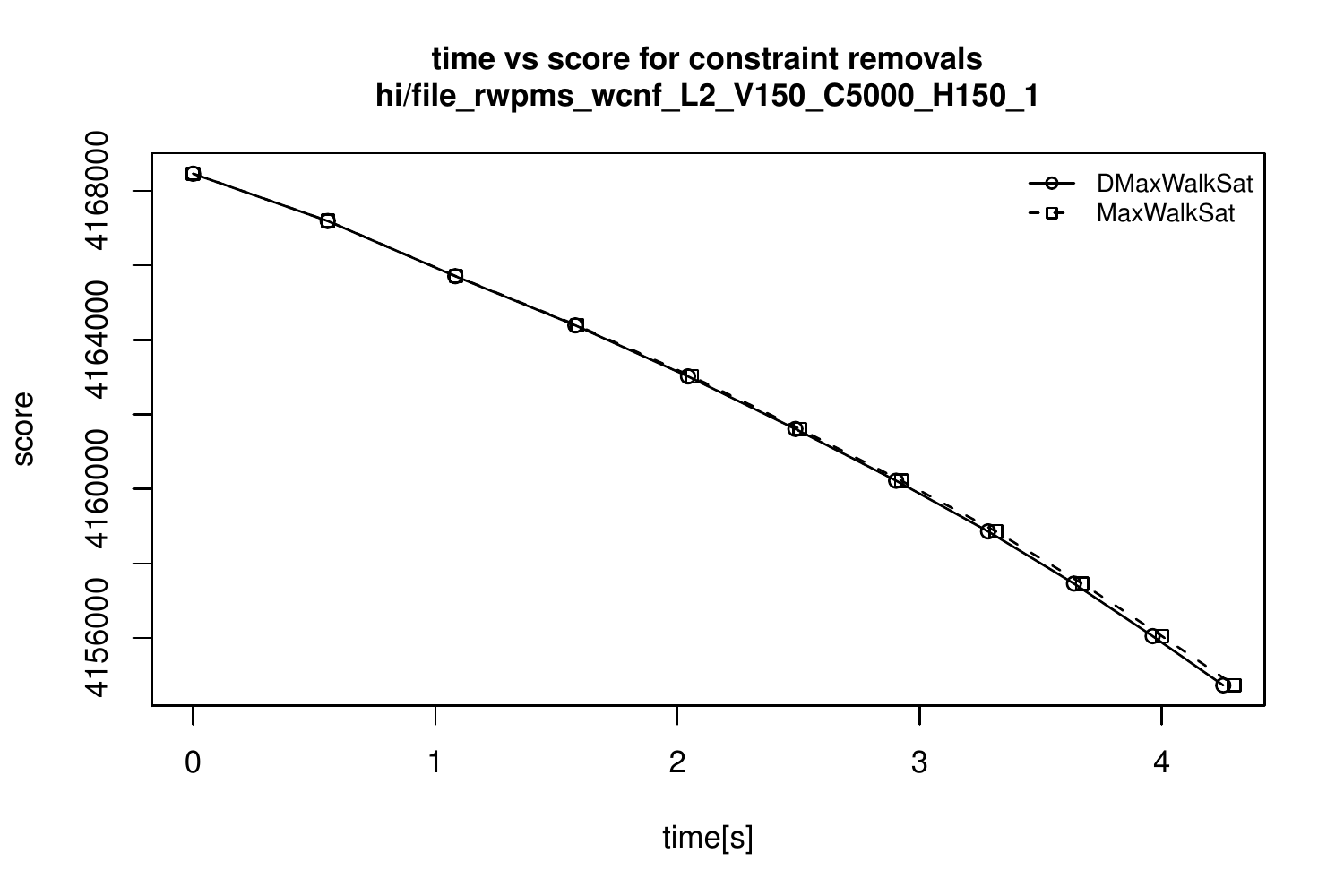}
        }

    \caption*{hi/file\_rwpms\_wcnf\_L2\_V150\_C5000\_H150\_1}
    \label{fig_hi/file_rwpms_wcnf_L2_V150_C5000_H150_1}
\end{figure}

\begin{figure}[H]
    \setcounter{subfigure}{0}
    \centering
        \subfloat[Constraint addition]
        {
            \includegraphics[width=2.7in]{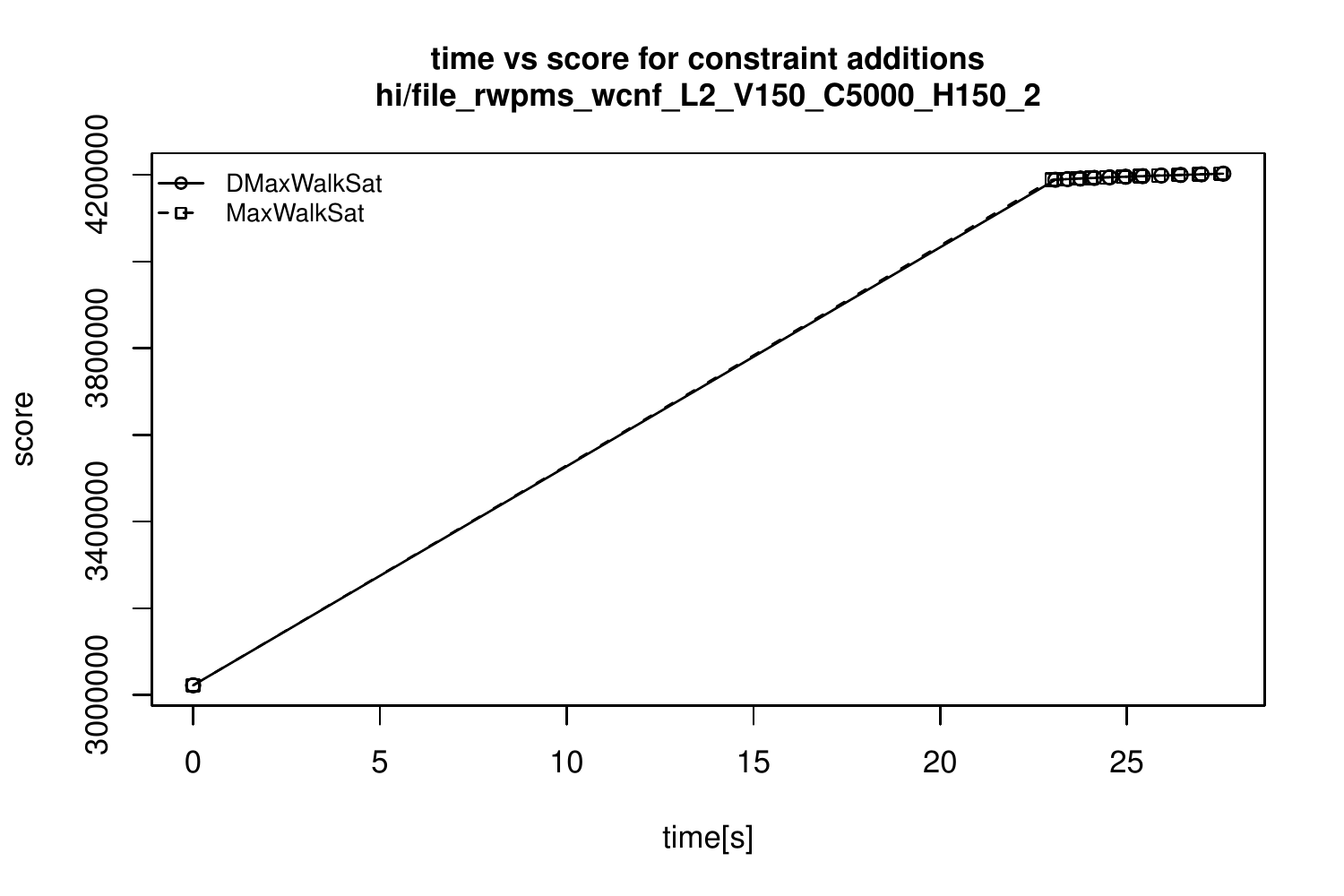}
        }
        \qquad
        \subfloat[Constraint removal]
        {
            \includegraphics[width=2.7in]{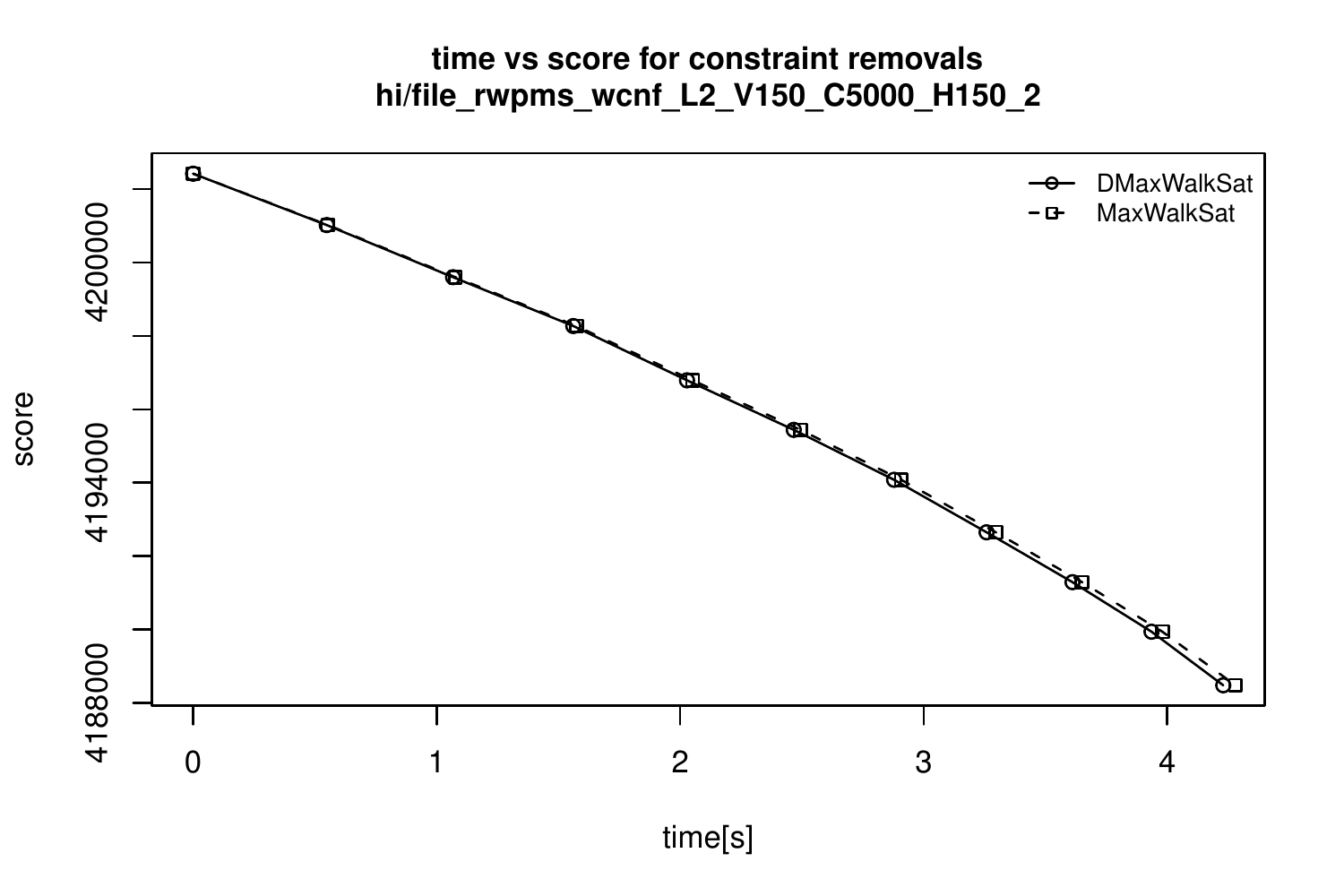}
        }

    \caption*{hi/file\_rwpms\_wcnf\_L2\_V150\_C5000\_H150\_2}
    \label{fig_hi/file_rwpms_wcnf_L2_V150_C5000_H150_2}
\end{figure}

\begin{figure}[H]
    \setcounter{subfigure}{0}
    \centering
        \subfloat[Constraint addition]
        {
            \includegraphics[width=2.7in]{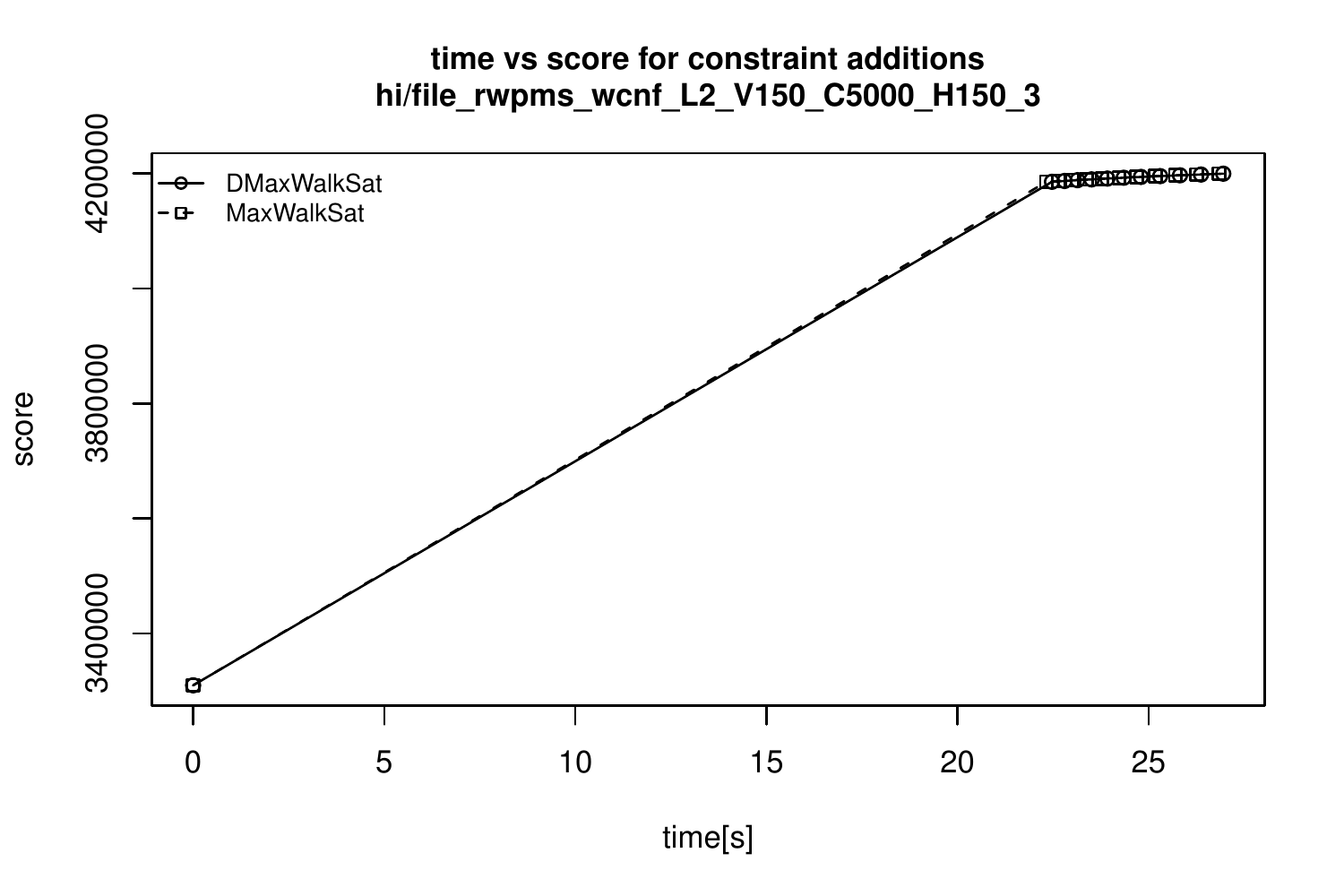}
        }
        \qquad
        \subfloat[Constraint removal]
        {
            \includegraphics[width=2.7in]{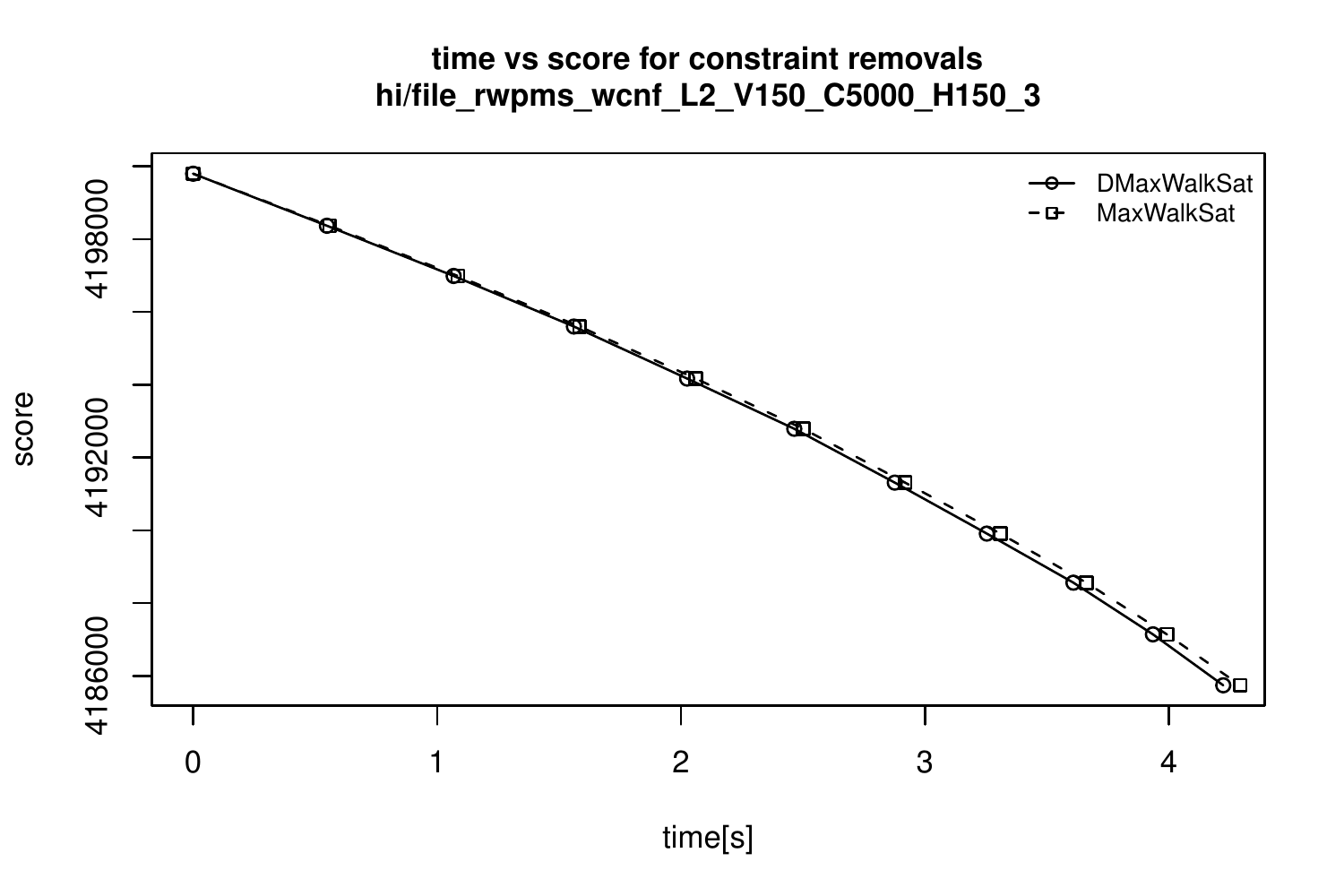}
        }

    \caption*{hi/file\_rwpms\_wcnf\_L2\_V150\_C5000\_H150\_3}
    \label{fig_hi/file_rwpms_wcnf_L2_V150_C5000_H150_3}
\end{figure}

\begin{figure}[H]
    \setcounter{subfigure}{0}
    \centering
        \subfloat[Constraint addition]
        {
            \includegraphics[width=2.7in]{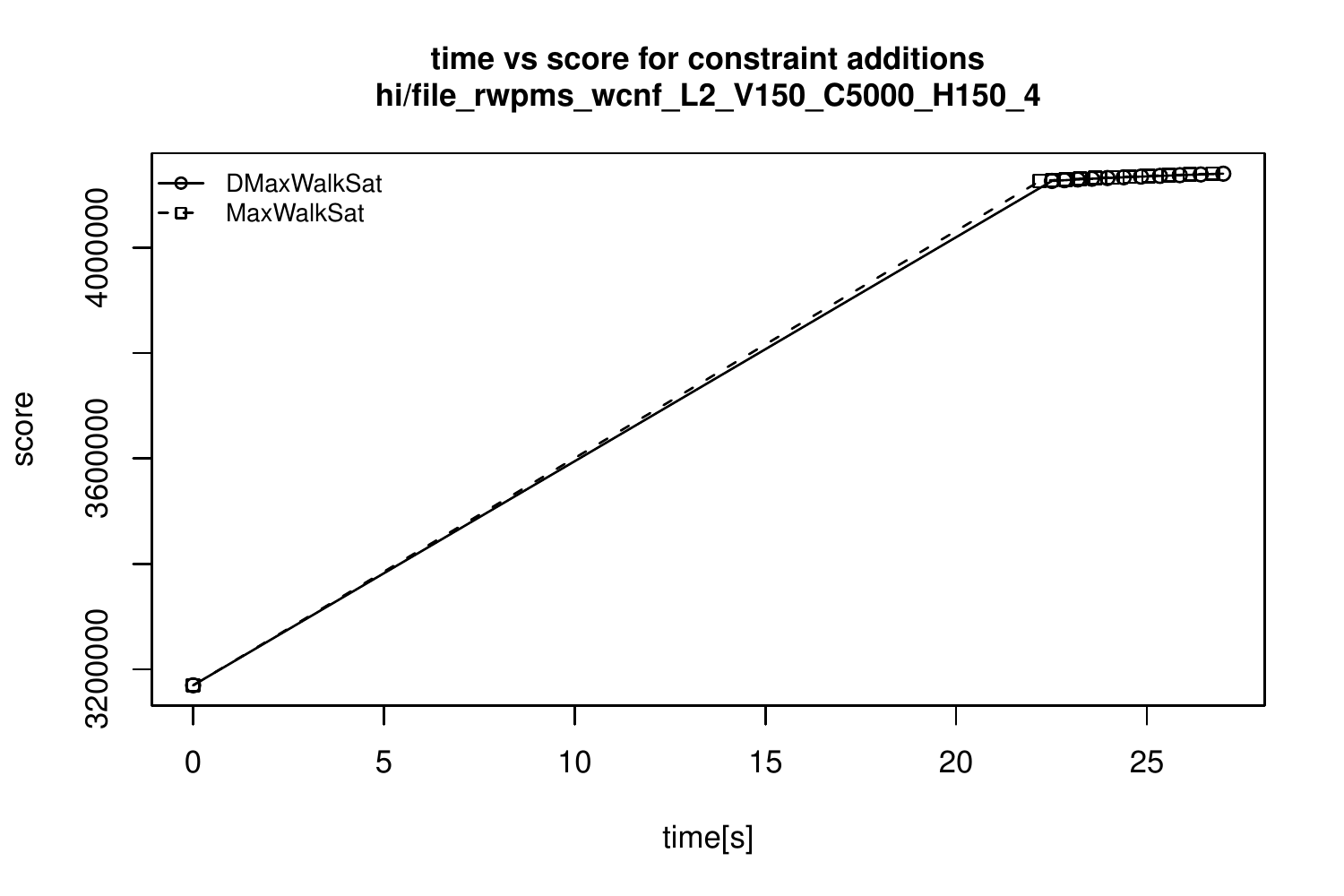}
        }
        \qquad
        \subfloat[Constraint removal]
        {
            \includegraphics[width=2.7in]{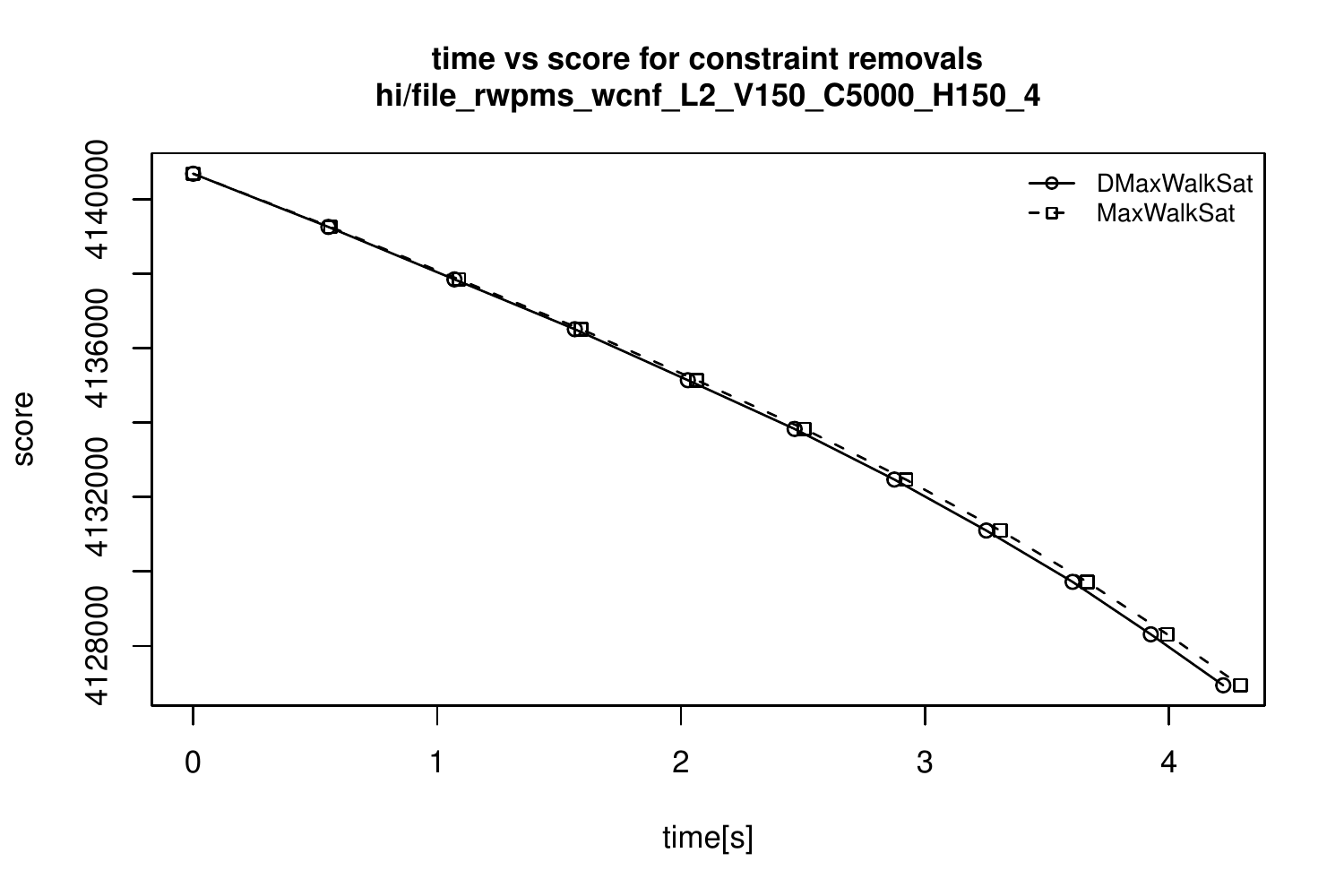}
        }

    \caption*{hi/file\_rwpms\_wcnf\_L2\_V150\_C5000\_H150\_4}
    \label{fig_hi/file_rwpms_wcnf_L2_V150_C5000_H150_4}
\end{figure}

\begin{figure}[H]
    \setcounter{subfigure}{0}
    \centering
        \subfloat[Constraint addition]
        {
            \includegraphics[width=2.7in]{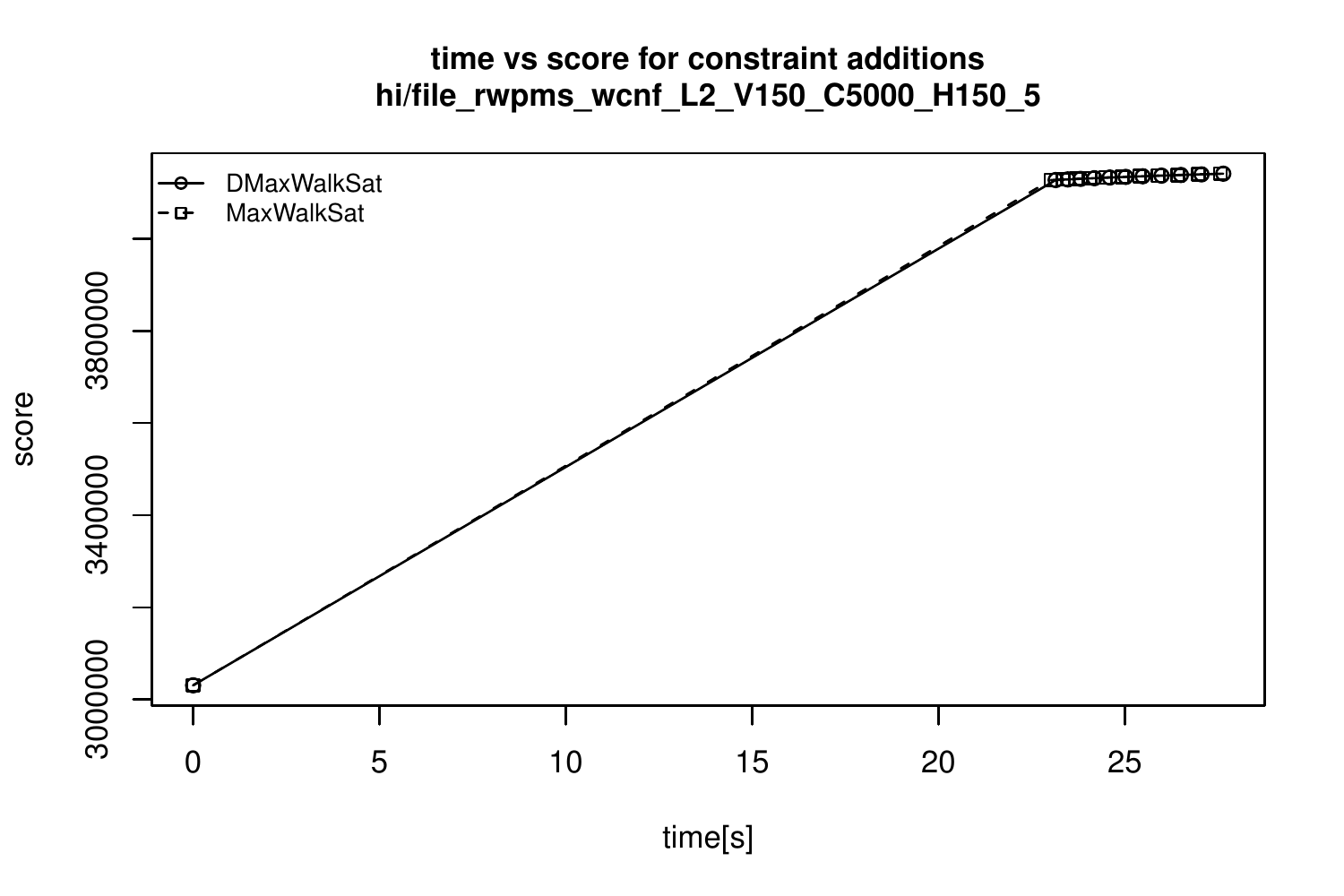}
        }
        \qquad
        \subfloat[Constraint removal]
        {
            \includegraphics[width=2.7in]{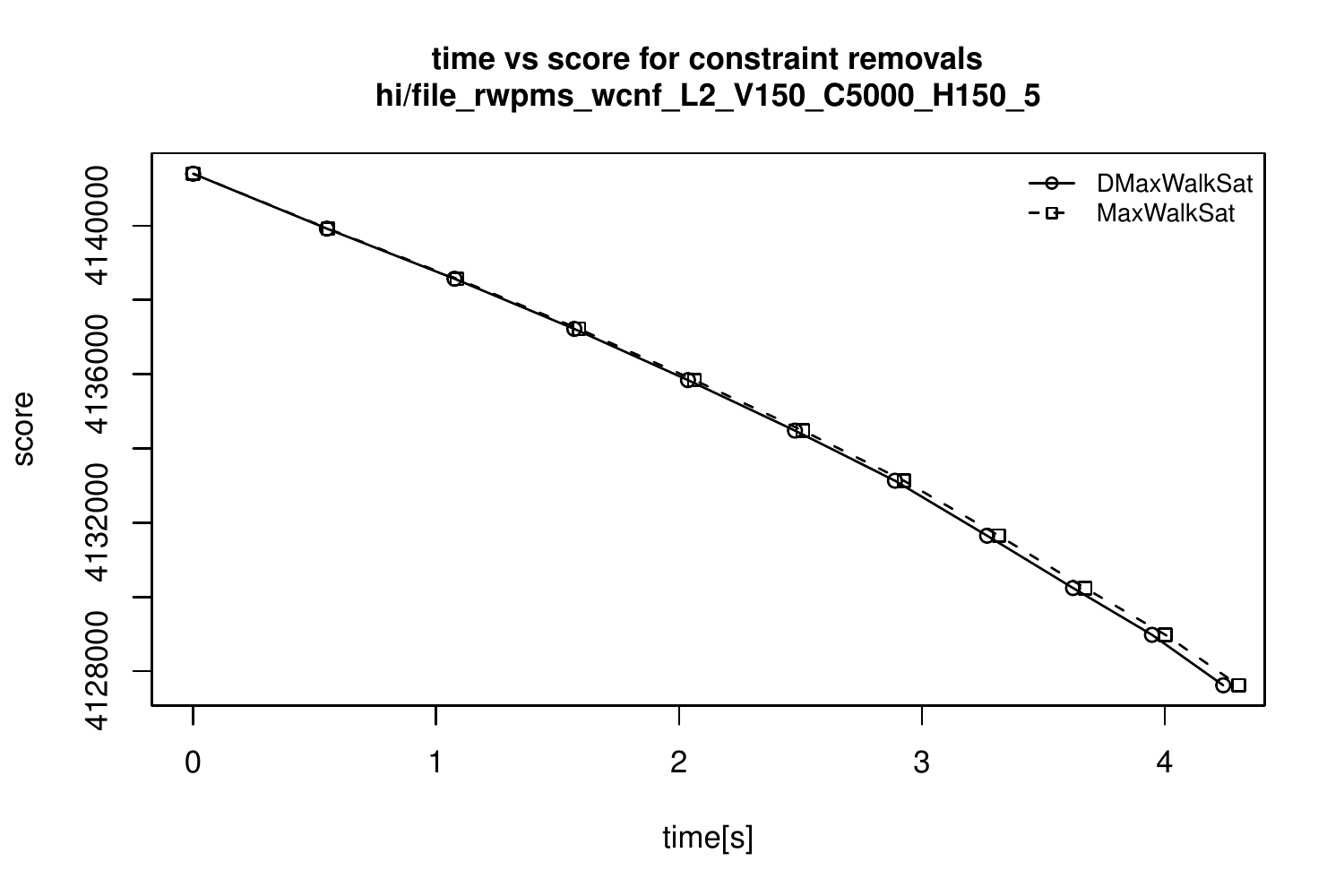}
        }

    \caption*{hi/file\_rwpms\_wcnf\_L2\_V150\_C5000\_H150\_5}
    \label{fig_hi/file_rwpms_wcnf_L2_V150_C5000_H150_5}
\end{figure}

\begin{figure}[H]
    \setcounter{subfigure}{0}
    \centering
        \subfloat[Constraint addition]
        {
            \includegraphics[width=2.7in]{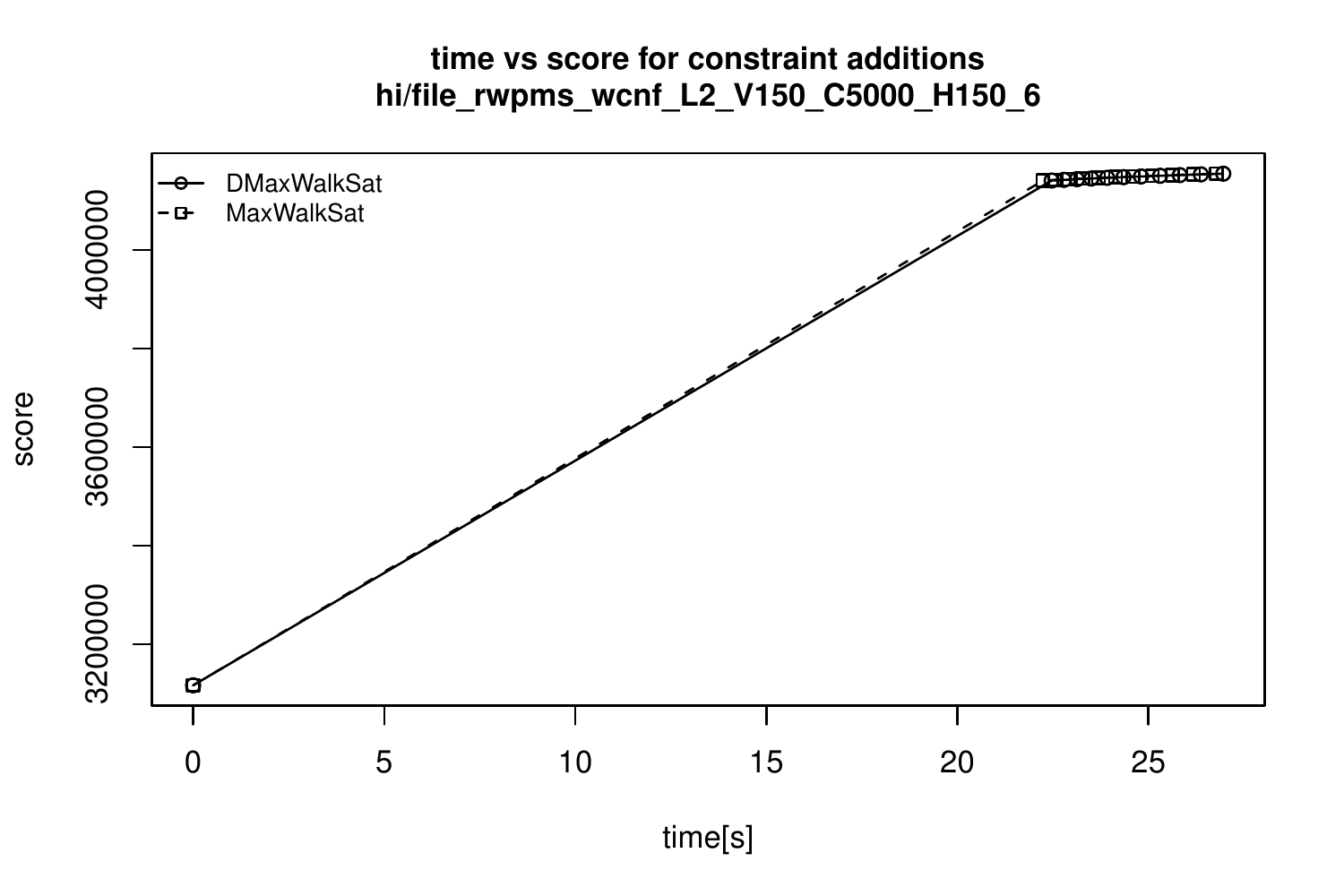}
        }
        \qquad
        \subfloat[Constraint removal]
        {
            \includegraphics[width=2.7in]{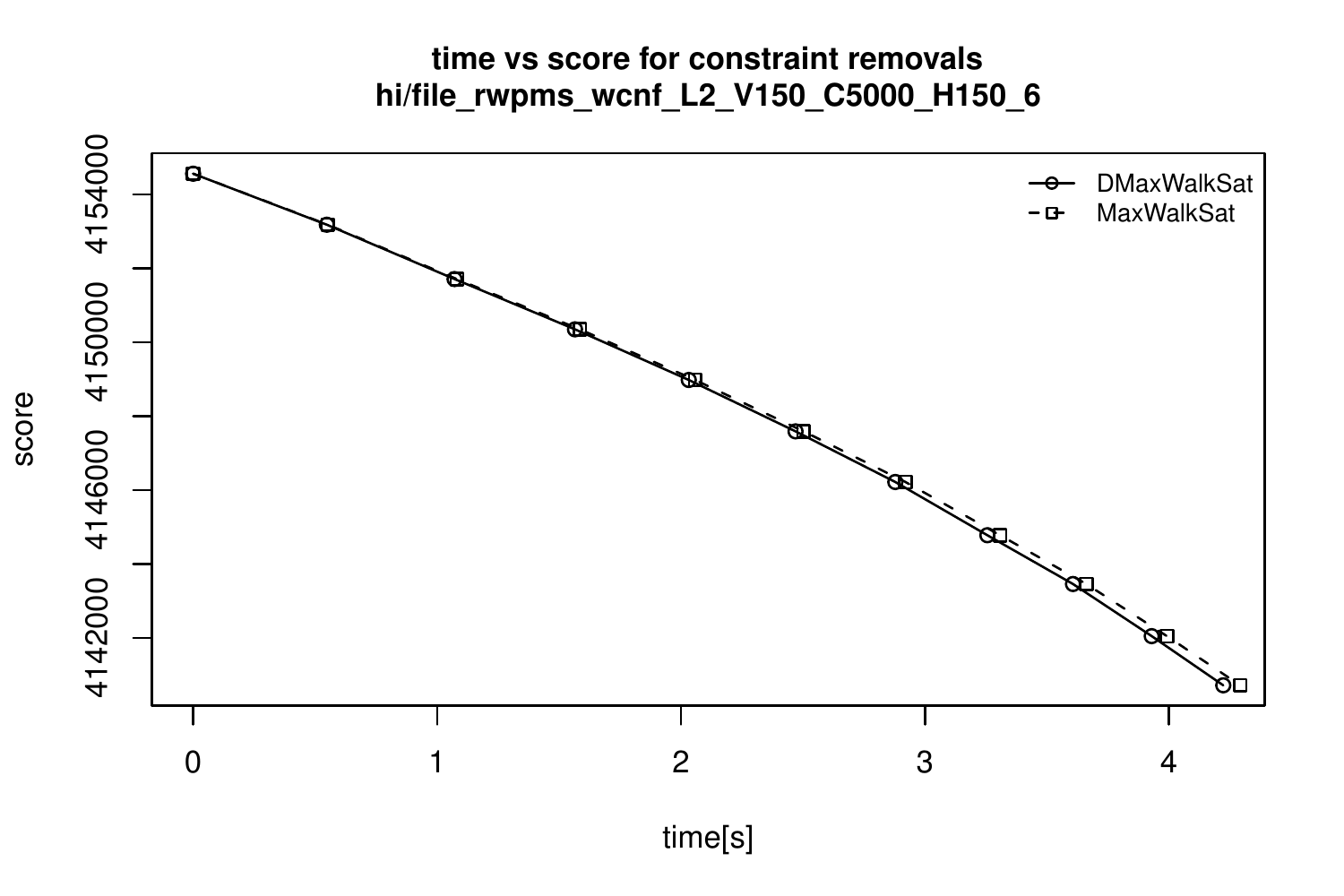}
        }

    \caption*{hi/file\_rwpms\_wcnf\_L2\_V150\_C5000\_H150\_6}
    \label{fig_hi/file_rwpms_wcnf_L2_V150_C5000_H150_6}
\end{figure}

\begin{figure}[H]
    \setcounter{subfigure}{0}
    \centering
        \subfloat[Constraint addition]
        {
            \includegraphics[width=2.7in]{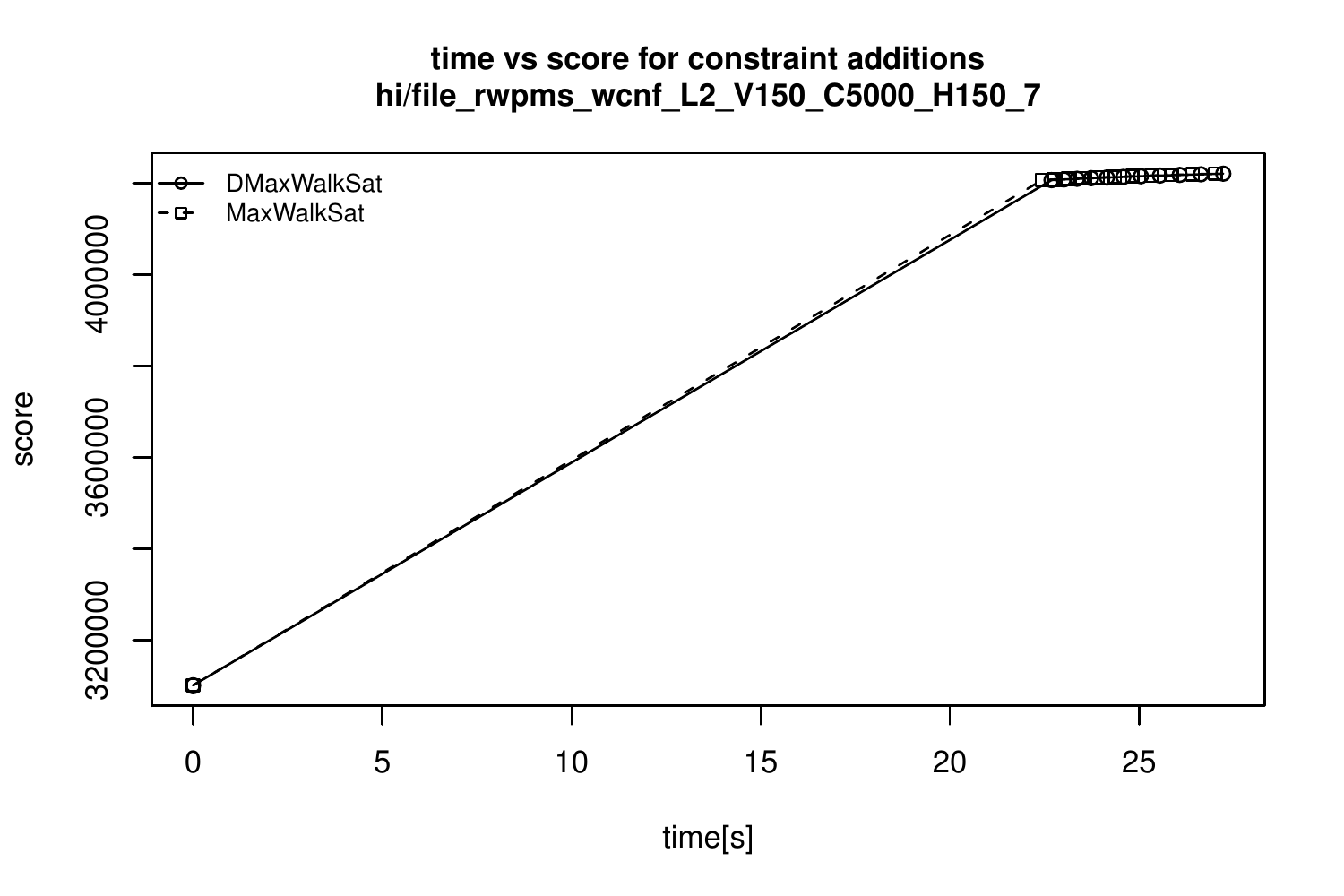}
        }
        \qquad
        \subfloat[Constraint removal]
        {
            \includegraphics[width=2.7in]{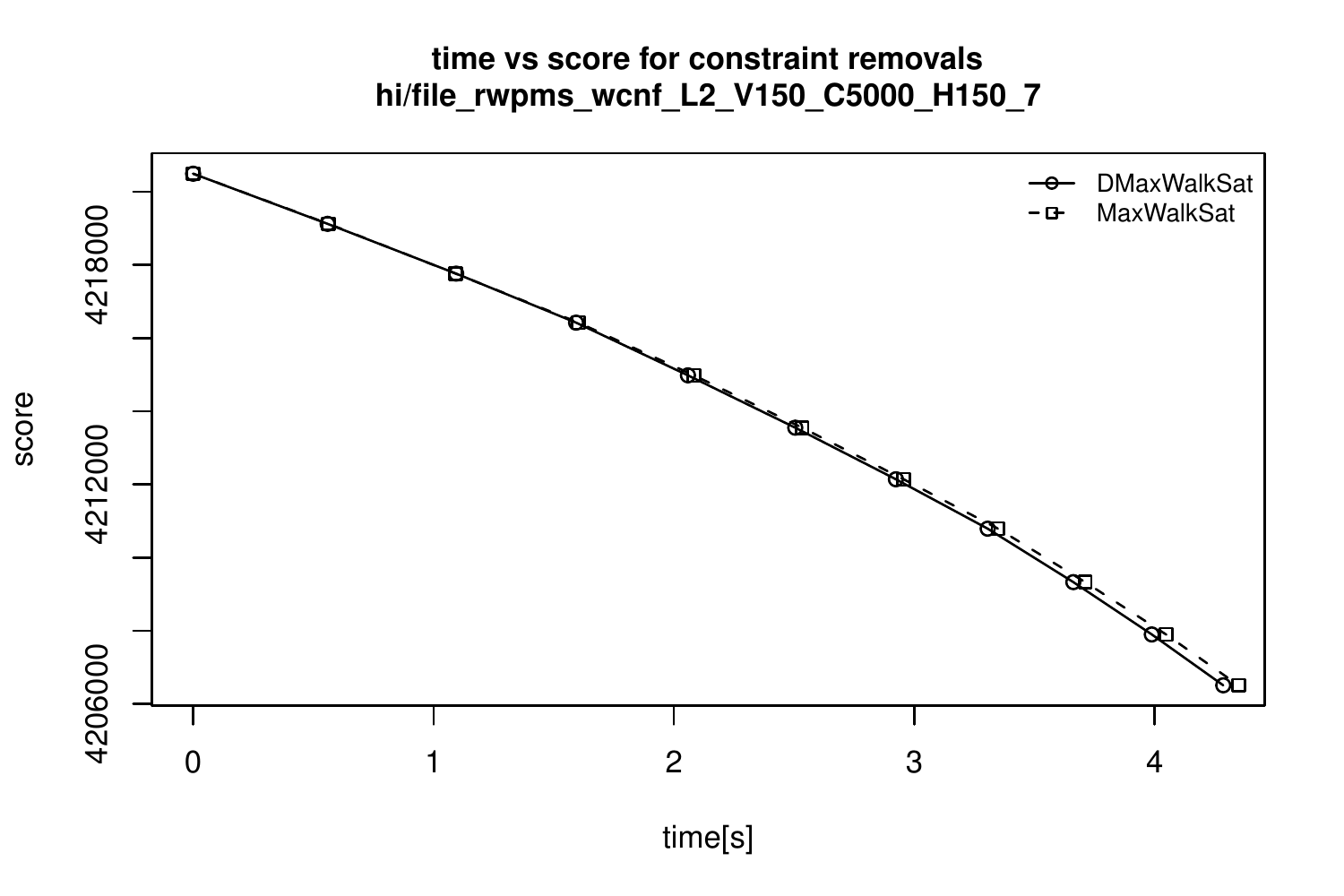}
        }

    \caption*{hi/file\_rwpms\_wcnf\_L2\_V150\_C5000\_H150\_7}
    \label{fig_hi/file_rwpms_wcnf_L2_V150_C5000_H150_7}
\end{figure}

\begin{figure}[H]
    \setcounter{subfigure}{0}
    \centering
        \subfloat[Constraint addition]
        {
            \includegraphics[width=2.7in]{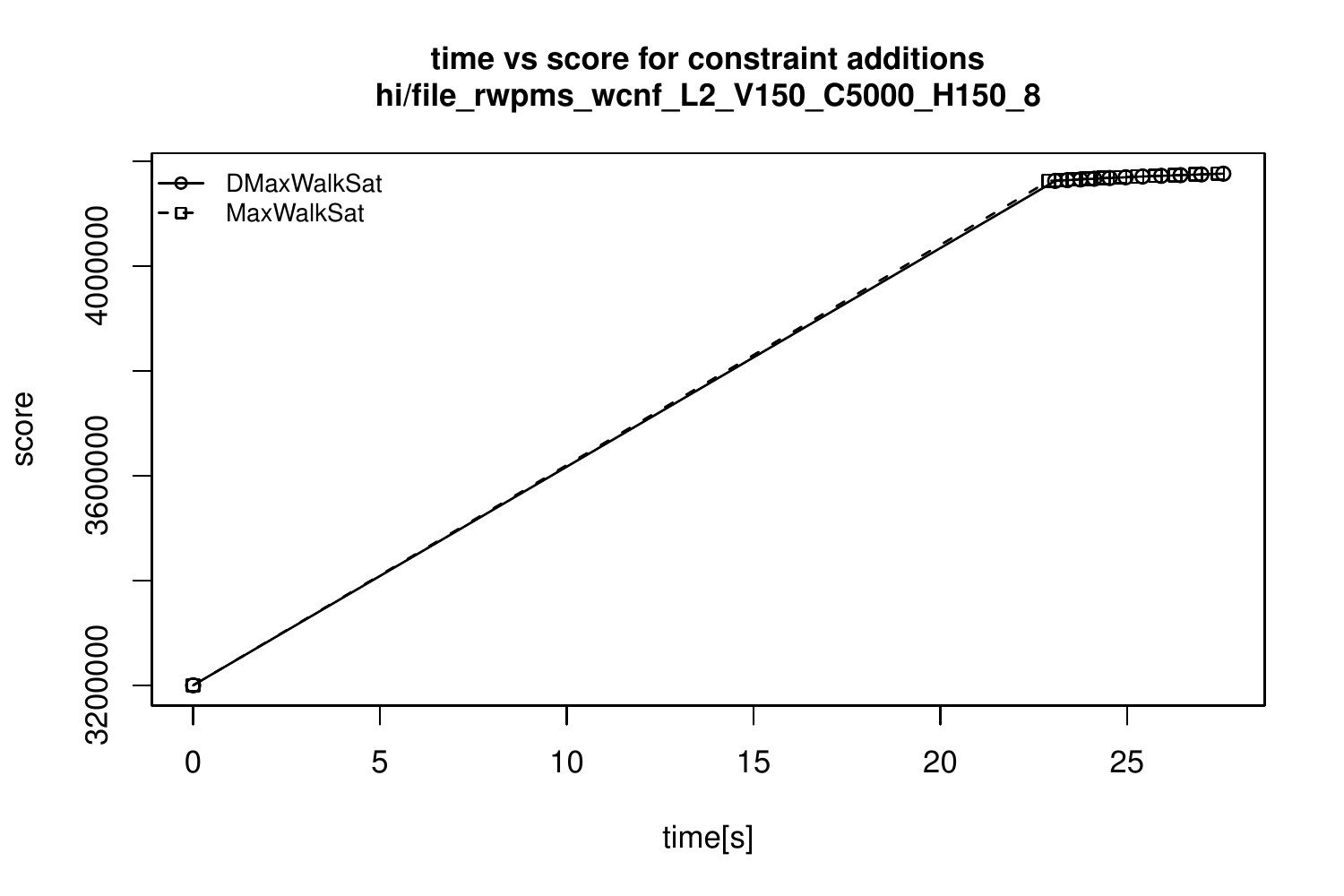}
        }
        \qquad
        \subfloat[Constraint removal]
        {
            \includegraphics[width=2.7in]{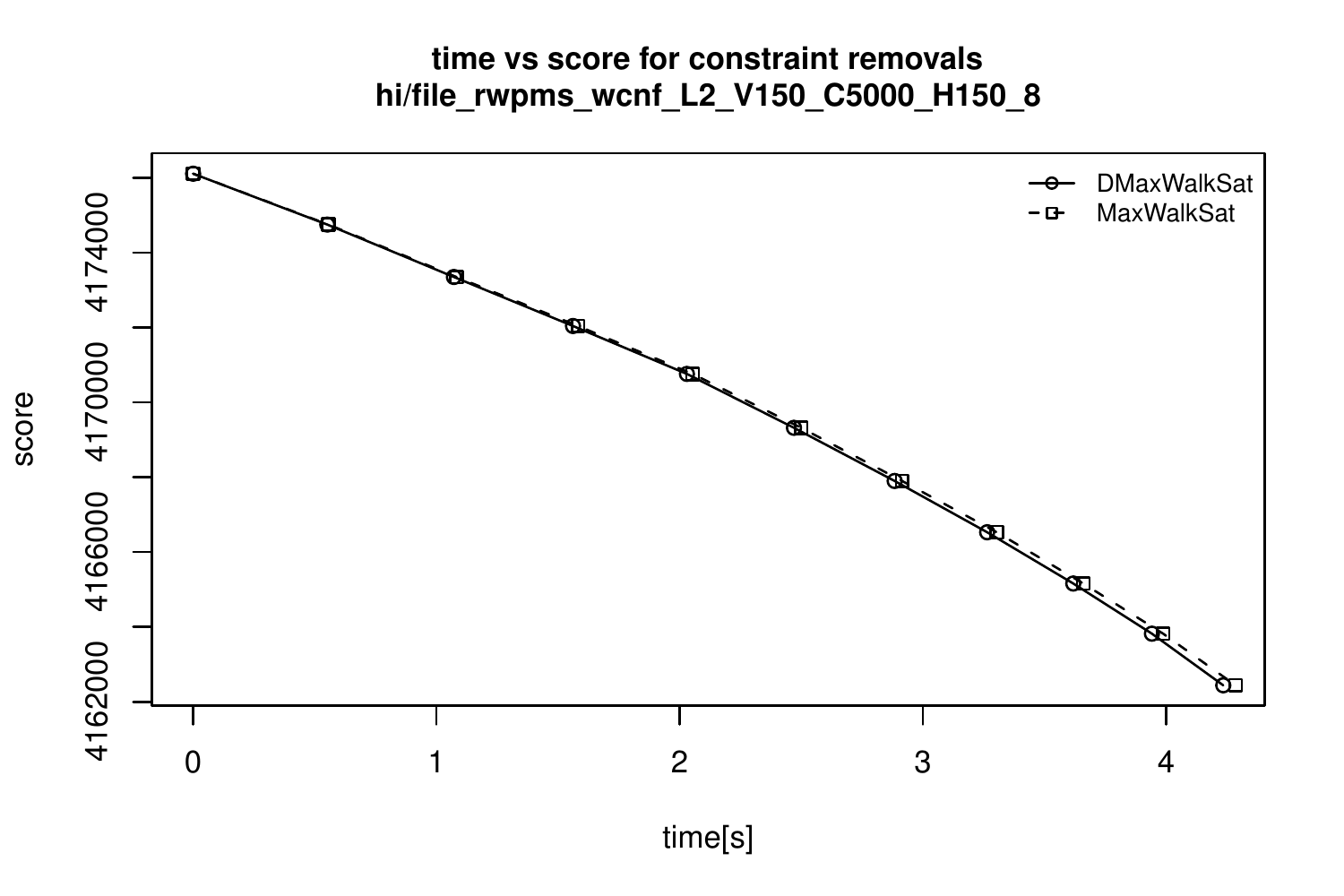}
        }

    \caption*{hi/file\_rwpms\_wcnf\_L2\_V150\_C5000\_H150\_8}
    \label{fig_hi/file_rwpms_wcnf_L2_V150_C5000_H150_8}
\end{figure}

\begin{figure}[H]
    \setcounter{subfigure}{0}
    \centering
        \subfloat[Constraint addition]
        {
            \includegraphics[width=2.7in]{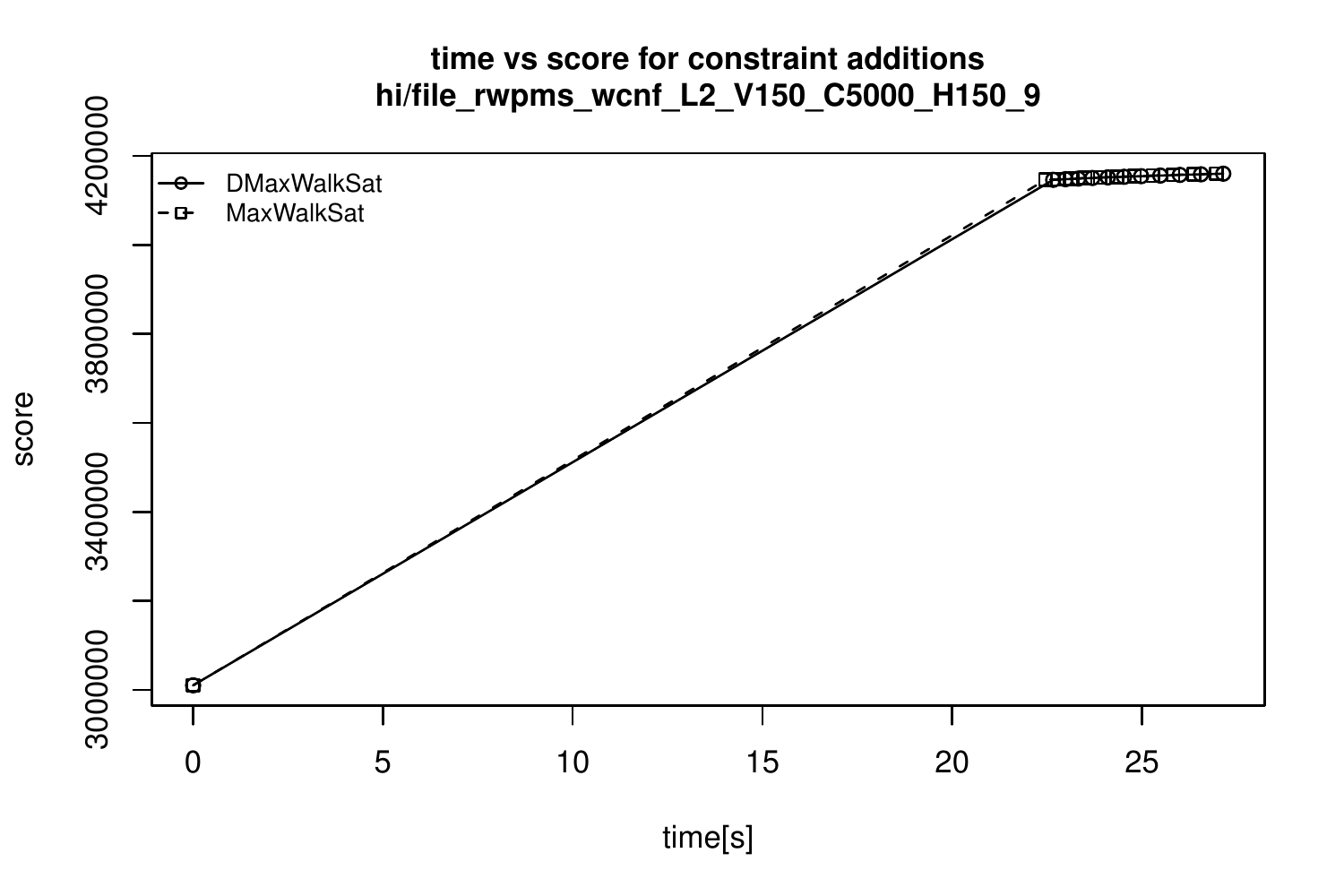}
        }
        \qquad
        \subfloat[Constraint removal]
        {
            \includegraphics[width=2.7in]{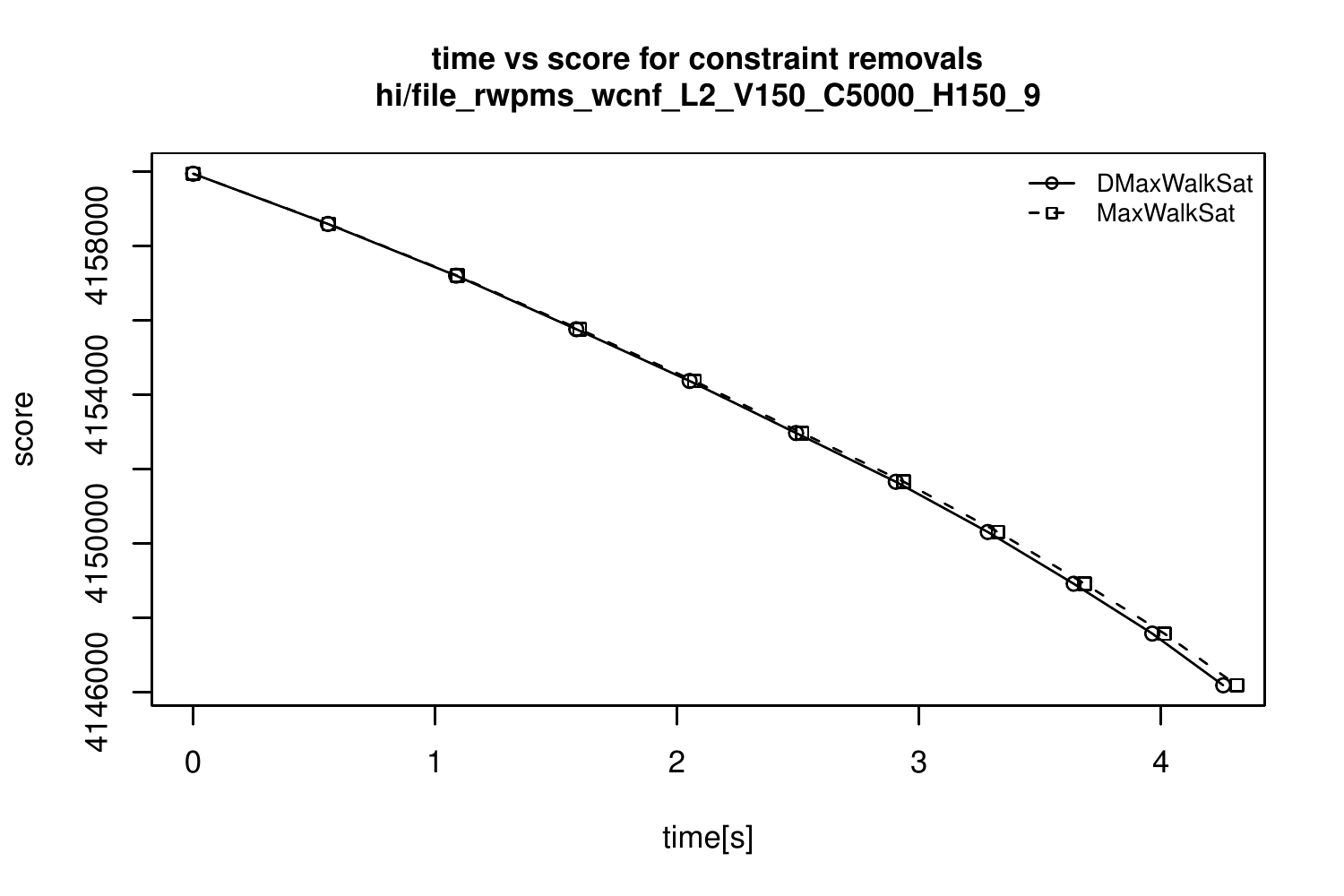}
        }

    \caption*{hi/file\_rwpms\_wcnf\_L2\_V150\_C5000\_H150\_9}
    \label{fig_hi/file_rwpms_wcnf_L2_V150_C5000_H150_9}
\end{figure}

\chapter{DWCSP Solution Velocities\label{dmaxwalksat_velocities}}

\begin{center}
    \begin{longtable}{|l|c|c|c|c|}
        \caption{Mean solution velocities $\left[\frac{score}{1.0 s}\right]$ for MaxWalkSat and DMaxWalkSat.} \label{tbl_dmaxwalksat_velocities} \\
        \hline
        \multirow{2}{*}{\textbf{Problem}} & \multicolumn{2}{|c|}{\textbf{MaxWalkSat}} & \multicolumn{2}{|c|}{\textbf{DMaxWalkSat}} \\
        & $\uparrow$ & $\downarrow$ & $\uparrow$ & $\downarrow$ \\
        \hline\hline
    lo/file\_rwpms\_wcnf\_L2\_V150\_C1000\_H150\_0 & 37801.4 & 17644.3 & 38096.2 & 18214.8 \\
    lo/file\_rwpms\_wcnf\_L2\_V150\_C1000\_H150\_1 & 44509.7 & 18494.4 & 45507.7 & 18748.3 \\
    lo/file\_rwpms\_wcnf\_L2\_V150\_C1000\_H150\_2 & 45686.5 & 18349.8 & 45783.7 & 18474.0 \\
    lo/file\_rwpms\_wcnf\_L2\_V150\_C1000\_H150\_3 & 45729.9 & 18963.6 & 46409.4 & 19006.8 \\
    lo/file\_rwpms\_wcnf\_L2\_V150\_C1000\_H150\_4 & 45681.6 & 18820.6 & 46522.7 & 18948.1 \\
    lo/file\_rwpms\_wcnf\_L2\_V150\_C1000\_H150\_5 & 49501.9 & 19163.7 & 50508.9 & 19250.0 \\
    lo/file\_rwpms\_wcnf\_L2\_V150\_C1000\_H150\_6 & 41534.6 & 19324.3 & 42449.4 & 19815.2 \\
    lo/file\_rwpms\_wcnf\_L2\_V150\_C1000\_H150\_7 & 41514.9 & 17644.3 & 41641.5 & 17965.8 \\
    lo/file\_rwpms\_wcnf\_L2\_V150\_C1000\_H150\_8 & 39498.1 & 18511.2 & 39998.0 & 18849.3 \\
    lo/file\_rwpms\_wcnf\_L2\_V150\_C1000\_H150\_9 & 47647.4 & 18800.9 & 47585.2 & 19059.6 \\
    lo/file\_rwpms\_wcnf\_L2\_V150\_C1500\_H150\_0 & 41112.5 & 11278.2 & 41562.4 & 11374.4 \\
    lo/file\_rwpms\_wcnf\_L2\_V150\_C1500\_H150\_1 & 39217.6 & 11750.2 & 39769.7 & 11629.2 \\
    lo/file\_rwpms\_wcnf\_L2\_V150\_C1500\_H150\_2 & 48013.7 & 12050.6 & 48231.9 & 12199.1 \\
    lo/file\_rwpms\_wcnf\_L2\_V150\_C1500\_H150\_3 & 40637.2 & 11532.0 & 41179.4 & 11607.5 \\
    lo/file\_rwpms\_wcnf\_L2\_V150\_C1500\_H150\_4 & 41143.5 & 11630.3 & 41741.6 & 11630.3 \\
    lo/file\_rwpms\_wcnf\_L2\_V150\_C1500\_H150\_5 & 44807.3 & 12125.2 & 45303.9 & 12033.9 \\
    lo/file\_rwpms\_wcnf\_L2\_V150\_C1500\_H150\_6 & 38802.2 & 11425.9 & 38766.1 & 11415.2 \\
    lo/file\_rwpms\_wcnf\_L2\_V150\_C1500\_H150\_7 & 46816.6 & 11621.1 & 47201.6 & 11731.7 \\
    lo/file\_rwpms\_wcnf\_L2\_V150\_C1500\_H150\_8 & 41694.6 & 11313.3 & 42291.8 & 11485.7 \\
    lo/file\_rwpms\_wcnf\_L2\_V150\_C1500\_H150\_9 & 38473.0 & 11649.4 & 38720.1 & 11660.4 \\
    lo/file\_rwpms\_wcnf\_L2\_V150\_C2000\_H150\_0 & 40383.6 & 8475.3 & 40939.1 & 8629.3 \\
    lo/file\_rwpms\_wcnf\_L2\_V150\_C2000\_H150\_1 & 45228.8 & 8551.8 & 45719.2 & 8662.2 \\
    lo/file\_rwpms\_wcnf\_L2\_V150\_C2000\_H150\_2 & 39768.2 & 8148.5 & 40271.3 & 8369.9 \\
    lo/file\_rwpms\_wcnf\_L2\_V150\_C2000\_H150\_3 & 47066.4 & 8479.6 & 47469.3 & 8602.5 \\
    lo/file\_rwpms\_wcnf\_L2\_V150\_C2000\_H150\_4 & 50656.3 & 8384.8 & 51246.0 & 8567.0 \\
    lo/file\_rwpms\_wcnf\_L2\_V150\_C2000\_H150\_5 & 45338.1 & 8556.1 & 46155.1 & 8631.0 \\
    lo/file\_rwpms\_wcnf\_L2\_V150\_C2000\_H150\_6 & 40695.3 & 8268.6 & 41571.4 & 8482.7 \\
    lo/file\_rwpms\_wcnf\_L2\_V150\_C2000\_H150\_7 & 41737.1 & 8339.3 & 42377.4 & 8477.9 \\
    lo/file\_rwpms\_wcnf\_L2\_V150\_C2000\_H150\_8 & 40560.0 & 8504.1 & 41125.9 & 8657.7 \\
    lo/file\_rwpms\_wcnf\_L2\_V150\_C2000\_H150\_9 & 37619.6 & 8534.2 & 38062.5 & 8640.1 \\
    me/file\_rwpms\_wcnf\_L2\_V150\_C2500\_H150\_0 & 42623.9 & 6625.6 & 43628.5 & 6688.5 \\
    me/file\_rwpms\_wcnf\_L2\_V150\_C2500\_H150\_1 & 38832.2 & 6648.2 & 39273.9 & 6700.7 \\
    me/file\_rwpms\_wcnf\_L2\_V150\_C2500\_H150\_2 & 42795.7 & 6731.7 & 43175.3 & 6831.7 \\
    me/file\_rwpms\_wcnf\_L2\_V150\_C2500\_H150\_3 & 36699.3 & 6873.4 & 37334.1 & 6934.5 \\
    me/file\_rwpms\_wcnf\_L2\_V150\_C2500\_H150\_4 & 37354.0 & 6524.7 & 38251.5 & 6582.0 \\
    me/file\_rwpms\_wcnf\_L2\_V150\_C2500\_H150\_5 & 38431.2 & 6716.6 & 38732.3 & 6793.9 \\
    me/file\_rwpms\_wcnf\_L2\_V150\_C2500\_H150\_6 & 36723.1 & 6660.1 & 37349.3 & 6684.1 \\
    me/file\_rwpms\_wcnf\_L2\_V150\_C2500\_H150\_7 & 38832.0 & 6764.7 & 39548.3 & 6769.1 \\
    me/file\_rwpms\_wcnf\_L2\_V150\_C2500\_H150\_8 & 34995.2 & 6638.6 & 35547.3 & 6686.2 \\
    me/file\_rwpms\_wcnf\_L2\_V150\_C2500\_H150\_9 & 43962.6 & 6526.4 & 44709.5 & 6644.7 \\
    me/file\_rwpms\_wcnf\_L2\_V150\_C3000\_H150\_0 & 35256.5 & 5478.2 & 35623.7 & 5551.0 \\
    me/file\_rwpms\_wcnf\_L2\_V150\_C3000\_H150\_1 & 43787.9 & 5453.3 & 44138.6 & 5522.8 \\
    me/file\_rwpms\_wcnf\_L2\_V150\_C3000\_H150\_2 & 41170.9 & 5456.4 & 41538.1 & 5510.2 \\
    me/file\_rwpms\_wcnf\_L2\_V150\_C3000\_H150\_3 & 40700.7 & 5424.1 & 41000.3 & 5539.4 \\
    me/file\_rwpms\_wcnf\_L2\_V150\_C3000\_H150\_4 & 38104.2 & 5412.5 & 38588.7 & 5510.8 \\
    me/file\_rwpms\_wcnf\_L2\_V150\_C3000\_H150\_5 & 33365.0 & 5390.4 & 33900.2 & 5437.5 \\
    me/file\_rwpms\_wcnf\_L2\_V150\_C3000\_H150\_6 & 40863.9 & 5496.9 & 41173.6 & 5588.6 \\
    me/file\_rwpms\_wcnf\_L2\_V150\_C3000\_H150\_7 & 41472.0 & 5495.8 & 41640.7 & 5565.1 \\
    me/file\_rwpms\_wcnf\_L2\_V150\_C3000\_H150\_8 & 40785.1 & 5473.8 & 41328.5 & 5521.2 \\
    me/file\_rwpms\_wcnf\_L2\_V150\_C3000\_H150\_9 & 39503.6 & 5417.9 & 39611.8 & 5536.1 \\
    me/file\_rwpms\_wcnf\_L2\_V150\_C3500\_H150\_0 & 44504.6 & 4663.0 & 44959.8 & 4727.8 \\
    me/file\_rwpms\_wcnf\_L2\_V150\_C3500\_H150\_1 & 39392.3 & 4714.4 & 40225.5 & 4750.1 \\
    me/file\_rwpms\_wcnf\_L2\_V150\_C3500\_H150\_2 & 38762.3 & 4638.5 & 39243.4 & 4713.0 \\
    me/file\_rwpms\_wcnf\_L2\_V150\_C3500\_H150\_3 & 39796.5 & 4646.4 & 40128.6 & 4720.4 \\
    me/file\_rwpms\_wcnf\_L2\_V150\_C3500\_H150\_4 & 41755.2 & 4631.3 & 42088.6 & 4703.5 \\
    me/file\_rwpms\_wcnf\_L2\_V150\_C3500\_H150\_5 & 35722.5 & 4512.7 & 36047.3 & 4609.3 \\
    me/file\_rwpms\_wcnf\_L2\_V150\_C3500\_H150\_6 & 36398.8 & 4651.1 & 36771.7 & 4714.1 \\
    me/file\_rwpms\_wcnf\_L2\_V150\_C3500\_H150\_7 & 38792.8 & 4609.1 & 39202.6 & 4689.3 \\
    me/file\_rwpms\_wcnf\_L2\_V150\_C3500\_H150\_8 & 40081.2 & 4647.5 & 40307.3 & 4693.8 \\
    me/file\_rwpms\_wcnf\_L2\_V150\_C3500\_H150\_9 & 41632.7 & 4509.1 & 41949.8 & 4570.5 \\
    hi/file\_rwpms\_wcnf\_L2\_V150\_C4000\_H150\_0 & 42500.4 & 4037.0 & 42715.0 & 4077.6 \\
    hi/file\_rwpms\_wcnf\_L2\_V150\_C4000\_H150\_1 & 34178.9 & 4021.2 & 34766.8 & 4057.7 \\
    hi/file\_rwpms\_wcnf\_L2\_V150\_C4000\_H150\_2 & 36554.7 & 3936.3 & 36894.2 & 3971.5 \\
    hi/file\_rwpms\_wcnf\_L2\_V150\_C4000\_H150\_3 & 40360.0 & 4093.6 & 40570.5 & 4136.7 \\
    hi/file\_rwpms\_wcnf\_L2\_V150\_C4000\_H150\_4 & 44064.3 & 4043.5 & 44299.6 & 4065.5 \\
    hi/file\_rwpms\_wcnf\_L2\_V150\_C4000\_H150\_5 & 37965.2 & 4134.7 & 38142.8 & 4180.6 \\
    hi/file\_rwpms\_wcnf\_L2\_V150\_C4000\_H150\_6 & 41900.3 & 4099.6 & 42016.9 & 4129.5 \\
    hi/file\_rwpms\_wcnf\_L2\_V150\_C4000\_H150\_7 & 40872.1 & 4040.4 & 41410.6 & 4063.4 \\
    hi/file\_rwpms\_wcnf\_L2\_V150\_C4000\_H150\_8 & 42596.6 & 4110.6 & 43120.6 & 4142.2 \\
    hi/file\_rwpms\_wcnf\_L2\_V150\_C4000\_H150\_9 & 41349.9 & 4145.9 & 41640.7 & 4191.1 \\
    hi/file\_rwpms\_wcnf\_L2\_V150\_C4500\_H150\_0 & 35331.8 & 3566.2 & 35850.1 & 3598.4 \\
    hi/file\_rwpms\_wcnf\_L2\_V150\_C4500\_H150\_1 & 33617.4 & 3542.2 & 34196.9 & 3583.8 \\
    hi/file\_rwpms\_wcnf\_L2\_V150\_C4500\_H150\_2 & 40389.5 & 3577.8 & 40473.9 & 3619.2 \\
    hi/file\_rwpms\_wcnf\_L2\_V150\_C4500\_H150\_3 & 37865.1 & 3558.1 & 38191.9 & 3612.4 \\
    hi/file\_rwpms\_wcnf\_L2\_V150\_C4500\_H150\_4 & 39815.5 & 3537.5 & 39791.0 & 3573.7 \\
    hi/file\_rwpms\_wcnf\_L2\_V150\_C4500\_H150\_5 & 41425.4 & 3560.5 & 41666.1 & 3609.7 \\
    hi/file\_rwpms\_wcnf\_L2\_V150\_C4500\_H150\_6 & 40623.1 & 3507.1 & 41069.6 & 3548.3 \\
    hi/file\_rwpms\_wcnf\_L2\_V150\_C4500\_H150\_7 & 37260.9 & 3569.7 & 37669.3 & 3612.2 \\
    hi/file\_rwpms\_wcnf\_L2\_V150\_C4500\_H150\_8 & 40539.8 & 3569.7 & 40823.3 & 3612.5 \\
    hi/file\_rwpms\_wcnf\_L2\_V150\_C4500\_H150\_9 & 37819.0 & 3591.0 & 38390.5 & 3630.4 \\
    hi/file\_rwpms\_wcnf\_L2\_V150\_C5000\_H150\_0 & 42205.0 & 3207.8 & 42324.3 & 3260.7 \\
    hi/file\_rwpms\_wcnf\_L2\_V150\_C5000\_H150\_1 & 40084.8 & 3193.5 & 40491.5 & 3228.1 \\
    hi/file\_rwpms\_wcnf\_L2\_V150\_C5000\_H150\_2 & 42377.1 & 3257.0 & 42785.7 & 3295.5 \\
    hi/file\_rwpms\_wcnf\_L2\_V150\_C5000\_H150\_3 & 32625.9 & 3275.4 & 33011.5 & 3329.4 \\
    hi/file\_rwpms\_wcnf\_L2\_V150\_C5000\_H150\_4 & 35842.3 & 3203.4 & 35978.1 & 3256.5 \\
    hi/file\_rwpms\_wcnf\_L2\_V150\_C5000\_H150\_5 & 39797.5 & 3201.9 & 40185.2 & 3249.7 \\
    hi/file\_rwpms\_wcnf\_L2\_V150\_C5000\_H150\_6 & 38254.7 & 3224.6 & 38506.0 & 3277.5 \\
    hi/file\_rwpms\_wcnf\_L2\_V150\_C5000\_H150\_7 & 40895.8 & 3214.8 & 41094.4 & 3262.5 \\
    hi/file\_rwpms\_wcnf\_L2\_V150\_C5000\_H150\_8 & 35086.4 & 3189.7 & 35398.1 & 3227.9 \\
    hi/file\_rwpms\_wcnf\_L2\_V150\_C5000\_H150\_9 & 42131.2 & 3187.3 & 42360.0 & 3229.5 \\
        \hline
    \end{longtable}
\end{center}

\chapter{RDMaxWalkSat Score vs. Time Graphs\label{rdmaxwalksat_graphs}}
\begin{figure}[H]
    \centering
    \includegraphics[height=3.5in]{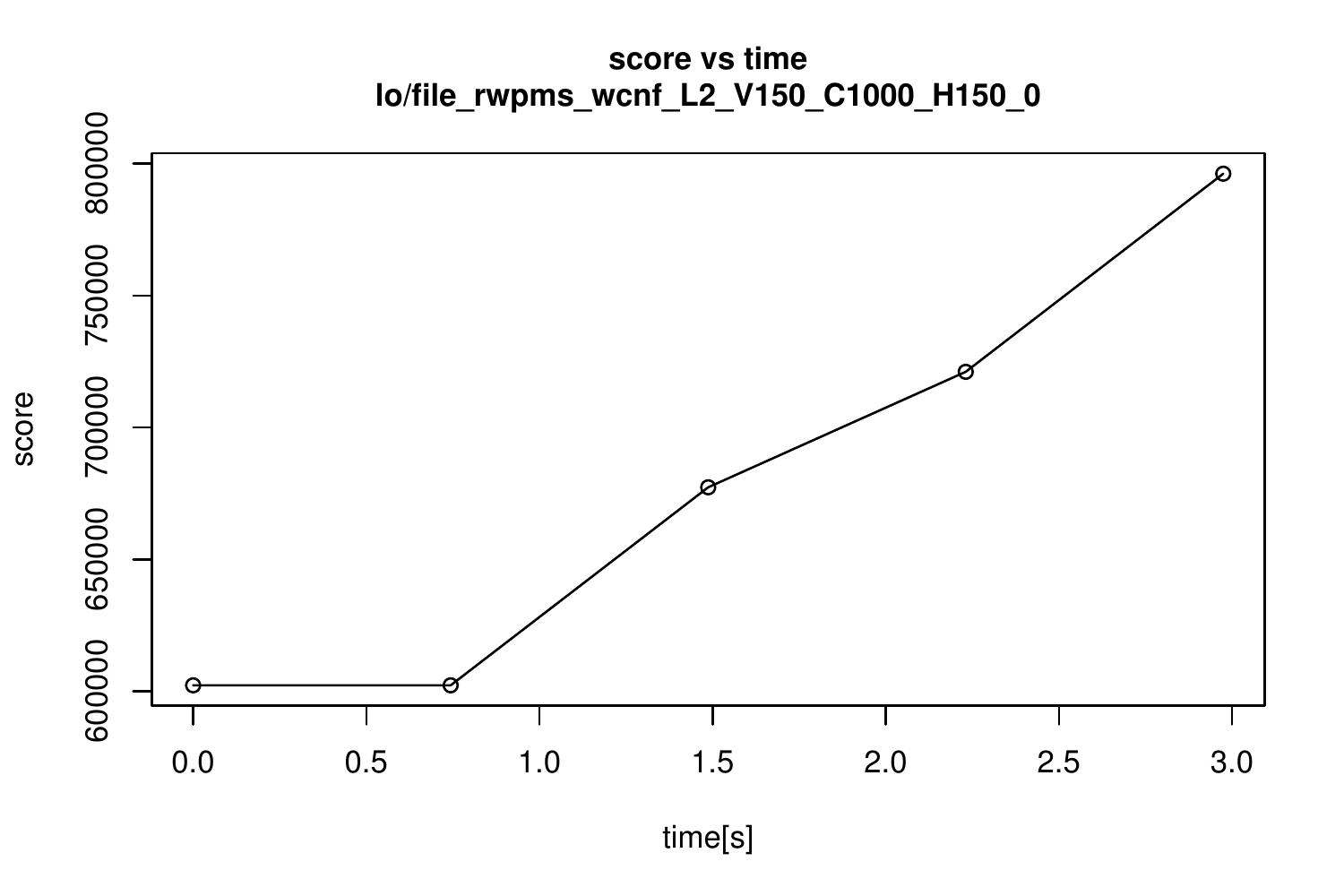}
    \label{fig_lo/file_rwpms_wcnf_L2_V150_C1000_H150_0/file_rwpms_wcnf_L2_V150_C1000_H150_0-score_vs_time}
\end{figure}

\begin{figure}[H]
    \centering
    \includegraphics[height=3.5in]{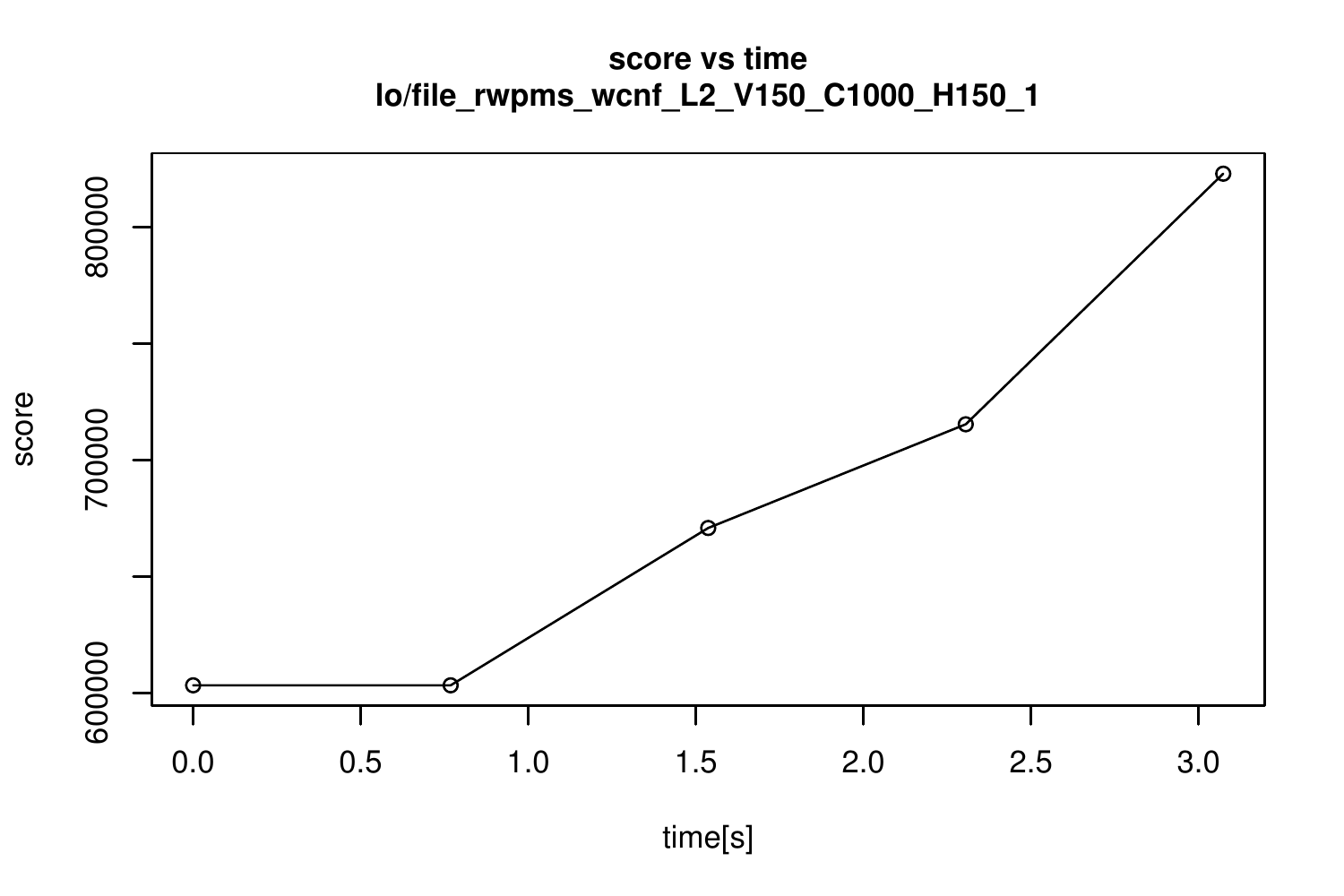}
    \label{fig_lo/file_rwpms_wcnf_L2_V150_C1000_H150_1/file_rwpms_wcnf_L2_V150_C1000_H150_1-score_vs_time}
\end{figure}

\begin{figure}[H]
    \centering
    \includegraphics[height=3.5in]{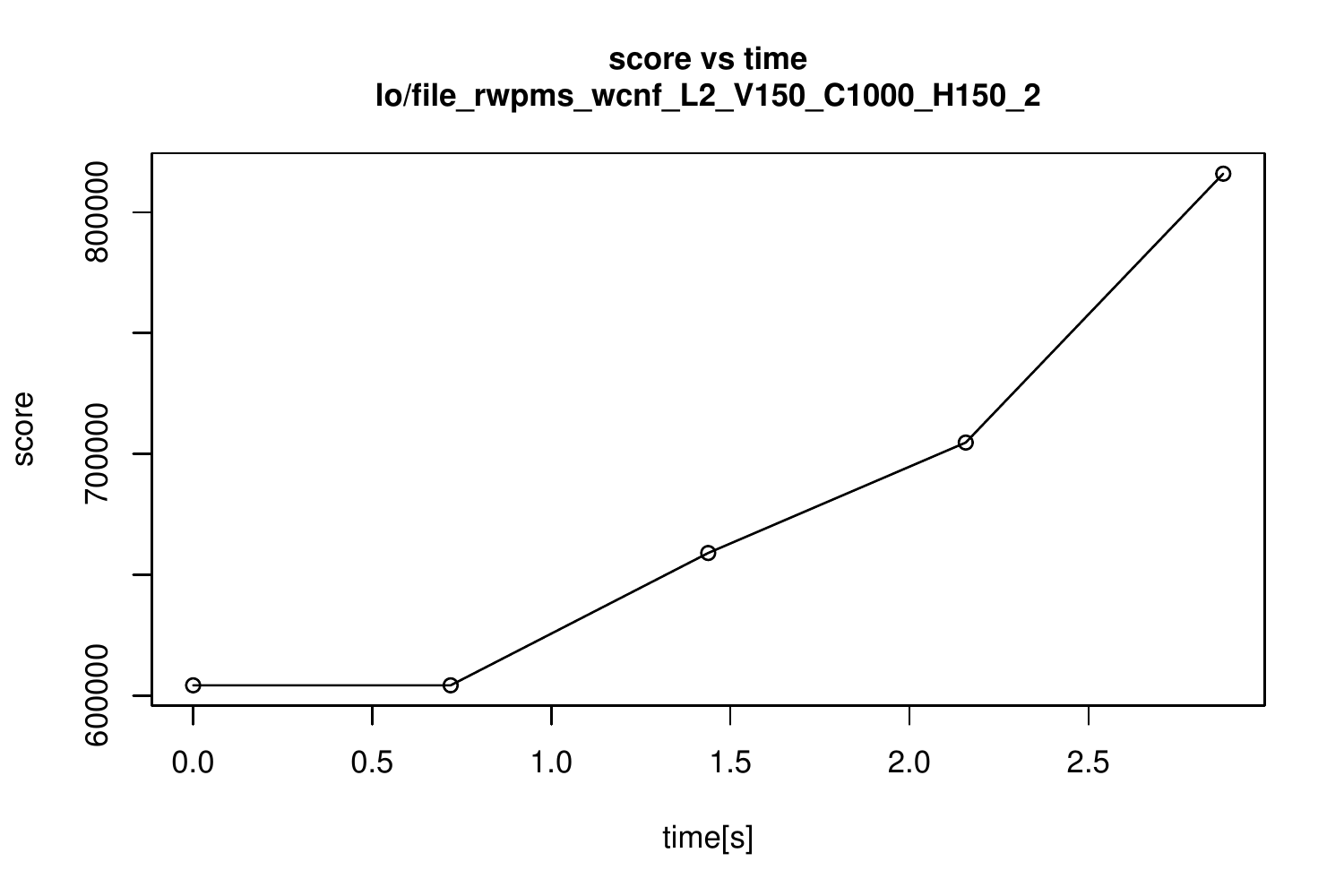}
    \label{fig_lo/file_rwpms_wcnf_L2_V150_C1000_H150_2/file_rwpms_wcnf_L2_V150_C1000_H150_2-score_vs_time}
\end{figure}

\begin{figure}[H]
    \centering
    \includegraphics[height=3.5in]{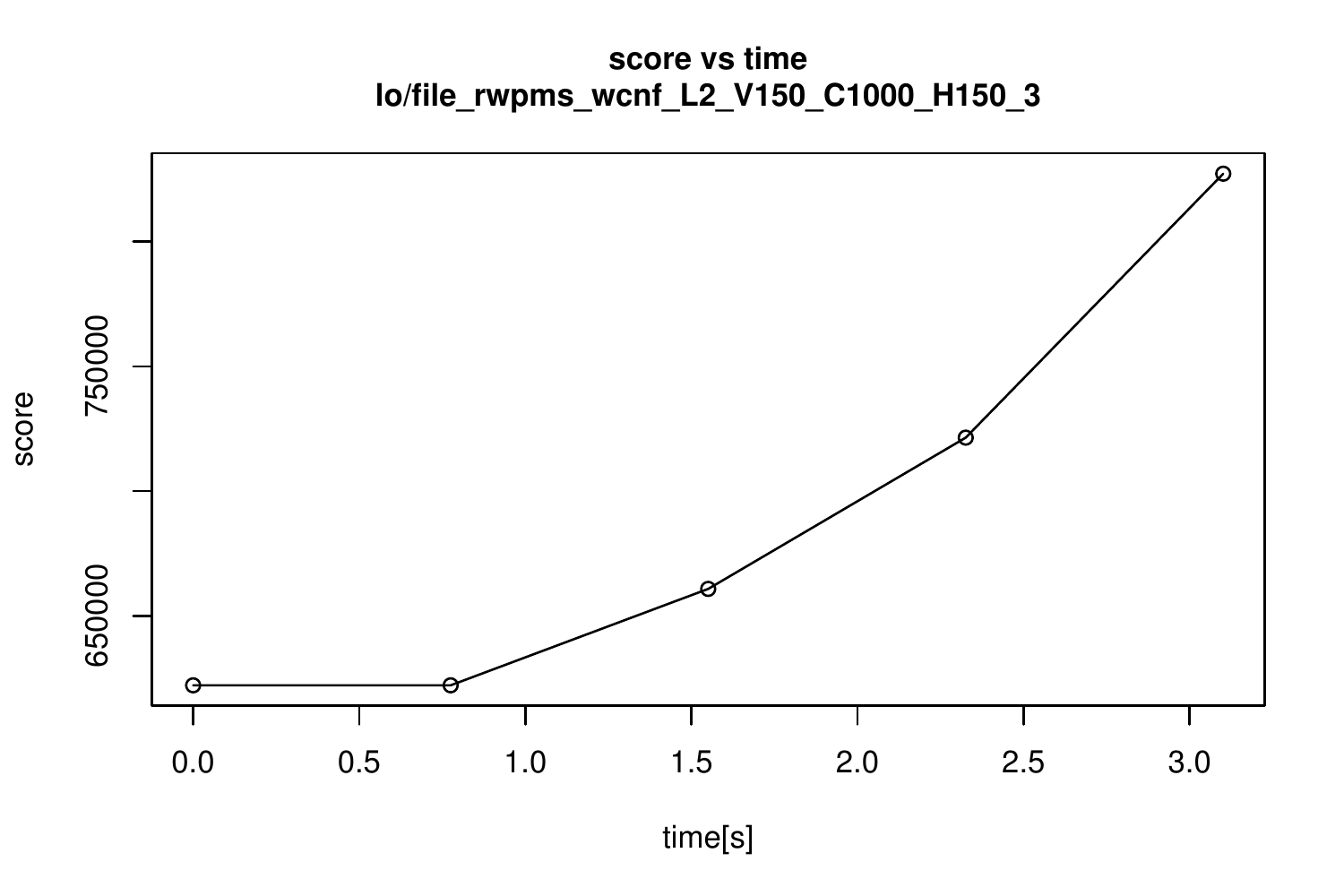}
    \label{fig_lo/file_rwpms_wcnf_L2_V150_C1000_H150_3/file_rwpms_wcnf_L2_V150_C1000_H150_3-score_vs_time}
\end{figure}

\begin{figure}[H]
    \centering
    \includegraphics[height=3.5in]{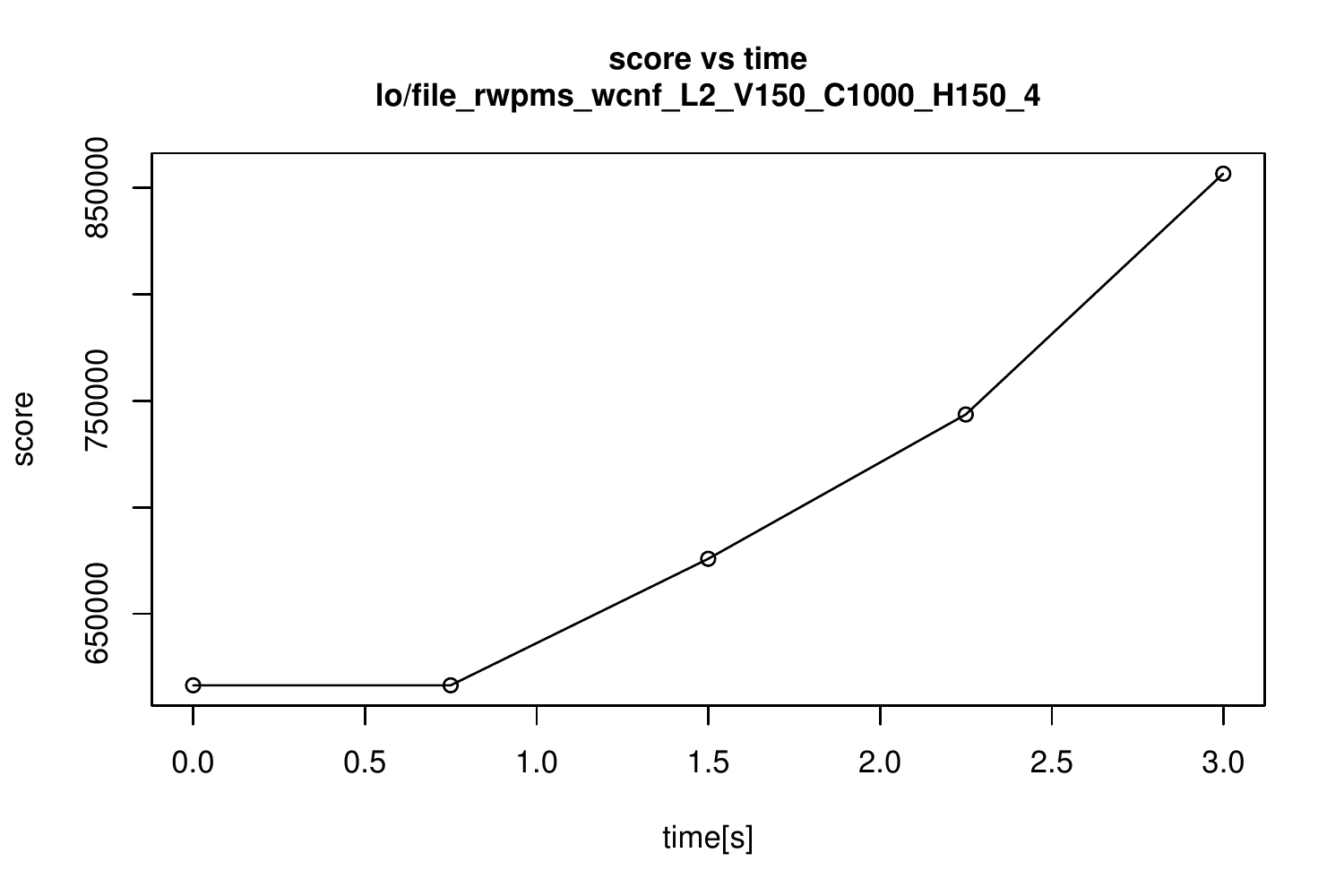}
    \label{fig_lo/file_rwpms_wcnf_L2_V150_C1000_H150_4/file_rwpms_wcnf_L2_V150_C1000_H150_4-score_vs_time}
\end{figure}

\begin{figure}[H]
    \centering
    \includegraphics[height=3.5in]{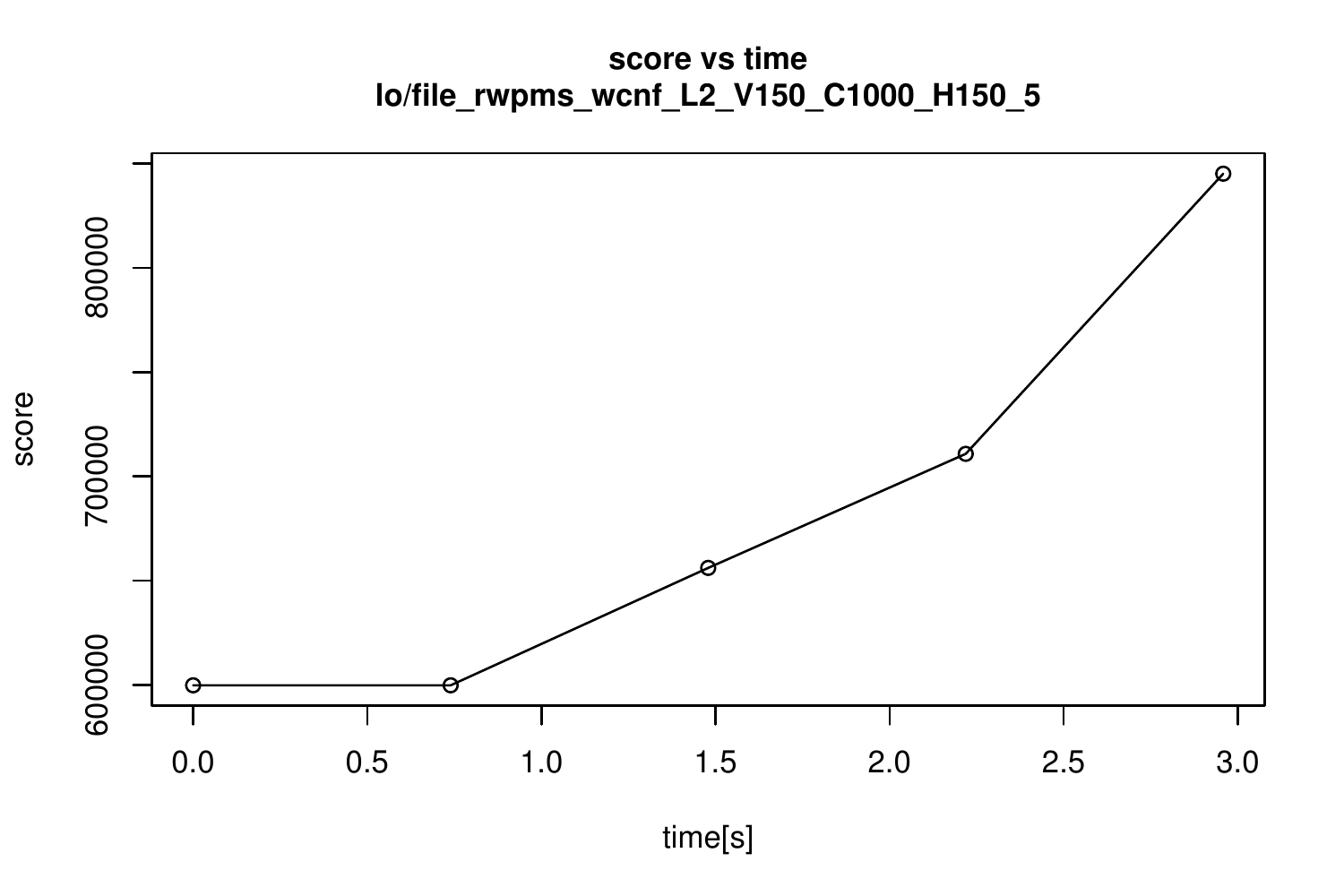}
    \label{fig_lo/file_rwpms_wcnf_L2_V150_C1000_H150_5/file_rwpms_wcnf_L2_V150_C1000_H150_5-score_vs_time}
\end{figure}

\begin{figure}[H]
    \centering
    \includegraphics[height=3.5in]{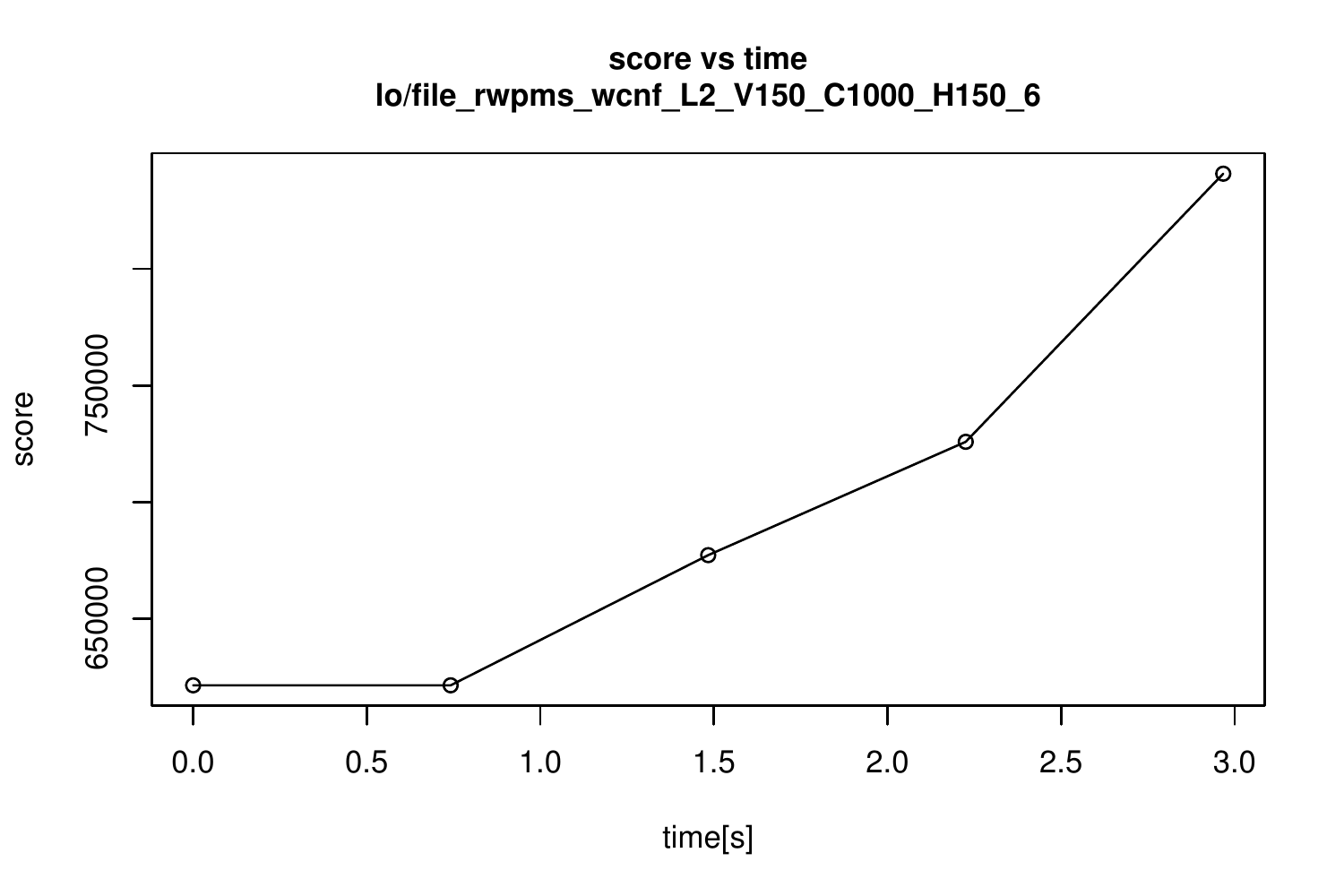}
    \label{fig_lo/file_rwpms_wcnf_L2_V150_C1000_H150_6/file_rwpms_wcnf_L2_V150_C1000_H150_6-score_vs_time}
\end{figure}

\begin{figure}[H]
    \centering
    \includegraphics[height=3.5in]{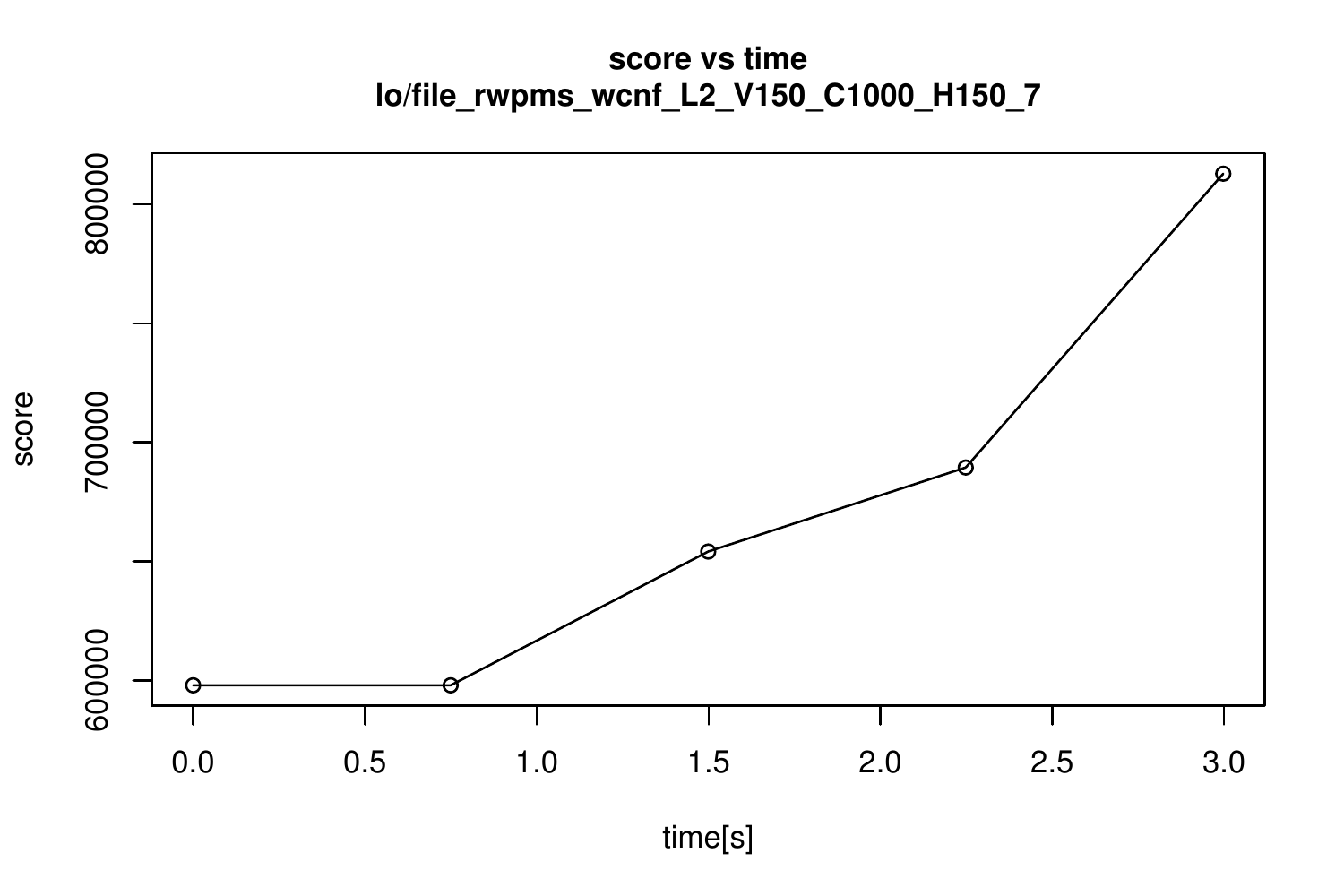}
    \label{fig_lo/file_rwpms_wcnf_L2_V150_C1000_H150_7/file_rwpms_wcnf_L2_V150_C1000_H150_7-score_vs_time}
\end{figure}

\begin{figure}[H]
    \centering
    \includegraphics[height=3.5in]{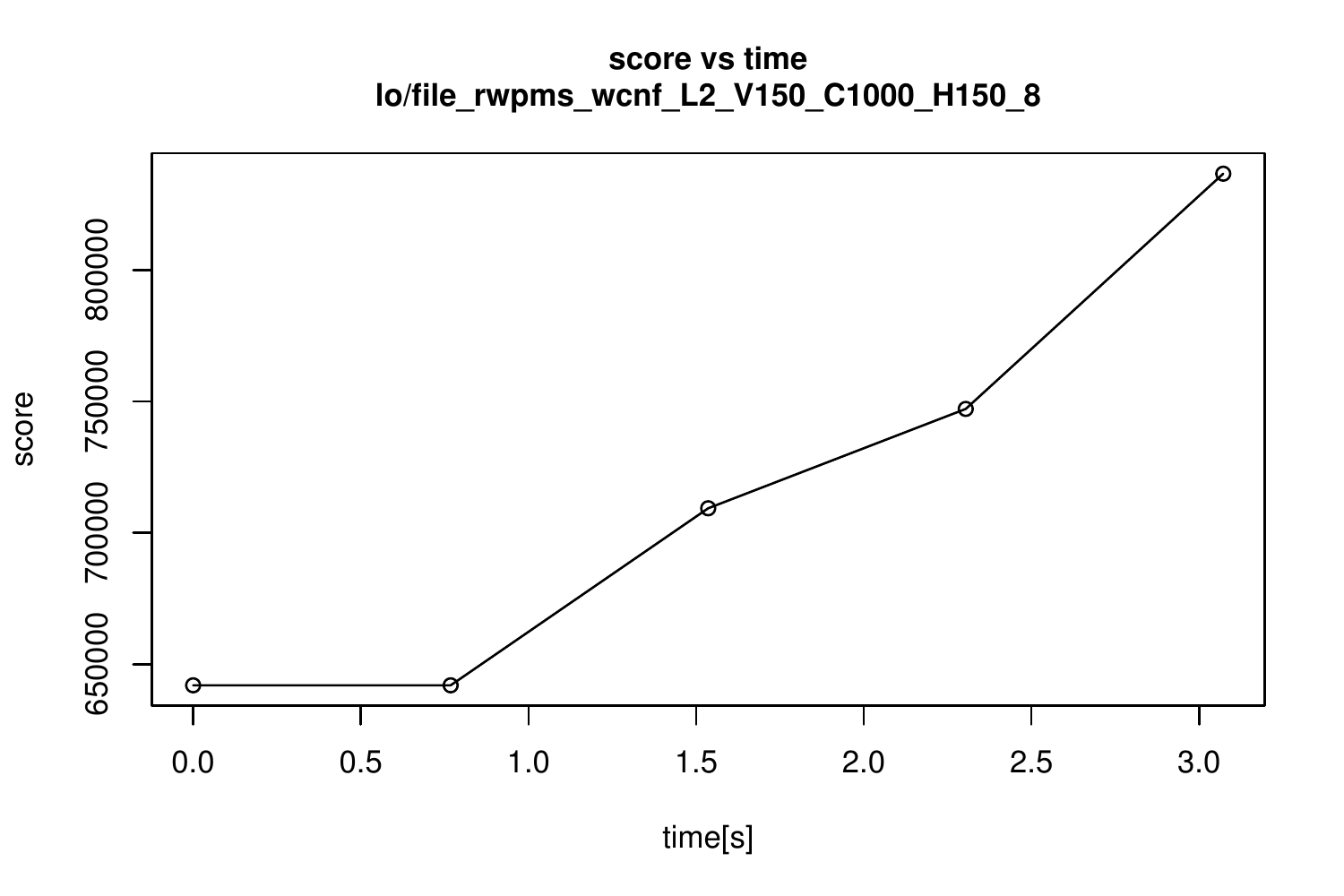}
    \label{fig_lo/file_rwpms_wcnf_L2_V150_C1000_H150_8/file_rwpms_wcnf_L2_V150_C1000_H150_8-score_vs_time}
\end{figure}

\begin{figure}[H]
    \centering
    \includegraphics[height=3.5in]{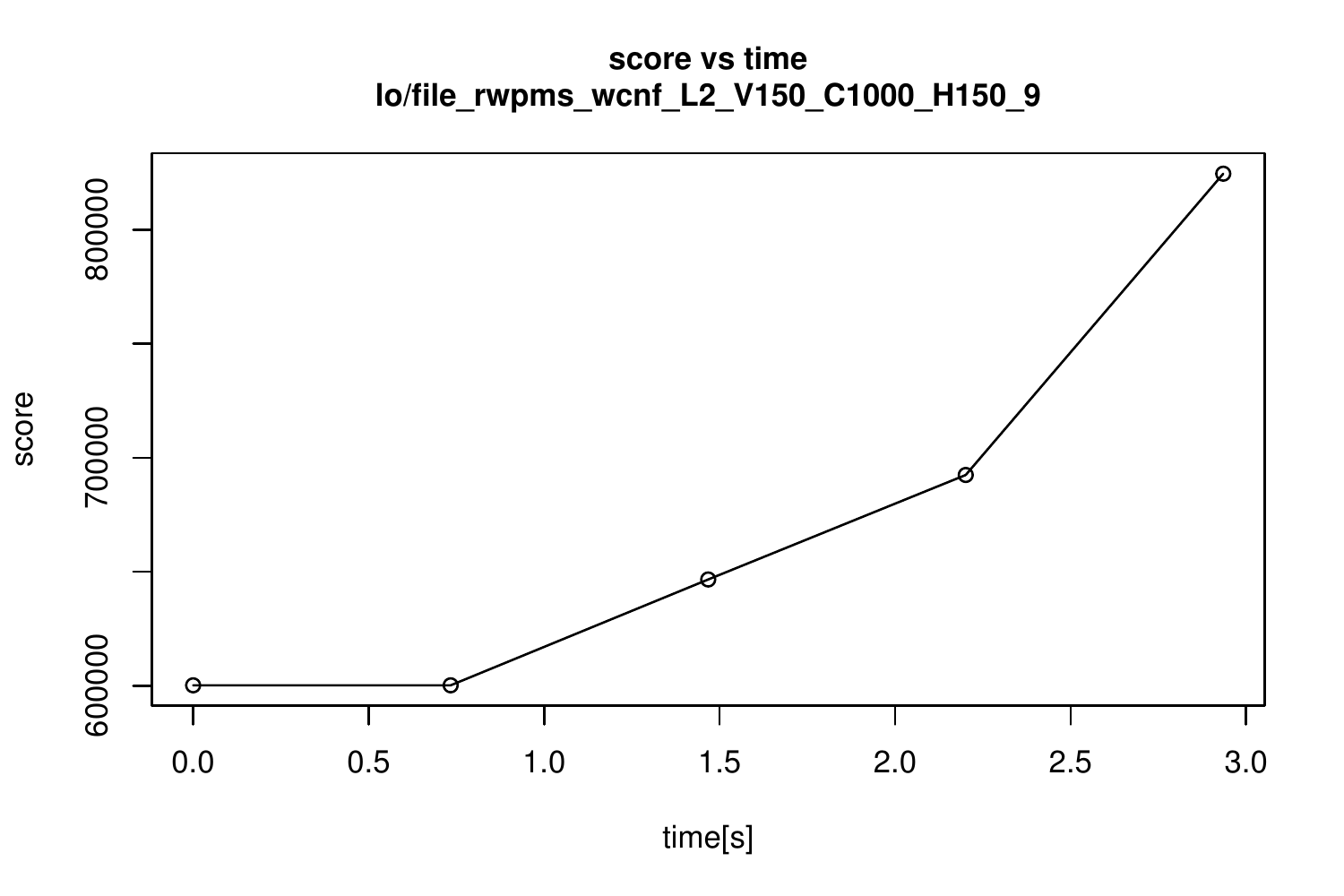}
    \label{fig_lo/file_rwpms_wcnf_L2_V150_C1000_H150_9/file_rwpms_wcnf_L2_V150_C1000_H150_9-score_vs_time}
\end{figure}

\begin{figure}[H]
    \centering
    \includegraphics[height=3.5in]{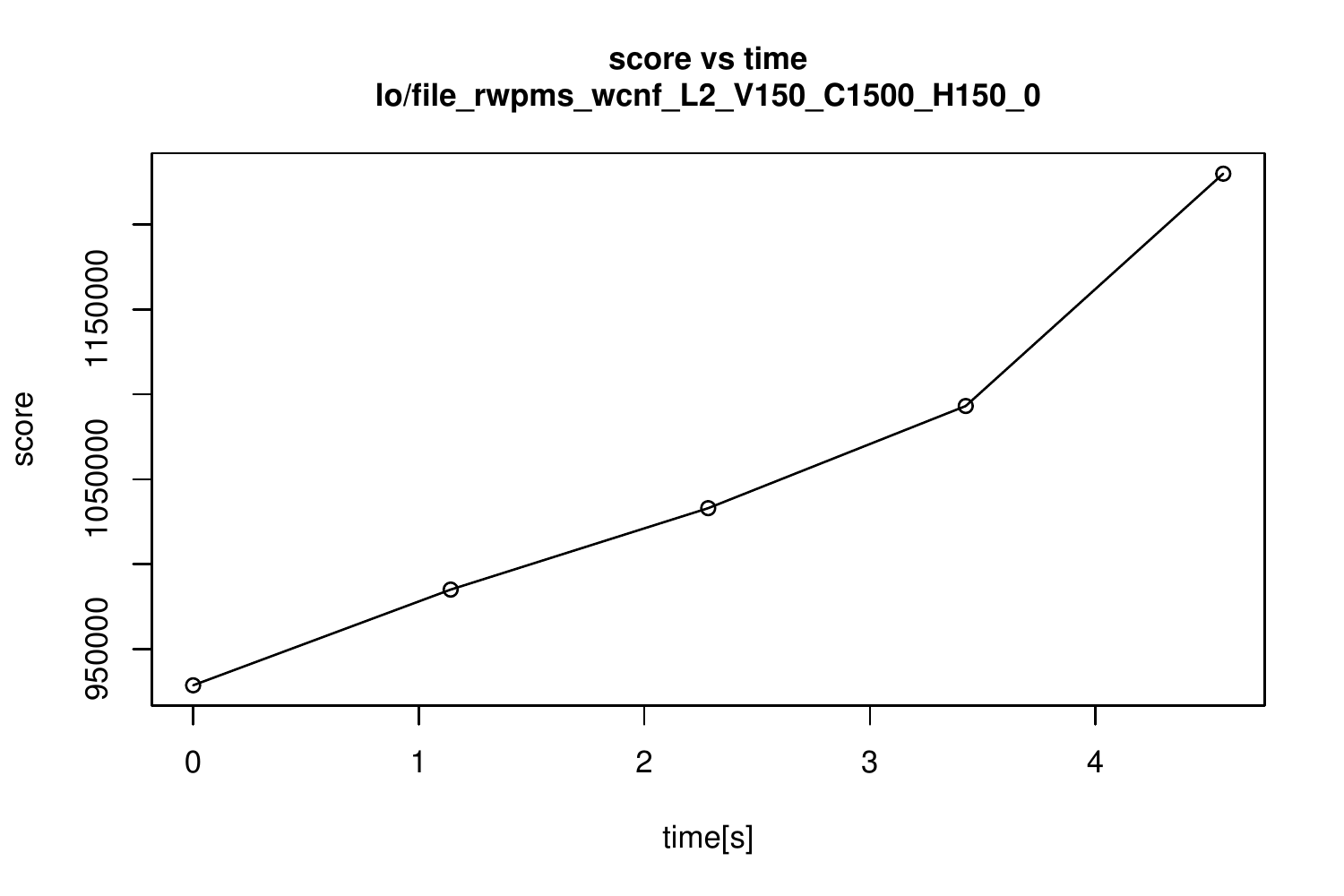}
    \label{fig_lo/file_rwpms_wcnf_L2_V150_C1500_H150_0/file_rwpms_wcnf_L2_V150_C1500_H150_0-score_vs_time}
\end{figure}

\begin{figure}[H]
    \centering
    \includegraphics[height=3.5in]{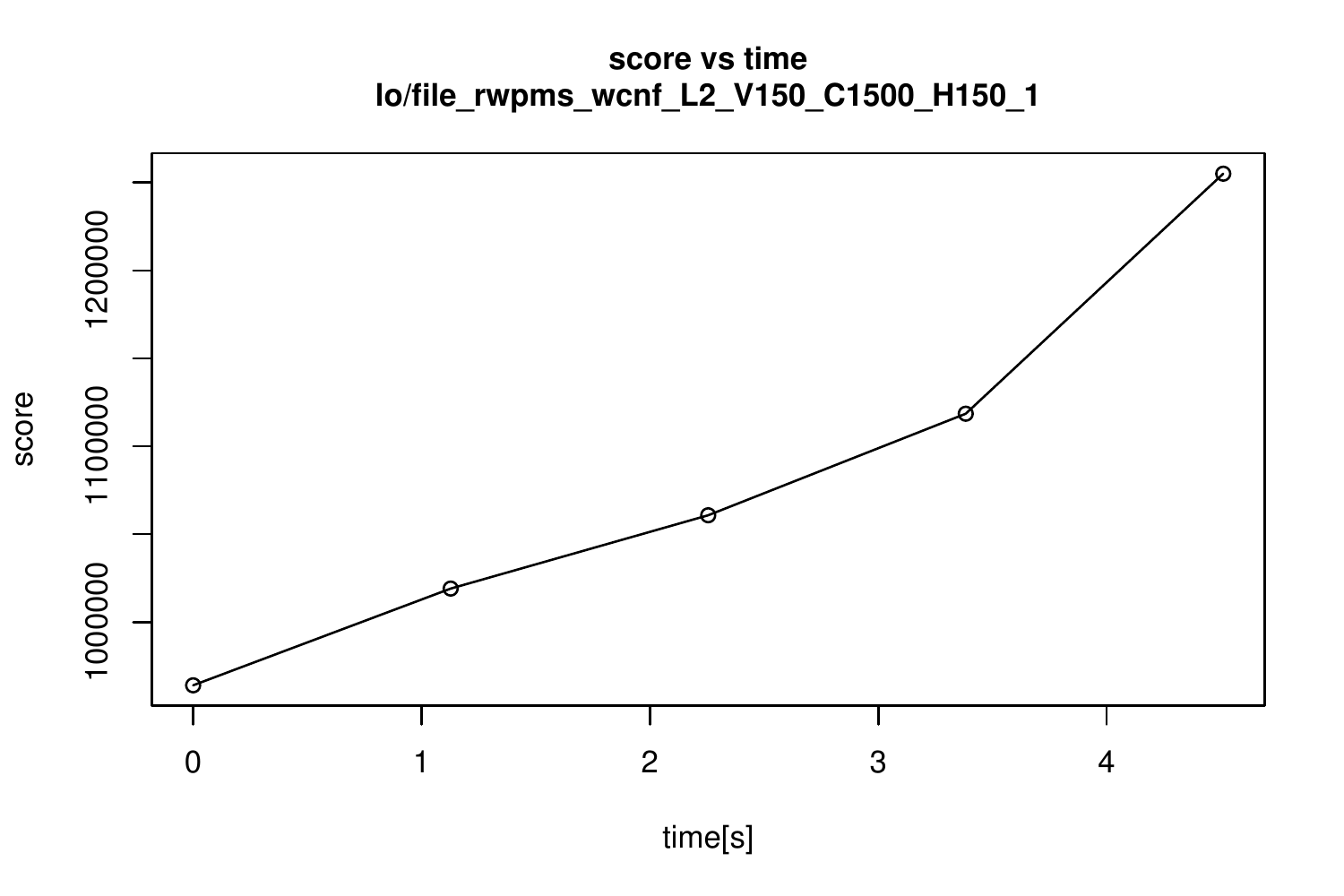}
    \label{fig_lo/file_rwpms_wcnf_L2_V150_C1500_H150_1/file_rwpms_wcnf_L2_V150_C1500_H150_1-score_vs_time}
\end{figure}

\begin{figure}[H]
    \centering
    \includegraphics[height=3.5in]{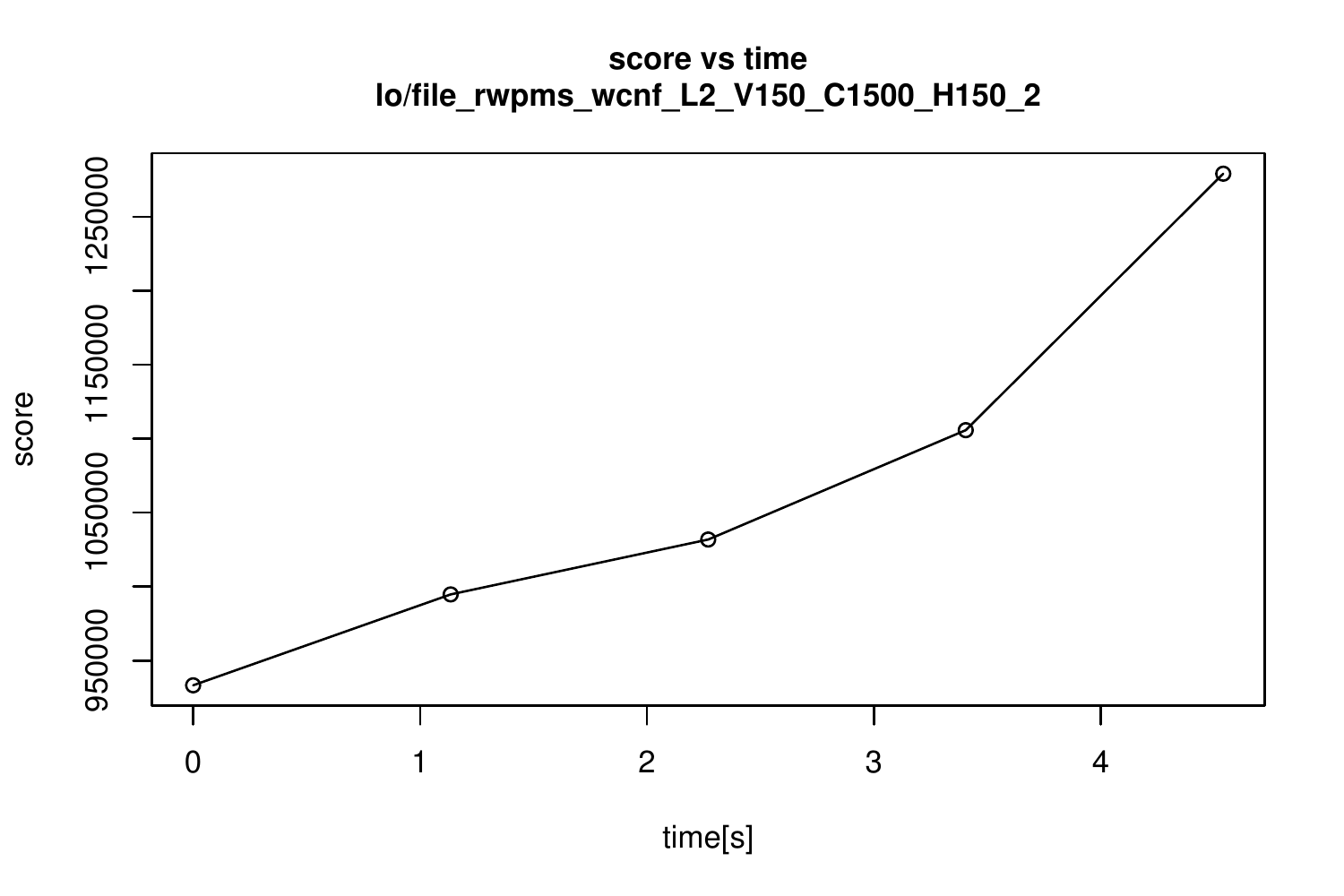}
    \label{fig_lo/file_rwpms_wcnf_L2_V150_C1500_H150_2/file_rwpms_wcnf_L2_V150_C1500_H150_2-score_vs_time}
\end{figure}

\begin{figure}[H]
    \centering
    \includegraphics[height=3.5in]{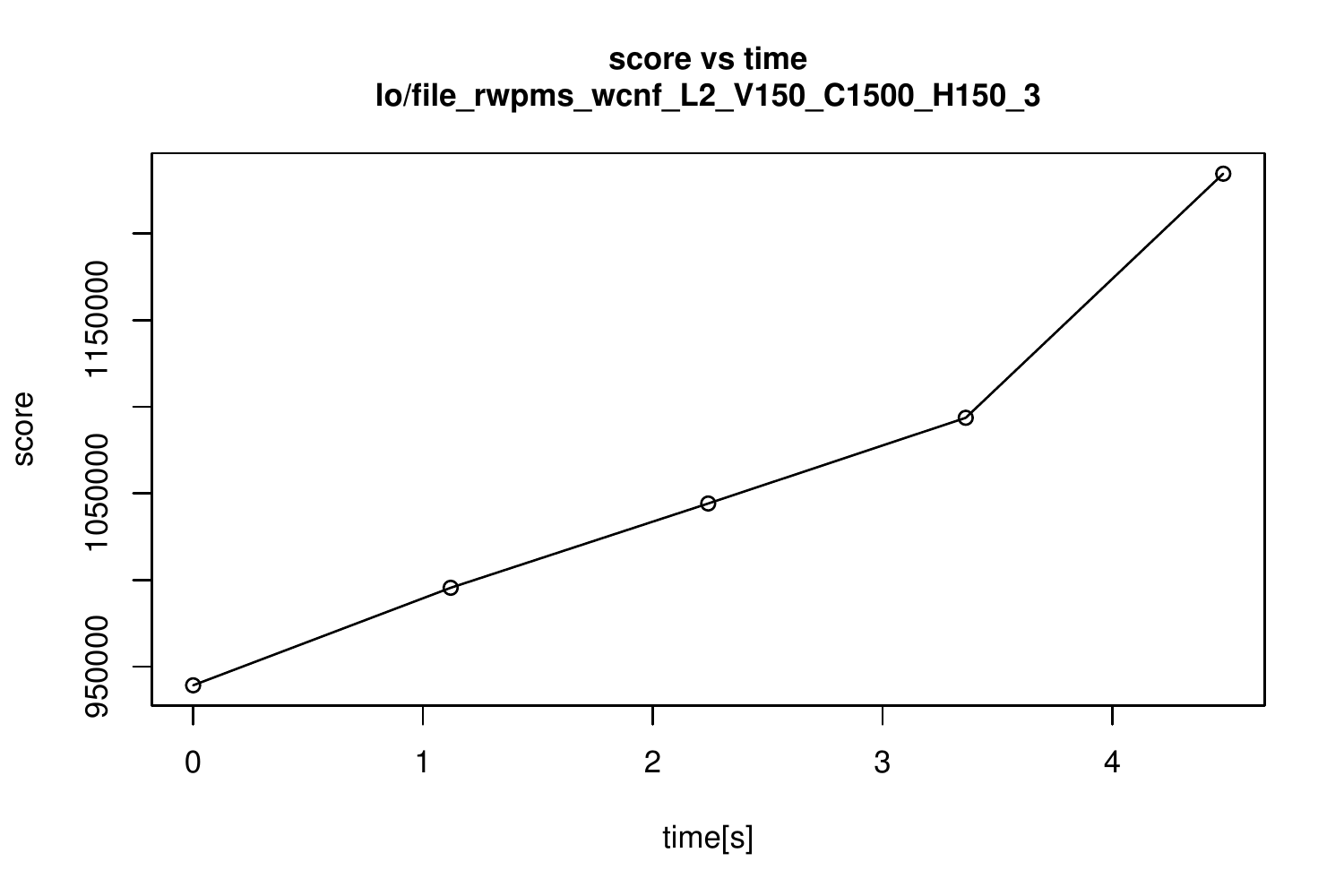}
    \label{fig_lo/file_rwpms_wcnf_L2_V150_C1500_H150_3/file_rwpms_wcnf_L2_V150_C1500_H150_3-score_vs_time}
\end{figure}

\begin{figure}[H]
    \centering
    \includegraphics[height=3.5in]{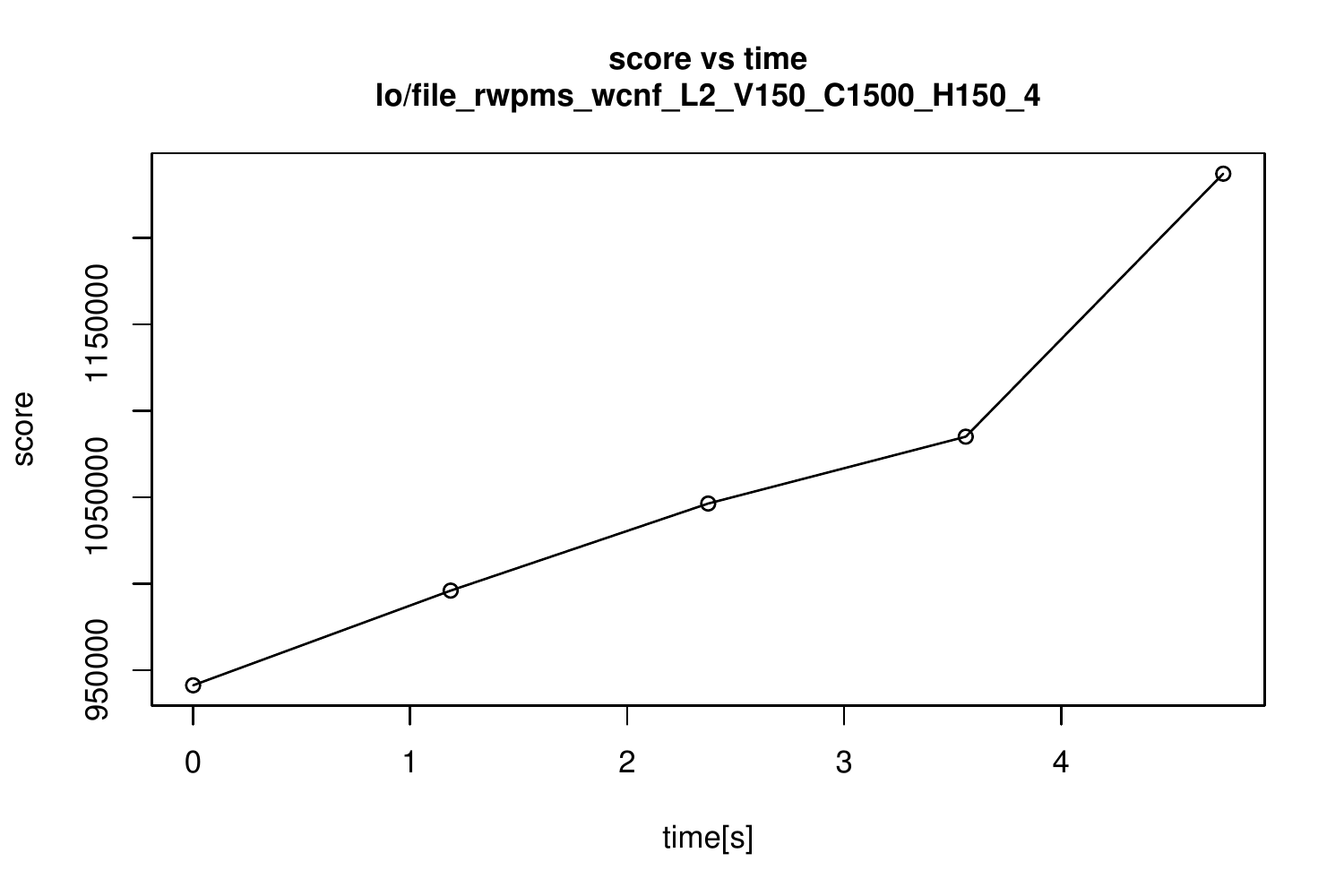}
    \label{fig_lo/file_rwpms_wcnf_L2_V150_C1500_H150_4/file_rwpms_wcnf_L2_V150_C1500_H150_4-score_vs_time}
\end{figure}

\begin{figure}[H]
    \centering
    \includegraphics[height=3.5in]{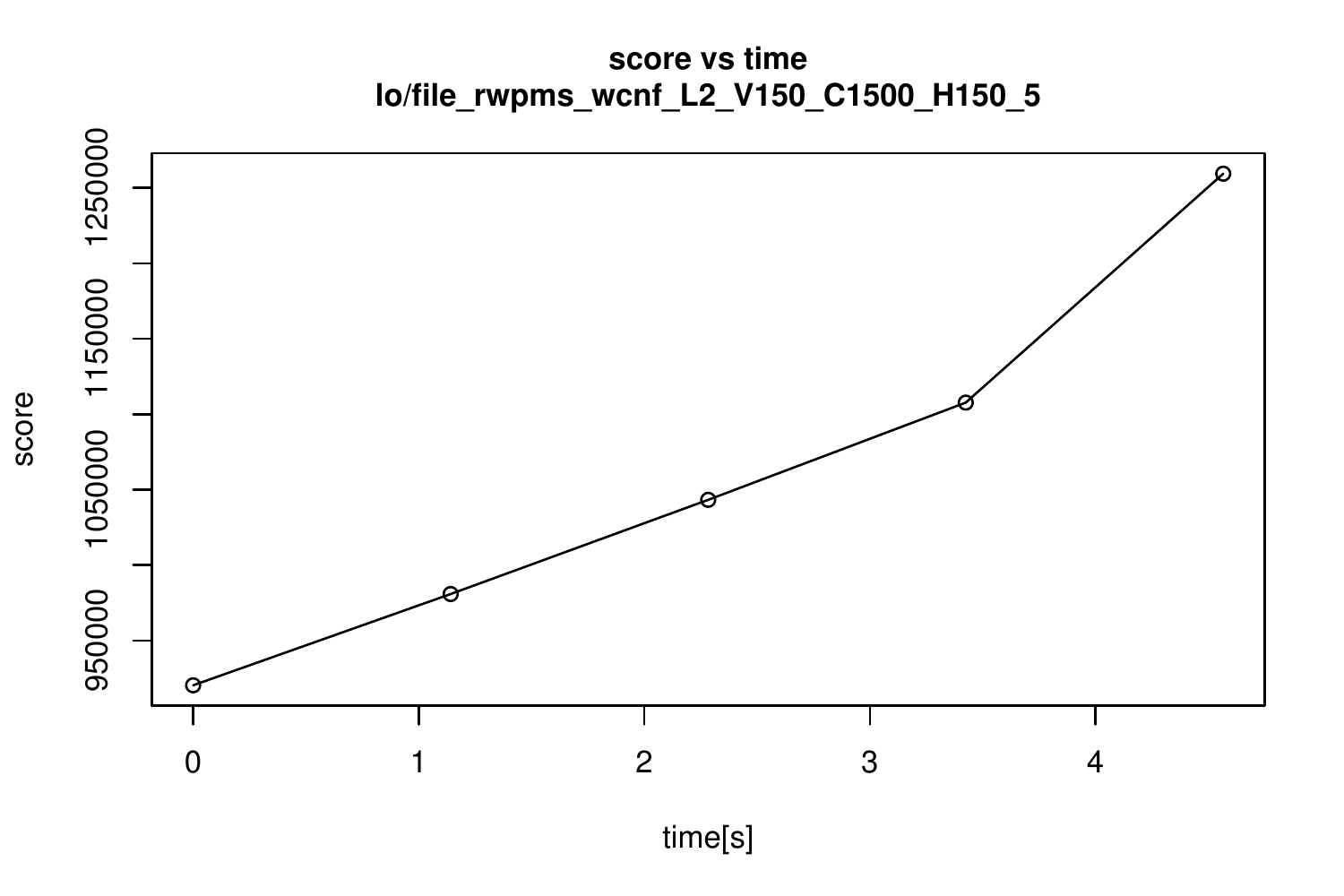}
    \label{fig_lo/file_rwpms_wcnf_L2_V150_C1500_H150_5/file_rwpms_wcnf_L2_V150_C1500_H150_5-score_vs_time}
\end{figure}

\begin{figure}[H]
    \centering
    \includegraphics[height=3.5in]{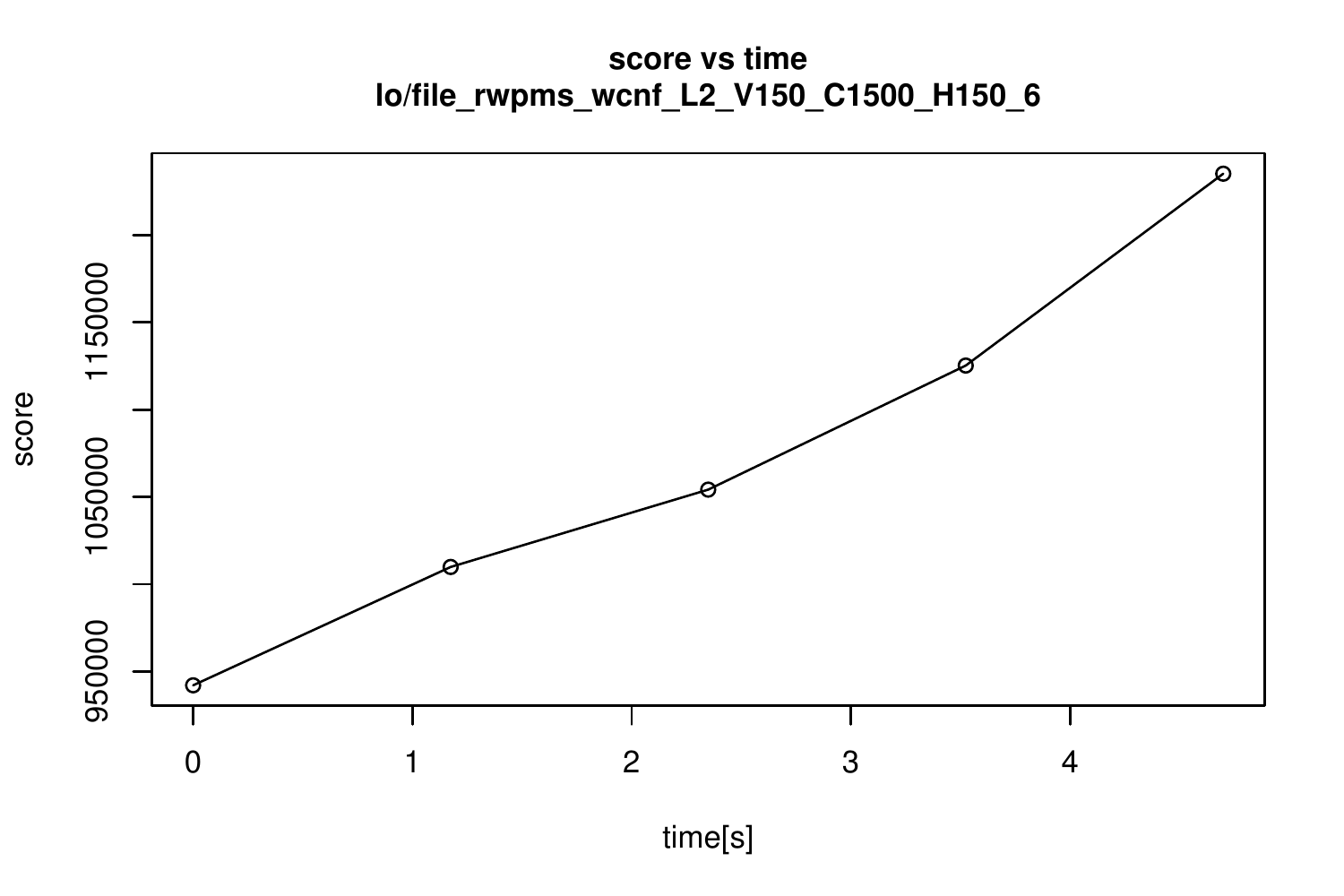}
    \label{fig_lo/file_rwpms_wcnf_L2_V150_C1500_H150_6/file_rwpms_wcnf_L2_V150_C1500_H150_6-score_vs_time}
\end{figure}

\begin{figure}[H]
    \centering
    \includegraphics[height=3.5in]{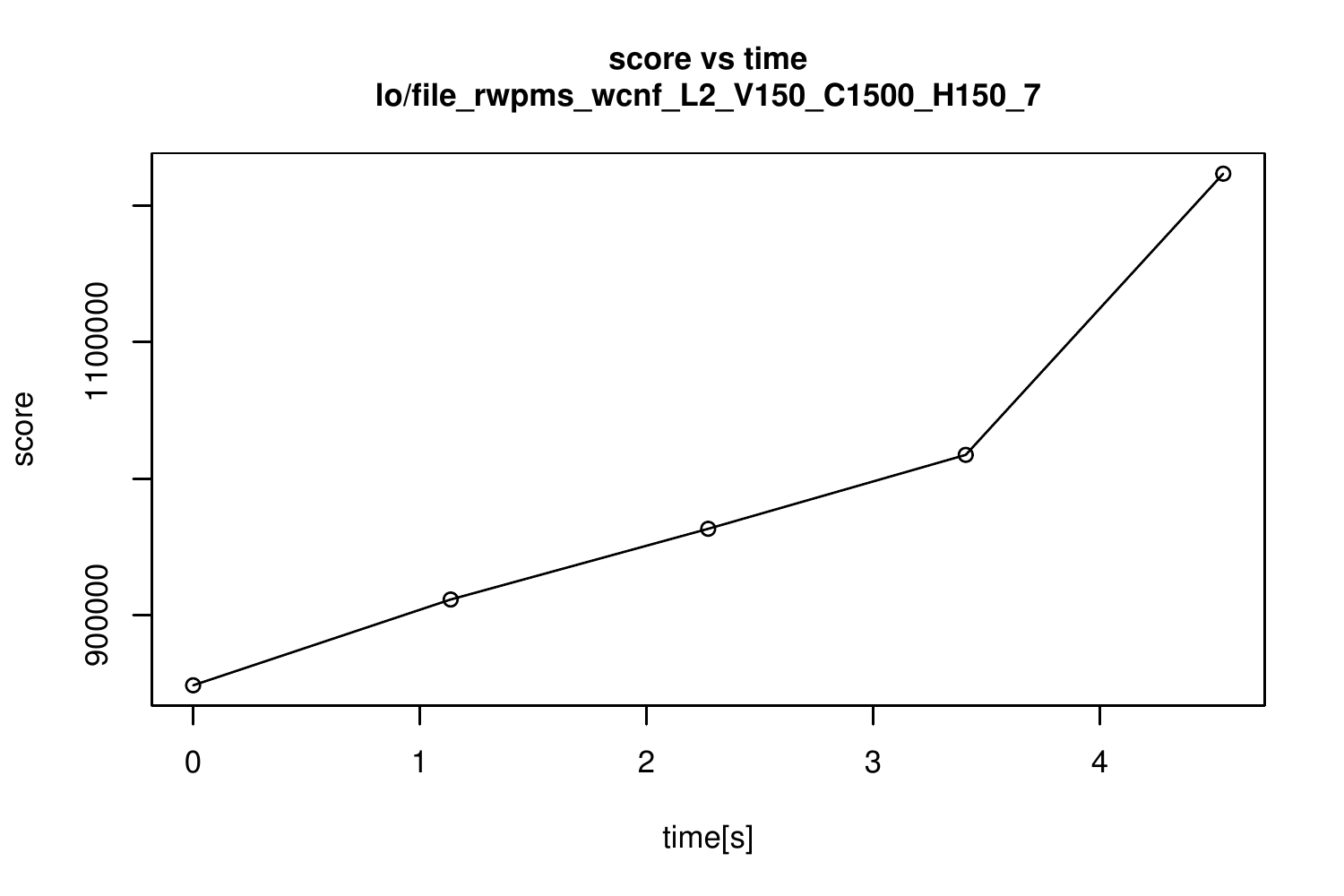}
    \label{fig_lo/file_rwpms_wcnf_L2_V150_C1500_H150_7/file_rwpms_wcnf_L2_V150_C1500_H150_7-score_vs_time}
\end{figure}

\begin{figure}[H]
    \centering
    \includegraphics[height=3.5in]{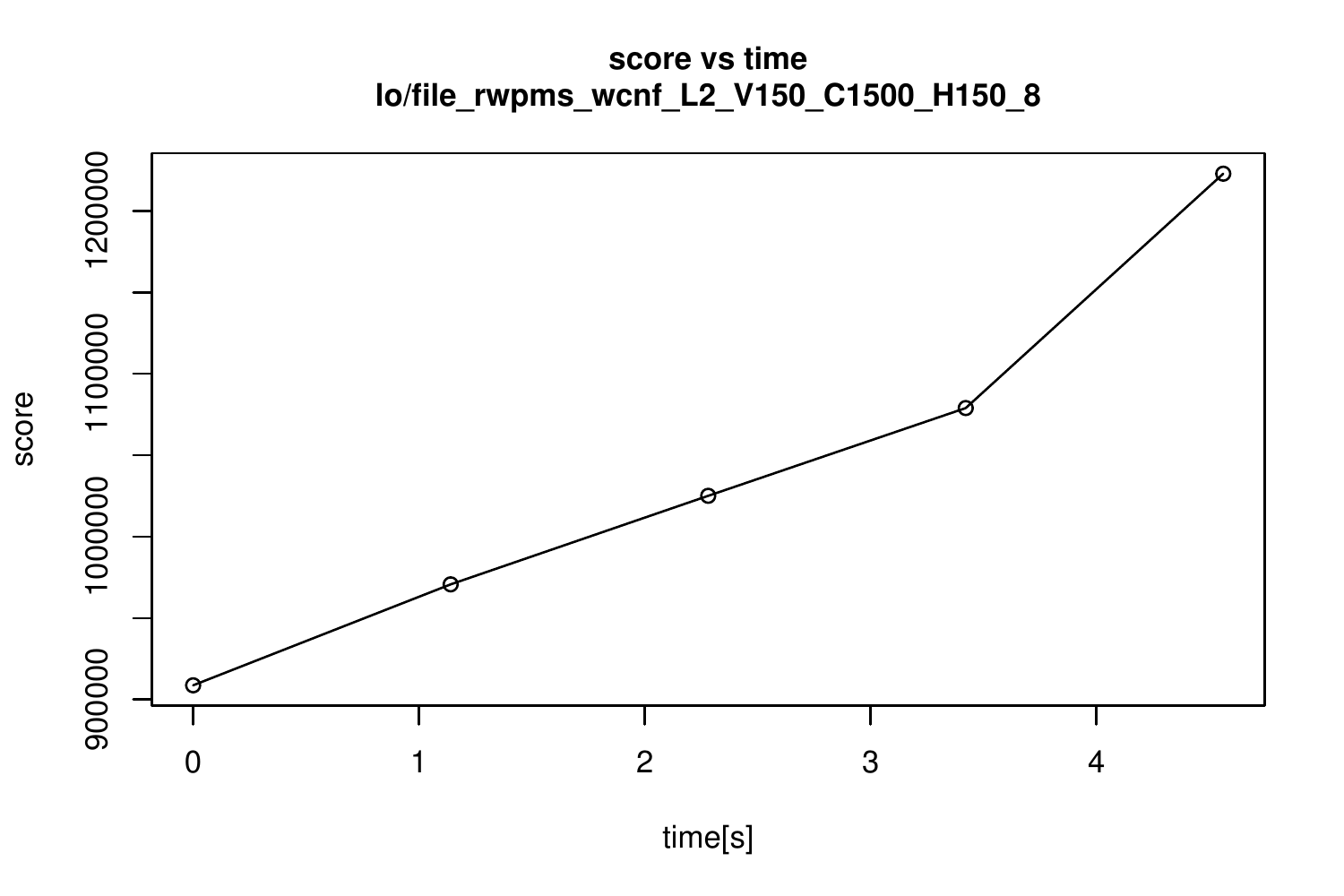}
    \label{fig_lo/file_rwpms_wcnf_L2_V150_C1500_H150_8/file_rwpms_wcnf_L2_V150_C1500_H150_8-score_vs_time}
\end{figure}

\begin{figure}[H]
    \centering
    \includegraphics[height=3.5in]{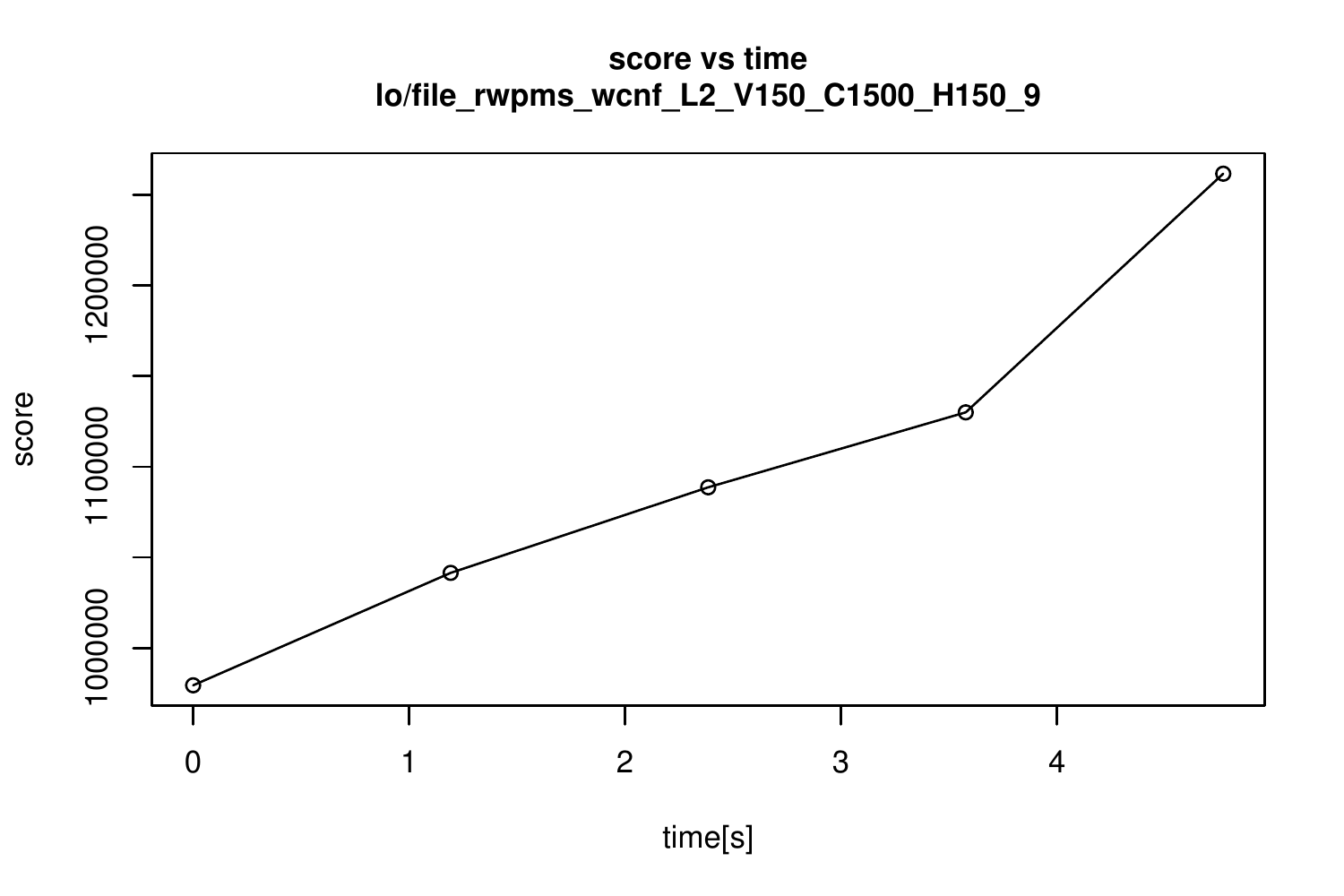}
    \label{fig_lo/file_rwpms_wcnf_L2_V150_C1500_H150_9/file_rwpms_wcnf_L2_V150_C1500_H150_9-score_vs_time}
\end{figure}

\begin{figure}[H]
    \centering
    \includegraphics[height=3.5in]{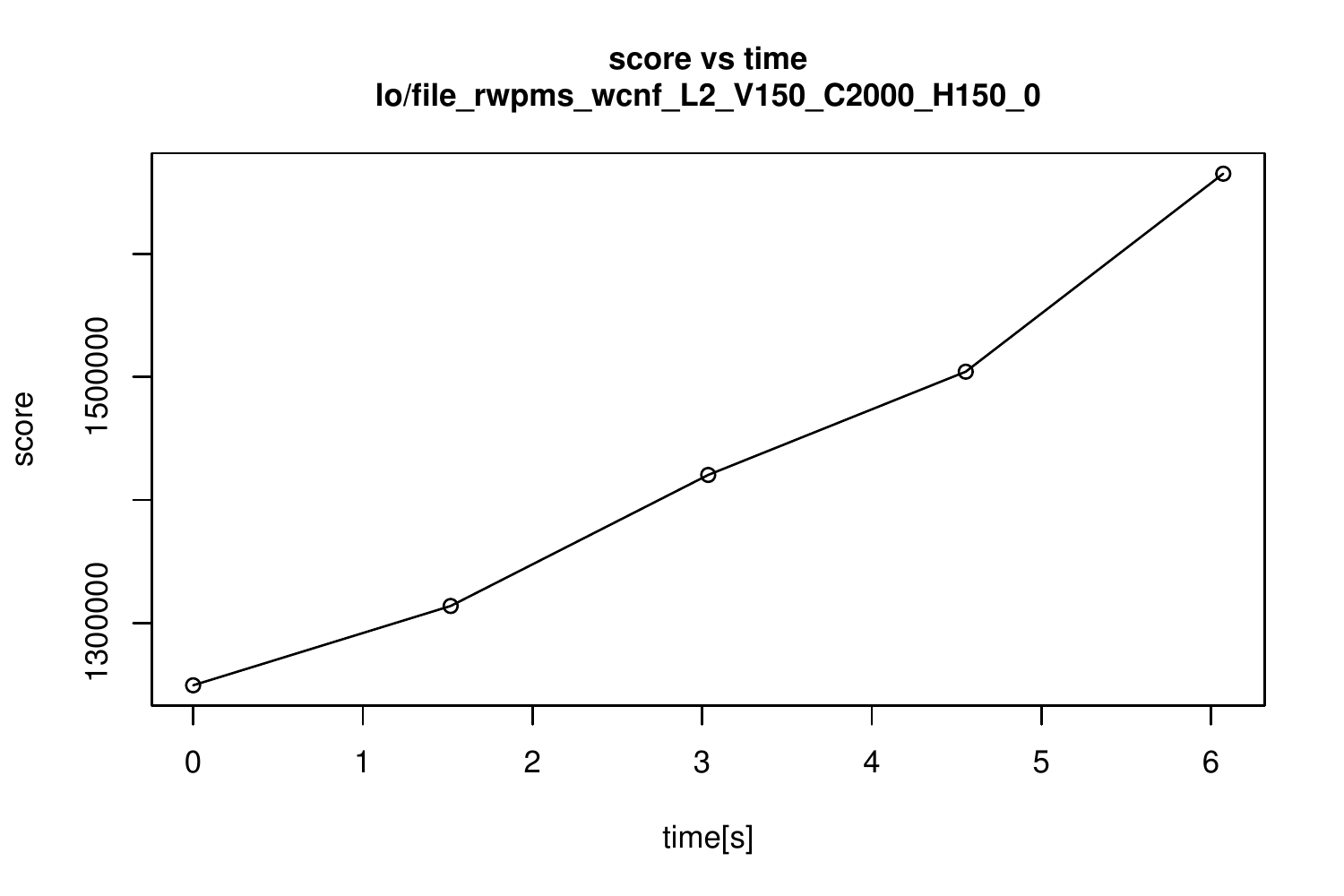}
    \label{fig_lo/file_rwpms_wcnf_L2_V150_C2000_H150_0/file_rwpms_wcnf_L2_V150_C2000_H150_0-score_vs_time}
\end{figure}

\begin{figure}[H]
    \centering
    \includegraphics[height=3.5in]{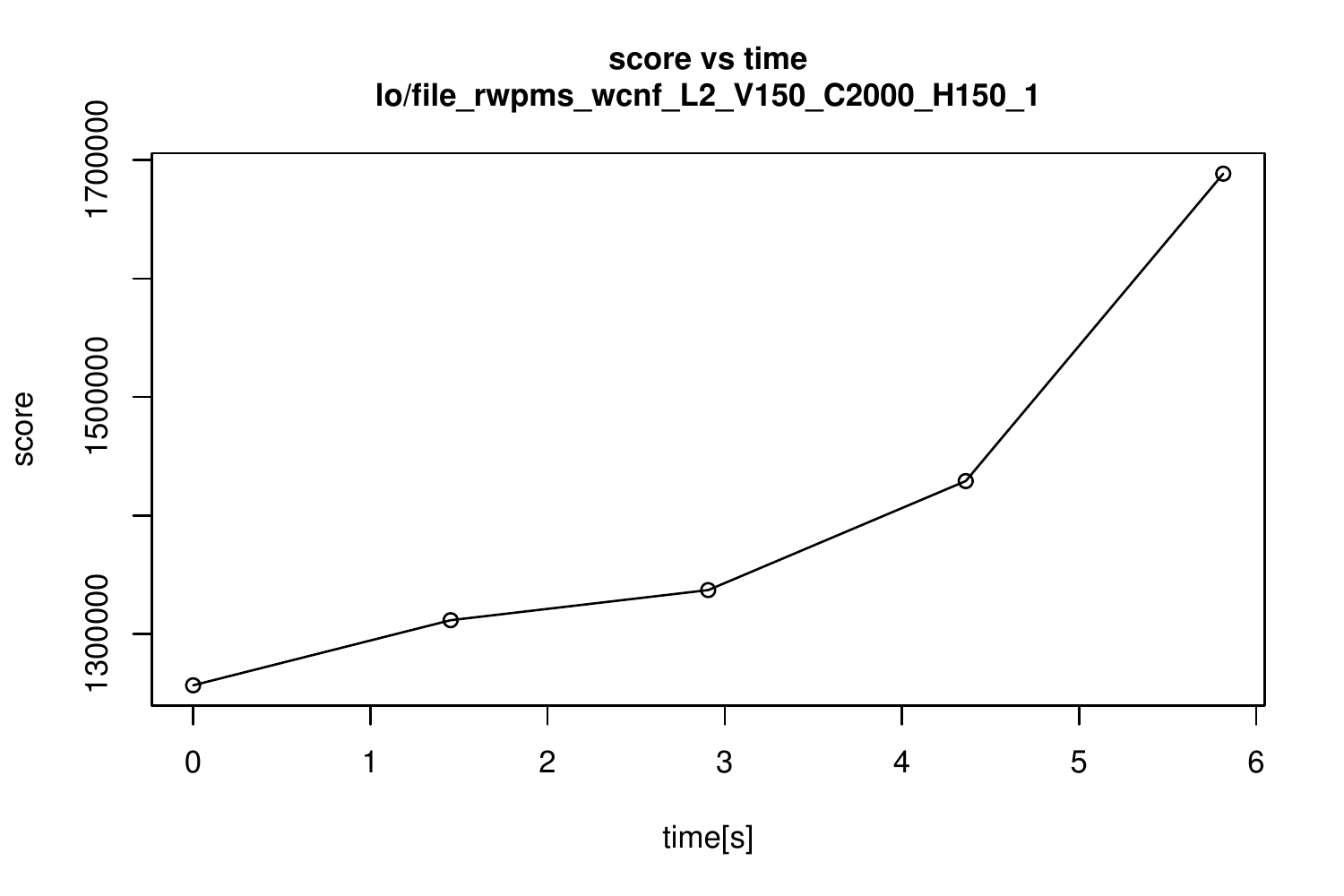}
    \label{fig_lo/file_rwpms_wcnf_L2_V150_C2000_H150_1/file_rwpms_wcnf_L2_V150_C2000_H150_1-score_vs_time}
\end{figure}

\begin{figure}[H]
    \centering
    \includegraphics[height=3.5in]{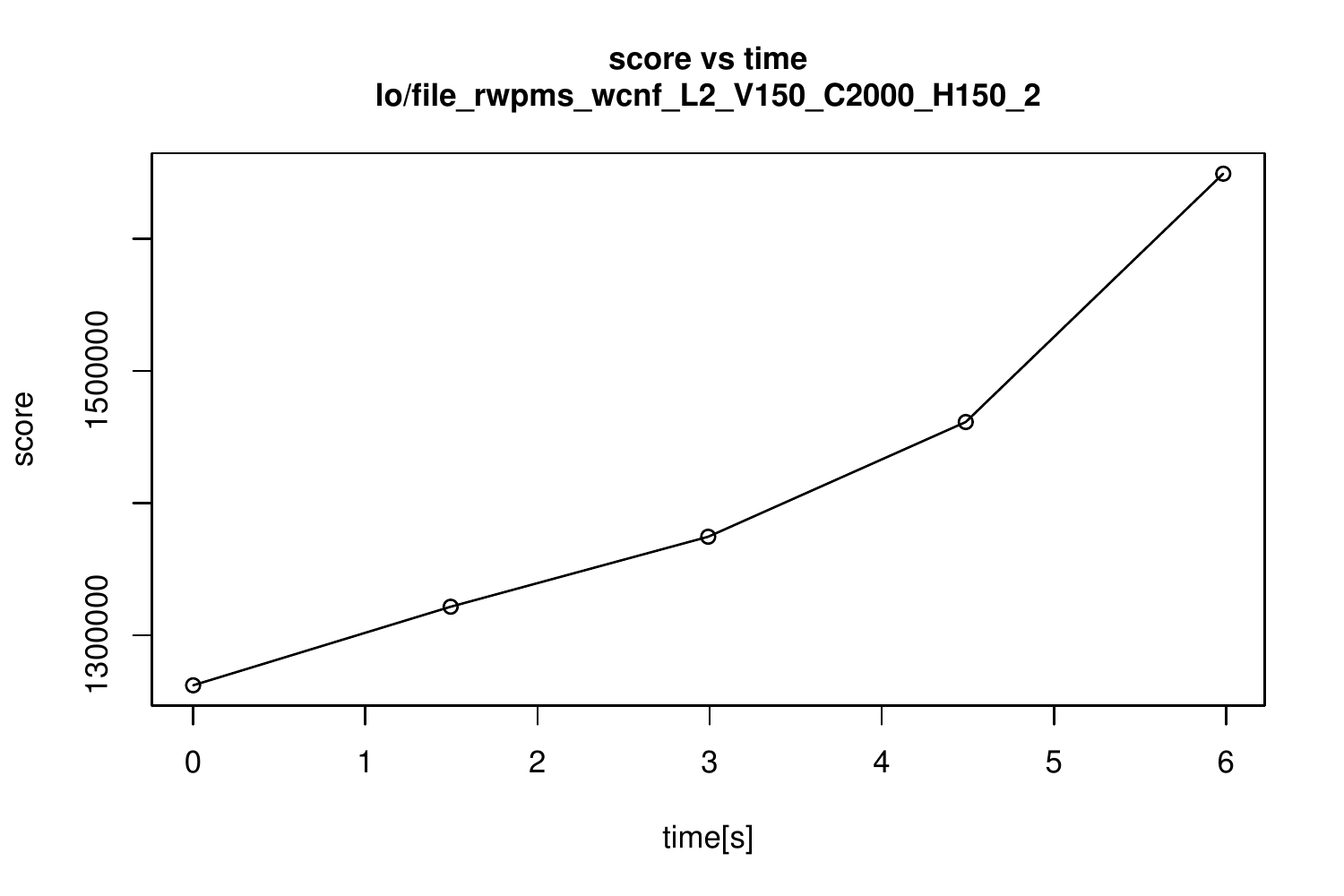}
    \label{fig_lo/file_rwpms_wcnf_L2_V150_C2000_H150_2/file_rwpms_wcnf_L2_V150_C2000_H150_2-score_vs_time}
\end{figure}

\begin{figure}[H]
    \centering
    \includegraphics[height=3.5in]{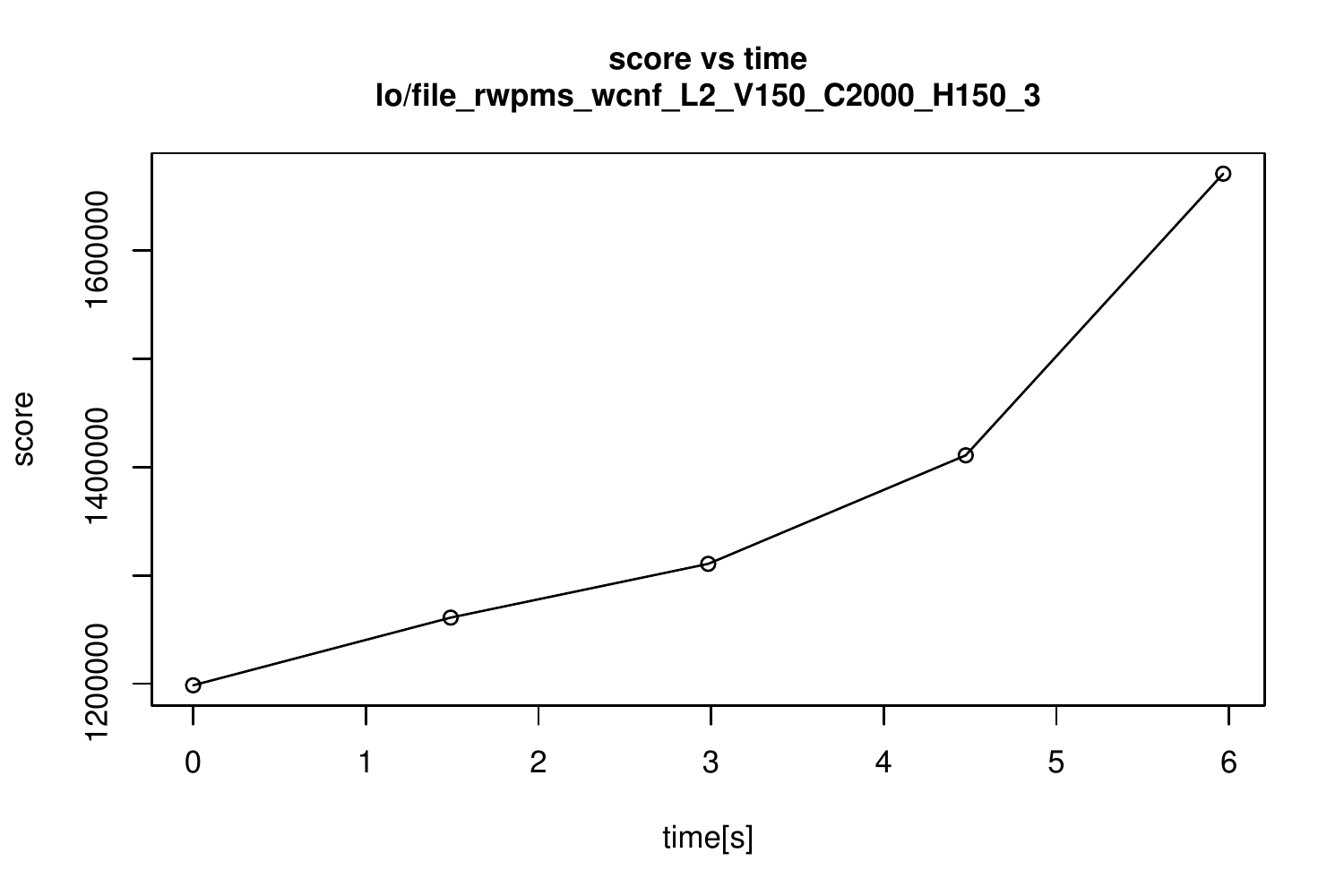}
    \label{fig_lo/file_rwpms_wcnf_L2_V150_C2000_H150_3/file_rwpms_wcnf_L2_V150_C2000_H150_3-score_vs_time}
\end{figure}

\begin{figure}[H]
    \centering
    \includegraphics[height=3.5in]{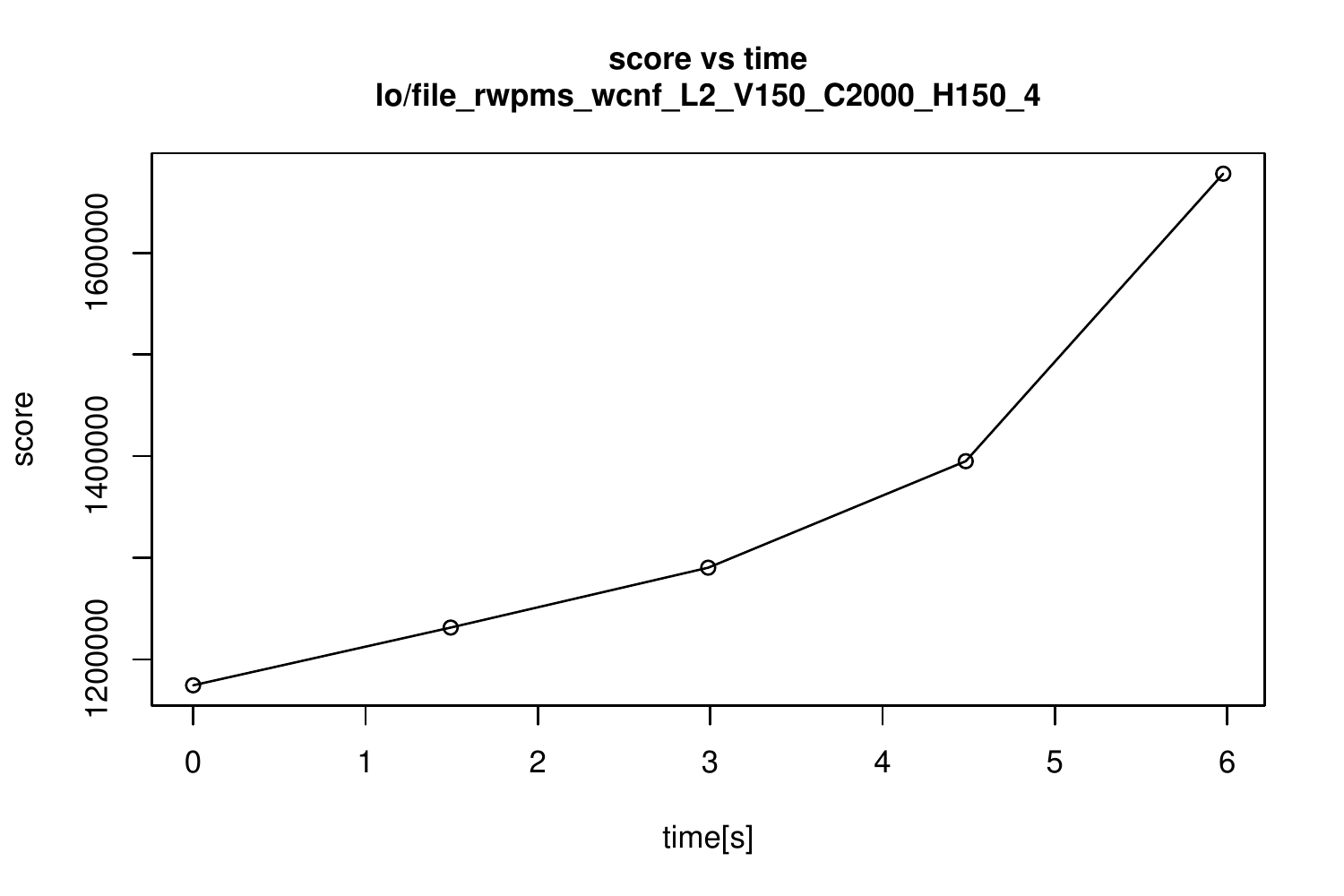}
    \label{fig_lo/file_rwpms_wcnf_L2_V150_C2000_H150_4/file_rwpms_wcnf_L2_V150_C2000_H150_4-score_vs_time}
\end{figure}

\begin{figure}[H]
    \centering
    \includegraphics[height=3.5in]{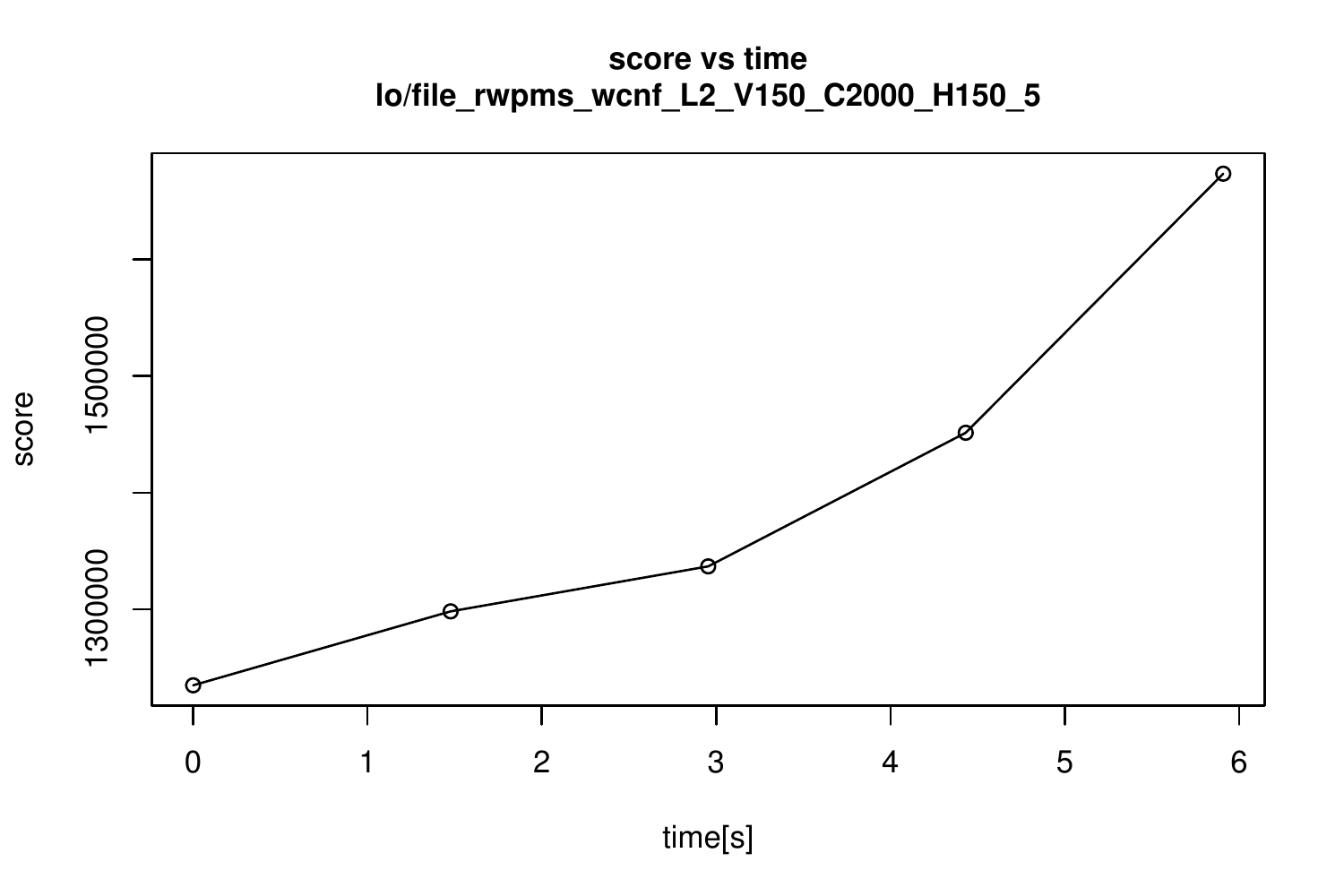}
    \label{fig_lo/file_rwpms_wcnf_L2_V150_C2000_H150_5/file_rwpms_wcnf_L2_V150_C2000_H150_5-score_vs_time}
\end{figure}

\begin{figure}[H]
    \centering
    \includegraphics[height=3.5in]{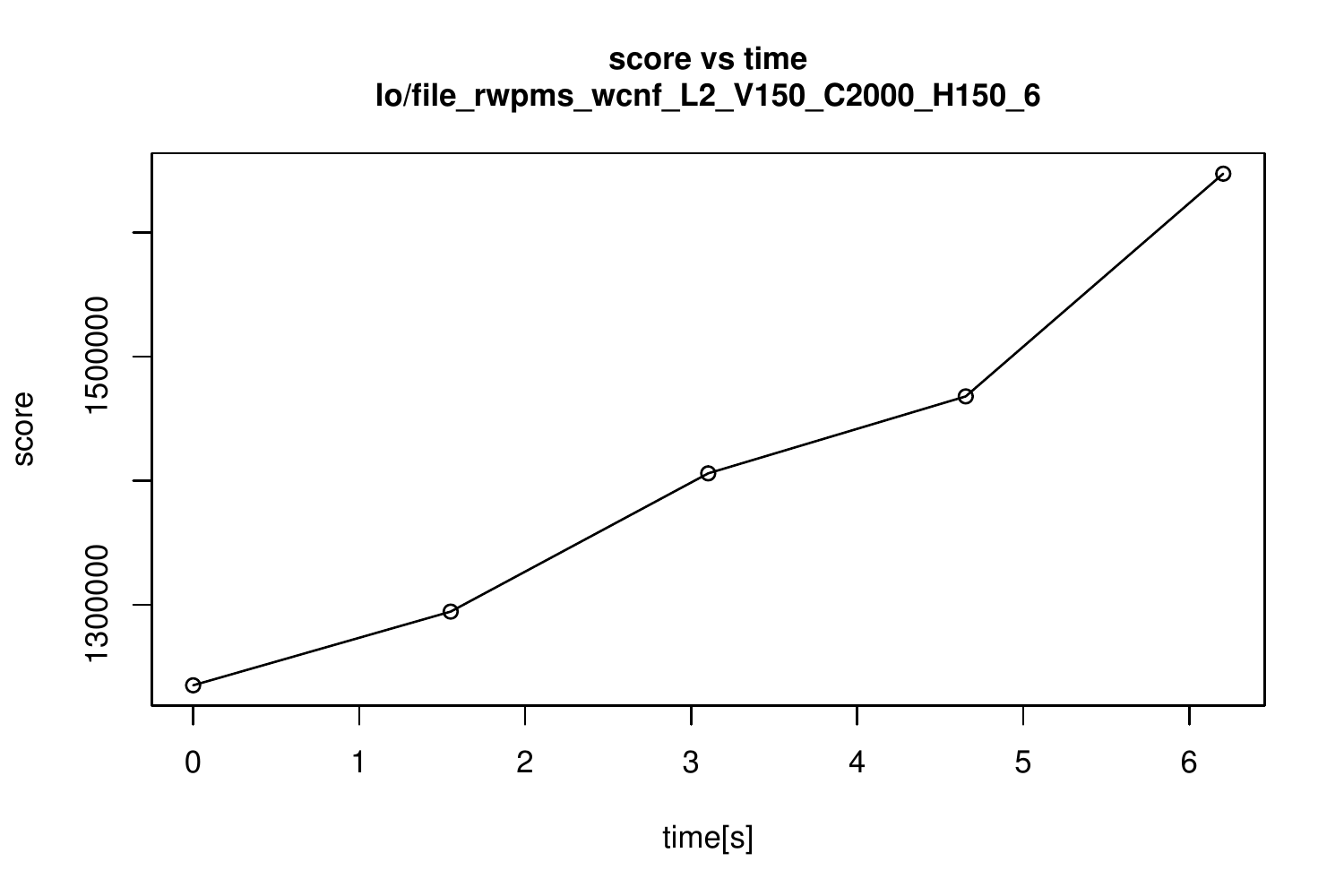}
    \label{fig_lo/file_rwpms_wcnf_L2_V150_C2000_H150_6/file_rwpms_wcnf_L2_V150_C2000_H150_6-score_vs_time}
\end{figure}

\begin{figure}[H]
    \centering
    \includegraphics[height=3.5in]{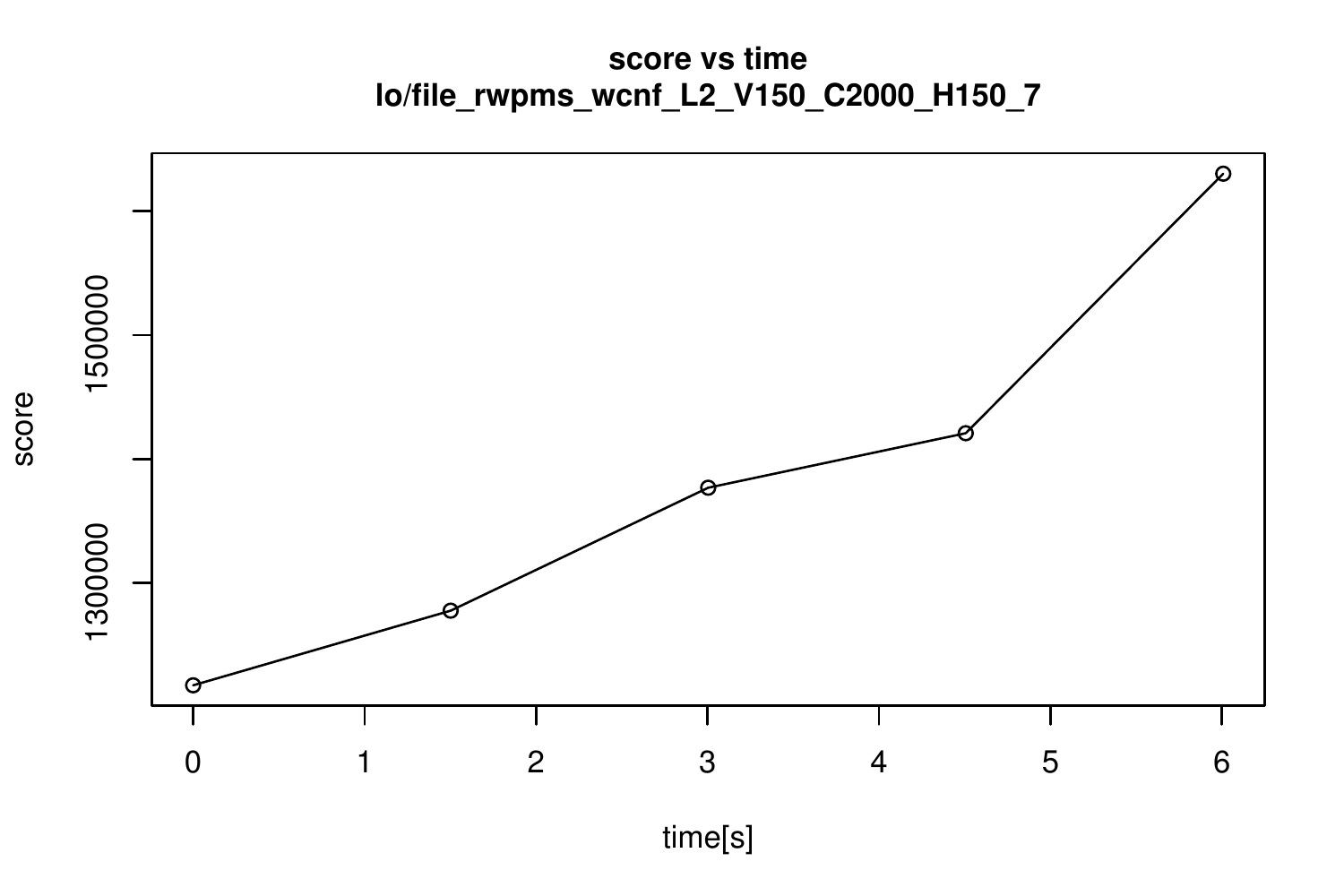}
    \label{fig_lo/file_rwpms_wcnf_L2_V150_C2000_H150_7/file_rwpms_wcnf_L2_V150_C2000_H150_7-score_vs_time}
\end{figure}

\begin{figure}[H]
    \centering
    \includegraphics[height=3.5in]{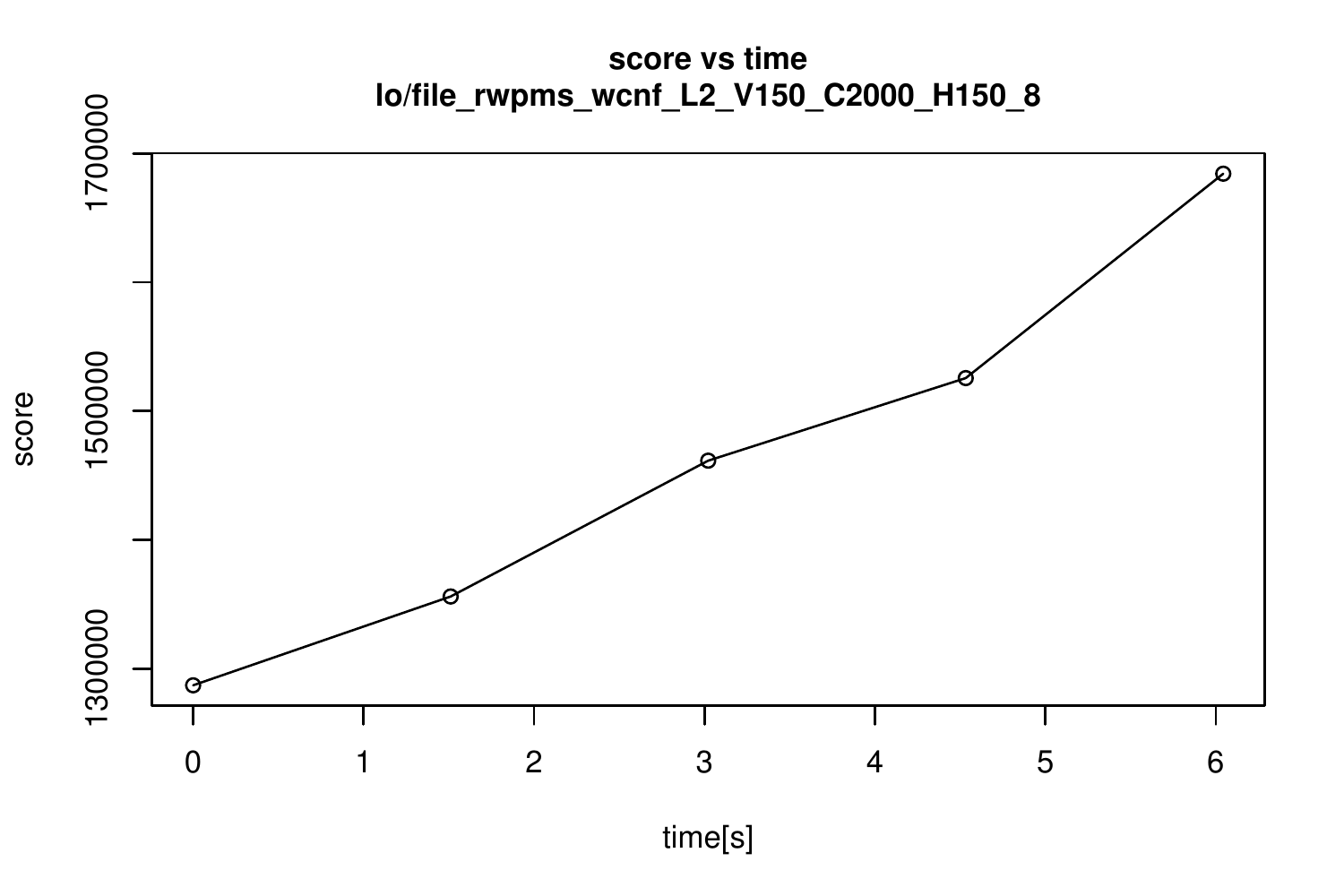}
    \label{fig_lo/file_rwpms_wcnf_L2_V150_C2000_H150_8/file_rwpms_wcnf_L2_V150_C2000_H150_8-score_vs_time}
\end{figure}

\begin{figure}[H]
    \centering
    \includegraphics[height=3.5in]{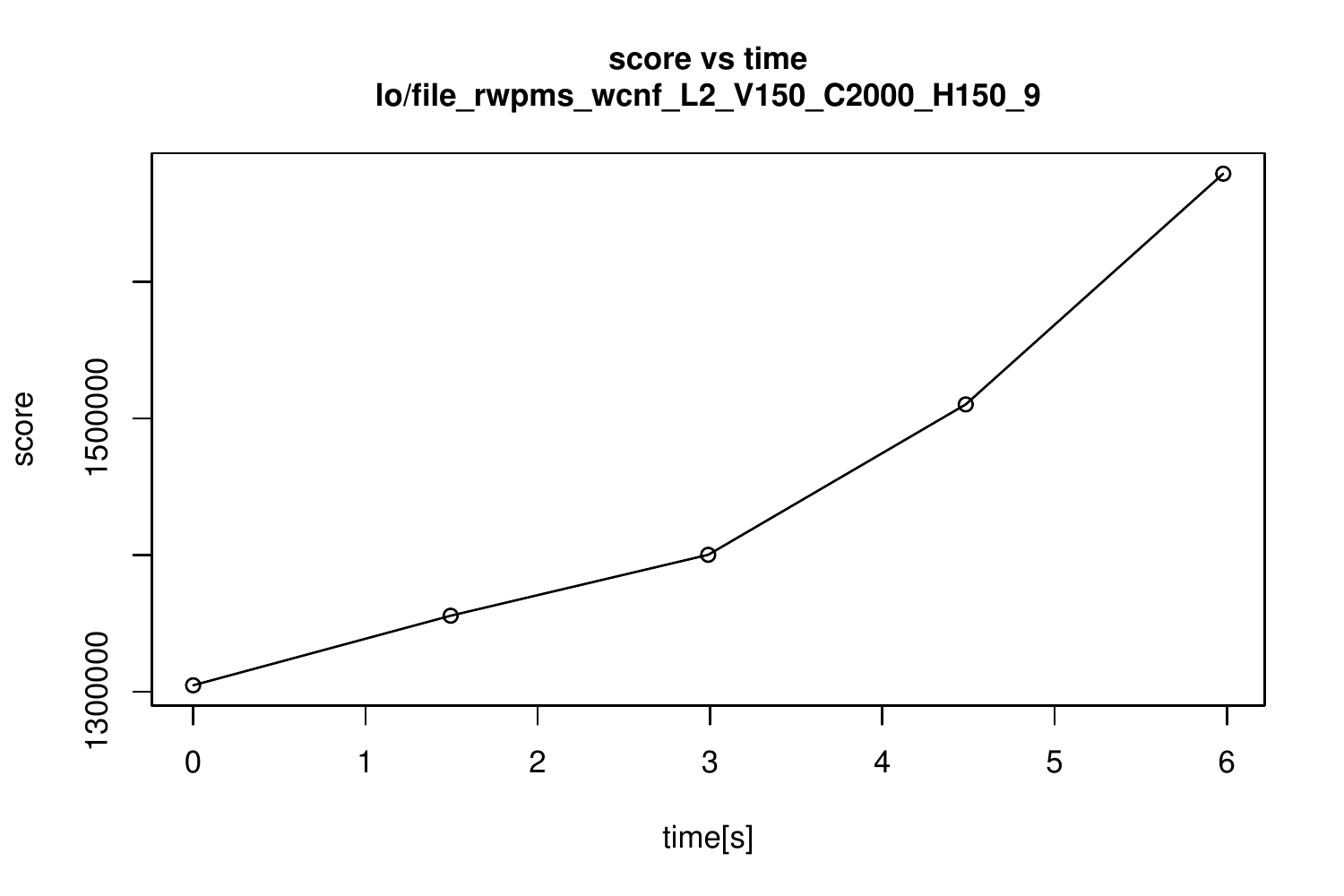}
    \label{fig_lo/file_rwpms_wcnf_L2_V150_C2000_H150_9/file_rwpms_wcnf_L2_V150_C2000_H150_9-score_vs_time}
\end{figure}

\begin{figure}[H]
    \centering
    \includegraphics[height=3.5in]{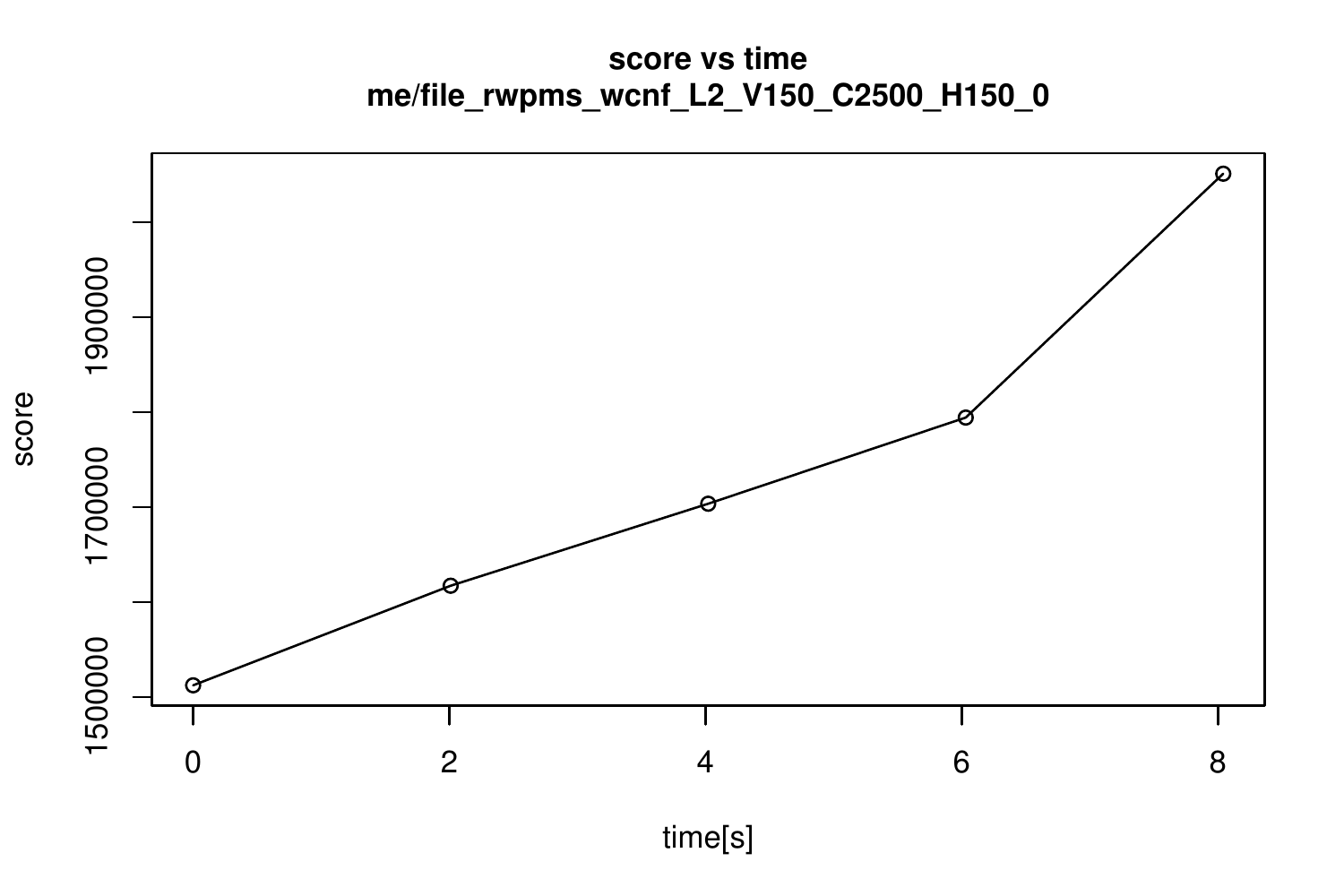}
    \label{fig_me/file_rwpms_wcnf_L2_V150_C2500_H150_0/file_rwpms_wcnf_L2_V150_C2500_H150_0-score_vs_time}
\end{figure}

\begin{figure}[H]
    \centering
    \includegraphics[height=3.5in]{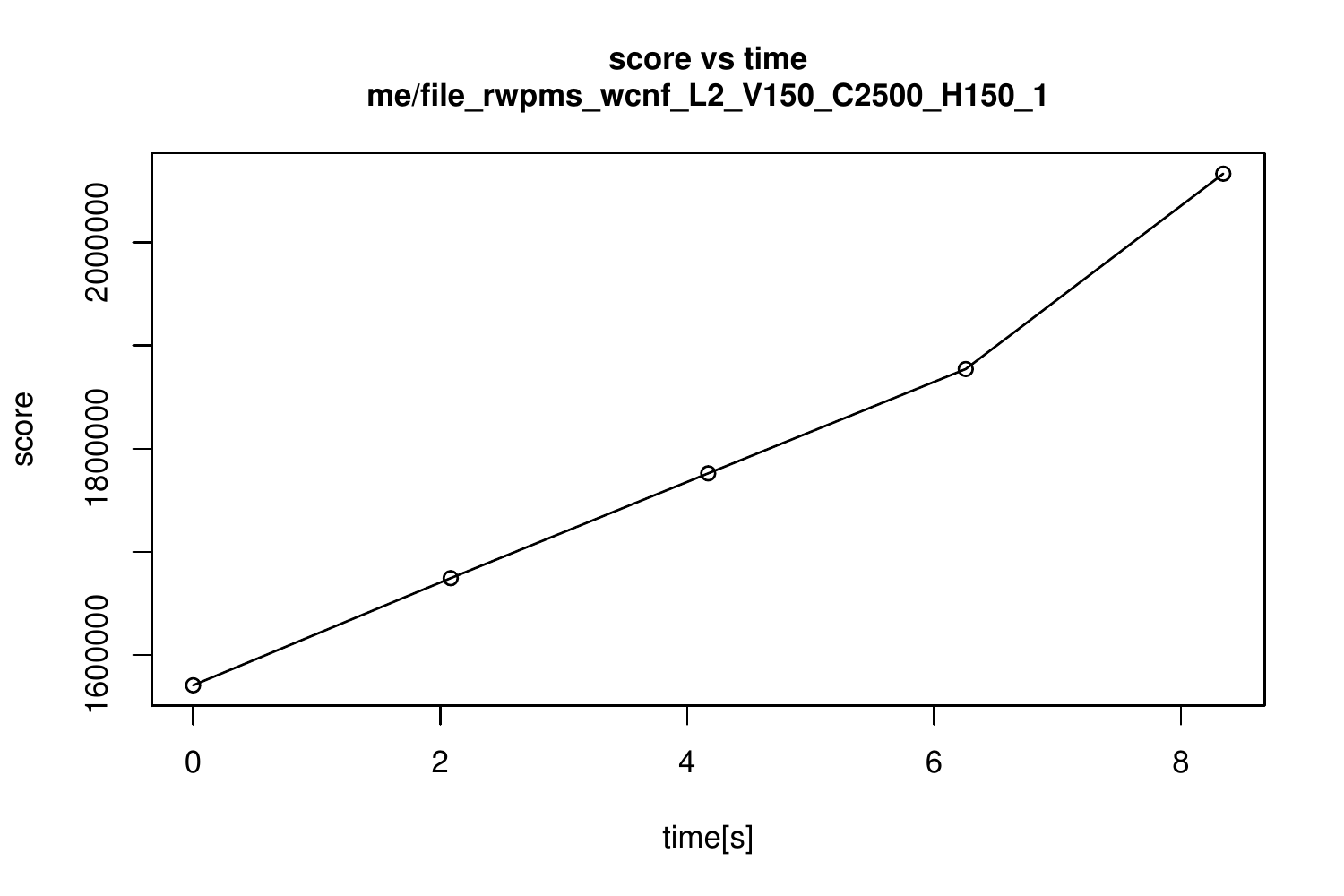}
    \label{fig_me/file_rwpms_wcnf_L2_V150_C2500_H150_1/file_rwpms_wcnf_L2_V150_C2500_H150_1-score_vs_time}
\end{figure}

\begin{figure}[H]
    \centering
    \includegraphics[height=3.5in]{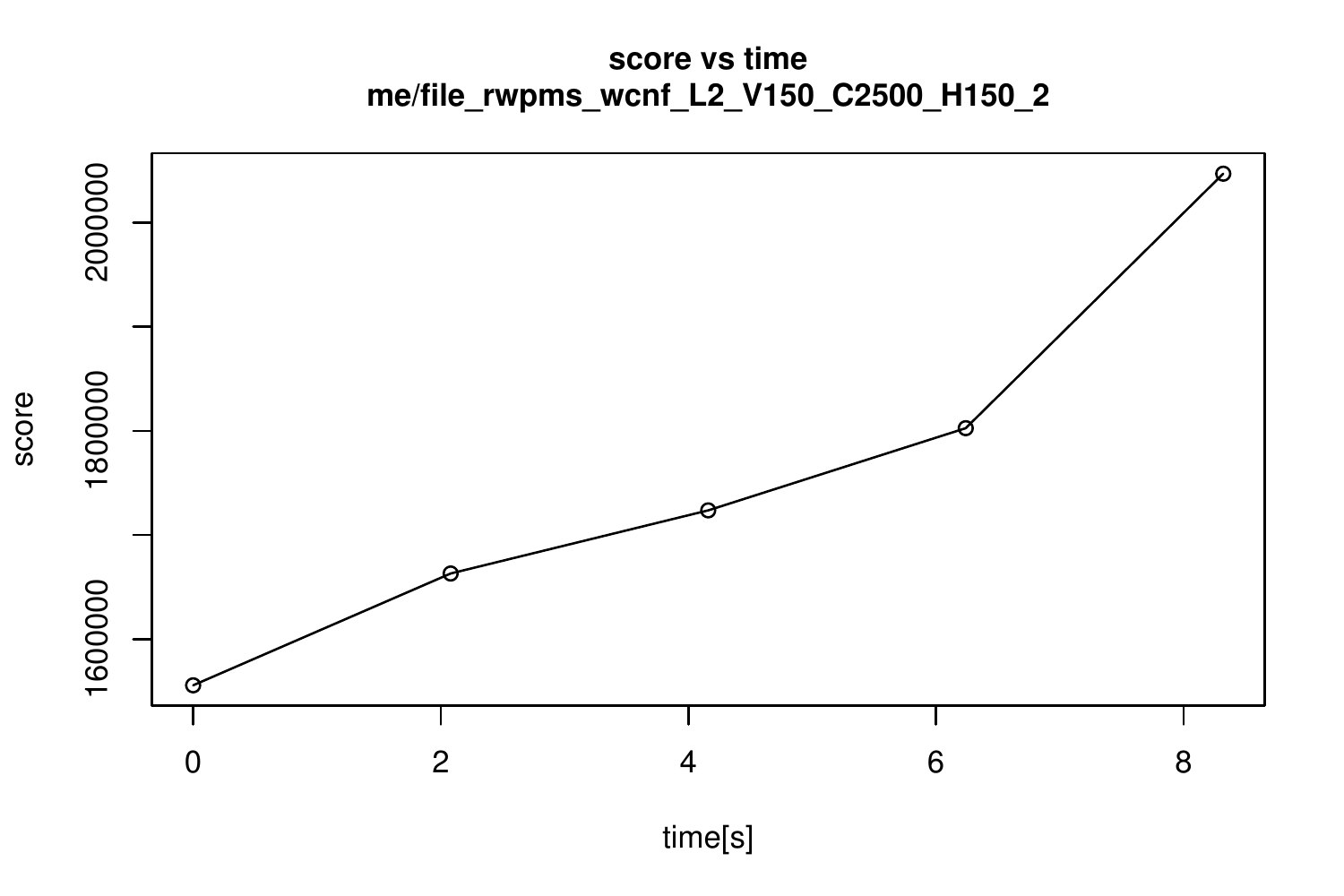}
    \label{fig_me/file_rwpms_wcnf_L2_V150_C2500_H150_2/file_rwpms_wcnf_L2_V150_C2500_H150_2-score_vs_time}
\end{figure}

\begin{figure}[H]
    \centering
    \includegraphics[height=3.5in]{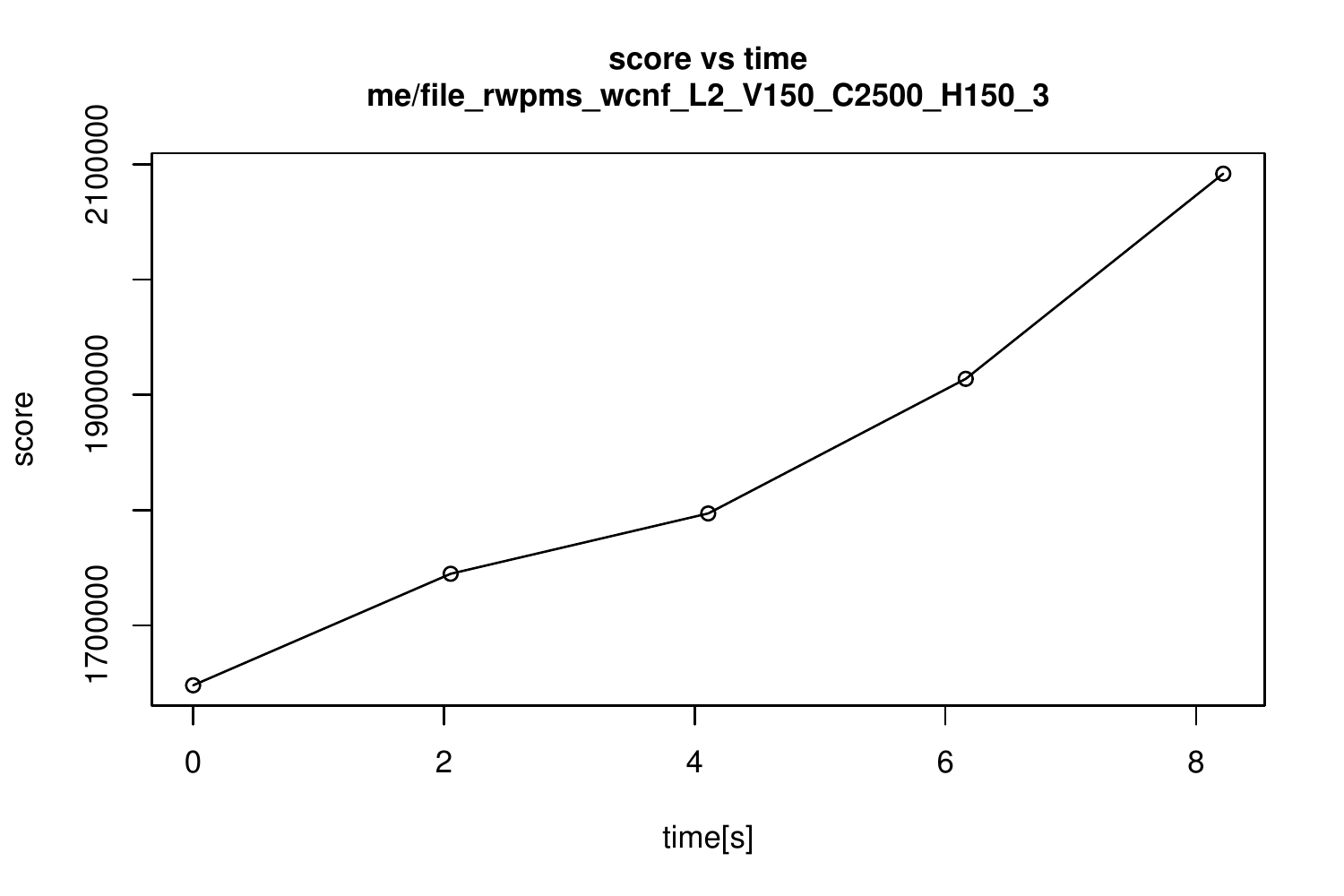}
    \label{fig_me/file_rwpms_wcnf_L2_V150_C2500_H150_3/file_rwpms_wcnf_L2_V150_C2500_H150_3-score_vs_time}
\end{figure}

\begin{figure}[H]
    \centering
    \includegraphics[height=3.5in]{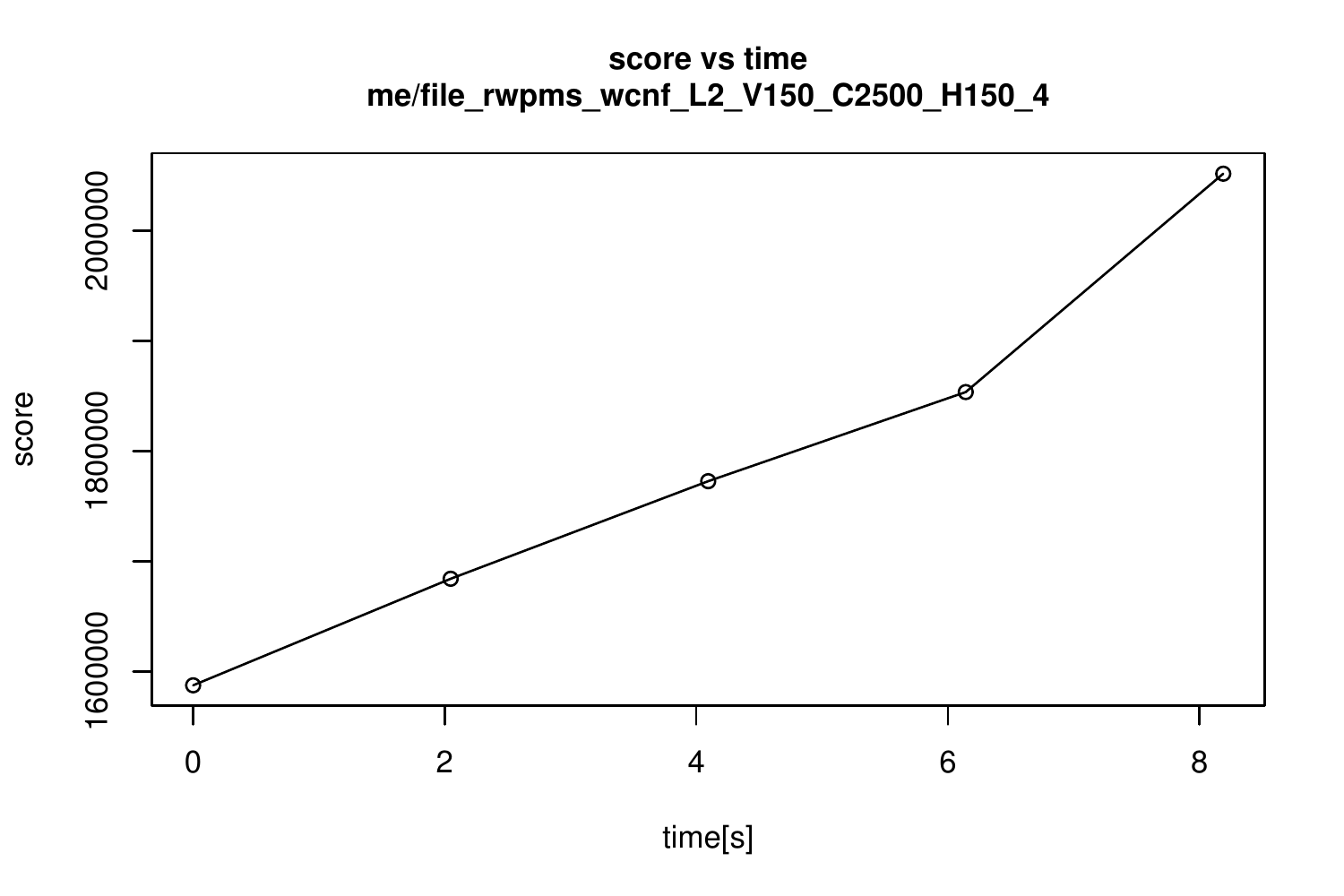}
    \label{fig_me/file_rwpms_wcnf_L2_V150_C2500_H150_4/file_rwpms_wcnf_L2_V150_C2500_H150_4-score_vs_time}
\end{figure}

\begin{figure}[H]
    \centering
    \includegraphics[height=3.5in]{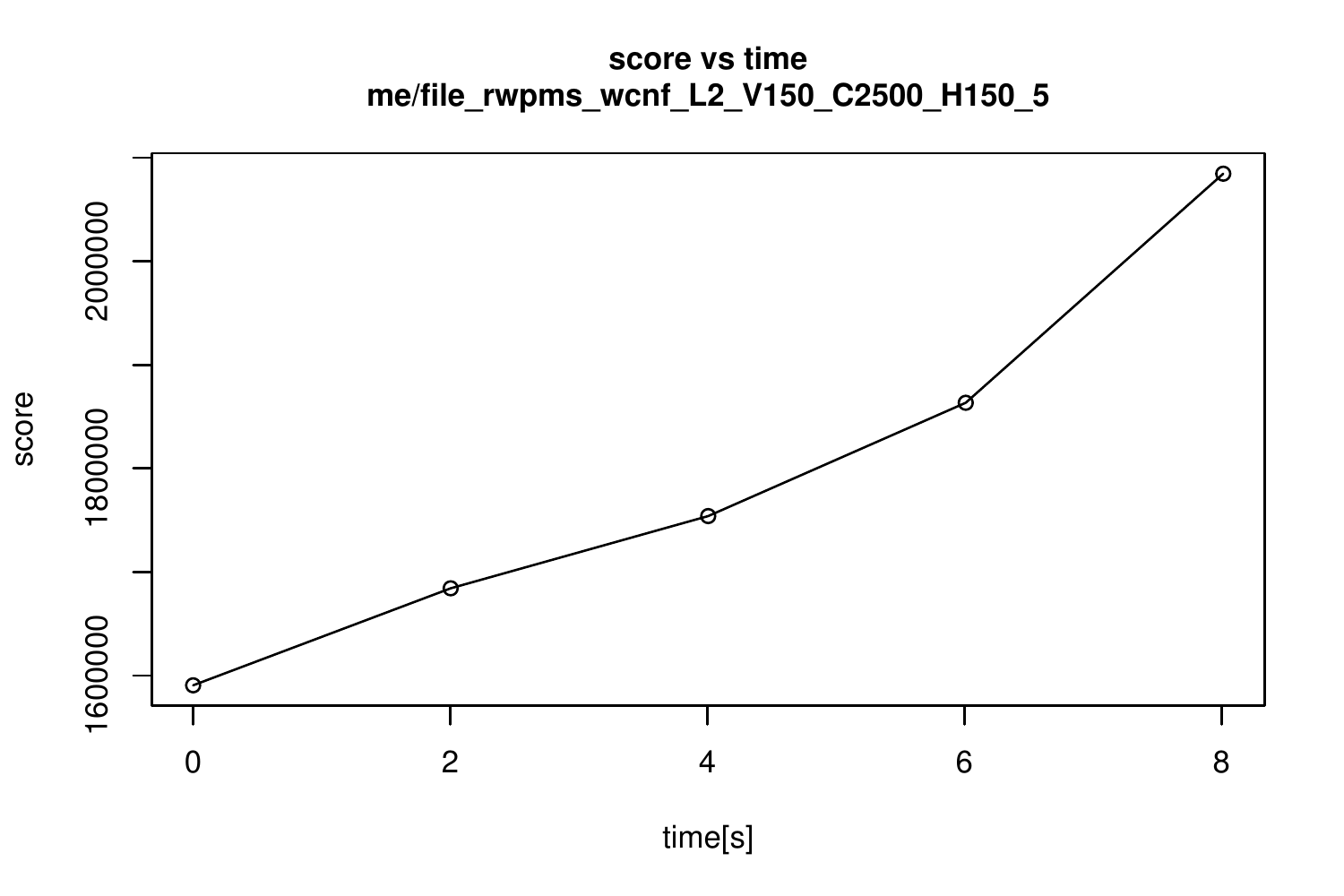}
    \label{fig_me/file_rwpms_wcnf_L2_V150_C2500_H150_5/file_rwpms_wcnf_L2_V150_C2500_H150_5-score_vs_time}
\end{figure}

\begin{figure}[H]
    \centering
    \includegraphics[height=3.5in]{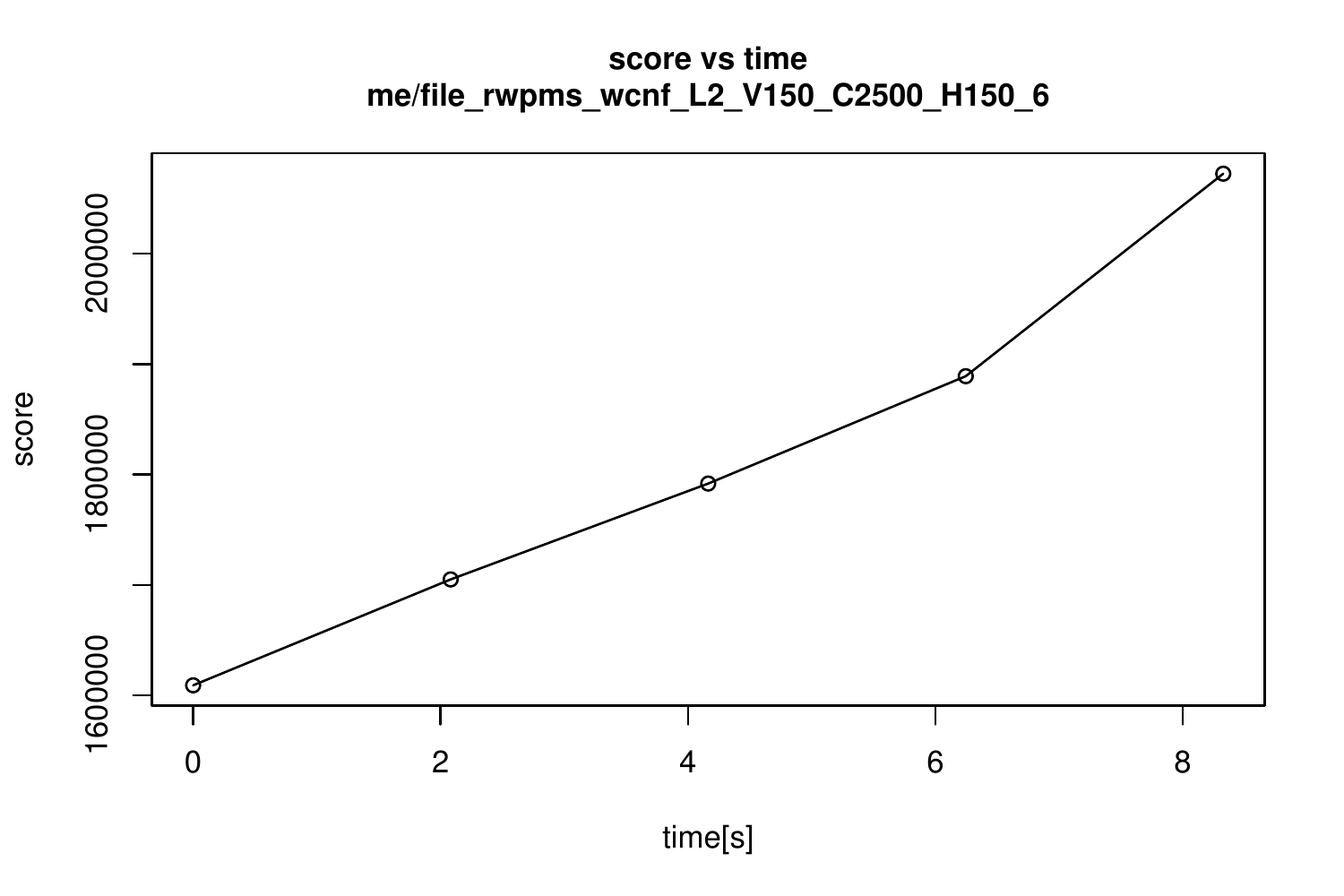}
    \label{fig_me/file_rwpms_wcnf_L2_V150_C2500_H150_6/file_rwpms_wcnf_L2_V150_C2500_H150_6-score_vs_time}
\end{figure}

\begin{figure}[H]
    \centering
    \includegraphics[height=3.5in]{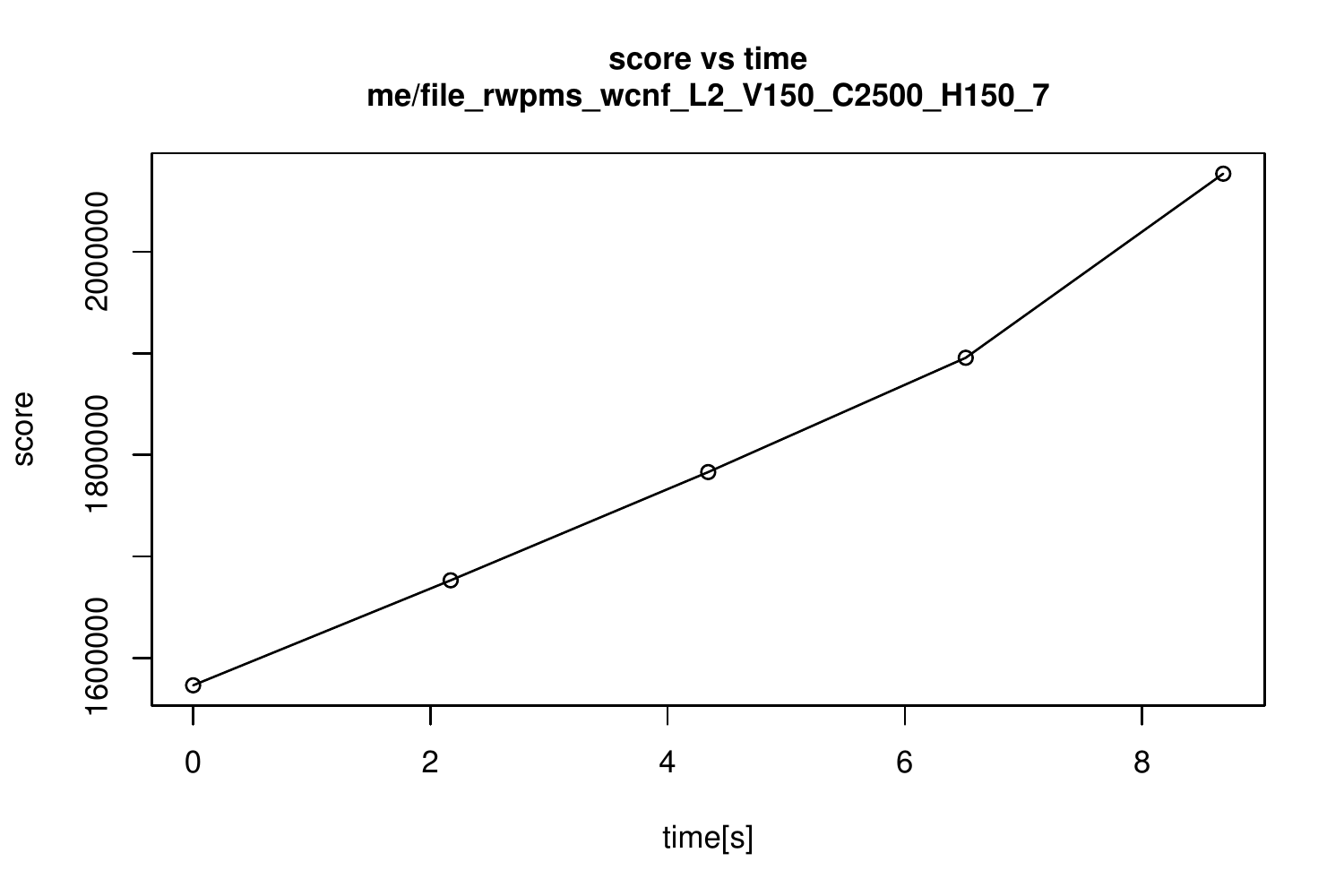}
    \label{fig_me/file_rwpms_wcnf_L2_V150_C2500_H150_7/file_rwpms_wcnf_L2_V150_C2500_H150_7-score_vs_time}
\end{figure}

\begin{figure}[H]
    \centering
    \includegraphics[height=3.5in]{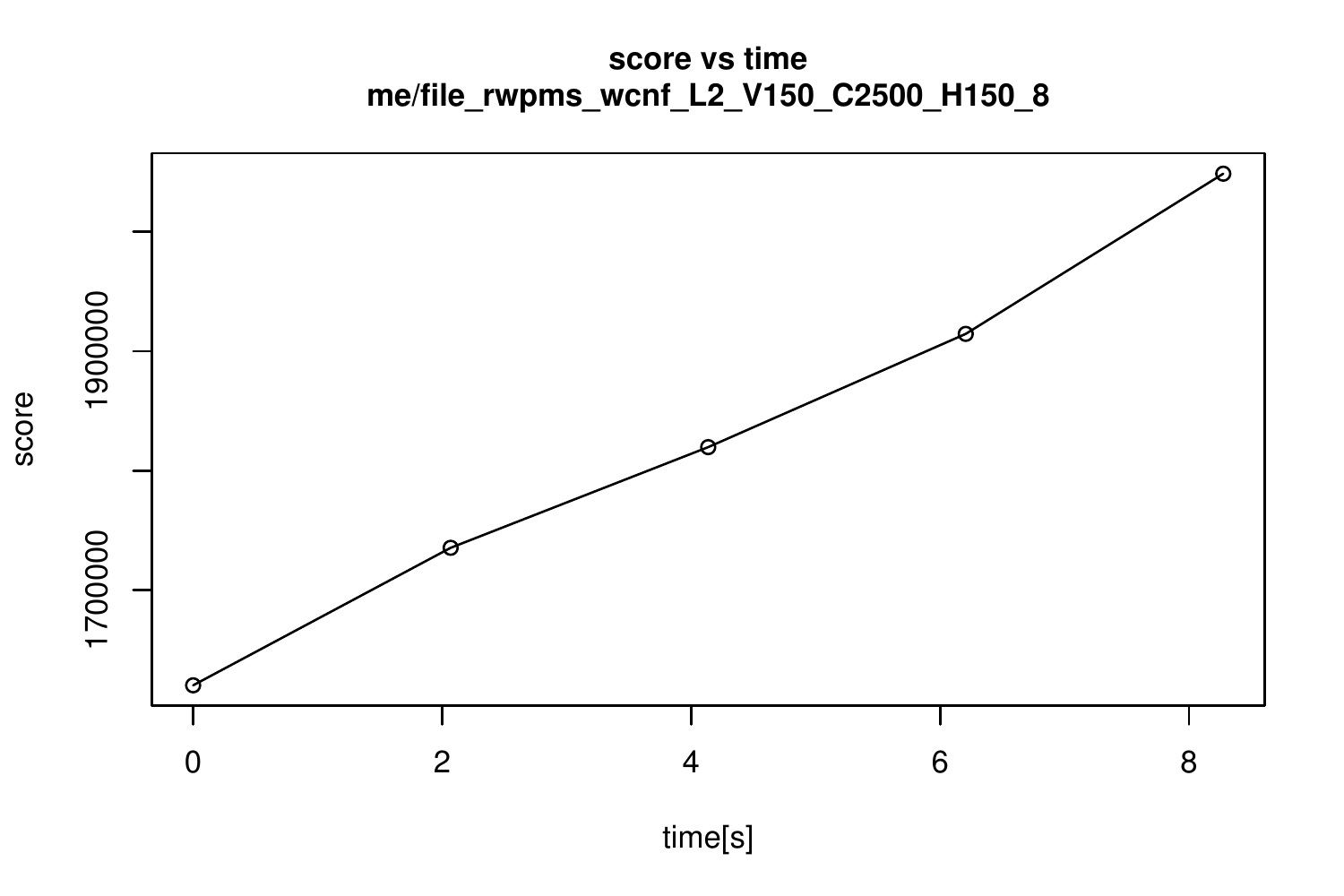}
    \label{fig_me/file_rwpms_wcnf_L2_V150_C2500_H150_8/file_rwpms_wcnf_L2_V150_C2500_H150_8-score_vs_time}
\end{figure}

\begin{figure}[H]
    \centering
    \includegraphics[height=3.5in]{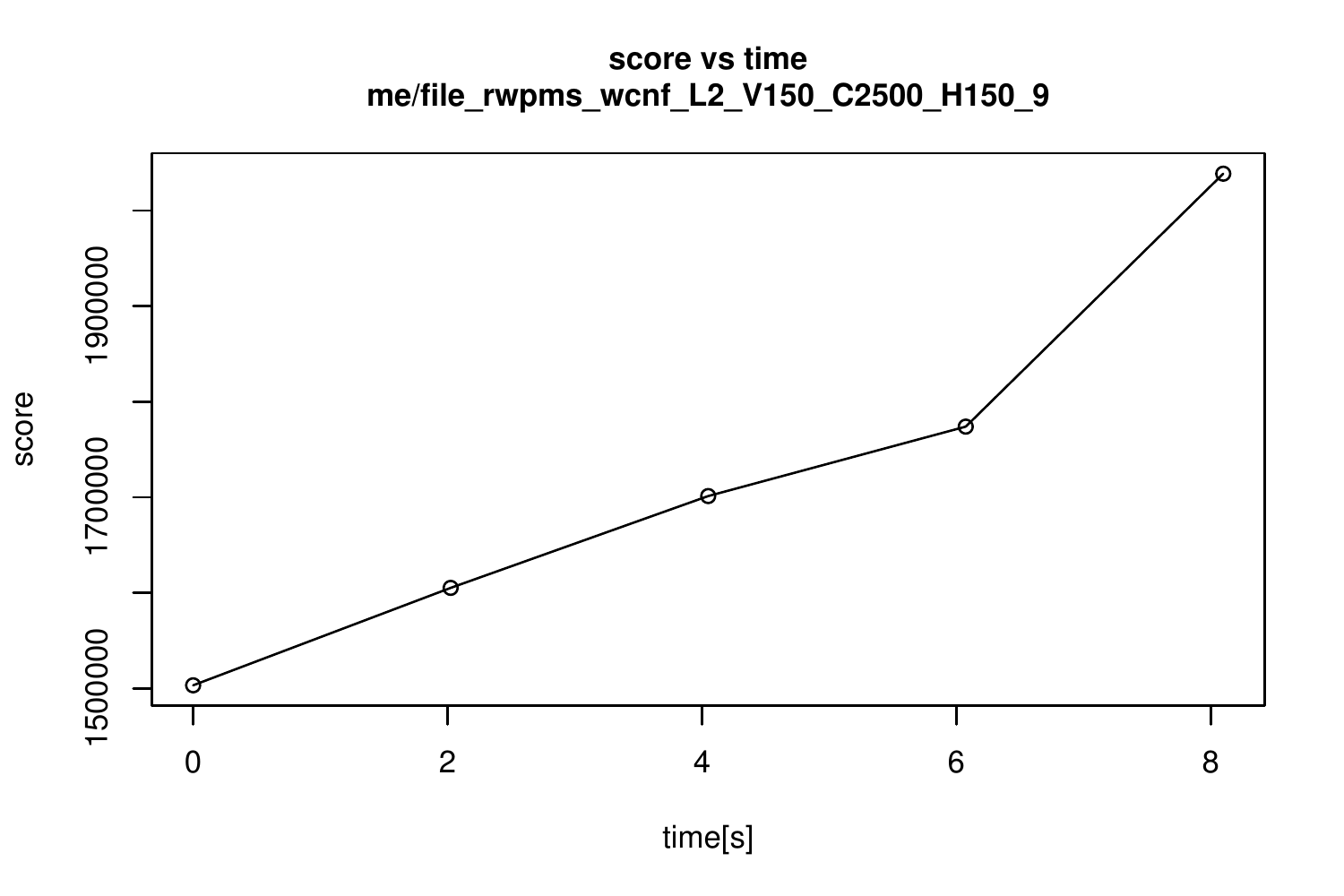}
    \label{fig_me/file_rwpms_wcnf_L2_V150_C2500_H150_9/file_rwpms_wcnf_L2_V150_C2500_H150_9-score_vs_time}
\end{figure}

\begin{figure}[H]
    \centering
    \includegraphics[height=3.5in]{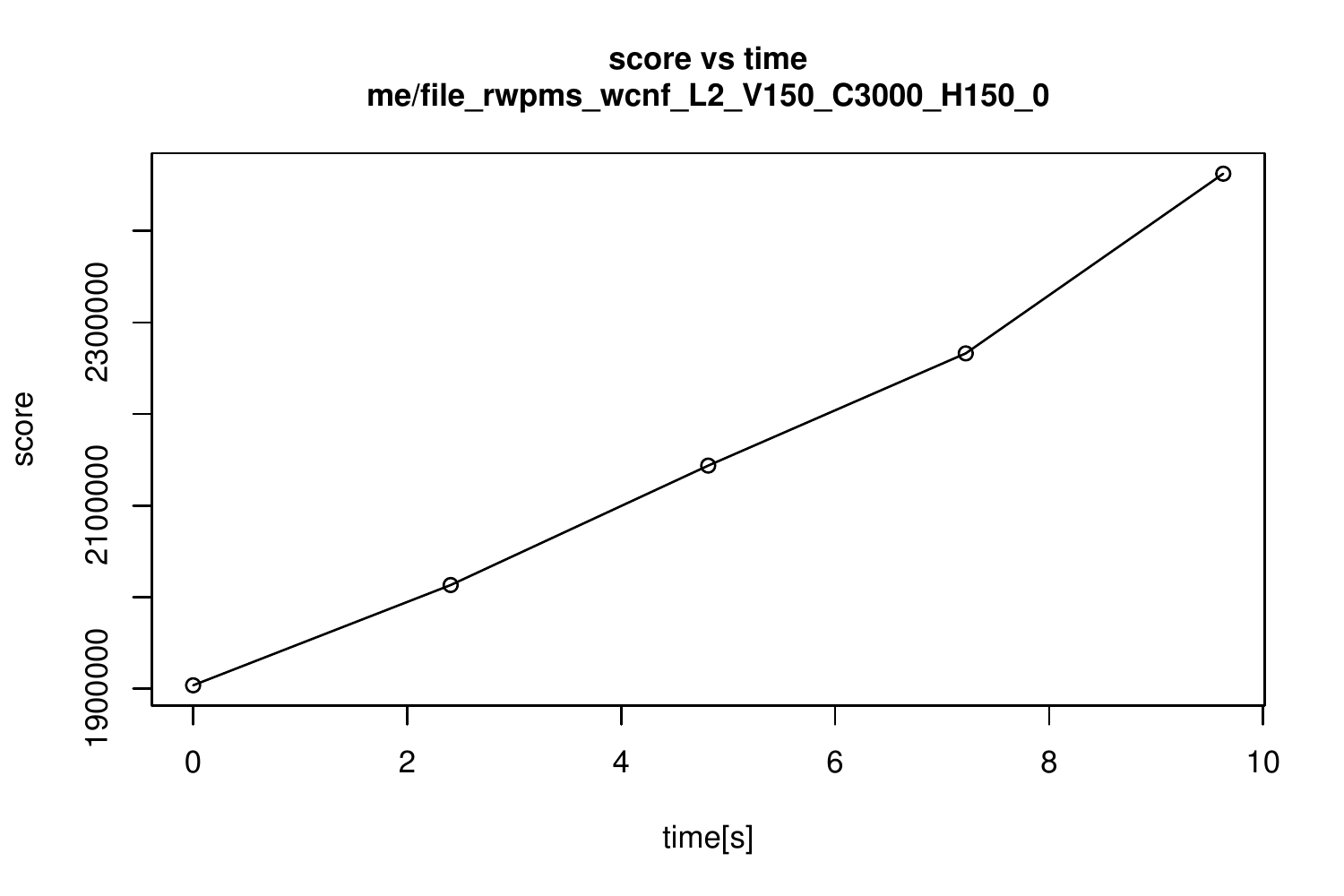}
    \label{fig_me/file_rwpms_wcnf_L2_V150_C3000_H150_0/file_rwpms_wcnf_L2_V150_C3000_H150_0-score_vs_time}
\end{figure}

\begin{figure}[H]
    \centering
    \includegraphics[height=3.5in]{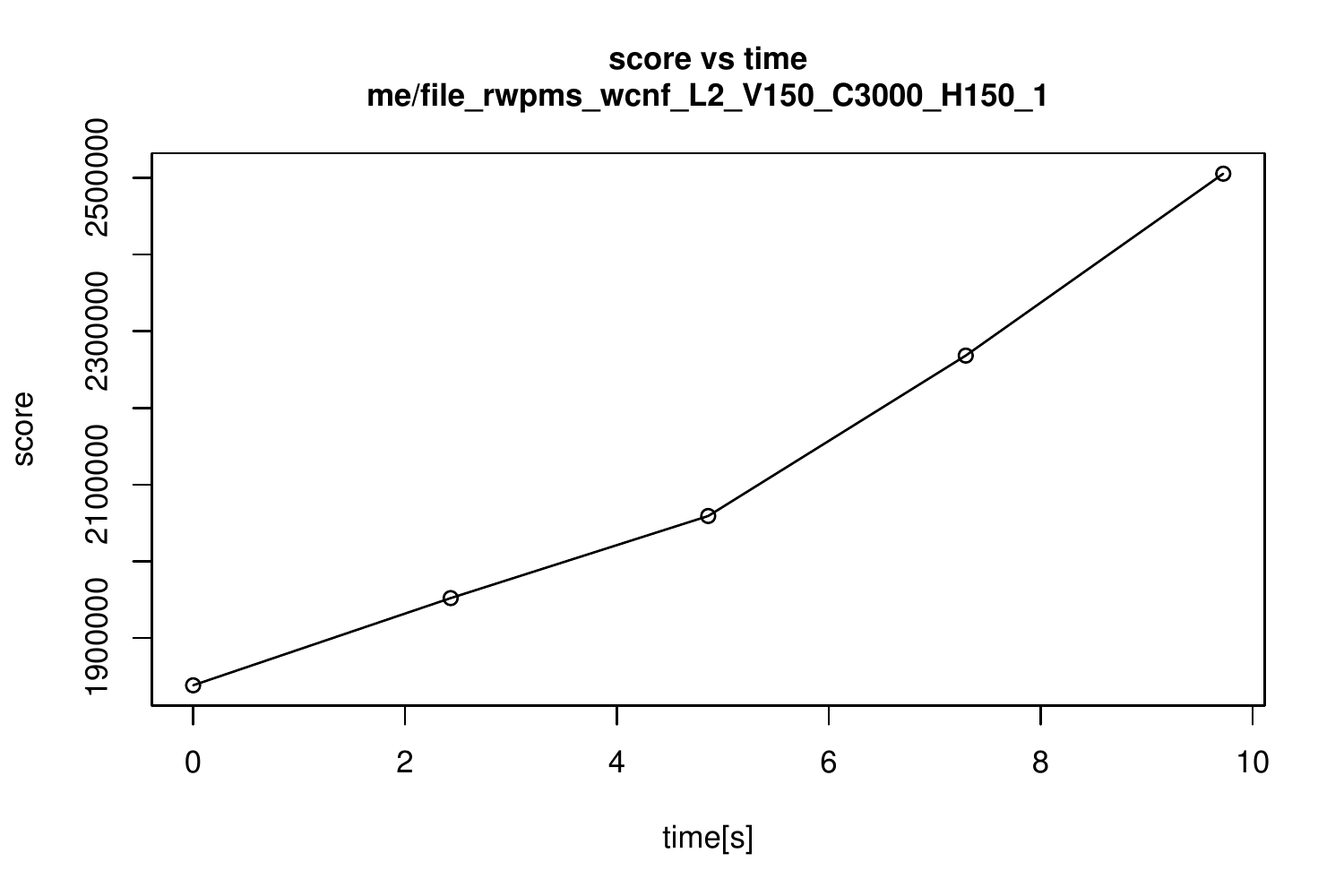}
    \label{fig_me/file_rwpms_wcnf_L2_V150_C3000_H150_1/file_rwpms_wcnf_L2_V150_C3000_H150_1-score_vs_time}
\end{figure}

\begin{figure}[H]
    \centering
    \includegraphics[height=3.5in]{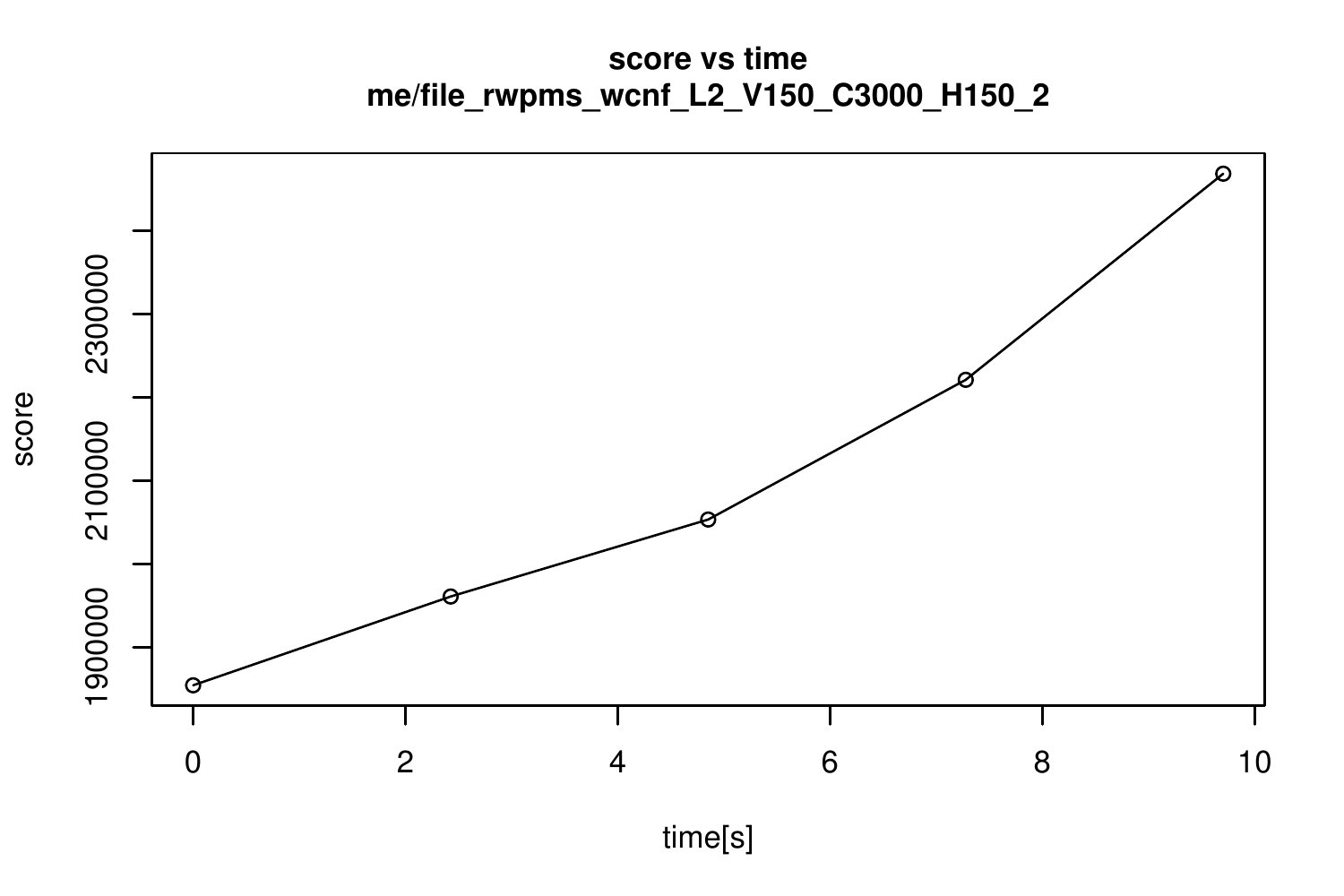}
    \label{fig_me/file_rwpms_wcnf_L2_V150_C3000_H150_2/file_rwpms_wcnf_L2_V150_C3000_H150_2-score_vs_time}
\end{figure}

\begin{figure}[H]
    \centering
    \includegraphics[height=3.5in]{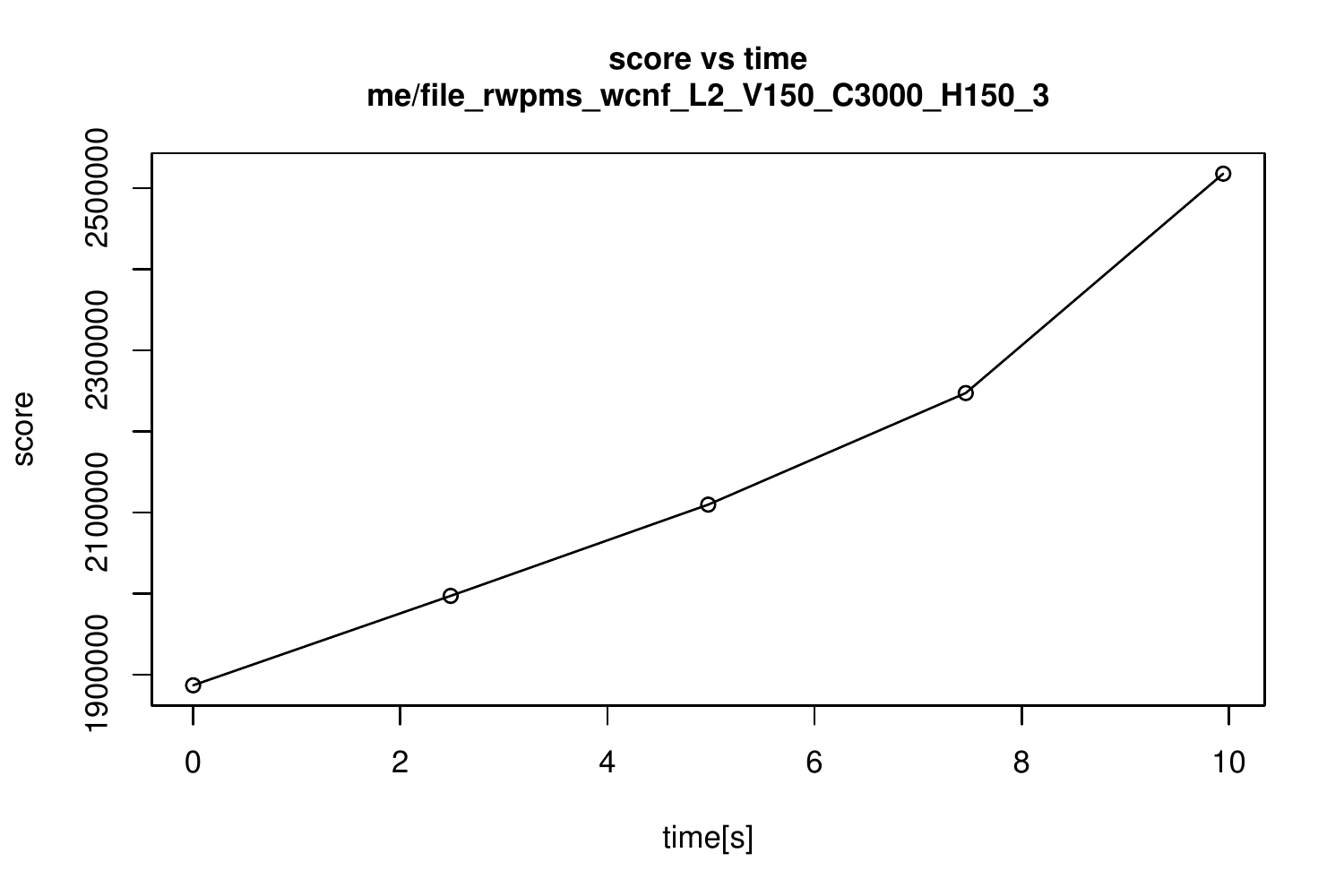}
    \label{fig_me/file_rwpms_wcnf_L2_V150_C3000_H150_3/file_rwpms_wcnf_L2_V150_C3000_H150_3-score_vs_time}
\end{figure}

\begin{figure}[H]
    \centering
    \includegraphics[height=3.5in]{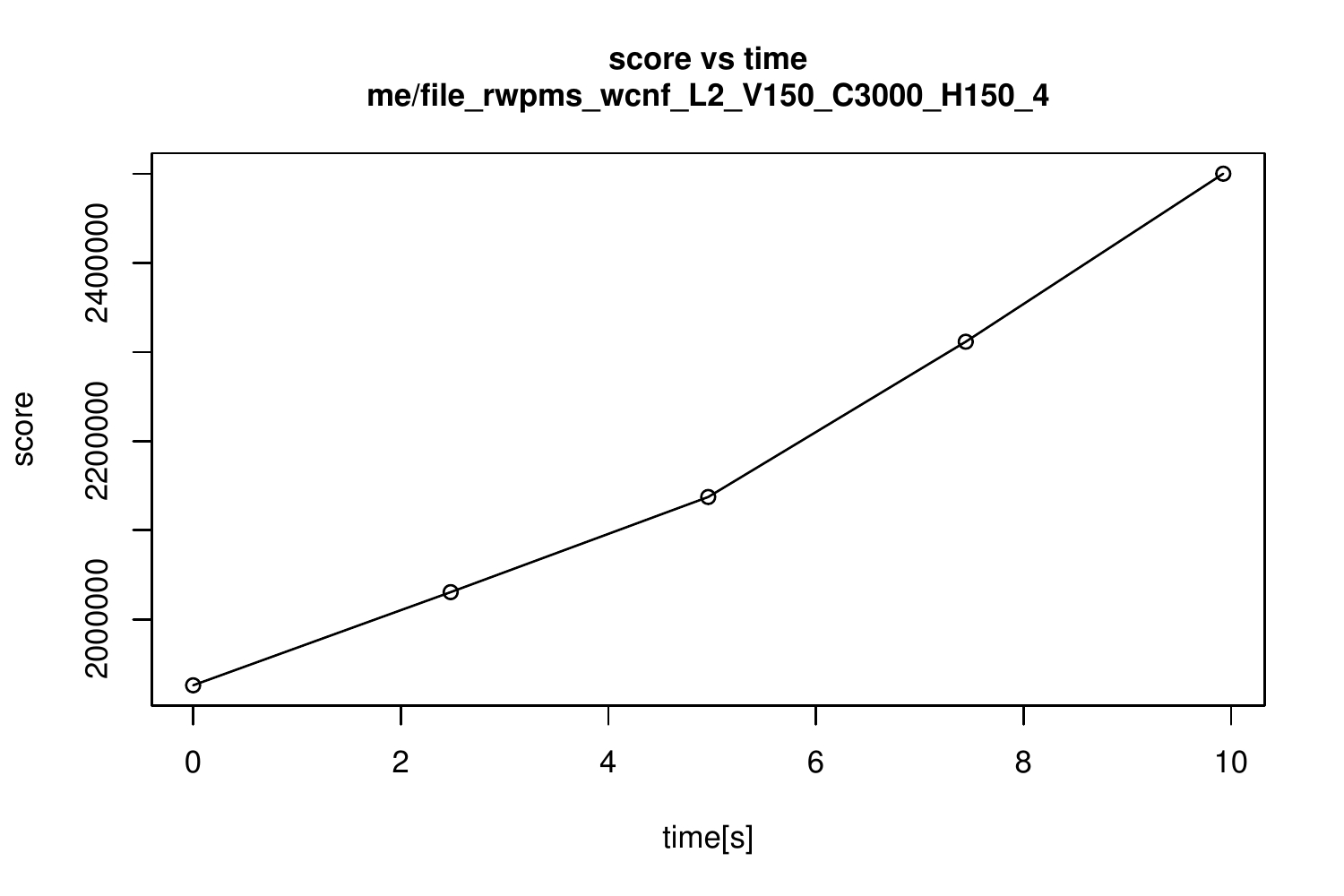}
    \label{fig_me/file_rwpms_wcnf_L2_V150_C3000_H150_4/file_rwpms_wcnf_L2_V150_C3000_H150_4-score_vs_time}
\end{figure}

\begin{figure}[H]
    \centering
    \includegraphics[height=3.5in]{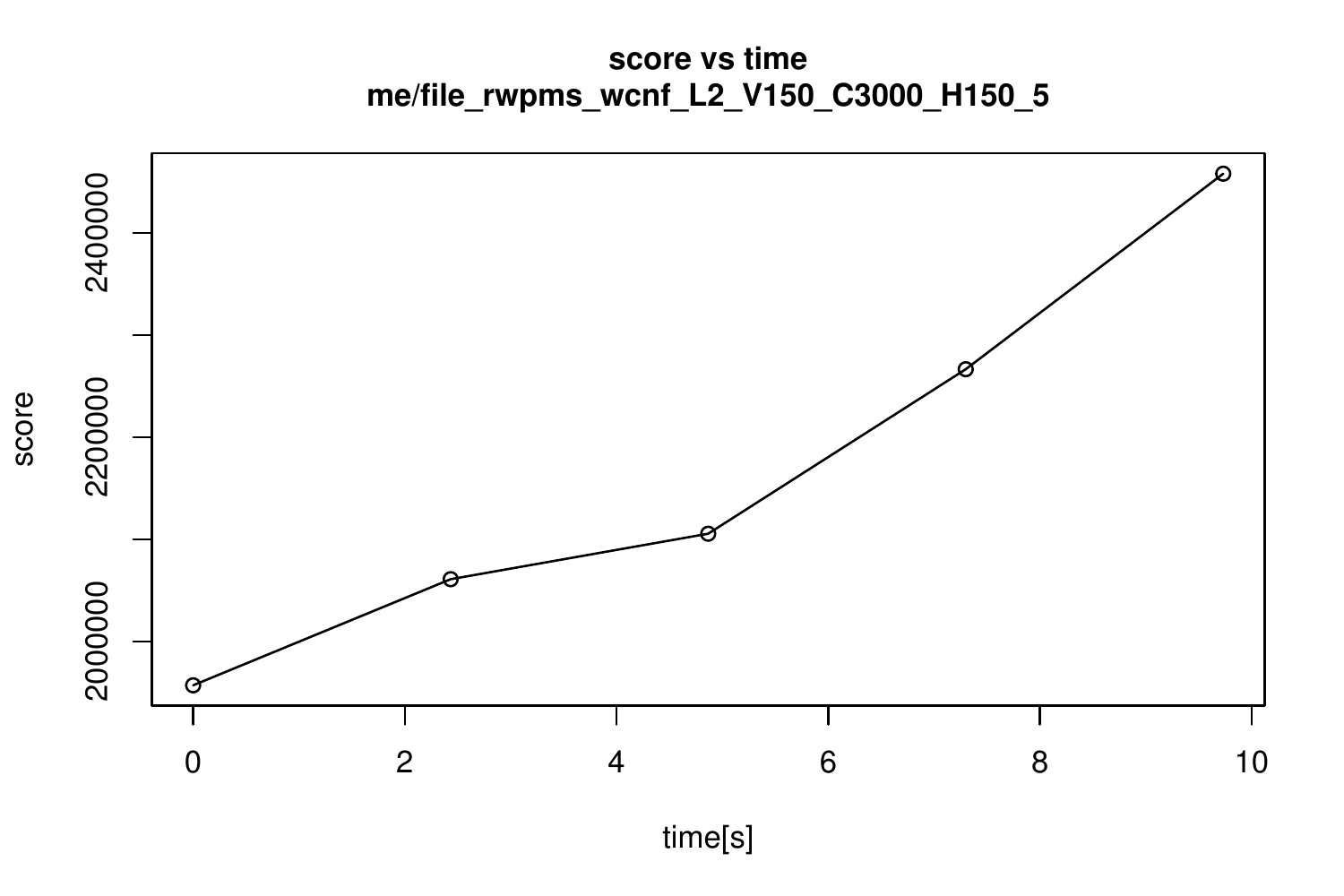}
    \label{fig_me/file_rwpms_wcnf_L2_V150_C3000_H150_5/file_rwpms_wcnf_L2_V150_C3000_H150_5-score_vs_time}
\end{figure}

\begin{figure}[H]
    \centering
    \includegraphics[height=3.5in]{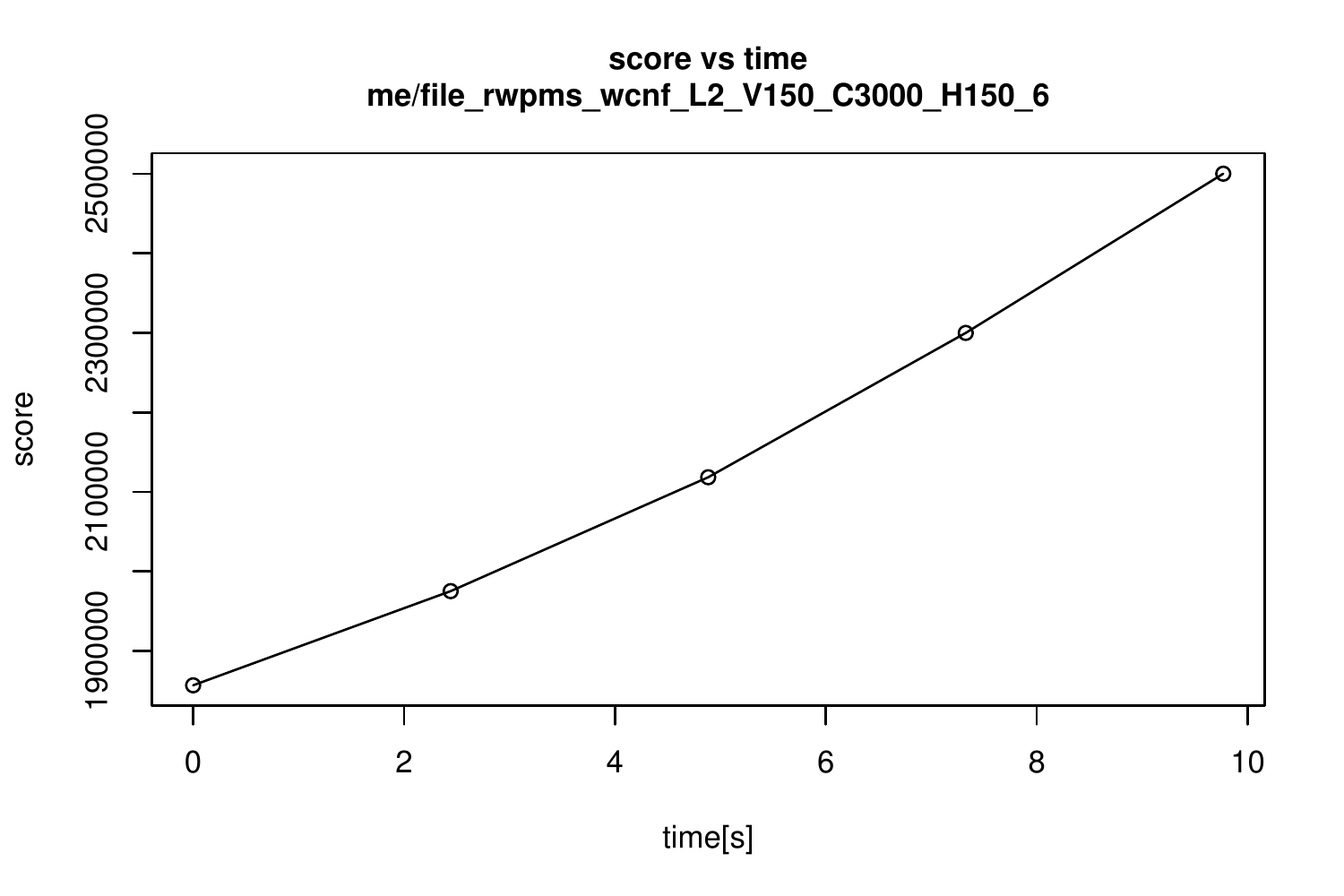}
    \label{fig_me/file_rwpms_wcnf_L2_V150_C3000_H150_6/file_rwpms_wcnf_L2_V150_C3000_H150_6-score_vs_time}
\end{figure}

\begin{figure}[H]
    \centering
    \includegraphics[height=3.5in]{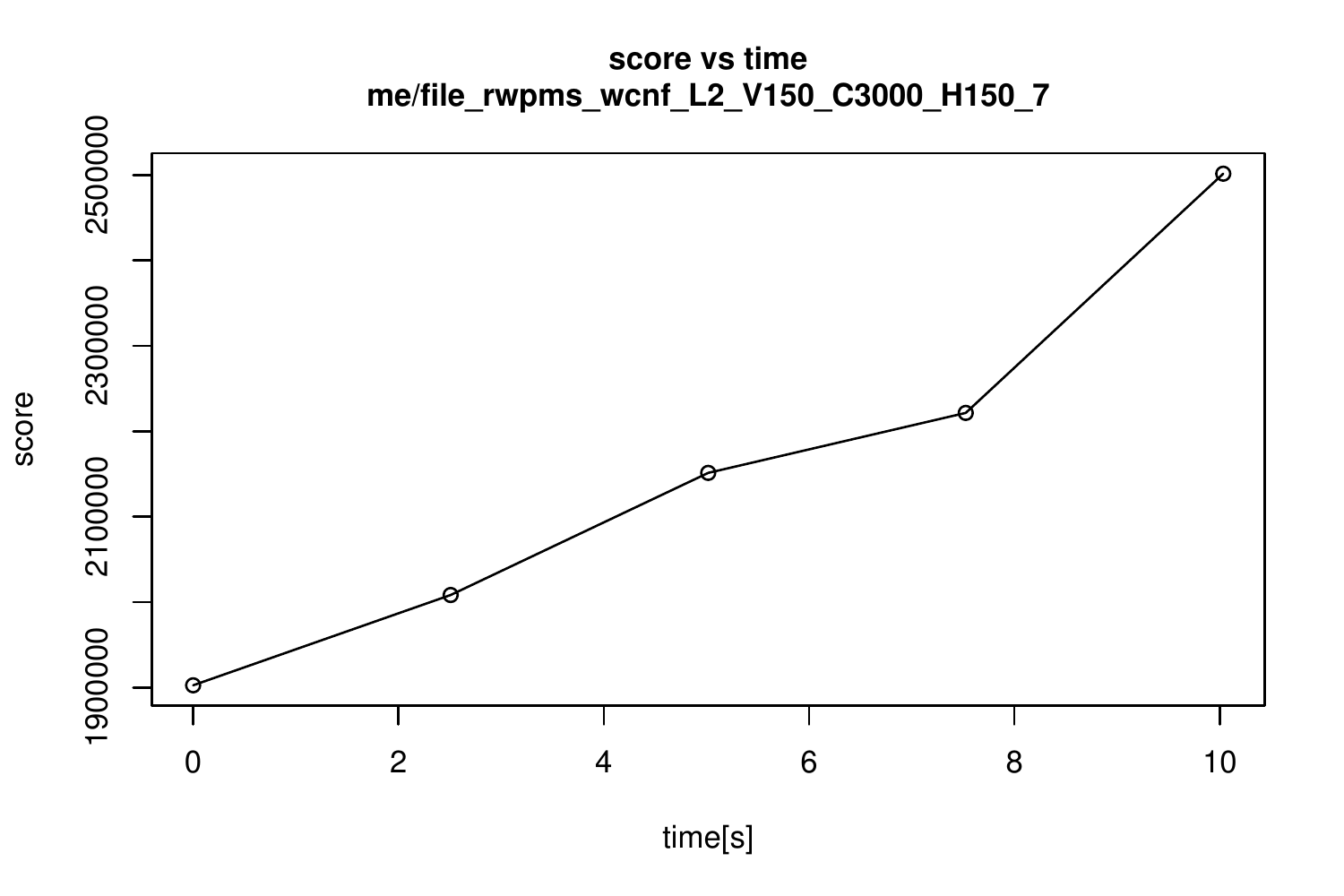}
    \label{fig_me/file_rwpms_wcnf_L2_V150_C3000_H150_7/file_rwpms_wcnf_L2_V150_C3000_H150_7-score_vs_time}
\end{figure}

\begin{figure}[H]
    \centering
    \includegraphics[height=3.5in]{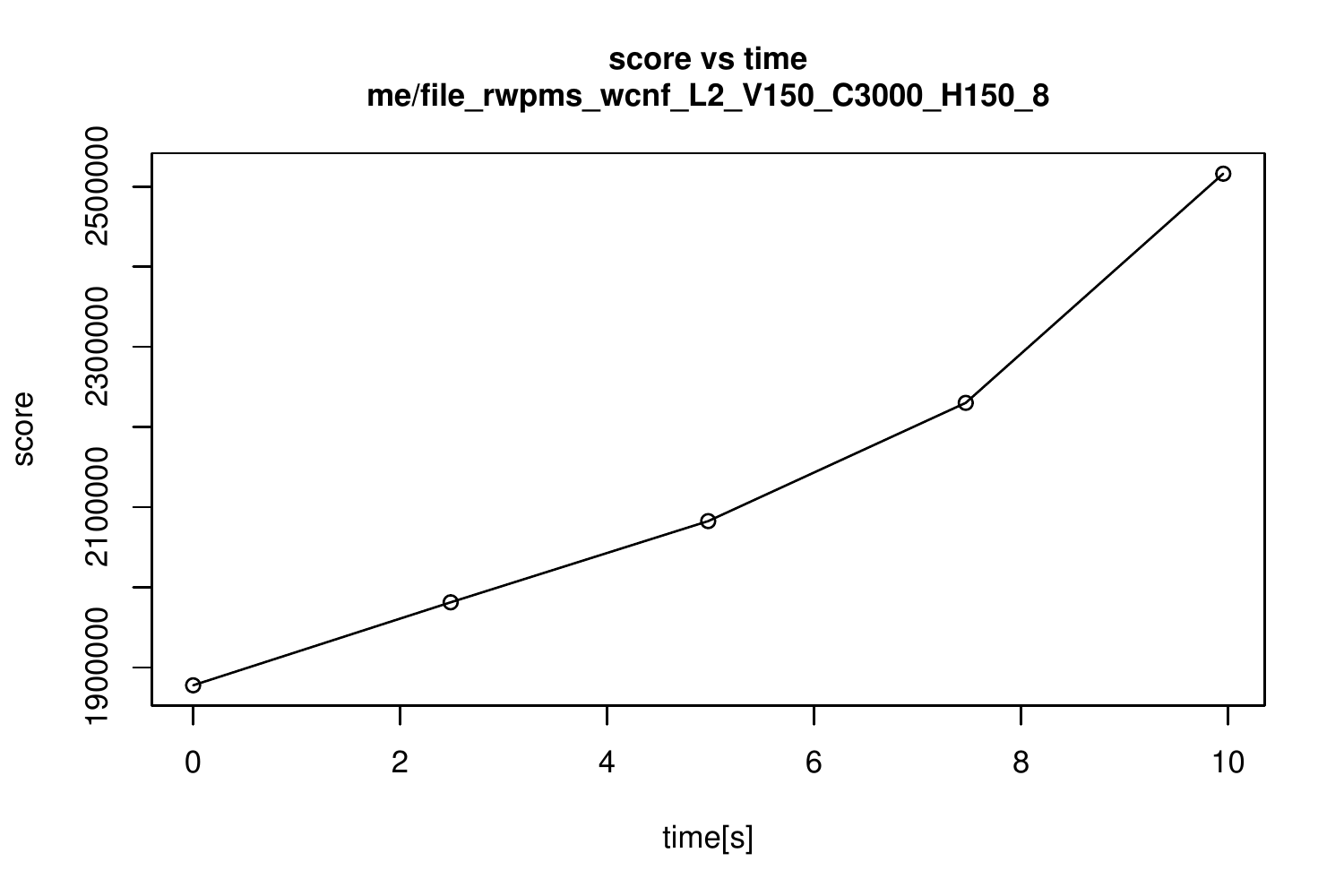}
    \label{fig_me/file_rwpms_wcnf_L2_V150_C3000_H150_8/file_rwpms_wcnf_L2_V150_C3000_H150_8-score_vs_time}
\end{figure}

\begin{figure}[H]
    \centering
    \includegraphics[height=3.5in]{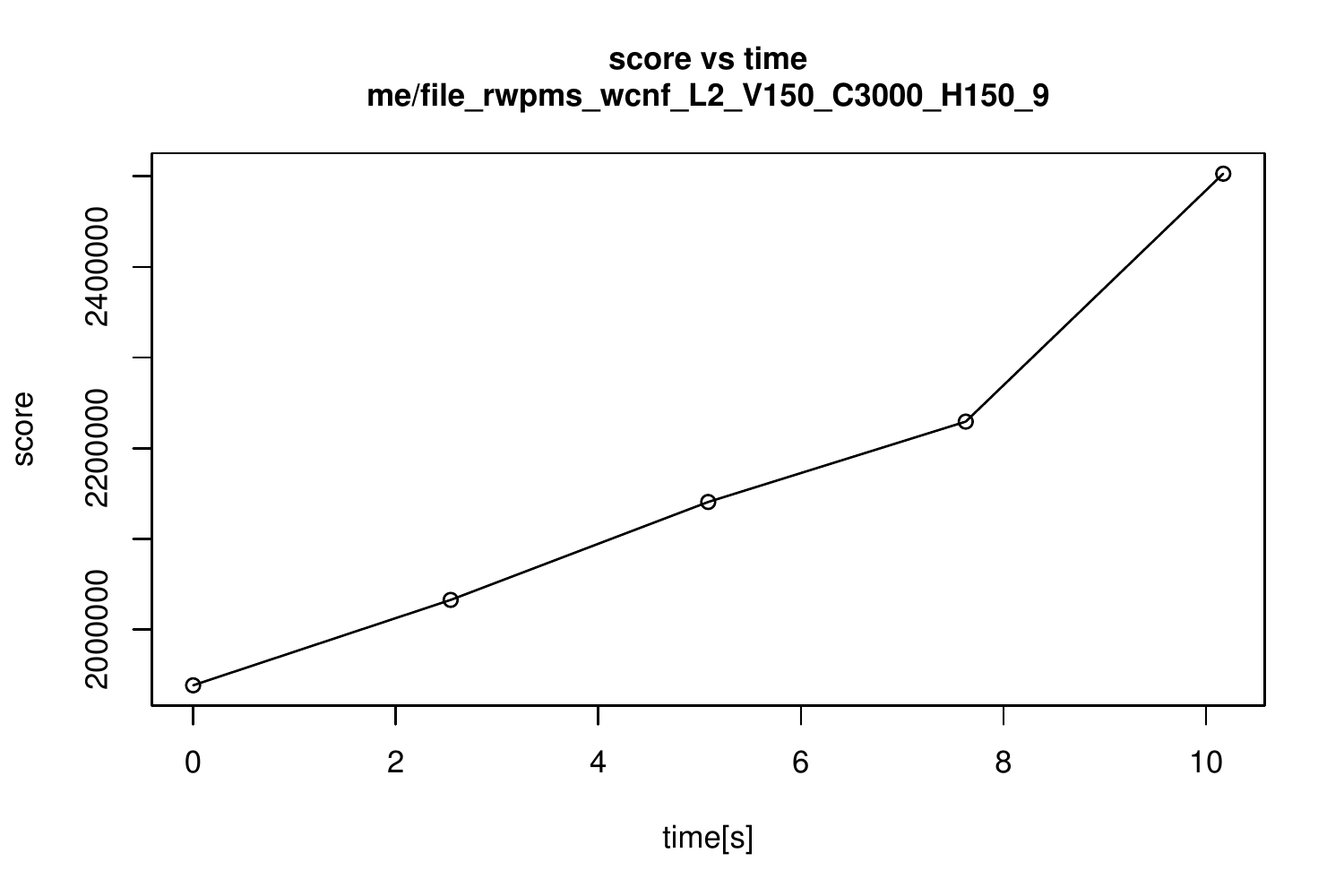}
    \label{fig_me/file_rwpms_wcnf_L2_V150_C3000_H150_9/file_rwpms_wcnf_L2_V150_C3000_H150_9-score_vs_time}
\end{figure}

\begin{figure}[H]
    \centering
    \includegraphics[height=3.5in]{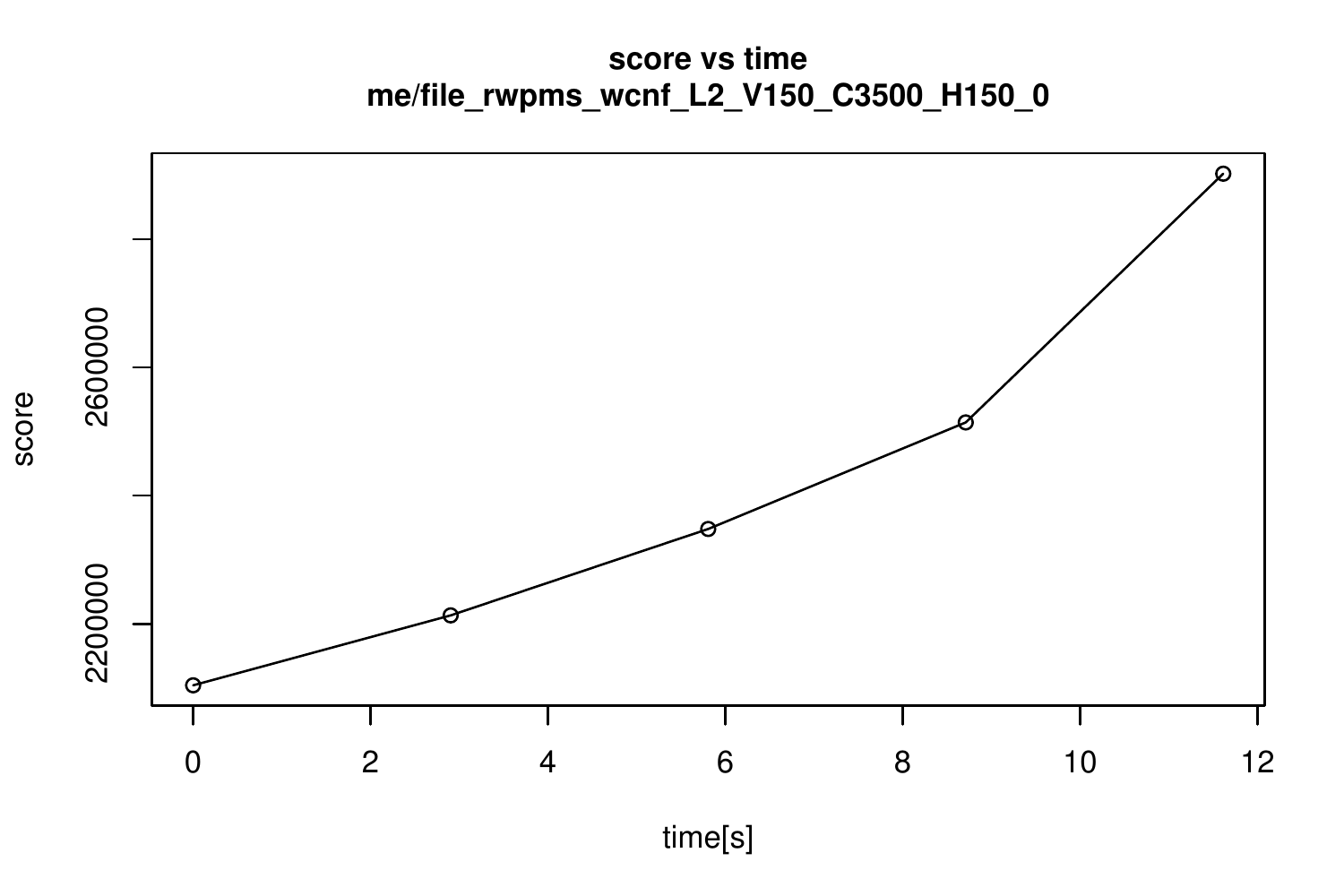}
    \label{fig_me/file_rwpms_wcnf_L2_V150_C3500_H150_0/file_rwpms_wcnf_L2_V150_C3500_H150_0-score_vs_time}
\end{figure}

\begin{figure}[H]
    \centering
    \includegraphics[height=3.5in]{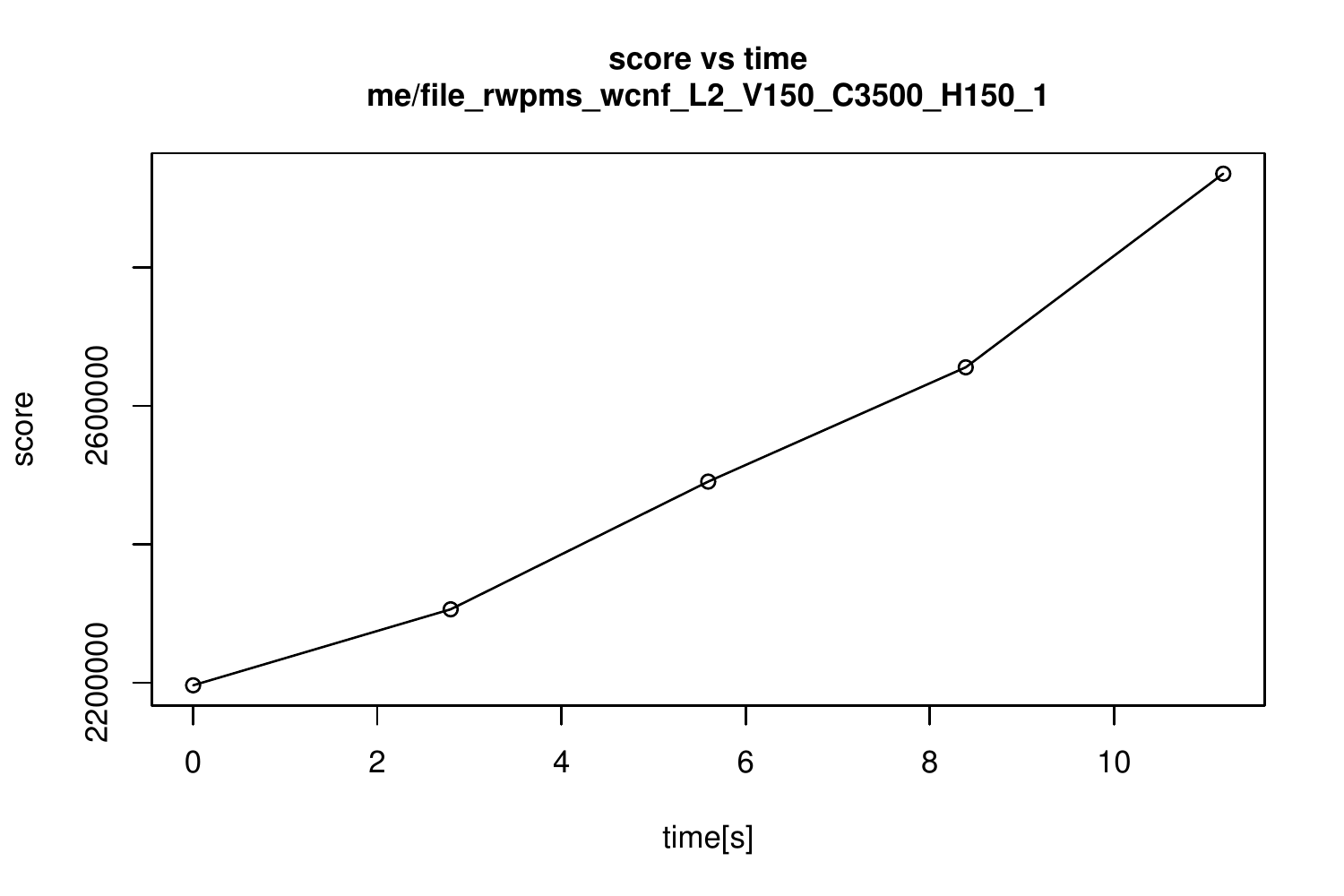}
    \label{fig_me/file_rwpms_wcnf_L2_V150_C3500_H150_1/file_rwpms_wcnf_L2_V150_C3500_H150_1-score_vs_time}
\end{figure}

\begin{figure}[H]
    \centering
    \includegraphics[height=3.5in]{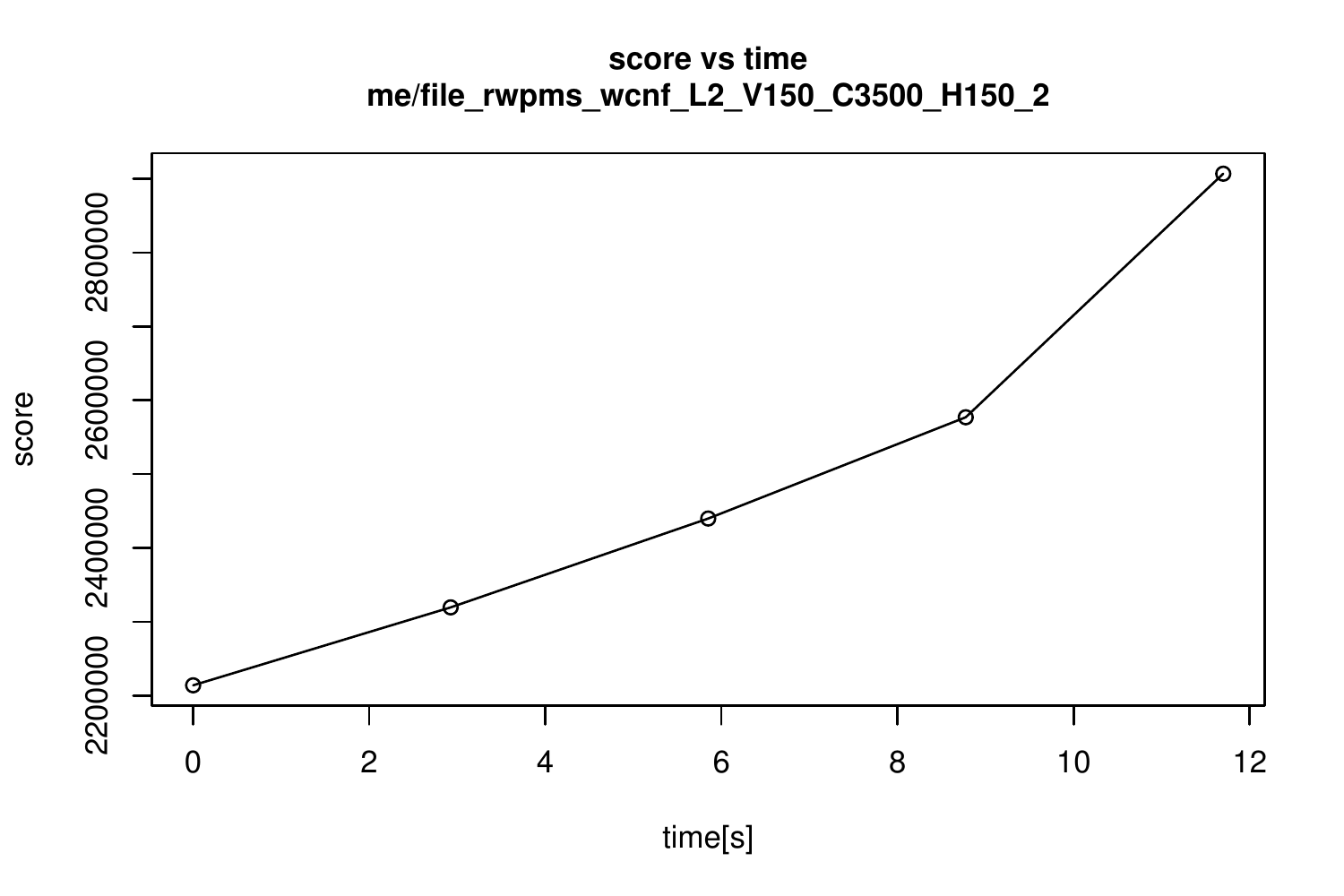}
    \label{fig_me/file_rwpms_wcnf_L2_V150_C3500_H150_2/file_rwpms_wcnf_L2_V150_C3500_H150_2-score_vs_time}
\end{figure}

\begin{figure}[H]
    \centering
    \includegraphics[height=3.5in]{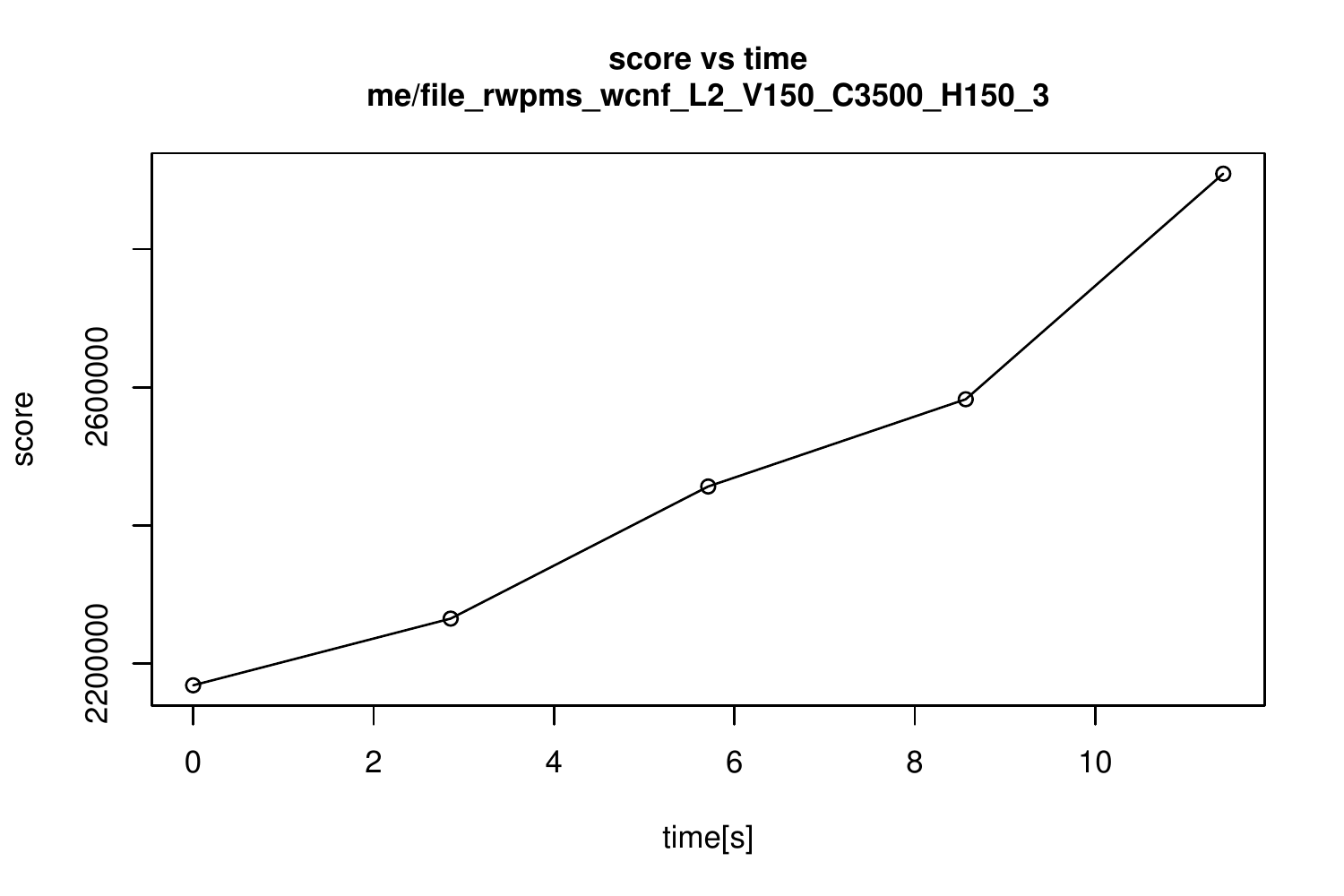}
    \label{fig_me/file_rwpms_wcnf_L2_V150_C3500_H150_3/file_rwpms_wcnf_L2_V150_C3500_H150_3-score_vs_time}
\end{figure}

\begin{figure}[H]
    \centering
    \includegraphics[height=3.5in]{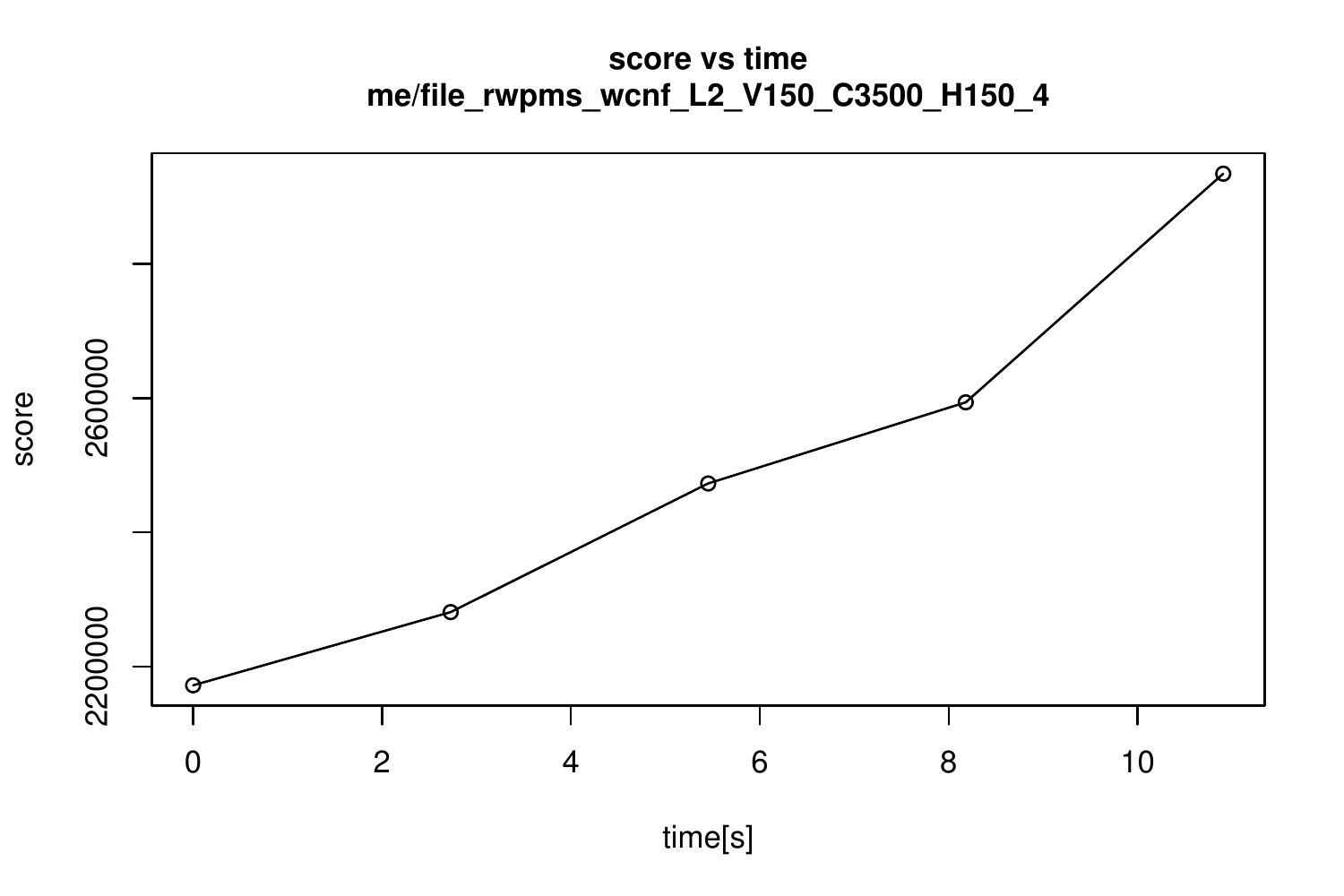}
    \label{fig_me/file_rwpms_wcnf_L2_V150_C3500_H150_4/file_rwpms_wcnf_L2_V150_C3500_H150_4-score_vs_time}
\end{figure}

\begin{figure}[H]
    \centering
    \includegraphics[height=3.5in]{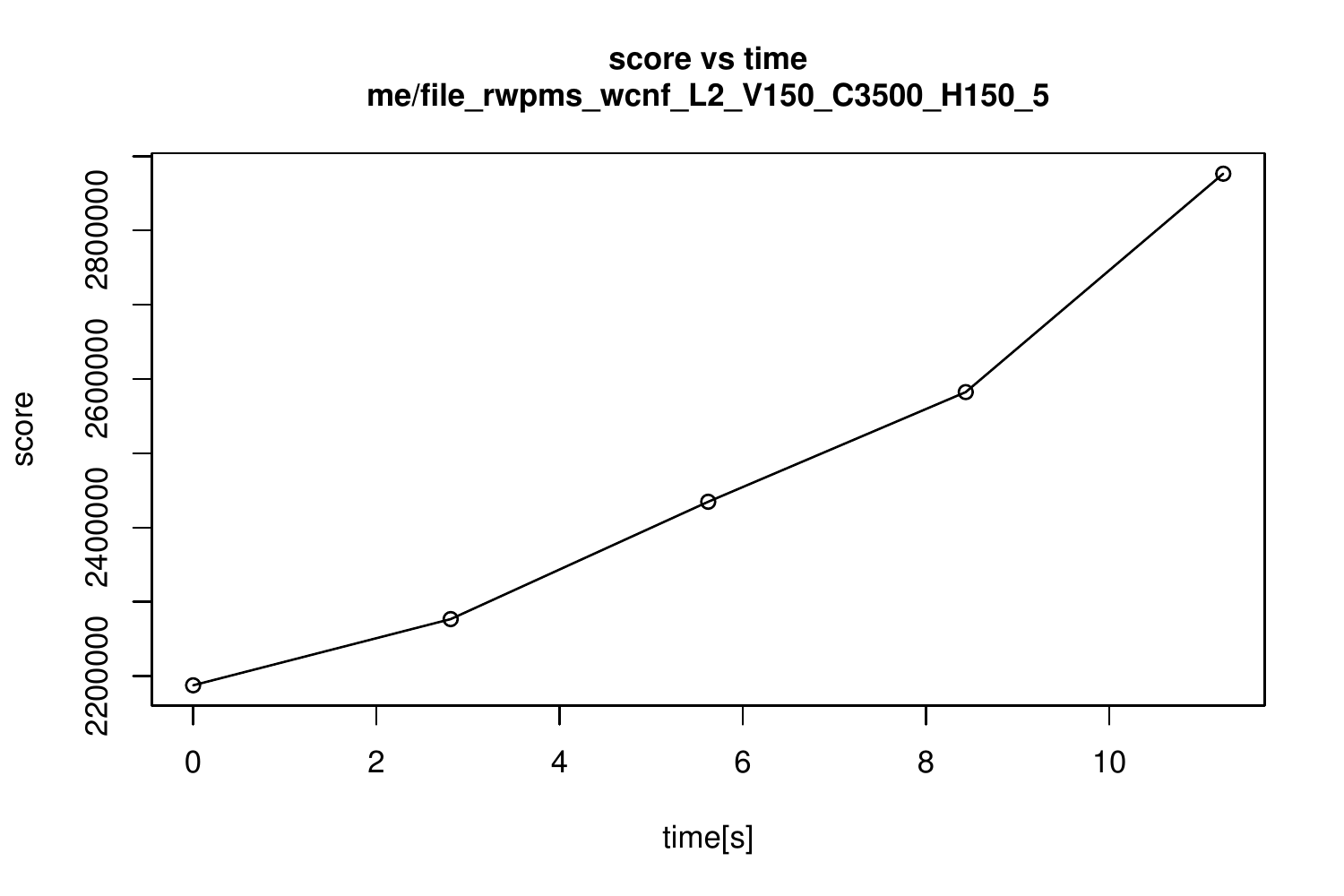}
    \label{fig_me/file_rwpms_wcnf_L2_V150_C3500_H150_5/file_rwpms_wcnf_L2_V150_C3500_H150_5-score_vs_time}
\end{figure}

\begin{figure}[H]
    \centering
    \includegraphics[height=3.5in]{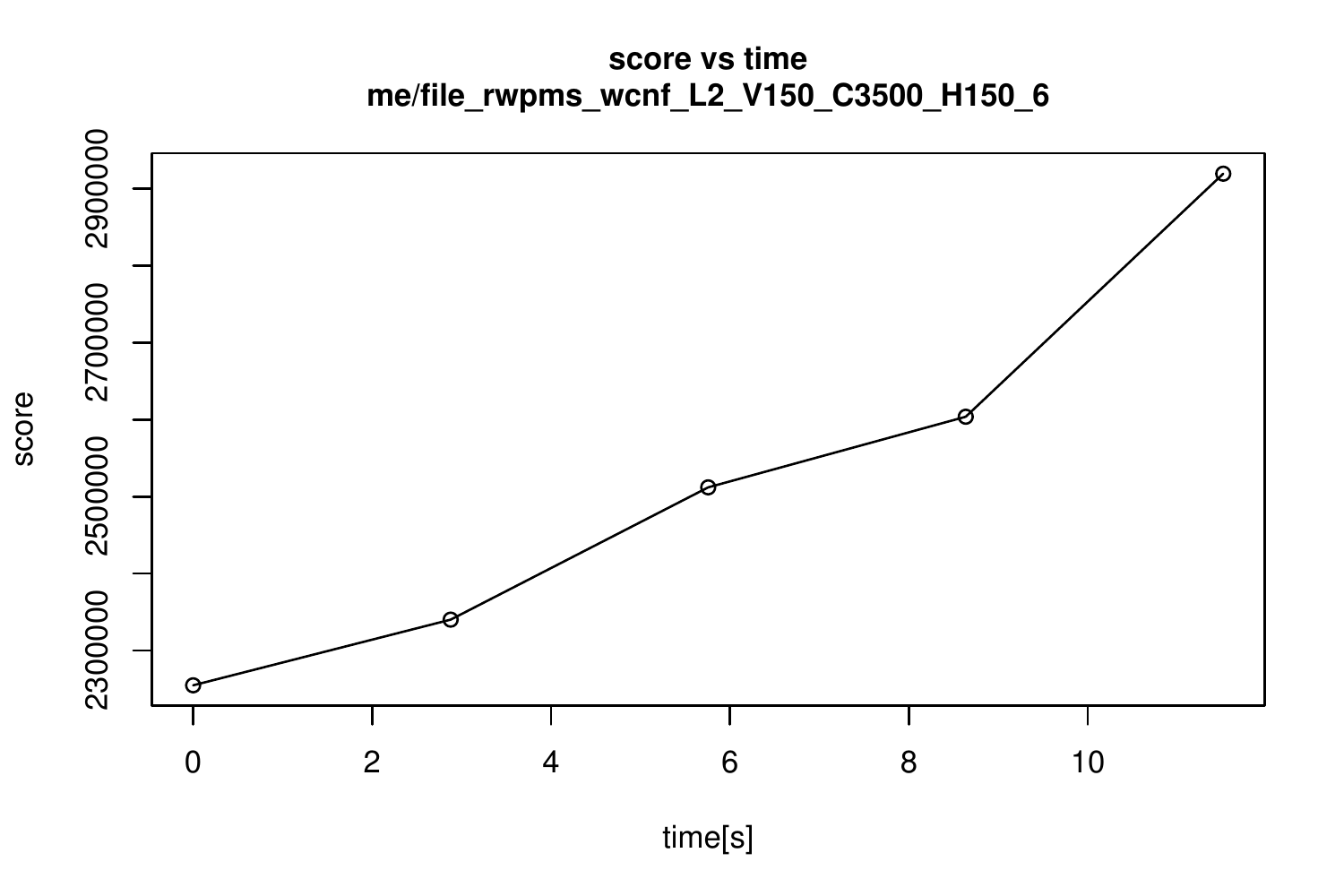}
    \label{fig_me/file_rwpms_wcnf_L2_V150_C3500_H150_6/file_rwpms_wcnf_L2_V150_C3500_H150_6-score_vs_time}
\end{figure}

\begin{figure}[H]
    \centering
    \includegraphics[height=3.5in]{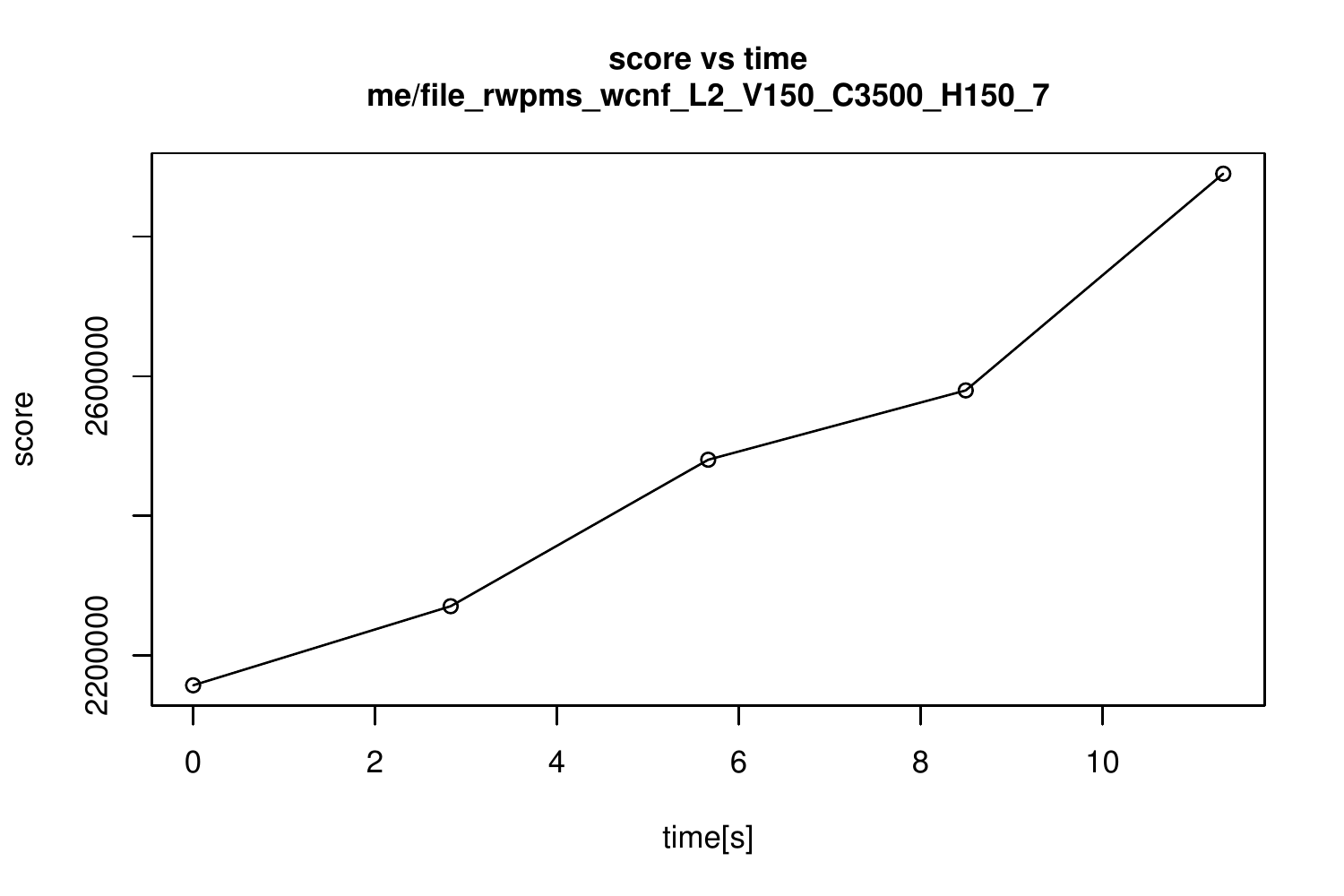}
    \label{fig_me/file_rwpms_wcnf_L2_V150_C3500_H150_7/file_rwpms_wcnf_L2_V150_C3500_H150_7-score_vs_time}
\end{figure}

\begin{figure}[H]
    \centering
    \includegraphics[height=3.5in]{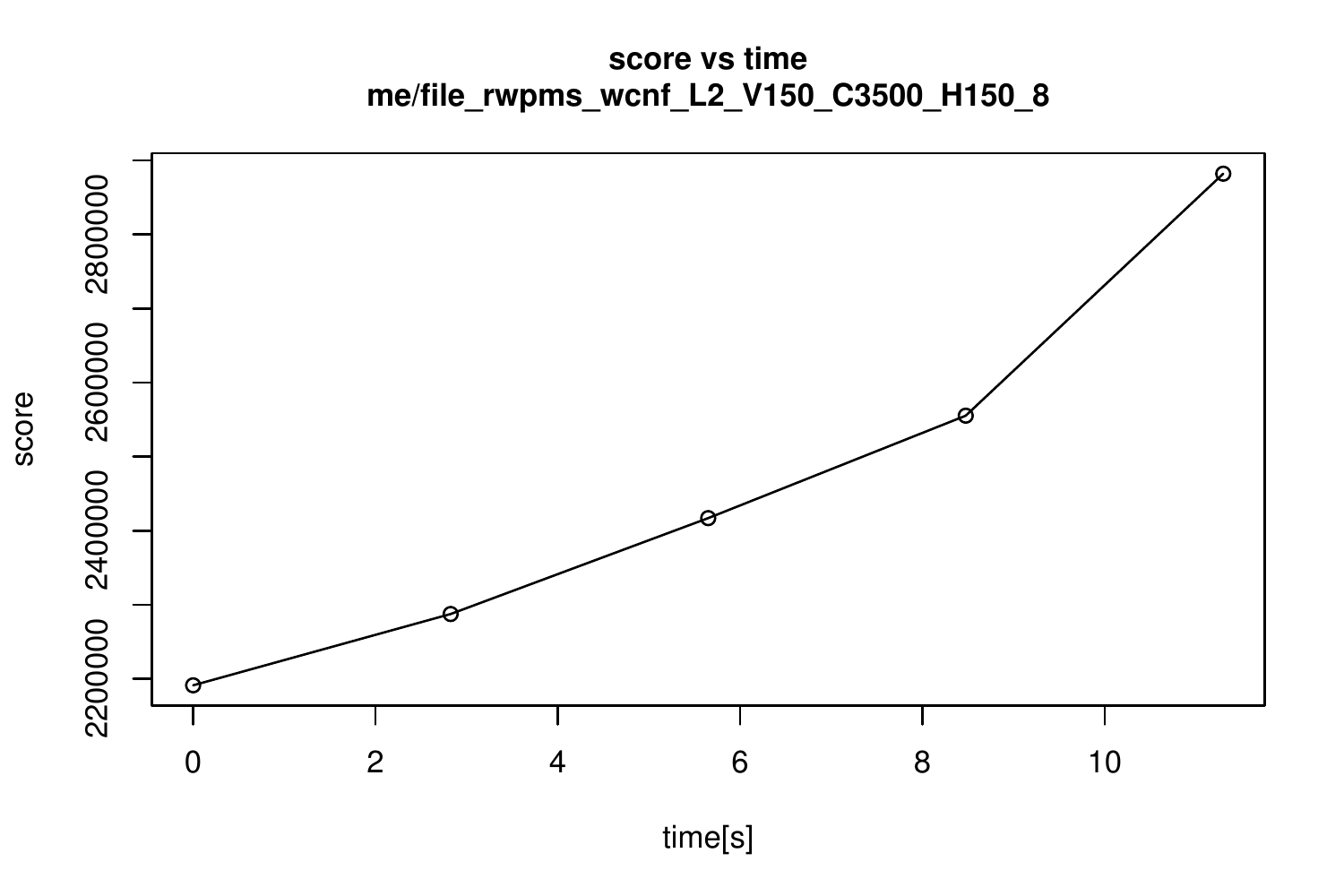}
    \label{fig_me/file_rwpms_wcnf_L2_V150_C3500_H150_8/file_rwpms_wcnf_L2_V150_C3500_H150_8-score_vs_time}
\end{figure}

\begin{figure}[H]
    \centering
    \includegraphics[height=3.5in]{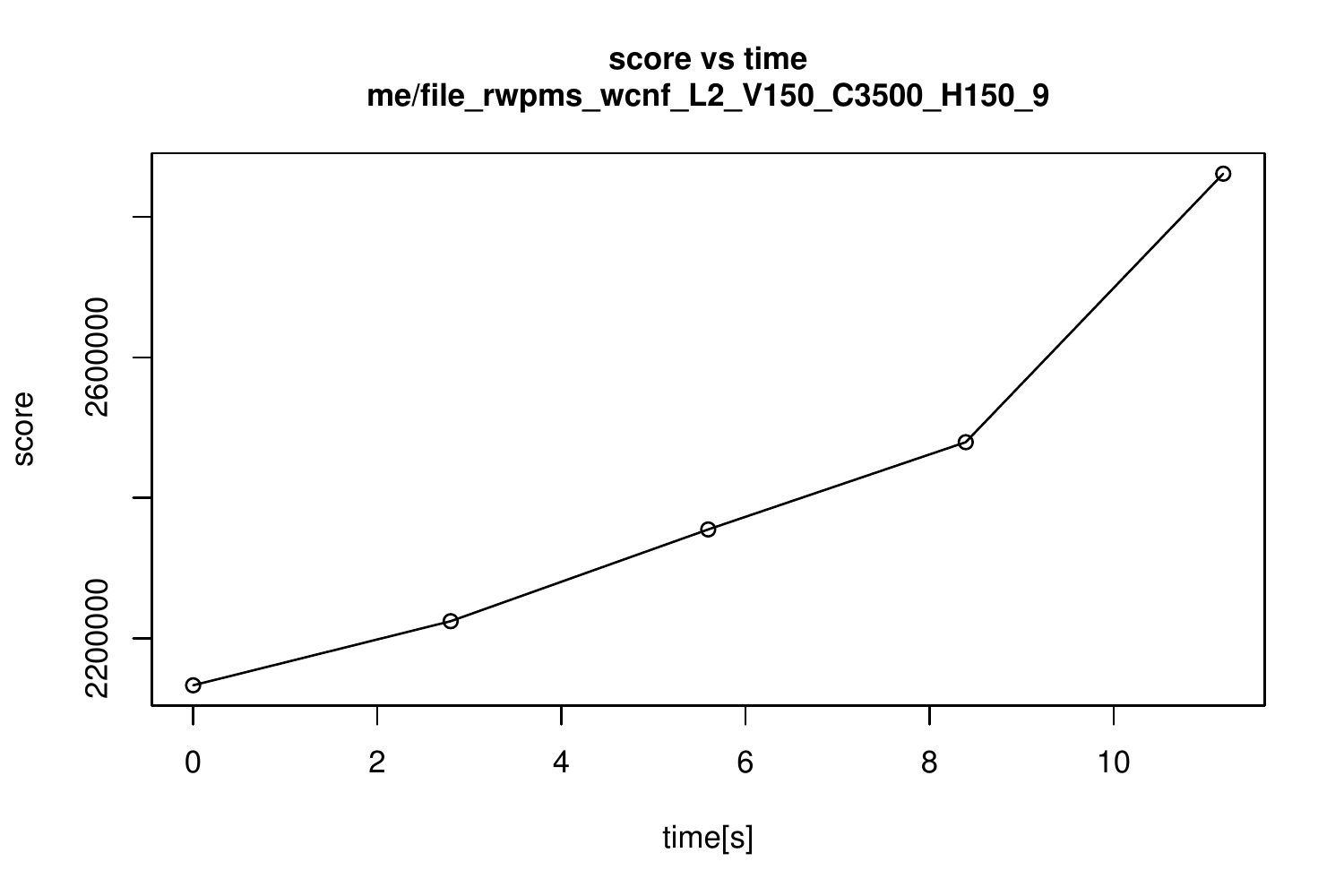}
    \label{fig_me/file_rwpms_wcnf_L2_V150_C3500_H150_9/file_rwpms_wcnf_L2_V150_C3500_H150_9-score_vs_time}
\end{figure}

\begin{figure}[H]
    \centering
    \includegraphics[height=3.5in]{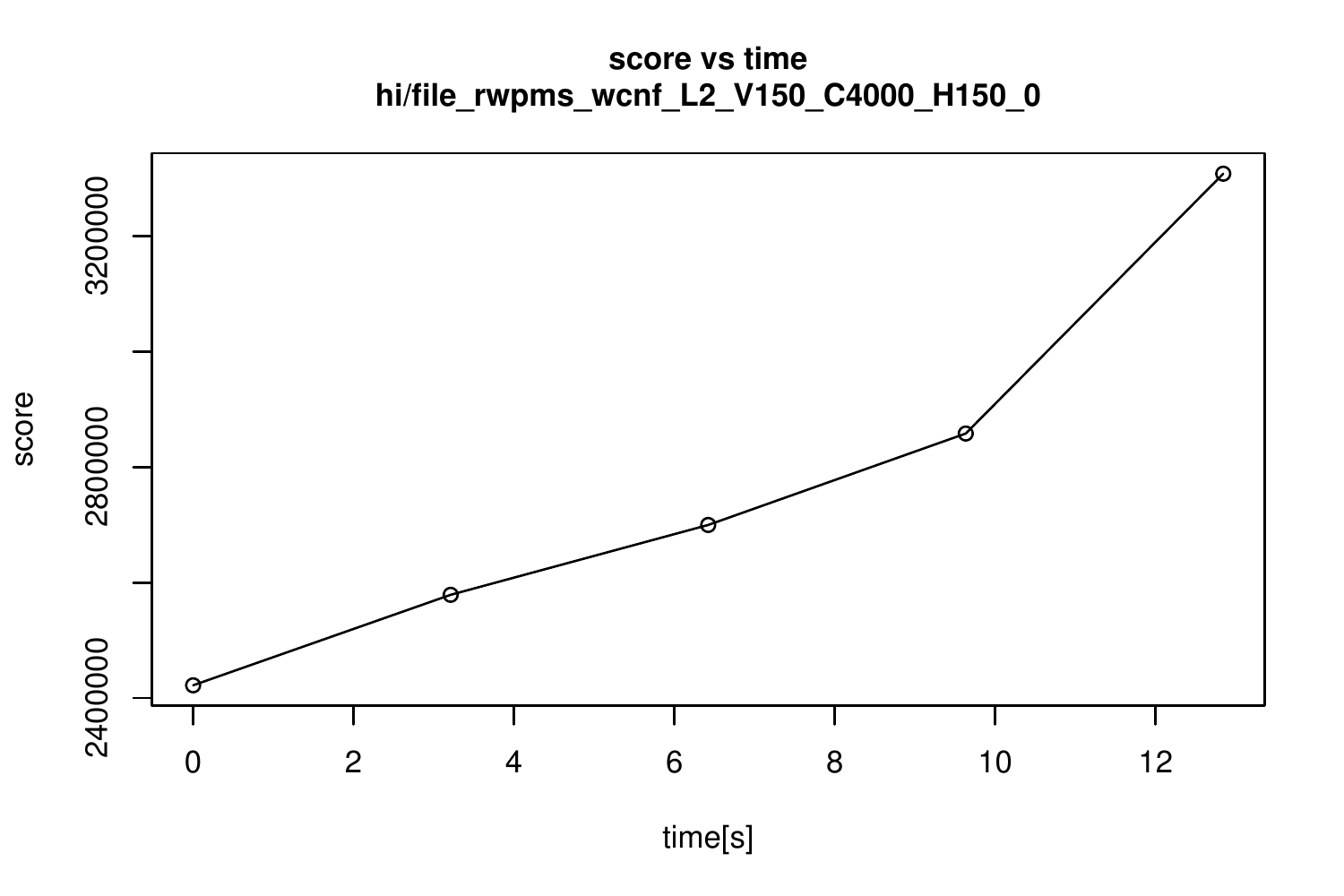}
    \label{fig_hi/file_rwpms_wcnf_L2_V150_C4000_H150_0/file_rwpms_wcnf_L2_V150_C4000_H150_0-score_vs_time}
\end{figure}

\begin{figure}[H]
    \centering
    \includegraphics[height=3.5in]{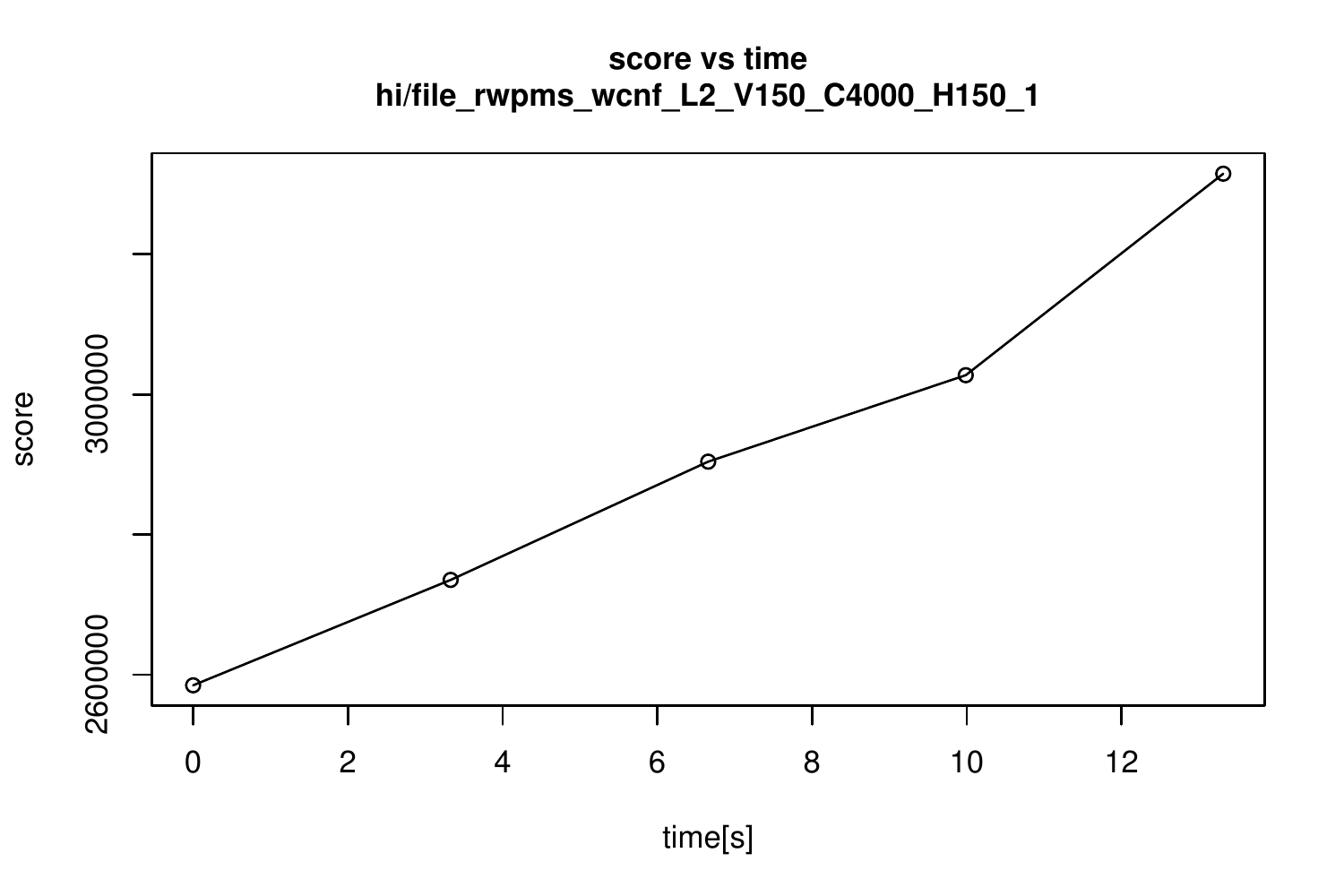}
    \label{fig_hi/file_rwpms_wcnf_L2_V150_C4000_H150_1/file_rwpms_wcnf_L2_V150_C4000_H150_1-score_vs_time}
\end{figure}

\begin{figure}[H]
    \centering
    \includegraphics[height=3.5in]{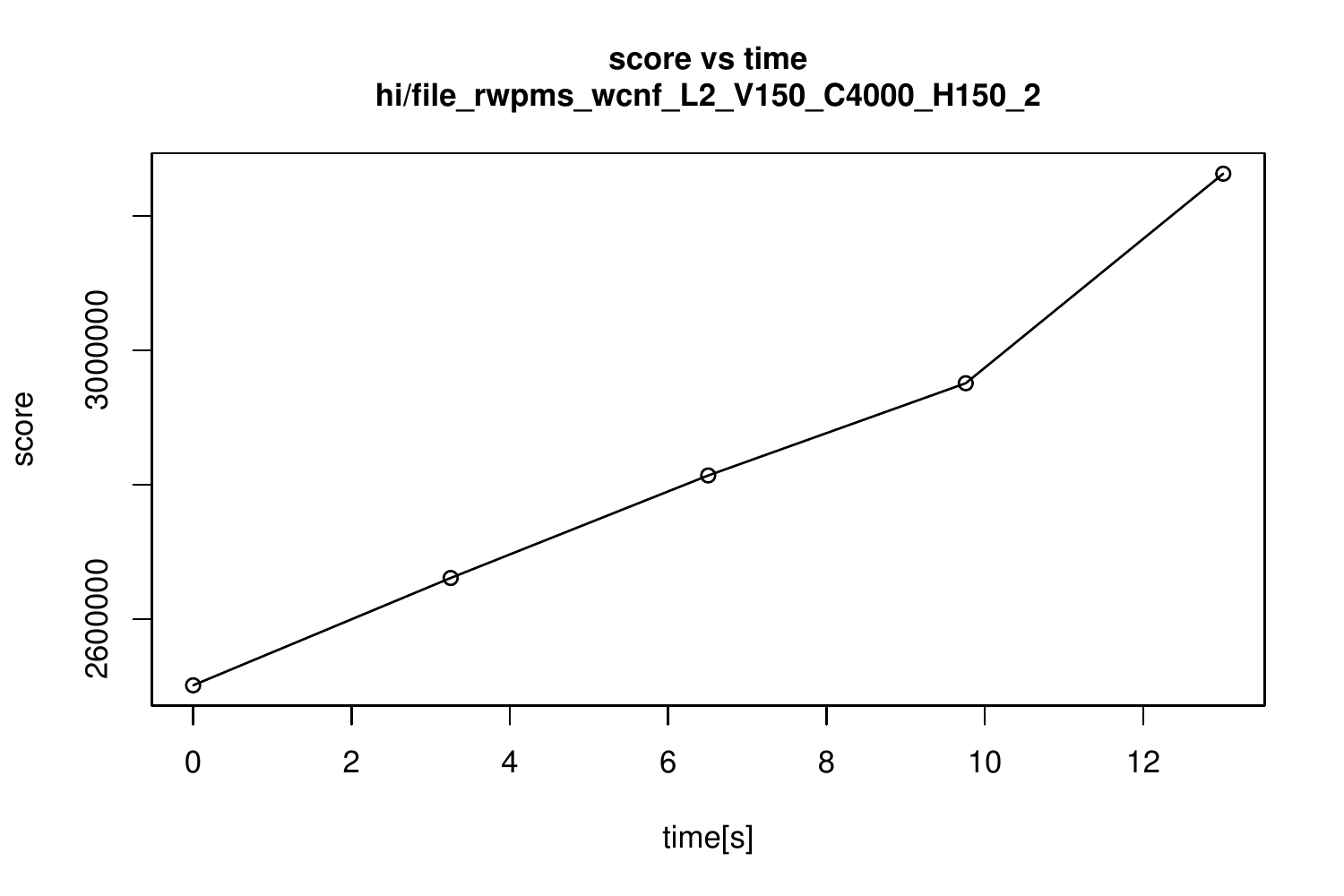}
    \label{fig_hi/file_rwpms_wcnf_L2_V150_C4000_H150_2/file_rwpms_wcnf_L2_V150_C4000_H150_2-score_vs_time}
\end{figure}

\begin{figure}[H]
    \centering
    \includegraphics[height=3.5in]{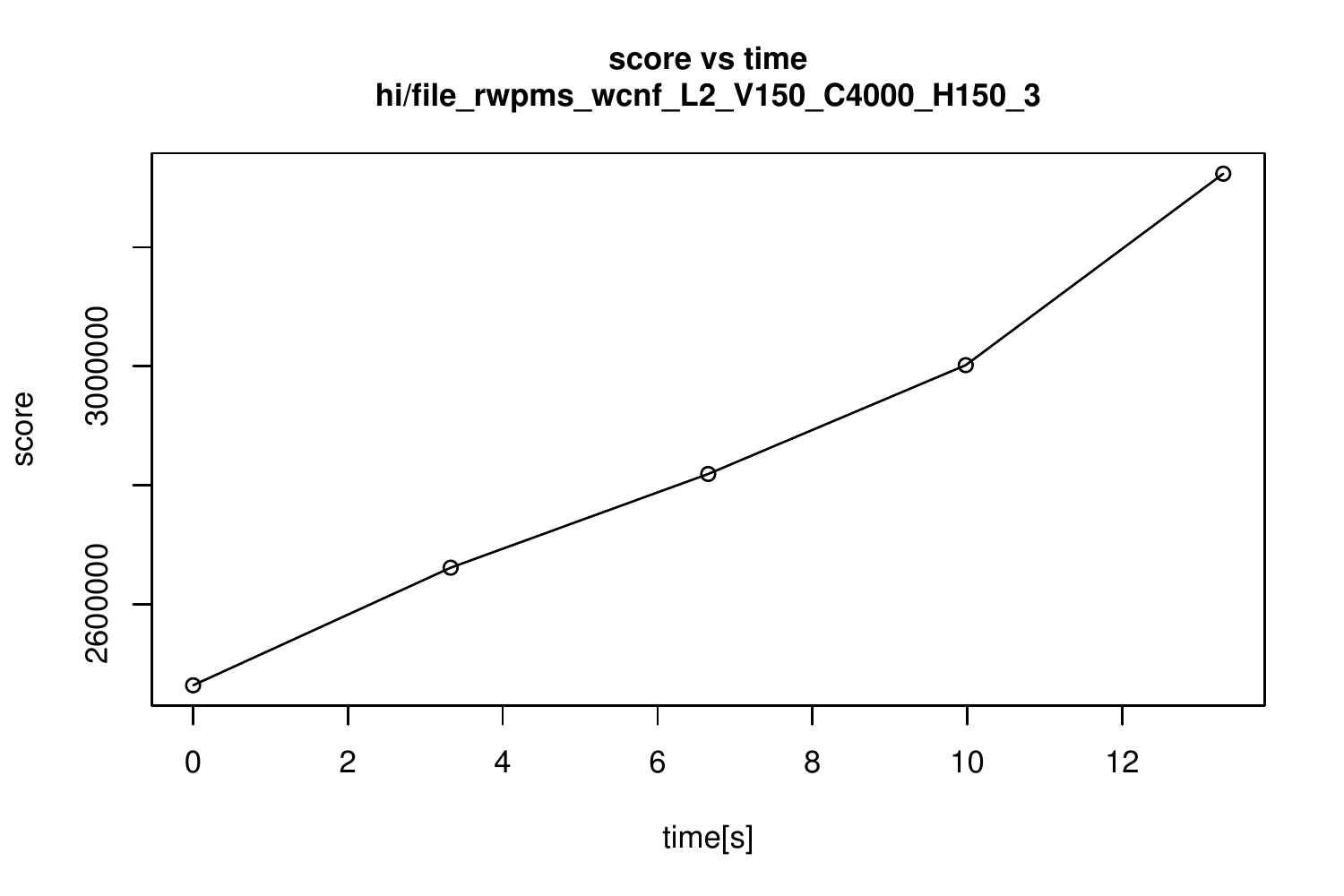}
    \label{fig_hi/file_rwpms_wcnf_L2_V150_C4000_H150_3/file_rwpms_wcnf_L2_V150_C4000_H150_3-score_vs_time}
\end{figure}

\begin{figure}[H]
    \centering
    \includegraphics[height=3.5in]{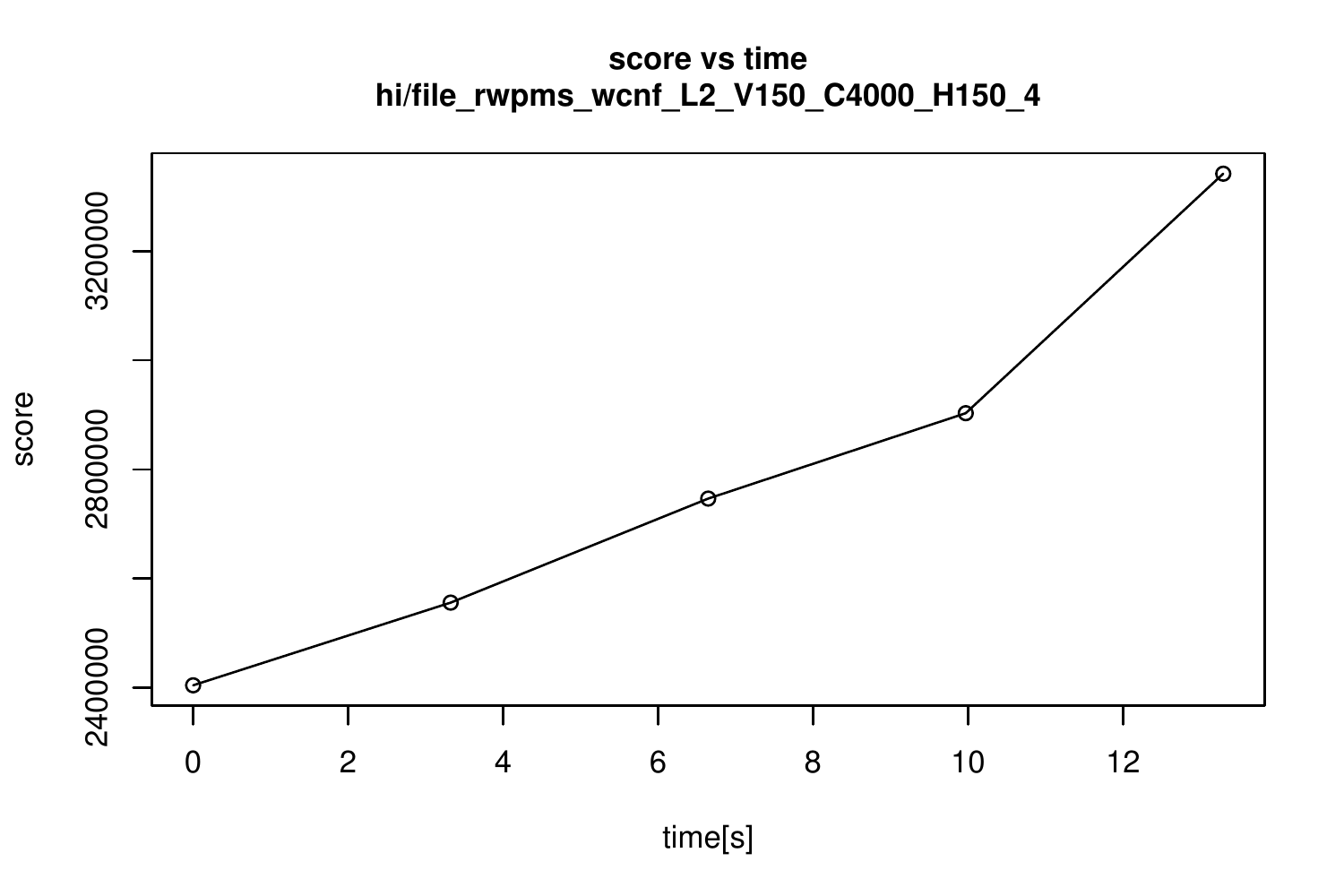}
    \label{fig_hi/file_rwpms_wcnf_L2_V150_C4000_H150_4/file_rwpms_wcnf_L2_V150_C4000_H150_4-score_vs_time}
\end{figure}

\begin{figure}[H]
    \centering
    \includegraphics[height=3.5in]{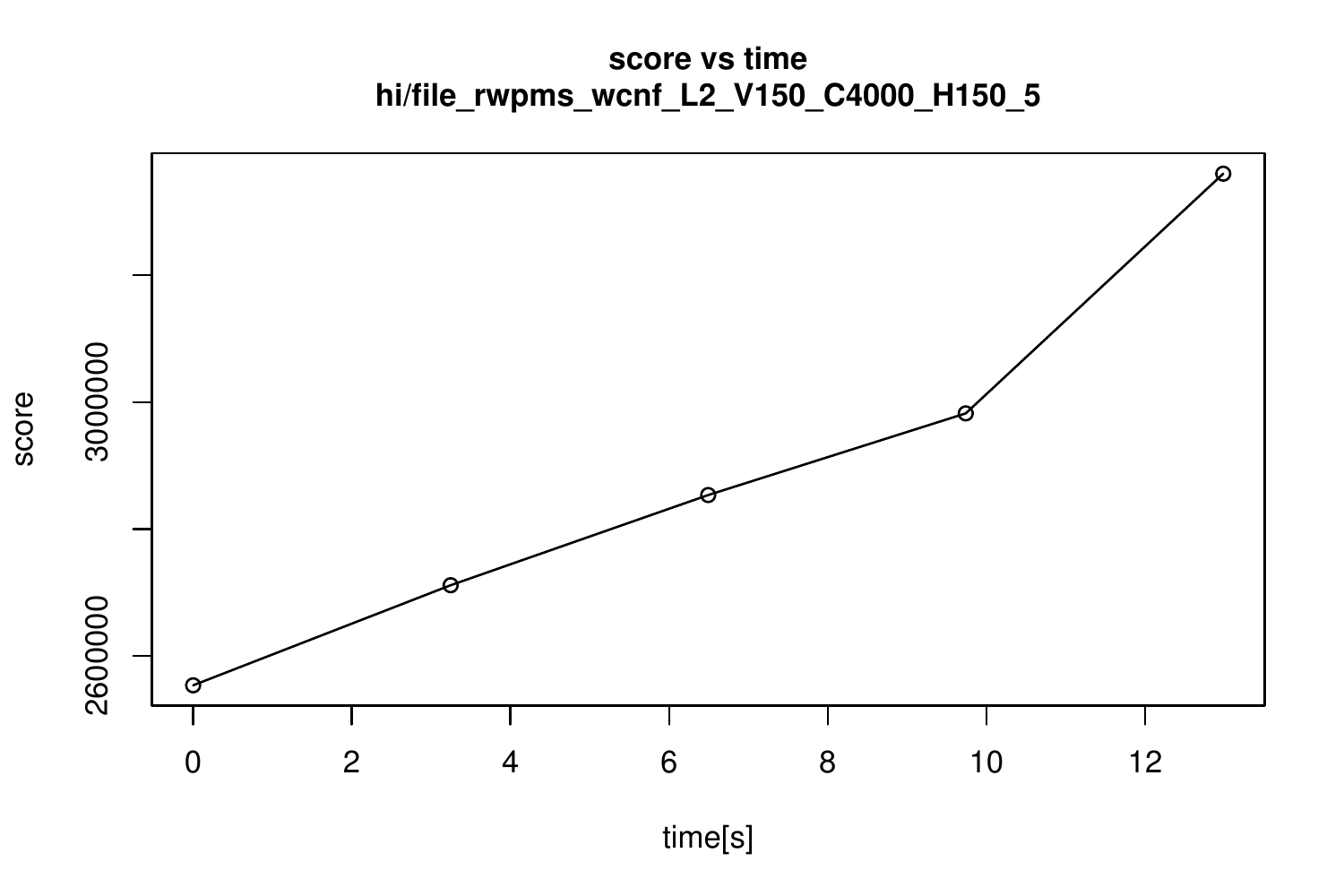}
    \label{fig_hi/file_rwpms_wcnf_L2_V150_C4000_H150_5/file_rwpms_wcnf_L2_V150_C4000_H150_5-score_vs_time}
\end{figure}

\begin{figure}[H]
    \centering
    \includegraphics[height=3.5in]{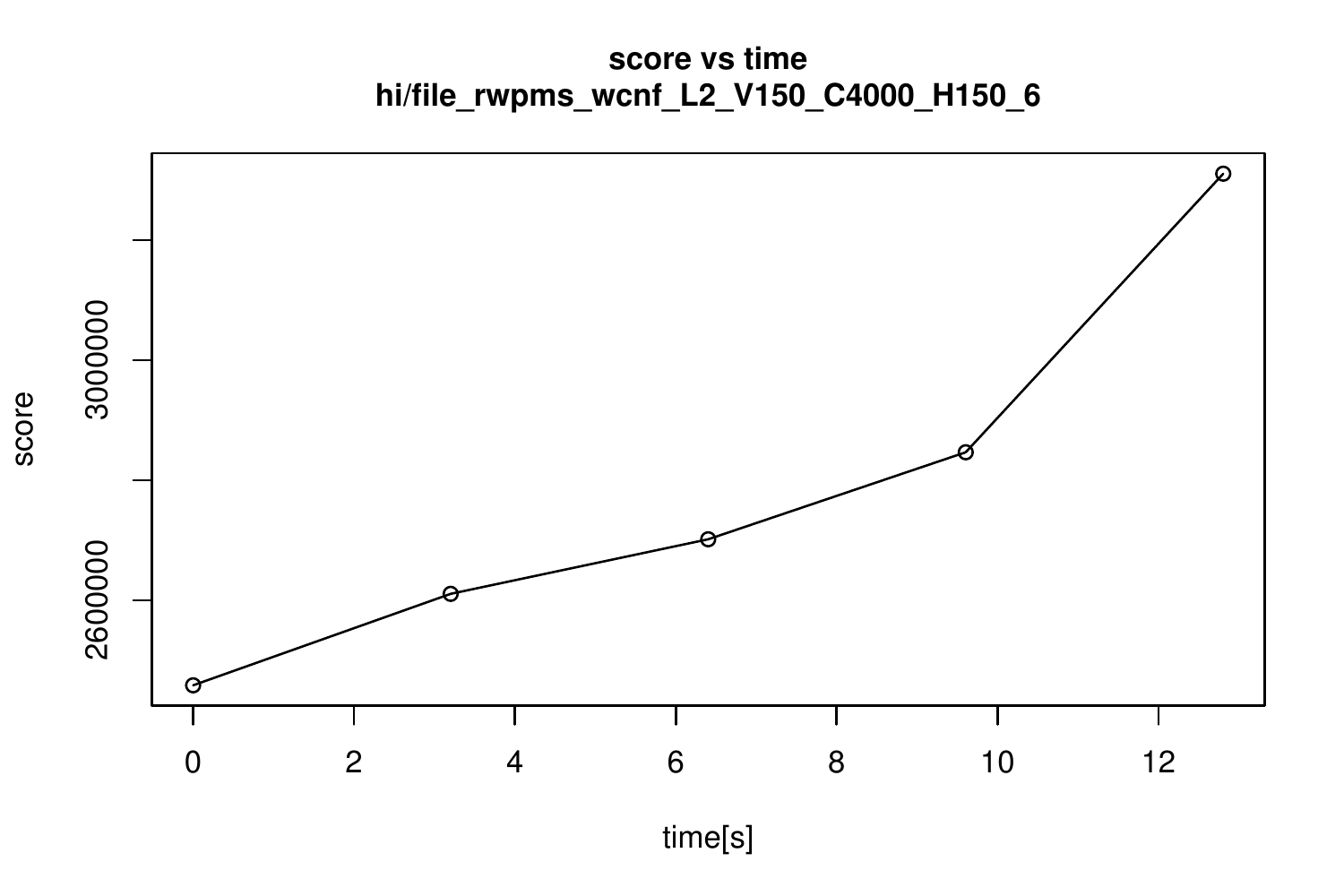}
    \label{fig_hi/file_rwpms_wcnf_L2_V150_C4000_H150_6/file_rwpms_wcnf_L2_V150_C4000_H150_6-score_vs_time}
\end{figure}

\begin{figure}[H]
    \centering
    \includegraphics[height=3.5in]{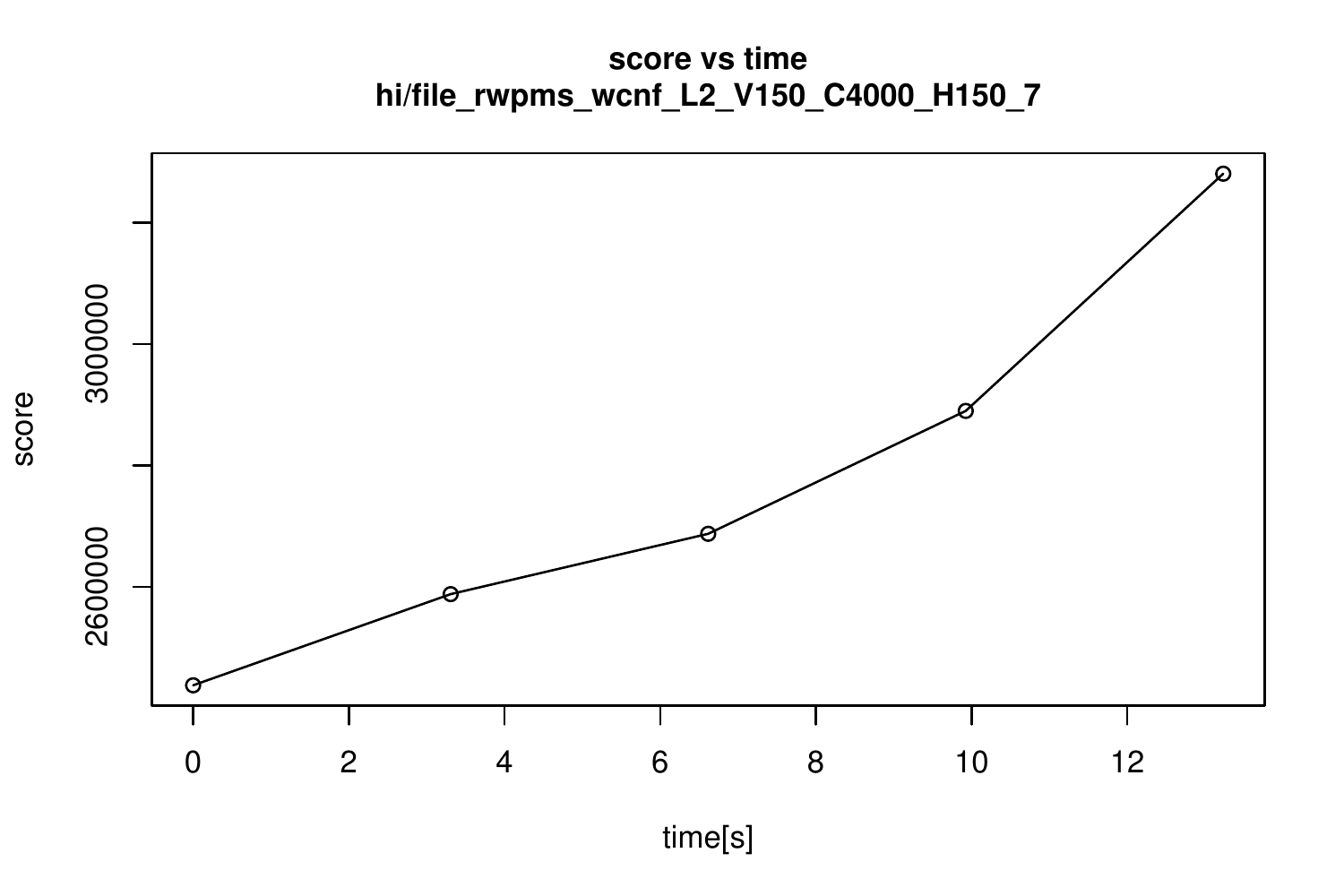}
    \label{fig_hi/file_rwpms_wcnf_L2_V150_C4000_H150_7/file_rwpms_wcnf_L2_V150_C4000_H150_7-score_vs_time}
\end{figure}

\begin{figure}[H]
    \centering
    \includegraphics[height=3.5in]{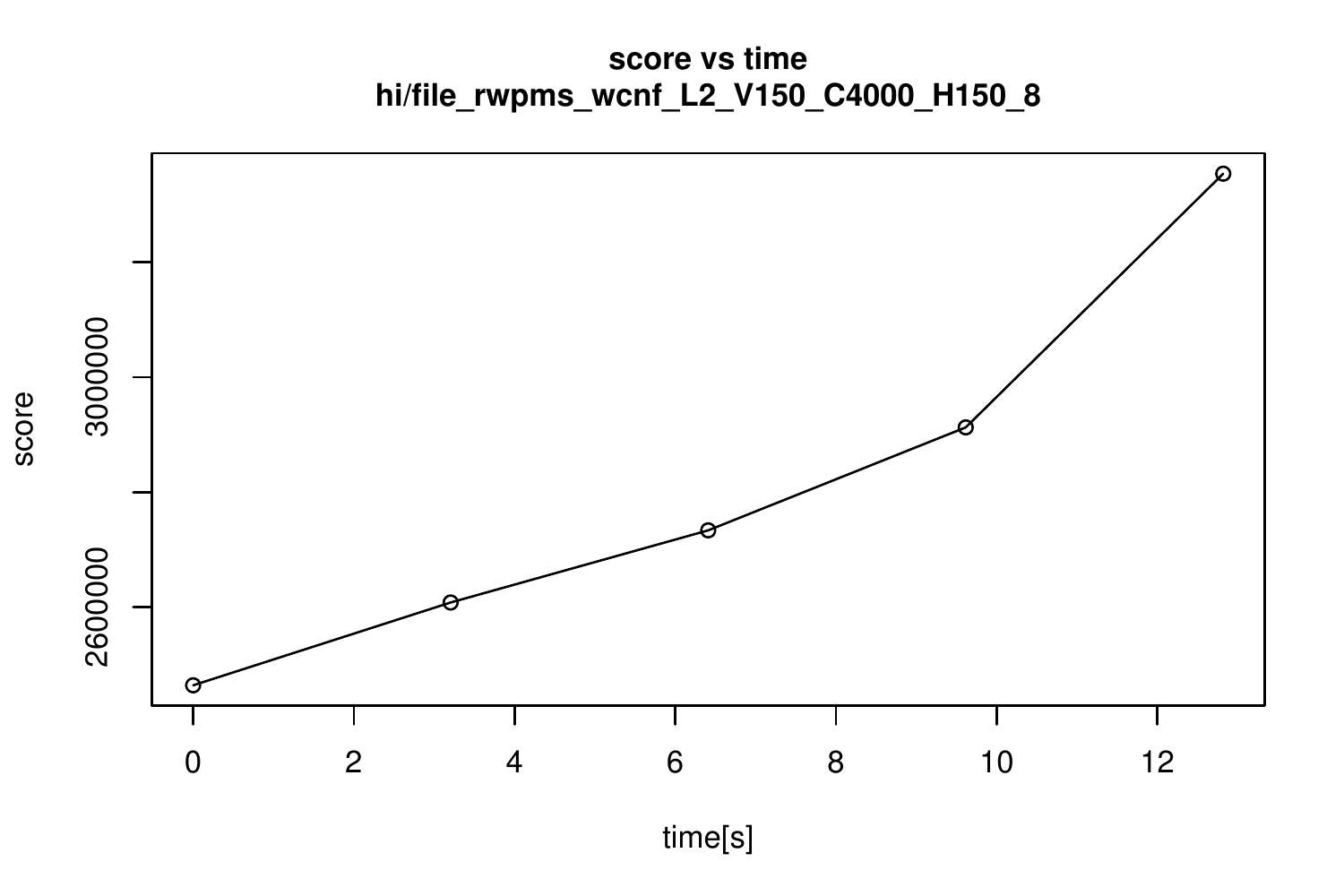}
    \label{fig_hi/file_rwpms_wcnf_L2_V150_C4000_H150_8/file_rwpms_wcnf_L2_V150_C4000_H150_8-score_vs_time}
\end{figure}

\begin{figure}[H]
    \centering
    \includegraphics[height=3.5in]{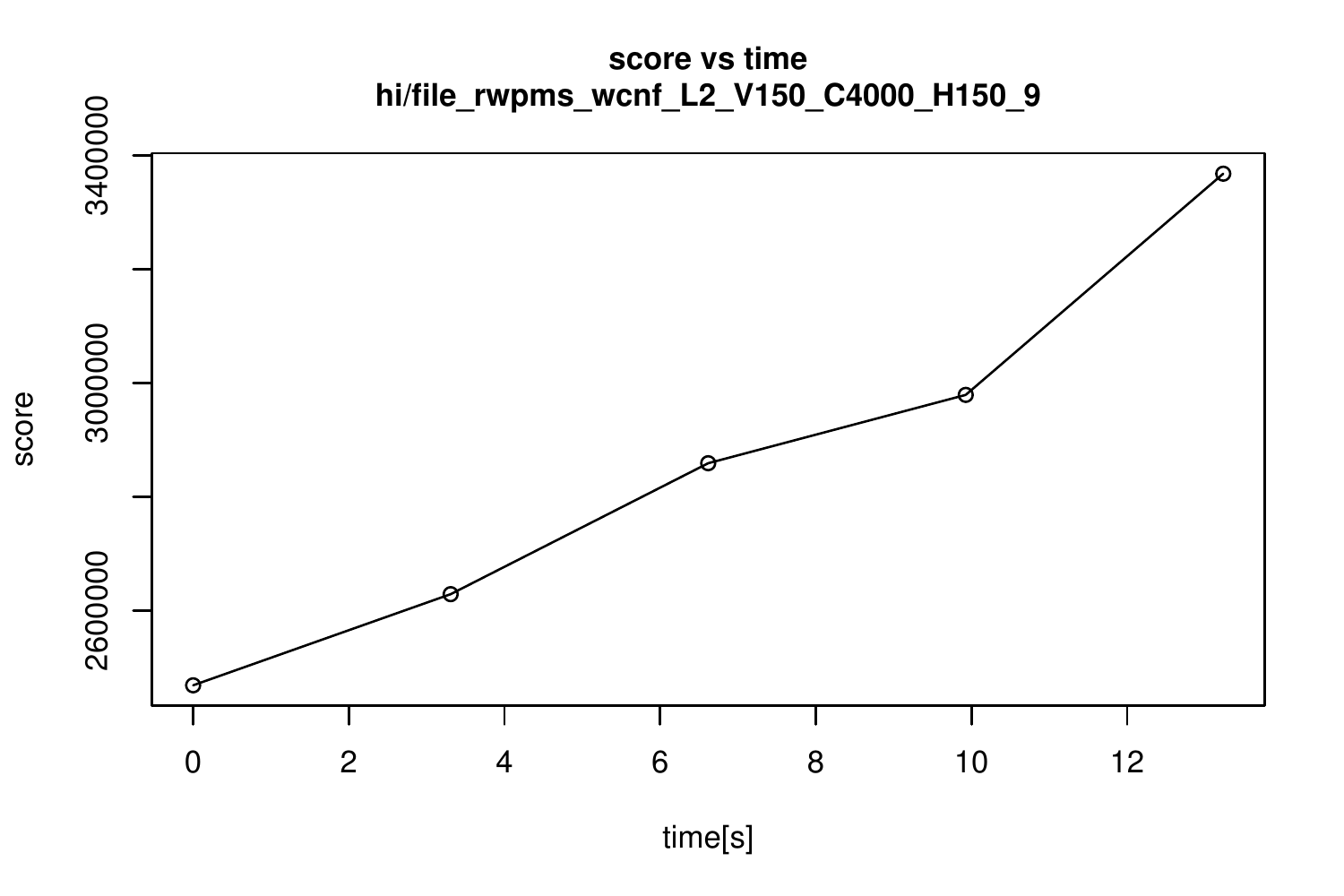}
    \label{fig_hi/file_rwpms_wcnf_L2_V150_C4000_H150_9/file_rwpms_wcnf_L2_V150_C4000_H150_9-score_vs_time}
\end{figure}

\begin{figure}[H]
    \centering
    \includegraphics[height=3.5in]{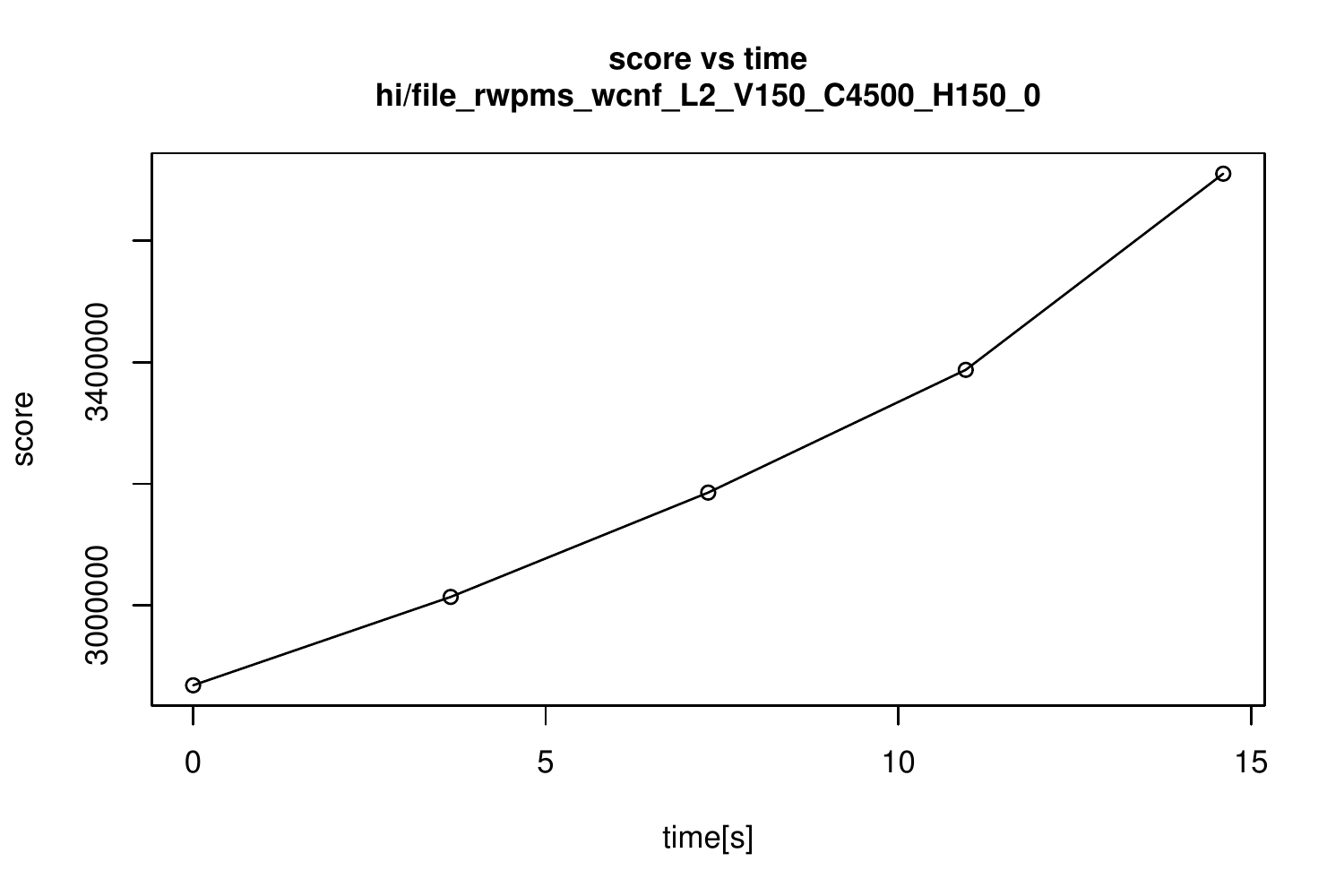}
    \label{fig_hi/file_rwpms_wcnf_L2_V150_C4500_H150_0/file_rwpms_wcnf_L2_V150_C4500_H150_0-score_vs_time}
\end{figure}

\begin{figure}[H]
    \centering
    \includegraphics[height=3.5in]{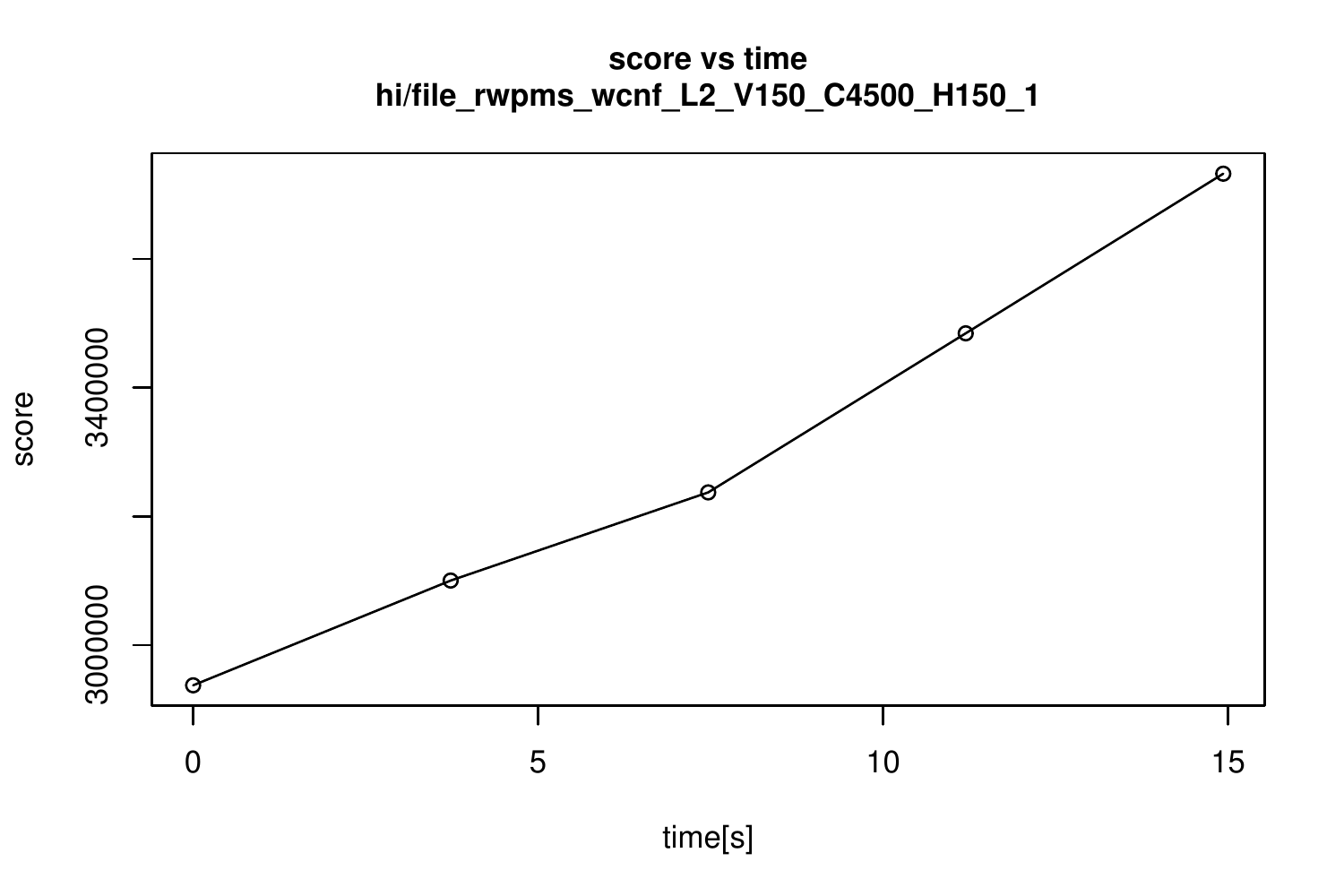}
    \label{fig_hi/file_rwpms_wcnf_L2_V150_C4500_H150_1/file_rwpms_wcnf_L2_V150_C4500_H150_1-score_vs_time}
\end{figure}

\begin{figure}[H]
    \centering
    \includegraphics[height=3.5in]{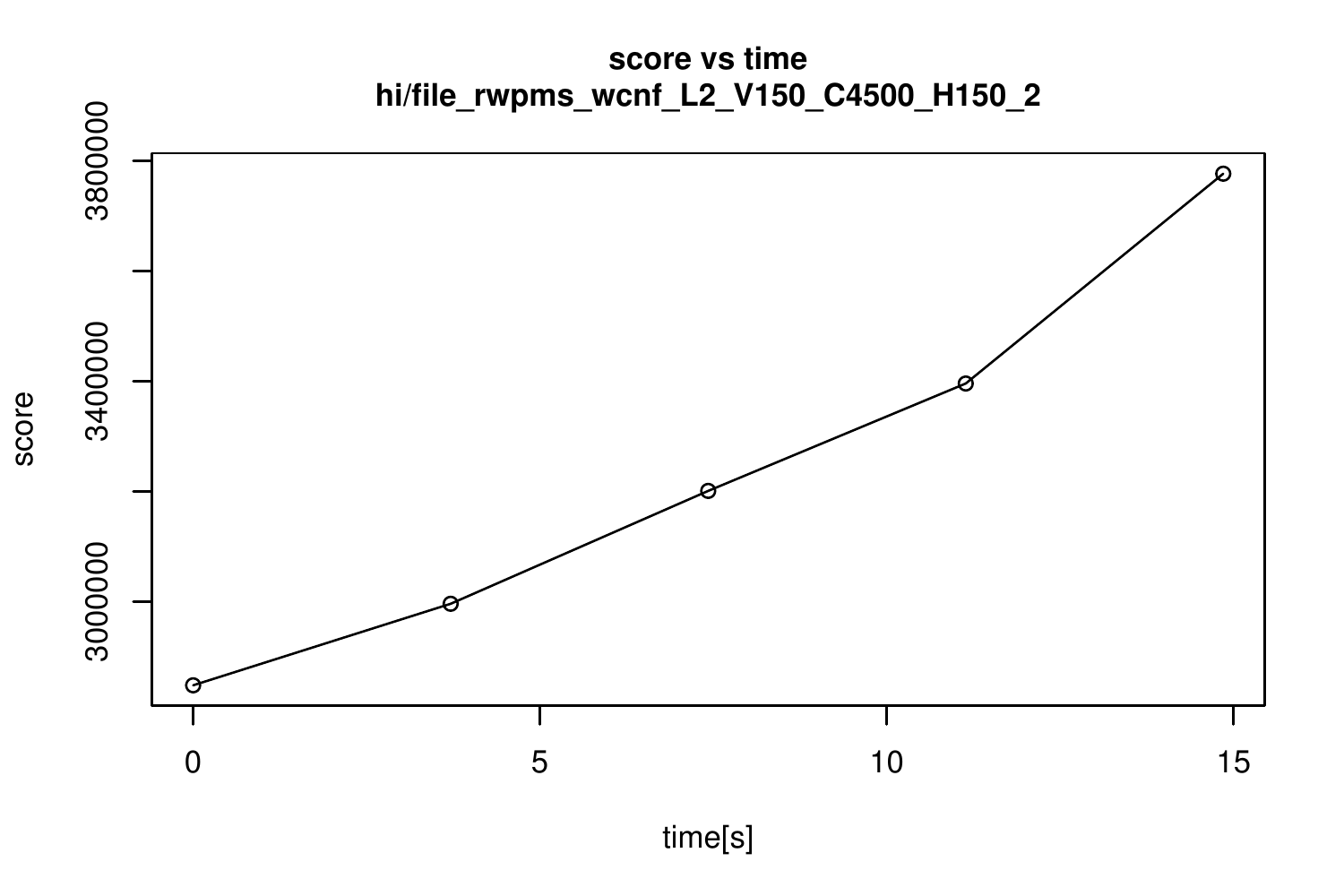}
    \label{fig_hi/file_rwpms_wcnf_L2_V150_C4500_H150_2/file_rwpms_wcnf_L2_V150_C4500_H150_2-score_vs_time}
\end{figure}

\begin{figure}[H]
    \centering
    \includegraphics[height=3.5in]{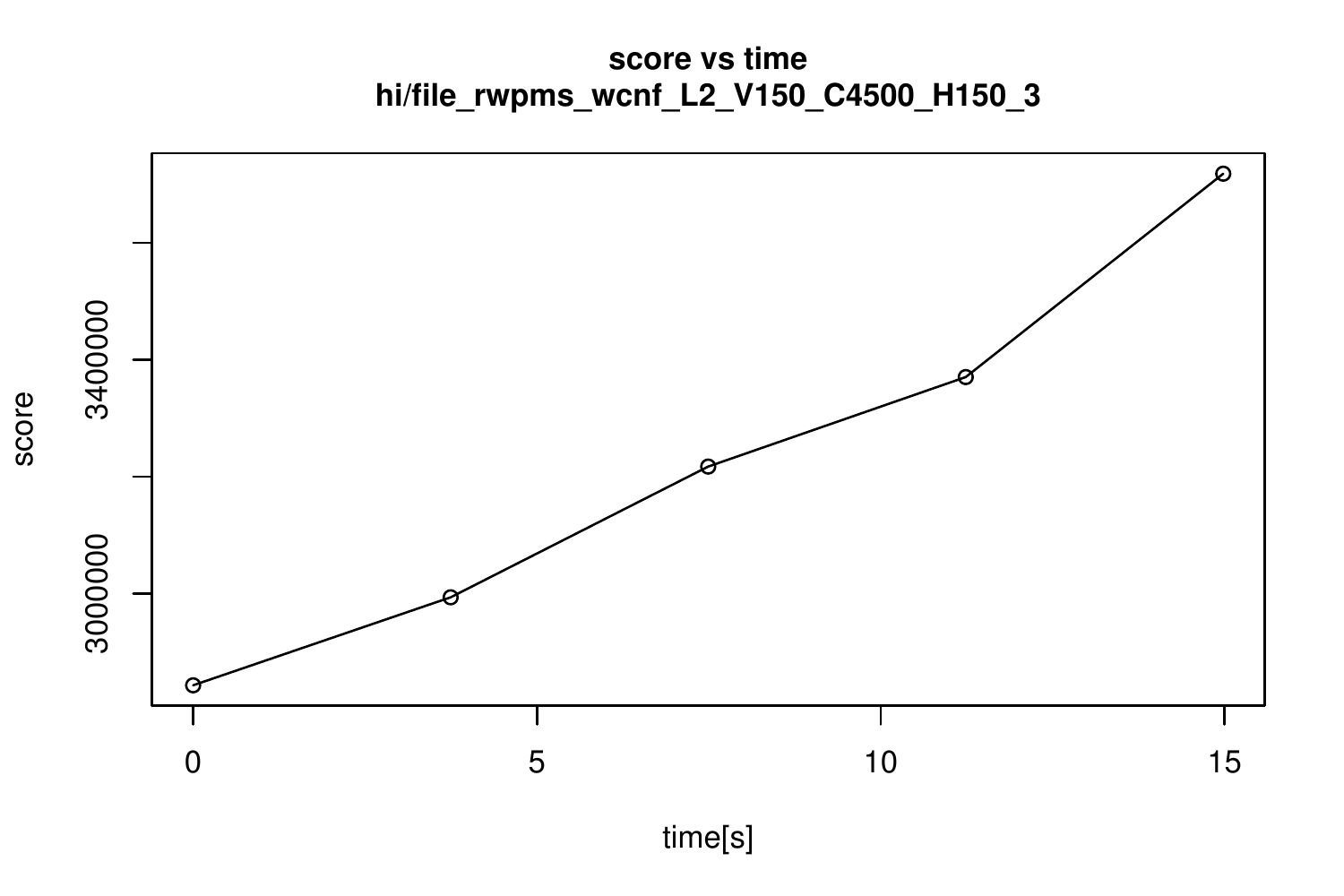}
    \label{fig_hi/file_rwpms_wcnf_L2_V150_C4500_H150_3/file_rwpms_wcnf_L2_V150_C4500_H150_3-score_vs_time}
\end{figure}

\begin{figure}[H]
    \centering
    \includegraphics[height=3.5in]{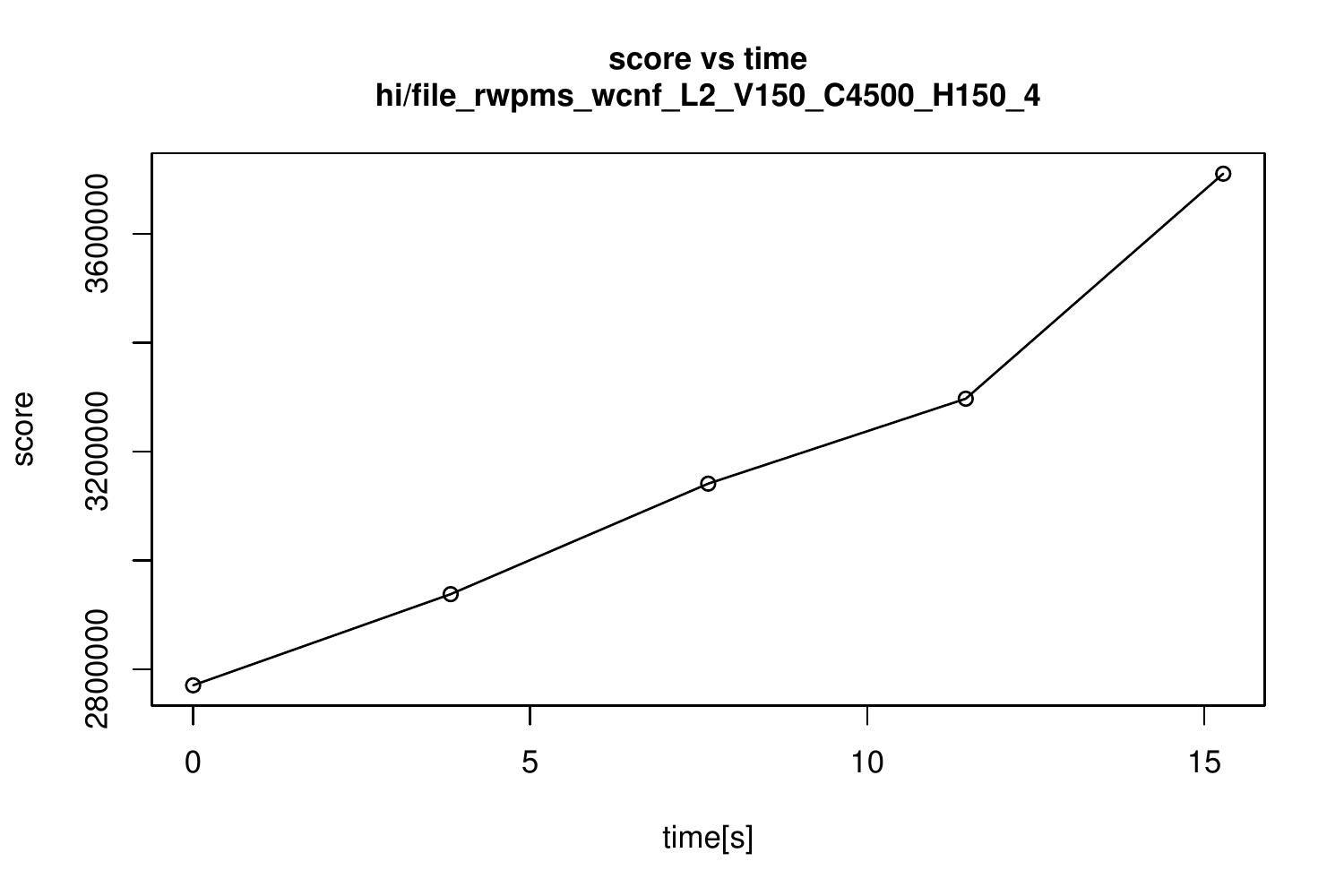}
    \label{fig_hi/file_rwpms_wcnf_L2_V150_C4500_H150_4/file_rwpms_wcnf_L2_V150_C4500_H150_4-score_vs_time}
\end{figure}

\begin{figure}[H]
    \centering
    \includegraphics[height=3.5in]{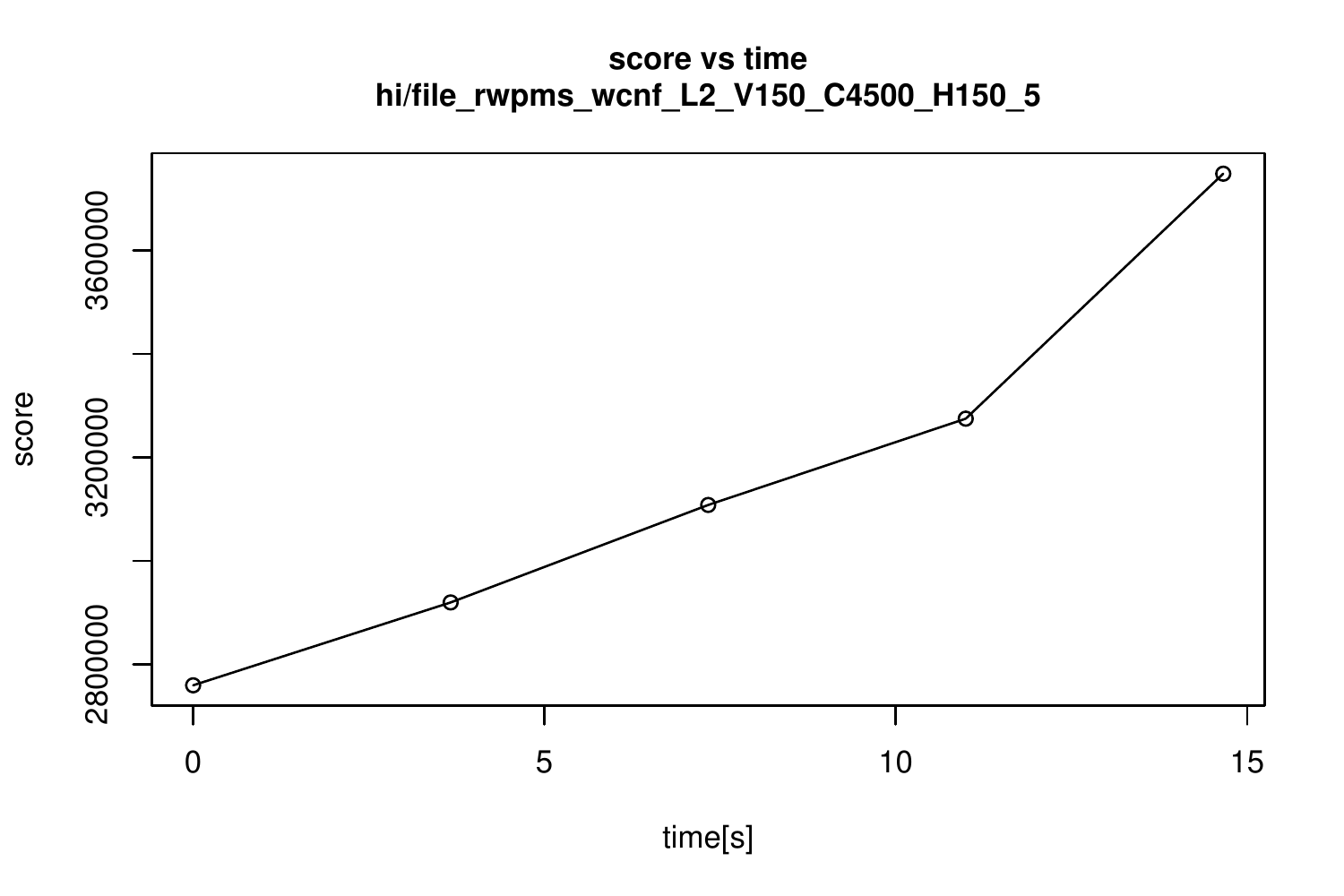}
    \label{fig_hi/file_rwpms_wcnf_L2_V150_C4500_H150_5/file_rwpms_wcnf_L2_V150_C4500_H150_5-score_vs_time}
\end{figure}

\begin{figure}[H]
    \centering
    \includegraphics[height=3.5in]{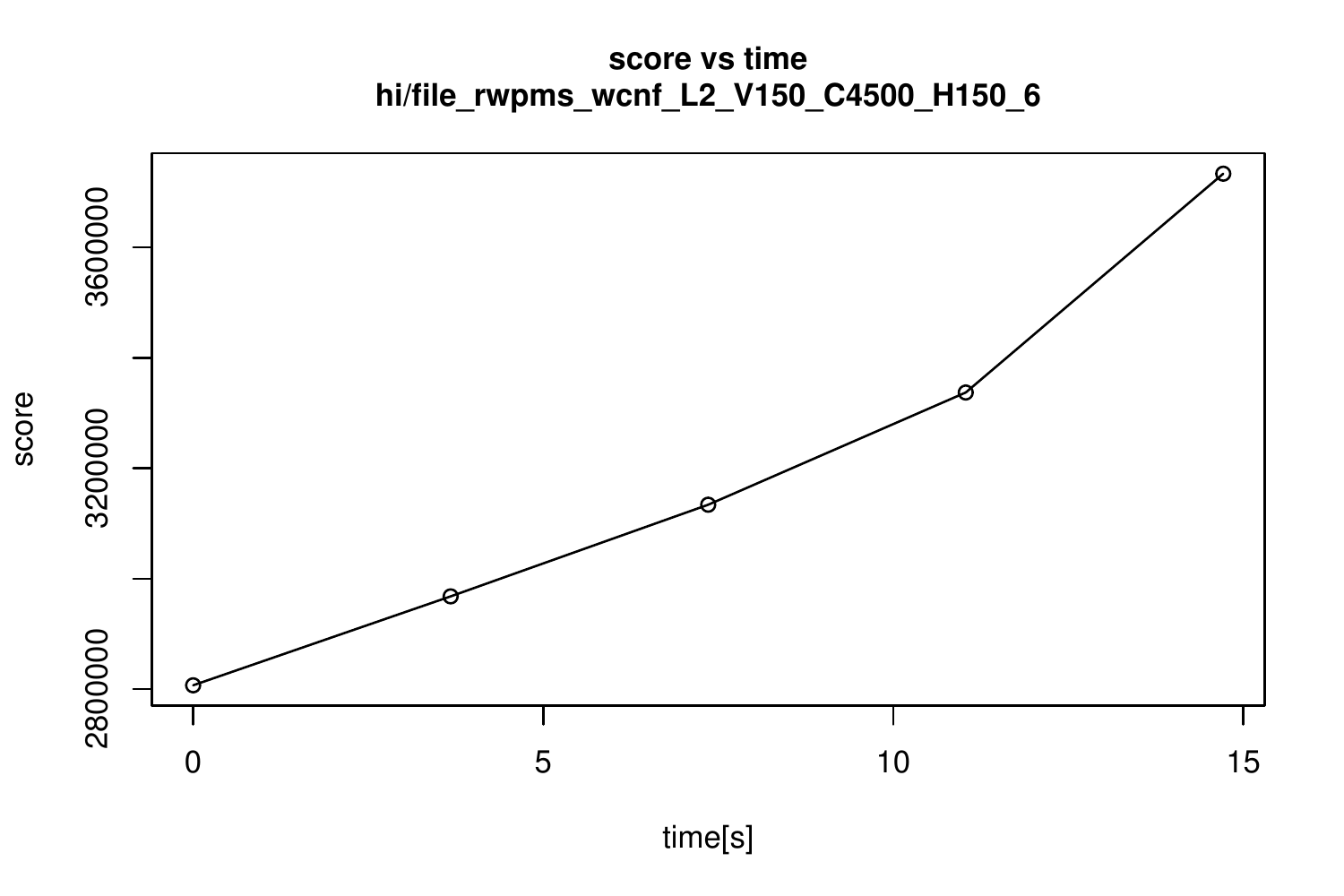}
    \label{fig_hi/file_rwpms_wcnf_L2_V150_C4500_H150_6/file_rwpms_wcnf_L2_V150_C4500_H150_6-score_vs_time}
\end{figure}

\begin{figure}[H]
    \centering
    \includegraphics[height=3.5in]{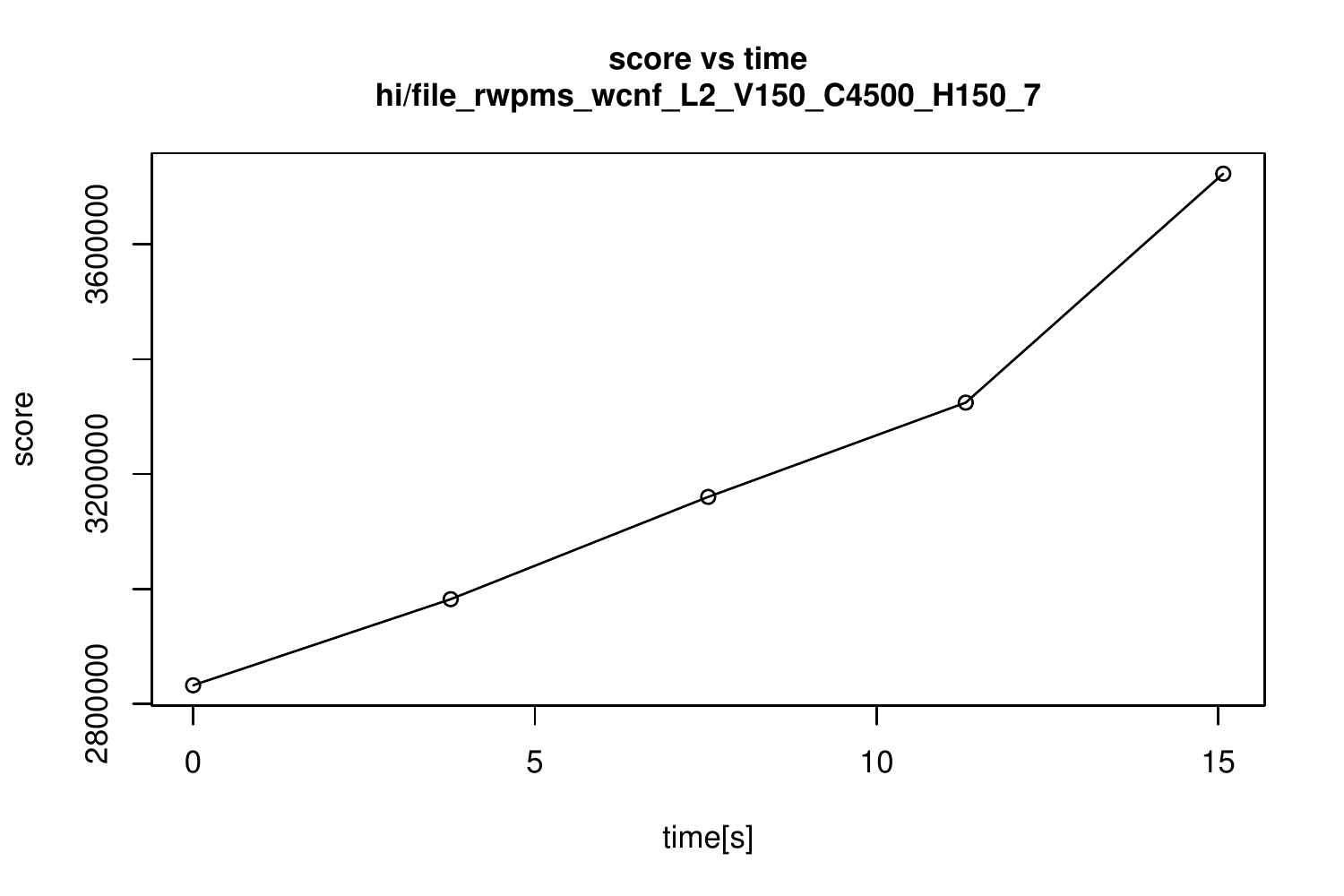}
    \label{fig_hi/file_rwpms_wcnf_L2_V150_C4500_H150_7/file_rwpms_wcnf_L2_V150_C4500_H150_7-score_vs_time}
\end{figure}

\begin{figure}[H]
    \centering
    \includegraphics[height=3.5in]{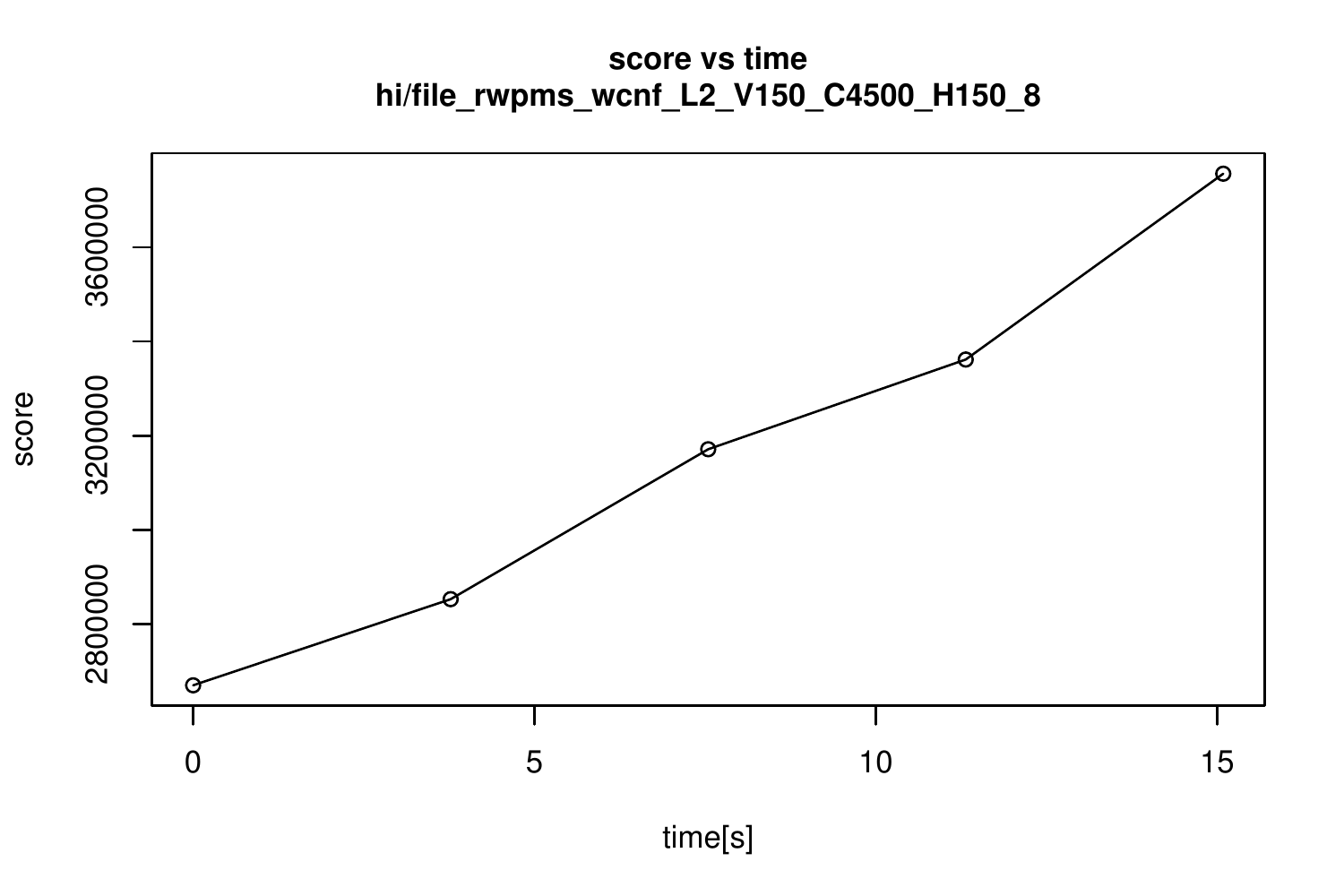}
    \label{fig_hi/file_rwpms_wcnf_L2_V150_C4500_H150_8/file_rwpms_wcnf_L2_V150_C4500_H150_8-score_vs_time}
\end{figure}

\begin{figure}[H]
    \centering
    \includegraphics[height=3.5in]{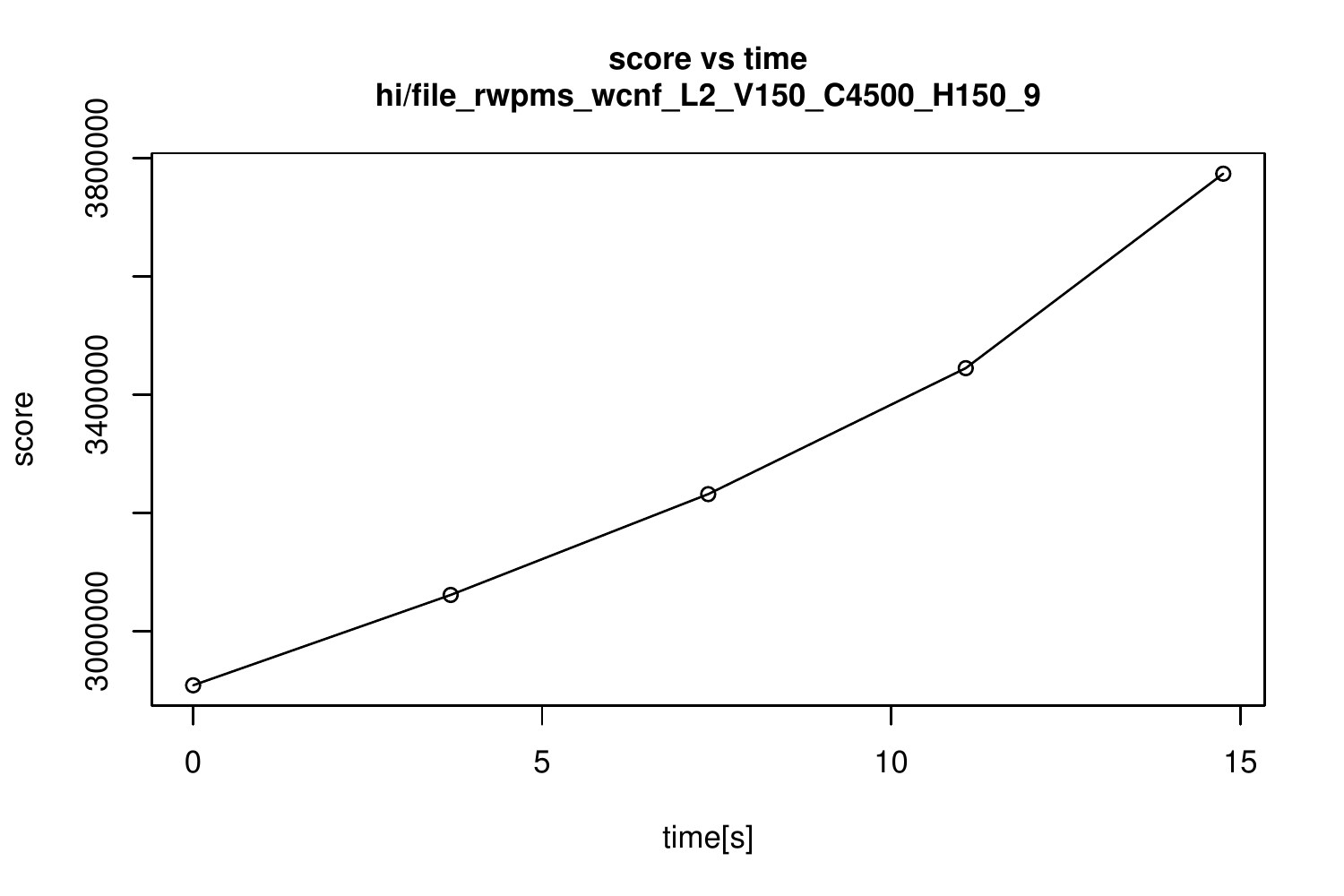}
    \label{fig_hi/file_rwpms_wcnf_L2_V150_C4500_H150_9/file_rwpms_wcnf_L2_V150_C4500_H150_9-score_vs_time}
\end{figure}

\begin{figure}[H]
    \centering
    \includegraphics[height=3.5in]{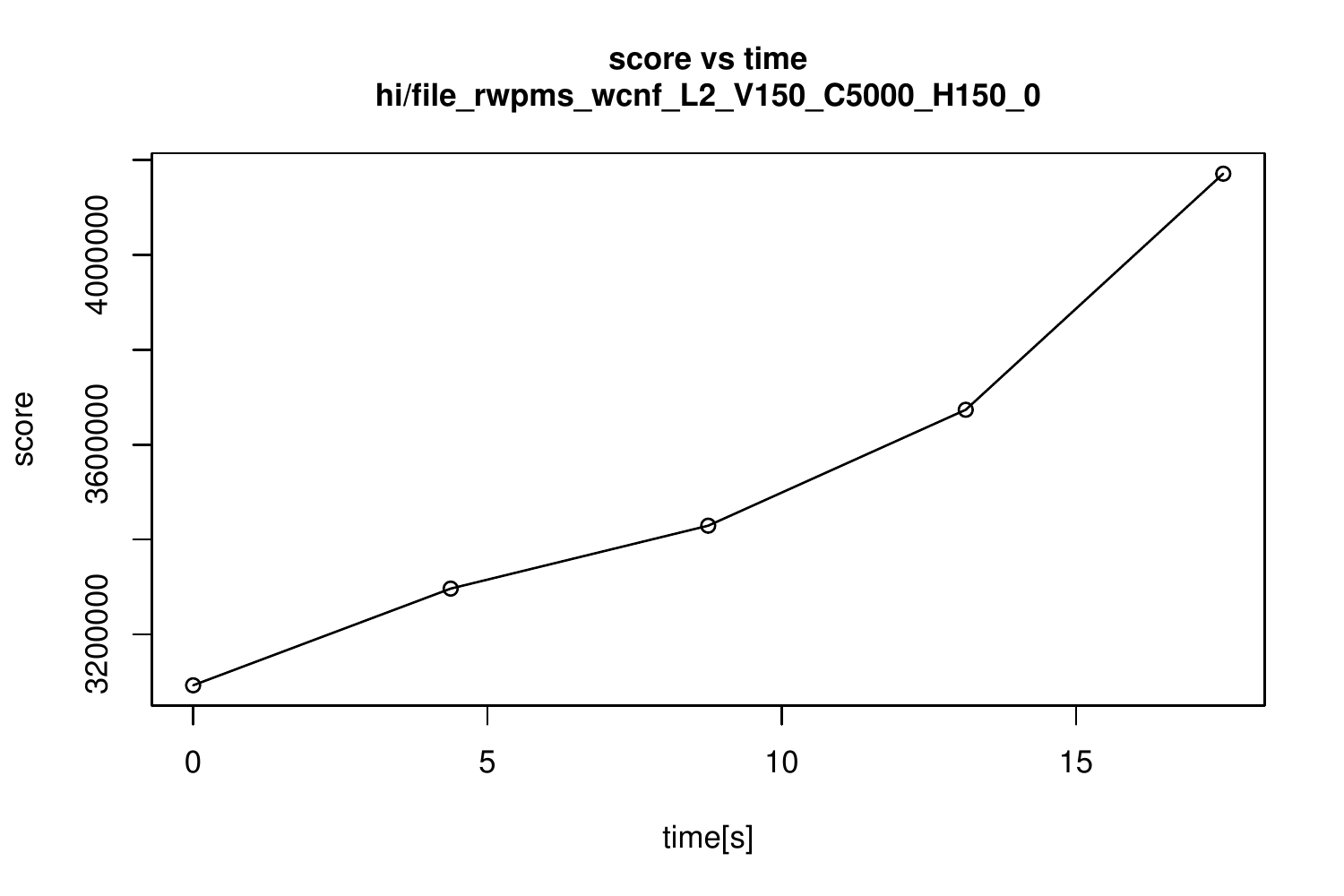}
    \label{fig_hi/file_rwpms_wcnf_L2_V150_C5000_H150_0/file_rwpms_wcnf_L2_V150_C5000_H150_0-score_vs_time}
\end{figure}

\begin{figure}[H]
    \centering
    \includegraphics[height=3.5in]{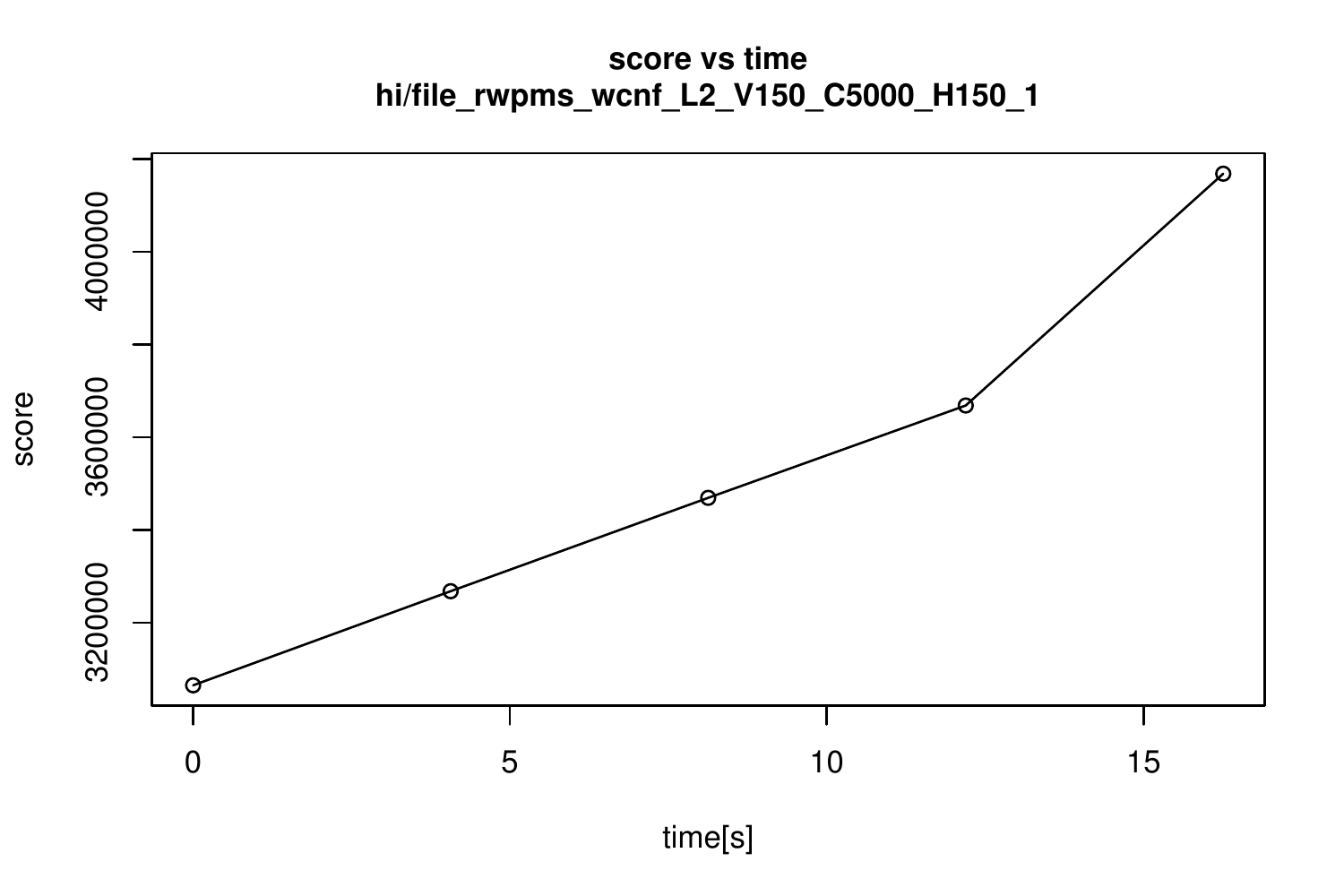}
    \label{fig_hi/file_rwpms_wcnf_L2_V150_C5000_H150_1/file_rwpms_wcnf_L2_V150_C5000_H150_1-score_vs_time}
\end{figure}

\begin{figure}[H]
    \centering
    \includegraphics[height=3.5in]{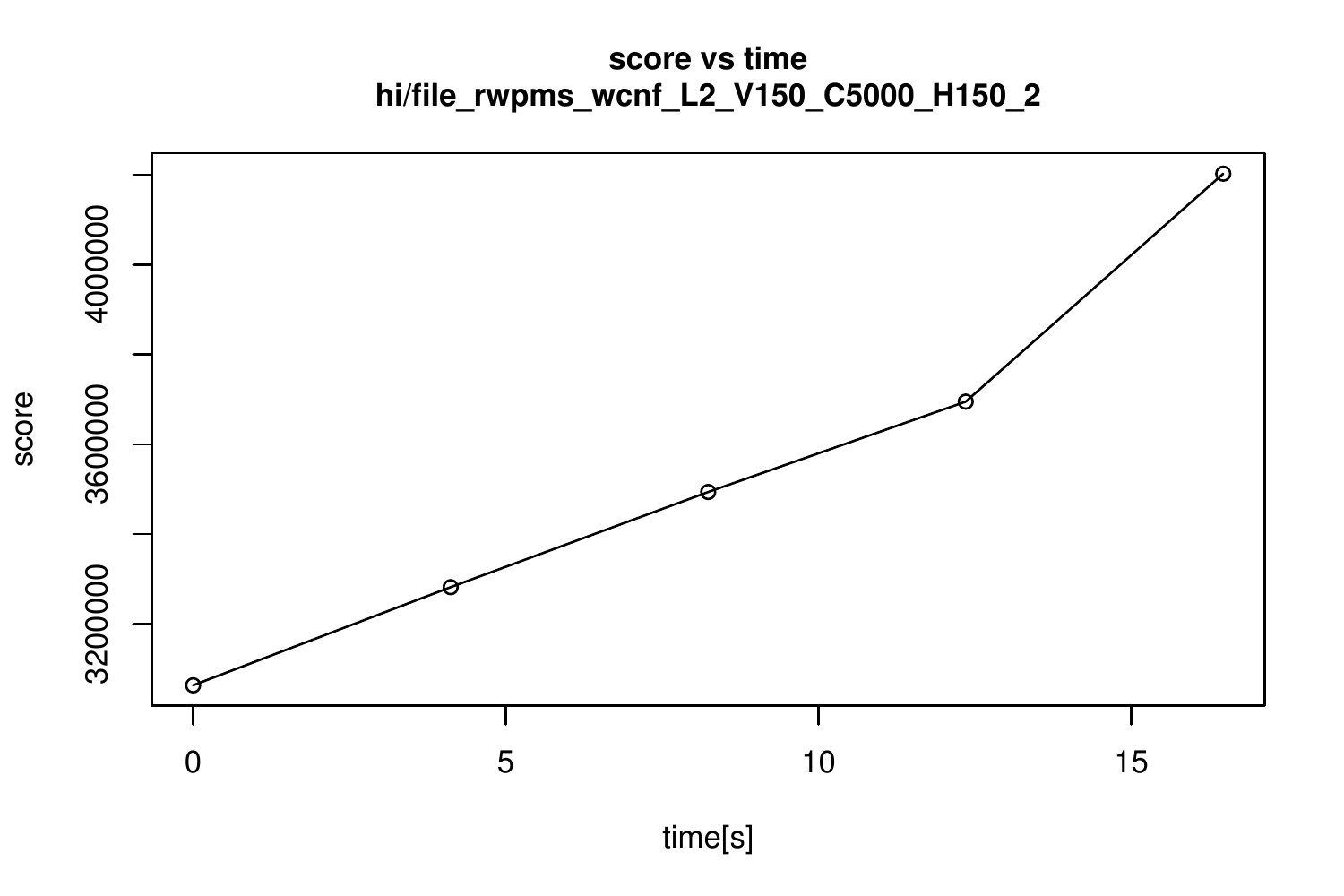}
    \label{fig_hi/file_rwpms_wcnf_L2_V150_C5000_H150_2/file_rwpms_wcnf_L2_V150_C5000_H150_2-score_vs_time}
\end{figure}

\begin{figure}[H]
    \centering
    \includegraphics[height=3.5in]{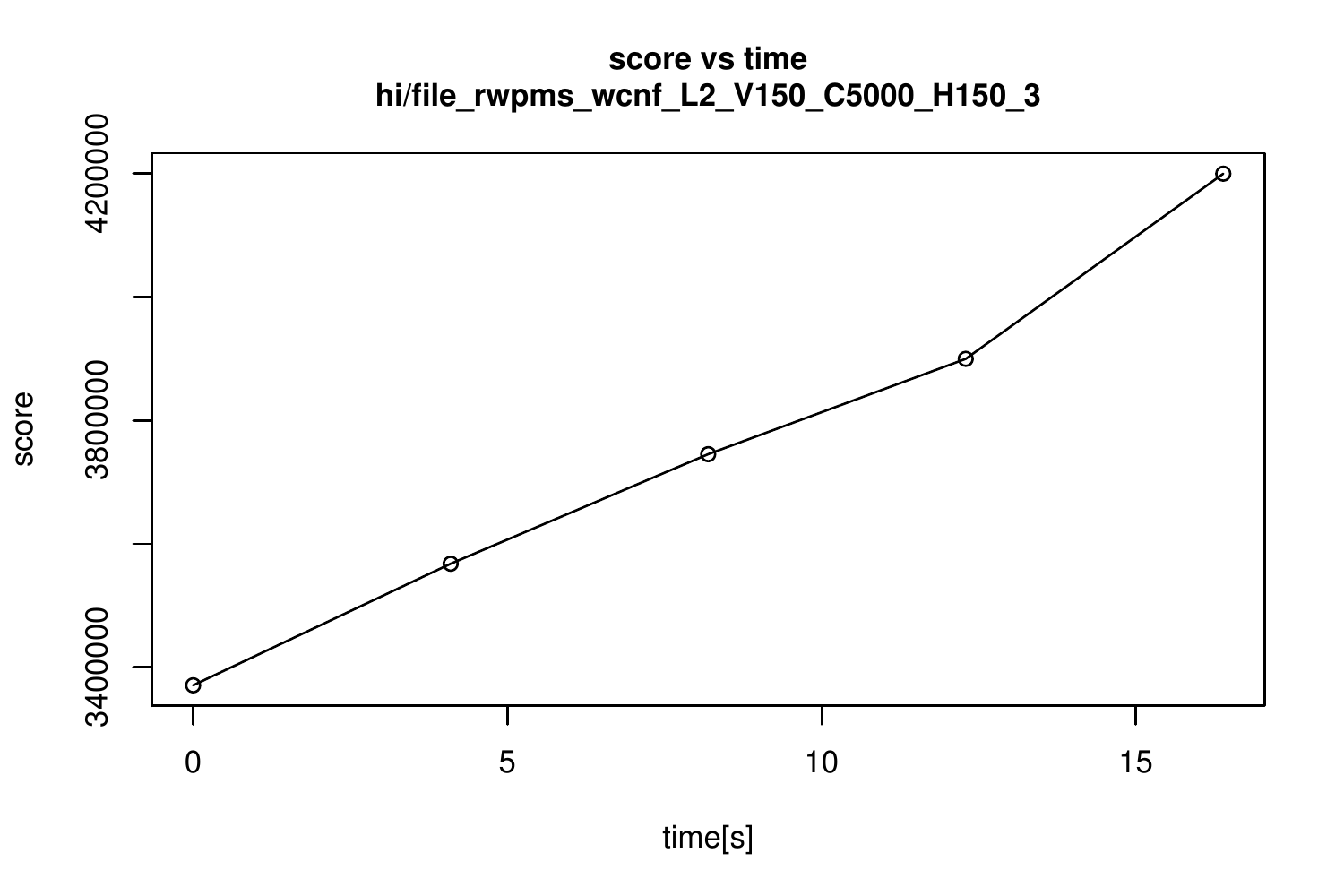}
    \label{fig_hi/file_rwpms_wcnf_L2_V150_C5000_H150_3/file_rwpms_wcnf_L2_V150_C5000_H150_3-score_vs_time}
\end{figure}

\begin{figure}[H]
    \centering
    \includegraphics[height=3.5in]{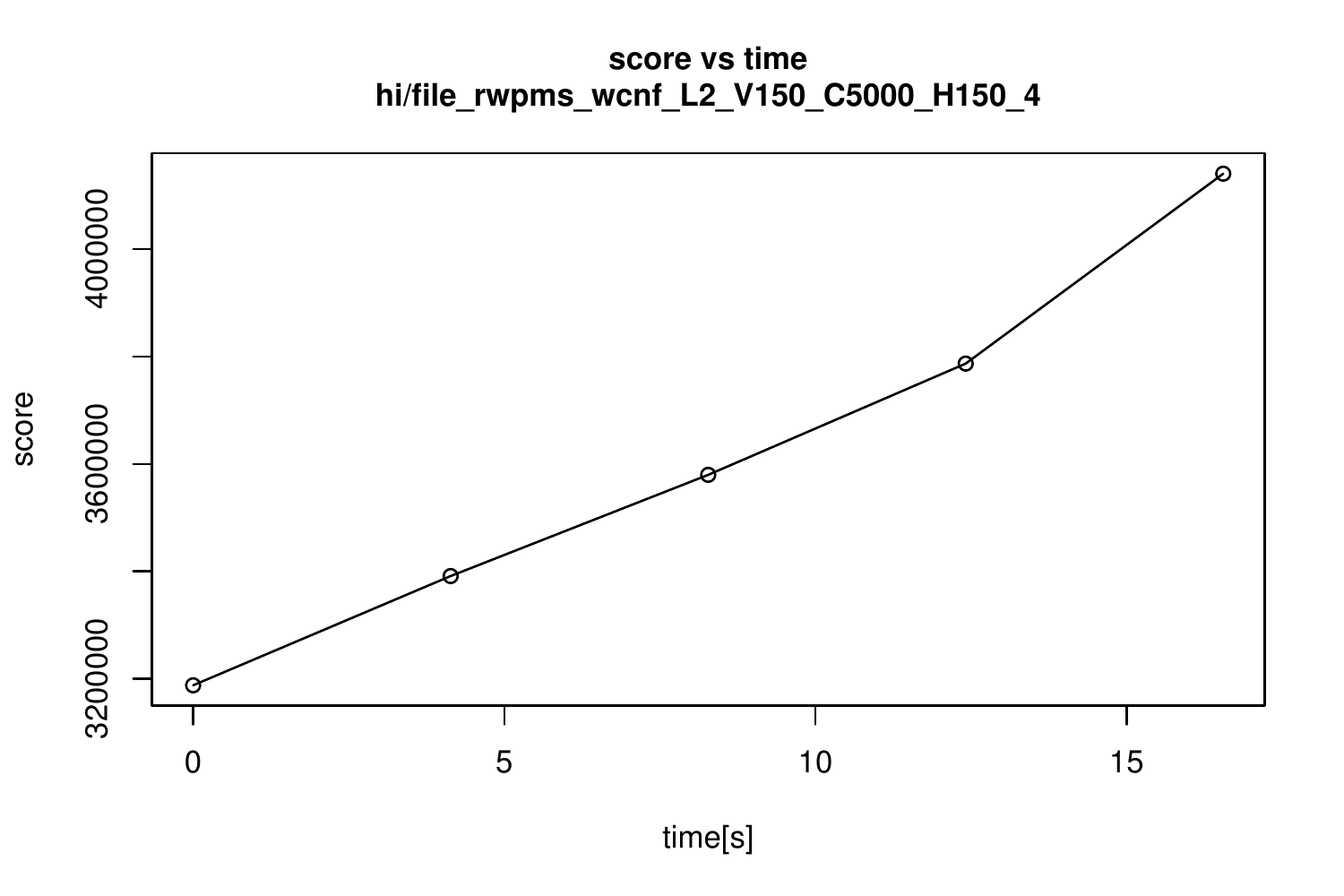}
    \label{fig_hi/file_rwpms_wcnf_L2_V150_C5000_H150_4/file_rwpms_wcnf_L2_V150_C5000_H150_4-score_vs_time}
\end{figure}

\begin{figure}[H]
    \centering
    \includegraphics[height=3.5in]{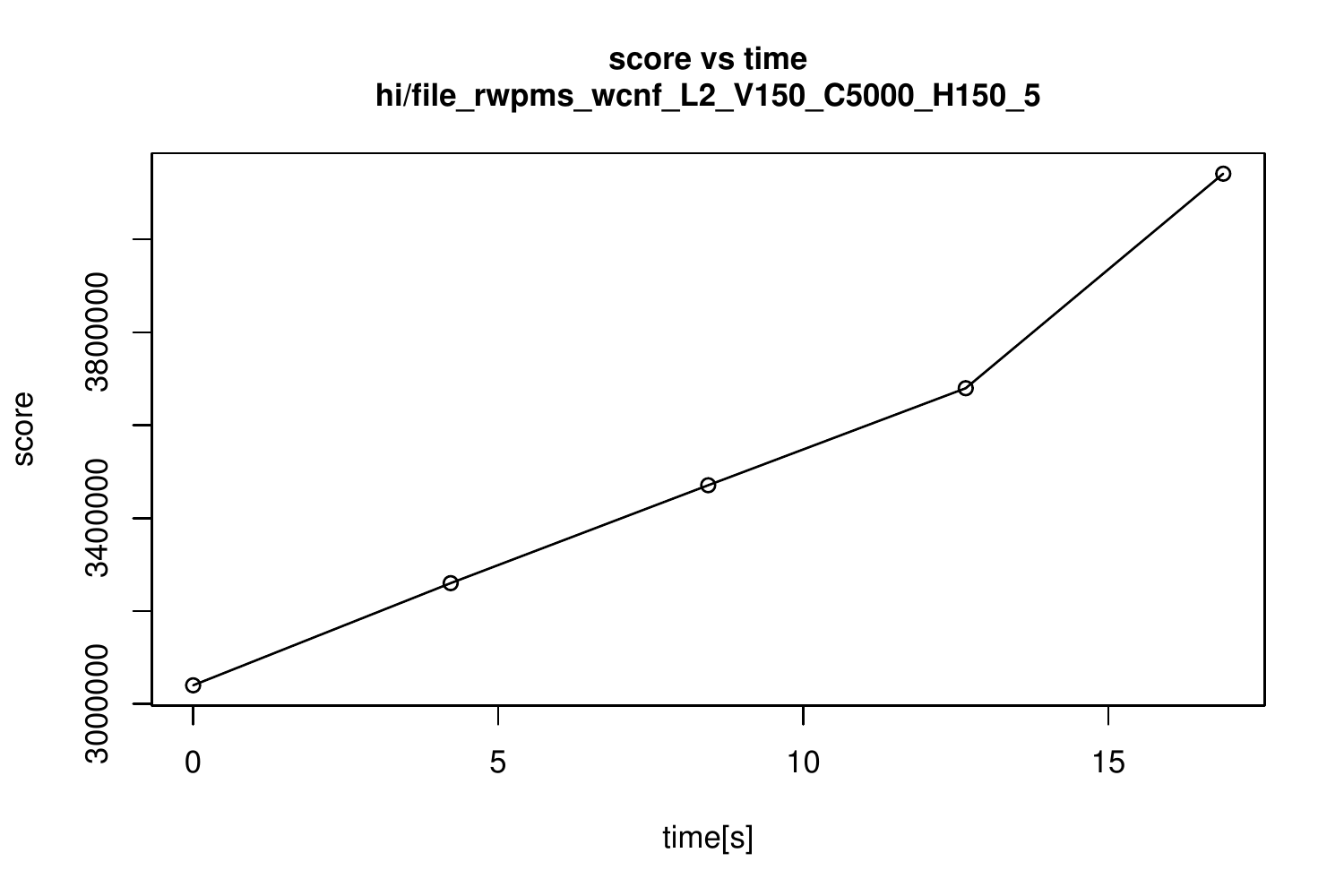}
    \label{fig_hi/file_rwpms_wcnf_L2_V150_C5000_H150_5/file_rwpms_wcnf_L2_V150_C5000_H150_5-score_vs_time}
\end{figure}

\begin{figure}[H]
    \centering
    \includegraphics[height=3.5in]{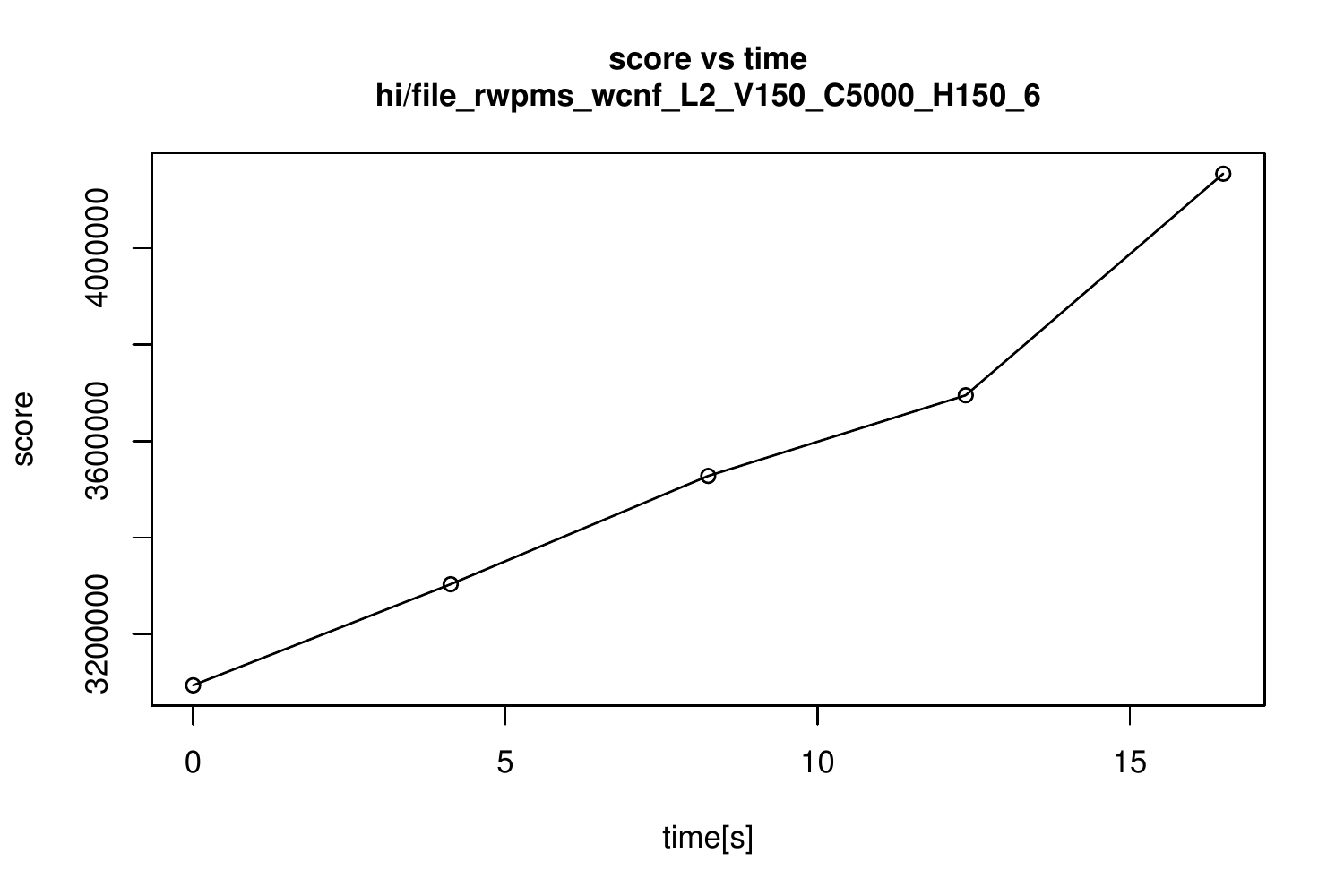}
    \label{fig_hi/file_rwpms_wcnf_L2_V150_C5000_H150_6/file_rwpms_wcnf_L2_V150_C5000_H150_6-score_vs_time}
\end{figure}

\begin{figure}[H]
    \centering
    \includegraphics[height=3.5in]{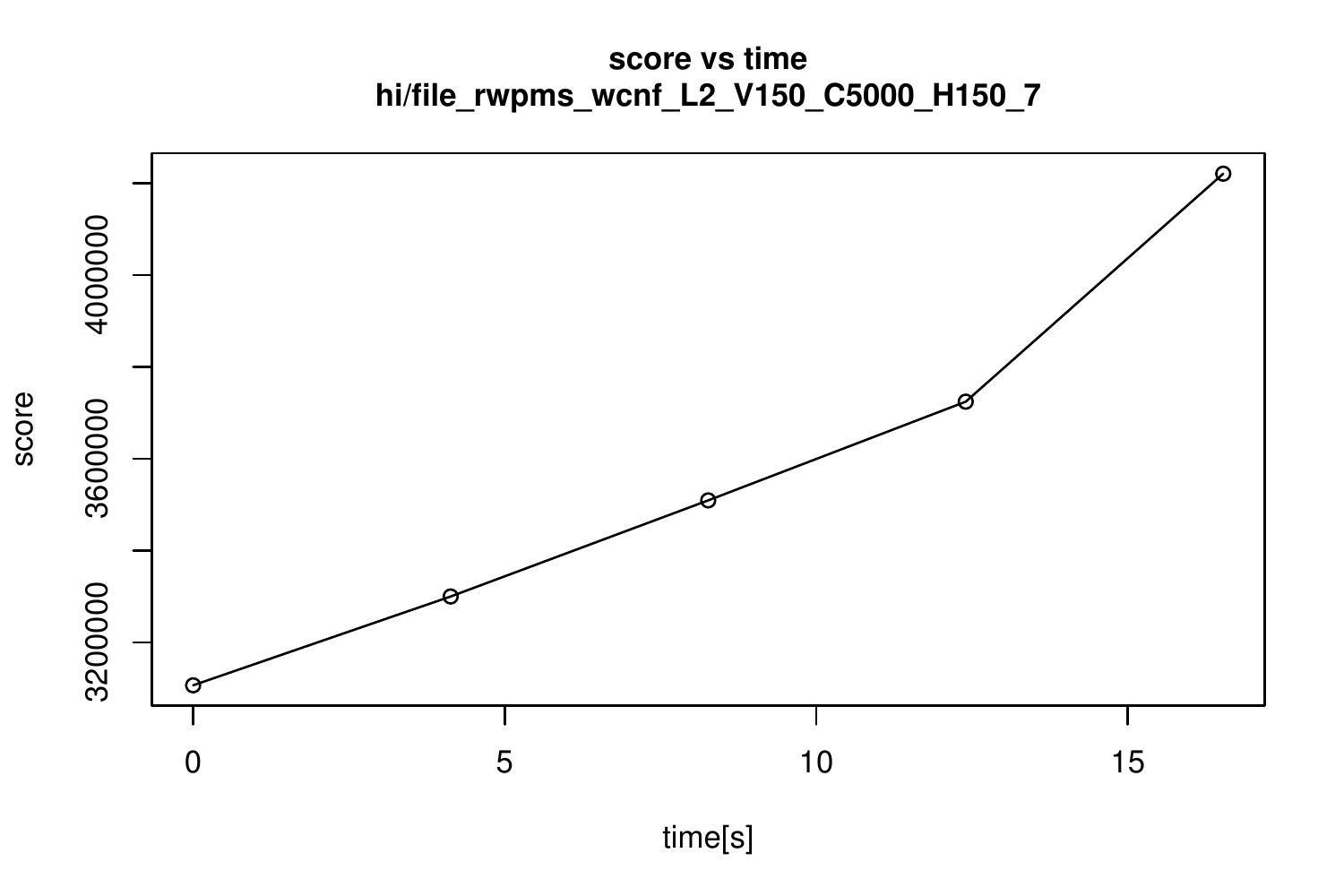}
    \label{fig_hi/file_rwpms_wcnf_L2_V150_C5000_H150_7/file_rwpms_wcnf_L2_V150_C5000_H150_7-score_vs_time}
\end{figure}

\begin{figure}[H]
    \centering
    \includegraphics[height=3.5in]{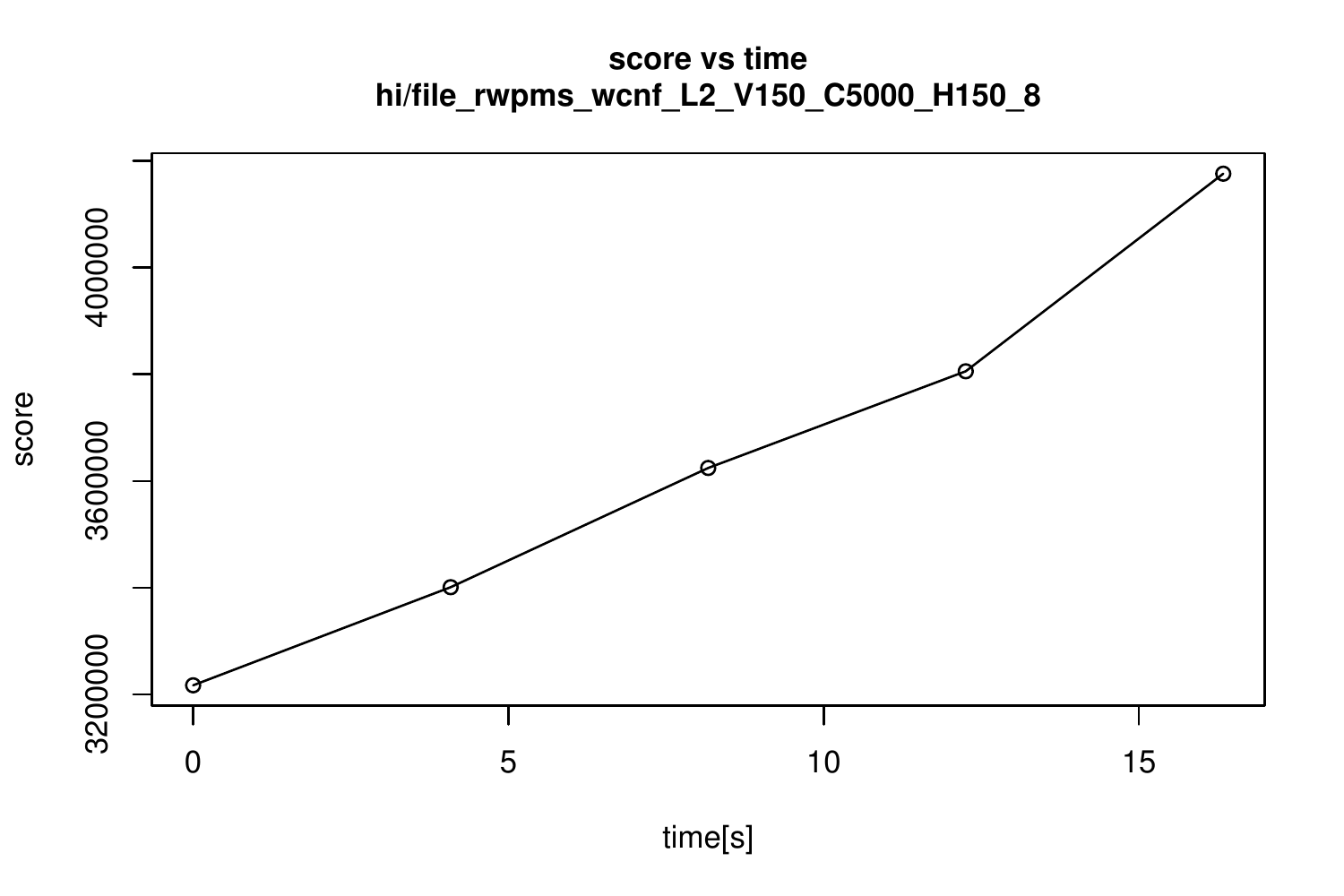}
    \label{fig_hi/file_rwpms_wcnf_L2_V150_C5000_H150_8/file_rwpms_wcnf_L2_V150_C5000_H150_8-score_vs_time}
\end{figure}

\begin{figure}[H]
    \centering
    \includegraphics[height=3.5in]{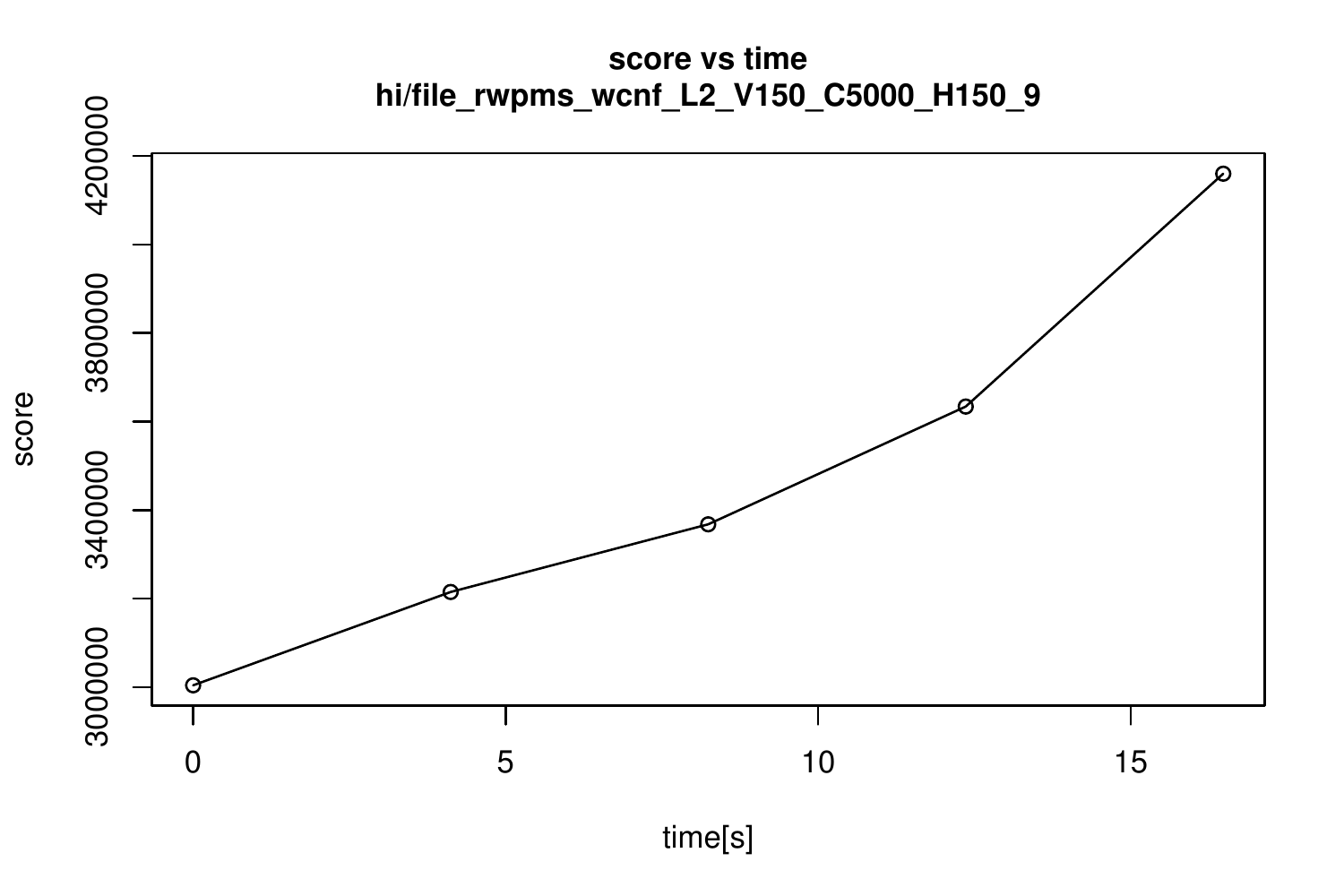}
    \label{fig_hi/file_rwpms_wcnf_L2_V150_C5000_H150_9/file_rwpms_wcnf_L2_V150_C5000_H150_9-score_vs_time}
\end{figure}

\chapter{RDMaxWalkSat Solution Deltas\label{rdmaxwalksat_deltas}}

\begin{center}
    \begin{longtable}{|l|c|c|c|c|}
        \caption{Table showing the quality of the solution RDMaxWalkSat has genereated at various points in time as a percentage of the maximum quality solution found by MaxWalkSat for the same problem.} \label{tbl_rdmaxwalksat_deltas} \\
        \hline
        & \multicolumn{4}{|c|}{\textbf{Threshold intervals}} \\
        \textbf{Problem} & $0-25\%$ & $26-50\%$ & $51-75\%$ & $76-100\%$ \\
        \hline\hline
    lo/file\_rwpms\_wcnf\_L2\_V150\_C1000\_H150\_0 & 0.0 & 9.4 & 5.5 & 9.4 \\
    lo/file\_rwpms\_wcnf\_L2\_V150\_C1000\_H150\_1 & 0.0 & 8.2 & 5.4 & 13.1 \\
    lo/file\_rwpms\_wcnf\_L2\_V150\_C1000\_H150\_2 & 0.0 & 6.7 & 5.6 & 13.6 \\
    lo/file\_rwpms\_wcnf\_L2\_V150\_C1000\_H150\_3 & 0.0 & 4.7 & 7.3 & 12.8 \\
    lo/file\_rwpms\_wcnf\_L2\_V150\_C1000\_H150\_4 & 0.0 & 6.9 & 7.9 & 13.2 \\
    lo/file\_rwpms\_wcnf\_L2\_V150\_C1000\_H150\_5 & 0.0 & 6.7 & 6.5 & 15.9 \\
    lo/file\_rwpms\_wcnf\_L2\_V150\_C1000\_H150\_6 & 0.0 & 6.6 & 5.8 & 13.7 \\
    lo/file\_rwpms\_wcnf\_L2\_V150\_C1000\_H150\_7 & 0.0 & 6.9 & 4.3 & 15.2 \\
    lo/file\_rwpms\_wcnf\_L2\_V150\_C1000\_H150\_8 & 0.0 & 8.1 & 4.5 & 10.7 \\
    lo/file\_rwpms\_wcnf\_L2\_V150\_C1000\_H150\_9 & 0.0 & 5.6 & 5.6 & 16.0 \\
    lo/file\_rwpms\_wcnf\_L2\_V150\_C1500\_H150\_0 & 4.6 & 3.9 & 4.9 & 11.1 \\
    lo/file\_rwpms\_wcnf\_L2\_V150\_C1500\_H150\_1 & 4.4 & 3.3 & 4.6 & 10.9 \\
    lo/file\_rwpms\_wcnf\_L2\_V150\_C1500\_H150\_2 & 4.8 & 2.9 & 5.8 & 13.6 \\
    lo/file\_rwpms\_wcnf\_L2\_V150\_C1500\_H150\_3 & 4.6 & 3.9 & 4.0 & 11.4 \\
    lo/file\_rwpms\_wcnf\_L2\_V150\_C1500\_H150\_4 & 4.4 & 4.1 & 3.1 & 12.3 \\
    lo/file\_rwpms\_wcnf\_L2\_V150\_C1500\_H150\_5 & 4.8 & 5.0 & 5.1 & 12.0 \\
    lo/file\_rwpms\_wcnf\_L2\_V150\_C1500\_H150\_6 & 5.5 & 3.6 & 5.8 & 8.9 \\
    lo/file\_rwpms\_wcnf\_L2\_V150\_C1500\_H150\_7 & 5.1 & 4.2 & 4.4 & 16.8 \\
    lo/file\_rwpms\_wcnf\_L2\_V150\_C1500\_H150\_8 & 5.1 & 4.4 & 4.4 & 11.8 \\
    lo/file\_rwpms\_wcnf\_L2\_V150\_C1500\_H150\_9 & 4.9 & 3.7 & 3.3 & 10.4 \\
    lo/file\_rwpms\_wcnf\_L2\_V150\_C2000\_H150\_0 & 3.9 & 6.4 & 5.0 & 9.7 \\
    lo/file\_rwpms\_wcnf\_L2\_V150\_C2000\_H150\_1 & 3.3 & 1.5 & 5.4 & 15.4 \\
    lo/file\_rwpms\_wcnf\_L2\_V150\_C2000\_H150\_2 & 3.6 & 3.2 & 5.3 & 11.4 \\
    lo/file\_rwpms\_wcnf\_L2\_V150\_C2000\_H150\_3 & 3.7 & 3.0 & 6.0 & 15.6 \\
    lo/file\_rwpms\_wcnf\_L2\_V150\_C2000\_H150\_4 & 3.4 & 3.5 & 6.2 & 16.9 \\
    lo/file\_rwpms\_wcnf\_L2\_V150\_C2000\_H150\_5 & 3.8 & 2.3 & 6.8 & 13.3 \\
    lo/file\_rwpms\_wcnf\_L2\_V150\_C2000\_H150\_6 & 3.6 & 6.8 & 3.8 & 10.9 \\
    lo/file\_rwpms\_wcnf\_L2\_V150\_C2000\_H150\_7 & 3.7 & 6.1 & 2.7 & 12.9 \\
    lo/file\_rwpms\_wcnf\_L2\_V150\_C2000\_H150\_8 & 4.1 & 6.3 & 3.8 & 9.4 \\
    lo/file\_rwpms\_wcnf\_L2\_V150\_C2000\_H150\_9 & 3.0 & 2.7 & 6.6 & 10.1 \\
    me/file\_rwpms\_wcnf\_L2\_V150\_C2500\_H150\_0 & 5.1 & 4.2 & 4.4 & 12.5 \\
    me/file\_rwpms\_wcnf\_L2\_V150\_C2500\_H150\_1 & 5.0 & 4.9 & 4.9 & 9.2 \\
    me/file\_rwpms\_wcnf\_L2\_V150\_C2500\_H150\_2 & 5.2 & 3.0 & 3.9 & 11.9 \\
    me/file\_rwpms\_wcnf\_L2\_V150\_C2500\_H150\_3 & 4.6 & 2.5 & 5.6 & 8.5 \\
    me/file\_rwpms\_wcnf\_L2\_V150\_C2500\_H150\_4 & 4.7 & 4.3 & 3.9 & 9.7 \\
    me/file\_rwpms\_wcnf\_L2\_V150\_C2500\_H150\_5 & 4.5 & 3.3 & 5.2 & 10.6 \\
    me/file\_rwpms\_wcnf\_L2\_V150\_C2500\_H150\_6 & 4.6 & 4.2 & 4.7 & 8.8 \\
    me/file\_rwpms\_wcnf\_L2\_V150\_C2500\_H150\_7 & 5.0 & 5.1 & 5.4 & 8.7 \\
    me/file\_rwpms\_wcnf\_L2\_V150\_C2500\_H150\_8 & 5.6 & 4.1 & 4.6 & 6.5 \\
    me/file\_rwpms\_wcnf\_L2\_V150\_C2500\_H150\_9 & 5.0 & 4.7 & 3.6 & 13.0 \\
    me/file\_rwpms\_wcnf\_L2\_V150\_C3000\_H150\_0 & 4.4 & 5.3 & 5.0 & 8.0 \\
    me/file\_rwpms\_wcnf\_L2\_V150\_C3000\_H150\_1 & 4.5 & 4.3 & 8.3 & 9.5 \\
    me/file\_rwpms\_wcnf\_L2\_V150\_C3000\_H150\_2 & 4.3 & 3.7 & 6.8 & 10.0 \\
    me/file\_rwpms\_wcnf\_L2\_V150\_C3000\_H150\_3 & 4.4 & 4.5 & 5.5 & 10.7 \\
    me/file\_rwpms\_wcnf\_L2\_V150\_C3000\_H150\_4 & 4.2 & 4.3 & 7.0 & 7.5 \\
    me/file\_rwpms\_wcnf\_L2\_V150\_C3000\_H150\_5 & 4.2 & 1.8 & 6.6 & 7.8 \\
    me/file\_rwpms\_wcnf\_L2\_V150\_C3000\_H150\_6 & 4.7 & 5.7 & 7.3 & 8.0 \\
    me/file\_rwpms\_wcnf\_L2\_V150\_C3000\_H150\_7 & 4.2 & 5.7 & 2.8 & 11.2 \\
    me/file\_rwpms\_wcnf\_L2\_V150\_C3000\_H150\_8 & 4.1 & 4.0 & 5.9 & 11.4 \\
    me/file\_rwpms\_wcnf\_L2\_V150\_C3000\_H150\_9 & 3.8 & 4.3 & 3.5 & 10.9 \\
    me/file\_rwpms\_wcnf\_L2\_V150\_C3500\_H150\_0 & 3.8 & 4.6 & 5.7 & 13.4 \\
    me/file\_rwpms\_wcnf\_L2\_V150\_C3500\_H150\_1 & 3.7 & 6.3 & 5.6 & 9.5 \\
    me/file\_rwpms\_wcnf\_L2\_V150\_C3500\_H150\_2 & 3.6 & 4.1 & 4.7 & 11.3 \\
    me/file\_rwpms\_wcnf\_L2\_V150\_C3500\_H150\_3 & 3.3 & 6.6 & 4.3 & 11.2 \\
    me/file\_rwpms\_wcnf\_L2\_V150\_C3500\_H150\_4 & 3.7 & 6.5 & 4.1 & 11.6 \\
    me/file\_rwpms\_wcnf\_L2\_V150\_C3500\_H150\_5 & 3.1 & 5.5 & 5.1 & 10.2 \\
    me/file\_rwpms\_wcnf\_L2\_V150\_C3500\_H150\_6 & 2.9 & 5.9 & 3.1 & 10.8 \\
    me/file\_rwpms\_wcnf\_L2\_V150\_C3500\_H150\_7 & 3.9 & 7.3 & 3.4 & 10.7 \\
    me/file\_rwpms\_wcnf\_L2\_V150\_C3500\_H150\_8 & 3.3 & 4.5 & 4.8 & 11.3 \\
    me/file\_rwpms\_wcnf\_L2\_V150\_C3500\_H150\_9 & 3.2 & 4.6 & 4.3 & 13.4 \\
    hi/file\_rwpms\_wcnf\_L2\_V150\_C4000\_H150\_0 & 4.7 & 3.7 & 4.8 & 13.6 \\
    hi/file\_rwpms\_wcnf\_L2\_V150\_C4000\_H150\_1 & 4.5 & 5.1 & 3.7 & 8.7 \\
    hi/file\_rwpms\_wcnf\_L2\_V150\_C4000\_H150\_2 & 4.9 & 4.7 & 4.2 & 9.5 \\
    hi/file\_rwpms\_wcnf\_L2\_V150\_C4000\_H150\_3 & 6.0 & 4.7 & 5.5 & 9.7 \\
    hi/file\_rwpms\_wcnf\_L2\_V150\_C4000\_H150\_4 & 4.5 & 5.7 & 4.7 & 13.1 \\
    hi/file\_rwpms\_wcnf\_L2\_V150\_C4000\_H150\_5 & 4.7 & 4.2 & 3.8 & 11.2 \\
    hi/file\_rwpms\_wcnf\_L2\_V150\_C4000\_H150\_6 & 4.6 & 2.7 & 4.4 & 14.0 \\
    hi/file\_rwpms\_wcnf\_L2\_V150\_C4000\_H150\_7 & 4.6 & 3.0 & 6.2 & 11.9 \\
    hi/file\_rwpms\_wcnf\_L2\_V150\_C4000\_H150\_8 & 4.3 & 3.7 & 5.3 & 13.2 \\
    hi/file\_rwpms\_wcnf\_L2\_V150\_C4000\_H150\_9 & 4.8 & 6.8 & 3.6 & 11.5 \\
    hi/file\_rwpms\_wcnf\_L2\_V150\_C4500\_H150\_0 & 3.9 & 4.6 & 5.4 & 8.7 \\
    hi/file\_rwpms\_wcnf\_L2\_V150\_C4500\_H150\_1 & 4.4 & 3.7 & 6.6 & 6.6 \\
    hi/file\_rwpms\_wcnf\_L2\_V150\_C4500\_H150\_2 & 3.9 & 5.4 & 5.2 & 10.1 \\
    hi/file\_rwpms\_wcnf\_L2\_V150\_C4500\_H150\_3 & 4.1 & 6.0 & 4.1 & 9.4 \\
    hi/file\_rwpms\_wcnf\_L2\_V150\_C4500\_H150\_4 & 4.5 & 5.5 & 4.2 & 11.1 \\
    hi/file\_rwpms\_wcnf\_L2\_V150\_C4500\_H150\_5 & 4.3 & 5.0 & 4.4 & 12.6 \\
    hi/file\_rwpms\_wcnf\_L2\_V150\_C4500\_H150\_6 & 4.3 & 4.4 & 5.4 & 10.6 \\
    hi/file\_rwpms\_wcnf\_L2\_V150\_C4500\_H150\_7 & 4.0 & 4.8 & 4.4 & 10.7 \\
    hi/file\_rwpms\_wcnf\_L2\_V150\_C4500\_H150\_8 & 4.9 & 8.5 & 5.1 & 10.5 \\
    hi/file\_rwpms\_wcnf\_L2\_V150\_C4500\_H150\_9 & 4.1 & 4.5 & 5.6 & 8.7 \\
    hi/file\_rwpms\_wcnf\_L2\_V150\_C5000\_H150\_0 & 4.9 & 3.2 & 5.9 & 11.9 \\
    hi/file\_rwpms\_wcnf\_L2\_V150\_C5000\_H150\_1 & 4.9 & 4.8 & 4.8 & 12.0 \\
    hi/file\_rwpms\_wcnf\_L2\_V150\_C5000\_H150\_2 & 5.2 & 5.0 & 4.8 & 12.1 \\
    hi/file\_rwpms\_wcnf\_L2\_V150\_C5000\_H150\_3 & 4.7 & 4.2 & 3.7 & 7.1 \\
    hi/file\_rwpms\_wcnf\_L2\_V150\_C5000\_H150\_4 & 4.9 & 4.6 & 5.0 & 8.5 \\
    hi/file\_rwpms\_wcnf\_L2\_V150\_C5000\_H150\_5 & 5.3 & 5.1 & 5.0 & 11.1 \\
    hi/file\_rwpms\_wcnf\_L2\_V150\_C5000\_H150\_6 & 5.0 & 5.4 & 4.0 & 11.1 \\
    hi/file\_rwpms\_wcnf\_L2\_V150\_C5000\_H150\_7 & 4.6 & 5.0 & 5.1 & 11.8 \\
    hi/file\_rwpms\_wcnf\_L2\_V150\_C5000\_H150\_8 & 4.4 & 5.4 & 4.3 & 8.9 \\
    hi/file\_rwpms\_wcnf\_L2\_V150\_C5000\_H150\_9 & 5.1 & 3.7 & 6.4 & 12.6 \\
        \hline
    \end{longtable}
\end{center}

\chapter{Score vs. Step Graph\label{highres_score_vs_step_graphs}}
\begin{figure}[!ht]
    \centering
    \includegraphics[height=6in]{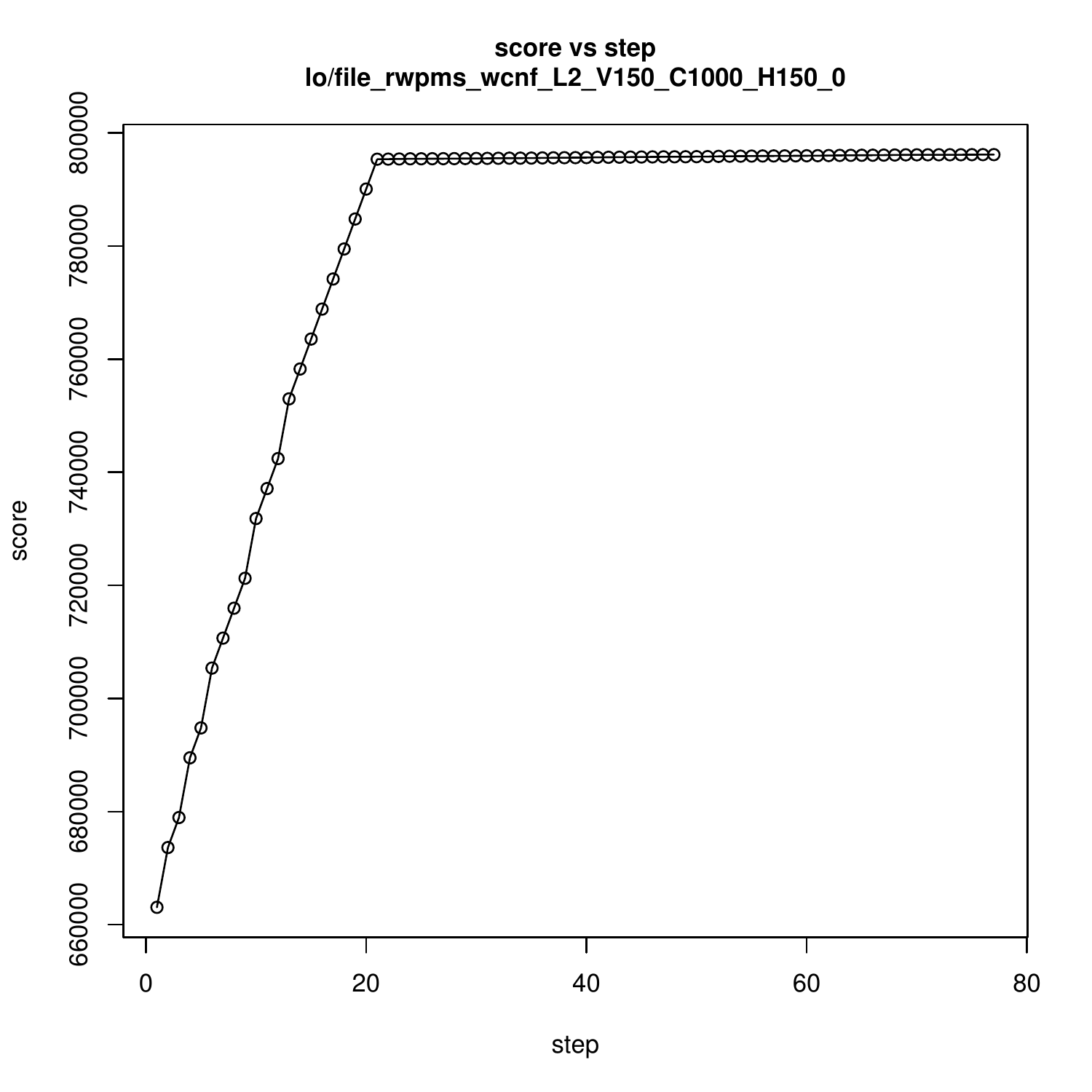}
    \label{fig_highres_search_map}
    \caption{High resolution view of MaxWalkSat's solution score as it searches the solution space of a problem (``lo/file\_rwpms\_wcnf\_L2\_V150\_C1000\_H150\_0'').}
\end{figure}

\end{singlespace}
\end{appendix}


\end{document}